\documentclass[11pt,leqno]{report}
\usepackage{amstex}
\usepackage{xspace}
\usepackage{avm}
\usepackage{examples}
\usepackage[dvips]{graphicx}
\usepackage{varioref}
\usepackage{makeidx}
\usepackage{epic}
\usepackage{eepic}
\usepackage{ecltree}

\newcommand{\feat}{\sc}                         
\newcommand{\fval}{\it}                         
\newcommand{\srt}{\it}                          
\newcommand{\pref}[1]{(\ref{#1})}               
\newcommand{\qit}[1]{\textsl{``#1''}}           
\newcommand{\sys}[1]{\texttt{#1}}               
\newcommand{\lexrule}[1]{\textsc{#1}}           
\newcommand{\principle}[1]{\textsc{#1}}         
\newcommand{\sql}[1]{{\small \texttt{#1}}}      
\newcommand{\proverb}[1]{\begin{flushright}``\textit{#1}''\end{flushright}}

\newcommand{\nlidb}{\textsc{Nlidb}\xspace}
\newcommand{\nlitdb}{\textsc{Nlitdb}\xspace}

\newcommand{\hpsg}{\textsc{Hpsg}\xspace}
\newcommand{\drt}{\textsc{Drt}\xspace}
\newcommand{\topl}{\textsc{Top}\xspace}
\newcommand{\tsql}{\textsc{Tsql2}\xspace}
\newcommand{\sqll}{\textsc{Sql}\xspace}
\newcommand{\sqlnt}{\textsc{Sql-92}\xspace}
\newcommand{\datalog}{\textsc{Datalog}\xspace}
\newcommand{\ale}{\textsc{Ale}\xspace}
\newcommand{\bcdm}{\textsc{Bcdm}\xspace}
\newcommand{\qed}{{\small Q.E.D.}\xspace}
\newcommand{\dbms}{\textsc{Dbms}\xspace}

\newcommand{\sul}{\textsc{Sul}\xspace}
\newcommand{\ptq}{\textsc{Ptq}\xspace}
\newcommand{\ils}{\textsc{Il}$_s$\xspace}
\newcommand{\hrdm}{\textsc{Hrdm}\xspace}

\newcommand{\supp}{\ensuremath{\surd}\xspace}        
\newcommand{\nosupp}{\ensuremath{\times}\xspace}     
\newcommand{\bad}{\sqz{*}}                           
\newcommand{\odd}{\sqz{?}}                           
\newcommand{\rej}{\sqz{$\times$}}                    

\newcommand{\subper}{\ensuremath{\sqsubseteq}}       
\newcommand{\propsubper}{\ensuremath{\sqsubset}}     
\newcommand{\union}{\cup}                            
\newcommand{\intersect}{\cap}                        
\newcommand{\defeq}{\overset{def}{=}}                
\newcommand{\denot}[2]{\|#2\|^{#1}}                  
\newcommand{\tup}[1]{\langle#1\rangle}               
\newcommand{\corn}[1]{\ulcorner#1\urcorner}          

\newcommand{\partop}{\ensuremath{\mathit{Part}}\xspace}
\newcommand{\pres}{\ensuremath{\mathit{Pres}}\xspace}
\newcommand{\past}{\ensuremath{\mathit{Past}}\xspace}
\newcommand{\culm}{\ensuremath{\mathit{Culm}}\xspace}
\newcommand{\perf}{\ensuremath{\mathit{Perf}}\xspace}
\newcommand{\progop}{\ensuremath{\mathit{Prog}}\xspace}
\newcommand{\at}{\ensuremath{\mathit{At}}\xspace}
\newcommand{\ntense}{\ensuremath{\mathit{Ntense}}\xspace}
\newcommand{\before}{\ensuremath{\mathit{Before}}\xspace}
\newcommand{\after}{\ensuremath{\mathit{After}}\xspace}
\newcommand{\lbegin}{\ensuremath{\mathit{Begin}}\xspace}
\newcommand{\lend}{\ensuremath{\mathit{End}}\xspace}
\newcommand{\fills}{\ensuremath{\mathit{Fills}}\xspace}
\newcommand{\for}{\ensuremath{\mathit{For}}\xspace}
\newcommand{\since}{\ensuremath{\mathbf{since}\xspace}}
\newcommand{\until}{\ensuremath{\mathbf{until}\xspace}}
\newcommand{\prevop}{\mathbf{\bullet}\xspace}
\newcommand{\nextop}{\mathbf{\circ}\xspace}
\newcommand{\tlpast}{\lozenge\xspace}

\newcommand{\pts}{\ensuremath{\mathit{PTS}}\xspace}
\newcommand{\periods}{\ensuremath{\mathit{PERIODS}}\xspace}
\newcommand{\telems}{\ensuremath{\mathit{TELEMS}}\xspace}
\newcommand{\instants}{\ensuremath{\mathit{INSTANTS}}\xspace}
\newcommand{\cons}{\ensuremath{\mathit{CONS}}\xspace}
\newcommand{\vars}{\ensuremath{\mathit{VARS}}\xspace}
\newcommand{\terms}{\ensuremath{\mathit{TERMS}}\xspace}
\newcommand{\pfuns}{\ensuremath{\mathit{PFUNS}}\xspace}
\newcommand{\parts}{\ensuremath{\mathit{PARTS}}\xspace}
\newcommand{\cparts}{\ensuremath{\mathit{CPARTS}}\xspace}
\newcommand{\gparts}{\ensuremath{\mathit{GPARTS}}\xspace}
\newcommand{\aforms}{\ensuremath{\mathit{AFORMS}}\xspace}
\newcommand{\forms}{\ensuremath{\mathit{FORMS}}\xspace}
\newcommand{\ynforms}{\ensuremath{\mathit{YNFORMS}}\xspace}
\newcommand{\whforms}{\ensuremath{\mathit{WHFORMS}}\xspace}
\newcommand{\objs}{\ensuremath{\mathit{OBJS}}\xspace}
\newcommand{\fcons}{\ensuremath{\mathit{f_{cons}}}\xspace}
\newcommand{\fpfuns}{\ensuremath{\mathit{f_{pfuns}}}\xspace}
\newcommand{\fculms}{\ensuremath{\mathit{f_{culms}}}\xspace}
\newcommand{\fgparts}{\ensuremath{\mathit{f_{gparts}}}\xspace}
\newcommand{\fcparts}{\ensuremath{\mathit{f_{cparts}}}\xspace}
\newcommand{\pow}{\ensuremath{\mathit{pow}}\xspace}
\newcommand{\mxlpers}{\ensuremath{\mathit{mxlpers}}\xspace}
\newcommand{\chrons}{\ensuremath{\mathit{CHRONS}}\xspace}
\newcommand{\hcons}{\ensuremath{\mathit{h_{cons}}}\xspace}
\newcommand{\hconsp}{\ensuremath{\mathit{h'_{cons}}}\xspace}
\newcommand{\hgparts}{\ensuremath{\mathit{h_{gparts}}}\xspace}
\newcommand{\hgpartsp}{\ensuremath{\mathit{h'_{gparts}}}\xspace}
\newcommand{\hcparts}{\ensuremath{\mathit{h_{cparts}}}\xspace}
\newcommand{\hcpartsp}{\ensuremath{\mathit{h'_{cparts}}}\xspace}
\newcommand{\hpfuns}{\ensuremath{\mathit{h_{pfuns}}}\xspace}
\newcommand{\hpfunsp}{\ensuremath{\mathit{h'_{pfuns}}}\xspace}
\newcommand{\hculms}{\ensuremath{\mathit{h_{culms}}}\xspace}
\newcommand{\hculmsp}{\ensuremath{\mathit{h'_{culms}}}\xspace}
\newcommand{\cvrel}{\ensuremath{\mathit{NVREL}_P}\xspace}
\newcommand{\vrel}{\ensuremath{\mathit{VREL}_P}\xspace}
\newcommand{\srel}{\ensuremath{\mathit{SREL}}\xspace}
\newcommand{\objsdb}{\ensuremath{\mathit{OBJS^{db}}}\xspace}
\newcommand{\fdi}{\ensuremath{f_D^{-1}}\xspace}
\newcommand{\fcn}{\ensuremath{\mathit{FCN}}\xspace}
\newcommand{\linit}{\ensuremath{\mathit{\lambda_{init}}}\xspace}
\newcommand{\vempty}{\ensuremath{v_\varepsilon}\xspace}

\newcommand{\dbtable}[5]{{\small\begin{tabular}[t]{#2}
      \hline                            
      \multicolumn{#1}{|l|}{#3} \\      
      \hline \hline                     
      #4 \\
      \hline
      #5 \\
      \hline 
   \end{tabular}}}

\newcommand{\dbtableb}[2]{{\small\begin{tabular}[t]{#1}
      \hline                            
      #2 \\
      \hline 
   \end{tabular}}}

\newcommand{\dbtablec}[3]{{\small\begin{tabular}[t]{#1}
      \hline                            
      #2 \\
      \hline
      #3 \\
      \hline 
   \end{tabular}}}

\newcommand{\adbtable}[5]        
{\vspace{-7mm}
 \begin{center}
 \dbtable{#1}{#2}{#3}{#4}{#5} 
 \end{center}}

\newcommand{\select}[1]                
{\sql{\hspace*{-2mm}\begin{tabular}[t]{l}
#1
\end{tabular}}}

\newtheorem{theorem}{Theorem}[chapter]
\newtheorem{lemma}{Lemma}[chapter]

\avmfont{\footnotesize\feat}
\avmvalfont{\footnotesize\fval}
\avmsortfont{\scriptsize\srt}

\newbox\avmboxa
\newbox\avmboxb
\newbox\avmboxc 

\makeatletter
\@twosidefalse
\columnwidth \textwidth 

\makeatother

\makeatletter
\long\def\@makefntext#1{\@setpar{\@@par\@tempdima \hsize 
  \advance\@tempdima-10pt\parshape \@ne 10pt \@tempdima}\par
  \parindent 1em\noindent \hbox to \z@{\hss$^{\@thefnmark}\ $}#1}
\makeatother

\newlength{\spacing}
\newcommand{\nspace}[1]{\setlength{\baselineskip}{#1\spacing}}

\newcommand{\doublespace}{\nspace{1.4}}
\newcommand{\singlespace}{\nspace{1.0}}

\newcommand{\ssmall}{\singlespace\small}
\newcommand{\dnorm}{\normalsize\doublespace}

\newenvironment{examps}[0]
   {\begin{examples*}
    \singlespace}
   {\end{examples*}}

\makeindex

\begin{document}
    \bibliographystyle{scribe}              
    \pagestyle{empty}

\begin{titlepage}
\begin{center}
\vspace*{1in}
{\LARGE
A Principled Framework for Constructing \\[4mm]
Natural Language Interfaces to Temporal Databases}\\
\vspace*{3cm}

{\large
{\bf Ioannis Androutsopoulos}

\vspace*{45mm}

\centerline{
\includegraphics[width=35mm]{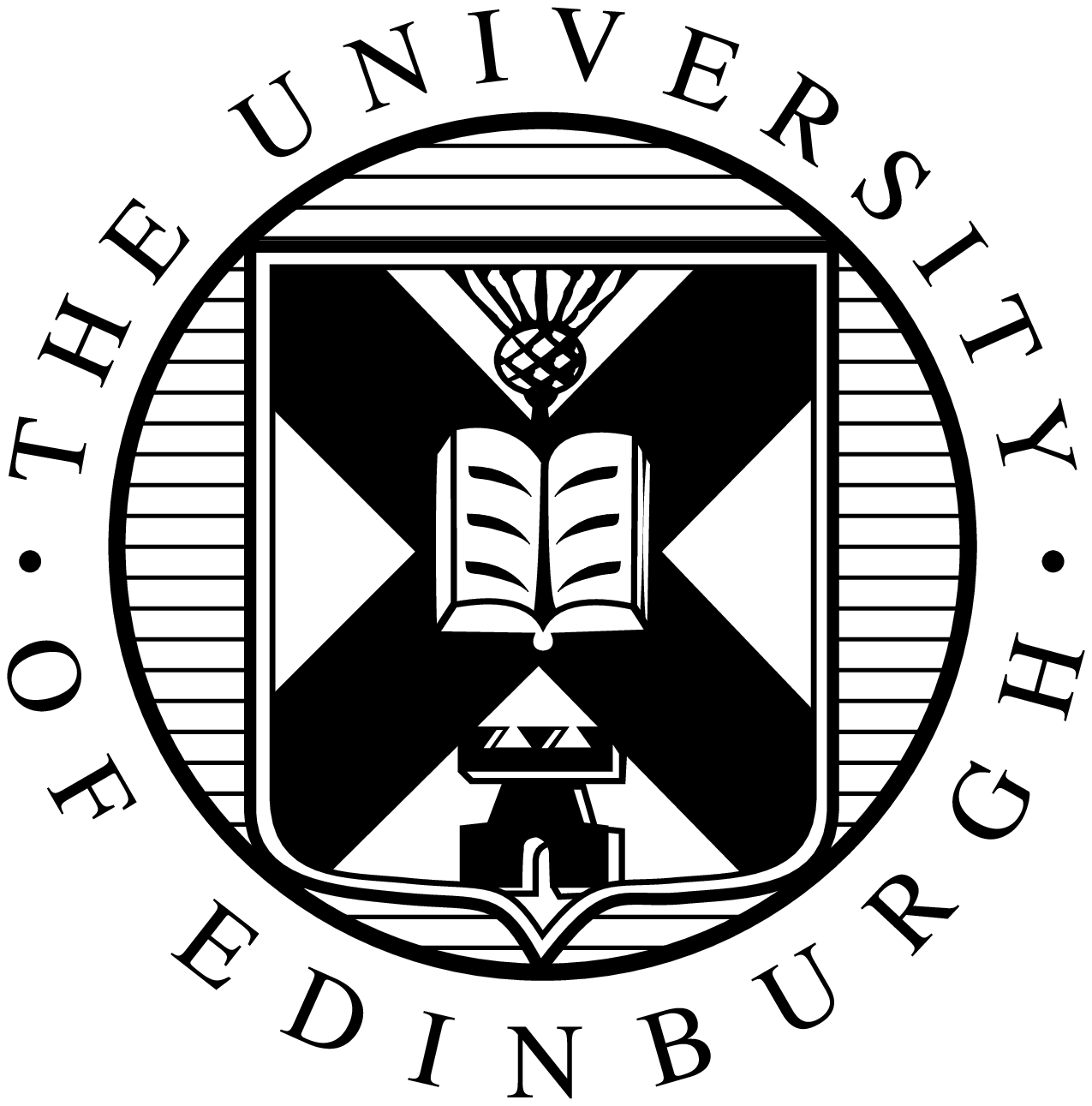}
}

\vspace*{45mm}

Ph.D. \\
University of Edinburgh \\
1996
}
\end{center}

\end{titlepage}

    \newpage

    \pagenumbering{roman}                   
    \pagestyle{plain}
    \setcounter{page}{2}
\vspace*{2cm}

\begin{center}
{\Large {\bf Abstract}}
\end{center}

Most existing natural language interfaces to databases (\nlidb{s})
were designed to be used with ``snapshot'' database systems, that
provide very limited facilities for manipulating time-dependent data.
Consequently, most \nlidb{s} also provide very limited support for the
notion of time. In particular, they were designed to answer questions
that refer mainly to the present, and do not support adequately the
mechanisms that natural language uses to express time. The database
community is becoming increasingly interested in \emph{temporal}
database systems. These are intended to store and manipulate in a
principled manner information not only about the present, but also
about the past and future. When interfacing to temporal databases, it
becomes crucial for \nlidb{s} to interpret correctly temporal
linguistic mechanisms (verb tenses, temporal adverbials, temporal
subordinate clauses, etc.)

I argue that previous approaches to natural language interfaces for
\emph{temporal} databases (\nlitdb{s}) are problematic, mainly because
they ignore important time-related linguistic phenomena, and/or they
assume idiosyncratic temporal database systems. This thesis develops a
principled framework for constructing English \nlitdb{s}, drawing on
research in tense and aspect theories, temporal logics, and temporal
databases. I first explore temporal linguistic phenomena that are
likely to appear in English questions to \nlitdb{s}. Drawing on
existing linguistic theories of time, I formulate an account for a
large number of these phenomena that is simple enough to be embodied
in practical \nlitdb{s}. Exploiting ideas from temporal logics, I then
define a temporal meaning representation language, \topl, and I show
how the \hpsg grammar theory can be modified to incorporate the tense
and aspect account of this thesis, and to map a wide range of English
questions involving time to appropriate \topl expressions. Finally, I
present and prove the correctness of a method to translate from \topl
to \tsql, \tsql being a temporal extension of the \sqlnt database
language. This way, I establish a sound route from English questions
involving time to a general-purpose temporal database language, that
can act as a principled framework for building \nlitdb{s}. To
demonstrate that this framework is workable, I employ it to develop a
prototype \nlitdb, implemented using the \ale grammar development
system and Prolog. The prototype \nlitdb can map temporal questions
from a non-trivial fragment of English to appropriate \tsql queries.


    \addcontentsline{toc}{chapter}{Abstract}
    \newpage

\vspace*{4cm}
\begin{center}
{\Large {\bf Acknowledgements}}
\end{center}

I would like to thank my supervisors, Dr.\ Graeme Ritchie and Dr.\
Peter Thanisch, for their guidance throughout my studies
in Edinburgh. The work that this thesis reports was supported by a
scholarship from the Greek State Scholarships Foundation, to which I
am grateful. 

The \ale grammar of the prototype \nlitdb (chapter
\ref{implementation}) is based on previous \ale encodings of \hpsg
fragments by Gerald Penn, Bob Carpenter, Suresh Manandhar, and Claire
Grover. I am indebted to all of them. I am also grateful to Chris
Brew, who provided additional \ale code for displaying
feature-structures, and to Jo Calder, for his help with \ale,
\textsc{Hpsg-Pl}, and \textsc{Pleuk}.

This thesis is dedicated to my parents, whose continuous support I
find invaluable.


    \addcontentsline{toc}{chapter}{Acknowledgements}
    \newpage

\vspace*{5cm}
\begin{center}
{\Large {\bf Declaration}}
\end{center}

I hereby declare that I composed this thesis entirely myself and that it
describes my own research.  

\vspace*{2in} 
\hspace*{3in}Ioannis Androutsopoulos \\    
\hspace*{3in}Edinburgh \\ 
\hspace*{3in}June 1996


    \addcontentsline{toc}{chapter}{Declaration}
    \newpage

    \tableofcontents                        
    \listoftables                           
    \addcontentsline{toc}{chapter}{List of Tables}
    \listoffigures                          
    \addcontentsline{toc}{chapter}{List of Figures}
    \newpage
    \addcontentsline{toc}{chapter}{Index of Symbols}
    \printindex
    \pagenumbering{arabic}                  
    \pagestyle{headings}
    \setcounter{page}{1}


    \doublespace                            

\chapter{Introduction} \label{intro_chapt}

\proverb{No time like the present.}


\section{Subject of this thesis} \label{thesis_subject}

Over the past thirty years, there have been significant advances in
the area of natural language interfaces to databases (\nlidb{s}).
\nlidb{s} allow users to access information stored in databases by
typing requests expressed in natural language (e.g.\ English). (The
reader is referred to \cite{Perrault}, \cite{Copestake}, and
\cite{Androutsopoulos1995} for surveys of \nlidb{s}.\footnote{The
  project described in this thesis began with an extensive survey of
  \nlidb{s}. The results of this survey were reported in
  \cite{Androutsopoulos1995}.}) Most of the existing \nlidb{s} were
designed to interface to ``snapshot'' database systems, that provide
very limited facilities for manipulating time-dependent data.
Consequently, most \nlidb{s} also provide very limited support for the
notion of time. In particular, they were designed to answer questions
that refer mainly to the present (e.g.\ \pref{intro:1} --
\pref{intro:3}), and do not support adequately the mechanisms that
natural language uses to express time. For example, very few (if any)
temporal adverbials (\qit{in 1991}, \qit{after 5:00pm}, etc.) and verb
forms (simple past, past continuous, past perfect, etc.) are typically
allowed, and their semantics are usually over-simplified or ignored.
\begin{examps}
\item What is the salary of each engineer? \label{intro:1}
\item Who is at site 4? \label{intro:2}
\item Which generators are in operation? \label{intro:3}
\end{examps}
The database community is becoming increasingly interested in
\emph{temporal} database systems. These are intended to store and
manipulate in a principled manner information not only about the
present, but also about the past and the future (see \cite{Tansel3} and 
\cite{tdbsglossary} for an introduction, and \cite{Bolour1983},
\cite{McKenzie2}, \cite{Stam}, \cite{Soo}, \cite{Kline1993},
and \cite{Tsotras1996} for bibliographies). When
interfacing to temporal databases, it becomes crucial for \nlidb{s} to
interpret correctly the temporal linguistic mechanisms (verb tenses,
temporal adverbials, temporal subordinate clauses, etc.) of questions
like \pref{intro:4} -- \pref{intro:6}.
\begin{examps}
\item What was the salary of each engineer while ScotCorp was
   building bridge 5? \label{intro:4} 
\item Did anybody leave site 4 before the chief engineer had inspected
   the control room? \label{intro:5}
\item Which systems did the chief engineer inspect on Monday after the
   auxiliary generator was in operation? \label{intro:6}
\end{examps}
In chapter \ref{comp_chapt}, I argue that previous approaches to
natural language interfaces for \emph{temporal} databases (\nlitdb{s})
are problematic, mainly because they ignore important time-related
linguistic phenomena, and/or they assume idiosyncratic temporal
database systems. This thesis develops a principled framework for
constructing English \nlitdb{s}, drawing on research in linguistic
theories of time, temporal logics, and temporal databases.


\section{Some background} 

This section introduces some ideas from \nlidb{s}, linguistic theories
of time, temporal logics, and temporal databases. Ideas from the four
areas will be discussed further in following chapters.


\subsection{Natural language interfaces to databases}
\label{domain_config}

\begin{figure}
\hrule
\medskip
\begin{center}
\includegraphics[scale=.7]{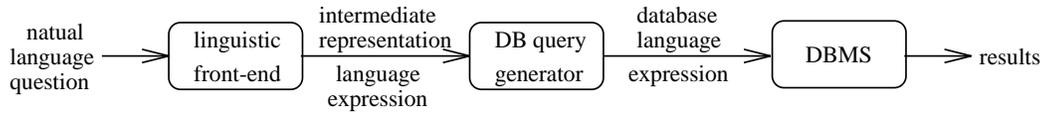}
\caption{Abstract architecture of many modern NLIDBs}
\label{pipeline_fig}
\end{center}
\hrule
\end{figure}

Past work on \nlidb{s} has shown the benefits of using the abstract
architecture of figure \ref{pipeline_fig}. The natural language
question is first parsed and analysed semantically by a linguistic
front-end, which translates the question into an intermediate meaning
representation language (typically, some form of logic). The generated
intermediate language expression captures formally what the system
understands to be the meaning of the natural language question,
without referring to particular database constructs. The intermediate
language expression is then translated into a database language
(usually \sqll \cite{Ullman} \cite{Melton1993}) that is supported by
the underlying \emph{database management system} (\dbms; this is the
part of the database system that manipulates the information in the
database). The resulting database language expression specifies what
information needs to be retrieved in terms of database constructs. The
\dbms retrieves this information by evaluating the database language
expression, and the obtained information is reported back to the user.

Most \nlidb{s} can only handle questions referring to a particular
knowledge-domain (e.g.\ questions about train departures, or about the
employees of a company), and need to be configured before they can be
used in a new domain. The configuration typically includes
``teaching'' the \nlidb words that can be used in the new domain, and
linking basic expressions of the formal intermediate language to
database constructs (see section 6 of \cite{Androutsopoulos1995}).

The architecture of figure \ref{pipeline_fig} has proven to have
several advantages (see sections 5.4 and 6 of
\cite{Androutsopoulos1995}), like modularity (e.g.\ the
linguistic-front end is shielded from database-level issues), and
\dbms portability (the same linguistic front-end can be used with
\dbms{s} that support different database languages). This thesis
examines how this architecture can be used to construct
\nlitdb{s}. 


\subsection{Tense and aspect theories} \label{tense_aspect_intro}

In English, temporal information can be conveyed by verb forms (simple
past, past continuous, present perfect, etc.), nouns (\qit{beginning},
\qit{predecessor}, \qit{day}), adjectives (\qit{earliest}, \qit{next},
\qit{annual}), adverbs (\qit{yesterday}, \qit{twice}), prepositional
phrases (\qit{at 5:00pm}, \qit{for two hours}), and subordinate
clauses (\qit{while tank 4 was empty}), to mention just some of the
available temporal mechanisms. A linguistic theory of time must
account for the ways in which these mechanisms are used (e.g.\ specify
what is the temporal content of each verb form, how temporal
adverbials or subordinate clauses affect the meaning of the overall
sentences, etc.). The term ``tense and aspect theories'' is often used
in the literature to refer to theories of this kind. (The precise
meanings of ``tense'' and ``aspect'' vary from one theory to the
other; see \cite{Comrie} and \cite{Comrie2} for some discussion.
Consult chapter 5 of \cite{Kamp1993} for an extensive introduction to
tense and aspect phenomena.)

It is common practice in tense and aspect theories to classify
natural language expressions or situations described by natural
language expressions into \emph{aspectual classes}. (The term
\qit{Aktionsarten} is often used to refer to these
classes.) Many aspectual classifications are similar to 
Vendler's taxonomy \cite{Vendler}, that distinguishes between
\emph{state} verbs, \emph{activity} verbs, \emph{accomplishment}
verbs, and \emph{achievement} verbs.\footnote{According to Mourelatos
\cite{Mourelatos1978}, a similar taxonomy was developed independently
in \cite{Kenny1963}, where Kenny notes that his classification is
similar to the distinction between \emph{kineseis} and \emph{energiai}
introduced by Aristotle in \textit{Metaphysics},
$\Theta$.1048\textit{b}, 18--36.} For example, \qit{to run} (as in
\qit{John ran.}) is said to be an activity verb, \qit{to know} (as in
\qit{John knows the answer.}) a state verb, \qit{to build} (as in
\qit{John built a house.}) an accomplishment verb, and \qit{to find}
(as in \qit{Mary found the treasure.}) an achievement verb.

Vendler's intuition seems to be that activity verbs describe actions
or changes in the world. For example, in \qit{John ran.} there is a
running action in the world. In contrast, state verbs do not refer to
any actions or changes. In \qit{John knows the answer.} there is no
change or action in the world. Accomplishment verbs are similar to
activity verbs, in that they denote changes or actions.  In the case
of accomplishment verbs, however, the action or change has an inherent
``climax'', a point that has to be reached for the action or change to
be considered complete.  In \qit{build a house} the climax is the
point where the whole of the house has been built. If the building
stops before the whole of the house has been built, the building
action is incomplete. In contrast, the action of the activity verb
\qit{to run} (with no object, as in \qit{John ran.}) does not seem to
have any climax. The runner can stop at any time without the running
being any more or less complete. If, however, \qit{to run} is used
with an object denoting a precise distance (e.g.\ \qit{to run a
  mile}), then the action \emph{does} have a climax: the point where
the runner completes the distance.  In this case, \qit{to run} is an
accomplishment verb. Finally, achievement verbs, like \qit{to find},
describe instantaneous events. In \qit{Mary found the treasure.} the
actual finding is instantaneous (according to Vendler, the time during
which Mary was searching for the treasure is not part of the actual
finding). In contrast, in \qit{John built a house.} (accomplishment
verb) the actual building action may have lasted many years.

Aspectual taxonomies are invoked to account for semantic differences
in similar sentences. The so-called ``imperfective paradox''
\cite{Dowty1977} \cite{Lascarides} is a well-known example (various
versions of the imperfective paradox have been proposed; see
\cite{Kent}). The paradox is that if the answer to a question like
\pref{taintro:1} is affirmative, then the answer to the
non-progressive \pref{taintro:2} must also be affirmative. In
contrast, an affirmative answer to \pref{taintro:3} does not
necessarily imply an affirmative answer to \pref{taintro:4} (John may
have abandoned the repair before completing it). The \nlitdb must
incorporate some account for this phenomenon. If the \nlitdb generates
an affirmative response to \pref{taintro:1}, there must be some
mechanism to guarantee that the \nlitdb's answer to \pref{taintro:2}
will also be affirmative. No such mechanism is needed in
\pref{taintro:3} and \pref{taintro:4}.
\begin{examps}
\item Was IBI ever advertising a new computer? \label{taintro:1}
\item Did IBI ever advertise a new computer? \label{taintro:2}
\item Was J.Adams ever repairing engine 2? \label{taintro:3}
\item Did J.Adams ever repair engine 2? \label{taintro:4}
\end{examps}
The difference between \pref{taintro:1} -- \pref{taintro:2} and
\pref{taintro:3} -- \pref{taintro:4} can be accounted for by
classifying \qit{to advertise} as an activity, \qit{to repair} as an
accomplishment, and by stipulating that: (i) the simple past of an
accomplishment requires the climax to have been reached; (ii) the past
continuous of an accomplishment or activity, and the simple past of an
activity impose no such requirement. Then, the fact that an
affirmative answer to \pref{taintro:3} does not necessarily imply an
affirmative answer to \pref{taintro:4} is accounted for by the fact
that \pref{taintro:4} requires the repair to have been completed,
while \pref{taintro:3} merely requires the repair to have been ongoing
at some past time. In contrast \pref{taintro:2} does not require any
climax to have been reached; like \pref{taintro:1}, it simply requires
the advertising to have been ongoing at some past time. Hence, an
affirmative answer to \pref{taintro:1} implies an affirmative answer
to \pref{taintro:2}. It will become clear in chapter
\ref{linguistic_data} that aspectual taxonomies pertain to the
semantics of almost all temporal linguistic mechanisms.


\subsection{Temporal logics} \label{temp_log_intro}

Time is an important research topic in logic, and many formal
languages have been proposed to express temporal information
\cite{VanBenthem} \cite{Gabbay1994b}. One of the simplest approaches
is to use the traditional first-order predicate logic, introducing
time as an extra argument of each predicate.  \pref{tlogi:1} would be
represented as \pref{tlogi:2}, where $t$ is a time-denoting variable,
$\prec$ stands for temporal precedence, $\sqsubseteq$ for temporal
inclusion, and $now$ is a special term denoting the present
moment. The answer to \pref{tlogi:1} would be affirmative iff
\pref{tlogi:2} evaluates to true, i.e.\ iff there is a time $t$, such
that $t$ precedes the present moment, $t$ falls within 1/10/95, and
tank 2 contained water at $t$. (Throughout this thesis, I use ``iff''
as a shorthand for ``if and only if''.)
\begin{examps}
\item Did tank 2 contain water (some time) on 1/10/95? \label{tlogi:1}
\item $\exists t \; contain(tank2, water, t) \land t \prec now \land
   t \sqsubseteq \mathit{1/10/95}$ \label{tlogi:2}
\end{examps}
An alternative approach is to employ \emph{temporal operators}, like
Prior's $P$ (past) and $F$ (future) \cite{Prior}. In that
approach, formulae are evaluated with respect to particular times. For
example, $contain(tank2, water)$ would be true at a time $t$ iff
tank 2 contained water at $t$. Assuming that $\phi$ is a formula,
$P\phi$ is true at a time $t$ iff there is a time $t'$,
such that $t'$ precedes $t$, and $\phi$ is true at $t'$. Similarly,
$F\phi$ is true at $t$ iff there is a $t'$, such that $t'$ follows
$t$, and $\phi$ is true at $t'$. \qit{Tank 2 contains water.}
can be expressed as $contain(tank2, water)$, \qit{Tank 2 contained
water.} as $P\, contain(tank2, water)$, \qit{Tank 2 will contain
water.} as $F \, contain(tank2, water)$, and \qit{Tank 2 will have
contained water.} as $F \, P \, contain(tank2, water)$. 
Additional operators can be introduced, to capture the semantics
of temporal adverbials, temporal subordinate clauses,
etc. For example, an $On$ operator could be introduced, with the
following semantics: if $\phi$ is a formula and $\kappa$ specifies a
day (e.g.\ the day 1/10/95), then $On[\kappa, \phi]$ is true at a
time $t$ iff $t$ falls within the day specified by $\kappa$, and
$\phi$ is true at $t$.  Then, \pref{tlogi:1} could be represented as
\pref{tlogi:4}.
\begin{examps}
\item $P \;\; On[\mathit{1/10/95}, contains(tank2, water)]$ \label{tlogi:4}
\end{examps}
The intermediate representation language of this thesis, called \topl,
adopts the operators approach (\topl stands for ``language with
Temporal OPerators''). Temporal operators have also been used in
\cite{Dowty1982}, \cite{Lascarides}, \cite{Richards}, \cite{Kent},
\cite{Crouch2}, \cite{Pratt1995}, and elsewhere.

Unlike logics designed to be used in systems that reason about what
changes or remains the same over time, what can or will happen, what
could or would have happened, or how newly arrived information fits
within already known facts or assumptions (e.g.\ the situation
calculus of \cite{McCarthy1969}, the event calculus of
\cite{Kowalski1986}, and the logics of \cite{Allen1983},
\cite{Allen1984}, and \cite{McDermott1982} -- see \cite{Vila1994} for a
survey), \topl is not intended to be used in reasoning. I provide no
inference rules for \topl, and this is why I avoid calling \topl a
logic. \topl is only a formal language, designed to facilitate the
systematic mapping of temporal English questions to formal expressions
(this mapping is not a primary consideration in the above mentioned
logics). The answers to the English questions are not generated by
carrying out reasoning in \topl, but by translating the \topl
expressions to database language expressions, which are then evaluated
by the underlying \dbms. The definition of \topl will be given in
chapter \ref{TOP_chapter}, where other ideas from temporal logics will
also be discussed.


\subsection{Temporal databases} \label{tdbs_general}

In the \emph{relational model} \cite{Codd1970}, currently the dominant
database model, information is stored in \emph{relations}.
Intuitively, relations can be thought of as tables, consisting of rows
(called \emph{tuples}) and columns (called \emph{attributes}). For
example, the $salaries$ relation below shows the present salaries of
the current employees of a company. In the case of $salaries$,
whenever the salary of an employee is changed, or whenever an employee
leaves the company, the corresponding tuple is modified or deleted.
Hence, the database ``forgets'' past facts, and does not contain
enough information to answer questions like \qit{What was the salary
  of T.Smith on 1/1/1992?}.
\adbtable{2}{|c|c|}{$salaries$}
{$employee$ & $salary$ }
{$J.Adams$ & $17000$ \\
 $T.Smith$ & $19000$ \\
 \ \dots   & \ \dots
}
It is certainly true that traditional database models and languages
\emph{can} and \emph{have} been used to store temporal information.
(This has led several researchers to question the need for special
temporal support in database systems; see \cite{Davies1995} for some
discussion.) For example, two extra attributes ($\mathit{from}$ and
$\mathit{to}$) could be added to $salaries$ (as in $salaries2$) to
\emph{time-stamp} its tuples, i.e.\ to show when each employee had the
corresponding salary. 
\adbtable{4}{|c|c|c|c|}{$salaries2$}
{$employee$ & $salary$ & $from$ & $to$ }
{$J.Adams$ & $17000$ & $1/1/88$  & $5/5/90$ \\
 $J.Adams$ & $18000$ & $6/5/90$  & $9/8/91$ \\
 $J.Adams$ & $21000$ & $10/8/91$ & $27/3/93$ \\
 \ \dots   & \ \dots & \ \dots & \ \dots \\
 $T.Smith$ & $17000$ & $1/1/89$  & $1/10/90$ \\
 $T.Smith$ & $21000$ & $2/10/90$ & $23/5/92$ \\
 \ \dots   & \ \dots & \ \dots & \ \dots 
}
The lack of special temporal support in traditional database models
and languages, however, complicates the task of expressing in database
language time-related data manipulations. We may want, for example,
to compute from $salaries2$ a new relation $same\_salaries$ that shows
the times when J.Adams and T.Smith had the same salary, along with
their common salary:
\adbtable{3}{|c|c|c|}{$same\_salaries$}
{$salary$ & $from$ & $to$ }
{$17000$ & $1/1/89$  & $5/5/90$ \\
 $21000$ & $10/8/91$ & $23/5/92$ \\
 \ \dots & \ \dots   & \ \dots
}
That is, for every tuple of J.Adams in $salaries2$, we need to check
if the period specified by the $\mathit{from}$ and $\mathit{to}$
values of that tuple overlaps the period specified by the
$\mathit{from}$ and $\mathit{to}$ values of a tuple for T.Smith which
has the same $salary$ value. If they overlap, we need to compute the
intersection of the two periods. This cannot be achieved easily in the
present version of \sqll (the dominant database language for
relational databases \cite{Ullman} \cite{Melton1993}), because \sqll
currently does not have any special commands to check if two periods
overlap, or to compute the intersection of two periods (in fact, it
does not even have a period datatype).

As a further example, the approach of adding a $\mathit{from}$ and a
$\mathit{to}$ attribute to every relation allows relations like $rel1$ and
$rel2$ to be formed. Although $rel1$ and $rel2$ contain
different tuples, they represent the same information.
\vspace{-9mm}
\begin{center}
\begin{tabular}{lr}
\dbtable{4}{|c|c|c|c|}{$rel1$}
{$employee$ & $salary$ & $from$ & $to$}
{$G.Foot$ & $17000$ & $1/1/88$  & $9/5/88$ \\
 $G.Foot$ & $17000$ & $10/5/88$ & $9/5/93$ \\
 $G.Foot$ & $18000$ & $10/5/93$ & $1/3/94$ \\
 $G.Foot$ & $18000$ & $2/3/94$  & $11/2/95$ \\
 $G.Foot$ & $17000$ & $12/2/95$ & \ $31/3/96$
}
&
\dbtable{4}{|c|c|c|c|}{$rel2$}
{$employee$ & $salary$ & $from$ & $to$}
{$G.Foot$ & $17000$ & $1/1/88$  & $31/5/89$ \\
 $G.Foot$ & $17000$ & $1/6/89$  & $10/8/92$ \\
 $G.Foot$ & $17000$ & $11/8/92$ & $9/5/93$ \\
 $G.Foot$ & $18000$ & $10/5/93$ & $11/2/95$ \\
 $G.Foot$ & $17000$ & $12/2/95$ & \ $31/3/96$
}
\end{tabular}
\end{center}
Checking if the two relations represent the same information is not
easy in the current \sqll version. This task would be greatly
simplified if \sqll provided some mechanism to ``normalise''
relations, by merging tuples that apart from their $from$ and $to$
values are identical (tuples of this kind are called
\emph{value-equivalent}). In our example, that mechanism would turn
both $rel1$ and $rel2$ into $rel3$. To check that $rel1$ and
$rel2$ contain the same information, one would check that the
normalised forms of the two relations are the same.
\adbtable{4}{|c|c|c|c|}{$rel3$}
{$employee$ & $salary$ & $from$ & $to$}
{$G.Foot$ & $17000$ & $1/1/88$  & $9/5/93$ \\
 $G.Foot$ & $18000$ & $10/5/93$ & $11/2/96$ \\
 $G.Foot$ & $17000$ & $12/2/95$ & \ $31/3/96$
}
Numerous temporal versions of \sqll and the relational model have been
proposed (e.g.\ \cite{Clifford2}, \cite{Ariav1986},
\cite{Tansel},\cite{Snodgrass}, \cite{Navathe1988}, \cite{Gadia1988},
\cite{Lorentzos1988}; see \cite{McKenzie} for a summary of some of the
proposals). These add special temporal facilities to \sqll (e.g.\
predicates to check if two periods overlap, functions to compute
intersections of periods, etc.), and often special types of relations
to store time-varying information (e.g.\ relations that force
value-equivalent tuples to be merged automatically). Until
recently there was little consensus on how temporal support should be
added to \sqll and the relational model (or other database languages
and models), with every researcher in the field adopting
his/her own temporal database language and model. Perhaps as a result
of this, very few temporal \dbms{s} have been implemented (these are
mostly early prototypes; see \cite{Boehlen1995c}).

This thesis adopts \tsql, a recently proposed temporal extension of
\sqlnt that was designed by a committee comprising most leading
temporal database researchers. (\sqlnt is the latest \sqll standard
\cite{Melton1993}. \tsql is defined in \cite{TSQL2book}. An earlier
definition of \tsql can be found in \cite{Tsql2Sigmod}.) \tsql and the
version of the relational model on which \tsql is based will be
presented in chapter \ref{tdb_chapter}, along with some modifications
that were introduced to them for the purposes of this thesis. Until
recently, there was no implemented \dbms supporting \tsql. A prototype
system, however, which is capable of evaluating \tsql queries now
exists. (This system is called \textsc{TimeDB}. See
\cite{Boehlen1995c} for a brief technical description of
\textsc{TimeDB}. \textsc{TimeDB} actually supports \textsc{Atsql2}, a
variant of \tsql. See \cite{Boehlen1996} for some information on
\textsc{Atsql2}.)

Researchers in temporal databases distinguish between \emph{valid
  time} and \emph{transaction time}.\footnote{I adopt the consensus
  terminology of \cite{tdbsglossary}. A third term, \emph{user-defined
    time}, is also employed in the literature to refer to temporal
  information that is stored in the database without the \dbms
  treating it in any special way.} The valid time of some information
is the time when that information was true in the \emph{world}. The
transaction time of some information is the time when the
\emph{database} ``believed'' some piece of information. In this
thesis, I ignore the transaction-time dimension. I assume that the
natural language questions will always refer to the information that
the database currently believes to be true. Questions like
\pref{dbi:5}, where \qit{on 2/1/95} specifies a transaction time other
than the present, will not be considered.
\begin{examps}
\item According to what the database believed on 2/1/95, what was the
   salary of J.Adams on 1/1/89? \label{dbi:5}
\end{examps}


\section{Contribution of this thesis} \label{contribution}

As mentioned in section \ref{thesis_subject}, most existing \nlidb{s}
were designed to interface to snapshot database systems. Although
there have been some proposals on how to build \nlidb{s} for temporal
databases, in chapter \ref{comp_chapt} I argue that these proposals
suffer from one or more of the following: (i) they ignore
important English temporal mechanisms, or assign to them
over-simplified semantics, (ii) they lack clearly defined meaning
representation languages, (iii) they do not provide complete
descriptions of the mappings from natural language to meaning
representation language, or (iv) from meaning representation language
to database language, (v) they adopt idiosyncratic and often not
well-defined temporal database models or languages, (vi) they do not
demonstrate that their ideas are implementable. In this thesis, I
develop a principled framework for constructing English \nlitdb{s},
attempting to avoid pitfalls (i) -- (vi). Building on the architecture
of figure \ref{pipeline_fig}:
\begin{itemize}
\item I explore temporal linguistic phenomena that are likely to
  appear in English questions to \nlitdb{s}. Drawing on existing
  linguistic theories of time, I formulate an account for many of
  these phenomena that is simple enough to be embodied in practical
  \nlitdb{s}.

\item Exploiting ideas from temporal logics, I define a temporal
  meaning representation language (\topl), which I use to represent the
  semantics of English questions.

\item I show how \hpsg \cite{Pollard1} \cite{Pollard2}, currently a
  highly regarded linguistic theory, can be modified to incorporate
  the tense and aspect account of this thesis, and to map a wide range
  of English questions involving time to appropriate \topl
  expressions.

\item I present and prove the correctness of a mapping that translates
  \topl expressions to \tsql queries.
\end{itemize}
This way, I establish a sound route from English questions involving time
to a general-purpose temporal database language, that can act as a principled
framework for constructing \nlitdb{s}. To ensure that this framework is 
workable:
\begin{itemize}

\item I demonstrate how it can be employed to implement a prototype
\nlitdb, using the \ale grammar development system
\cite{Carpenter1992} \cite{Carpenter1994} and Prolog
\cite{Clocksin1994} \cite{Sterling1994}. I configure the prototype \nlitdb
for a hypothetical air traffic control domain, similar to that of
\cite{Sripada1994}.

\end{itemize}

Unfortunately, during most of the work of this thesis no \dbms
supported \tsql. As mentioned in section \ref{tdbs_general}, a
prototype \dbms (\textsc{TimeDB}) that supports a version of \tsql
(\textsc{Atsql2}) was announced recently. Although it would be
obviously very interesting to link the \nlitdb of this thesis to
\textsc{TimeDB}, there is currently very little documentation on
\textsc{TimeDB}. The task of linking the two systems is further
complicated by the fact that both adopt their own versions of \tsql
(\textsc{TimeDB} supports \textsc{Atsql2}, and the \nlitdb of this
thesis adopts a slightly modified version of \tsql, to be discussed in
chapter \ref{tdb_chapter}). One would have to bridge the differences
between the two \tsql versions. Due to shortage of time, I made no
attempt to link the \nlitdb of this thesis to \textsc{TimeDB}. The
\tsql queries generated by the \nlitdb are currently not executed, and
hence no answers are produced.

Although several issues (summarised in section \ref{to_do}) remain to
be addressed, I am confident that this thesis will prove valuable to
both those wishing to implement \nlitdb{s} for practical applications,
and those wishing to carry out further research on \nlitdb{s},
because: (a) it is essentially the first in-depth exploration of time-related
problems the \nlitdb designer has to face, from the linguistic
level down to the database level, (b) it proposes a clearly defined
framework for building \nlitdb{s} that addresses a great number of
these problems, and (c) it shows how this framework was used to
implement a prototype \nlitdb on which more elaborate \nlitdb{s} can
be based.

Finally, I note that: (i) the work of this thesis is one of the first
to use \tsql, and one of the first to generate feedback to the
\tsql designers (a number of obscure points and possible improvements in
the definition of \tsql were revealed during this project; these were
reported in \cite{Androutsopoulos1995b}); (ii) the prototype \nlitdb
of this thesis is currently one of the very few \nlidb{s} (at least
among \nlidb{s} whose grammar is publicly documented) that adopt
\hpsg.\footnote{See also \cite{Cercone1993}. A version of
the \hpsg grammar of this thesis, stripped of its temporal mechanisms,
was used in \cite{Seldrup1995} to construct a \nlidb for snapshot
databases.}


\section{Issues that will not be addressed} \label{no_issues}

To allow the work of this thesis to be completed within the available
time, the following issues were not considered. 

\paragraph{Updates:} This thesis focuses on
\emph{questions}. Natural language requests to \emph{update} the
database (e.g.\ \pref{noiss:1}) are not considered
(see \cite{Davidson1983} for work on natural language updates.)
\begin{examps}
\item Replace the salary of T.Smith for the period 1/1/88 to 5/5/90
by 17000. \label{noiss:1}
\end{examps}
Assertions like \pref{noiss:2} will be treated as yes/no
questions, i.e.\ \pref{noiss:2} will be treated in the same way as
\pref{noiss:3}.
\begin{examps}
\item On 1/1/89 the salary of T.Smith was 17000. \label{noiss:2}
\item Was the salary of T.Smith 17000 on 1/1/89? \label{noiss:3}
\end{examps}

\paragraph{Schema evolution:} This term refers to cases where the
\emph{structure}, not only the \emph{contents}, of the database change
over time (new relations are created, old deleted, attributes are added or
removed from relations, etc.; see \cite{McKenzie1990}). Schema evolution
is not considered in this thesis. The structure of the
database is assumed to be static, although the information in the
database may change over time.

\paragraph{Modal questions:} Modal questions ask if 
something could have happened, or could never have happened, or will
necessarily happen, or can possibly happen. For example,
\qit{Could T.Smith have been an employee of IBI in 1985?} does not ask
if T.Smith was an IBI employee in 1985, but if it would have
been possible for T.Smith to be an IBI employee at that time. Modal
questions are not examined in this thesis (see \cite{Mays1986} and
\cite{Lowden1991} for related work). 

\paragraph{Future questions:} A temporal database may contain
predictions about the future. At some company, for example, it may
have been decided that T.Smith will retire two years from the present,
and that J.Adams will replace him. These decisions may have been
recorded in the company's database. In that context, one may want to
submit questions referring to the future, like \qit{When will T.Smith
retire?} or \qit{Who will replace T.Smith?}. To simplify the
linguistic data that the work of this thesis had to address, future
questions were not considered. The database may contain information
about the future, but the framework of this thesis does not currently
allow this information to be accessed through natural
language. Further work could extend the framework of this thesis to
handle future questions as well (see section \ref{to_do}).

\paragraph{Cooperative responses:} In many cases, it is helpful
for the user if the \nlidb reports more information than what the
question literally asks for. In the dialogue below (from
\cite{Johnson1985}), for example, the system has reasoned that the
user would be interested to know about the United flight, and has
included information about that flight in its answer although this
was not requested. 
\begin{examps}
\item Do American Airlines have a night flight to Dallas? \label{noiss:4}
\item \sys{No, but United have flight 655.} \label{noiss:5}
\end{examps}
In other cases, the user's requests may be based on false
presumptions. \pref{noiss:4a}, for example, presumes that there is a
flight called BA737. If this is not true, it would be useful if the
\nlidb could generate a response like \pref{noiss:4b}.
\begin{examps}
\item Does flight BA737 depart at 5:00pm? \label{noiss:4a}
\item \sys{Flight BA737 does not exist.} \label{noiss:4b}
\end{examps}
The term \emph{cooperative responses} \cite{Kaplan1982} is used to
refer to responses like \pref{noiss:5} and \pref{noiss:4b}. The
framework of this thesis includes no mechanism to generate cooperative
responses. During the work of this thesis, however, it became clear
that such a mechanism is particularly important in questions to
\nlitdb{s}, and hence a mechanism of this kind should be added (this
will be discussed further in section \ref{to_do}).

\paragraph{Anaphora:}
Pronouns (\qit{she}, \qit{they}, etc.), possessive determiners
(\qit{his}, \qit{their}), and some noun phrases (\qit{the project},
\qit{these people}) are used anaphorically, to refer to contextually
salient entities. The term \emph{nominal anaphora} is frequently used
to refer to this phenomenon (see \cite{Hirst1981} for an overview of
nominal anaphora, and \cite{Hobbs1986} for methods that can be used to
resolve pronoun anaphora). Verb tenses and other temporal expressions
(e.g.\ \qit{on Monday}) are often used in a similar anaphoric manner
to refer to contextually salient times (this will be discussed in
section \ref{temporal_anaphora}). The term \emph{temporal anaphora}
\cite{Partee1984} is used in that case. Apart from a temporal
anaphoric phenomenon related to noun phrases like \qit{the sales
  manager} (to be discussed in section \ref{noun_anaphora}), for which
support is provided, the framework of this thesis currently provides
no mechanism to resolve anaphoric expressions (i.e.\ to determine the
entities or times these expressions refer to). Words introducing
nominal anaphora (e.g.\ pronouns) are not allowed, and (excluding the
phenomenon of section \ref{noun_anaphora}) temporal anaphoric
expressions are treated as denoting any possible referent (e.g.\ 
\qit{on Monday} is taken to refer to any Monday).

\paragraph{Elliptical sentences:} Some \nlidb{s} allow
elliptical questions to be submitted as follow-ups to previous
questions (e.g.\ \qit{What is the salary of J.Adams?}, followed
by \qit{His address?}; see section 4.6 of
\cite{Androutsopoulos1995} for more examples). Elliptical questions
are not considered in this thesis. 


\section{Outline of the remainder of this thesis}

The remainder of this thesis is organised as follows:

Chapter 2 explores English temporal mechanisms, delineating
the set of linguistic phenomena that this thesis attempts to
support. Drawing on existing ideas from tense and aspect theories, an
account for these phenomena is formulated that is suitable to the
purposes of this thesis.

Chapter 3 defines formally \topl, discussing how it can be used to
represent the semantics of temporal English expressions, and how
it relates to other existing temporal representation
languages.

Chapter 4 provides a brief introduction to \hpsg, and discusses how
\hpsg can be modified to incorporate the tense and aspect account of
this thesis, and to map English questions involving time to appropriate
\topl expressions. 

Chapter 5 defines the mapping from \topl to \tsql, and proves its
correctness (parts of this proof are given in appendix
\ref{trans_proofs}). It also discusses the modifications to \tsql that
are adopted in this thesis.

Chapter 6 describes the architecture of the prototype \nlitdb,
provides information about its implementation, and explains which
additional modules would have to be added if the system were to be
used in real-life applications. Several sample English questions
directed to a hypothetical temporal database of an airport are shown,
discussing the corresponding output of the prototype \nlitdb. 

Chapter 7 discusses previous proposals in the area of \nlitdb{s},
comparing them to the framework of this thesis. 

Chapter 8 summarises and proposes directions for further research.



\chapter{The Linguistic Data and an Informal Account} \label{linguistic_data}

\proverb{There is a time for everything.}


\section{Introduction}

This chapter explores how temporal information is conveyed in English,
focusing on phenomena that are relevant to \nlitdb{s}.
There is a wealth of temporal English mechanisms (e.g.\ verb tenses,
temporal adverbials, temporal adjectives, etc.), and it would be
impossible to consider all of those in this thesis. Hence,
several English temporal mechanisms will be ignored, and 
simplifying assumptions will be introduced in some of the mechanisms
that will be considered. One of the goals of this chapter is to specify
exactly which linguistic phenomena this thesis attempts to
support. For the phenomena that will be supported, a further goal is 
to provide an informal account of how they will be treated.

Although this chapter draws on existing tense and aspect theories, I
stress that it is in no way an attempt to formulate an improved tense
and aspect theory. The aim is more practical: to explore how
ideas from existing tense and aspect theories can be integrated into
\nlitdb{s}, in a way that leads to provably implementable systems.


\section{Aspectual taxonomies} \label{asp_taxes}

As mentioned in section \ref{tense_aspect_intro}, many tense and
aspect theories employ aspectual classifications, which are often
similar to Vendler's distinction between states (e.g.\ \qit{to know},
as in \qit{John knows the answer.}), activities (e.g.\ \qit{to run},
as in \qit{John ran.}), accomplishments (e.g.\ \qit{to build}, as in
\qit{John built a house.}), and achievements (e.g.\ \qit{to find}, as
in \qit{Mary found the treasure.}). 

Vendler proposes a number of linguistic tests to determine the
aspectual classes of verbs. For example, according to Vendler,
activity and accomplishment verbs can appear in the progressive (e.g.\
\qit{John is running}, \qit{John is building a house}), while state
and achievement verbs cannot (*\qit{John is knowing the answer.},
*\qit{Mary is finding the treasure}). Activity verbs are said to
combine felicitously with \qit{for~\dots} adverbials specifying
duration (\qit{John ran for two minutes.}), but sound odd with
\qit{in~\dots} duration adverbials (?\qit{John ran in two minutes.}).
Accomplishment verbs, in contrast, combine felicitously with
\qit{in~\dots} adverbials (\qit{John built a house in two weeks.}),
but sound odd with \qit{for~\dots} adverbials (?\qit{John built a
house for two weeks.}). Finally, according to Vendler state verbs
combine felicitously with \qit{for~\dots} adverbials (e.g.\ \qit{John
knew the answer for ten minutes (but then forgot it).}), while
achievement verbs sound odd with \qit{for~\dots} adverbials
(?\qit{Mary found the treasure for two hours.}). 

The exact nature of the objects classified by Vendler is unclear. In
most cases, Vendler's wording suggests that his taxonomy classifies
verbs. However, some of his examples (e.g.\ the fact that \qit{to run}
with no object is said to be an activity, while \qit{to run a mile} is
said to be an accomplishment) suggest that the natural language
expressions being classified are not always verbs, but sometimes
larger syntactic constituents (perhaps verb phrases). In other cases,
Vendler's arguments suggest that the objects being classified are not
natural language expressions (e.g.\ verbs, verb phrases), but world
situations denoted by natural language expressions. According to
Vendler, \qit{Are you smoking?}  ``asks about an activity'', while
\qit{Do you smoke?} ``asks about a state''. In this case, the terms
``activity'' and ``state'' seem to refer to types of situations in the
world, rather than types of natural language expressions. (The first
question probably asks if somebody is actually smoking at the present
moment. The second one has a \emph{habitual} meaning: it asks if
somebody has the habit of smoking.  Vendler concludes that habits
``are also states in our sense''.)

Numerous variants of Vendler's taxonomy have been proposed. These
differ in the number of aspectual classes they assume, the names of
the classes, the nature of the objects being classified, and the
properties assigned to each class. Vlach \cite{Vlach1993}
distinguishes four aspectual classes of sentences, and assumes that
there is a parallel fourfold taxonomy of world situations. Moens
\cite{Moens} distinguishes between ``states'', ``processes'',
``culminated processes'', ``culminations'', and ``points'', commenting
that his taxonomy does not classify real world situations, but ways
people use to describe world situations.  Parsons \cite{Parsons1989}
distinguishes three kinds of ``eventualities'' (``states'',
``activities'', and ``events''), treating eventualities as entities in
the world.  Lascarides \cite{Lascarides} classifies propositions
(functions from time-periods to truth values), distinguishing between
``state'', ``process'', and ``event'' propositions.


\section{The aspectual taxonomy of this thesis} \label{aspectual_classes}

Four aspectual classes are employed in this thesis: \emph{states},
\emph{activities}, \emph{culminating activities}, and \emph{points}.
(Culminating activities and points correspond to Vendler's
``accomplishments'' and ``achievements'' respectively. Similar terms
are used in \cite{Moens} and \cite{Blackburn1994}.) These aspectual
classes correspond to ways of \emph{viewing world situations} that
people seem to use: a situation can be viewed as involving no change
or action (state view), as an instantaneous change or action (point
view), as a change or action with no climax (activity view), or as a
change or action with a climax (culminating activity view).
(Throughout this thesis, I use ``situation'' to refer collectively to
elements of the world that other authors call ``events'',
``processes'', ``states'', etc.)  Determining which view the speaker
has in mind is important to understand what the speaker means. For
example, \qit{Which tanks contained oil?} is typically uttered with a
state view. When an \qit{at \dots} temporal adverbial (e.g.\ \qit{at
  5:00pm}) is attached to a clause uttered with a state view, the
speaker typically means that the situation of the clause simply holds
at the time of the adverbial.  There is normally no implication that
the situation starts or stops holding at the time of the adverbial.
For example, in \qit{Which tanks contained oil at 5:00pm?} there is
normally no implication that the tanks must have started or stopped
containing oil at 5:00pm.  In contrast, \qit{Who ran to the station?}
is typically uttered with a culminating activity view. In this case,
an \qit{at \dots} adverbial usually specifies the time when the
situation starts or is completed. \qit{Who ran to the station at
  5:00pm?}, for example, probably asks for somebody who started
running to the station or reached it at 5:00pm.

Some linguistic markers seem to signal which view the speaker has in
mind. For example, the progressive usually signals a state view (e.g.\
unlike \qit{Who ran to the station at 5:00pm?}, \qit{Who was running
to the station at 5:00pm?} is typically uttered with a state view; in
this case, the running is simply ongoing at 5:00pm, it does not start
or finish at 5:00pm). Often, however, there are no such explicit
markers. The processes employed in those cases by hearers to determine
the speaker's view are not yet fully understood. In an \nlitdb,
however, where questions refer to a restricted domain,
reasonable guesses can be made by observing that in each domain, each
verb tends to be associated mainly with one particular view. Certain
agreements about how situations are to be viewed (e.g.\ that some
situations are to be treated as instantaneous -- point view) will also
have been made during the design of the database. These agreements
provide additional information about how the situations of the various
verbs are viewed in each domain.

More precisely, the following approach is adopted in this thesis. 
Whenever the \nlitdb is configured for a new application domain, the
base form of each verb is assigned to one of the four aspectual
classes, using criteria to be discussed in section
\ref{aspect_criteria}. These criteria are intended to detect the view
that is mainly associated with each verb in the particular domain
that is being examined. Following \cite{Dowty1986}, \cite{Moens},
\cite{Vlach1993}, and others, aspectual class is treated as a property
of not only verbs, but also verb phrases, clauses, and
sentences. Normally, all verb forms will inherit the aspectual classes
of the corresponding base forms. Verb phrases, clauses, or sentences
will normally inherit the aspectual classes of their main verb
forms. Some linguistic mechanisms (e.g.\ the progressive or some
temporal adverbials), however, may cause the aspectual class of a verb
form to differ from that of the base form, or the aspectual class of a
verb phrase, clause, or sentence to differ from that of its main verb
form. The aspectual class of each verb phrase, clause, or sentence is
intended to reflect the view that users typically have in mind when
using that expression in the particular domain.

In the case of a verb like \qit{to run}, that typically involves a
culminating activity view when used with an expression that specifies
a destination or specific distance (e.g.\ \qit{to run to the
station/five miles}), but an activity view when used on its own, it
will be assumed that there are two different homonymous verbs \qit{to
run}. One has a culminating activity base form, and requires a
complement that specifies a destination or specific distance. The
other has an activity base form, and requires no such complement. A
similar distinction would be introduced in the case of verbs whose
aspectual class depends on whether or not the verb's object denotes a
countable or mass entity (e.g.\ \qit{to drink a bottle of wine} vs.\
\qit{to drink wine}; see \cite{Mourelatos1978}). 

Similarly, when a verb can be used in a domain with both habitual and
non-habitual meanings (e.g. \qit{BA737 (habitually) departs from
  Gatwick.} vs.\ \qit{BA737 (actually) departed from Gatwick five
  minutes ago.}), a distinction will be made between a homonym with a
habitual meaning, and a homonym with a non-habitual
meaning.\footnote{When discussing sentences with multiple readings, I often use
  parenthesised words (e.g.\ \qit{(habitually)}) to indicate which
  reading is being considered.} The base forms of habitual homonyms
are classified as states. (This agrees with Vendler, Vlach
\cite{Vlach1993}, and Moens and Steedman \cite{Moens2}, who all
classify habituals as states.) The aspectual classes of non-habitual
homonyms depend on the verb and the application domain.  Approaches
that do not postulate homonyms are also possible (e.g.\ claiming that
\qit{to run} is an activity which is transformed into a culminating
activity by \qit{the station}). The homonyms method, however, leads to
a more straight forward treatment in the \hpsg grammar of chapter
\ref{English_to_TOP} (where the base form of each homonym is mapped to
a different sign).

In the rest of this thesis, I refer to verbs whose base forms are
classified as states, activities, culminating activities, or points as
\emph{state verbs}, \emph{activity verbs}, \emph{culminating activity
verbs}, and \emph{point verbs}.


\section{Criteria for classifying base verb forms} \label{aspect_criteria}

\begin{figure}[tb]
  \hrule
  \medskip
  \begin{center}
    \includegraphics[scale=.5]{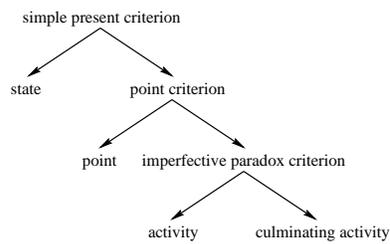}
    \caption{Determining the aspectual class of a verb's base form}
    \label{decision_tree}
  \end{center}
  \hrule
\end{figure}

This section discusses the criteria that determine the
aspectual class of a verb's base form in a particular \nlitdb
domain. Three criteria are employed, and they are
applied in the order of figure \ref{decision_tree}.

\subsection{The simple present criterion} \label{simple_present_criterion}

The first criterion distinguishes state verbs (verbs whose base forms
are states) from point, activity, and culminating activity verbs. If
the simple present of a verb can be used (in the particular domain) in
single-clause questions with non-futurate meanings, the verb is a
state one; otherwise it is a point, activity, or culminating activity
verb. For example, in domains where \pref{crit:1} and \pref{crit:2}
are possible, \qit{to contain} and \qit{to own} are state verbs.
\begin{examps}
\item Does any tank contain oil? \label{crit:1} 
\item Which employees own a car? \label{crit:2}
\end{examps}

Some clarifications are needed. First, the simple present sometimes
refers to something that is scheduled to happen. For example,
\pref{crit:2.7} could refer to a scheduled assembling (in that case,
\pref{crit:2.7} is very similar to \pref{crit:2.7.2}). I consider this
meaning of \pref{crit:2.7} futurate. Hence, this use of
\pref{crit:2.7} does not constitute evidence that \qit{to assemble}
is a state verb.
\begin{examps}
\item When does J.Adams assemble engine 5? \label{crit:2.7}
\item When will J.Adams assemble engine 5? \label{crit:2.7.2}
\end{examps}
In reporting contexts, the simple present of
verbs whose base forms I would not want to be classified as states can be used
with a non-futurate meaning. For example, in a context where the speaker
reports events as they happen, \pref{crit:2.8} is possible. (This use
of the simple present is unlikely in \nlitdb questions.)
\begin{examps}
\item J.Adams arrives. He moves the container. He fixes the engine.
 \label{crit:2.8}
\end{examps}
The simple present criterion examines questions directed to a \nlitdb,
not sentences from other contexts. Hence, \pref{crit:2.8} does not
constitute evidence that \qit{to arrive}, \qit{to move}, and \qit{to
fix} are state verbs.

The reader is reminded that when verbs have both habitual and
non-habitual meanings, I distinguish between habitual and non-habitual
homonyms (section \ref{aspectual_classes}). Ignoring
scheduled-to-happen meanings (that do not count for the simple present
criterion), \pref{crit:3} and \pref{crit:4} can only have
habitual meanings. 
\begin{examps}
\item Which flight lands on runway 2? \label{crit:3}
\item Does any doctor smoke? \label{crit:4}
\end{examps}
\pref{crit:3} asks for a flight that 
habitually lands on runway 2, and \pref{crit:4} for doctors that are
smokers. That is, \pref{crit:3} and \pref{crit:4} can only be
understood as involving the habitual homonyms of \qit{to land} and
\qit{to smoke}. (In contrast, \pref{crit:5} and \pref{crit:6} can be
understood with non-habitual meanings, i.e.\ as involving the
non-habitual homonyms.) 
\begin{examps}
\item Which flight is landing on runway 2? \label{crit:5}
\item Is any doctor smoking? \label{crit:6}
\end{examps}
Therefore, in domains where \pref{crit:3} and \pref{crit:4} are
possible, the habitual \qit{to land} and \qit{to smoke} are state
verbs. \pref{crit:3} and \pref{crit:4} do not constitute evidence that
the non-habitual \qit{to land} and \qit{to smoke} are state verbs.

\subsection{The point criterion} \label{point_criterion}

The second criterion, the \emph{point criterion}, distinguishes point
verbs from activity and culminating activity ones (state verbs will
have already been separated by the simple present criterion; see
figure \ref{decision_tree}).  The point criterion is based on the fact
that some verbs will be used to describe kinds of world situations
that are modelled in the database as being always instantaneous. If a
verb describes situations of this kind, its base form should be
classified as point; otherwise, it should be classified as activity or
culminating activity.

In section \ref{aspect_examples}, for example, I consider a
hypothetical airport database. That database does not distinguish
between the times at which a flight starts or stops entering an
airspace sector. Entering a sector is modelled as instantaneous.
Also, in the airport domain \qit{to enter} is only
used to refer to flights entering sectors. Consequently, in that 
domain \qit{to enter} is a point verb. If \qit{to enter} were also
used to refer to, for example, groups of passengers entering planes,
and if situations of this kind were modelled in the database as
non-instantaneous, one would have to distinguish between two
homonyms \qit{to enter}, one used with flights entering sectors, and
one with passengers entering planes. The first would be a point verb;
the second would not. 

The person applying the criterion will often have to decide exactly
what is or is not part of the situations described by the verbs.  The
database may store, for example, the time-points at which a flight
starts to board, finishes boarding, starts to taxi to a runway,
arrives at the runway, and leaves the ground. Before classifying the
non-habitual \qit{to depart}, one has to decide exactly what is or is
not part of departing. Is boarding part of departing,
i.e.\ is a flight departing when it is boarding? Is taxiing to a
runway part of departing? Or does departing include only the time at
which the flight actually leaves the ground?  If a flight starts to
depart when it starts to board, and finishes departing when it leaves
the ground, then the base form of \qit{to depart} should not be
classified as point, because the database does not treat departures as
instantaneous (it distinguishes between the beginning of the boarding
and the time when the flight leaves the ground). If, however,
departing starts when the front wheels of the aircraft leave the
ground and finishes when the rear wheels leave the ground, the base
form of \qit{to depart}
\emph{should} be classified as point, because the database does not
distinguish the two times. In any case, the user should be aware of
what \qit{to depart} is taken to mean. 

The point criterion is similar to claims in 
\cite{Vendler}, \cite{Singh}, \cite{Vlach1993},
and elsewhere that achievement (point) verbs denote instantaneous
situations.

\subsection{The imperfective paradox criterion} \label{ip_criterion}

The third criterion distinguishes activity from 
culminating activity verbs (state and point verbs will have already
been separated by the point and simple present criteria). The
criterion is based on the imperfective paradox (section
\ref{tense_aspect_intro}). Assertions containing the past
continuous and simple past of the verbs, like \pref{crit:20} --
\pref{crit:23}, are considered. 
\begin{examps}
\item John was running. \label{crit:20}
\item John ran. \label{crit:21}
\item John was building a house.  \label{crit:22}
\item John built a house. \label{crit:23}
\end{examps}
The reader is reminded that
assertions are treated as yes/no questions (section \ref{no_issues}).
If an affirmative answer to the past continuous assertion implies an
affirmative answer to the simple past assertion (as in 
\pref{crit:20} -- \pref{crit:21}), the verb
is an activity one; otherwise (e.g.\ \pref{crit:22} --
\pref{crit:23}), it is a culminating activity one.

As will be discussed in section \ref{progressives}, the past
continuous sometimes has a futurate meaning. Under this reading,
\pref{crit:20} means \qit{John was going to run.}, and an affirmative
answer to \pref{crit:20} does not necessarily imply an affirmative
answer to \pref{crit:21}. When applying the imperfective paradox
criterion, the past continuous must not have its futurate meaning.
In various forms, the imperfective paradox criterion has been
used in \cite{Vendler}, \cite{Vlach1993}, 
\cite{Kent}, and elsewhere.

\subsection{Other criteria} \label{other_aspect_criteria}

The three criteria above are not the only ones that could be used. The
behaviour of verbs when appearing in various forms or when combining
with some temporal adverbials varies depending on their aspectual
classes.  Alternative criteria can be formulated by observing this
behaviour.  For example, some authors classify verbs (or situations
denoted by verbs) by observing how easily they appear in progressive
forms (to be discussed in section \ref{progressives}), how easily they
combine with \qit{for~\dots} and \qit{in~\dots} duration adverbials
(sections \ref{for_adverbials} and \ref{in_adverbials} below), or what
the verbs entail about the start or the end of the described situation
when they combine with \qit{at \dots} temporal adverbials (section
\ref{point_adverbials} below). In some cases, the person classifying
the base verb forms may be confronted with a verb for which the three
criteria of sections \ref{simple_present_criterion} --
\ref{ip_criterion} do not yield a clear verdict. In such cases,
additional evidence for or against classifying a base verb form into a
particular class can be found by referring to following sections,
where the typical behaviour of each class is examined.

\subsection{Classifying base verb forms in the airport domain}
\label{aspect_examples} 

To illustrate the use of the criteria of sections
\ref{simple_present_criterion} -- \ref{ip_criterion}, I 
now consider a hypothetical \nlitdb to a temporal database that
contains information about the air-traffic of an airport. (I borrow
some terminology from \cite{Sripada1994}. The airport domain will be
used in examples throughout this thesis.) The airport database shows
the times when flights arrived at, or departed from, the airport, the
times flights spent circling around the airport while waiting for
permission to land, the runways they landed on or took off from, the
gates where the flights boarded, etc. The database is
occasionally queried using the \nlitdb to determine the causes of
accidents, and to collect data that are used to
optimise the airport's traffic-handling strategies. 

The airport's airspace is divided into sectors. Flights
approaching or leaving the airport cross the boundaries of 
sectors, each time \emph{leaving} a sector and \emph{entering}
another one. The airport is very busy, and some of its runways may
also \emph{be closed} for maintenance. Hence, approaching flights are
often instructed to \emph{circle} around the airport until a runway
\emph{becomes} free.  When a runway is freed,  flights 
\emph{start} to \emph{land}. Landing
involves following a specific procedure. In some cases,
the pilot may abort the landing procedure before completing
it. Otherwise, the flight lands on a runway, and it then
\emph{taxies} to a gate that \emph{is free}.  The moment at which the
flight \emph{reaches} the gate is considered the time at which the
flight \emph{arrived} (reaching a location and arriving are modelled
as instantaneous). Normally (habitually) each flight arrives at the
same gate and time every day. Due to traffic congestion, however, a
flight may sometimes arrive at a gate or time other than its normal
ones.

Before \emph{taking off}, each flight is \emph{serviced} by a service
company. This involves carrying out a specific set of tasks. Unless
all tasks have been carried out, the service is incomplete. Each
service company normally (habitually) services particular
flights. Sometimes, however, a company may be asked to service a
flight that it does not normally service. After being serviced, a
flight may be \emph{inspected}. Apart from flights, inspectors also
inspect gates and runways. In all cases, there are particular tasks to
be carried out for the inspections to be considered complete.

Shortly before taking off, flights start to \emph{board}. Unless all
the passengers that have checked in enter the aircraft, the boarding
is not complete, and the flight cannot depart. (There are special
arrangements for cases where passengers are too late.) The flight then
\emph{leaves} the gate, and that moment is considered the time at
which the flight \emph{departed} (leaving a location and departing are
modelled as instantaneous). Normally (habitually) each flight departs
from the same gate at the same time every day. Sometimes, however, flights
depart from gates, or at times, other than their normal ones. After
leaving its gate, a flight may be told to \emph{queue} for a
particular runway, until that runway becomes free.  When the runway is free,
the flight starts to \emph{take off}, which involves following a
specific procedure. As with landings, the pilot may 
abort the taking off procedure before completing it.

The database also records the states of parts of the airport's
emergency system. There are, for example, emergency tanks, used by the
fire-brigade. Some of those may
\emph{contain} water, others may contain foam, and others may
\emph{be empty} for maintenance. 

\begin{table}
\begin{center}
{\small
\begin{tabular}{|l|l|l|l|}
\hline
state verbs & activity verbs & culm.\ activity verbs & point verbs \\
\hline
 service (habitually) &  circle &  land &  cross \\
 arrive (habitually) &  taxi (no destination) &  take off &  enter \\
 depart (habitually) &  queue &  service (actually) &  become \\
 contain & &  inspect &  start/begin \\
 be (non-auxiliary) & &  board &  stop/finish \\
& &  taxi (to destination) &  reach \\
&&&  leave \\
&&&  arrive (actually) \\
&&&  depart (actually) \\
\hline
\end{tabular}
} 
\end{center}
\caption{Verbs of the airport domain}
\label{airport_verbs}
\end{table}

Table \ref{airport_verbs} shows some of the verbs that are used in the airport
domain. \qit{To depart}, \qit{to arrive}, and \qit{to service} are
used with both habitual and non-habitual meanings. 
\pref{criteg:1.3} and \pref{criteg:1.4}, for example, can have
habitual meanings. In \pref{criteg:1.3.2} and
\pref{criteg:1.4.1}, the verbs are probably used with their
non-habitual meanings. I distinguish between
habitual and non-habitual homonyms of \qit{to depart}, \qit{to
arrive}, and \qit{to service} (section \ref{aspectual_classes}). 
\begin{examps}
\item Which flights depart/arrive at 8:00am? \label{criteg:1.3}  
\item Which flight departed/arrived at 8:00am yesterday? \label{criteg:1.3.2}
\item Which company services BA737? \label{criteg:1.4}
\item Which company serviced BA737 yesterday? \label{criteg:1.4.1}
\end{examps}
I also distinguish between two homonyms of \qit{to taxi}, one that
requires a destination-denoting complement (as in \qit{BA737 was
taxiing to gate 2.}), and one that requires no such complement (as in
\qit{BA737 was taxiing.}).

The simple present criterion and sentences like \pref{criteg:1.1},
\pref{criteg:1.2}, \pref{criteg:1.3}, and
\pref{criteg:1.4} imply that the non-auxiliary \qit{to be}, \qit{to
contain}, and the habitual \qit{to depart}, \qit{to arrive}, and
\qit{to service} are state verbs.
\begin{examps}
\item Which gates are free? \label{criteg:1.1}
\item Does any tank contain foam? \label{criteg:1.2}
\end{examps}
All other verbs of table \ref{airport_verbs} are not state verbs.
For example, (excluding habitual and futurate meanings)
\pref{criteg:4} -- \pref{criteg:12} sound unlikely or odd in the
airport domain. \pref{criteg:19} -- \pref{criteg:19.1} would be used instead. 
\begin{examps}
\item \odd Which flights circle? \label{criteg:4}
\item \odd Which flight taxies to gate 2? \label{criteg:9}
\item \odd Which flight departs? \label{criteg:12} 
\item Which flights are circling? \label{criteg:19}
\item Which flight is taxiing to gate 2? \label{criteg:20}
\item Which flight is departing? \label{criteg:19.1}
\end{examps}
The verbs in the rightmost column of table \ref{airport_verbs} are
used in the airport domain to refer to situations which I assume 
are modelled as instantaneous in the database. Consequently, by 
the point criterion these are all point verbs. In contrast, I assume
that the situations of the verbs in the two middle columns are not
modelled as instantaneous. Therefore, those are activity or
culminating activity verbs. 

In the airport domain, a sentence like \pref{criteg:31} means that
BA737 spent some time circling around the airport. It does not imply
that BA737 completed any circle around the airport. Hence, an
affirmative answer to \pref{criteg:30} implies an affirmative answer
to \pref{criteg:31}. By the imperfective paradox criterion, \qit{to
circle} is an activity verb.
\begin{examps}
\item BA737 was circling. \label{criteg:30}
\item BA737 circled. \label{criteg:31}
\end{examps}
Similar assertions and the imperfective paradox criterion imply that
\qit{to taxi} (no destination) and \qit{to queue} are also activity
verbs. In contrast, the verbs in the third column of table
\ref{airport_verbs} are culminating activity verbs. For example, in
the airport domain an affirmative answer to \pref{criteg:43.2} does
not imply an affirmative answer to \pref{criteg:43.3}: J.Adams may
have aborted the inspection before completing all the inspection
tasks, in which case the inspection is incomplete. 
\begin{examps}
\item J.Adams was inspecting runway 5. \label{criteg:43.2}
\item J.Adams inspected runway 5. \label{criteg:43.3}
\end{examps}


\section{Verb tenses} \label{verb_tenses}

I now turn to verb tenses. I use ``tense'' with the meaning it has in
traditional English grammar textbooks (e.g.\ \cite{Thomson}). For
example, \qit{John sings.} and \qit{John is singing.} will be said to
be in the simple present and present continuous tenses
respectively. In linguistics, ``tense'' is not always used in this
way. According to \cite{Comrie2}, for example, \qit{John sings.} and
\qit{John is singing.} are in the same tense, but differ aspectually.

Future questions are not examined in this thesis (section
\ref{no_issues}). Hence, future tenses and futurate meanings of other
tenses (e.g.\ the scheduled-to-happen meaning of the simple present;
section \ref{simple_present_criterion}) will be ignored. To
simplify further the linguistic data, the present perfect continuous
and past perfect continuous (e.g.\ \qit{has/had been inspecting})
were also not considered: these tenses combine problems from both
continuous and perfect tenses. This leaves six tenses to be
discussed: simple present, simple past, present continuous, past
continuous, present perfect, and past perfect.

\subsection{Simple present} \label{simple_present}

The framework of this thesis allows the simple present to be used only
with state verbs, to refer to a situation that holds at the present
(e.g.\ \pref{sp:1}, \pref{sp:2}). 
\begin{examps}
\item Which runways are closed? \label{sp:1}
\item Does any tank contain water? \label{sp:2}
\end{examps}
Excluding the scheduled-to-happen meaning 
(which is ignored in this thesis), \pref{sp:3} can only be understood
as asking for the current normal (habitual) servicer of
BA737. Similarly, \pref{sp:4} can only be 
asking for the current normal departure gate of BA737. \pref{sp:3}
would not be used to refer to a company that is actually servicing
BA737 at the present moment (similar comments apply to \pref{sp:4}). That is,
\pref{sp:3} and \pref{sp:4} can only involve the habitual
homonyms of \qit{to service} and \qit{to depart} (which are state
verbs), not the non-habitual ones (which are culminating activity and
point verbs respectively; see table \ref{airport_verbs}). This is
consistent with the assumption that the simple present can only be used with
state verbs. 
\begin{examps}
\item Which company services BA737? \label{sp:3}
\item Which flights depart from gate 2? \label{sp:4}
\end{examps}
In the airport domain, \qit{to circle} is an activity verb (there is
no state habitual homonym). Hence, \pref{sp:3.7} is rejected. This is
as it should be, because \pref{sp:3.7} can only be understood with a
habitual meaning, a meaning which is not available in the airport
domain (there are no circling habits). 
\begin{examps}
\item Does BA737 circle? \label{sp:3.7}
\end{examps}

The simple present can also be used with non-state verbs to describe
events as they happen (section \ref{simple_present_criterion}), or
with a historic meaning (e.g.\ \qit{In 1673 a fire destroys the
palace.}), but these uses are extremely unlikely in \nlitdb questions.

\subsection{Simple past} \label{simple_past}

Like the simple present, the simple past can be used with verbs from
all four classes (e.g.\ \pref{spa:2} -- \pref{spa:7}). 
\begin{examps}
\item Which tanks contained water on 1/1/95? \label{spa:2}
\item Did BA737 circle on 1/1/95? \label{spa:5}
\item Which flights (actually) departed from gate 2 on 1/1/95? \label{spa:3}
\item Which flights (habitually) departed from gate 2 in 1994? \label{spa:1}
\item Which company (actually) serviced BA737 yesterday? \label{spa:6}
\item Which company (habitually) serviced BA737 last year? \label{spa:7}
\end{examps}
\pref{spa:3} -- \pref{spa:7} show that both the habitual and the
non-habitual homonyms of verbs like \qit{to depart} or \qit{to
service} are generally possible in the simple past. 
\pref{spa:8} is ambiguous. It may refer either to flights that
actually departed (perhaps only once) from gate 2 in 1994, or to
flights that normally (habitually) departed from gate 2 in 1994.
\begin{examps}
\item Which flights departed from gate 2 in 1994? \label{spa:8}
\end{examps}

The simple past of culminating activity verbs normally implies that
the situation of the verb reached its climax. For example, in
\pref{spa:13} the service must have been completed, and in
\pref{spa:12} BA737 must have reached gate 2 for the answer to be
affirmative.
\begin{examps}
\item Did Airserve service BA737? \label{spa:13}
\item Did BA737 taxi to gate 2? \label{spa:12}
\item BA737 was taxiing to gate 2 but never reached it. \label{spa:12.5}
\end{examps}
Some native English speakers consider simple negative answers to
\pref{spa:13} and \pref{spa:12} unsatisfactory, if for example BA737
was taxiing to gate 2 but never reached it. Although they agree that
strictly speaking the answer should be negative, they consider
\pref{spa:12.5} a much more appropriate answer. To
generate answers like \pref{spa:12.5}, a mechanism for
\emph{cooperative responses} is needed, an issue not addressed
in this thesis (section \ref{no_issues}). 

The simple past (and other tenses) often has an \emph{anaphoric}
nature. For example, \pref{spa:13} probably does not ask if Airserve
serviced BA737 at \emph{any} time in the past. \pref{spa:13} would
typically be used with a particular time in mind (e.g.\ the present day),
to ask if Airserve serviced BA737 during that time. As will be
discussed in section \ref{temporal_anaphora}, a temporal anaphora resolution
mechanism is needed to determine the time the speaker has in mind.
The framework of this thesis currently provides no such mechanism, and
\pref{spa:13} is taken to refer to any past time. (The same approach
is adopted with all other tenses that refer to past situations.)

\subsection{Present continuous and past continuous} \label{progressives}

\paragraph{Futurate meanings:} The present and past continuous can be used
with futurate meanings. In that case, 
\pref{prog:13} is similar to \pref{prog:14}. 
\begin{examps}
\item Who is/was inspecting BA737? \label{prog:13}
\item Who will/would inspect BA737? \label{prog:14}
\end{examps}
Futurate meanings of tenses are not examined in this thesis. Hence,
this use of the present and past continuous will be ignored. 

\paragraph{Activity and culminating activity verbs:}

The present and past continuous can be used with activity and
culminating activity verbs to refer to a situation that is or was in
progress (e.g.\ \pref{prog:1} -- \pref{prog:7} from the airport
domain).
\begin{examps}
\item Are/Were any flights circling? \label{prog:1}
\item Is/Was BA737 taxiing to gate 2? \label{prog:7}
\end{examps}
In the case of culminating activity verbs, there is no requirement for
the climax to be reached at any time. The past continuous version of
\pref{prog:7}, for example, does not require BA737 to have reached
gate 2 (cf.\ \pref{spa:12}).

\paragraph{Point verbs:}
The present and past continuous of point verbs 
often refers to a preparatory process that is or was ongoing, and that 
normally leads to the instantaneous situation of the verb. For
example, in the airport domain where \qit{to
depart} is a point verb and departing includes
only the moment where the flight leaves its gate, one could utter
\pref{prog:19.110} when the checking-in is ongoing or when 
the flight is boarding. 
\begin{examps}
\item BA737 is departing. \label{prog:19.110}
\end{examps}
The framework of this thesis provides no mechanism for determining
exactly which preparatory process is asserted to be in
progress. (Should the checking-in be in progress for the response to
\pref{prog:19.110} to be affirmative? Should the boarding be ongoing?)
Hence, this use of the present and past continuous of point verbs is
not allowed. The response to \pref{prog:19.110} will be affirmative
only at the time-point where BA737 is leaving its gate (as will be
discussed in section \ref{tsql2_time}, the database may model
time-points as corresponding to minutes or even whole days). To avoid
misunderstandings, the \nlitdb should warn the user that
\pref{prog:19.110} is taken to refer to the actual (instantaneous)
departure, not to any preparatory process. This is again a case for
cooperative responses, an issue not examined in this thesis. 

\paragraph{State verbs:}
It has often been observed (e.g.\ Vendler's tests in
section \ref{asp_taxes}) that state verbs are not normally
used in progressive forms. For example, \pref{prog:28} and
\pref{prog:30} are easily rejected by most native
speakers. (I assume that \qit{to own} and \qit{to consist} would be
classified as state verbs, on the basis that simple present questions
like \pref{prog:29} and \pref{prog:31} are possible.)
\begin{examps}
\item \bad Who is owning five cars? \label{prog:28}
\item Who owns five cars? \label{prog:29}
\item \bad Which engine is consisting of 34 parts? \label{prog:30}
\item Which engine consists of 34 parts? \label{prog:31}
\end{examps}
The claim that state verbs do not appear in the progressive is
challenged by sentences like \pref{prog:35} (from \cite{Smith1986},
cited in \cite{Passonneau}; \cite{Kent} and \cite{Vlach1993} provide
similar examples). \pref{prog:35} shows that the non-auxiliary \qit{to
be}, which is typically classified as state verb, can be used in the
progressive.
\begin{examps}
\item My daughter is being very naughty. \label{prog:35}
\end{examps}
Also, some native English speakers find \pref{prog:32} and
\pref{prog:36} acceptable (though they would use
the non-progressive forms instead). (I assume that \qit{to border}
would be classified as state verb, on the basis that \qit{Which
countries border France?} is possible.) 
\begin{examps}
\item \odd Tank 4 was containing water when the bomb exploded. \label{prog:32}
\item \odd Which countries were bordering France in 1937? \label{prog:36}
\end{examps}
Not allowing progressive forms of state verbs also seems problematic
in questions like \pref{prog:40}. \pref{prog:40} has a reading which
is very similar to the habitual reading of \pref{prog:41} (habitually
serviced BA737 in 1994).
\begin{examps}
\item Which company was servicing BA737 in 1994? \label{prog:40}
\item Which company serviced (habitually) BA737 in 1994? \label{prog:41}
\end{examps}
The reader is reminded that in the airport domain I distinguish
between a habitual and a non-habitual homonym of \qit{to service}. The
habitual homonym is a state verb, while the non-habitual one is a
culminating activity verb. If progressive forms of state verbs are not
allowed, then only the non-habitual homonym (actually servicing) is
possible in \pref{prog:40}. This does not account for the apparently
habitual meaning of \pref{prog:40}.

One could argue that the reading of \pref{prog:40} is not really habitual 
but \emph{iterative} (servicing many times, as opposed to having a
servicing habit). As pointed out in 
\cite{Comrie} (p.~27), the mere repetition of a situation does not
suffice for the situation to be considered a habit. \pref{prog:44},
for example, can be used when John is banging his hand on the table
repeatedly. In this case, it seems odd to claim that
\pref{prog:44} asserts that John has the habit of banging his hand on
the table, i.e.\ \pref{prog:44} does not seem to be equivalent to the
habitual \pref{prog:45}. 
\begin{examps}
\item John is banging his hand on the table.  \label{prog:44}
\item John (habitually) bangs his hand on the table.  \label{prog:45}
\end{examps}
In sentences like \pref{prog:40} -- \pref{prog:41}, however, the
difference between habitual and iterative meaning is hard to
define. For simplicity, I do not distinguish between habitual and
iterative readings, and I allow state verbs to be used in progressive
forms (with the same meanings as the non-progressive forms). This
causes \pref{prog:40} to receive two readings: one involving the
habitual \qit{to service} (servicing habitually in 1994), and one
involving the non-habitual \qit{to service} (actually servicing at
some time in 1994; this reading is more likely without the \qit{in
1994}). \pref{prog:28} and \pref{prog:30} are treated as equivalent to
\pref{prog:29} and \pref{prog:31}. 

As will be discussed in section \ref{temporal_adverbials}, I assume
that progressive tenses cause an aspectual shift from activities and
culminating activities to states. In the airport domain, for example,
although the base form of \qit{to inspect} is a culminating activity,
\qit{was inspecting} is a state.

\subsection{Present perfect} \label{present_perfect}

Like the simple past, the present perfect can be
used with verbs of all four classes to refer to past situations (e.g.\ 
\pref{prep:5} -- \pref{prep:10}). With culminating activity verbs, the
situation must have reached its climax (e.g.\ in \pref{prep:10}
the service must have been completed). 
\begin{examps}
\item Has BA737 (ever) been at gate 2? \label{prep:5}
\item Which flights have circled today? \label{prep:8}
\item Has BA737 reached gate 2? \label{prep:7}
\item Which company has (habitually) serviced BA737 this year? \label{prep:9}
\item Has Airserve (actually) serviced BA737? \label{prep:10}
\end{examps}
It has often been claimed (e.g.\ \cite{Moens}, \cite{Vlach1993},
\cite{Blackburn1994}) that the English present perfect asserts that
some consequence of the past situation holds at the present. For
example, \pref{prep:11} seems to imply that there is a consequence of
the fact that engine 5 caught fire that still holds (e.g.\ engine 5 is
still on fire, or it was damaged by the fire and has not been
repaired). In contrast, \pref{prep:12} does not seem to imply (at least not
as strongly) that some consequence still holds.
\begin{examps}
\item Engine 5 has caught fire. \label{prep:11}
\item Engine 5 caught fire. \label{prep:12}
\end{examps}
Although these claims are intuitively appealing, it is
difficult to see how they could be used in a \nlitdb. Perhaps in
\pref{prep:15} the \nlitdb should check not only that
the landing was completed, but also that some consequence
of the landing still holds.
\begin{examps}
\item Has BA737 landed? \label{prep:15}
\end{examps}
It is unclear, however, what this consequence should be. Should the
\nlitdb check that the plane is still at the airport? And should the
answer be negative if the plane has departed since the landing?
Should the \nlitdb check that the passengers of BA737 are still at the
airport? And should the answer be negative if the passengers have left
the airport? Given this uncertainty, the framework of this thesis does
not require the past situation to have present consequences.

When the present perfect combines with \qit{for~\dots} duration
adverbials (to be discussed in section \ref{for_adverbials}), there is
often an implication that the past situation still holds at the
present (this seems related to claims that the past situation must
have present consequences). For example, there is a reading of
\pref{prep:16.10} where J.Adams is still a manager. (\pref{prep:16.10}
can also mean that J.Adams was simply a manager for two years, without
the two years ending at the present moment.) In contrast,
\pref{prep:16.12} carries no implication that J.Adams is still a
manager (in fact, it seems to imply that he is no longer a
manager).
\begin{examps}
\item J.Adams has been a manager for two years. \label{prep:16.10}
\item J.Adams was a manager for two years. \label{prep:16.12}
\end{examps}
Representing in \topl the still-holding reading of
sentences like \pref{prep:16.10} has proven difficult. Hence, I 
ignore the possible implication that the past situation
still holds, and I treat \pref{prep:16.10} as equivalent to
\pref{prep:16.12}. 

The present perfect does not combine
felicitously with some temporal adverbials.  For example,
\pref{prep:16} and \pref{prep:19} sound at least odd
to most native English speakers (they would use \pref{prep:16.1} and
\pref{prep:19.1} instead).  In contrast, \pref{prep:17} and
\pref{prep:20} are acceptable.
\begin{examps}
\item \odd Which flights have landed yesterday? \label{prep:16}
\item Which flights landed yesterday? \label{prep:16.1}
\item Which flights have landed today? \label{prep:17}
\item \odd Which flights has J.Adams inspected last week? \label{prep:19}
\item Which flights did J.Adams inspect last week? \label{prep:19.1}
\item Which flights has J.Adams inspected this week? \label{prep:20}
\end{examps}
\pref{prep:16} -- \pref{prep:20} suggest that the present perfect can
only be used if the time of the adverbial contains not only
the time where the past situation occurred, but also the speech
time, the time when the sentence was uttered. (A similar
explanation is given on p.~167 of \cite{Thomson}.)  \pref{prep:17}
is felicitous, because \qit{today} contains the speech time.
In contrast, \pref{prep:16} is unacceptable, because \qit{yesterday}
cannot contain the speech time.  
The hypothesis, however, that the time of the adverbial must include
the speech time does not account for the fact that \pref{prep:22} is
acceptable by most native English speakers (especially
if the \qit{ever} is added), even if the question is not uttered on a
Sunday. 
\begin{examps}
\item Has J.Adams (ever) inspected BA737 on a Sunday? \label{prep:22}
\end{examps}
As pointed out in \cite{Moens2}, a superstitious person could also
utter \pref{prep:23} on a day other than Friday the 13th.
\begin{examps}
\item They have married on Friday 13th! \label{prep:23}
\end{examps}
One could attempt to formulate more elaborate restrictions, to predict
exactly when temporal adverbials can be used with the present
perfect. In the case of a \nlitdb, however, it is difficult
to see why this would be worth the effort, as opposed to simply
accepting questions like \pref{prep:16} as equivalent to
\pref{prep:16.1}. I adopt the latter approach.

Given that the framework of this thesis does not associate present
consequences with the present perfect, that the
still-holding reading of sentences like \pref{prep:16.10} is not
supported, and that questions like \pref{prep:16} are allowed, there
is not much left to distinguish the present perfect from the simple
past. Hence, I treat the present perfect as equivalent to the simple past.

\subsection{Past perfect} \label{past_perfect}

The past perfect is often used to refer to a situation that occurred at
some past time before some other past time. Following Reichenbach
\cite{Reichenbach} and many others, let us call the latter time the
\emph{reference time}. \pref{pap:1} and \pref{pap:4} have readings
where \qit{at 5:00pm} specifies the reference time. In that case,
\pref{pap:1} asks for flights that Airserve serviced before 5:00pm,
and \pref{pap:4} asks if BA737 was at gate 2 some time before
5:00pm. (When expressing these meanings, \qit{by 5:00pm} is probably
preferable. I claim, however, that \qit{at 5:00pm} can also be used in this
way. \qit{By~\dots} adverbials are not examined in this thesis.)
\begin{examps}
\item Which flights had Airserve serviced at 5:00pm? \label{pap:1}
\item Had BA737 been at gate 2 at 5:00pm? \label{pap:4}
\end{examps}
With culminating activity verbs, the climax must have been reached
before (or possibly at) the reference time. For example, in
\pref{pap:1} the services must have been completed up to 5:00pm.
Perhaps some consequence of the past situation must still hold at the
reference time (see similar comments about the present perfect in
section \ref{present_perfect}). As with the present perfect, however, I
ignore such consequential links.

When the past perfect combines with temporal adverbials, it is often
unclear if the adverbial is intended to specify the reference time or
directly the time of the past situation. For example, \pref{pap:6}
could mean that BA737 had already reached gate 2 at 5:00pm, or that it
reached it at 5:00pm. In the latter case, \pref{pap:6} is similar to
the simple past \pref{pap:6.1}, except that \pref{pap:6} creates the
impression of a longer distance between the time of the reaching and
the speech time.
\begin{examps}
\item BA737 had reached gate 2 at 5:00pm. \label{pap:6}
\item BA737 reached gate 2 at 5:00pm. \label{pap:6.1}
\end{examps}
When the past perfect combines with \qit{for~\dots} duration
adverbials and a reference time is specified, there is often an
implication that the past situation still held at the reference
time. (A similar implication arises in the case of the present
perfect; see section \ref{present_perfect}.) For example, \pref{pap:8}
seems to imply that J.Adams was still a manager on 1/1/94. As in the
case of the present perfect, I ignore this implication, for reasons
related to the difficulty of representing it in \topl.
\begin{examps}
\item J.Adams had been a manager for two years on 1/1/94. \label{pap:8}
\end{examps}
As will be discussed in section \ref{temporal_adverbials}, I assume
that the past perfect triggers an aspectual shift to state (e.g.\ the
base form of \qit{to inspect} is a culminating activity, but \qit{had
inspected} is a state). This shift seems to be a property of all
perfect tenses. For reasons, however, related to the fact that I treat
the present perfect as equivalent to the simple past (section
\ref{present_perfect}), I do not postulate any shift in the case of
the present perfect.


\section{Special temporal verbs} \label{special_verbs}

Through their tenses, all verbs can convey temporal
information. Some verbs, however, like \qit{to begin} or 
\qit{to precede}, are of a more
inherently temporal nature. These verbs 
differ from ordinary ones (e.g.\ \qit{to build}, \qit{to contain}) in
that they do not describe directly situations, but rather
refer to situations introduced by other
verbs or nouns. (A similar observation is made in \cite{Passonneau}.) In \pref{spv:1}, for example, \qit{to begin} refers to the
situation of \qit{to build}. \qit{To start}, \qit{to end}, \qit{to
finish}, \qit{to follow}, \qit{to continue}, and \qit{to happen} all
belong to this category of special temporal verbs. 
\begin{examps}
\item They began to build terminal 9 in 1985. \label{spv:1}
\end{examps}

From all the special temporal verbs, I have considered only \qit{to
start}, \qit{to begin}, \qit{to stop}, and \qit{to finish}. I
allow \qit{to start}, \qit{to begin}, \qit{to end}, and \qit{to
finish} to be used with state and activity verbs, even though with
state verbs \qit{to begin} and \qit{to finish} usually sound unnatural
(e.g.\ \pref{spv:11}), and with activity verbs (e.g.\ \pref{spv:13})
it could be argued that the use of \qit{to begin} or \qit{to finish}
signals that the speaker has in mind a culminating activity (not
activity) view of the situation.
\begin{examps}
\item \odd Which tank began to contain/finished containing water on
27/7/95? \label{spv:11} 
\item Which flight began/finished circling at 5:00pm? \label{spv:13}
\end{examps}
When combining with culminating activity verbs, \qit{to start} and
\qit{to begin} have the same meanings. \qit{To stop} and \qit{to
finish}, however, differ: \qit{to finish} requires the climax to be
reached, while \qit{to stop} requires the action or change to simply stop
(possibly without being completed). For example, in \pref{spv:20} the
service must have simply stopped, while in \pref{spv:21} it must have
been completed.
\begin{examps}
\item Which company stopped servicing (actually) BA737 at 5:00pm?
  \label{spv:20} 
\item Which company finished servicing (actually) BA737 at 5:00pm?
  \label{spv:21} 
\end{examps}
With point verbs (like \qit{to enter} and \qit{to leave} in the airport
domain), the use of \qit{to start}, \qit{to begin}, \qit{to stop}, and
\qit{to finish} (e.g.\ \pref{spv:22}, \pref{spv:23}) typically signals
that the person submitting the question is unaware that the situation
of the point verb is taken to be instantaneous. In these cases, I
ignore the temporal verbs (e.g.\ \pref{spv:22} is treated as
\pref{spv:24}). Ideally, the \nlitdb would also warn the user that the
temporal verb is ignored, and that the situation is modelled as
instantaneous (another case for cooperative responses; see section
\ref{no_issues}). The framework of this thesis, however, provides no
mechanism for generating such warnings. 
\begin{examps}
\item Which flight started to enter sector 2 at 5:00pm? \label{spv:22}
\item Which flight finished leaving gate 2 at 5:00pm? \label{spv:23}
\item Which flight entered sector 2 at 5:00pm? \label{spv:24}
\end{examps}


\section{Temporal nouns} \label{temporal_nouns}

Some nouns have a special temporal nature.  Nouns like
\qit{development} or \qit{inspection}, for example, are similar to
verbs, in that they introduce world situations that occur in time.
The role of \qit{development} in \pref{tn:0} is very similar to that
of \qit{to develop} in \pref{tn:0.1}. 
\begin{examps}
\item When did the development of \textsc{Masque} start?
   \label{tn:0} 
\item When did they start to develop \textsc{Masque}? \label{tn:0.1}
\end{examps}
Other nouns indicate temporal order (e.g.\ \qit{predecessor},
\qit{successor}), or refer to start or end-points (e.g.\ \qit{beginning},
\qit{end}). Finally, many nouns (and proper names) refer to time
periods, points, or generally entities of the temporal ontology (e.g.\
\qit{minute}, \qit{July}, \qit{event}). 

From all the temporal nouns (and proper names), this
thesis examines only nouns like \qit{year}, \qit{month}, \qit{week},
\qit{day}, \qit{minute}, \qit{second}, \qit{1993}, \qit{July},
\qit{1/1/85}, \qit{Monday}, \qit{3:05pm}. Temporal nouns of a more
abstract nature (e.g.\ \qit{period}, \qit{point}, \qit{interval},
\qit{event}, \qit{time}, \qit{duration}), nouns referring to start or
end-points, nouns introducing situations, and nouns of temporal
precedence are not considered.


\section{Temporal adjectives} \label{temporal_adjectives}

There are also adjectives of a special temporal nature. Some refer to
a temporal order (e.g.\ \qit{current}, \qit{previous},
\qit{earliest}), others refer to durations (e.g.\ \qit{brief},
\qit{longer}), and others specify frequencies (e.g.\ \qit{annual},
\qit{daily}). Adjectives of this kind are not examined in this thesis,
with the exception of \qit{current} which is supported to illustrate
some points related to temporal anaphora (these points will be discussed 
in section \ref{noun_anaphora}). (``Normal'' adjectives, like \qit{open} and
\qit{free}, are also supported.)


\section{Temporal adverbials} \label{temporal_adverbials}

This section discusses adverbials that convey temporal
information. 

\subsection{Punctual adverbials} \label{point_adverbials}

Some adverbials are understood as specifying time-points.  Following
\cite{Vlach1993}, I call these \emph{punctual adverbials}. In English,
punctual adverbials are usually prepositional phrases introduced by
\qit{at} (e.g.\ \qit{at 5:00pm}, \qit{at the end of the
inspection}). In this thesis, only punctual adverbials consisting of
\qit{at} and clock-time expressions (e.g.\ \qit{at 5:00pm}) are
considered.

\paragraph{With states:}
When combining with state expressions, punctual adverbials specify a
time where the situation of the state expression holds. There is
usually no implication that the situation of the state starts or stops
at the time of the adverbial. \pref{pa:1}, for example, asks if tank 5
was empty at 5:00pm. There is no requirement that the tank must have
started or stopped being empty at 5:00pm. Similar comments apply to 
\pref{pa:2}. 
\begin{examps}
\item Was tank 5 empty at 5:00pm? \label{pa:1}
\item Which flight was at gate 2 at 5:00pm? \label{pa:2}  
\end{examps}
In other words, with states punctual adverbials normally have an
\emph{interjacent} meaning, not an \emph{inchoative} or
\emph{terminal} one. (``Interjacent'', ``inchoative'',
and ``terminal'' are borrowed from \cite{Kent}. Kent explores the
behaviour of \qit{at}, \qit{for}, and \qit{in} adverbials, and arrives
at conclusions similar to the ones presented here.)

In narrative contexts, punctual adverbials combining with
states sometimes have inchoative meanings. For example, the \qit{at 8:10am} in
\pref{pa:4} most probably specifies the time when J.Adams arrived
(started being) in Glasgow. In \nlitdb questions, however, this
inchoative meaning seems unlikely. For example, it seems unlikely that
\pref{pa:5} would be used to enquire about persons that
\emph{arrived} in Glasgow at 8:10am. Hence, for the purposes of this
thesis, it seems reasonable to assume that punctual adverbials
combining with states always have interjacent meanings.
\begin{examps}
\item J.Adams left Edinburgh early in the morning, and at 8:10am he
  was in Glasgow. \label{pa:4}
\item Who was in Glasgow at 8:10am? \label{pa:5}
\end{examps}

\paragraph{With points:}
With point expressions, punctual adverbials specify the time where
the instantaneous situation of the point expression takes
place (e.g.\ \pref{pa:8}, \pref{pa:9}; \qit{to enter} and \qit{to
reach} are point verbs in the airport domain).
\begin{examps}
\item Which flight entered sector 2 at 23:02? \label{pa:8}
\item Which flight reached gate 5 at 23:02? \label{pa:9}
\end{examps}
\pref{pa:10} is ambiguous. It may either involve the
non-habitual homonym of \qit{to depart} (this homonym is a point verb
in the airport domain), in which case 5:00pm is
the time of the actual departure, or the state habitual homonym (to
depart habitually at some time), in which case 5:00pm is the habitual
departure time. In the latter case, I treat \qit{at 5:00pm} as a
prepositional phrase complement of the habitual \qit{to depart}, not
as a temporal adverbial. This reflects the fact that the \qit{at
5:00pm} does not specify a time when 
the habit holds, but it forms part of the
description of the habit, i.e.\ it is used in a way very similar to
how \qit{from gate 2} is used in \pref{pa:3}.
\begin{examps}
\item Which flight departed at 5:00pm? \label{pa:10}
\item Which flight departed (habitually) from gate 2 (last year)? \label{pa:3}
\end{examps}

\paragraph{With activities:}

With activities, punctual adverbials usually have an inchoative
meaning, but an interjacent one is also possible in some cases.
\pref{pa:11}, for example, could refer to a flight that either joined
the queue of runway 2 at 5:00pm or was simply in
the queue at 5:00pm. (In the airport domain, \qit{to queue} and
\qit{to taxi} (no destination) are activity verbs.) The
inchoative meaning seems the preferred one in \pref{pa:11}. It
also seems the preferred one in \pref{pa:13}, though an interjacent
meaning is (arguably) also possible. The interjacent meaning is
easier to accept in \pref{pa:14}.
\begin{examps}
\item Which flight queued for runway 2 at 5:00pm? \label{pa:11}
\item BA737 taxied at 5:00pm. \label{pa:13}
\item Which flights circled at 5:00pm?  \label{pa:14}
\end{examps}
With past continuous forms of activity verbs, however, punctual
adverbials normally have only interjacent meanings (compare
\pref{pa:11} and \pref{pa:13} to \pref{pa:17} and 
\pref{pa:19}). (One would not normally use punctual adverbials with
present continuous forms, since in that case the situation is known to
take place at the present.)
\begin{examps}
\item Which flight was queueing for runway 2 at 5:00pm? \label{pa:17}
\item BA737 was taxiing at 5:00pm. \label{pa:19}
\end{examps}
To account for sentences like \pref{pa:17} and \pref{pa:19} (and other
phenomena to be discussed in following sections), I classify the
progressive tenses (present and past continuous) of activity and
culminating activity verbs as states. For example, in the airport
domain, the base form of \qit{to queue} is an activity. Normally, all
other forms of the verb (e.g.\ the simple past) inherit the aspectual
class of the base form. The progressive tenses (e.g.\ \qit{is
queueing}, \qit{was queueing}) of the verb, however, are states. (The
progressive can be seen as forcing an aspectual shift from activities
or culminating activities to states. No such aspectual shift is needed
in the case of point verbs.) This arrangement, along with the assumption
that punctual adverbials combining with states have only interjacent
meanings, accounts for the fact that \pref{pa:17} and \pref{pa:19}
have only interjacent meanings. In various forms, assumptions that
progressive tenses cause  aspectual shifts to states have also been
used in \cite{Dowty1986}, \cite{Moens}, \cite{Vlach1993}, \cite{Kent},
and elsewhere. 

\paragraph{With culminating activities:}
When combining with culminating activities, punctual adverbials
usually have inchoative or terminal meanings (when they have terminal
meanings, they specify the time when the climax was
reached). The terminal reading is the preferred one in \pref{pa:23}.
In \pref{pa:25} both readings seem possible. In \pref{pa:24} the
inchoative meaning seems the preferred one. (In the airport domain,
\qit{to land}, \qit{to taxi} (to destination), and \qit{to inspect}
are culminating activity verbs.)
\begin{examps}
\item Which flight landed at 5:00pm? \label{pa:23}
\item Which flight taxied to gate 4 at 5:00pm? \label{pa:25}
\item Who inspected BA737 at 5:00pm? \label{pa:24}
\end{examps}
Perhaps, as with activities, an interjacent meaning
is sometimes also possible (e.g.\ \pref{pa:25} would refer to a flight
that was on its way to gate 4 at 5:00pm). This may be true, but with
culminating activities the inchoative or terminal reading is usually
much more dominant. For simplicity, I ignore the possible interjacent
meaning in the case of culminating activities.

With past continuous forms of culminating activity verbs, punctual
adverbials normally have only interjacent meanings.
Compare, for example, \pref{pa:23} -- \pref{pa:24} to \pref{pa:28} --
\pref{pa:30}. This is in accordance with the assumption that the
progressive tenses of activity and culminating activity verbs are
states. 
\begin{examps}
\item Which flight was landing at 5:00pm? \label{pa:28}
\item Which flight was taxiing to gate 4 at 5:00pm? \label{pa:29}
\item Who was inspecting BA737 at 5:00pm? \label{pa:30}
\end{examps}

\paragraph{With past perfect:}
As discussed in section \ref{past_perfect}, in sentences like
\pref{pa:31} the adverbial can be taken to refer either directly to
the taxiing (the taxiing started or ended at 5:00pm) or to the 
reference time (the taxiing had already finished at 5:00pm).
\begin{examps}
\item BA737 had taxied to gate 2 at 5:00pm. \label{pa:31}  
\item BA737 had [taxied to gate 2 at 5:00pm]. \label{pa:32}
\item BA737 [had taxied to gate 2] at 5:00pm. \label{pa:33}
\end{examps}
The way in which sentences like \pref{pa:31} are treated in this
thesis will become clearer in chapter \ref{TOP_chapter}. A rough
description, however, can be given here. \pref{pa:31} is treated as
syntactically ambiguous between \pref{pa:32} (where the adverbial
applies to the past participle \qit{taxied}) and \pref{pa:33} (where
the adverbial applies to the past perfect \qit{had taxied}). The past
participle (\qit{taxied}) inherits the aspectual class of the base
form, and refers directly to the situation of the verb (the
taxiing). In contrast, the past perfect (\qit{had taxied}) is always
classified as state (the past perfect can be seen as causing an
aspectual shift to state), and refers to a time-period that starts
immediately after the end of the situation of the past participle (the
end of the taxiing), and extends up to the end of time. Let us call
this period the \emph{consequent period}.

In the airport domain, the base form of \qit{to taxi} (to destination)
is a culminating activity. Hence, the past participle \qit{taxied}
(which refers directly to the taxiing) is also a culminating
activity. In \pref{pa:32}, a punctual adverbial combines with the
(culminating activity) past participle.  According to the discussion
above, two readings arise: an inchoative (the taxiing started at
5:00pm) and a terminal one (the taxiing finished at 5:00pm).  In
contrast, in \pref{pa:33} the punctual adverbial combines with the
past perfect \qit{had taxied}, which is a state expression that refers
to the consequent period. Hence, only an interjacent reading arises:
the time of the adverbial must simply be within the consequent period
(there is no need for the adverbial's time to be the beginning or end
of the consequent period). This requires the taxiing to have been
completed at the time of the adverbial. A similar arrangement is used
when the past perfect combines with period adverbials, duration
\qit{for~\dots} and \qit{in~\dots} adverbials, or temporal subordinate
clauses (to be discussed in following sections).
The assumption that the past perfect causes an aspectual shift to
state is similar to claims in \cite{Moens},
\cite{Vlach1993}, and elsewhere, that English perfect forms are (or refer
to) states. 

\paragraph{Lexical, consequent, and progressive states:}
There is sometimes a need to distinguish between expressions
that are states because they have inherited the state aspectual class
of a base verb form, and expressions that are states because of an aspectual
shift introduced by a past perfect or a progressive
tense. Following \cite{Moens}, I use the terms
\emph{lexical state}, \emph{consequent state}, and \emph{progressive
state} to distinguish the three genres. In the airport domain, for
example, the base form of \qit{to queue} is a lexical state. The
simple past \qit{queued} and the past participle \qit{queued} are also
lexical states. The past perfect \qit{had queued} is a consequent
state, while the present continuous form \qit{is queueing} is a
progressive state.

Finally, for reasons that will be discussed in section
\ref{hpsg:mult_mods}, I assume that punctual adverbials cause the
aspectual class of the syntactic constituent they modify to become
point. In \pref{pa:33}, for example, the \qit{taxied to gate 2}
inherits the culminating activity aspectual class of the base form.
The past perfect causes the aspectual class of \qit{had taxied to gate
  2} to become consequent state. Finally, the \qit{at 5:00pm} causes
the aspectual class of \qit{had departed at 5:00pm} to become point.
Table \ref{punctual_adverbials_table} summarises the main points of
this section.

\begin{table}
\begin{center}
{\small
\begin{tabular}{|l|l|}
\hline
\multicolumn{2}{|c|}{meanings of punctual adverbials} \\
\hline \hline
with state & interjacent \\
\hline
with activity & inchoative or interjacent \\
\hline 
with culm.\ activity & inchoative or terminal \\
\hline
with point & specifies time of instantaneous situation \\
\hline \hline
\multicolumn{2}{|l|}{The resulting aspectual class is point.}\\
\hline
\end{tabular}
}
\end{center}
\caption{Punctual adverbials in the framework of this thesis} 
\label{punctual_adverbials_table}
\end{table}

\subsection{Period adverbials} \label{period_adverbials}

Unlike punctual adverbials, which are understood as specifying points
in time, adverbials like \qit{in 1991}, \qit{on Monday},
\qit{yesterday} (e.g.\ \pref{padv:1} -- \pref{padv:2}) are usually
understood as specifying longer, non-instantaneous periods of time.
In \pref{padv:1}, for example, the period of \qit{in 1991} covers the
whole 1991. I call adverbials of this kind \emph{period adverbials}.
\begin{examps}
\item Who was the sales manager in 1991? \label{padv:1}
\item Did BA737 circle on Monday? \label{padv:3}
\item Which flights did J.Adams inspect yesterday? \label{padv:2}
\end{examps}
\qit{Before~\dots} and \qit{after~\dots} adverbials (e.g.\ \pref{padv:3.2})
can also be considered period adverbials, except that in this case one
of the boundaries of the period is left unspecified. (I model time as
bounded; see section \ref{temporal_ontology} below. In the absence of other
constraints, I treat the unspecified boundary as the beginning or end
of time.) In \pref{padv:3.2}, for example, the end of the period is
the beginning of 2/5/95; the beginning of the period is left
unspecified. \qit{Before} and \qit{after} can also introduce
temporal subordinate clauses; this will be discussed in section
\ref{before_after_clauses}. 
\begin{examps}
\item Which company serviced BA737 before 2/5/95? \label{padv:3.2}
\end{examps}

This thesis examines only period adverbials introduced by \qit{in},
\qit{on}, \qit{before}, and \qit{after}, as well as \qit{today} and
\qit{yesterday}. (\qit{In~\dots} adverbials can also 
specify durations, e.g.\ \qit{in two hours}; this will be discussed in
section \ref{in_adverbials}.) Other period adverbials, like \qit{from 1989 to
1990}, \qit{since 1990}, \qit{last week}, or \qit{two days ago}, are
not considered. Extending the framework of this thesis to support
more period adverbials should not be difficult.

\paragraph{With states:}

When period adverbials combine with state expressions, the situation of
the state expression must hold for at least some time during the
period of the adverbial. In \pref{padv:10}, for example, the person must
have been a manager for at least some time in 1995. Similarly, in
\pref{padv:11}, the person must have been at gate 2 for at least some
time on the previous day. 
\begin{examps}
\item Who was a manager in 1995? \label{padv:10}  
\item Who was at gate 2 yesterday? \label{padv:11}
\end{examps}
There is often, however, an implication that the situation 
holds \emph{throughout} the period of the adverbial. 
\pref{padv:13}, for example, could mean that the tank was empty
throughout January, not at simply some part of January. Similarly, in
\pref{padv:12} the user could be referring to tanks that were empty
\emph{throughout} January. In that case, if a tank was empty only some days in
January and the \nlitdb included that tank in the answer, the user
would be misled to believe that the tank was empty throughout
January. Similar comments can be made for
\pref{padv:13.9} and \pref{padv:15}. 
\begin{examps}
\item Tank 4 was empty in January. \label{padv:13}
\item Which tanks were empty in January? \label{padv:12}
\item Was runway 2 open on 6/7/95? \label{padv:13.9}
\item Which flights departed (habitually) from gate 2 in 1993? \label{padv:15}
\end{examps}
The same implication is possible in sentences with \qit{before~\dots}
or \qit{after~\dots} adverbials.  \pref{padv:20.1}, for example, could
mean that the runway was open all the time from some unspecified time
up to immediately before 5:00pm (and possibly longer).
\begin{examps}
\item Runway 2 was open before 5:00pm. \label{padv:20.1}
\end{examps}
One way to deal with such implications is to treat sentences where
period adverbials combine with states as ambiguous. That is, to
distinguish between a reading where the situation holds throughout the
adverbial's period, and a reading where the situation holds at simply
some part of the adverbial's period. \cite{Vlach1993} (p.~256) uses
the terms \emph{durative} and \emph{inclusive} to refer to the two
readings. (A \nlitdb could paraphrase both readings and ask the user
to select one, or it could provide answers to both readings,
indicating which answer corresponds to which reading.)  This approach
has the disadvantage of always generating two readings, even in cases
where the durative reading is clearly impossible. For example, when
the state expression combines not only with a period adverbial but
also with a \qit{for~\dots} duration adverbial, the meaning can never
be that the situation must necessarily hold all the time of the
adverbial's period.  For example, \pref{padv:29} can never mean that
the tank must have been empty throughout January (cf.\ 
\pref{padv:12}).
\begin{examps}
\item Which tank was empty for two days in January? \label{padv:29}
\item When on 6/7/95 was tank 5 empty? \label{padv:27}
\end{examps}
Similarly, in time-asking questions like \pref{padv:27}, the durative
reading is impossible. \pref{padv:27} can never mean that the tank
must have been empty throughout 6/7/95 (cf.\
\pref{padv:13.9}). Formulating an account of exactly when the durative
reading is possible is a task which I have not undertaken. 
Although in chapter \ref{TOP_chapter} I discuss how the distinction
between durative and inclusive readings could be captured in \topl,
for simplicity in the rest of this thesis I consider only the
inclusive readings, ignoring the durative ones. 

\paragraph{With points:}
When period adverbials combine with point expressions, the period of
the adverbial must contain the time where the instantaneous situation of
the point expression occurs (e.g.\ \pref{padv:62}). 
\begin{examps}
\item Did BA737 enter sector 5 on Monday? \label{padv:62}
\end{examps}

\paragraph{With culminating activities:}
When period adverbials combine with culminating activity expressions,
I allow two possible readings: (a) that the situation of the
culminating activity expression both starts and reaches its completion
within the adverbial's period, or (b) that the situation simply
reaches its completion within the adverbial's period. In the second
reading, I treat the culminating activity expression as referring to
only the completion of the situation it would normally describe, and
the aspectual class is changed to point.

The first reading is the preferred one in \pref{padv:42} which is
most naturally understood as referring to a runner who both started
and finished running the 40 miles on Wednesday (\qit{to run} (a 
distance) is typically classified as culminating activity verb).
\begin{examps}
\item Who ran 40 miles on Wednesday? \label{padv:42}
\end{examps}
In the airport domain, the first reading is the preferred one in
\pref{padv:47.1} (the inspection both started and was completed on
Monday). 
\begin{examps}
\item J.Adams inspected BA737 on Monday. \label{padv:47.1}
\end{examps}
The second reading (the situation simply reaches its
completion within the adverbial's period) is needed in questions
like \pref{padv:48} and \pref{padv:49}. In the airport domain,
\qit{to land} and \qit{to take off} are culminating
activity verbs (landings and taking offs involve following particular
procedures; the landing or taking off starts when the pilot starts the
corresponding procedure, and is completed when that procedure is
completed). If only the first reading were available (both start and
completion within the adverbial's period), in 
\pref{padv:48} the \nlitdb would report only flights that both started
and finished landing on Monday. If a flight started the
landing procedure at 23:55 on Sunday and finished it at 00:05
on Monday, that flight would not be reported. This seems over-restrictive. In
\pref{padv:48} the most natural reading is that the flights must have
simply touched down on Monday, i.e.\ the landing must have simply been
completed within Monday. Similar comments can be made for
\pref{padv:49} and \pref{padv:52} (in domains where \qit{to fix} is a
culminating activity verb). 
\begin{examps}
\item Which flights landed on Monday? \label{padv:48}
\item Which flights took off after 5:00pm? \label{padv:49}
\item Did J.Adams fix any faults yesterday? \label{padv:52}  
\end{examps}
The problem in these cases is that \qit{to land}, \qit{to take off},
and \qit{to fix} need to be treated as point verbs (referring to only
the time-points where the corresponding situations are completed),
even though they have been classified as culminating activity verbs
(section \ref{aspect_examples}). The second reading allows exactly
this. The culminating activity expression is taken to refer to only
the completion point of the situation it would normally describe, its
aspectual class is changed to point, and the completion point is
required to fall within the adverbial's period.

The fact that two readings are allowed when period adverbials combine
with culminating activities means that sentences like
\pref{padv:47.1} -- \pref{padv:52} are treated as ambiguous. In
all ambiguous sentences, I assume that a \nlitdb would present
all readings to the user asking them to choose one, or that it would
provide answers to all readings, showing which answer corresponds to
which reading. (The prototype \nlitdb of this thesis adopts the
second strategy, though the mechanism for explaining which answer
corresponds to which reading is primitive: the readings are shown as
\topl formulae.)

In the case of \qit{before~\dots} adverbials (e.g.\ \pref{padv:49b}),
the two readings are semantically equivalent: requiring the situation
to simply reach its completion before some time is equivalent to
requiring the situation to both start and reach its completion before
that time. To avoid generating two equivalent readings, I allow only
the reading where the situation both starts and reaches its completion
within the adverbial's period.
\begin{examps}
\item Which flights took off before 5:00pm? \label{padv:49b}
\end{examps}

Even with the second reading, the answers of the \nlitdb may not
always be satisfactory. Let us assume, for example, that J.Adams
started inspecting a flight late on Monday, and finished the
inspection early on Tuesday. None of the two readings would include
that flight in the answer to \pref{padv:47}, because both require the
completion point to fall on Monday. While strictly speaking this seems
correct, it would be better if the \nlitdb could also include in the
answer inspections that \emph{partially} overlap the adverbial's 
period, warning the user about the fact that these inspections are not
completely contained in the adverbial's period. This is another case
where cooperative responses (section \ref{no_issues}) are needed. 
\begin{examps}
\item Which flights did J.Adams inspect on Monday? \label{padv:47}  
\end{examps}

Finally, I note that although in the airport domain \qit{to taxi} (to
destination) is a culminating activity verb, in \pref{padv:61.7} the
verb form is a (progressive) state. Hence, the \nlitdb's answer would
be affirmative if BA737 was taxiing to gate 2 some time within the
adverbial's period (before 5:00pm), even if BA737 did not reach the
gate during that period. This captures correctly the most natural
reading of
\pref{padv:61.7}. 
\begin{examps}
\item Was BA737 taxiing to gate 2 before 5:00pm? \label{padv:61.7}
\end{examps}

\paragraph{With activities:}
When period adverbials combine with activities, I require the
situation of the verb to hold for at least some time within
the adverbial's period (same meaning as with
states). In \pref{padv:66}, for example, the flight must have
circled for at least some time on Monday, and in \pref{padv:67} the
flights must have taxied for at least some time after 5:00pm. 
\begin{examps}
\item Did BA737 circle on Monday? \label{padv:66}
\item Which flights taxied after 5:00pm? \label{padv:67}
\end{examps}
Another stricter reading is sometimes possible (especially with
\qit{before} and \qit{after}): that the situation 
does not extend past the boundaries of the adverbial's period.
For example, \pref{padv:67} would refer to flights that \emph{started}
to taxi after 5:00pm (a flight that started to taxi
at 4:55pm and continued to taxi until 5:05pm would not be
reported). This reading is perhaps also possible with states (e.g.\
\pref{padv:20.1}), though with activities it seems easier to
accept. As a simplification, such readings are ignored in this thesis.

\paragraph{Elliptical forms:} \qit{Before} and \qit{after} are sometimes
followed by noun phrases that do not denote entities of the temporal
ontology (e.g.\ \pref{padv:71}). 
\begin{examps}
\item Did J.Adams inspect BA737 before/after UK160? \label{padv:71}
\item Did J.Adams inspect BA737 before/after he inspected UK160?
  \label{padv:71.1}
\end{examps}
Questions like \pref{padv:71} can be considered elliptical forms of
\pref{padv:71.1}, i.e.\ in these cases \qit{before} and \qit{after}
could be treated as when they introduce subordinate clauses
(section \ref{before_after_clauses} below). Questions like \pref{padv:71}
are currently not supported by the framework of this thesis. 

Table \ref{period_adverbials_table} summarises the main points of this
section. 

\begin{table}
\begin{center}
{\small
\begin{tabular}{|l|l|}
\hline
\multicolumn{2}{|c|}{meanings of period adverbials} \\
\hline \hline
with state or activity & situation holds for at least part of adverbial's period \\
\hline 
with culm.\ activity & situation starts and is completed within
                       adverbial's period,\\ 
                     & or situation is simply completed within adverbial's 
                       period$^{*\dagger}$\\
\hline 
with point & instantaneous situation occurs within adverbial's period \\
\hline \hline
\multicolumn{2}{|l|}{$^*$Not with \qit{before~\dots} adverbials.} \\
\multicolumn{2}{|l|}{$^{\dagger}$The resulting aspectual class is point. (In
  all other cases the aspectual class} \\
\multicolumn{2}{|l|}{\ \ remains the same.)}\\
\hline
\end{tabular}
}
\end{center}
\caption{Period adverbials in the framework of this thesis} 
\label{period_adverbials_table}
\end{table}

\subsection{Duration \qit{for~\dots} adverbials} \label{for_adverbials}

This section discusses \qit{for~\dots} adverbials that 
specify durations (e.g.\ \pref{dura:1}). 
\begin{examps}
\item Runway 2 was open for five days. \label{dura:1}
\end{examps}

\paragraph{With states and activities:}

When \qit{for~\dots} adverbials combine with states or activities, one
reading is that there must be a period with the duration of the
\qit{for~\dots} adverbial, such that the situation of the state or
activity holds throughout that period. According to this reading, in
\pref{dura:3} there must be a five-year
period, throughout which the person was a manager, and in
\pref{dura:3} a twenty-minute period throughout which
the flight was circling. If J.Adams was a manager for six
consecutive years (e.g.\ 1981 -- 1986), he would
be included in the answer to \pref{dura:3}, because there is a five-year
period (e.g.\ 1981 -- 1985) throughout which he was a manager.
\begin{examps}
\item Who was a manager for five years? \label{dura:3}
\item Did BA737 circle for twenty minutes? \label{dura:4}
\end{examps}
In some cases, however, \qit{for~\dots} adverbials are used with a stricter
meaning: they specify the duration of a \emph{maximal}
period where a situation held. In that case, if J.Adams 
started to be a manager at the beginning of 1981 and stopped being a
manager at the end of 1986 (six consecutive years), he would
\emph{not} be included in the answer to \pref{dura:3}. For simplicity,
this stricter reading is ignored in this thesis . 

In other cases, a \qit{for~\dots} adverbial does not necessarily
specify the duration of a \emph{single} period, but a \emph{total
duration}. According to this reading, if J.Adams was a manager during
several non-overlapping periods, and the total duration of these
periods is five years, he would be included in the answer to
\pref{dura:3}, even if he was never a manager for a continuous
five-year period.  This reading of \qit{for} adverbials is also not
supported in this thesis.

There is a problem if \qit{for~\dots} adverbials are allowed to
combine with consequent states (section \ref{point_adverbials}). This
problem will be discussed in section \ref{duration_adverbials}, once
some formal apparatus has been established. For the moment, I note
that the solution involves disallowing \qit{for~\dots} adverbials to
be used with consequent states.

\paragraph{With points:}
\qit{For~\dots} adverbials sometimes specify the duration of a
situation that \emph{follows} the situation of the verb. This is
particularly common when \qit{for~\dots} adverbials combine with point
expressions. For instance, \pref{dura:11} (based on an example from
\cite{Hwang1994}), probably does not mean that J.Adams was actually
leaving his office for fifteen minutes. It means that he stayed (or
intended to stay) out of his office for fifteen minutes. (I assume
here that \qit{to leave} is a point verb, as in the airport domain.)
This use of \qit{for~\dots} adverbials is not supported in this thesis.
\begin{examps}
\item J.Adams left his office for fifteen minutes. \label{dura:11}
\end{examps}

\qit{For~\dots} adverbials also give rise to iterative readings (section 
\ref{progressives}). This is again particularly common with point
expressions. \pref{dura:12} (from \cite{Hwang1994})
probably means that Mary won several times (\qit{to win} is typically
classified as point verb). Such iterative uses of \qit{for~\dots}
adverbials are not supported in this thesis. 
\begin{examps}
\item Mary won the competition for four years. \label{dura:12}  
\end{examps}
Excluding iterative readings and readings where \qit{for~\dots}
adverbials refer to consequent situations (both
are not supported in this thesis), sentences where 
\qit{for~\dots} adverbials combine with point expressions either
sound odd or signal that the user is unaware that the situation
of the point expression is modelled as instantaneous (an explanatory
message to the user is needed in the latter case; this thesis,
however, provides no mechanism to generate such messages). Hence,
for the purposes of this thesis it seems reasonable not to allow
\qit{for~\dots} adverbials to combine with point expressions.

\paragraph{With culminating activities:}
When \qit{for~\dots} adverbials combine with culminating activities,
the resulting sentences sometimes sound odd or even unacceptable. For
example, \pref{dura:38} (based on an example from 
\cite{Moens}) sounds odd or unacceptable to most native English speakers
(\qit{to build} is typically classified as culminating
activity verb). In contrast, \pref{dura:37} where the adverbial combines
with a (progressive) state is easily acceptable. 
\begin{examps}
\item \odd Housecorp built a shopping centre for two years. \label{dura:38}
\item Housecorp was building a shopping centre for two years. \label{dura:37}
\end{examps}
Based on similar examples, Vendler (section \ref{asp_taxes})
concludes that accomplishments (culminating activities) do not combine
with \qit{for~\dots} adverbials. This, however, seems
over-restrictive. \pref{dura:40} and \pref{dura:42}, for example, seem
acceptable. 
\begin{examps}
\item BA737 taxied to gate 2 for two minutes. \label{dura:40}
\item Did J.Adams inspect BA737 for ten minutes? \label{dura:42}
\end{examps}
Unlike \pref{dura:43}, in \pref{dura:40}
there is no requirement that the taxiing must have been completed,
i.e.\ that BA737 must have reached the gate. Similar comments can be
made for \pref{dura:42} and \pref{dura:45}. \qit{For~\dots} adverbials seem to
cancel any requirement that the climax must have been
reached. (Similar observations are made in \cite{Dowty1986},
\cite{Moens2}, and \cite{Kent}.)
\begin{examps}
\item BA737 taxied to gate 2. \label{dura:43}
\item Did J.Adams inspect BA737?  \label{dura:45}
\end{examps}
In the framework of this thesis, I allow \qit{for~\dots} adverbials to 
combine with culminating activities, with the same meaning that 
I adopted in the case of states and activities, and with the
proviso that any requirement that the climax
must have been reached should be cancelled. That is, in \pref{dura:42}
there must be a ten-minute period throughout which J.Adams 
was inspecting BA737. 

Table \ref{for_adverbials_table} summarises the main points of this
section.

\begin{table}
\begin{center}
{\small 
\begin{tabular}{|l|l|}
\hline
\multicolumn{2}{|c|}{meanings of duration \qit{for~\dots} adverbials} \\
\hline \hline
with lexical or progressive state & situation holds continuously for at least 
                     that long\\
\hline
with consequent state & (not allowed in the framework of this thesis) \\
\hline
with activity & situation holds continuously for at least that long \\
\hline 
with culminating activity & situation holds continuously for at
                       least that long \\
                     & (no need for climax to be reached)\\ 
\hline
with point & (not allowed in the framework of this thesis) \\
\hline
\end{tabular}
}
\end{center}
\caption{Duration \qit{for~\dots} adverbials in the framework of this thesis}
\label{for_adverbials_table}
\end{table}

\subsection{Duration \qit{in~\dots} adverbials} \label{in_adverbials}

This section discusses \qit{in~\dots} adverbials that specify 
durations (e.g.\ \pref{inad:1}, \pref{inad:2}). \qit{In} can also
introduce period adverbials (e.g.\ \qit{in 1995}; see section 
section \ref{period_adverbials}). 
\begin{examps}
\item Airserve serviced BA737 in two hours. \label{inad:1}
\item Which flight did J.Adams inspect in one hour? \label{inad:2}
\end{examps}

\paragraph{With culminating activities:} 

With culminating activity expressions, \qit{in~\dots} adverbials
usually specify the length of a period that ends at the time-point
where the situation of the culminating activity expression is
completed. In \pref{inad:1}, for example, two hours is probably the
length of a period that ends at the time-point where the service was
completed. \pref{inad:2} is similar. The period whose length is
specified by the \qit{in~\dots} adverbial usually starts at the
time-point where the situation of the culminating activity expression
begins. In \pref{inad:1}, for example, the two hours probably start at
the time-point where the service began.  The period of the adverbial,
however, may sometimes not start at the beginning of the situation of
the culminating activity expression, but at some other earlier
time. In \pref{inad:1}, the start of the two hours could be the
time-point where Airserve was asked to service BA737, not the
beginning of the actual service.  The framework of this thesis
supports only the case where the period of the adverbial starts at the
beginning of the situation described by the culminating activity expression.

\paragraph{With points:}

With point expressions, the period of the \qit{in~\dots} adverbial
starts before the (instantaneous) situation of the point expression,
and ends at the time-point where the situation of the point expression
occurs. In \pref{inad:10} the ten minutes end at the point where BA737
arrived at gate 2, and start at some earlier time-point (e.g.\ when
BA737 started to taxi to gate 2). \pref{inad:11} is similar.
\begin{examps}
\item BA737 reached gate 2 in ten minutes. \label{inad:10}
\item BA737 entered sector 2 in five minutes. \label{inad:11}
\end{examps}
Determining exactly when the period of the adverbial starts is often
difficult. It is not clear, for example, when the five minutes of
\pref{inad:11} start. As a simplification, I do not allow duration
\qit{in~\dots} adverbials to combine with point expressions.

\paragraph{With states and activities:}
\qit{In~\dots} adverbials are sometimes used with activity
expressions, with the \qit{in~\dots} duration adverbial intended to
specify the duration of the situation described by the activity
expression. Typically, in these cases the speaker has a culminating
activity view in mind. For example, \pref{inad:17} can be used in this
way if the speaker has a particular destination (say gate 2) in
mind. In that case, \pref{inad:17} can be thought as an elliptical
form of \pref{inad:19}. The framework of this thesis does not support
this use of \pref{inad:17}.
\begin{examps}
\item BA737 taxied in ten minutes. \label{inad:17}
\item BA737 taxied to gate 2 in ten minutes. \label{inad:19}  
\end{examps}

With state and activity expressions, \qit{in~\dots} adverbials can also
specify the duration of a period that ends at the beginning of the
situation of the state or activity expression. In \pref{inad:5}, for
example, the two hours probably end at the time-point where tank 5
started to be empty. The beginning of the two hours could be, for
example, a time-point where a pump started to empty the tank, or a time-point
where a decision to empty the tank was taken. Similar
comments apply to \pref{inad:17}. 
\begin{examps}
\item Tank 5 was empty in two hours. \label{inad:5}  
\end{examps}
As with point expressions, determining exactly when the period of the
adverbial starts is often difficult. As a simplification, I do not
allow duration \qit{in~\dots} adverbials to combine with state or
activity expressions.

Table \ref{in_adverbials_table} summarises the main points of this section.

\begin{table}
\begin{center}
{\small
\begin{tabular}{|l|l|}
\hline
\multicolumn{2}{|c|}{meanings of duration ``in \dots'' adverbials} \\
\hline \hline
with state, activity, or point & (not allowed in the framework of this
   thesis) \\ 
\hline 
with culminating activity & distance from the start to the completion of
                       the situation \\
\hline
\end{tabular}
}
\end{center}
\caption{Duration \qit{in~\dots} adverbials in the framework of this thesis}
\label{in_adverbials_table}
\end{table}

\subsection{Other temporal adverbials} \label{other_adverbials}

Other temporal adverbials, that are not supported by the framework of
this thesis, include some adverbials that specify boundaries (e.g.\
\qit{until 1/5/95}, \qit{since 1987}, \qit{by Monday}), frequency
adverbials (\qit{always}, \qit{twice}, \qit{every Monday}), and
adverbials of temporal order (\qit{for the second time},
\qit{earlier}). 


\section{Temporal subordinate clauses} \label{subordinate_clauses}

Three kinds of temporal subordinate clauses are examined in this
thesis: clauses introduced by \qit{while}, \qit{before}, and
\qit{after} (e.g.\ clauses introduced by \qit{since}, \qit{until}, or
\qit{when} are not examined). From the temporal subordinate clauses
that are not examined, \qit{when~\dots} clauses are generally
considered the most difficult to support (see \cite{Ritchie},
\cite{Yip1985}, \cite{Hinrichs1986}, \cite{Moens}, 
\cite{Moens2}, and \cite{Lascarides1993} for explorations of
\qit{when~\dots} clauses). 

\subsection{\qit{While~\dots} clauses} \label{while_clauses}

\paragraph{Subordinate clause:}
As with period adverbials (section \ref{period_adverbials}), each
\qit{while~\dots} clause is understood as specifying a time period.
This is a maximal period throughout which the situation of the
\qit{while~\dots} clause holds. Let us assume, for example, that
J.Adams was a manager only from 1/1/1980 to 31/12/1983, and from
1/1/1987 to 31/12/1990. Then, in \pref{whc:1} the period of the
\qit{while~\dots} clause can be either one of these two periods. The
user may have in mind a particular one of the two periods. In that
case, a temporal anaphora resolution mechanism is needed to determine
that period (temporal anaphora is discussed in section
\ref{temporal_anaphora}). The framework of this thesis, however,
provides no such mechanism (the answer to \pref{whc:1} includes
anybody who was fired during any of the two periods).
\begin{examps}
\item Who was fired while J.Adams was a manager? \label{whc:1}  
\end{examps}

Sentences where the aspectual class of the \qit{while~\dots} clause is
point (e.g.\ \pref{whc:4} in the airport domain) typically signal
that the user is unaware that the situation of the \qit{while~\dots}
clause is modelled as instantaneous. In the framework of
this thesis, the answer to \pref{whc:4} includes any flight that was
circling at the time-point where BA737 entered sector 2. Ideally, a
message would also be generated to warn the user that entering a
sector is modelled as instantaneous (no warning is currently
generated). This is another case where cooperative responses (section
\ref{no_issues}) are needed. 
\begin{examps}
\item Which flights were circling while BA737 entered sector 2? \label{whc:4}  
\end{examps}
Sentences containing \qit{while~\dots} clauses whose aspectual class
is consequent state (section \ref{point_adverbials}) usually sound
unnatural or unacceptable. For example, \pref{whc:5.2} --
\pref{whc:5.8} sound at least unnatural (e.g.\ instead of
\pref{whc:5.2} one would normally use \pref{whc:5.10} or
\pref{whc:5.11}). Hence, I do not allow
\qit{while~\dots} clauses whose aspectual class is consequent
state. This also avoids some complications in the English to \topl mapping. 
\begin{examps}
\item \odd Did any flight depart while BA737 had landed?
  \label{whc:5.2}
\item \odd Did ABM fire anybody while J.Adams had been
  the manager? \label{whc:5.5} 
\item \odd Had any flight departed while J.Adams had inspected BA737?
  \label{whc:5.8} 
\item Did any flight depart while BA737 was landing? \label{whc:5.10}
\item Did any flight depart after BA737 had landed? \label{whc:5.11}
  \label{whc:5.12} 
\end{examps}

When the aspectual class of the \qit{while~\dots} clause is
culminating activity, there is no requirement that the climax
of the situation of the \qit{while~\dots} clause must
have been reached, even if the tense of that clause
normally requires this. In \pref{whc:8} and
\pref{whc:6}, for example, there does not seem to be any requirement that the
service or the boarding must have been completed (cf.\ \pref{whc:10}
and \pref{whc:11}). \pref{whc:8} and \pref{whc:6} appear to have the
same meanings as \pref{whc:9} and \pref{whc:7} (in progressive tenses,
there is no requirement for the climax to be reached; see section 
\ref{progressives}). 

Table \ref{while_clauses_table}
summarises the main points about \qit{while~\dots} clauses so far.
\begin{examps}
\item Did Airserve service BA737? \label{whc:10}
\item Which flights departed while Airserve serviced BA737? \label{whc:8}
\item Which flights departed while Airserve was servicing BA737? \label{whc:9}
\item Did BA737 board? \label{whc:11}
\item Which flights departed while BA737 boarded? \label{whc:6}
\item Which flights departed while BA737 was boarding? \label{whc:7}
\end{examps}

\begin{table}
\begin{center}
{\small
\begin{tabular}{|l|l|}
\hline
aspectual class of        & \\
\qit{while~\dots} clause  & period specified by \qit{while~\dots} clause \\
\hline \hline
consequent state & (not allowed in the framework of this thesis) \\
\hline
lexical/progressive state & maximal period where situation of
               \qit{while~\dots} clause holds \\
or activity & \\
\hline 
culminating activity  &  maximal period where situation of
                    \qit{while~\dots} clause holds \\
                 & (no need for climax of \qit{while~\dots} clause
                   to be reached)  \\
\hline
point & instant.\ period where situation of \qit{where~\dots}
   clause occurs \\
\hline 
\end{tabular}
}
\end{center}
\caption{Periods of \qit{while~\dots} clauses in the framework of this thesis}
\label{while_clauses_table}
\end{table}

\paragraph{Main clause:}
Once the periods of the \qit{while~\dots} clauses have been 
established (following table \ref{while_clauses_table}), the behaviour
of \qit{while~\dots} clauses appears to be the same as that of period
adverbials (i.e.\ it follows table \vref{period_adverbials_table}).
With main clauses whose aspectual class is point, the
instantaneous situation of the main clause must occur within
the period of the \qit{while~\dots} clause (e.g.\ in \pref{whc:30} the 
departures must have occurred during a maximal period where runway 5
was closed; \qit{to depart} is a point verb in the airport domain).
\begin{examps}
\item Did any flight depart from gate 2 while runway 5 was closed?
  \label{whc:30} 
\end{examps}
With activity main clauses, the situation of the main clause must be
ongoing some time during the period of the \qit{while~\dots} clause. In
\pref{whc:33}, for example, the flights must have taxied
some time during a maximal period where BA737 was circling. As with
period adverbials, stricter readings are sometimes
possible with activity main clauses. \pref{whc:33}, for example, could
refer to flights that both started and stopped taxiing during a
maximal period where BA737 was circling. As with period adverbials, I
ignore such stricter readings.
\begin{examps}
\item Which flights taxied while BA737 circled? \label{whc:33}  
\end{examps}
As in the case of period adverbials, with culminating activity main
clauses I allow two readings: (a) that the situation of the main
clause both starts and reaches its completion within the period of the 
\qit{while~\dots} clause, or (b) that the situation of the main clause
simply reaches its completion within the period of the \qit{while~\dots}
clause. In the second reading, the main clause is taken
to refer to only the completion point of the situation it would
normally describe, and its aspectual is changed to 
point. In the airport domain, the first reading is the preferred one in
\pref{whc:34}. The second reading allows the
answer to \pref{whc:35} to contain flights that simply
touched down during the service, even if their
landing procedures did not start during the service. 
\begin{examps}
\item J.Adams inspected BA737 while Airserve was servicing
  UK160. \label{whc:34}
\item Which flights landed while Airserve was servicing UK160?
  \label{whc:35}   
\end{examps}
With state main clauses, I require the situation of the main clause to
hold some time during the period of the \qit{while~\dots} clause
(inclusive reading; see section \ref{period_adverbials}). For
example, the answer to \pref{whc:20} must contain anybody who was a
lecturer some time during a maximal period where J.Adams was a
professor (the non-auxiliary \qit{to be} is typically classified as
state verb). As with period adverbials, there is often an
implication that the situation of the main clause holds
\emph{throughout} the period of the 
\qit{while~\dots} clause (durative reading). The durative reading is
unlikely in \pref{whc:20}, but seems the preferred one in \pref{whc:21}
(progressive state main clause). According to the durative reading,
\pref{whc:21} refers to a flight that was circling \emph{throughout} a
maximal period where runway 2 was closed.
\begin{examps}
\item Who was a lecturer while J.Adams was a professor? \label{whc:20} 
\item Which flight was circling while runway 2 was
  closed? \label{whc:21} 
\end{examps}

The treatment of \qit{while~\dots} clauses of this thesis is
similar to that of \cite{Ritchie}. Ritchie also views
\qit{while~\dots} clauses as establishing periods,
with the exact relations between these periods and the situations of
the main clauses depending on the aspectual classes of the main
clauses.  Ritchie uses only two aspectual classes (``continuing'' and
``completed''), which makes presenting a direct comparison between his
treatment of \qit{while~\dots} clauses and the treatment of this
thesis difficult. Both approaches, however, lead to similar results,
with the following two main exceptions. (a) In \pref{whc:20} and
\pref{whc:21} (state main clause), Ritchie's treatment admits only durative
readings. In contrast, the framework of this thesis admits only
inclusive ones. (b) In \pref{whc:35} (culminating activity main
clause), Ritchie's arrangements allow only one reading, where the
landings must have both started and been completed during the
service. The framework of this thesis allows an additional reading,
whereby it is enough if the landings were simply completed during the
service.

\subsection{\qit{Before~\dots} and \qit{after~\dots} clauses}
\label{before_after_clauses}

I treat \qit{before~\dots} and \qit{after~\dots} clauses as
establishing periods, as in the case of the \qit{before~\dots} and
\qit{after~\dots} adverbials of section \ref{period_adverbials}. In
\qit{before~\dots} clauses, the period starts at some
unspecified time-point (in the absence of other constraints, the
beginning of time), and ends at a time-point provided by the
\qit{before~\dots} clause. In \qit{after~\dots} clauses, the period
starts at a time-point provided by the \qit{after~\dots} clause, and
ends at some unspecified time-point (the end of time, in the absence
of other constraints). I use the terms \emph{before-point} and
\emph{after-point} to refer to the time-points provided by
\qit{before~\dots} and \qit{after~\dots} clauses respectively. Once
the periods of the \qit{before~\dots} and \qit{after~\dots} clauses
have been established, the behaviour of the clauses appears to be the
same as that of period adverbials (i.e.\ it follows table
\vref{period_adverbials_table}).

\paragraph{State \qit{before/after~\dots} clause:}
Let us first examine sentences where the aspectual class of the
\qit{before~\dots} or \qit{after~\dots} clause is state. With 
\qit{before~\dots} clauses, the before-point is a time-point where
the situation of the \qit{before~\dots} clause starts (table
\ref{before_clauses_table}). In
\pref{bac:1}, for example, the before-point is a time-point where
runway 2 started to be open. The aspectual class of the main clause is
point (\qit{to depart} is a point verb in the airport domain). Hence,
according to table \vref{period_adverbials_table}, the departures must
have occurred within the period of the \qit{before~\dots} clause, i.e.\
before the time-point where runway 2 started to be open. Similar
comments apply to \pref{bac:1.1}, \pref{bac:2} (progressive 
\qit{before~\dots} clause), and \pref{bac:3} (consequent state
\qit{before~\dots} clause). In \pref{bac:3}, the 
before-point is the beginning of the consequent period of the
inspection (the period that contains all the time after the completion of
the inspection; see section \ref{point_adverbials}), i.e.\ the
departures must have happened before the inspection was completed.
\begin{examps}
\item Which flights departed before runway 2 was open? \label{bac:1}
\item Which flights departed before the emergency system was in
  operation? \label{bac:1.1}
\item Which flights departed before BA737 was circling? \label{bac:2}
\item Which flights departed before J.Adams had inspected BA737? \label{bac:3}
\end{examps}

\begin{table}[t]
\begin{center}
{\small 
\begin{tabular}{|l|l|}
\hline aspectual class of & before-point \\ 
\qit{before~\dots} clause & (right boundary of period specified by
                            \qit{before~\dots} clause) \\  
\hline \hline 
state & time-point where situation of \qit{before~\dots} clause starts \\
\hline activity & time-point where situation of \qit{before~\dots} 
                  clause starts \\ 
\hline culm.\ activity & time-point where situation of 
                         \qit{before~\dots} clause \\ 
                       & starts or is completed \\ 
\hline point & time-point where situation of \qit{before~\dots} 
               clause occurs \\ 
\hline
\end{tabular}
}
\end{center}
\caption{Boundaries of \qit{before~\dots} clauses in the
   framework of this thesis}
\label{before_clauses_table}
\end{table}

According to table \vref{period_adverbials_table}, in \pref{bac:12}
where the main clause is a state, the flight must have been at gate 2
some time during the period of the \qit{before~\dots} clause,
i.e.\ for some time before runway 2 started to be open. In
\pref{bac:10} (activity main clause), the flight must
have circled for some time before runway 2 started to be open, and in
\pref{bac:11} (culminating activity main clause) the inspections must
have both started and been completed before runway 2 started to be
open. (As with the \qit{before~\dots} adverbials of section
\ref{period_adverbials}, in \pref{bac:11} it would be better if the
\nlitdb could also report inspections that started but were not
completed before runway 2 opened, warning the user that these
inspections were not completed before runway 2 opened.)
\begin{examps}
\item Was any flight at gate 2 before runway 2 was open? \label{bac:12}
\item Did any flight circle before runway 2 was open? \label{bac:10}
\item Which flights did J.Adams inspect before runway 2 was open?
  \label{bac:11}   
\end{examps}
In the case of \qit{after~\dots} clauses, when the aspectual class of
the \qit{after~\dots} clause is state, the after-point is a time-point
where the situation of the \qit{after~\dots} clause either starts or
ends. \pref{bac:1a}, for example, has two readings: that the
flights must have departed after runway 2 \emph{started} to be 
open, or that the flights must have departed after
runway 2 \emph{stopped} being open. Similar comments apply to 
\pref{bac:1.1a} and \pref{bac:2a}.
\begin{examps}
\item Which flights departed after runway 2 was open? \label{bac:1a}
\item Which flights departed after the emergency system was in
  operation? \label{bac:1.1a}
\item Which flights departed after BA737 was circling? \label{bac:2a}
\end{examps}
In sentences like \pref{bac:3a}, where the aspectual class of the
\qit{after~\dots} clause is consequent state, the after-point can only
be the beginning of the consequent period (the first time-point after
the completion of the inspection). It cannot be the end of the
consequent period: the end of the consequent period is the end of
time; if the after-point were the end of the consequent period, the
departures of \pref{bac:3a} would have to occur after the end of time,
which is impossible. This explains the distinction between
lexical/progressive and consequent states in table
\ref{after_clauses_table}. 
\begin{examps}
\item Which flights departed after J.Adams had inspected BA737? \label{bac:3a}
\end{examps}

\begin{table}[t]
\begin{center}
{\small
\begin{tabular}{|l|l|}
\hline
aspectual class of         & after-point \\
\qit{before~\dots} clause  & (left boundary of period specified by
                             \qit{after~\dots} clause) \\ 
\hline \hline
lexical/progressive state & time-point where situation of
                            \qit{after~\dots} clause 
                            starts or ends \\
\hline 
consequent state & time-point where consequent period of
                   \qit{after~\dots} clause starts \\
\hline 
activity     & time-point where situation of \qit{after~\dots} clause
                    ends \\
\hline 
culm.\ activity  & time-point where situation of \qit{after~\dots}
                   clause is completed \\
\hline
point & time-point where situation of \qit{before~\dots} clause occurs \\
\hline 
\end{tabular}
}
\end{center}
\caption{Boundaries of \qit{after~\dots} clauses in the framework of this thesis}
\label{after_clauses_table}
\end{table}

\paragraph{Point \qit{before/after~\dots} clause:}
If the aspectual class of the \qit{before~\dots} or \qit{after~\dots}
clause is point, the before/after-point is the
time-point where the instantaneous situation of the subordinate
clause occurs. In \pref{bac:20}, for example, the before/after-point is
the point where BA737 reached gate2. 
\begin{examps}
\item Which flights departed before/after BA737 reached gate 2?
   \label{bac:20}   
\end{examps}

\paragraph{Activity \qit{before/after~\dots} clause:}
With activity \qit{before/after~\dots} clauses, I
consider the before-point to be a time-point where the situation of
the \qit{before~\dots} clause starts, and the after-point to be a
point where the situation of the \qit{after~\dots} clause
ends. In \pref{bac:23} and \pref{bac:24}, for example, the departures
must have occurred before BA737 \emph{started} to taxi or circle. In
\pref{bac:25} and \pref{bac:26}, the departures must have occurred
after BA737 \emph{stopped} taxiing or circling.
\begin{examps}
\item Which flights departed before BA737 taxied? \label{bac:23}
\item Which flights departed before BA737 circled? \label{bac:24}
\item Which flights departed after BA737 taxied? \label{bac:25}
\item Which flights departed after BA737 circled? \label{bac:26}
\end{examps}
Perhaps another reading is sometimes possible with
\qit{after~\dots} clauses: that the after-point is a time-point where
the situation of the \qit{after~\dots} clause \emph{starts} (e.g.\ 
\pref{bac:26} would refer to departures that occurred
after BA737 \emph{started} to circle). This reading, however, does not seem
very likely, and for simplicity I ignore it. 

\paragraph{Culminating activity \qit{before/after~\dots} clause:} With
\qit{after~\dots} clauses whose aspectual class is culminating
activity, I consider the after-point to be a time-point where the 
situation of the \qit{after~\dots} clause reaches its completion. In
\pref{bac:27}, the departures must have occurred after
the completion of the inspection, and in \pref{bac:28} they must have
occurred after the time-point where BA737 reached gate 2.
\begin{examps}
\item Which flights departed after J.Adams inspected BA737?
  \label{bac:27}
\item Which flights departed after BA737 taxied to gate 2?
  \label{bac:28}  
\end{examps}
With culminating activity \qit{before~\dots} clauses, I allow the
before-point to be a time-point where the situation of the
\qit{before~\dots} clause either starts or reaches its completion. In
the airport domain, the
first reading seems the preferred one in \pref{bac:30} (the flights
must have departed before the \emph{beginning} of the inspection). The
second reading seems the preferred one in \pref{bac:31} (the flights
must have departed before the \emph{completion} of the landing). Both
readings seems possible in \pref{bac:31}. 
\begin{examps}
\item Which flights departed before J.Adams inspected BA737?
  \label{bac:30}
\item Which flights departed before BA737 landed? \label{bac:33}
\item Which flights departed before BA737 taxied to gate 2?
  \label{bac:31}
\end{examps}
If the first reading is adopted (the situation of the
\qit{before~\dots} clause \emph{starts} at the before-point) and the
\qit{before~\dots} clause is in the simple past, it is unclear if
the situation of the \qit{before~\dots} clause must have necessarily
reached its climax (the simple past of culminating
activity verbs normally requires this;
see section \ref{simple_past}). For example, let us assume that the
first reading is adopted in \pref{bac:30}. Should the before-point be
the beginning of an inspection that was necessarily completed, or can
it also be the beginning of an inspection that was never completed?
The framework of this thesis currently adopts the first approach, but
this is perhaps over-restrictive. It would probably be better if the
\nlitdb allowed the before-point to be the beginning of both inspections
that were and were not completed, warning the user
about inspections that were not completed. This is another case for 
cooperative responses (section \ref{no_issues}). 

\paragraph{Other uses:}
\qit{Before} and \qit{after} can be preceded by expressions specifying
durations (e.g.\ \pref{bac:40}). This use of \qit{before} and
\qit{after} is not considered in this thesis.
\begin{examps}
\item BA737 reached gate 2 five minutes after UK160 departed. \label{bac:40}
\end{examps}
\qit{Before~\dots} clauses also have counter-factual uses. For
example, in \pref{bac:43} (from \cite{Crouch}) the situation where the
car runs into the tree never takes place. This use of \qit{before} is
not considered in this thesis. 
\begin{examps}
\item Smith stopped the car before it ran into the tree. \label{bac:43}
\end{examps}

The treatment of \qit{before~\dots} and \qit{after~\dots} clauses of
this thesis is similar to that of \cite{Ritchie}. Ritchie also views
\qit{before~\dots} and \qit{after~\dots} clauses as providing before
and after-points. As noted in section \ref{while_clauses}, however,
Ritchie uses only two aspectual classes.  According to
Ritchie, in the case of \qit{before~\dots} clauses, the before-point
is a time-point where the situation of the \qit{before~\dots}
clause starts, and the situation of the main clause must simply start
before that point. In \pref{bac:11},
this requires the inspections to have simply
\emph{started} before the time-point where runway 2 started to be
open. In contrast, the framework of this thesis requires the
inspections to have been \emph{completed} before that time-point.

In the case of \qit{after~\dots} clauses, the main difference between
Ritchie's treatment and the treatment of this thesis concerns state
\qit{after~\dots} clauses. In that case, Ritchie allows the
after-point to be only the beginning of the situation of the
\qit{after~\dots} clause. In \pref{bac:1.1a}, this requires the
flights to have departed after the time-point where the system \emph{started}
to be in operation. The framework of this thesis
allows an additional reading, where the flights must have departed after
the time-point where the system \emph{stopped} being in operation.

\subsection{Tense coordination} \label{tense_coordination}

Some combinations of tenses in the main and subordinate clauses are
unacceptable (e.g.\ \pref{coo:0}, \pref{coo:2}). 
This thesis makes no attempt to account for the unacceptability of
such combinations. The reader is referred
to \cite{Harper} and \cite{Brent1990} for methods that could be used
to detect and reject sentences like \pref{coo:0} and \pref{coo:2}. 
\begin{examps}
\item \bad BA737 left gate 2 before runway 2 is free. \label{coo:0}
\item \bad Which runways are closed while runway 2 was circling? \label{coo:2}
\end{examps}


\section{Noun phrases and temporal reference} \label{noun_anaphora}

A question like \pref{nana:1} can refer either to the present sales
manager (asking the 1991 salary of the present sales manager) or to
the 1991 sales manager (asking the 1991 salary of the 1991 sales
manager). Similarly, \pref{nana:2} may refer either to present
students or last year's students. In \pref{nana:3.1}, \qit{which
closed runway} probably refers to a runway that is \emph{currently}
closed, while in \pref{nana:3.5} \qit{a closed runway} probably refers
to a runway that was closed at the time of the landing. 
\begin{examps}
\item What was the salary of the sales manager in 1991? \label{nana:1} 
\item Which students failed in physics last year? \label{nana:2} 
\item Which closed runway was open yesterday? \label{nana:3.1}
\item Did BA737 ever land on a closed runway in 1991? \label{nana:3.5} 
\end{examps}
It seems that noun phrases (e.g.\ \qit{the sales manager}, \qit{which
students}, \qit{a closed runway}) generally refer either to the
present or to the time of the verb tense (if this time is different
than the present). In \pref{nana:1}, the simple past tense refers to
some time in 1991. Therefore, there are two options:
\qit{the sales manager} can refer either to the present sales manager
or to somebody who was the sales manager in 1991. Similar comments
apply to \pref{nana:2}. In contrast, in \pref{nana:3} the
verb tense refers to the present. Hence, there is only one
possibility: \qit{the sales manager} refers to the present sales manager.
\begin{examps}
\item What is the salary of the sales manager? \label{nana:3}
\end{examps}
In \pref{nana:3.1}, the verb tense refers to a time (within the
previous day) where the runway was open. There should be two readings:
it should be possible for \qit{which closed runway} to refer either to
a currently closed runway, or to a runway that was closed at the time
it was open. Since a runway cannot be closed at the same time where it
is open, the second reading is ruled out. (This clash, however, cannot
be spotted easily by a \nlitdb without some inferential capability.)

The hypothesis that noun phrases refer either to the present or to the
time of the verb tense is not always adequate. For example, a person
submitting \pref{nana:5} to the \nlitdb of a university most probably
refers to \emph{previous} students of the university. In contrast, the
hypothesis predicts that the question can refer only to \emph{current}
students. (Similar examples can be found in \cite{Enc1986}.)
\begin{examps}
\item How many of our students are now professors? \label{nana:5}
\end{examps}
The hypothesis also predicts that \pref{nana:6} can refer only to
current Prime Ministers or to persons that were Prime Ministers at the
time they were born (an extremely unlikely reading). There is, however, a reading where the question refers to all past and present Prime
Ministers. This reading is incorrectly ruled out by the hypothesis. 
\begin{examps}
\item Which Prime Ministers were born in Scotland? \label{nana:6}
\end{examps}

Hinrichs \cite{Hinrichs} argues that determining the
times to which noun phrases refer is part of a more general
problem of determining the entities to which noun phrases refer.
According to Hinrichs, a noun phrase like \qit{every
admiral} generally refers to anybody who was, is, or will be an admiral
of any fleet in the world at any time. If, however, \pref{nana:8}
is uttered in a context where the current personnel of the U.S.\
Pacific fleet is being discussed, the temporal scope of \qit{every
admiral} is restricted to current admirals, in the same way that the
scope of \qit{every admiral} is restricted to admirals of the U.S.\
Pacific fleet (e.g.\ \pref{nana:8} does not mean that all Russian
admirals also graduated from Annapolis).
\begin{examps}
\item Every admiral graduated from Annapolis. \label{nana:8}
\end{examps}
The fact that Hinrichs does not limit the times of the noun phrases to
the present and the time of the verb tense is in accordance with the
fact that \qit{our students} in \pref{nana:5} is not limited to
present students, and the fact that \qit{which Prime
Ministers} in \pref{nana:6} may refer to all past and present Prime
Ministers. Hinrichs' approach,
however, requires some mechanism to restrict the scope of noun phrases
as the discourse evolves. Hinrichs offers only a very limited sketch of
how such a mechanism could be constructed. Also, in the absence of
previous discourse, Hinrichs' treatment suggests that
\pref{nana:1} refers to the sales managers of all times, an unlikely
interpretation. The hypothesis that noun phrases refer either to the
present or to the time of the verb tense performs better in this case.
Given these deficiencies of Hinrichs' approach, I adopt the initial
hypothesis that noun phrases refer to the present or the time of the
verb tense. (An alternative approach would be to attempt to merge this
hypothesis with Hinrichs' method. \cite{Dalrymple1988} goes towards
this direction.)

A further improvement can be made to the hypothesis that noun phrases
refer to the present or the time of the verb tense. When a noun phrase
is the complement of the predicative \qit{to be}, it seems that the
noun phrase can refer only to the time of the verb
tense. \pref{nana:11}, for example, can only be a request to report
the 1991 sales manager, not the current sales manager. Similarly,
\pref{nana:11.5} cannot mean that J.Adams is the current sales
manager. This also accounts for the fact that in \pref{nana:1}, unlike
\qit{the sales manager} which can refer either to the present or 
1991, \qit{the salary of the sales manager} (the complement of
\qit{was}) can refer only to a 1991 salary, not to a present
salary. (I assume that the restriction that the complement of the
predicative \qit{to be} must refer to the time of the verb tense does
not extend to noun phrases that are subconstituents of that
complement, like \qit{the sales manager} in \pref{nana:1}.) The same
restriction applies to bare adjectives used as complements of the
predicative \qit{to be}. In \pref{nana:3.1}, for example, \qit{open} can only
refer to runways that were open on the previous day. It cannot refer
to currently open runways.
\begin{examps}
\item Who was the sales manager in 1991? \label{nana:11}
\item J.Adams was the sales manager in 1991. \label{nana:11.5}
\end{examps}

The hypothesis that noun phrases refer to the present or the time of
the verb tense does not apply when a temporal adjective (e.g.\
\qit{current}) specifies explicitly the time of the noun phrase (e.g.\
\pref{nana:9}). (Although temporal adjectives are not considered in
this thesis, I support \qit{current} to be able to illustrate this point.)
\begin{examps}
\item Which current students failed in Physics last year?
   \label{nana:9} 
\end{examps}

In chapter \ref{English_to_TOP}, an additional mechanism will be
introduced, that allows the person configuring the \nlitdb to force
some noun phrases to be treated as always referring to the time of the
verb tense, or as always referring to the present. 


\section{Temporal anaphora} \label{temporal_anaphora}
 
There are several English expressions (e.g.\ \qit{that time}, \qit{the
following day}, \qit{then}, \qit{later}) that refer implicitly to
contextually salient times, in a way that is similar to how pronouns,
possessive determiners, etc.\ refer to contextually salient world
entities (the terms \emph{temporal} and \emph{nominal anaphora} were
used in section \ref{no_issues} to refer to these two phenomena; the
parallels between temporal and nominal anaphora are discussed in
\cite{Partee1984}). For example, the user of a
\nlitdb may submit \pref{tan:1}, followed by \pref{tan:2}. In
\pref{tan:2}, \qit{at that time} refers to the time when John became
manager (temporal anaphora). In a similar manner, \qit{he} refers to
John (nominal anaphora). 
\begin{examps}
\item When did John become manager? \label{tan:1}
\item Was he married at that time? \label{tan:2}
\end{examps}
Names of months, days, etc.\ often have a similar temporal anaphoric
nature. For example, in a context where several questions about the
1990 status of a company have just been asked, \pref{tan:6} most
probably refers to the January of 1990, not any other January. In the
absence of previous questions, \pref{tan:6} most probably refers to
the January of the current year. (See section 5.5.1 of \cite{Kamp1993}
for related discussion.)
\begin{examps}
\item Who was the sales manager in January? \label{tan:6}
\end{examps}
Verb tenses also seem to have a temporal anaphoric nature (the term
\emph{tense anaphora} is often used in this case). For example, the
user may ask \pref{tan:7} (let us assume that the response is
``\sys{no}''), followed by \pref{tan:8}. In that case, the simple past
\qit{was} of \pref{tan:8} does not refer to an arbitrary past time, it
refers to the past time of the previous question, i.e.\ 1993.
\begin{examps}
\item Was Mary the personnel manager in 1993? \label{tan:7}
\item Who was the personnel manager? \label{tan:8}
\end{examps}
The anaphoric nature of verb tenses is clearer in multi-sentence text
(see \cite{Hinrichs1986}, \cite{Webber1988}, \cite{Kamp1993},
\cite{Kameyama1993} for related work). In \pref{tan:9}, for example,
the simple past \qit{landed} refers to a landing that happened
immediately after the permission of the first sentence was given. It
does not refer to an arbitrary past time where BA737 landed on runway
2. Similar comments apply to the \qit{taxied}.
\begin{examps}
\item BA737 was given permission to land at 5:00pm. It landed on
   runway 2, and taxied to gate 4. \label{tan:9}
\end{examps}

In dialogues like the one in \pref{tan:7} -- \pref{tan:8}, a
simplistic treatment of tense anaphora is to 
store the time of the adverbial of \pref{tan:7}, and to require the
simple past of \pref{tan:8} to refer to 
that time. (A more elaborate version of this approach will be
discussed in section \ref{lt_anaphora}.)

The behaviour of noun phrases like \qit{the sales manager} of section
\ref{noun_anaphora} can be seen as a case of temporal
anaphora. This is the only type of temporal anaphora that is supported
by the framework of this thesis. Expressions like \qit{at that time},
\qit{the following day}, etc.\ are not supported, and tenses referring
to the past are taken to refer to \emph{any} past time. For example,
\pref{tan:8} is taken to refer to anybody who was the personnel
manager at any past time. The reader is also reminded (section
\ref{no_issues}) that nominal anaphora is not considered in this
thesis.


\section{Other phenomena that are not supported} \label{ling_not_supported}

This section discusses some further phenomena that are not supported
by the framework of this thesis.

\paragraph{Cardinality and duration questions:}
Questions about the cardinality of a set or the duration of a
situation (e.g.\ \pref{oiss:1}, \pref{oiss:2}) are not
supported. (\topl is currently not powerful enough to express the
meanings of these questions.)
\begin{examps}
\item How many flights have landed today? \label{oiss:1}
\item For how long was tank 2 empty? \label{oiss:2}
\end{examps}

\paragraph{Cardinalities and plurals:}
Expressions specifying cardinalities of sets (e.g.\ \qit{eight
passengers}, \qit{two airplanes}) are not supported (this does not
include duration expressions like \qit{five hours}, which are
supported). Expressions of this kind give rise to a distinction
between \emph{distributive} and \emph{collective} readings
\cite{Stirling1985} \cite{Crouch2}. \pref{oiss:10}, for example, has a
collective reading where the eight passengers arrive at the same
time, and a distributive one where there are eight separate
arrivals. This distinction was not explored during the work of this thesis.
\topl is also currently not powerful enough to express cardinalities of
sets.
\begin{examps}
\item Eight passengers arrived. \label{oiss:10}
\end{examps}
The framework of this thesis accepts plural noun phrases
introduced by \qit{some} and \qit{which} (e.g.\ \qit{some flights},
\qit{which passengers}), but it treats them semantically as
singular. For example, \pref{oiss:11} and \pref{oiss:12} are treated
as having the same meanings as \pref{oiss:11.1} and \pref{oiss:12.1}
respectively.
\begin{examps}
\item Which flights landed? \label{oiss:11}
\item Which flight landed? \label{oiss:11.1}
\item Some flights entered sector 2. \label{oiss:12}
\item A flight entered sector 2. \label{oiss:12.1}
\end{examps}

\paragraph{Quantifiers:}
Expressions introducing universal quantifiers at the logical level
(e.g.\ \qit{every}, \qit{all}) are not supported. This leaves only
existential quantifiers (and an interrogative version of them, to be
discussed in chapter \ref{TOP_chapter}) at the logical level,
avoiding issues related to quantifier scoping (see also section
\ref{quantif_scoping}). It also simplifies the semantics of \topl and the
mapping from \topl to \tsql.

\paragraph{Conjunction, disjunction, and negation:}
Conjunctions of words or phrases are not supported. Among other
things, this avoids phenomena related to sequencing of events. For
example, \pref{oiss:15} is understood as saying that the patient died
\emph{after} (and probably as a result of) being given Qdrug (cf.\ 
\pref{oiss:16} which sounds odd). In contrast, in \pref{oiss:17} the
patient was given Qdrug \emph{while} he had high fever. (See, for
example, \cite{Hinrichs1986}, \cite{Hinrichs}, \cite{Webber1988},
\cite{Kamp1993}, \cite{terMeulen1994}, and \cite{Hwang1994} for
related work.)
\begin{examps}
\item Which patient was given Qdrug and died?  \label{oiss:15}
\item \odd Which patient died and was given Qdrug? \label{oiss:16}
\item Which patient had high fever and was given Qdrug? \label{oiss:17}
\end{examps}

Expressions introducing disjunction or negation (e.g.\ \qit{or},
\qit{either}, \qit{not}, \qit{never}) are also not supported. This
simplifies the semantics of \topl and the \topl to \tsql mapping. Not
supporting negation also avoids various temporal phenomena related
to negation (see section 5.2.5 of \cite{Kamp1993}), and claims that
negation causes aspectual shifts (see, for example, \cite{Dowty1986}
and \cite{Moens}).

\paragraph{Relative clauses:}
Relative clauses are also not supported.  Relative clauses require
special temporal treatment. \pref{oiss:20}, for example, most probably
does not refer to a runway that was closed at an \emph{arbitrary} past
time; it probably refers to a runway that was closed at the time of
the landing. The relation between the time of the relative clause and
that of the main clause can vary. In \pref{oiss:21} (from
\cite{Dowty1986}), for example, the woman may have seen John during,
before, or even after the stealing.
\begin{examps}
\item Which flight landed on a runway that was closed? \label{oiss:20}
\item The woman that stole the book saw John. \label{oiss:21}
\end{examps} 
Relative clauses can also be used with nouns that refer to the
temporal ontology (e.g.\ \qit{period} in \pref{oiss:22}). Additional
temporal phenomena involving relative clauses are discussed in section
5.5.4.2 of \cite{Kamp1993}.
\begin{examps}
\item Who was fired during the period that J.Adams was
    personnel manager?  \label{oiss:22}
\end{examps}

\paragraph{Passives:}
Finally, I have concentrated on active voice verb forms. This
simplifies the \hpsg grammar of chapter 4. It should be easy to extend
the framework of this thesis to cover passive forms as well.


\section{Summary}

The framework of this thesis uses an aspectual taxonomy of four
classes (states, points, activities, and culminating activities). This
taxonomy classifies verb forms, verb phrases, clauses, and
sentences. Whenever the \nlitdb is configured for a new application,
the base form of each verb is assigned to one of the four aspectual
classes. All other verb forms normally inherit the aspectual class of
the base form. Verb phrases, clauses, and sentences normally inherit
the aspectual classes of their main verb forms.  Some linguistic mechanisms
(e.g.\ progressive tenses, or some temporal adverbials), however, may
cause the aspectual class of a verb form to differ from that of the
base form, or the aspectual class of a verb phrase, clause, or
sentence to differ from that of its main verb form. The
aspectual taxonomy plays an important role in most time-related
linguistic phenomena.

Six tenses (simple present, simple past, present continuous, past
continuous, present perfect, and past perfect) are supported, with
various simplifications introduced in their meanings. Some
special temporal verbs were identified (e.g.\ \qit{to happen}, \qit{to
start}); from these only \qit{to start}, \qit{to begin}, \qit{to
stop}, and \qit{to finish} are supported.

Some nouns have a special temporal nature. For example,
some introduce situations (e.g.\ \qit{inspection}), others specify 
temporal order (e.g.\ \qit{predecessor}), and others refer to
entities of the temporal ontology (e.g.\ \qit{day}, \qit{period},
\qit{event}). From all these, only nouns like \qit{year}, \qit{month},
\qit{day}, etc.\ (and proper names like \qit{Monday}, \qit{January}, and
\qit{1/5/92}) are supported. Nouns referring to more abstract temporal
entities (e.g.\ \qit{period}, \qit{event}) are not supported.
No temporal adjectives (e.g.\ \qit{first}, \qit{earliest}) are
handled, with the only exception of \qit{current} which is supported
to demonstrate the anaphoric behaviour of some noun phrases. 

Among temporal adverbials, only punctual adverbials (e.g.\ \qit{at
5:00pm}), \qit{for~\dots} and \qit{in~\dots}
duration adverbials, and period adverbials introduced by \qit{on}, \qit{in},
\qit{before}, or \qit{after}, as well as \qit{today} and
\qit{yesterday} are handled. Frequency, order, or other adverbials
that specify boundaries (e.g.\ \qit{twice}, \qit{for the second time},
\qit{since 1992}) are not supported.

Only subordinate clauses introduced by \qit{while}, \qit{before}, and
\qit{after} are handled (e.g.\ clauses introduced by \qit{when} or
\qit{since} and relative clauses are not supported). The issue of
tense coordination between main and subordinate clauses is ignored.

Among temporal anaphora phenomena, only the temporal anaphoric nature
of noun phrases like \qit{the sales manager} is
supported. Proper names like \qit{May} or \qit{Monday} are 
taken to refer to \emph{any} May or Monday. Similarly, past tenses
are treated as referring to \emph{any} past time. Temporal anaphoric
expressions like \qit{that time} or \qit{the following day} are not
allowed. (Nominal anaphoric expressions, e.g.\ \qit{he}, \qit{her
salary}, are also not allowed.)

The framework of this thesis does not support cardinality or duration
queries (\qit{How many~\dots?}, \qit{How long~\dots?}) and cardinality
expressions (e.g.\ \qit{five flights}). Plurals introduced by
\qit{which} and \qit{some} (e.g.\ \qit{which flights}, \qit{some
gates}) are treated semantically as singular. Conjunctions of words or
phrases, and expressions introducing universal quantifiers,
disjunction, or negation are also not supported. Finally, only active
voice verb forms have been considered, though it should be easy to
extend the mechanisms of this thesis to support passive voice as well.

Table \ref{coverage_table} summarises the linguistic coverage of the
framework of this thesis. 

\begin{table}
\begin{tabular}{|l|l|}
\hline
verb tenses & \supp   simple present (excluding scheduled meaning) \\
            & \supp   simple past \\
            & \supp   present continuous (excluding futurate meaning) \\
            & \supp   past continuous (excluding futurate meaning) \\
            & \supp   present perfect (treated as simple past) \\
            & \supp   past perfect \\
            & \nosupp other tenses \\
\hline
temporal verbs & \supp \qit{to start}, \qit{to begin}, \qit{to stop},
                       \qit{to finish} \\
               & \nosupp other temporal verbs (e.g.\ \qit{to happen},
                         \qit{to follow}) \\
\hline
temporal nouns & \supp   \qit{year}, \qit{month}, \qit{day}, etc.\ \\
               & \nosupp \qit{period}, \qit{event}, \qit{time}, etc.\ \\
               & \nosupp nouns introducing situations (e.g.\
                         \qit{inspection}) \\
               & \nosupp nouns of temporal order (e.g.\ \qit{predecessor}) \\
\hline
temporal adjectives & \nosupp (only \qit{current}) \\
\hline
temporal adverbials & \supp   punctual adverbials (e.g.\ \qit{at 5:00pm})\\
                    & \supp   period adverbials (only those introduced by
                              \qit{on}, \qit{in}, \\
                    & \ \ \ \ \qit{before}, or \qit{after}, and
                              \qit{today}, \qit{yesterday}) \\
                    & \supp   \qit{for~\dots} adverbials \\
                    & \supp   \qit{in~\dots} duration adverbials (only
                              with culm.\ act.\ verbs) \\
                    & \nosupp frequency adverbials (e.g.\ \qit{twice}) \\
                    & \nosupp order adverbials (e.g.\ \qit{for the
                              second time}) \\
                    & \nosupp other boundary adverbials (e.g.\
                              \qit{since 1987}) \\
\hline 
subordinate clauses & \supp   \qit{while~\dots} clauses \\
                    & \supp   \qit{before~\dots} clauses \\
                    & \supp   \qit{after~\dots} clauses \\
                    & \nosupp relative clauses \\
                    & \nosupp other subordinate clauses (e.g.\ introduced 
                              by \qit{when}) \\
                    & \nosupp tense coordination between
                      main-subordinate clauses \\
\hline
anaphora  & \supp   noun phrases and temporal reference \\
          & \nosupp \qit{January}, \qit{August}, etc.\ \\
          & \ \ \ \ (taken to refer to any January, August, etc.) \\
          & \nosupp tense anaphora \\
          & \ \ \ \ (past tenses taken to refer to any past time) \\ 
          & \nosupp \qit{that time}, \qit{the following day}, etc.\ \\
          & \nosupp nominal anaphora (e.g.\ \qit{he}, \qit{her salary}) \\
\hline
other phenomena  & \nosupp cardinality and duration queries \\
                 & \ \ \ \ (\qit{How many~\dots?}, \qit{How long~\dots?}) \\
                 & \nosupp cardinality expressions (e.g.\ \qit{five flights})\\
                 & \nosupp plurals (treated as singulars) \\
                 & \nosupp conjunctions of words or phrases \\
                 & \nosupp expressions introducing universal quantifiers, \\
                 & \ \ \ \ disjunction, negation \\
                 & \nosupp passive voice \\
\hline   
\end{tabular}
\caption{The linguistic coverage of the framework of this thesis} 
\label{coverage_table}
\end{table}


\chapter{The TOP Language} \label{TOP_chapter}

\proverb{Time will tell.}


\section{Introduction} \label{top_intro}

This chapter defines \topl, the intermediate representation language
of this thesis.  As noted in section \ref{temp_log_intro}, \topl
employs temporal operators. \pref{tintro:1}, for example, is
represented in \topl as \pref{tintro:2}. Roughly speaking, the \past
operator requires $contain(tank2, water)$ to be true at some past time
$e^v$, and the \at operator requires that time to fall within
1/10/95. The answer to \pref{tintro:1} is affirmative iff
\pref{tintro:2} evaluates to true.
\begin{examps}
\item Did tank 2 contain water (some time) on 1/10/95? \label{tintro:1}
\item $\at[\mathit{1/10/95}, \past[e^v, contain(tank2, water)]]$ 
   \label{tintro:2}
\end{examps}
An alternative operator-less approach is to introduce time as an
extra argument of each predicate (section \ref{temp_log_intro}). I use
temporal operators because they lead to more compact formulae, and
because they make the semantic contribution of each linguistic
mechanism easier to see (in \pref{tintro:2}, the simple
past tense contributes the \past operator, while the \qit{on~\dots}
adverbial contributes the \at operator). 

\topl is period-based, in the sense that the truth of a \topl formula
is checked with respect to a time-period (a segment of the time-axis)
rather than an individual time-point. (The term ``period'' is used
here to refer to what other authors call ``intervals''; see section
\ref{temporal_ontology} below.) A \topl formula may be true at a
time-period without being true at the subperiods of that period.
Actually, following the Reichenbachian tradition \cite{Reichenbach},
\topl formulae are evaluated with respect to more than one times:
\emph{speech time} (time at which the question is submitted),
\emph{event time} (time where the situation described by the formula
holds), and \emph{localisation time} (a temporal window within which
the event time must be placed; this is different from Reichenbach's
reference time, and similar to the ``location time'' of
\cite{Kamp1993}). While speech time is always a time-point, the event
and localisation times are generally periods, and this is why I
consider \topl period-based. Period-based languages have been used in
\cite{Dowty1982}, \cite{Allen1984}, \cite{Lascarides},
\cite{Richards}, \cite{Pratt1995}, and elsewhere.  Multiple temporal
parameters have been used by several researchers (e.g.\ 
\cite{Dowty1982}, \cite{Hinrichs}, \cite{Brent1990}, \cite{Crouch2}).
The term ``localisation time'' is borrowed from \cite{Crouch2}, where
$lt$ is a temporal window for $et$ as in \topl.

Although the aspectual classes of linguistic expressions affect how
these expressions are represented in \topl, it is not always possible
to tell the aspectual class of a linguistic expression by examining
the corresponding \topl formula. The approach here is different from
those of \cite{Dowty1977}, \cite{Dowty1986}, \cite{Lascarides}, and
\cite{Kent}, where aspectual class is a property of formulae (or
denotations of formulae).

\topl was greatly influenced by the representation language of
Pirie et al.\ \cite{Pirie1990} \cite{Crouch} \cite{Crouch2}, that was
used in a natural language front-end to a planner. \topl, however,
differs in numerous ways from the language of Pirie et al.\ (several
of these differences will be mentioned in following sections).


\section{Syntax of TOP} \label{top_syntax}

This section defines the syntax of \topl. Some informal comments about
the semantics of the language are also given to make the syntax
definition easier to follow. The semantics of \topl will be defined
formally in following sections.

\paragraph{Terms:} Two disjoint sets of strings, \cons 
\index{cons@\cons (set of all \topl constants)}
(constants) and \vars 
\index{vars@\vars (set of all \topl variables)} 
(variables), are assumed. I use the suffix ``$^v$'' to distinguish
variables from constants. For example, $\mathit{runway^v},
\mathit{gate1^v} \in \vars$, while $\mathit{ba737}, \mathit{1/5/94}
\in \cons$. \terms 
\index{terms@\terms (set of all \topl terms)} 
(\topl terms) is the set $\cons \union \vars$. (\topl has no function
symbols.)

\paragraph{Predicate functors:} A set of strings \pfuns
\index{pfuns@\pfuns (set of all \topl predicate functors)} 
is assumed. These strings are used as predicate functors (see atomic
formulae below).

\paragraph{Complete partitioning names:} A set of strings \cparts
\index{cparts@\cparts (set of all \topl complete partitioning names)}
is assumed. These strings represent \emph{complete partitionings} of
the time-axis. A complete partitioning of the time-axis is a set of
consecutive non-overlapping periods, such that the union of all the
periods covers the whole time-axis. (A formal definition will be given
in section \ref{top_model}.) For example, the word \qit{day}
corresponds to the complete partitioning that contains the period that
covers exactly the day 13/10/94, the period that covers exactly
14/10/94, etc. No day-period overlaps another one, and together all
the day-periods cover the whole time-axis. Similarly, \qit{month}
corresponds to the partitioning that contains the period for October
1994, the period for November 1994, etc. I use the suffix ``$^c$'' for
elements of \cparts. For example, $\mathit{day}^c$ could represent the
partitioning of day-periods, and $\mathit{month}^c$ the partitioning
of month-periods.

\paragraph{Gappy partitioning names:} A set of strings \gparts
\index{gparts@\gparts (set of all \topl gappy partitioning names)} 
is assumed. These strings represent \emph{gappy partitionings} of the
time-axis. A gappy partitioning of the time-axis is a set of
non-overlapping periods, such that the union of all the periods does
\emph{not} cover the whole time-axis. For example, \qit{Monday}
corresponds to the gappy partitioning that contains the period which
covers exactly the Monday 17/10/94, the period that covers exactly the
Monday 24/10/94, etc. No Monday-period overlaps another Monday-pariod,
and all the Monday-periods together do not cover the whole
time-axis. I use the suffix ``$^g$'' for elements of \gparts. For
example, $\mathit{monday}^g$ could represent the partitioning of
Monday-periods, and $\text{\textit{5:00pm}}^g$ the partitioning of all
5:00pm-periods (the period that covers exactly the 5:00pm minute of
24/10/94, the period that covers the 5:00pm minute of 25/10/94, etc.).

\paragraph{Partitioning names:} 
\index{parts@\parts (set of all \topl partitioning names)} 
\parts (partitioning names) is the set $\cparts \union \gparts$.

\paragraph{Atomic formulae:} 
\index{aforms@\aforms (set of all \topl atomic formulae)} 
\aforms (atomic formulae) is the smallest possible set, such that: 
\begin{itemize}

\item If $\pi \in \pfuns$, and $\tau_1, \tau_2, \dots, \tau_n \in
\terms$, then $\pi(\tau_1, \tau_2, \dots, \tau_n) \in
\aforms$. $\pi(\tau_1, \tau_2, \dots, \tau_n)$ is called a
\emph{predicate}. $\tau_1, \tau_2, \dots, \tau_n$ are the
\emph{arguments} of the predicate.

\item 
\index{part@$\partop[\;]$ (used to select periods from partitionings)} 
If $\sigma \in \parts$, $\beta \in \vars$, and
$\nu_{ord} \in \{\dots, -3, -2, -1, 0\}$, then $\partop[\sigma, \beta,
\nu_{ord}] \in \aforms$ and $\partop[\sigma, \beta] \in \aforms$. 

\end{itemize}
Greek letters are used as meta-variables, i.e.\ they stand for
expressions of \topl. Predicates (e.g.\ $be\_at(ba737,
gate^v)$) describe situations in the world. $\partop[\sigma, \beta,
\nu_{ord}]$ means that $\beta$ is a period in the partitioning
$\sigma$. The $\nu_{ord}$ is used to select a particular period from
the partitioning. If $\nu_{ord} = 0$, then $\beta$ is the current
period of the partitioning (the one that contains the present
moment). If $\nu_{ord} < 0$, then $\beta$ is the $-\nu_{ord}$-th
period of the partitioning before the current one. When there is no
need to select a particular period from a partitioning, the
$\partop[\sigma, \beta]$ form is used.

\paragraph{Yes/no formulae:} Yes/no formulae represent questions that
are to be answered with a \qit{yes} 
or \qit{no} (e.g.\ \qit{Is BA737 circling?}). \ynforms
\index{ynforms@\ynforms (set of all \topl yes/no formulae)}
is the set of all yes/no formulae. It is the smallest possible set,
such that if $\pi \in \pfuns$, $\tau_1, \dots,
\tau_n \in \terms$, $\phi, \phi_1, \phi_2 \in \forms$, $\sigma_c \in
\cparts$, $\nu_{qty} \in \{1,2,3,\dots\}$, $\beta$ is a
\topl variable that does not occur in $\phi$, and $\tau$ is a \topl
variable that does not occur in $\phi$ or a \topl constant, all the
following hold. (The restriction that $\beta$ and $\tau$ must not be
variables that occur in $\phi$ is needed in the translation from \topl
to \tsql of chapter \ref{tdb_chapter}.)
\begin{itemize}
\item $\aforms \subseteq \ynforms$

\item $\phi_1 \land \phi_2 \in \ynforms$ 
\index{^@$\land$ (\topl's conjunction)}

\item 
\index{pres@$\pres[\;]$ (used to refer to the present)}
\index{past@$\past[\;]$ (used to refer to the past)}
\index{perf@$\perf[\;]$ (used to express the past perfect)}
$\pres[\phi]$, $\past[\beta, \phi]$, $\perf[\beta, \phi] \in \ynforms$

\item 
\index{at@$\at[\;]$ (narrows the localisation time)}
$\at[\tau, \phi]$, $\at[\phi_1, \phi_2] \in \ynforms$

\item 
\index{before@$\before[\;]$ (used to express \qit{before})}
\index{after@$\after[\;]$ (used to express \qit{after})}
$\before[\tau, \phi]$, $\before[\phi_1, \phi_2]$, $\after[\tau,
   \phi]$, $\after[\phi_1, \phi_2] \in \ynforms$

\item 
\index{ntense@$\ntense[\;]$ (used when expressing nouns or adjectives)}
$\ntense[\beta, \phi]$, $\ntense[\mathit{now}^*, \phi] \in \ynforms$

\item 
\index{for@$\for[\;]$ (used to express durations)}
\index{fills@$\fills[\;]$ (requires $et = lt$)}
$\for[\sigma_c, \nu_{qty}, \phi]$, $\fills[\phi] \in \ynforms$

\item 
\index{begin@$\lbegin[\;]$ (used to refer to start-points of situations)}
\index{end@$\lend[\;]$ (used to refer to end-points of situations)}
$\lbegin[\phi]$, $\lend[\phi] \in \ynforms$

\item 
\index{culm@$\culm[\;]$ (used to express non-progressives of culminating activity verbs)}
$\culm[\pi(\tau_1, \dots, \tau_n)] \in \ynforms$

\end{itemize}

No negation and disjunction connectives are defined, because English
expressions introducing these connectives are not considered (section
\ref{ling_not_supported}). For the same reason no universal
quantifiers are defined. All variables can be thought of as
existentially quantified. Hence, no explicit existential quantifier is
needed.
 
An informal explanation of \topl's operators follows (\topl's semantics
will be defined formally in following sections). 
$\pres[\phi]$ means that $\phi$ is true at the
present. For example, \qit{Runway 2 is open.} is represented as
$\pres[open(runway2)]$.  $\past[\beta, \phi]$ means that
$\phi$ is true at some past time $\beta$. The \perf operator is used
along with the \past operator to express the past perfect. For
example, \qit{Runway 2 was open.} is represented as $\past[e^v,
open(runway2)]$, and \qit{Runway 2 had been open.} as:
\[\past[e1^v,\perf[e2^v, open(runway2)]] \]

$\at[\tau, \phi]$ means that $\phi$ holds some time
within a period $\tau$, and $\at[\phi_1, \phi_2]$ means that
$\phi_2$ holds at some time where $\phi_1$ holds. For example,
\qit{Runway 2 was open (some time) on 1/1/94.} is represented as
$\at[\mathit{1/1/94}, \past[e^v, open(runway2)]]$,
and \qit{Runway 2 was open (some time) while BA737 was circling.} as:
\[\at[\past[e1^v, circling(ba737)], \past[e2^v, open(runway2)]]\]

$\before[\tau, \phi]$ means that $\phi$ is true at some
time before a period $\tau$, and $\before[\phi_1, \phi_2]$ means that
$\phi_2$ is true at some time before a time where $\phi_1$ is
true. $\after[\tau, \phi]$ and $\after[\phi_1, \phi_2]$ have similar
meanings. For example, \qit{Tank 2 was empty (some time) after
1/1/92.} is represented as $\after[\mathit{1/1/92}, \past[e^v,
empty(tank2)]]$, and \qit{Tank 2 was empty (some time) before the bomb
exploded.} as:
\[
\before[\past[e1^v, explode(bomb)], \past[e2^v, empty(tank2)]]
\]

\ntense is used when expressing noun phrases (see section
\ref{noun_anaphora}). $\ntense[\beta, \phi]$ means that
at a time $\beta$ something has the property specified by $\phi$. 
$\ntense[\mathit{now}^*, \phi]$ means that something has the
property specified by $\phi$ at the present. The reading of \qit{The
president was visiting Edinburgh.} that refers to the person who was
the president during the visit is represented as 
$\ntense[e1^v, president(p^v)] \land \past[e1^v, visiting(p^v, edinburgh)]$.
In contrast, the reading that refers to the current president is
represented as: 
\[\ntense[\mathit{now}^*,president(p^v)] \land 
  \past[e1^v, visiting(p^v, edinburgh)]
\]

$\for[\sigma_c, \nu_{qty}, \phi]$ means that
$\phi$ holds throughout a period that is $\nu_{qty}$ $\sigma_c$-periods long.
\qit{Runway 2 was open for two days.} is represented as: 
\[\for[day^c,2, \past[e^v, open(runway2)]]
\]
The \fills operator is currently not
used in the framework of this thesis, but it could be used to capture
readings of sentences like \qit{Tank 2 was empty in 1992.} whereby the
situation of the verb holds \emph{throughout} the period of the
adverbial (see section \ref{period_adverbials}).  $\at[1992, \past[e^v,
\fills[empty(tank2)]]]$ means that the tank was empty
\emph{throughout} 1992, while $\at[1992, \past[e^v, empty(tank2)]]$
means that the tank was empty some time in 1992, but not necessarily
throughout 1992.

$\lbegin[\phi]$ means that $\phi$ starts to hold, and
$\lend[\phi]$ means that $\phi$ stops holding. For example, \qit{BA737
started to land.} can be represented as $\past[e^v,
\lbegin[landing(ba737)]]$, and \qit{Tank 2 stopped being empty.} as
$\past[e^v, \lend[empty(tank2)]]$.

Finally, \culm is used to represent sentences where verbs whose
base forms are culminating activities appear in tenses that require
some inherent climax to have been reached. The \culm operator will be
discussed in section \ref{culm_op}.

\paragraph{Wh-formulae:} \emph{Wh-formulae} are used to represent
questions that contain interrogatives (e.g.\ \qit{Which~\dots?},
\qit{Who~\dots?}, \qit{When~\dots}?). \whforms is the set of all
wh-formulae. $\whforms \defeq \whforms_1 \union \whforms_2$,
\index{whforms@\whforms (set of all \topl wh-formulae)}
\index{whforms1@$\whforms_1$ (set of all \topl wh-formulae with no $?_{mxl}$)}
\index{whforms2@$\whforms_2$ (set of all \topl wh-formulae with a $?_{mxl}$)}
where: 
\begin{itemize}

\item 
\index{?@$?$ (\topl's interrogative quantifier)}
$\whforms_1$ is the set of all expressions of the form $?\beta_1
\; ?\beta_2 \; \dots \; ?\beta_n \; \phi$, where $\beta_1, \beta_2,
\dots, \beta_n \in \vars$, $\phi \in \ynforms$, and each one of
$\beta_1, \beta_2, \dots, \beta_n$ occurs at least once within
$\phi$.

\item 
\index{?@$?$ (\topl's interrogative quantifier)}
\index{?mxl@$?_{mxl}$ (\topl's interrogative-maximal quantifier)}
$\whforms_2$ is the set of all expressions of the form
$?_{mxl}\beta_1 \; ?\beta_2 \; ?\beta_3 \; \dots \; ?\beta_n \; \phi$,
where $\beta_1, \beta_2, \beta_3, \dots, \beta_n \in \vars$, $\phi \in
\ynforms$, each one of $\beta_2$, $\beta_3$, \dots, $\beta_n$ occurs
at least once within $\phi$, and $\beta_1$ occurs at least once within
$\phi$ as the first argument of a \past, \perf, \at, \before, \after,
or \ntense operator, or as the second argument of a \partop operator. 

\end{itemize}

``$?$'' is the \emph{interrogative quantifier}, and $?_{mxl}$ the
\emph{interrogative-maximal quantifier}. The interrogative quantifier
is similar to an explicit existential quantifier, but it has the
additional effect of reporting the values of its variables that
satisfy its scope. Intuitively, $?\beta_1 \; ?\beta_2 \; ?\beta_n \;
\phi$ means \qit{report all $\beta_1, \beta_2, \dots, \beta_n$ such
that $\phi$}. For example, \qit{Which runways are open?} is
represented as $?r^v \; \ntense[\mathit{now}^*, runway(r^v)] \land
\pres[open(r^v)]$. The constraint that each one of $\beta_1, \dots,
\beta_n$ must occur at least once within $\phi$ rules out meaningless
formulae like $?o^v \; \past[manager(john)]$, where the $o^v$ does not
have any relation to the rest of the formula. This constraint is
similar to the notion of \emph{safety} in \datalog \cite{Ullman}, and
it is needed in the translation from \topl to \tsql of chapter
\ref{tdb_chapter}.

The interrogative-maximal quantifier is similar, except that it
reports only \emph{maximal periods}. $?_{mxl}$ is intended to be used
only with variables that denote periods, and this is why in the case
of $?_{mxl}$, $\beta_1$ is required to occur within $\phi$ as the
first argument of a \past, \perf, \at, \before, \after, or \ntense
operator, or as the second argument of a \partop operator (the
semantics of these operators ensure that variables occurring at these
potitions denote periods).  Intuitively, $?_{mxl}\beta_1 \;
?\beta_2 \; ?\beta_n \; \phi$ means \qit{report all the maximal
  periods $\beta_1$, and all $\beta_2$, \dots, $\beta_n$, such that
  $\phi$}. The interrogative-maximal quantifier is used in \qit{When
  \dots?} questions, where we want the answer to contain only the
\emph{maximal} periods during which a situation held, not all the
periods during which the situation held. If, for example, gate 2 was
open from 9:00am to 11:00am and from 3:00pm to 5:00pm, we want the
answer to \qit{When was gate 2 open?} to contain only the two maximal
periods 9:00am to 11:00am and 3:00pm to 5:00pm; we do not want the
answer to contain any subperiods of these two maximal periods (e.g.\ 
9:30am to 10:30am). To achieve this, the question is represented as
$?_{mxl}e^v \; \past[e^v, open(gate2)]$.

\paragraph{Formulae:} 
\index{forms@\forms (set of all \topl formulae)}
\forms is the set of all \topl formulae. $\forms \defeq \ynforms
\union \whforms$. 


\section{The temporal ontology} \label{temporal_ontology}

\paragraph{Point structure:} A \emph{point structure} for \topl is an
ordered pair $\tup{\pts, \prec}$, such that \pts 
\index{pts@\pts (set of all time-points)}
is a non-empty set, $\prec$ \index{<@$\prec$ (precedes)} is a binary
relation over $\pts \times \pts$, and $\tup{\pts, \prec}$ has the
following five properties: 
\begin{description}

\item[transitivity:] If $t_1, t_2, t_3 \in \pts$, $t_1 \prec t_2$, and
$t_2 \prec t_3$, then $t_1 \prec t_3$.

\item[irreflexivity:] If $t \in \pts$, then $t \prec t$ does not hold.

\item[linearity:] If $t_1, t_2 \in \pts$ and $t_1 \not= t_2$, then
exactly one of the following holds: $t_1 \prec t_2$ or $t_2 \prec t_1$.

\item[left and right boundedness:] There is a $t_{first} \in \pts$,
\index{tfirst@$t_{first}$ (earliest time-point)} 
such that for all $t \in \pts$, $t_{first} \preceq t$. Similarly, there is
a $t_{last} \in \pts$, 
\index{tlast@$t_{last}$ (latest time-point)}
such that for all $t \in \pts$, $t \preceq t_{last}$.

\item[discreteness:] For every $t_1, t_2 \in \pts$, with $t_1 \not=
t_2$, there is at most a finite number of $t_3 \in \pts$, such
that $t_1 \prec t_3 \prec t_2$.

\end{description}

Intuitively, a point structure $\tup{\pts, \prec}$ for \topl is a
model of time. \topl models time as being discrete, linear, bounded,
and consisting of time-points (see \cite{VanBenthem} for other time
models.) \pts is the set of all time-points, and $p_1
\prec p_2$ means that the time-point $p_1$ precedes the
time-point $p_2$. 

\paragraph{prev(t) and next(t):} 
\index{prev@$prev()$ (previous time-point)} 
\index{next@$next()$ (next time-point)} 
If $t_1 \in \pts - \{t_{last}\}$, then $next(t_1)$ denotes a $t_2 \in
\pts$, such that $t_1 \prec t_2$ and for no $t_3 \in \pts$ is it true
that $t_1 \prec t_3 \prec t_2$. Similarly, if $t_1 \in \pts -
\{t_{first}\}$, then $prev(t_1)$ denotes a $t_2 \in \pts$, such that
$t_2 \prec t_1$ and for no $t_3 \in \pts$ is it true that $t_2 \prec
t_3 \prec t_1$. In the rest of this thesis, whenever $next(t)$ is
used, it is assumed that $t \not= t_{last}$. Similarly, whenever
$prev(t)$ is used, it is assumed that $t \not= t_{first}$.

\paragraph{Periods and instantaneous periods:}
A \emph{period} $p$ over $\tup{\pts, \prec}$ is a non-empty subset of
\pts with the following property:
\begin{description}

\item[convexity:] If $t_1, t_2 \in p$, $t_3 \in \pts$, and $t_1 \prec
t_3 \prec t_2$, then $t_3 \in p$.

\end{description}
The term ``interval'' is often used in the literature instead of
``period''. Unfortunately, \tsql uses ``interval'' to refer to a
duration (see chapter \ref{tdb_chapter}). To avoid confusing the
reader when \tsql will be discussed, I follow the \tsql terminology
and use the term ``period'' to refer to convex sets of time-points.

\index{periods1@$\periods_{\tup{\pts, \prec}}$ (set of all periods over $\tup{\pts, \prec}$)} 
$\periods_{\tup{\pts, \prec}}$ is the set of all periods over
$\tup{\pts, \prec}$.  If $p \in \periods_{\tup{\pts, \prec}}$ and $p$
contains only one time-point, then $p$ is an \emph{instantaneous
period over $\tup{\pts, \prec}$}. $\instants_{\tup{\pts, \prec}}$ 
\index{instants1@$\instants_{\tup{\pts, \prec}}$ (set of all instantaneous periods over $\tup{\pts, \prec}$)}
is the set of all instantaneous periods over $\tup{\pts, \prec}$. For
simplicity, I often write \periods 
\index{periods@$\periods$ (set of all periods)} 
and \instants  
\index{instants@$\instants$ (set of all instantaneous periods)}
instead of $\periods_{\tup{\pts, \prec}}$ and $\instants_{\tup{\pts,
\prec}}$, and I often refer to simply ``periods'' and
``instantaneous periods'' instead of ``periods over $\tup{\pts,
\prec}$'' and ``instantaneous periods over $\tup{\pts, \prec}$''.

\index{periods*@$\periods^*$ ($\periods \union \emptyset$)}
$\periods^*_{\tup{\pts, \prec}}$ (or simply $\periods^*$) 
is the set $\periods \union \{\emptyset\}$, i.e.\ $\periods^*$ 
is the same as $\periods$, except that it also contains the
empty set. (The reader is reminded that periods are non-empty sets.) 

\paragraph{Subperiods:} $p_1$ is a \emph{subperiod} of
$p_2$, iff $p_1, p_2 \in \periods$ and $p_1 \subseteq p_2$. In this
case I write $p_1 \subper p_2$. 
\index{<sq@$\subper$ (subperiod)}
($p_1 \subseteq p_2$ is weaker than $p_1 \subper p_2$, because it
does not guarantee that $p_1, p_2 \in \periods$.)
Similarly, $p_1$ is a \emph{proper 
subperiod} of $p_2$, iff $p_1, p_2 \in \periods$ and $p_1 \subset
p_2$. In this case I write $p_1 \propsubper
p_2$. 
\index{<sq@$\propsubper$ (proper subperiod)}

\paragraph{Maximal periods:} 
\index{mxlpers@$mxlpers()$ (maximal periods of a set or temporal element)}
If $S$ is a set of periods, then $\mxlpers(S)$ is the set of
\emph{maximal periods} of $S$.  $\mxlpers(S) \defeq \{p \in S \mid
\text{for no } p' \in S \text{ is it true that } p \propsubper p'\}$.

\paragraph{minpt(S) and maxpt(S):} 
\index{minpt@$minpt()$ (earliest time-point in a set)}
\index{maxpt@$maxpt()$ (latest time-point in a set)}
If $S \subseteq \pts$, $minpt(S)$ denotes
the time-point $t \in S$, such that for every $t' \in S$, 
$t \preceq t'$. Similarly, if $S \subseteq \pts$, $maxpt(S)$
denotes the time-point $t \in S$, such that for every $t' \in S$, 
$t' \preceq t$. 

\paragraph{Notation:} Following standard conventions, $[t_1,
t_2]$ denotes the set $\{t \in \pts \mid t_1 \preceq t \preceq t_2
\}$. (This is not always a period. If $t_2 \prec t_1$, then $[t_1,
t_2]$ is the empty set, which is not a period.) Similarly, $(t_1,
t_2]$ denotes the set $\{t \in \pts \mid t_1 \prec t \preceq t_2
\}$. $[t_1, t_2)$ and $(t_1,t_2)$ are defined similarly.


\section{TOP model} \label{top_model}

A \topl model $M$ is an ordered 7-tuple:
\[ M = \tup{\tup{\pts, \prec}, \objs, 
            \fcons, \fpfuns, \fculms, \fgparts, \fcparts}
\]
such that $\tup{\pts, \prec}$ is a point structure for \topl 
(section \ref{temporal_ontology}), $\periods_{\tup{\pts, \prec}}
\subseteq \objs$, and \fcons, \fpfuns, \fculms, \fgparts, and \fcparts
are as specified below. 

\paragraph{$\mathbf{OBJS}$:} 
\index{objs@\objs (\topl's world objects)}
\objs is a set containing all the objects in the
modelled world that can be denoted by \topl terms. For example, in the
airport domain \objs contains all the gates and runways of the
airport, the inspectors, the flights, etc. The constraint
$\periods_{\tup{\pts, \prec}} \subseteq \objs$ ensures that all
periods are treated as world objects. This simplifies the semantics of
\topl.

\paragraph{$\mathbf{f_{cons}}$:}
\index{fcons@$\fcons()$ (maps \topl constants to world objects)}
\fcons is a function $\cons \mapsto \objs$. (I use the notation
$D \mapsto R$ to refer to a function whose domain and range are $D$
and $R$ respectively.) \fcons specifies which world object each
constant denotes. In the airport domain, for example, \fcons may map
the constants $gate2$ and $ba737$ to some gate of the airport and some
flight respectively.

\paragraph{$\mathbf{f_{pfuns}}$:}
\index{fpfuns@$\fpfuns()$ (returns the maximal periods where predicates hold)}
\fpfuns is a function that maps each pair $\tup{\pi, n}$, where
$\pi \in \pfuns$ and $n \in \{1,2,3,\dots\}$, to another function
$(\objs)^n \mapsto \pow(\periods)$. ($\pow(S)$\/ 
\index{pow@$\pow()$ (powerset)} 
denotes the powerset of $S$, i.e.\ the set of all subsets of $S$. 
$(\objs)^n$ 
\index{objsn@$(\objs)^n$ ($\objs \times \dots \times \objs$)}
is the $n$-ary cartesian product $\objs \times \objs
\times \dots \times \objs$.) That is, for every $\pi \in \pfuns$ and
each $n \in \{1,2,3,\dots\}$, $\fpfuns(\pi,n)$ is a function that maps
each $n$-tuple of elements of $\objs$ to a set of periods (an element
of $\pow(\periods)$). 

Intuitively, if $\tau_1, \tau_2, \dots, \tau_n$ are \topl terms
denoting the world objects $o_1, o_2, \dots, o_n$,
$\fpfuns(\pi, n)(o_1, o_2, \dots, o_n)$ is the set of the maximal
periods throughout which the situation described by $\pi(\tau_1,
\tau_2, \dots, \tau_n)$ is true.  For example, if the constant
$ba737$ denotes a flight-object $o_1$, $gate2$ denotes a gate-object $o_2$, and
$be\_at(ba737,gate2)$ describes the situation whereby the flight $o_1$
is located at the gate $o_2$, then $\fpfuns(be\_at, 2)(o_1, o_2)$ will
be the set that contains all the maximal periods throughout which the
flight $o_1$ is located at the gate $o_2$. 

For every $\pi \in \pfuns$ and $n \in \{1,2,3,\dots\}$, 
$\fpfuns(\pi, n)$ must have the following property: for every
$\tup{o_1,o_2,\dots,o_n} \in (\objs)^n$, it must be
the case that: 
\[
\text{if } p_1, p_2 \in \fpfuns(\pi, n)(o_1,o_2,\dots,o_n) \text{ and }
p_1 \union p_2 \in \periods, \text{ then } p_1 = p_2
\]
This ensures that no two different periods $p_1, p_2$ in $\fpfuns(\pi,
n)(o_1,\dots,o_n)$ overlap or are adjacent (because if they overlap or
they are adjacent, then their union is also a period, and then it must
be true that $p_1 = p_2$). Intuitively, if $p_1$ and $p_2$ overlap or are
adjacent, we want $\fpfuns(\pi, n)(o_1,o_2,\dots,o_n)$ to contain
their union $p_1 \union p_2$ instead of $p_1$ and $p_2$.

\paragraph{$\mathbf{f_{culms}}$:}
\index{fculms@$\fculms()$ (shows if the situation of a predicate reaches its climax)}
\fculms is a function that maps each pair $\tup{\pi, n}$, where $\pi
\in \pfuns$ and $n \in \{1,2,3,\dots\}$, to another function $(\objs)^n
\mapsto \{T,F\}$ ($T,F$ are the two truth values). That is, for every
$\pi \in \pfuns$ and each $n \in \{1,2,3,\dots\}$, $\fculms(\pi,n)$ is
a function that maps each $n$-tuple of elements of
\objs to $T$ or $F$. 

\fculms is only consulted in the case of predicates that represent
actions or changes that have inherent climaxes. If $\pi(\tau_1,
\tau_2, \dots, \tau_n)$ represents such an action or change, and
$\tau_1, \tau_2, \dots, \tau_n$ denote the world objects $o_1, o_2,
\dots, o_n$, then $\fpfuns(\pi, n)(o_1, o_2, \dots, o_n)$ is
the set of all maximal periods throughout which the action or change
is ongoing.  $\fculms(\pi, n)(o_1, o_2, \dots, o_n)$ shows whether or
not the change or action reaches its climax at the latest time-point
at which the change or action is ongoing.
For example, if the constant $j\_adams$ denotes a person $o_1$
in the world, $bridge2$ denotes an object
$o_2$, and $building(j\_adams, ba737)$ describes the
situation whereby $o_1$ is building $o_2$, 
$\fpfuns(building,2)(o_1,o_2)$ will be the set of all maximal periods
where $o_1$ is building $o_2$. $\fculms(building,2)(o_1,o_2)$ will be
$T$ if the building is completed at the end-point of the latest
maximal period in $\fpfuns(building,2)(o_1,o_2)$, and $F$
otherwise. The role of \fculms will become clearer in section
\ref{culm_op}.

\paragraph{$\mathbf{f_{gparts}}$:}
\index{fgparts@$\fgparts()$ (assigns gappy partitionings to elements of \gparts)}
\fgparts is a function that maps each element of \gparts to a
\emph{gappy partitioning}. A gappy partitioning is a subset $S$ of
\periods, such that for every $p_1, p_2 \in S$, $p_1 \intersect p_2 =
\emptyset$, and $\bigcup_{p \in S}p \not= \pts$. For example,
$\fgparts(monday^g)$ could be the gappy partitioning of all Monday-periods.

\paragraph{$\mathbf{f_{cparts}}$:}
\index{fcparts@$\fcparts()$ (assigns complete partitionings to elements of \cparts)}
\fcparts is a function that maps each element of \cparts to a
\emph{complete partitioning}. A complete partitioning is a subset
$S$ of \periods, such that for every $p_1, p_2 \in S$, $p_1 \intersect
p_2 = \emptyset$, and $\bigcup_{p \in S}p = \pts$. For example,
$\fcparts(day^c)$ could be the complete partitioning of all
day-periods.


\section{Variable assignment} \label{var_assign}

A variable assignment with respect to (w.r.t.) a \topl model $M$ 
is a function $g: \vars \mapsto \objs$
\index{g@$g()$, $g^\beta_o()$ (variable assignment)}
($g$ assigns to each variable an element of \objs). $G_M$,
or simply $G$, 
\index{G@$G$, $G_M$ (set of all variable assignments)}
is the set of all possible variable assignments w.r.t.\ $M$, i.e.\ $G$
is the set of all functions $\vars \mapsto \objs$.

If $g \in G$, $\beta \in \vars$, and $o \in \objs$, then $g^\beta_o$
\index{g@$g()$, $g^\beta_o()$ (variable assignment)}
is the variable assignment defined as follows: $g^\beta_o(\beta) = o$,
and for every $\beta' \in \vars$ with $\beta' \not= \beta$,
$g^\beta_o(\beta') = g(\beta)$.


\section{Denotation of a TOP expression} \label{denotation}

\paragraph{Index of evaluation:} An index of evaluation is an ordered 3-tuple
$\tup{st,et,lt}$, such that $st \in \pts$, $et \in \periods$, and $lt
\in \periods^*$. 

$st$ 
\index{st@$st$ (speech time)}
(\emph{speech time}) is the time-point at which the English question
is submitted to the \nlitdb.  $et$ 
\index{et@$et$ (event time)}
(\emph{event time}) is a period where the situation described by a
\topl expression takes place.  $lt$ 
\index{lt@$lt$ (localisation time)}
(\emph{localisation time}) can be thought of as a temporal window,
within which $et$ must be located. When computing the denotation of a
\topl formula that corresponds to an English question, $lt$ is
initially set to
\pts. That is, the temporal window covers the whole
time-axis, and $et$ is allowed to be located anywhere. Various
operators, however, may narrow down $lt$, imposing constraints on
where $et$ can be placed.

\paragraph{Denotation w.r.t.\ M, st, et, lt, g:} 
The denotation of a \topl expression $\xi$ w.r.t.\ a model $M$,
an index of evaluation $\tup{st,et,lt}$, and a variable assignment $g$,
is written $\denot{M,st,et,lt,g}{\xi}$ or simply
$\denot{st,et,lt,g}{\xi}$. When the denotation of $\xi$ does not
depend on $st$, $et$, and $lt$, I often write $\denot{M,g}{\xi}$
or simply $\denot{g}{\xi}$. 

The denotations w.r.t.\ $M,st,et,lt,g$ of \topl expressions are defined
recursively, starting with the denotations of terms and atomic
formulae which are defined below.
\begin{itemize}

\item If $\kappa \in \cons$, then $\denot{g}{\kappa} =
  \fcons(\kappa)$.

\item If $\beta \in \vars$, then $\denot{g}{\beta} =
  g(\beta)$.

\item If $\phi \in \ynforms$, then $\denot{st,et,lt,g}{\phi} \in
\{T,F\}$. 
\end{itemize}

The general rule above means that in the case of yes/no formulae, we only
need to define when the denotation is $T$. In all other cases the
denotation is $F$. 

\begin{itemize}
\item If $\phi_1, \phi_2 \in \ynforms$, then
\index{^@$\land$ (\topl's conjunction)}
$\denot{st,et,lt,g}{\phi_1 \land \phi_2} = T$ iff 
$\denot{st,et,lt,g}{\phi_1} = T$ and $\denot{st,et,lt,g}{\phi_2} = T$. 

\item 
\index{part@$\partop[\;]$ (used to select periods from partitionings)} 
If $\sigma \in \parts$, $\beta \in \vars$, and
$\nu_{ord} \in \{\dots, -3, -2, -1, 0\}$, then
$\denot{g}{\partop[\sigma, \beta, \nu_{ord}]}$ is $T$, iff
all the following hold (below $f = \fcparts$ if $\sigma \in
\cparts$, and $f = \fgparts$ if $\sigma \in \gparts$):
  \begin{itemize}
  \item $g(\beta) \in f(\sigma)$, 

  \item if $\nu_{ord} = 0$, then $st \in g(\beta)$, 

  \item if $\nu_{ord} \leq -1$, then the
    following set contains exactly $-\nu_{ord} - 1$ elements:
    \[
    \{ p \in f(\sigma) \mid
       maxpt(g(\beta)) \prec minpt(p) \text{ and } maxpt(p) \prec st \} 
    \]
  \end{itemize}
\end{itemize}

Intuitively, if $\nu_{ord} = 0$, then $\beta$ must denote a period in
the partitioning that contains $st$. If $\nu_{ord} \leq -1$, $\beta$
must denote the $-\nu_{ord}$-th period of the partitioning
that is completely situated before the speech time (e.g.\ if
$\nu_{ord} = -4$, $\beta$ must denote the 4th period which is
completely situated before $st$); that is, there must be $-\nu_{ord} -
1$ periods in the partitioning that fall completely between
the end of the period denoted by $\beta$ and $st$ ($-(-4) - 1 = 3$
periods if $\nu_{ord} = -4$).

For example, if $\fcparts(day^c)$ is the partitioning of
all day-periods, then $\denot{g}{\partop[day^c, \beta, 0]}$ is $T$
iff $g(\beta)$ covers exactly the whole current
day. Similarly, $\denot{g}{\partop[day^c, \beta, -1]}$ 
is $T$ iff $g(\beta)$ covers exactly the whole
previous day. ($\partop[day^c, \beta, 0]$ and $\partop[day^c, \beta,
-1]$ can be used to represent the meanings of \qit{today} and
\qit{yesterday}; see section \ref{at_before_after_op}.)
The definition of \partop could be extended to allow positive values
as its third argument. This would allow expressing \qit{tomorrow},
\qit{next January}, etc.

\begin{itemize}
\index{part@$\partop[\;]$ (used to select periods from partitionings)} 
\item If $\sigma \in \parts$ and $\beta \in \vars$,
then $\denot{g}{\partop[\sigma, \beta]} = T$ iff $g(\beta)
\in f(\sigma)$ (where $f = \fcparts$ if $\sigma \in \cparts$, and $f
= \fgparts$ if $\sigma \in \gparts$).
\end{itemize}

$\partop[\sigma, \beta]$ is a simplified version of $\partop[\sigma,
\beta, \nu_{ord}]$, used when we want to ensure that $g(\beta)$ is
simply a period in the partitioning of $\sigma$.

\begin{itemize}
\item If $\pi \in \pfuns$ and $\tau_1, \tau_2, \dots, \tau_n \in
\terms$, then $\denot{st,et,lt,g}{\pi(\tau_1, \tau_2, \dots, \tau_n)}$
is $T$ iff $et \subper lt$ and for some $p_{mxl} \in \fpfuns(\pi,
n)(\denot{g}{\tau_1}, \denot{g}{\tau_2}, \dots, \denot{g}{\tau_n})$,
$et \subper p_{mxl}$.

\end{itemize}

Intuitively, for the denotation of a predicate to be $T$, $et$ must
fall within $lt$, and $et$ must be a subperiod of a maximal period
where the situation described by the predicate holds. It is trivial to
prove that the definition above causes all \topl predicates to have
the following property:

\paragraph{Homogeneity:} A \topl formula $\phi$ is \emph{homogeneous}, iff
for every $st \in \pts$, $et \in \periods$, $lt \in \periods^*$, and $g \in
G$, the following implication holds:\footnote{The term ``homogeneity''
is also used in the temporal databases literature, but with a
completely different meaning; see \cite{tdbsglossary}.} 
\[
\text{if } et' \subper et \text{ and } \denot{st,et,lt,g}{\phi} = T, 
\text{ then } \denot{st,et',lt,g}{\phi} = T
\]
Intuitively, if a predicate is true at some $et$, then it is also true
at any subperiod $et'$ of $et$. Although \topl predicates are
homogeneous, more complex formulae are not always homogeneous.
Various versions of homogeneity have been used in \cite{Allen1984},
\cite{Lascarides}, \cite{Richards}, \cite{Kent}, \cite{Pratt1995}, and
elsewhere.

The denotation of a wh-formula w.r.t.\ $st$, $et$, $lt$, and $g$ is
defined below. It is assumed that $\beta_1, \beta_2, \beta_3, \dots,
\beta_n \in \vars$ and $\phi \in \ynforms$.
\begin{itemize}
\item
 \index{?@$?$ (\topl's interrogative quantifier)}
     $\denot{st,et,lt,g}{?\beta_1 \; ?\beta_2 \; \dots \; ?\beta_n \; \phi} 
      = \{\tup{g(\beta_1), g(\beta_2), \dots, g(\beta_n)} \mid
          \denot{st,et,lt,g}{\phi} = T\}$
\end{itemize}
That is, if $\denot{st,et,lt,g}{\phi} = T$, then
$\denot{st,et,lt,g}{?\beta_1 \; ?\beta_2 \; \dots \; ?\beta_n \;
\phi}$ is a one-element set; it contains one tuple that holds the
world-objects assigned to $\beta_1, \beta_2, \dots, \beta_n$ by
$g$. Otherwise, $\denot{st,et,lt,g}{?\beta_1 \; ?\beta_2 \; \dots \;
?\beta_n \; \phi}$ is the empty set.

\begin{itemize}
\item 
\index{?@$?$ (\topl's interrogative quantifier)}
\index{?mxl@$?_{mxl}$ (\topl's interrogative-maximal quantifier)}
      $
      \denot{st,et,lt,g}{?_{mxl}\beta_1 \; ?\beta_2 \; ?\beta_3 \;
      \dots \; ?\beta_n \; \phi} = 
      $ \\
      $
      \{\tup{g(\beta_1), g(\beta_2), g(\beta_3), \dots, g(\beta_n)} \mid 
      \denot{st,et,lt,g}{\phi} = T \text{, and }
      $ \\
      $
      \text{ for no } et' \in \periods \text{ and } g' \in G 
      \text{ is it true that  } 
      \denot{st,et',lt,g'}{\phi} = T, 
      $ \\
      $
      g(\beta_1) \propsubper g'(\beta_1), \; 
      g(\beta_2) = g'(\beta_2), \; g(\beta_3) = g'(\beta_3), \;
      \dots, \; g(\beta_n) = g'(\beta_n)\}
      $
\end{itemize}
The denotation $\denot{st,et,lt,g}{?_{mxl}\beta_1 \; ?\beta_2 \;
?\beta_3 \; \dots \; ?\beta_n \; \phi}$ is either a one-element
set that contains a tuple holding the world-objects $g(\beta_1),
g(\beta_2), \dots, g(\beta_n)$, or the empty set. Intuitively, the
denotation of $?_{mxl}\beta_1 \; ?\beta_2 \; ?\beta_3 \; \dots \;
?\beta_n \; \phi$ contains the values assigned to $\beta_1, \beta_2,
\beta_3, \dots, \beta_n$ by $g$, if these values satisfy $\phi$, and
there is no other variable assignment $g'$ that assigns the
same values to $\beta_2, \beta_3, \dots, \beta_n$, a superperiod of
$g(\beta_1)$ to $\beta_1$, and that satisfies $\phi$ (for any $et' \in
\periods$). That is, it must not be possible to extend any further the
period assigned to $\beta_1$ by $g$, preserving at the same time the
values assigned to $\beta_2, \beta_3, \dots, \beta_n$, and satisfying
$\phi$. Otherwise, the denotation of $?_{mxl}\beta_1 \; ?\beta_2 \;
?\beta_3 \; \dots \; ?\beta_n \; \phi$ is the empty set.

The syntax of \topl (section \ref{top_syntax})
requires $\beta_1$ to appear at least once within $\phi$ as the first
argument of a \past, \perf, \at, \before, \after, or \ntense operator,
or as the second argument of a \partop operator. The semantics of
these operators require variables occurring at these positions 
to denote periods.  Hence, variable assignments $g$ that do not
assign a period to $\beta_1$ will never satisfy $\phi$, and no tuples
for these variable assignments will be included in
$\denot{st,et,lt,g}{?_{mxl}\beta_1 \; ?\beta_2 \; ?\beta_3 \; \dots \;
?\beta_n \; \phi}$.

The rules for computing the denotations w.r.t.\ $M,st,et,lt,g$ of
other \topl expressions will be given in following sections.

\paragraph{Denotation w.r.t.\ M, st:} I now define the
denotation of a \topl expression with respect to only $M$ and
$st$. The denotation w.r.t.\ $M, st$ is similar to the denotation
w.r.t.\ $M, st, et, lt, g$, except that there is an implicit
existential quantification over all $g \in G$ and all $et \in
\periods$, and $lt$ is set to \pts (the whole time-axis). The
denotation of $\phi$ w.r.t.\ $M, st$, written $\denot{M,st}{\phi}$ or simply
$\denot{st}{\phi}$, is defined only for \topl formulae:
\begin{itemize}

\item If $\phi \in \ynforms$, then $\denot{st}{\phi} =$
\begin{itemize}
\item $T$, if for some $g \in G$ and $et \in \periods$, 
$\denot{st,et,\pts,g}{\phi} = T$, 
\item $F$, otherwise
\end{itemize}

\item If $\phi \in \whforms$, then $\denot{st}{\phi} = 
   \bigcup_{g \in G, \; et \in \periods}\denot{st,et,\pts,g}{\phi}$. 

\end{itemize}
Each question will be mapped to a \topl formula $\phi$
(if the question is ambiguous, multiple formulae will
be generated, one for each reading). $\denot{st}{\phi}$ specifies what
the \nlitdb's answer should report. When $\phi \in \ynforms$,
$\denot{st}{\phi} = T$ (i.e.\ the answer should be \qit{yes}) if for
some assignment to the variables of $\phi$ and for some event time,
$\phi$ is satisfied; otherwise
$\denot{st}{\phi} = F$ (the answer should be \qit{no}). The
localisation time is set to \pts (the whole time-axis) to
reflect the fact that initially there is no restriction on where 
$et$ may be located. As mentioned in section \ref{denotation},
however, when computing the denotations of the subformulae of
$\phi$, temporal operators may narrow down the localisation
time, placing restrictions on $et$.

In the case where $\phi \in \whforms$ (i.e 
$\phi = ?\beta_1 \; \dots \; ?\beta_n \; \phi'$ or
$\phi = ?_{mxl}\beta_1 \; \dots \; ?\beta_n \; \phi'$ 
with $\phi' \in \ynforms$), $\denot{st}{\phi}$ is the union
of all $\denot{st,et,\pts,g}{\phi}$, for any $g \in G$ and $et \in
\periods$. For each $g \in G$ and $et \in \periods$,
$\denot{st,et,\pts,g}{\phi}$ is either an empty set or a one-element
set containing a tuple that holds values of $\beta_1, \beta_2,
\beta_3, \dots, \beta_n$ that satisfy $\phi'$ ($\beta_1$ must be
maximal if $\phi \in \whforms_2$). Hence, $\denot{st}{\phi}$ (the
union of all $\denot{st,et,\pts,g}{\phi}$) is the set of all tuples
that hold values of $\beta_1, \beta_2, \beta_3, \dots, \beta_n$ that
satisfy $\phi'$. The answer should report these tuples to the user (or
be a message like \qit{No answer found.}, if $\denot{st}{\phi} =
\emptyset$).


\section{The Pres operator} \label{pres_op}

\begin{figure}[tb]
  \hrule
  \medskip
  \begin{center}
    \includegraphics[scale=.6]{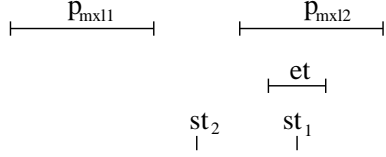}
    \caption{\qit{Is BA737 at gate 2?}}
    \label{pres_op_fig}
  \end{center}
  \hrule
\end{figure}

The \pres operator is used to express the simple present and present
continuous tenses. For $\phi \in \ynforms$:
\begin{itemize}
\item 
\index{pres@$\pres[\;]$ (used to refer to the present)}
$\denot{st,et,lt,g}{\pres[\phi]} = T$, iff $st \in et$ and
  $\denot{st,et,lt,g}{\phi} = T$. 
\end{itemize}
\pref{presop:1}, for example, is represented as \pref{presop:2}.
\begin{examps}
\item Is BA737 at gate 2? \label{presop:1}
\item $\pres[be\_at(ba737, gate2)]$ \label{presop:2}
\end{examps}
Let us assume that the only maximal periods where BA737
was/is/will be at gate 2 are $p_{mxl_1}$ and $p_{mxl_2}$ (i.e.\ 
\pref{presop:3} holds; see section \ref{top_model}), and that \pref{presop:1}
is submitted at a time-point $st_1$, such that 
\pref{presop:4} holds (figure \ref{pres_op_fig}). 
\begin{gather}
\fpfuns(be\_at, 2)(\fcons(ba737), \fcons(gate2)) =
   \{p_{mxl_1}, p_{mxl_2}\}  \label{presop:3}  \\
st_1 \in p_{mxl_2}  \label{presop:4} 
\end{gather}

The answer to \pref{presop:1} will be affirmative iff \pref{presop:5}
is $T$.
\begin{equation}
\denot{st_1}{\pres[be\_at(ba737, gate2)]} \label{presop:5}
\end{equation}
According to section \ref{denotation}, \pref{presop:5} is $T$ iff for
some $g \in G$ and $et \in \periods$, \pref{presop:6} holds.
\begin{equation}
\denot{st_1, et, \pts, g}{\pres[be\_at(ba737, gate2)]} = T \label{presop:6}
\end{equation}
By the definition of \pres, \pref{presop:6}
holds iff both \pref{presop:7} and \pref{presop:8} hold.
\begin{gather}
st_1 \in et \label{presop:7} \\
\denot{st_1,et,\pts,g}{be\_at(ba737, gate2)} = T \label{presop:8}
\end{gather}
By the definitions of $\denot{st,et,lt,g}{\pi(\tau_1, \dots, \tau_n)}$
and $\denot{g}{\kappa}$ (section \ref{denotation}), \pref{presop:8} holds
iff for some $p_{mxl}$, \pref{presop:9} -- \pref{presop:11} hold.
\begin{gather}
et \subper \pts \label{presop:9} \\
p_{mxl} \in \fpfuns(be\_at, 2)(\fcons(ba737), \fcons(gate2))
   \label{presop:10} \\
et \subper p_{mxl} \label{presop:11}
\end{gather}
By \pref{presop:3}, \pref{presop:10} is equivalent to \pref{presop:13}.
\begin{gather}
p_{mxl} \in \{p_{mxl_1}, p_{mxl_2}\} \label{presop:13}
\end{gather}
The answer to \pref{presop:1} will be affirmative iff for some $et
\in \periods$ and some $p_{mxl}$, \pref{presop:7},
\pref{presop:9}, \pref{presop:11}, and \pref{presop:13} hold.  For
$p_{mxl} = p_{mxl_2}$, and $et$ any subperiod of $p_{mxl_2}$ that
contains $st_1$ (figure \ref{pres_op_fig}), \pref{presop:7},
\pref{presop:9}, \pref{presop:11}, and \pref{presop:13} hold. Hence,
the answer to \pref{presop:1} will be affirmative, as one would
expect. If the question is submitted at an 
$st_2$ that falls outside $p_{mxl_1}$ and $p_{mxl_2}$
(figure \ref{pres_op_fig}), then the answer will be negative,
because in that case there is no subperiod $et$ of $p_{mxl_1}$ or
$p_{mxl_2}$ that contains $st_2$. 

The present continuous is expressed similarly. For example, the
reading of \pref{presop:14} where Airserve is actually
servicing BA737 at the present moment is expressed as
\pref{presop:15}. Unlike \cite{Dowty1977}, 
\cite{Lascarides}, \cite{Pirie1990}, and \cite{Crouch2}, 
in \topl progressive tenses do not introduce any special
progressive operator. This will be discussed in section \ref{culm_op}.
\begin{examps}
\item Airserve is (actually) servicing BA737. \label{presop:14}
\item $\pres[servicing(airserve, ba737)]$ \label{presop:15}
\end{examps}
The habitual \pref{presop:16} is represented using a different
predicate functor from that of \pref{presop:14}, as in
\pref{presop:17}. As will be explained in chapter
\ref{English_to_TOP}, \pref{presop:14} is taken to involve a
non-habitual homonym of \qit{to service}, while \pref{presop:16} is
taken to involve a habitual homonym. The two homonyms introduce
different predicate functors.
\begin{examps}
\item Airserve (habitually) services BA737. \label{presop:16}
\item $\pres[hab\_server\_of(airserve, ba737)]$ \label{presop:17}
\end{examps}
\topl's \pres operator is similar to that of 
\cite{Pirie1990}. The main difference is that the \pres
of Pirie et al.\ does not require $st$ to fall within
$et$. Instead, it narrows $lt$ to start at or after $st$. This,
in combination with the requirement $et \subper lt$, requires $et$ to
start at or after $st$. Using this version of \pres in \pref{presop:2}
would cause the answer to be affirmative if \pref{presop:1} is
submitted at $st_2$ (figure \ref{pres_op_fig}), i.e.\ at a point
where BA737 is not at gate 2, because there is an $et$
at which BA737 is at gate 2 (e.g.\ the $et$ of figure
\ref{pres_op_fig}), and this $et$ starts after $st_2$. This version of
\pres was adopted by Pirie et al.\ to cope with sentences
like \qit{J.Adams inspects BA737 tomorrow.}, where the simple present
refers to a future inspection (section \ref{simple_present}). In
this case, $et$ (inspection time) must be allowed to start after $st$.

The \pres of Pirie et al.\ is often over-permissive (e.g.\ it causes
the answer to be affirmative if \pref{presop:1} is submitted at
$st_2$). Pirie et al.\ employ a post-processing mechanism, which is
invoked after the English sentence is translated into logic, and which
attempts to restrict the semantics of \pres in cases where it is
over-permissive. In effect, this mechanism introduces modifications in
only one case: if the \pres is introduced by a state verb (excluding
progressive states) which is not modified by a temporal adverbial,
then $et$ is set to $\{st\}$. For example, in \qit{J.Adams is at site
2.} where the verb is a state, the mechanism causes $et$ to be set to
$\{st\}$, which correctly requires J.Adams to be at gate 2 at $st$.
In \qit{J.Adams is at site 2 tomorrow.}, where the state verb is
modified by a temporal adverbial, the post-processing has no effect,
and $et$ (the time where J.Adams is at site 2) is allowed to start at
or after $st$. This is again correct, since in this case $et$ must be
located within the following day, i.e.\ after $st$. In \qit{J.Adams is
inspecting site 2.}, where the verb is a progressive state, the
post-processing has again no effect, and $et$ (inspection time) can
start at or after $st$. The rationale in this case is that $et$ cannot
be set to $\{st\}$, because there is a reading where the present
continuous refers to a future inspection (section
\ref{progressives}). For the purposes of this project, where the
futurate readings of the simple present and the present continuous are
ignored, \topl's \pres is adequate. If, however, these futurate
readings were to be supported, a more permissive \pres operator, like
that of Pirie et al., might have to be adopted.


\section{The Past operator} \label{past_op}

The \past operator is used when expressing the simple past, the past
continuous, the past perfect, and the present perfect (the latter is
treated as equivalent to the simple past; section
\ref{present_perfect}). For $\phi \in \ynforms$ and $\beta \in \vars$:
\begin{itemize}
\item 
\index{past@$\past[\;]$ (used to refer to the past)}
$\denot{st,et,lt,g}{\past[\beta, \phi]} = T$, iff 
  $g(\beta) = et$ and 
  $\denot{st,et, lt \intersect [t_{first}, st), g}{\phi} = T$.
\end{itemize}
The \past operator narrows the localisation time, so that the latter
ends before $st$. $et$ will eventually be required to be a subperiod
of the localisation time (this requirement will be introduced by the
rules that compute the denotation of $\phi$). Hence, $et$ will be
required to end before $st$.  $\beta$ is used as a pointer to $et$
(the definition of $\past[\beta, \phi]$ makes sure that the value of
$\beta$ is $et$). $\beta$ is useful in formulae that contain
\ntense{s} (to be discussed in section \ref{ntense_op}). It is also
useful in time-asking questions, where $et$ has to be reported. For
example, \qit{When was gate 2 open?}  is represented as $?_{mxl}e^v \;
\past[e^v, open(gate2)]$, which reports the maximal $et$s that end
before $st$, such that gate 2 is open throughout $et$.

\topl's \past operator is essentially the same as that of
\cite{Pirie1990}. (A slightly different \past operator is adopted in
\cite{Crouch2}.) 


\section{Progressives, non-progressives, and the Culm operator} \label{culm_op}

\begin{figure}[tb]
  \hrule
  \medskip
  \begin{center}
    \includegraphics[scale=.6]{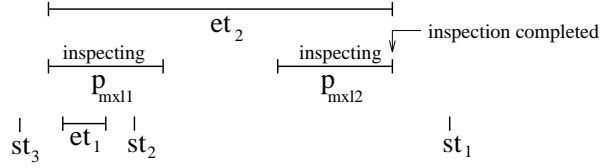}
    \caption{\qit{Was J.Adams inspecting BA737?} vs.\ 
             \qit{Did J.Adams inspect BA737?}}
    \label{culm_op_fig}
  \end{center}
  \hrule
\end{figure}

Let us now examine in more detail how \topl represents the simple past
and the past continuous. Let us start
from verbs whose base forms are culminating activities,
like \qit{to inspect} in the airport domain. The past continuous
\pref{culmop:1} is represented as \pref{culmop:2}.
\begin{examps}
\item Was J.Adams inspecting BA737? \label{culmop:1}
\item $\past[e^v, inspecting(j\_adams, ba737)]$ \label{culmop:2}
\end{examps}
Let us assume that the inspection of BA737 by J.Adams started at the
beginning of $p_{mxl_1}$ (figure \ref{culm_op_fig}), that it
stopped temporarily at the end of $p_{mxl_1}$, that it was resumed at
the beginning of $p_{mxl_2}$, and that it was completed at the end of
$p_{mxl_2}$. Let us also assume that there is no other
time at which J.Adams was/is/will be inspecting BA737. Then,
\pref{culmop:3} and \pref{culmop:3.2} hold.
\begin{gather}
\fpfuns(inspecting, 2)(\fcons(j\_adams), \fcons(ba737)) = 
  \{p_{mxl_1},p_{mxl_2}\} \label{culmop:3} \\
\fculms(inspecting, 2)(\fcons(j\_adams), \fcons(ba737)) = T \label{culmop:3.2}
\end{gather}

The reader can check that \pref{culmop:4} is $T$ iff there is an $et$
that is a subperiod of $p_{mxl_1}$ or $p_{mxl_2}$, and that ends
before $st$.
\begin{equation}
\label{culmop:4}
\denot{st}{\past[e^v, inspecting(j\_adams, ba737)]}
\end{equation}
If \pref{culmop:1} is submitted at
$st_1$ or $st_2$ (figure \ref{culm_op_fig}), then \pref{culmop:4}
is $T$ (the answer to \pref{culmop:1} will be \qit{yes}),
because in both cases there is an $et$ (e.g.\ the $et_1$ of figure
\ref{culm_op_fig}) that ends before $st_1$ and $st_2$, and that is a
subperiod of $p_{mxl_1}$. In contrast, if the question is submitted at
$st_3$, \pref{culmop:4} is $F$ (the answer will be
negative), because in this case there is no subperiod of $p_{mxl_1}$ or
$p_{mxl_2}$ that ends before $st_3$. This is what one would expect: at
$st_1$ and $st_2$ the answer to \pref{culmop:1} should be affirmative,
because J.Adams has already spent some time inspecting BA737. In
contrast, at $st_3$ J.Adams has not yet spent any time inspecting
BA737, and the answer should be negative. 

Let us now consider the simple past \pref{culmop:5}. We want the
answer to be affirmative if \pref{culmop:5}  is
submitted at $st_1$ (or any other time-point after the end of
$p_{mxl_2}$), but not if it is submitted at $st_2$ (or any other
time-point before the end of $p_{mxl_2}$), because at
$st_2$ J.Adams has not yet completed the inspection (section
\ref{simple_past}).
\begin{examps}
\item Did J.Adams inspect BA737? \label{culmop:5}
\end{examps}
\pref{culmop:5} cannot be represented as \pref{culmop:2}, because this
would cause the answer to \pref{culmop:5} to be affirmative if the
question is submitted at $st_2$. Instead, \pref{culmop:5} is
represented as \pref{culmop:6}. The same predicate
$inspecting(j\_adams, ba737)$ of \pref{culmop:2} is used, but an
additional \culm operator is inserted.
\begin{equation}
\label{culmop:6}
\past[e^v, \culm[inspecting(j\_adams, ba737)]]
\end{equation}
Intuitively, the \culm requires the event time to be the $et_2$
of figure \ref{culm_op_fig}, i.e.\ to cover the whole time from
the point where the inspection starts to the point where the 
inspection is completed. (If the inspection is never completed, 
\culm causes the denotation of \pref{culmop:6} to be $F$.)
Combined with the \past operator, the \culm causes the answer
to be affirmative if \pref{culmop:5} is submitted at
$st_1$ (because $et_2$ ends before $st_1$), and negative if the
question is submitted at $st_2$ (because $et_2$ does not end before
$st_2$). More formally, for $\pi \in \pfuns$ and $\tau_1, \dots,
\tau_n \in \terms$: 
\begin{itemize}
\item 
\index{culm@$\culm[\;]$ (used to express non-progressives of culminating activity verbs)}
$\denot{st,et,lt,g}{\culm[\pi(\tau_1, \dots, \tau_n)]} = T$, iff
$et \subper lt$, $\fculms(\pi,n)(\denot{g}{\tau_1}, \dots,
\denot{g}{\tau_n}) = T$, $S \not= \emptyset$, and $et = [minpt(S),
maxpt(S)]$, where: 
\[
S = \bigcup_{p \in \fpfuns(\pi, n)(\denot{g}{\tau_1}, \dots,
             \denot{g}{\tau_n})}p
\]
\end{itemize}
The $et = [minpt(S), maxpt(S)]$ 
requires $et$ to start at the first time-point where the change or
action of $\pi(\tau_1, \dots, \tau_n)$ is ongoing, and to
end at the latest time-point where the change or action is
ongoing. The $\fculms(\pi)(\denot{g}{\tau_1}, \dots,
\denot{g}{\tau_n}) = T$ means that the change or action must
reach its climax at the latest time-point where it
is ongoing.

Let us now check formally that the denotation \pref{culmop:10} of
\pref{culmop:6} is in order.
\begin{equation}
\label{culmop:10}
\denot{st}{\past[e^v, \culm[inspecting(j\_adams, ba737)]]}
\end{equation}
\pref{culmop:10} is $T$ iff for some $g \in G$ and
$et \in \periods$, \pref{culmop:11} holds.
\begin{equation}
\label{culmop:11}
\denot{st,et,\pts,g}{\past[e^v, \culm[inspecting(j\_adams, ba737)]]} = T
\end{equation}
By the definition of \past, \pref{culmop:11} holds iff
\pref{culmop:12} and \pref{culmop:13} hold ($\pts \intersect
[t_{first}, st) = [t_{first}, st)$).
\begin{gather}
g(e^v) = et \label{culmop:12} \\
\denot{st,et, [t_{first}, st), g}{\culm[inspecting(j\_adams, ba737)]} = T
\label{culmop:13}
\end{gather}
By the definition of \culm, \pref{culmop:13} holds iff
\pref{culmop:14} -- \pref{culmop:18} hold. 
\begin{gather}
et \subper [t_{first}, st) \label{culmop:14} \\
\fculms(inspecting, 2)(\fcons(j\_adams), \fcons(ba737)) = T
   \label{culmop:15} \\
S \not= \emptyset \label{culmop:17} \\
et = [minpt(S), maxpt(S)] \label{culmop:16} \\
S = \bigcup_{p \in \fpfuns(inspecting, 2)(\fcons(j\_adams), \fcons(ba737))}p
   \label{culmop:18}
\end{gather}
By \pref{culmop:3}, and assuming that $maxpt(p_{mxl_1}) \prec
minpt(p_{mxl_2})$ (as in figure \ref{culm_op_fig}), \pref{culmop:17}
-- \pref{culmop:18} are equivalent to \pref{culmop:19} --
\pref{culmop:20}. \pref{culmop:19} holds (the union of two periods is
never the empty set), and \pref{culmop:15} is the same as
\pref{culmop:3.2}, which was assumed to hold.
\begin{gather}
p_{mxl_1} \union p_{mxl_2} \not= \emptyset \label{culmop:19} \\
et = [minpt(p_{mxl_1}), maxpt(p_{mxl_2})] \label{culmop:20}
\end{gather}
Hence, \pref{culmop:10} is $T$ (i.e.\ the answer to
\pref{culmop:5} is affirmative) iff for some $g \in G$ and $et
\in \pts$, \pref{culmop:12}, \pref{culmop:14}, and \pref{culmop:20}
hold. Let $et_2 = [minpt(p_{mxl_1}), maxpt(p_{mxl_2})]$ (as in
figure \ref{culm_op_fig}). 

Let us assume that \pref{culmop:5} is submitted at an $st$ that
follows the end of $et_2$ (e.g.\ $st_1$ in figure
\ref{culm_op_fig}). For $et = et_2$,
\pref{culmop:14} and \pref{culmop:20} are satisfied. \pref{culmop:12}
is also satisfied by choosing $g = g_1$, where $g_1$ as below. Hence, 
the answer to \pref{culmop:5} will be affirmative, as required.
\[g_1(\beta) = 
\begin{cases}
et_2 & \text{if } \beta = e^v \\
o & \text{otherwise ($o$ is an arbitrary element of \objs)}
\end{cases}
\]
In contrast, if the question is submitted before the end of $et_2$
(e.g.\ $st_2$ or $st_3$ in figure \ref{culm_op_fig}), then the answer
to \pref{culmop:5} will be negative, because there is no $et$ that
satisfies \pref{culmop:14} and \pref{culmop:20}.

In the case of verbs whose base forms are processes, states, or
points, the simple past does not introduce a \culm operator. In this
case, when both the simple past and the past continuous are possible,
they are represented using the same \topl formula. (A  similar
approach is adopted in \cite{Parsons1989}.) For example, in the
airport domain where \qit{to circle} is classified as process, both
\pref{culmop:28} and \pref{culmop:29} are represented as \pref{culmop:30}. 
\begin{examps}
\item Was BA737 circling? \label{culmop:28}
\item Did BA737 circle? \label{culmop:29}
\item $\past[e^v, circling(ba737)]$ \label{culmop:30}
\end{examps}
The reader can check that the denotation of
\pref{culmop:30} w.r.t.\ $st$ is $T$ (i.e.\ the answer to 
\pref{culmop:28} and \pref{culmop:29} is affirmative) iff there
is an $et$ which is a subperiod of a maximal period where BA737 was
circling, and $et$ ends before $st$. That is, the answer is
affirmative iff BA737 was circling at some time before $st$.
There is no requirement that any climax must
have been reached. 

The reader will have noticed that in the case of verbs whose base
forms are culminating activities, the (non-progressive) simple past is
represented by adding a \culm operator to the expression that
represents the (progressive) past continuous. For example, assuming
that \qit{to build (something)} is a culminating activity,
\pref{culmop:36} is represented as \pref{culmop:37}, and
\pref{culmop:38} as \pref{culmop:39}).
\begin{examps}
\item Housecorp was building bridge 2. \label{culmop:36}
\item $\past[e^v, building(housecorp, bridge2)]$ \label{culmop:37}
\item Housecorp built bridge 2. \label{culmop:38}
\item $\past[e^v, \culm[building(housecorp, bridge2)]]$ \label{culmop:39}
\end{examps}
In contrast, in \cite{Dowty1977}, \cite{Lascarides}, \cite{Pirie1990},
\cite{Crouch2}, and \cite{Kamp1993}, progressive tenses are
represented by adding a progressive operator to the expressions that
represent the non-progressive tenses. For example, ignoring some
details, Pirie et al.\ represent \pref{culmop:36} and \pref{culmop:38}
as \pref{culmop:44} and \pref{culmop:46} respectively.
\begin{examps}
\item $\past[e^v, \progop[build(housecorp, bridge2)]]$ \label{culmop:44}
\item $\past[e^v, build(housecorp, bridge2)]$ \label{culmop:46}
\end{examps}
In \pref{culmop:46}, the semantics that Pirie et al.\ assign to
$build(housecorp, bridge2)$ require $et$ to cover the whole
building of the bridge by Housecorp, from its beginning to the point
where the building is complete. (The semantics of
\topl's $building(housecorp, bridge2)$ in \pref{culmop:37} 
require $et$ to be simply a period throughout which Housecorp is
building bridge 2.) The \past of \pref{culmop:46} requires $et$ (start
to completion of inspection) to end before $st$. Hence, the answer to
\pref{culmop:38} is affirmative iff the building was completed before
$st$.

\begin{figure}[tb]
  \hrule
  \medskip
  \begin{center}
    \includegraphics[scale=.6]{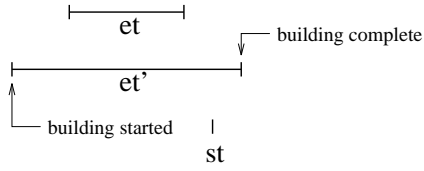}
    \caption{A flawed Prog operator}
    \label{prog_op_fig}
  \end{center}
  \hrule
\end{figure}

In \pref{culmop:44}, the semantics that Pirie et al.\ assign to 
\progop require $et$ to be a subperiod of another period
$et'$ that covers the whole building (from start to completion; 
figure \ref{prog_op_fig}). The \past of \pref{culmop:44}
requires $et$ to end before $st$. If, for
example, \pref{culmop:36} is submitted at an $st$ that falls between
the end of $et$ and the end of $et'$ (figure
\ref{prog_op_fig}), the answer will be affirmative. This is
correct, because at that $st$ Housecorp has already been building the
bridge for some time (although the bridge is not yet complete).
The \progop of Pirie et al., however, has a flaw (acknowledged in
\cite{Crouch2}): \pref{culmop:44} implies that there is a period
$et'$, such that the building is completed at the end of $et'$; i.e.\
according to \pref{culmop:44} the building was or will be necessarily
completed at some time-point. This does not capture correctly the
semantics of \pref{culmop:36}. \pref{culmop:36} carries no
implication that the building was or will ever be
completed. (\topl's representation of \pref{culmop:36}, i.e.\
\pref{culmop:37}, does not suffer from this problem: it contains
no assumption that the building is ever completed.) To overcome
similar problems with \progop operators, ``branching'' models
of time or ``possible worlds'' have been employed (see, for
example, \cite{Dowty1977}, \cite{McDermott1982}, \cite{Mays1986},
\cite{Kent}; see also \cite{Lascarides} for criticism of
possible-worlds approaches to progressives.) Approaches based on
branching time and possible worlds, however, seem unnecessarily
complicated for the purposes of this thesis.


\section{The At, Before, and After operators} \label{at_before_after_op}

The \at, \before, and \after operators are used to express punctual
adverbials, period adverbials, and \qit{while~\dots},
\qit{before~\dots}, and \qit{after~\dots} subordinate clauses (sections \ref{temporal_adverbials} and \ref{subordinate_clauses}).
For $\phi, \phi_1, \phi_2 \in \ynforms$ and $\tau \in \terms$:
\begin{itemize}
\item 
\index{at@$\at[\;]$ (narrows the localisation time)}
   $\denot{st,et,lt,g}{\at[\tau, \phi]} = T$, iff 
   $\denot{g}{\tau} \in \periods$ and 
   $\denot{st,et,lt \intersect \denot{g}{\tau},g}{\phi} = T$.
\item 
\index{at@$\at[\;]$ (narrows the localisation time)}
$\denot{st,et,lt,g}{\at[\phi_1, \phi_2]} = T$, iff
   for some $et'$ \\
   $et' \in mxlpers(\{e \in \periods \mid \denot{st,e,\pts,g}{\phi_1} = T\})$ 
   and 
   $\denot{st,et,lt \intersect et',g}{\phi_2} = T$.
\end{itemize}
If the first argument of \at is a term $\tau$, then
$\tau$ must denote a period. The localisation time is
narrowed to the intersection of the original $lt$ with the period
of $\tau$. If the first argument of \at is a formula $\phi_1$,
the localisation time of $\phi_2$ is narrowed to the
intersection of the original $lt$ with a maximal event time period
$et'$ at which $\phi_1$ holds.  For example, \pref{atop:1} is
represented as \pref{atop:2}.
\begin{examps}
\item Was tank 2 empty (some time) on 25/9/95? \label{atop:1}
\item $\at[\mathit{25/9/95}, \past[e^v, empty(tank2)]]$ \label{atop:2}
\end{examps}
In \pref{atop:2}, $lt$ initially covers the whole time-axis. The \at
operator causes $lt$ to become the 25/9/95 period (I assume that the
constant $\mathit{25/9/95}$ denotes the obvious period), and the \past
operator narrows $lt$ to end before $st$ (if 25/9/95 is entirely in
the past, the \past operator has not effect). The answer to
\pref{atop:1} is affirmative iff it is possible to find an $et$ that is
a subperiod of the narrowed $lt$, such that tank 2 was empty during
$et$. 

If \pref{atop:1} is submitted before 25/9/95 (i.e.\ 25/9/95 starts
after $st$), the \nlitdb's answer will be negative, because the \at
and \past operators cause $lt$ to become the empty set, and hence it
is impossible to find a subperiod $et$ of $lt$ where tank 2 is empty. A
simple negative response is unsatisfactory in this case: \pref{atop:1}
is unacceptable if uttered before 25/9/95, and the system should warn
the user about this. The unacceptability of \pref{atop:1} in this case
seems related to the unacceptability of \pref{atop:20}, which would be
represented as \pref{atop:21} (the definition of \partop would have to
be extended to allow positive values of its third argument; see
section \ref{denotation}). 
\begin{examps}
\item \bad Was tank 2 empty tomorrow? \label{atop:20}
\item $\partop[day^c, tom^v, 1] \land \at[tom^v, \past[e^v, empty(tank2)]]$ 
   \label{atop:21}
\end{examps}
In both cases, the combination of the simple past and the adverbial
causes $lt$ to become the empty set. In \pref{atop:20}, $tom^v$
denotes the period that covers exactly the day after $st$. The \at and
\past operators set $lt$ to the intersection of that period with
$[t_{first}, st)$. The two periods do not overlap, and hence $lt =
\emptyset$, and it is impossible to find a subperiod $et$ of $lt$.
This causes the answer to be always negative, no matter what happens
in the world (i.e.\ regardless of when tank 2 is empty).  Perhaps the
questions sound unacceptable because people, using a concept similar
to \topl's $lt$, realise that the answers can never be affirmative.
This suggests that the \nlitdb should check if $lt = \emptyset$, and
if this is the case, generate a cooperative response (section
\ref{no_issues}) explaining that the question is problematic (this is
similar to the ``overlap rule'' of \cite{Harper} and the
``non-triviality constraint'' on p.~653 of \cite{Kamp1993}). The
framework of this thesis currently provides no such mechanism.

Moving to further examples, \pref{atop:3} and \pref{atop:22} are
represented as \pref{atop:4} and \pref{atop:23}.
Unlike the \qit{on 25/9/95} of
\pref{atop:1}, which is represented using a constant
($\mathit{25/9/95}$), the \qit{on Monday} of \pref{atop:3} is
represented using a variable ($mon^v$) that ranges over the periods of
the partitioning of Monday-periods.  Similarly,
the \qit{at 5:00pm} of \pref{atop:22} is represented using a variable
($fv^v$) that ranges over the 5:00pm minute-periods.
\begin{examps}
\item Was tank 2 empty on Monday? \label{atop:3} 
\item Was tank 2 empty on a Monday? \label{atop:24}
\item $\partop[monday^g, mon^v] \land \at[mon^v, \past[e^v, empty(tank2)]]$ 
   \label{atop:4}
\item Was tank 2 empty on Monday at 5:00pm? \label{atop:22}
\item $\partop[monday^g, mon^v] \land
\partop[\text{\textit{5:00pm}}^g, fv^v] \; \land$ \\
      $\at[mon^v, \at[fv^v, \past[e^v, empty(tank2)]]]$ \label{atop:23}
\end{examps}
\pref{atop:4} requires tank 2 to have been empty at some
past $et$ that falls within some Monday. No attempt is made to
determine exactly which Monday the user has in mind in \pref{atop:3}
(\pref{atop:3} is treated as equivalent to \pref{atop:24}; section
\ref{temporal_anaphora}). Similarly, \pref{atop:23} requires tank 2 to
have been empty at some past $et$ that falls within the intersection
of some 5:00pm-period with some Monday-period.

Assuming that \qit{to inspect} is a culminating activity (as in the
airport application), the reading of \pref{atop:28} that requires the
inspection to have both started and been completed within the previous
day (section \ref{period_adverbials}) is represented as
\pref{atop:29}. The \culm requires $et$ to cover exactly the whole
inspection, from its beginning to its completion. The \past requires
$et$ to end before $st$, and the \at requires $et$ to fall within the
day before $st$.
\begin{examps}
\item Did J.Adams inspect BA737 yesterday? \label{atop:28}
\item $\partop[day^c, y^v, -1] \land 
   \at[y^v, \past[e^v, \culm[inspecting(j\_adams, ba737)]]]$ \label{atop:29}
\end{examps}
In contrast, \pref{atop:26} is represented as \pref{atop:27}. In this
case, $et$ must be simply a subperiod of a maximal period where 
J.Adams was inspecting BA737, and also be located 
within the previous day. 
\begin{examps}
\item Was J.Adams inspecting BA737 yesterday? \label{atop:26}
\item $\partop[day^c, y^v, -1] \land 
   \at[y^v, \past[e^v, inspecting(j\_adams, ba737)]]$ \label{atop:27}
\end{examps}
Finally, \pref{atop:30} is represented as \pref{atop:31}, which
intuitively requires BA737 to have been circling at some past period
$e2^v$, that falls within some past maximal period $e1^v$ where
gate 2 was open.
\begin{examps}
\item Did BA737 circle while gate 2 was open? \label{atop:30}
\item $\at[\past[e1^v, open(gate2)], \past[e2^v, circling(ba737)]]$
   \label{atop:31}
\end{examps}

The \before and \after operators are similar. They are used to express
adverbials and subordinate clauses introduced by \qit{before} and
\qit{after}. For $\phi, \phi_1, \phi_2 \in \ynforms$ and $\tau \in \terms$:
\begin{itemize}
\item 
\index{before@$\before[\;]$ (used to express \qit{before})}
   $\denot{st,et,lt,g}{\before[\tau, \phi]} = T$, iff 
   $\denot{g}{\tau} \in \periods$ and \\
   $\denot{st,et,
           lt \intersect [t_{first}, minpt(\denot{g}{\tau})),
           g}{\phi} = T$.
\item 
\index{before@$\before[\;]$ (used to express \qit{before})}
   $\denot{st,et,lt,g}{\before[\phi_1, \phi_2]} = T$, iff
   for some $et'$ \\
   $et' \in mxlpers(\{e \in \periods \mid \denot{st,e,\pts,g}{\phi_1} = T\})$, 
   and \\
   $\denot{st,et,
           lt \intersect [t_{first}, minpt(et')),
           g}{\phi_2} = T$.
\item 
\index{after@$\after[\;]$ (used to express \qit{after})}
   $\denot{st,et,lt,g}{\after[\tau, \phi]} = T$, iff 
   $\denot{g}{\tau} \in \periods$ and \\
   $\denot{st,et,
           lt \intersect (maxpt(\denot{g}{\tau}), t_{last}],
           g}{\phi} = T$.
\item 
\index{after@$\after[\;]$ (used to express \qit{after})}
   $\denot{st,et,lt,g}{\after[\phi_1, \phi_2]} = T$, iff
   for some $et'$ \\
   $et' \in mxlpers(\{e \mid \denot{st,e,\pts,g}{\phi_1} = T\})$
   and \\
   $\denot{st,et,
           lt \intersect (maxpt(et'), t_{last}],
           g}{\phi_2} = T$.
\end{itemize}
If the first argument of \before is a term
$\tau$, $\tau$ must denote a period. The localisation time 
is required to end before the beginning of $\tau$'s period. If the
first argument of \before is a formula $\phi_1$, the localisation time
of $\phi_2$ is required to end before the beginning of a maximal event
time period $et'$ where $\phi_1$ holds. The \after operator is similar.

For example, \pref{atop:35} is expressed as \pref{atop:36}, and the
reading of \pref{atop:37} that requires BA737 to have departed after
the \emph{end} of a maximal period where the emergency system was in
operation is expressed as \pref{atop:38}. (I assume here that \qit{to
depart} is a point, as in the airport application.)
\begin{examps}
\item Was tank 2 empty before 25/9/95? \label{atop:35}
\item $\before[\mathit{25/9/95}, \past[e^v, empty(tank2)]]$ \label{atop:36}
\item BA737 departed after the emergency system was in
   operation. \label{atop:37} 
\item $\after[\past[e1^v, in\_operation(emerg\_sys)], 
              \past[e2^v, depart(ba737)]]$ \label{atop:38}
\end{examps}
\pref{atop:37} also has a reading where BA737 must have departed after
the emergency system \emph{started} to be in operation (section
\ref{before_after_clauses}). To express this reading, we need the
\lbegin operator of section \ref{begin_end_op} below.

\topl's \at, \before, and \after operators are similar to those of 
\cite{Pirie1990}. The operators of Pirie et
al., however, do not narrow $lt$ as in
\topl. Instead, they place directly restrictions on $et$. For example,
ignoring some details, the $\after[\phi_1, \phi_2]$ 
of Pirie et al.\ requires $\phi_2$ to hold at an event time $et_2$
that follows an $et_1$ where $\phi_1$ holds (both
$et_1$ and $et_2$ must fall within $lt$). Instead, \topl's
$\after[\phi_1, \phi_2]$ requires $et_1$ to be a maximal period where
$\phi_1$ holds ($et_1$ does not need to fall within the original $lt$), and
evaluates $\phi_2$ with respect to a narrowed $lt$, which is the
intersection of the original $lt$ with $et_1$. In most cases, both
approaches lead to similar results. \topl's approach, however, is
advantageous in sentences like \pref{atop:60}, where one may want to
express the reading whereby the tank was empty \emph{throughout}
26/9/95 (section \ref{period_adverbials}).
\begin{examps}
\item Tank 2 was empty on 26/9/95. \label{atop:60}
\item $\at[\mathit{26/9/95}, \past[e^v, empty(tank2)]]$ \label{atop:61}
\end{examps}
In these cases one
wants $et$ (time where the tank was empty)
to cover all the available time, where by ``available time'' I mean
the part of the time-axis where the tense and the 
adverbial allow $et$ to be placed. This notion of ``available
time'' is captured by \topl's $lt$: the simple past and
the \qit{on 26/9/95} of \pref{atop:60} introduce \at and
\past operators that, assuming that \pref{atop:60} is submitted after
26/9/95, cause $lt$ to become the period that covers exactly the
day 26/9/95. The intended reading can be expressed easily in \topl by
including an additional operator that forces $et$ to cover the whole
$lt$ (this operator will be discussed in section \ref{fills_op}). This
method cannot be used in the language of Pirie et al. Their \past
operator narrows the $lt$ to the part of the time-axis up to $st$, but
their \at does not narrow $lt$ any further; instead, it imposes a
direct restriction on $et$ (the semantics of Pirie et al.'s \at is not
very clear, but it seems that this restriction requires $et$ to be a
subperiod of 26/9/95). Hence, $lt$ is left to be the time-axis up to
$st$, and one cannot require $et$ to cover the whole $lt$, because
this would require the tank to be empty all the time from $t_{first}$
to $st$.

The \at operator of Pirie et al.\ also does not allow its first
argument to be a formula, and it is unclear how they represent
\qit{while~\dots} clauses. Finally, Pirie et al.'s \before allows
counter-factual uses of \qit{before} to be expressed (section
\ref{before_after_clauses}). Counter-factuals are not considered in this
thesis, and hence Pirie et al.'s \before will not be discussed any
further.


\section{The Fills operator} \label{fills_op}

As discussed in section \ref{period_adverbials}, when states combine
with period adverbials, there is often a reading where the situation
of the verb holds \emph{throughout} the adverbial's period. For
example, there is a reading of \pref{fop:1} where tank 2
was empty throughout 26/9/95, not at simply some part of that day.
\begin{examps}
\item Tank 2 was empty on 26/9/95. \label{fop:1}
\end{examps}
Similar behaviour was observed in cases where states
combine with \qit{while~\dots} subordinate clauses (section
\ref{while_clauses}). For example, there is a reading of
\pref{fop:3} whereby BA737 was at gate 2 throughout the
entire inspection of UK160 by J.Adams, not at simply some time during
the inspection. 
\begin{examps}
\item BA737 was at gate 2 while J.Adams was inspecting UK160. \label{fop:3}
\end{examps}
It is also interesting that \pref{fop:5} cannot be understood as
saying that tank 2 was empty throughout the period of \qit{last
summer}. There is, however, a reading of \pref{fop:5} where tank 2 was
empty throughout the August of the previous summer.
\begin{examps}
\item Tank 2 was empty in August last summer. \label{fop:5}
\end{examps}
It seems that states give rise to readings where the situation of the
verb covers the whole available localisation time.  \pref{fop:2} --
\pref{fop:10} would express the readings of \pref{fop:1} --
\pref{fop:5} that are under discussion, if there
were some way to force the event times of the predicates
$empty(tank2)$, $be\_at(ba737, gate2)$, and $empty(tank2)$ to cover their
whole localisation times. 
\begin{examps}
\item $\at[\mathit{26/9/95}, \past[e^v, empty(tank2)]]$ \label{fop:2}
\item $\at[\past[e1^v, inspecting(j\_adams, uk160)], 
           \past[e2^v, be\_at(ba737, gate2)]]$ \label{fop:4}
\item $\partop[august^g, aug^v] \land \partop[summer^g, sum^v, -1] \; \land$ \\
      $\at[aug^v, \at[sum^v, \past[e^v, empty(tank2)]]]$ \label{fop:10}
\end{examps}
The \fills operator achieves exactly this: it sets $et$ to the whole
of $lt$. For $\phi \in \ynforms$:
\begin{itemize}
\item 
\index{fills@$\fills[\;]$ (requires $et = lt$)}
   $\denot{st,et,lt,g}{\fills[\phi]} = T$, iff $et = lt$ and
   $\denot{st,et,lt,g}{\phi} = T$.  
\end{itemize}
The readings of \pref{fop:1} -- \pref{fop:5} that are under discussion
can be expressed as \pref{fop:2b} -- \pref{fop:11} respectively. 
\begin{examps}
\item $\at[\mathit{26/9/95}, \past[e^v, \fills[empty(tank2)]]]$ \label{fop:2b}
\item $\at[\past[e1^v, inspecting(j\_adams, uk160)],$ \\
      $    \past[e2^v, \fills[be\_at(ba737, gate2)]]]$ \label{fop:4b}
\item $\partop[august^g, aug^v]  \land \partop[summer^g, sum^v, -1]$ \\
      $\at[aug^v, \at[sum^v, \past[e^v, \fills[empty(tank2)]]]]$ \label{fop:11}
\end{examps}
This suggests that when state expressions
combine with period-specifying subordinate clauses or adverbials, the
\nlitdb could generate two formulae, one with and one without a
\fills, to capture the readings where $et$ covers
the whole or just part of $lt$. As mentioned
in section \ref{period_adverbials}, this approach (which was tested in
one version of the prototype \nlitdb) has the disadvantage that it
generates a formula for the reading where $et$ covers the whole $lt$
even in cases where this reading is impossible. In time-asking
questions like \pref{fop:35}, for example, the reading where $et$
covers the whole $lt$ (the whole 1994) is impossible, and hence the
corresponding formula should not be generated. 
\begin{examps}
\item When was tank 5 empty in 1994? \label{fop:35}
\end{examps}
Devising an algorithm to decide when the formulae that contain
\fills should or should not be generated is a task which I have not
addressed. For simplicity, the prototype \nlitdb
and the rest of this thesis ignore the readings that require $et$
to cover the whole $lt$, and hence the \fills
operator is not used. The \fills operator, however, may prove useful
to other researchers who may attempt to explore further the topic of this
section.


\section{The Begin and End operators} \label{begin_end_op}

The \lbegin and \lend operators are used to refer to the time-points
where a situation starts or ends. For $\phi \in \ynforms$:
\begin{itemize}
\item 
\index{begin@$\lbegin[\;]$ (used to refer to start-points of situations)}
   $\denot{st,et,lt,g}{\lbegin[\phi]} = T$, iff $et \subper lt$\\
   $et' \in mxlpers(\{e \in \periods \mid \denot{st,e,\pts,g}{\phi} = T\})$ 
   and $et = \{minpt(et')\}$.
\item 
\index{end@$\lend[\;]$ (used to refer to end-points of situations)}
   $\denot{st,et,lt,g}{\lend[\phi]} = T$, iff $et \subper lt$ \\
   $et' \in mxlpers(\{e \in \periods \mid \denot{st,e,\pts,g}{\phi} = T\})$ 
   and $et = \{maxpt(et')\}$.
\end{itemize}
$\lbegin[\phi]$ is true only at instantaneous event times $et$ that
are beginnings of maximal event times $et'$ where $\phi$
holds. The \lend operator is similar.

The \lbegin and \lend operators can be used to express \qit{to
start}, \qit{to stop}, \qit{to begin}, and \qit{to finish} (section
\ref{special_verbs}). For example, \pref{beop:3} is expressed as
\pref{beop:4}. Intuitively, in \pref{beop:4} the
$\culm[inspecting(j\_adams, uk160)]$ 
refers to an event-time period that covers exactly a complete
inspection of UK160 by J.Adams (from start to
completion). $\lend[\culm[inspecting(j\_adams, uk160)]]$ refers to the
end of that period, i.e.\ the completion point of J.Adams'
inspection. $\lbegin[inspecting(t\_smith, ba737)]$ refers to the
beginning of an inspection of BA737 by T.Smith. The beginning of
T.Smith's inspection must precede the completion point of J.Adams'
inspection, and both points must be in the past.
\begin{examps}
\item Did T.Smith start to inspect BA737 before J.Adams finished
   inspecting UK160? \label{beop:3}
\item $\begin{aligned}[t]
       \before[&\past[e1^v, \lend[\culm[inspecting(j\_adams, uk160)]]], \\
               &\past[e2^v, \lbegin[inspecting(t\_smith, ba737)]]]
       \end{aligned}$
   \label{beop:4}
\end{examps}
The reading of \pref{atop:37} (section \ref{at_before_after_op}) that
requires BA737 to have departed after the emergency system
\emph{started} to be in operation can be expressed as
\pref{beop:6}. (The reading of \pref{atop:37} where BA737 must have
departed after the system \emph{stopped} being in operation is
expressed as \pref{atop:38}.)
\begin{examps}
\item $\after[\past[e1^v, \lbegin[in\_operation(emerg\_sys)]], 
              \past[e2^v, depart(ba737)]]$ \label{beop:6}
\end{examps}


\section{The Ntense operator} \label{ntense_op}

The framework of this thesis (section \ref{noun_anaphora}) allows noun
phrases like \qit{the sales manager} in \pref{ntop:1} to refer either
to the present (current sales manager) or the time of the verb tense
(1991 sales manager). The \ntense operator is used to represent these
two possible readings.  
\begin{examps}
\item What was the salary of the sales manager in 1991?
   \label{ntop:1} 
\item $?slr^v \; \ntense[now^*, manager\_of(mgr^v, sales)] \; \land$ \\
   $\at[1991, \past[e^v, salary\_of(mgr^v, slr^v)]]$ \label{ntop:1.1}
\item $?slr^v \; \ntense[e^v, manager\_of(mgr^v, sales)] \; \land$ \\
   $\at[1991, \past[e^v, salary\_of(mgr^v, slr^v)]]$ \label{ntop:1.2}
\end{examps}
The reading of \pref{ntop:1} where \qit{the sales manager} refers to
the present is represented as \pref{ntop:1.1}, while the reading where
it refers to the time of
the verb tense is represented as \pref{ntop:1.2}. Intuitively,
\pref{ntop:1.1} reports any $slr^v$, such that $slr^v$ was the
salary of $mgr^v$ at some past time $e^v$ that falls within 1991, and
$mgr^v$ is the manager of the sales department \emph{at the present}.
In contrast, \pref{ntop:1.2} reports any $slr^v$, such
that $slr^v$ was the salary of $mgr^v$ at some past time $e^v$ that falls
within 1991, and $mgr^v$ was the manager of the sales department \emph{at}
$e^v$. Notice that in \pref{ntop:1.2} the first
argument of the \ntense is the same as the first argument of the
\past, which is a pointer to the past event time where
$salary\_of(mgr^v, slr^v)$ is true (see the semantics of \past in
section \ref{past_op}). 

For $\phi \in \ynforms$ and $\beta \in \vars$ :
\begin{itemize}
\item 
\index{ntense@$\ntense[\;]$ (used when expressing nouns or adjectives)}
  $\denot{st,et,lt,g}{\ntense[\beta, \phi]} = T$, iff for some
  $et' \in \periods$, it is true that $g(\beta)= et'$ and
  $\denot{st,et',\pts,g}{\phi} = T$. 
\item 
\index{ntense@$\ntense[\;]$ (used when expressing nouns or adjectives)}
  $\denot{st,et,lt,g}{\ntense[now^*, \phi]} = T$, iff 
  $\denot{st,\{st\},\pts,g}{\phi} = T$.
\end{itemize}
\ntense evaluates $\phi$ with
respect to a new event time $et'$, which may be different from the
original event time $et$ that is used to evaluate the part of the
formula outside the \ntense. Within the \ntense, the localisation time
is reset to \pts (whole time-axis) freeing $et'$ from restrictions
imposed on the original $et$. If the first argument of \ntense is
$now^*$, the new event time is the instantaneous period that contains
only $st$, i.e.\ the object to which the noun phrase refers must have
at $st$ the property described by $\phi$. If the first argument of
\ntense is a variable $\beta$, the new event time $et'$ can generally
be any period, and $\beta$ denotes $et'$. In
\pref{ntop:1.2}, however, $\beta$ is the same as the first
argument of the \past, which denotes the original
$et$ that the \past requires to be placed before $st$. This
means that $manager\_of(mgr^v, sales)$
must hold at the same event time where $salary\_of(mgr^v, slr^v)$
holds, i.e.\ the person $mgr^v$ must be the sales manager at the same
time where the salary of $mgr^v$ is $slr^v$. If the first argument of
the \ntense in \pref{ntop:1.2} and the first argument of the \past
were different variables, the answer would contain any 1991 salary of
anybody who was, is, or will be the sales manager at any time.
This would be useful in \pref{ntop:13}, where one may want to allow
\qit{Prime Minister} to refer to the Prime Ministers of all times, a
reading that can be expressed as \pref{ntop:16}.
\begin{examps}
\item Which Prime Ministers were born in Scotland? \label{ntop:13}
\item $?pm^v \; \ntense[e1^v, pminister(pm^v)] \land 
         \past[e2^v, birth\_in(pm^v,scotland)]$ \label{ntop:16}
\end{examps}
The framework of this thesis, however, does not currently generate
\pref{ntop:16}. \pref{ntop:13} would receive only two formulae, one
for current Prime Ministers, and one for persons that were Prime
Ministers at the time they were born (the latter reading is, of
course, unlikely). 

Questions like \pref{ntop:5} and \pref{ntop:7}, where temporal
adjectives specify explicitly the times to which the noun phrases
refer, can be represented as \pref{ntop:6} and \pref{ntop:8}.
(The framework of this thesis, however, does not support
temporal adjectives other than \qit{current}; see section
\ref{temporal_adjectives}.)
\begin{examps}
\item What was the salary of the current sales manager in 1991?
   \label{ntop:5} 
\item $?slr^v \; \ntense[now^*, manager\_of(mgr^v, sales)] \; \land$\\ 
   $\at[1991, \past[e^v, salary\_of(mgr^v, slr^v)]]$ \label{ntop:6}
\item What was the salary of the 1988 sales manager in
   1991? \label{ntop:7} 
\item $?slr^v \; \ntense[e1^v, \at[1988, manager\_of(mgr^v, sales)]]
   \; \land$\\  
   $\at[1991, \past[e^v, salary\_of(mgr^v, slr^v)]]$ \label{ntop:8}
\end{examps}

The \ntense operator of \topl is the same as the \ntense operator of
\cite{Crouch} and \cite{Crouch2}. 


\section{The For operator} \label{for_op}

The \for operator is used to express \qit{for~\dots} and
duration \qit{in~\dots} adverbials (sections \ref{for_adverbials} and
\ref{in_adverbials}). For $\sigma_c \in \cparts$, $\nu_{qty} \in
\{1,2,3,\dots\}$, and $\phi \in \ynforms$:
\begin{itemize}
\item 
\index{for@$\for[\;]$ (used to express durations)}
$\denot{st,et,lt,g}{\for[\sigma_c, \nu_{qty}, \phi]} = T$, iff
$\denot{st,et,lt,g}{\phi} = T$, and for some $p_1,p_2,\dots,p_{\nu_{qty}}
\in \fcparts(\sigma_c)$, it is true that $minpt(p_1) = minpt(et)$,
$next(maxpt(p_1)) = minpt(p_2)$, $next(maxpt(p_2)) = minpt(p_3)$,
\dots, $next(maxpt(p_{\nu_{qty} - 1})) = minpt(p_{\nu_{qty}})$, and
$maxpt(p_{\nu_{qty}}) = maxpt(et)$.
\end{itemize}
$\for[\sigma_c, \nu_{qty}, \phi]$ requires $\phi$ to be true at an event
time period that is $\nu_{qty}$ $\sigma_c$-periods long. For example,
assuming that $month^c$ denotes the partitioning of month-periods (the
period that covers exactly the August of 1995, the period for
September of 1995, etc.), \pref{forop:3} can be expressed as
\pref{forop:4}.
\begin{examps}
\item Was tank 2 empty for three months? \label{forop:3}
\item $\for[month^c, 3, \past[e^v, empty(tank2)]]$ \label{forop:4}
\end{examps}
\pref{forop:4} requires an event time $et$ to exist, such that $et$
covers exactly three continuous months, and tank 2 was empty
throughout $et$. 
As noted in section \ref{for_adverbials}, \qit{for~\dots}
adverbials are sometimes used to specify the duration of a
\emph{maximal} period where a situation holds, or to refer to the
\emph{total duration} of possibly non-overlapping periods where some
situation holds. The current version of \topl cannot express such
readings.

Expressions like \qit{one week}, \qit{three months}, \qit{two years},
\qit{two hours}, etc., are often used to specify a duration of seven
days, $3 \times 30$ days, $2 \times 365$ days, $2 \times 60$ minutes,
etc. \pref{forop:4} expresses \pref{forop:3} if \qit{three months}
refers to \emph{calendar} months (e.g.\ from the beginning of a June
to the end of the following August). If \qit{three months} means $3
\times 30$ days, \pref{forop:10} has to be used instead. (I assume that
$day^c$ denotes the partitioning of day-periods: the period that
covers exactly 26/9/95, the period for 27/9/95, etc.)
\begin{examps}
\item $\for[day^c, 90, \past[e^v, empty(tank2)]]$ \label{forop:10}
\end{examps}
Assuming that \qit{to inspect} is a culminating
activity (as in the airport application), \pref{forop:13} represents the
reading of \pref{forop:12} where 42 minutes is the
duration from the beginning of the inspection to the
inspection's completion (section \ref{in_adverbials}). \pref{forop:13}
requires $et$ to cover the whole inspection (from beginning to
completion), $et$ to be in the past, and the duration of $et$ to be 42
minutes. 
\begin{examps}
\item J.Adams inspected BA737 in 42 minutes. \label{forop:12}
\item $\for[minute^c, 42, \past[e^v, \culm[inspecting(j\_adams, ba737)]]]$ 
   \label{forop:13}
\end{examps}
Unlike \pref{forop:12}, \pref{forop:14} does not require the
inspection to have been completed (section
\ref{for_adverbials}). \pref{forop:14} is represented as 
\pref{forop:15}, which contains no \culm.
In this case, $et$ must simply be a 
period throughout which J.Adams was inspecting BA737, it 
must be located in the past, and it must be 42 minutes long.
\begin{examps}
\item J.Adams inspected BA737 for 42 minutes. \label{forop:14}
\item $\for[minute^c, 42, \past[e^v, inspecting(j\_adams, ba737)]]$ 
   \label{forop:15}
\end{examps}


\section{The Perf operator} \label{perf_op}

\begin{figure}[tb]
  \hrule
  \medskip
  \begin{center}
    \includegraphics[scale=.6]{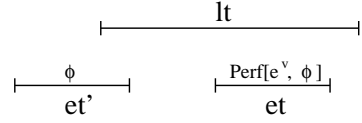}
    \caption{The Perf operator}
    \label{perf_op_fig}
  \end{center}
  \hrule
\end{figure}

The \perf operator is used when expressing the past perfect. For
example, \pref{perfop:3} is expressed as \pref{perfop:4}. \perf could
also be used to express the present perfect (e.g.\ \pref{perfop:1}
could be represented as \pref{perfop:2}). This thesis, however, treats
the present perfect in the same way as the simple past (section
\ref{present_perfect}), and \pref{perfop:1} is mapped to
\pref{perfop:6}, the same formula that expresses \pref{perfop:5}. 
\begin{examps}
\item BA737 had departed. \label{perfop:3}
\item $\past[e1^v, \perf[e2^v, depart(ba737)]]$ \label{perfop:4}
\item BA737 has departed. \label{perfop:1}
\item $\pres[\perf[e^v, depart(ba737)]]$ \label{perfop:2}
\item BA737 departed. \label{perfop:5}
\item $\past[e^v, depart(ba737)]$ \label{perfop:6}
\end{examps}
For $\phi \in \ynforms$ and $\beta \in \vars$:
\begin{itemize}
\item 
\index{perf@$\perf[\;]$ (used to express the past perfect)}
$\denot{st,et,lt,g}{\perf[\beta, \phi]} = T$, iff $et \subper
lt$, and for some $et' \in \periods$, it is true that $g(\beta) =
et'$, $maxpt(et') \prec minpt(et)$, and $\denot{st,et',\pts,g}{\phi}
= T$. 
\end{itemize}
$\perf[\beta, \phi]$ holds at the event time $et$, only if $et$ is
preceded by a new event time $et'$ where $\phi$ holds (figure
\ref{perf_op_fig}). The original $et$ must be a subperiod of $lt$. In
contrast $et'$ does not need to be a subperiod of $lt$ (the
localisation time in $\denot{st,et',\pts,g}{\phi}$ is reset to \pts,
the whole time-axis). The $\beta$ of $\perf[\beta, \phi]$ is a pointer
to $et'$, similar to the $\beta$ of $\past[\beta,\phi]$.
Ignoring constraints imposed by
$lt$, the event time $et$ where $\perf[\beta, \phi]$ is true can be
placed anywhere within the period that starts immediately after the
end of $et'$ ($et'$ is where $\phi$ is true) and that extends up to
$t_{last}$. The informal term ``consequent period'' was used in section
\ref{point_adverbials} to refer to this period.

\begin{figure}[tb]
  \hrule
  \medskip
  \begin{center}
    \includegraphics[scale=.6]{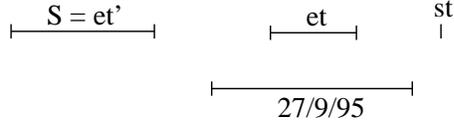}
    \caption{First reading of \qit{J.Adams had inspected gate 2 on
             27/9/95}} 
    \label{perf_op2_fig}
  \end{center}
  \hrule
\end{figure}

Using the \perf operator, the reading of \pref{perfop:7} where
the inspection happens at some time before (or possibly on 27/9/95) is
expressed as \pref{perfop:8} (in this case, \qit{on 27/9/95} provides
a ``reference time''; see section \ref{past_perfect}). In contrast,
the reading of \pref{perfop:7} where the inspection happens on
27/9/95 is expressed as \pref{perfop:9}. 
\begin{examps}
\item J.Adams had inspected gate 2 on 27/9/95. \label{perfop:7}
\item $\at[\mathit{27/9/95}, 
           \past[e1^v, \perf[e2^v, \culm[inspecting(ja, g2)]]]]$ 
   \label{perfop:8}
\item $\past[e1^v, \perf[e2^v, \at[\mathit{27/9/95},
                                   \culm[inspecting(ja, g2)]]]]$
   \label{perfop:9}
\end{examps}
Let us explore formally the denotations of
\pref{perfop:8} and \pref{perfop:9}. The denotation of \pref{perfop:8}
w.r.t.\ $st$ is $T$ iff for some $et \in \periods$ and $g \in G$,
\pref{perfop:10} holds.
\begin{examps}
\item $\denot{st,et,\pts,g}
   {\at[\mathit{27/9/95},
    \past[e1^v, \perf[e2^v, \culm[inspecting(ja, g2)]]]]} = T$
\label{perfop:10}
\end{examps}
Assuming that $\mathit{27/9/95}$ denotes the obvious period, by 
the definition of \at, \pref{perfop:10} holds iff \pref{perfop:11} is
true ($\pts \intersect \fcons(\mathit{27/9/95}) = \fcons(\mathit{27/9/95})$).
\begin{examps}
\item $\denot{st,et,\fcons(\mathit{27/9/95}),g}
      {\past[e1^v,
             \perf[e2^v, \culm[inspecting(ja, g2)]]]} = T$
\label{perfop:11}
\end{examps}
By the definition of \past, ignoring $e1^v$ which
does not play any interesting role here, and assuming that $st$
follows 27/9/95, \pref{perfop:11} is true iff
\pref{perfop:12} holds.
\begin{examps}
\item $\denot{st, et, \fcons(\mathit{27/9/95}), g}
      {\perf[e2^v, \culm[inspecting(ja, g2)]]} = T$
\label{perfop:12}
\end{examps}
By the definition of \perf (ignoring $e2^v$),
\pref{perfop:12} holds iff for some $et' \in \periods$,
\pref{perfop:14}, \pref{perfop:15}, and \pref{perfop:16} hold.
\begin{gather}
et \subper \fcons(\mathit{27/9/95}) \label{perfop:14} \\
maxpt(et') \prec minpt(et) \label{perfop:15} \\
\denot{st,et',\pts,g}{\culm[inspecting(ja, g2)]} = T \label{perfop:16}
\end{gather}
By the definition of \culm, \pref{perfop:16} holds iff
\pref{perfop:17} -- \pref{perfop:21} hold.
\begin{gather}
et' \subper \pts \label{perfop:17} \\
\fculms(inspecting, 2)(\fcons(ja), \fcons(g2)) = T
   \label{perfop:18} \\
S = \bigcup_{p \in \fpfuns(inspecting, 2)(\fcons(ja), \fcons(g2))}p
   \label{perfop:19} \\
S \not= \emptyset \label{perfop:20} \\
et' = [minpt(S), maxpt(S)] \label{perfop:21}
\end{gather}
Let us assume that there is only one maximal period where J.Adams is
inspecting BA737, and that the inspection is completed at the end of
that period. Then, the $S$ of \pref{perfop:19} is the 
maximal period, and \pref{perfop:18} and \pref{perfop:20}
hold. \pref{perfop:21} requires $et'$ to be the same period as $S$, in
which case \pref{perfop:17} is trivially satisfied. The denotation of
\pref{perfop:8} w.r.t.\ $st$ is $T$ (i.e.\ the answer to
\pref{perfop:7} is affirmative) iff for some $et$, $et' = S$, and
\pref{perfop:14} and \pref{perfop:15} hold, i.e.\ iff there is an
$et$ within 27/9/95, such that $et$ follows $S$
($S = et'$ is the period that covers the whole inspection). The
situation is depicted in figure \ref{perf_op2_fig}. In other words,
27/9/95 must contain an $et$ where the inspection has already
been completed.

Let us now consider \pref{perfop:9}. Its denotation w.r.t.\ $st$ will
be true iff for some $et \in \periods$ and $g \in G$, \pref{perfop:22}
holds.
\begin{examps}
\item $\denot{st,et,\pts,g}
      {\past[e1^v, \perf[e2^v, \at[\mathit{27/9/95},
                                   \culm[inspecting(ja, g2)]]]]}
   = T$
\label{perfop:22}
\end{examps}
By the definition of \past, \pref{perfop:22} holds iff
\pref{perfop:23} is true. (For simplicity, I ignore again $e1^v$ and
$e2^v$.)
\begin{examps}
\item $\denot{st,et,[t_{first}, st),g}
      {\perf[e2^v, \at[\mathit{27/9/95},
                                   \culm[inspecting(ja, g2)]]]}$
\label{perfop:23}
\end{examps}
By the definition of \perf, \pref{perfop:23} is true iff
for some $et' \in \periods$, \pref{perfop:24}, \pref{perfop:25}, and
\pref{perfop:26} hold.
\begin{gather}
et \subper [t_{first}, st) \label{perfop:24} \\
maxpt(et') \prec minpt(et) \label{perfop:25} \\
\denot{st,et',\pts,g}{\at[\mathit{27/9/95},
                          \culm[inspecting(ja, g2)]]} = T
   \label{perfop:26} 
\end{gather}
By the definition of the \at operator, \pref{perfop:26}
holds iff \pref{perfop:27} holds. (I assume again that
$\mathit{27/9/95}$ denotes the obvious period.)
\begin{equation}
\denot{st,et',\fcons(\mathit{27/9/95}),g}{\culm[inspecting(ja,
g2)]} 
   = T \label{perfop:27}
\end{equation}
By the definition of \culm, \pref{perfop:27} holds iff
\pref{perfop:28} -- \pref{perfop:32} are true. 
\begin{gather}
et' \subper \fcons(\mathit{27/9/95}) \label{perfop:28} \\
\fculms(inspecting, 2)(\fcons(ja), \fcons(g2)) = T
   \label{perfop:29} \\
S = \bigcup_{p \in \fpfuns(inspecting, 2)(\fcons(ja), \fcons(g2))}p
   \label{perfop:30} \\
S \not= \emptyset \label{perfop:31} \\
et' = [minpt(S), maxpt(S)] \label{perfop:32}
\end{gather}
Assuming again that there is only one maximal period where J.Adams is
inspecting BA737, and that the inspection is completed at the end of
that period, the $S$ of \pref{perfop:30} is the 
maximal period, and \pref{perfop:29} and \pref{perfop:31} hold. 
\pref{perfop:32} requires $et'$ to be the same as $S$. The
denotation of \pref{perfop:9} w.r.t.\ $st$ is $T$ (i.e.\ the answer to
\pref{perfop:7} is affirmative) iff for some $et$, $et' = S$, and
\pref{perfop:24}, \pref{perfop:25}, and \pref{perfop:28} hold. That is
there must be some past $et$ that follows $S$ ($S = et'$ is the period
that covers the whole inspection), with $S$ falling within 27/9/95
(figure \ref{perf_op2_fig}). The inspection must have been completed
within 27/9/95.

\begin{figure}[tb]
  \hrule
  \medskip
  \begin{center}
    \includegraphics[scale=.6]{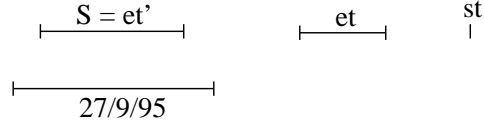}
    \caption{Second reading of \qit{J.Adams had inspected gate 2 on
             27/9/95}} 
    \label{perf_op3_fig}
  \end{center}
  \hrule
\end{figure}

In \pref{perfop:33}, where there are no temporal adverbials, the
corresponding formula \pref{perfop:34} requires some past 
$et$ (pointed to by $e1^v$) to exist, such that $et$ follows an
$et'$ (pointed to by $e2^v$) that covers exactly the whole
(from start to completion) inspection of gate 2 by
J.Adams. The net effect is that the inspection must have been
completed in the past.
\begin{examps}
\item J.Adams had inspected gate 2 \label{perfop:33} 
\item $\past[e1^v, \perf[e2^v, \culm[inspecting(ja, g2)]]]$
   \label{perfop:34} 
\end{examps}

As noted in section \ref{past_perfect}, there is a reading of 
\pref{perfop:35} (probably the preferred one)
whereby the two-year period ends on 1/1/94, i.e.\ J.Adams was still a
manager on 1/1/94. Similarly, there is a reading of \pref{perfop:37},
whereby the two-year period ends at $st$, i.e.\ J.Adams is still a
manager (section \ref{present_perfect}). These readings cannot be
captured in \topl.
\begin{examps}
\item On 1/1/94, J.Adams had been a manager for two years. \label{perfop:35}
\item J.Adams has been a manager for two years. \label{perfop:37}
\end{examps}
For example, \pref{perfop:36} requires some past $et$
(pointed to by $e1^v$) to exist, such that $et$ falls within 1/1/94,
$et$ follows a period $et'$ (pointed to by $e2^v$), $et'$
is a period where J.Adams is a manager, and
the duration of $et'$ is two years. If, for example, J.Adams was a
manager only from 1/1/88 to 31/12/89, \pref{perfop:36}
causes the answer to \pref{perfop:35} to be affirmative. 
\pref{perfop:36} does not require the two-year period to end on 1/1/94.
\begin{examps}
\item $\at[\mathit{1/1/94}, 
           \past[e1^v, 
                 \perf[e2^v, \for[year^c, 2, be(ja, manager)]]]]$
   \label{perfop:36}
\end{examps}
Various versions of \perf operators have been used in
\cite{Dowty1982}, \cite{Richards}, \cite{Pirie1990}, \cite{Crouch2},
and elsewhere. 


\section{Occurrence identifiers} \label{occurrence_ids}

\begin{figure}[tb]
  \hrule
  \medskip
  \begin{center}
    \includegraphics[scale=.58]{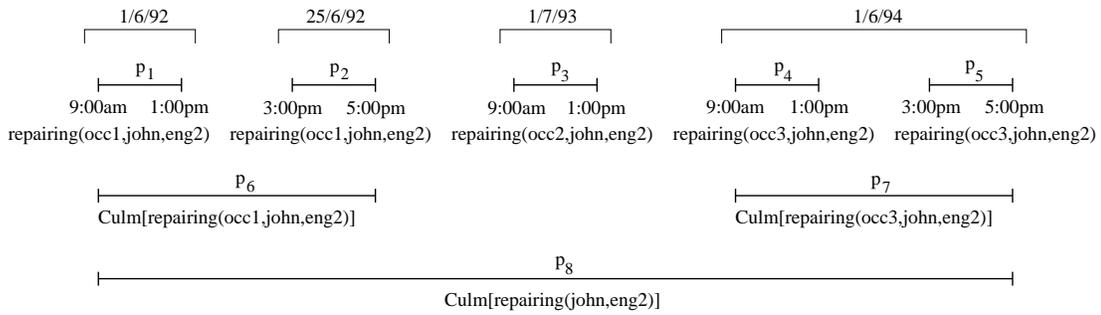}
    \caption{Occurrence identifiers}
    \label{episodes_fig}
  \end{center}
  \hrule
\end{figure}

Predicates introduced by verbs whose base forms are culminating
activities often have an extra argument that acts as an
\emph{occurrence identifier}. Let us consider a scenario involving an
engineer, John, who worked on engine 2 repairing 
faults of the engine at several past times (figure
\ref{episodes_fig}). John started repairing a
fault of engine 2 on 1/6/92 at 9:00am. He continued to work on this
fault up to 1:00pm on the same day, at which point he temporarily
abandoned the repair without completing it. He resumed the repair at
3:00pm on 25/6/92, and completed it at 5:00pm on the same day.

In 1993, John was asked to repair another fault of engine 2. He
started the repair on 1/7/93 at 9:00am, and continued to work on that
fault up to 1:00pm on the same day without completing the repair. He
then abandoned the repair for ever (John was not qualified to fix that
fault, and the repair was assigned to another engineer). Finally, in
1994 John was asked to repair a third fault of engine 2. He started
to repair the third fault on 1/6/94 at 9:00am, and continued to
work on that fault up to 1:00pm on the same day, without completing
the repair. He resumed the repair at 3:00pm, and completed it at
5:00pm on the same day.

There is a problem if \pref{epid:1} is represented as
\pref{epid:2}. Let us assume that the question is submitted after
1/6/94. One would expect the answer to be affirmative, since a
complete past repair of engine 2 by John is situated within 1/6/94. In
contrast, \pref{epid:2} causes the answer to be negative. The
semantics of \culm (section \ref{culm_op}) requires 
$et$ to start at the beginning of the earliest maximal
period where $repairing(john, eng2)$ holds (i.e.\ at the beginning of
$p_1$ in figure \ref{episodes_fig}) and to end at the end of the
latest maximal period where $repairing(john, eng2)$ holds (i.e.\ at
the end of $p_5$ in figure \ref{episodes_fig}). That is, $et$ must
be $p_8$ of figure \ref{episodes_fig}. The \at requires
$et$ ($p_8$) to be also a subperiod of 1/6/94. Since this is not the
case, the answer is negative.
\begin{examps}
\item Did John repair engine 2 on 1/6/94? \label{epid:1}
\item $\at[\mathit{1/6/94}, \past[e^v, \culm[repairing(john,eng2)]]]$
   \label{epid:2}
\end{examps}
The problem is that although John was repairing engine 2 during all
five periods ($p_1$, $p_2$, $p_3$, $p_4$, and $p_5$), the five periods
intuitively belong to different occurrences of the situation where
John is repairing engine 2. The first two periods have to do with the
repair of the first fault (occurrence 1), the third period has to do with
the repair of the second fault (occurrence 2), and the last two periods
relate to the repair of the third fault (occurrence 3). The
$\culm[repairing(john,eng2)]$ of \pref{epid:2}, however, does not
distinguish between the three occurrences, and forces $et$ to start at
the beginning of $p_1$ and to end at the end of $p_5$. Instead, we
would like $\culm[repairing(john,eng2)]$ to distinguish between the
three occurrences: to require $et$ to start at the beginning of $p_1$
(beginning of the first repair) and to end at the end of $p_2$
(completion of the first repair), or to require $et$ to start at
the beginning of $p_4$ (beginning of the third repair) and to end at
the end of $p_5$ (completion of the third
repair). ($\culm[repairing(john,eng2)]$ should not allow $et$ to be
$p_3$, because the second repair does not reach its completion at the
end of $p_3$.)

To achieve this, an occurrence-identifying argument is added to
$fixing(john,eng2)$. If $occ1$, $occ2$, and $occ3$ denote
the three repairing-occurrences,
$fixing(occ1,john,eng2)$ will be true only at $et$s that are
subperiods of $p_1$ or $p_2$, $fixing(occ2, john, eng2)$ only at
$et$s that are subperiods of $p_3$, and $fixing(occ3, john,
eng2)$ only at $et$s that are subperiods of $p_4$ or $p_5$. In
practice, the occurrence-identifying argument is always a
variable. For example, \pref{epid:1} is now represented as
\pref{epid:3} instead of \pref{epid:2}.
\begin{examps}
\item $\at[\mathit{1/6/94}, \past[e^v, \culm[repairing(occ^v,john,eng2)]]]$
   \label{epid:3}
\end{examps}
Intuitively, according to \pref{epid:3} the answer should be
affirmative if there is an $et$ and a particular occurrence $occ^v$ of
the situation where John is repairing engine 2, such that $et$ starts
at the beginning of the first period where $occ^v$ is ongoing, $et$
ends at the end of the last period where $occ^v$ is ongoing, $occ^v$
reaches its completion at the end of $et$, and $et$ falls within the
past and 1/6/94. Now if \pref{epid:1} is submitted after 1/6/94, the
answer is affirmative.

To see that \pref{epid:3} generates the correct result, let us
examine the denotation of \pref{epid:3}. The denotation of
\pref{epid:3} w.r.t.\ $st$ is affirmative if for some $et \in
\periods$ and $g \in G$, \pref{epid:4} holds.
\begin{examps}
\item $\denot{st,et,\pts,g}
       {\at[\mathit{1/6/94}, \past[e^v, \culm[repairing(occ^v,john,eng2)]]]}
       = T$ \label{epid:4}
\end{examps}
Assuming that the question is submitted after 1/6/94, and that
$\mathit{1/6/94}$ denotes the obvious period, by the definitions of
\at and \past, \pref{epid:4} holds iff \pref{epid:5} and \pref{epid:6} hold.
\begin{gather}
g(e^v) = et \label{epid:5} \\
\denot{st,et,\fcons(\mathit{1/6/94}),g}{\culm[repairing(occ^v,john,eng2)]}
   = T \label{epid:6}
\end{gather}
By the definition of \culm, \pref{epid:6} holds iff
\pref{epid:7} -- \pref{epid:10} hold, where $S$ is as in \pref{epid:11}.
\begin{gather}
et \subper \fcons(\mathit{1/6/94}) \label{epid:7} \\
\fculms(repairing,2)(g(occ^v), \fcons(john), \fcons(eng2)) = T 
   \label{epid:8} \\
S \not= \emptyset \label{epid:9} \\
et = [minpt(S), maxpt(S)] \label{epid:10} \\
S = 
\bigcup_{p \in \fpfuns(repairing, 2)(g(occ^v), \fcons(john), \fcons(eng2))}p
   \label{epid:11}
\end{gather}
The denotation of \pref{epid:3} w.r.t.\ $st$ is $T$ (i.e.\ the
answer to \pref{epid:1} is affirmative), iff for some $et \in
\periods$ and $g \in G$, \pref{epid:5} and \pref{epid:7} --
\pref{epid:10} hold.
For $et$ as in \pref{epid:21} and $g$ the variable assignment of
\pref{epid:20}, \pref{epid:5} and \pref{epid:7}
hold. \pref{epid:11} becomes \pref{epid:13}, and \pref{epid:9} holds. 
\pref{epid:10} becomes \pref{epid:21}, which holds ($et$ was chosen to
satisfy it). \pref{epid:8} also holds, because
the third repair is completed at the end of $p_5$. 
\begin{gather}
et = [minpt(p_4), maxpt(p_5)] = p_7 \label{epid:21} \\
g(\beta) = 
\begin{cases}
et & \text{if } \beta = e^v \\
\fcons(occ3) & \text{if } \beta = occ^v \\
o & \text{otherwise ($o$ is an arbitrary element of \objs)}
\end{cases}
\label{epid:20} \\
S = p_4 \union p_5 \label{epid:13} 
\end{gather}
Hence, there is some $et \in \periods$ and $g \in G$ for which
\pref{epid:5} and \pref{epid:7} -- \pref{epid:10} hold, i.e.\ the
answer to \pref{epid:1} will be affirmative as wanted.

Occurrence identifiers are a step towards formalisms that treat
occurrences of situations (or ``events'' or ``episodes'') as objects
in the modelled world (e.g.\ \cite{Parsons1990}, \cite{Kamp1993},
\cite{Blackburn1994}, \cite{Hwang1994}). In \topl all terms (constants
and variables) denote elements of \objs, i.e.\ objects of the modelled
world. Thus, allowing occurrence-identifying terms (like $occ^v$ in
\pref{epid:3}) implies that occurrences of situations are also world
objects.  Unlike other formalisms (e.g.\ those mentioned above),
however, \topl does not treat these occurrence-identifying terms in
any special way, and there is nothing in the definition of \topl to
distinguish objects denoted by occurrence-identifiers from objects
denoted by other terms.


\section{Tense anaphora and localisation time} \label{lt_anaphora}

Although tense anaphora (section
\ref{temporal_anaphora}) was not considered during the work of this thesis,
it seems that \topl's localisation time could prove useful if
this phenomenon were to be supported. As noted in section
\ref{temporal_anaphora}, some cases of tense anaphora can be
handled by storing the temporal window established by adverbials and
tenses of previous questions, and by requiring the situations of
follow-up questions to fall within that window. \topl's $lt$ can
capture this notion of previous window. Assuming that \pref{ltan:1} is
submitted after 1993, the \at and \past operators of the corresponding
formula \pref{ltan:2} narrow $lt$ to the period that covers exactly
1993. This period could be stored, and used as the initial value of
$lt$ in \pref{ltan:4}, that expresses the follow-up question
\pref{ltan:3}. In effect, \pref{ltan:3} would be taken to mean \pref{ltan:5}.
\begin{examps}
\item Was Mary the personnel manager in 1993? \label{ltan:1}
\item $\at[1993, \past[e^v, manager\_of(mary, personnel)]]$ \label{ltan:2}
\item Who was the personnel manager? \label{ltan:3}
\item $?wh^v \; \past[e^v, manager\_of(wh^v, personnel)]$ \label{ltan:4}
\item Who was the personnel manager in 1993? \label{ltan:5}
\end{examps}

Substantial improvements are needed to make these ideas workable.  For
example, if \pref{ltan:1} and \pref{ltan:3} are followed by
\pref{ltan:6} (expressed as \pref{ltan:7}), and the dialogue takes
place after 1993, the \nlitdb must be intelligent enough to reset $lt$
to the whole time axis. Otherwise, no person will ever be reported,
because the \pres of \pref{ltan:7} requires $et$ to contain $st$, and
an $et$ that contains $st$ can never fall within the past year 1993
(the $lt$ of the previous question).
\begin{examps}
\item Who is (now) the personnel manager? \label{ltan:6}
\item $?wh^v \; \pres[manager\_of(wh^v, personnel)]$ \label{ltan:7}
\end{examps}


\section{Expressing habituals} \label{hab_problems}

As noted in section \ref{pres_op}, habitual readings of
sentences are taken to involve habitual homonyms of verbs. 
Habitual homonyms introduce different predicates than the
corresponding non-habitual ones. For example, \pref{habp:1} and
\pref{habp:3} would be expressed as \pref{habp:2} and \pref{habp:4}
respectively. Different predicates would used in the two cases.
\begin{examps}
\item Last month BA737 (habitually) departed from gate 2. \label{habp:1}
\item $\partop[month^c, mon^v, -1] \; \land$\\
      $\at[mon^v, \past[e^v, hab\_depart\_from(ba737, gate2)]]$
    \label{habp:2} 
\item Yesterday BA737 (actually) departed from gate 2. \label{habp:3}
\item $\partop[day^c, y^v, -1] \; \land$\\
      $\at[y^v, \past[e^v, actl\_depart\_from(ba737, gate2)]]$ 
   \label{habp:4}
\end{examps}
$hab\_depart\_from(ba737, gate2)$ is intended to hold at $et$s that
fall within periods where BA737 has the habit of departing from gate
2. If BA737 departed habitually from gate 2 throughout 1994,
$hab\_depart\_from(ba737, gate2)$ would be true at any $et$ that is a
subperiod of 1994. In contrast, $actl\_depart\_from(ba737, gate2)$ is
intended to hold only at $et$s where BA737 actually departs from gate
2. If departures are modelled as instantaneous (as in the airport
application), $actl\_depart\_from(ba737, gate2)$ is true only at
instantaneous $et$s where BA737 leaves gate 2.
One would expect that if BA737 had the habit of departing from gate 2
during some period, it would also have actually departed from gate 2
at least some times during that period:
if $hab\_depart\_from(ba737, gate2)$ is true at an $et$,
$actl\_depart\_from(ba737, gate2)$ would also be true at
some subperiods $et'$ of $et$. There is nothing in the
definition of \topl, however, to guarantee that this implication
holds. The event times where $hab\_depart\_from(ba737, gate2)$ and
$actl\_depart\_from(ba737, gate2)$ hold are ultimately determined by
\fpfuns (that specifies the maximal
periods where the two predicates hold; see section
\ref{top_model}). There is no restriction in the definition of \topl
to prohibit whoever defines \fpfuns from specifying that
$hab\_depart\_from(ba737, gate2)$ is true at some $et$ that does not
contain any $et'$ where $actl\_depart\_from(ba737, gate2)$ is true.

Another issue is how to represent \pref{habp:5}.
\pref{habp:5} cannot be represented as \pref{habp:6}. \pref{habp:6}
says that at 5:00pm on some day in the previous month BA737 had the
habit of departing. I have found no elegant solution to this
problem. \pref{habp:5} is mapped to \pref{habp:7}, where the constant
$\text{\textit{5:00pm}}$ is intended to denote a generic
representative of 5:00pm-periods. This generic representative is taken
to be an entity in the world.
\begin{examps}
\item Last month BA737 (habitually) departed at 5:00pm. \label{habp:5}
\item $\partop[month^c, mon^v, -1] \land 
       \partop[\text{\textit{5:00pm}}^g, fv^v] \; \land$\\
      $\at[mon^v, \at[fv^v, \past[e^v, hab\_depart(ba737)]]$ \label{habp:6}
\item $\partop[month^c, mon^v, -1] \; \land$ \\
      $\at[mon^v, \past[e^v, hab\_depart\_time(ba737,
      \text{\textit{5:00pm}})]]$ \label{habp:7} 
\end{examps}
Unlike \pref{habp:5}, where \qit{at 5:00pm} introduces a
constant ($\text{\textit{5:00pm}}$) as a predicate-argument in \pref{habp:7},
the \qit{at 5:00pm} of \pref{habp:8} introduces
an \at operator in \pref{habp:9}.
\begin{examps}
\item Yesterday BA737 (actually) departed at 5:00pm. \label{habp:8}
\item $\partop[day^c, y^v, -1] \land 
       \partop[\text{\textit{5:00pm}}^g, fv^v] \; \land$ \\
      $\at[y^v, \at[fv^v, \past[e^v, actl\_depart(ba737)]]]$ \label{habp:9}
\end{examps}
The fact that \qit{at 5:00pm} is treated in such different ways in
the two cases is admittedly counter-intuitive, and it also complicates
the translation from English to \topl (to be discussed in chapter
\ref{English_to_TOP}). 


\section{Summary}

\topl is a formal language, used to represent the meanings of the
English questions that are submitted to the \nlitdb. The denotation
with respect to $st$ of a \topl formula specifies what the answer to
the corresponding English question should report ($st$ is the
time-point where the question is submitted to the \nlitdb). The
denotations with respect to $st$ of \topl formulae are defined in
terms of the denotations of \topl formulae with respect to $st$, $et$,
and $lt$. $et$ (event time) is a time period where the situation
described by the formula holds, and $lt$ (localisation time) is a
temporal window within which $et$ must be placed.

Temporal linguistic mechanisms are expressed in \topl using temporal
operators that manipulate $st$, $et$, and $lt$. There are thirteen
operators in total. \partop picks a period from a 
partitioning. \pres and \past are used when
expressing present and past tenses. \perf is
used in combination with \past to express the past
perfect. \culm is used to represent non-progressive forms
of verbs whose base forms are culminating activities. \at,
\before, and \after are employed when expressing punctual
and period adverbials, and when expressing \qit{while~\dots},
\qit{before~\dots}, and \qit{after~\dots} subordinate
clauses. Duration \qit{in~\dots} and \qit{for~\dots} adverbials are
expressed using \for. \fills can be used to
represent readings of sentences where the situation of the verb 
covers the whole localisation time; \fills, however,
is not used in the rest of this thesis, nor in the prototype
\nlitdb. \lbegin and \lend are used to 
refer to time-points where situations start or stop.  Finally, \ntense
allows noun phrases to refer either to $st$ or to the time of the
verb's tense.



\chapter{From English to TOP} \label{English_to_TOP}

\proverb{One step at a time.}


\section{Introduction} 

This chapter shows how \hpsg \cite{Pollard1} \cite{Pollard2} was
modified to map English questions directed to a \nlitdb to appropriate
\topl formulae.\footnote{The \hpsg version of this thesis is based on
  the revised \hpsg version of chapter 9 of \cite{Pollard2}.} Although
several modifications to \hpsg were introduced, the \hpsg version of
this thesis remains very close to \cite{Pollard2}. The main
differences from \cite{Pollard2} are that: (a) \hpsg mechanisms for
phenomena not examined in this thesis (e.g.\ pronouns, relative
clauses) were removed, and (b) the situation-theoretic semantic
constructs of \hpsg were replaced by feature structures that represent
\topl expressions.

Readers with a rudimentary grasp of modern unification-based grammars
\cite{Shieber} should be able to follow most of the discussion in this
chapter. Some of the details, however, may be unclear to readers not
familiar with \hpsg. The \hpsg version of this thesis was implemented
as a grammar for the \textsc{Ale} system (see chapter
\ref{implementation}).


\section{HPSG basics} \label{HPSG_basics}

In \hpsg, each word and syntactic constituent is mapped to a
\emph{sign}, a feature structure of a particular form, that provides
information about the word or syntactic constituent. An \hpsg grammar
consists of signs for words (I call these \emph{lexical signs}),
\emph{lexical rules}, \emph{schemata}, \emph{principles}, and a
\emph{sort hierarchy}, all discussed below.

\subsection{Lexical signs and sort hierarchy}

Lexical signs provide information about individual words. (Words with
multiple uses may receive more than one lexical sign.)
\pref{lentr:1} shows a lexical sign for the base form of \qit{to land}
in the airport domain.
\begin{examps}
\item 
\avmoptions{active}
\begin{avm}
[\avmspan{phon \; \<\fval land\>} \\
 synsem & [loc & [cat & [head  & \osort{verb}{
                                 [vform & bse \\
                                  aux   & $-$ ]} \\
                         aspect & culmact \\
                         spr    & \<\> \\
                         subj   & \< \feat np[-prd]$_{@1}$ \>  \\
                         comps  & \< 
                                    \feat pp[-prd, pform {\fval on}]$_{@2}$
                                  \> ]\\
                  cont & \sort{landing\_on}{
                         [arg1 & occr\_var \\
                          arg2 & @1 \\
                          arg3 & @2]} ]]]
\end{avm}
\label{lentr:1}
\end{examps}
The $<$ and $>$ delimiters denote lists. The {\feat phon} feature
shows the list of words to which the sign corresponds
(\pref{lentr:1} corresponds to the single word \qit{land}). Apart from
{\feat phon}, every sign has a {\feat synsem} feature (as well as
other features not shown in \pref{lentr:1}; I often omit features that
are not relevant to the discussion). The value of {\feat synsem} in
\pref{lentr:1} is a feature structure that has a feature {\feat loc}.
The value of {\feat loc} is in turn a feature
structure that has the features {\feat cat} (intuitively, syntactic
category) and {\feat cont} (intuitively, semantic content).

Each \hpsg feature structure belongs to a particular sort. The sort
hierarchy of \hpsg shows the available sorts, as well as which sort is
a subsort of which other sort. It also specifies which features the
members of each sort must have, and the sorts to which the values of
these features must belong. (Some modifications were made to the sort
hierarchy of \cite{Pollard2}. These will be discussed in sections
\ref{TOP_FS} and \ref{more_ind}.) In \pref{lentr:1}, for example, the
value of {\feat head} is a feature structure of sort {\srt verb}. The
value of {\feat head} signals that the word is the base form ({\feat
  vform} {\fval bse}) of a non-auxiliary ({\feat aux}~{\fval $-$})
verb. The sort hierarchy of \cite{Pollard2} specifies that the value
of {\feat head} must be of sort {\srt head}, and that {\srt verb}\/ is
a subsort of {\srt head}. This allows feature structures of sort {\srt
  verb}\/ to be used as values of {\feat head}. The value of {\feat
  vform} in \pref{lentr:1} is an \emph{atomic feature structure} (a
feature structure of no features) of sort {\srt bse}.  For simplicity,
when showing feature structures I often omit uninteresting sort names.

{\feat aspect} \index{aspect@{\feat aspect} (\hpsg feature)} is
the only new \hpsg feature of this thesis. It is a feature of feature
structures of sort {\srt cat}\/ (feature structures that can be used
as values of {\feat cat}), and its values are feature structures of
sort {\srt aspect}. {\srt aspect}\/ contains only atomic feature
structures, and has the subsorts: {\srt state},
\index{state@{\srt state}\/ (\hpsg sort, state aspectual class)}
{\srt activity},
\index{activity@{\srt activity}\/ (\hpsg sort, activity aspectual class)}
{\srt culmact}\/
\index{culmact@{\srt culmact}\/ (\hpsg sort, culminating activity)}
(culminating activity), and {\srt point}.
\index{point@{\srt point}\/ (\hpsg sort, point aspectual class)}
{\srt state}\/ is in turn partitioned into: {\srt lex\_state}\/
\index{lexstate@{\srt lex\_state}\/ (\hpsg sort, lexical state)}
(lexical state), {\srt progressive}\/
\index{progressive@{\srt progressive}\/ (\hpsg sort, progressive state)}
(progressive state), and {\srt cnsq\_state}\/ 
\index{cnsqstate@{\srt cnsq\_state}\/ (\hpsg sort, consequent state)}
(consequent state). This agrees with the aspectual
taxonomy of chapter \ref{linguistic_data}. Following table
\vref{airport_verbs}, \pref{lentr:1} classifies the base form of
\qit{to land} as culminating activity.

The {\feat spr}, {\feat subj}, and {\feat comps} features of
\pref{lentr:1} provide information about the specifier, subject, and
complements with which the verb has to combine. Specifiers are
determiners (e.g.\ \qit{a}, \qit{the}), and words like \qit{much} (as
in \qit{much more}) and \qit{too} (as in \qit{too late}). Verbs do not
admit specifiers, and hence the value of {\feat spr} in \pref{lentr:1}
is the empty list.

The {\feat subj} value of \pref{lentr:1} means that the verb requires
a noun-phrase as its subject. The {\feat np[-prd]$_{\avmbox{1}}$} in
\pref{lentr:1} has the same meaning as in \cite{Pollard2}. Roughly
speaking, it is an abbreviation for a sign that corresponds to a noun
phrase. The {\feat -prd} means that the noun phrase must be
non-predicative (see section \ref{hpsg:nouns} below).  The \avmbox{1}
is intuitively a pointer to the world entity described by the noun
phrase. Similarly, the {\feat comps} value of \pref{lentr:1} means
that the verb requires as its complement a non-predicative
prepositional phrase (section \ref{hpsg:pps} below), introduced by
\qit{on}. The \avmbox{2} is intuitively a pointer to the world entity
of the prepositional phrase (e.g.\ if the prepositional phrase is
\qit{on a runway}, the \avmbox{2} is a pointer to the runway).

The value of {\feat cont} in \pref{lentr:1} represents the \topl
predicate $landing\_on(\beta, \tau_1, \tau_2)$, where $\tau_1$ and
$\tau_2$ are \topl terms corresponding to \avmbox{1} and \avmbox{2},
and $\beta$ is a \topl variable acting as an occurrence identifier
(section \ref{occurrence_ids}).\footnote{I follow the approach of
  section 8.5.1 of \cite{Pollard2}, whereby the {\feat relation}
  feature is dropped, and its role is taken up by the sort of the
  feature structure.} The exact relation between \hpsg feature
structures and \topl expressions will be discussed in the following sections.

\subsection{Lexical rules}

Lexical rules generate new lexical signs from existing ones. In section
\ref{hpsg:verb_forms}, for example, I introduce lexical rules that
generate automatically lexical signs for (single-word) non-base
verb forms (e.g.\ a sign for the simple past \qit{landed}) from signs
for base forms (e.g.\ \pref{lentr:1}). This reduces the number of
lexical signs that need to be listed in the grammar. 

\subsection{Schemata and principles} \label{schemata_principles}

\hpsg schemata specify basic patterns that are used when words or
syntactic constituents combine to form larger constituents. For
example, the \emph{head-complement schema} is the pattern that is used
when a verb combines with its complements (e.g.\ when \qit{landed}
combines with its complement \qit{on runway 2}; in this case, the verb
is the ``head-daughter'' of the constituent \qit{landed on runway 2}).
The \emph{head-subject schema} is the one used when a verb phrase (a
verb that has combined with its complements but not its subject)
combines with its subject (e.g.\ when \qit{landed on runway 2}
combines with \qit{BA737}; in this case, the verb phrase is the
head-daughter of \qit{BA737 landed on runway 2}). No modifications to
the schemata of \cite{Pollard2} are introduced in this thesis, and
hence schemata will not be discussed further. 

\hpsg principles control the propagation of feature values from the
signs of words or syntactic constituents to the signs of their
super-constituents. The \emph{head feature principle}, for example,
specifies that the sign of the super-constituent inherits the {\feat
  head} value of the head-daughter's sign. This causes the sign of
\qit{landed on runway 2} to inherit the {\feat head} value of the sign
of \qit{landed}, and the same value to be inherited by the sign of
\qit{BA737 landed on runway 2}. This thesis uses simplified versions
of Pollard and Sag's semantics principle and constituent ordering
principle (to be discussed in sections \ref{non_pred_nps} and
\ref{fronted}), and introduces one new principle (the aspect principle,
to be discussed in section \ref{hpsg:punc_adv}). All other principles
are as in \cite{Pollard2}.


\section{Representing TOP yes/no formulae in HPSG} \label{TOP_FS}

According to \cite{Pollard2}, the {\feat cont} value of \pref{lentr:1}
should actually be \pref{lentr:2}.
\begin{examps}
\item 
\avmoptions{active}
\begin{avm}
\sort{psoa}{
[quants  & \<\> \\
 nucleus & \sort{landing\_on}{
           [arg1 & occr\_var \\
            arg2 & @1 \\
            arg3 & @2]} ]}
\end{avm}
\label{lentr:2}
\end{examps}
In \cite{Pollard2}, feature structures of sort {\srt psoa} have two
features: {\feat quants} and {\feat nucleus}.\footnote{``{\srt Psoa}''
  stands for ``parameterised state of affairs'', a term from situation
  theory \cite{Cooper1990}. The semantic analysis here is not
  situation-theoretic, but the term ``psoa'' is still used for
  compatibility with \cite{Pollard2}.} {\feat quants}, which is part
of \hpsg's quantifier storage mechanism, is not used in this thesis.
This leaves only one feature ({\feat nucleus}) in {\srt psoa}s. For
simplicity, {\feat nucleus} was also dropped, and the {\srt psoa}\/
sort was taken to contain the feature structures that would be values
of {\feat nucleus} in \cite{Pollard2}.

\begin{figure}
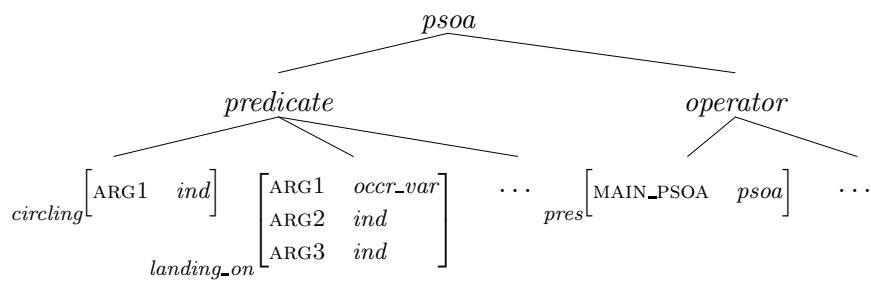

\avmoptions{}
\setlength{\GapWidth}{5mm}
\hrule
\begin{center}
\begin{bundle}{{\srt psoa}}
\chunk{
   \begin{bundle}{{\srt predicate}}
   \chunk{\begin{avm}
          \sort{circling}{
          \[arg1 & ind\]}
          \end{avm}}
   \chunk{\begin{avm}
          \osort{landing\_on}{ 
          \[arg1 & occr\_var \\
            arg2 & ind \\
            arg3 & ind\]}
          \end{avm}}
   \chunk{\dots}
   \end{bundle}}
\chunk{
   \begin{bundle}{{\srt operator}}
   \chunk{\begin{avm}
          \osort{pres}{
          \[main\_psoa & psoa\]}
          \end{avm}}
   \chunk{\dots}
   \end{bundle}}
\end{bundle}
\caption{Subsorts of {\srt psoa}}
\label{psoa_fig}
\index{pres2@{\srt pres}\/ (\hpsg sort, corresponds to \topl's \pres)}
\index{past2@{\srt past}\/ (\hpsg sort, corresponds to \topl's \past)}
\index{perf2@{\srt perf}\/ (\hpsg sort, corresponds to \topl's \perf)}
\index{atop2@{\srt at\_op}\/ (\hpsg sort, corresponds to \topl's \at)}
\index{beforeop2@{\srt before\_op} (\hpsg sort, corresponds to \topl's \before)}
\index{afterop2@{\srt after\_op} (\hpsg sort, corresponds to \topl's \after)}
\index{part2@{\srt part}\/ (\hpsg sort, corresponds to \topl's \partop)}
\index{culm2@{\srt culm}\/ (\hpsg sort, corresponds to \topl's \culm)}
\index{end2@{\srt end}\/ (\hpsg sort, corresponds to \topl's \lend)}
\index{begin2@{\srt begin}\/ (\hpsg sort, corresponds to \topl's \lbegin)}
\index{and2@{\srt and}\/ (\hpsg sort, corresponds to \topl's conjunction)}
\index{ntense2@{\srt ntense}\/ (\hpsg sort, corresponds to \topl's \ntense)}
\index{forop2@{\srt for\_op}\/ (\hpsg sort, corresponds to \topl's \for)}
\index{mainpsoa@{\feat main\_psoa} (\hpsg feature, used in the representation of \topl formulae)}
\index{ethandle@{\feat et\_handle} (\hpsg feature, used in the representation of \topl formulae)}
\index{timespec@{\feat time\_spec} (\hpsg feature, used in the representation of \topl formulae)}
\index{partng@{\feat partng} (\hpsg feature, used in the representation of \topl formulae)}
\index{partvar@{\feat part\_var} (\hpsg feature, used in the representation of \topl formulae)}
\index{conjunct12@{\feat conjunct1}, {\feat 2} (\hpsg features, used in the representation of \topl formulae)}
\index{durunit@{\feat dur\_unit} (\hpsg feature, used in the representation of \topl formulae)}
\index{duration@{\feat duration} (\hpsg feature, used in the representation of \topl formulae)}
\end{center}
\hrule
\end{figure}

More precisely, in this thesis {\srt psoa}\/ has two subsorts: {\srt
  predicate}\/
\index{predicate@{\srt predicate}\/ (\hpsg sort, represents \topl predicates)}
and {\srt operator}\/ 
\index{operator@{\srt operator}\/ (\hpsg sort, represents \topl operators)}
(figure \ref{psoa_fig}). {\srt predicate}\/ contains feature
structures that represent \topl predicates, while {\srt operator}\/
contains feature structures that represent all other \topl yes/no
formulae. (Hence, {\srt psoa}\/ corresponds to all yes/no formulae.)
{\srt predicate}\/ has domain-specific subsorts, corresponding to
predicate functors used in the domain for which the \nlitdb is
configured. In the airport domain, for example, {\srt landing\_on}\/
is a subsort of {\srt predicate}.  The feature structures in the
subsorts of {\srt predicate}\/ have features named {\feat arg1},
{\feat arg2}, {\feat arg3}, 
\index{arg123@{\feat arg1}, {\feat 2}, {\feat 3} (new
  \hpsg features, correspond to \topl predicate arguments)}
etc. These represent the first, second,
third, etc.\ arguments of the predicates.  The values of {\feat arg1},
{\feat arg2}, etc.\ are of sort {\srt ind}\/ ({\srt occr\_var}\/
\index{occrvar@{\srt occr\_var}\/ (\hpsg sort, represents occurrence identifiers)}
is a subsort of {\srt ind}). {\srt ind}\/ will be discussed further below.

\begin{figure}
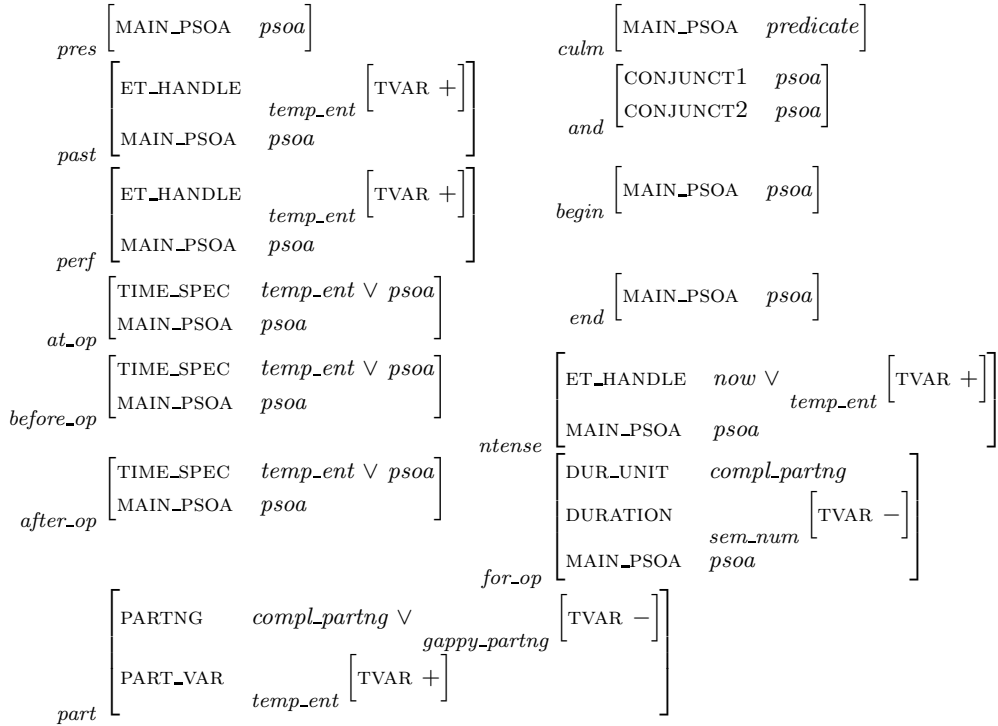

\avmoptions{active}
\hrule
\medskip
\hspace*{9mm}
\begin{tabular}{lll}
\begin{avm}
\osort{pres}{
[main\_psoa & psoa]}
\end{avm}
&&
\hspace*{7mm}
\begin{avm}
\osort{culm}{
[main\_psoa & predicate]}
\end{avm}
\\
\begin{avm}
\osort{past}{
[et\_handle & \sort{temp\_ent}{
              [tvar $+$]} \\
 main\_psoa & psoa]}
\end{avm}
&&
\hspace*{7mm}
\begin{avm}
\osort{and}{
[conjunct1 & psoa \\
 conjunct2 & psoa]}
\end{avm}
\\
\begin{avm}
\osort{perf}{
[et\_handle & \sort{temp\_ent}{
              [tvar $+$]} \\
 main\_psoa & psoa]}
\end{avm}
&&
\hspace*{7mm}
\begin{avm}
\osort{begin}{
[main\_psoa & psoa]}
\end{avm}
\\
\begin{avm}
\osort{at\_op}{
[time\_spec & temp\_ent $\lor$ psoa \\
 main\_psoa & psoa]}
\end{avm}
&&
\hspace*{7mm}
\begin{avm}
\osort{end}{
[main\_psoa & psoa]}
\end{avm}
\\
\begin{avm}
\osort{before\_op}{
[time\_spec & temp\_ent $\lor$ psoa \\
 main\_psoa & psoa]}
\end{avm}
&&
\begin{avm}
\osort{ntense}{
[et\_handle & now $\lor$ \sort{temp\_ent}{
                         [tvar $+$]} \\
 main\_psoa & psoa]}
\end{avm}
\\
\begin{avm}
\osort{after\_op}{
[time\_spec & temp\_ent $\lor$ psoa \\
 main\_psoa & psoa]}
\end{avm}
&&
\begin{avm}
\osort{for\_op}{
[dur\_unit  & compl\_partng \\
 duration   & \sort{sem\_num}{
              [tvar $-$]} \\
 main\_psoa & psoa]}
\end{avm}
\\
\multicolumn{3}{l}{\avmoptions{}\begin{avm}
\osort{part}{
\[partng       & compl\_partng $\lor$ \sort{gappy\_partng}{
                                      \[tvar $-$ \]} \\
  part\_var    & \sort{temp\_ent}{
                 \[tvar $+$ \]} \]}
\end{avm}
}
\end{tabular}
\caption{Subsorts of {\srt operator}}
\label{operator_sorts}
\medskip
\hrule
\end{figure}

The {\srt operator}\/ sort has thirteen subsorts, shown in figure
\ref{operator_sorts}. These correspond to the twelve \topl operators
(\fills is ignored), plus one sort for conjunction.\footnote{The sorts
  that correspond to the \at, \before, \after, and \for operators are
  called {\srt at\_op}, {\srt before\_op}, {\srt after\_op}, and {\srt
    for\_op}\/ to avoid name clashes with existing \hpsg sorts.} The
order of the features in figure \ref{operator_sorts} corresponds to
the order of the arguments of the \topl operators. For example, the
{\feat et\_handle} and {\feat main\_psoa} features of the {\srt
  past}\/ sort correspond to the first and second arguments
respectively of \topl's $\past[\beta, \phi]$. For simplicity, in the
rest of this thesis I drop the $\partop[\sigma, \beta, \nu_{ord}]$
version of \partop (section \ref{denotation}), and I represent words
like \qit{yesterday} using \topl constants (e.g.\ $yesterday$) rather
than expressions like $\partop[day^c, \beta, -1]$. This is why there
is no sort for $\partop[\sigma, \beta, \nu_{ord}]$ in figure
\ref{operator_sorts}.

\begin{figure}
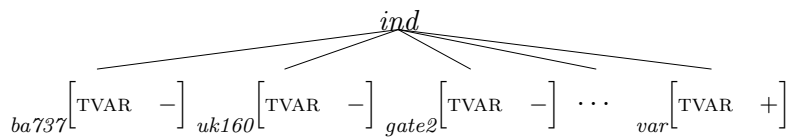

\avmoptions{}
\hrule
\begin{center}
\begin{bundle}{{\srt ind}}
\chunk{\begin{avm}
       \sort{ba737}{
       \[tvar & $-$\]}
       \end{avm}}
\chunk{\begin{avm}
       \sort{uk160}{
       \[tvar & $-$\]}
       \end{avm}}
\chunk{\begin{avm}
       \sort{gate2}{
       \[tvar & $-$\]}
       \end{avm}}
\chunk{\dots}
\chunk{\begin{avm}
       \sort{var}{
       \[tvar & $+$\]}
       \end{avm}}
\end{bundle}
\caption{{\srt ind} and its subsorts -- simplified version}
\label{simple_ind_hierarchy}
\end{center}
\hrule
\end{figure}

In \cite{Pollard2}, feature structures of sort {\srt ind}\/ (called
indices) have the features {\srt person}, {\srt number}, and {\srt
  gender}, which are used to enforce person, number, and gender
agreement. For simplicity, these features are ignored here, and no
agreement checks are made. Pollard and Sag's subsorts of {\srt ind}\/
({\srt ref}, {\srt there}, {\srt it}\/), which are used in \hpsg's
binding theory, are also ignored here. In this thesis, indices
represent \topl terms (they also represent gappy partitioning names,
but let us ignore this temporarily). The situation is roughly
speaking as in figure \ref{simple_ind_hierarchy}. For each \topl
constant (e.g.\ $ba737$, $gate2$), there is a subsort of {\srt ind}\/
that represents that constant. There is also a subsort {\srt var}\/ of
{\srt ind}, whose indices represent \topl variables. A {\feat tvar}
\index{tvar@{\feat tvar} (\hpsg feature, shows if an index represents a \topl variable)}
feature is used to distinguish indices that represent constants from
indices that represent variables. All indices of constant-representing
sorts (e.g.\ {\srt ba737}, {\srt uk160}\/) have their {\feat tvar} set
to $-$. Indices of {\srt var}\/ have their {\feat tvar} set to $+$.

The fact that there is only one subsort ({\srt var}\/) for \topl
variables in figure \ref{simple_ind_hierarchy} does not mean that only
one \topl variable can be represented. {\srt var} is a \emph{sort} of
feature structures, containing infinitely many feature-structure
members. Although the members of {\srt var} cannot be distinguished by
their feature values (they all have {\feat tvar} set to $+$), they are
still different; i.e.\ they are ``structurally identical'' but not
``token-identical'' (see chapter 1 of \cite{Pollard2}). Each one of
the feature-structure members of {\srt var}\/ represents a different
\topl variable. The subsorts that correspond to \topl constants also
contain infinitely many different feature-structure members. In this
case, however, all members of the same subsort are taken to represent
the same constant. For example, any feature structure of sort {\srt
  gate2} represents the \topl constant $gate2$.


\section{More on the subsorts of ind} \label{more_ind}

The subsorts of {\srt ind}\/ are actually more complicated than in
figure \ref{simple_ind_hierarchy}. Natural language front-ends (e.g.\ 
\textsc{Masque} \cite{Auxerre2}, \textsc{Team} \cite{Grosz1987},
\textsc{Cle} \cite{Alshawi}, \textsc{SystemX} \cite{Cercone1993})
often employ a domain-dependent hierarchy of types of world entities.
This hierarchy is typically used in disambiguation, and to detect
semantically anomalous sentences like \qit{Gate 2 departed from runway
  1}. Here, a hierarchy of this kind is mounted under the {\srt ind}\/
sort. (Examples illustrating the use of this hierarchy are given in
following sections.)

\begin{figure}
\begin{center}
\includegraphics[scale=.6]{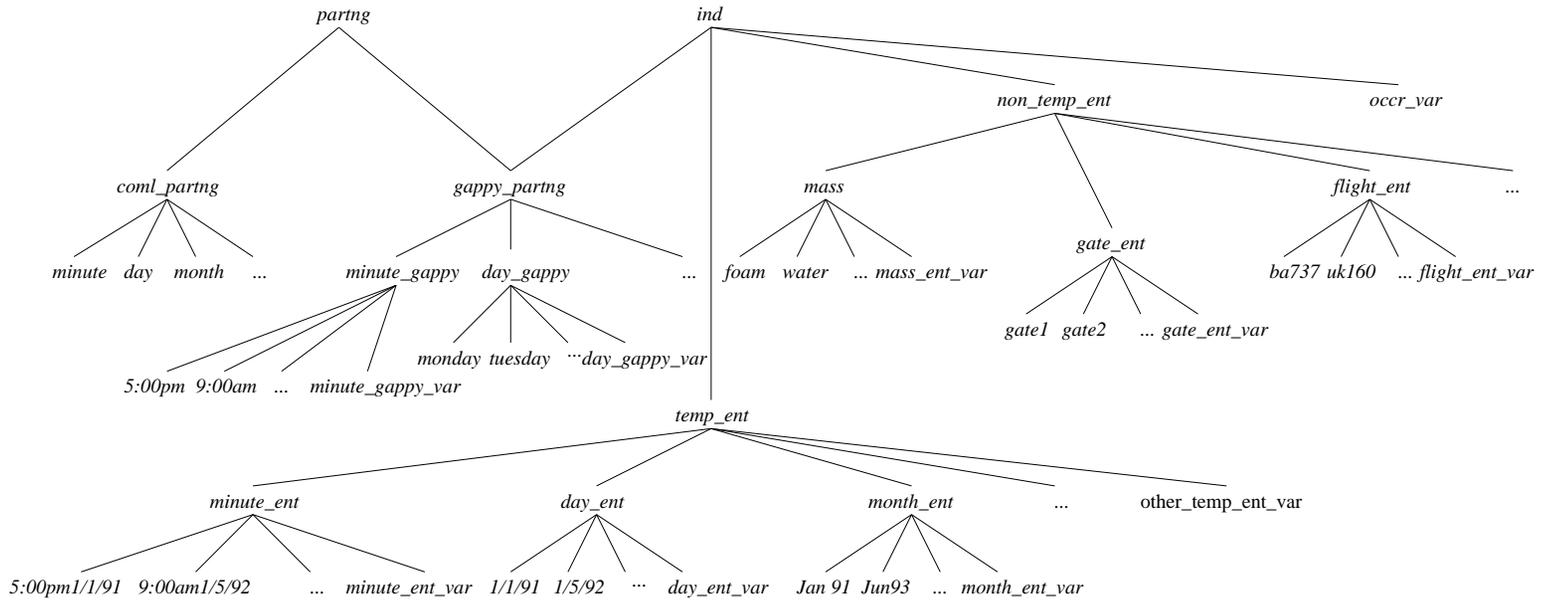}
\caption{{\srt partng}, {\srt ind}, and their subsorts}
\label{ind_hierarchy}
\end{center}
\end{figure}

In the airport domain, there are temporal world entities (the Monday
16/10/95, the year 1995, etc.), and non-temporal world entities
(flight BA737, gate 2, etc.). Indices representing temporal entities
are classified into a {\srt temp\_ent}\/ 
\index{tempent@{\srt temp\_ent}\/ (\hpsg sort, represents temporal
  entities)}
subsort of {\srt ind}, while indices representing non-temporal
entities are classified into {\srt non\_temp\_ent}\/
\index{nontempent@{\srt non\_temp\_ent} (\hpsg sort, represents
  non-temporal entities)}
(see figure \ref{ind_hierarchy}; ignore {\srt partng}\/ and its
subsorts for the moment). {\srt non\_temp\_ent}\/ has in turn
subsorts like {\srt mass}\/ (indices representing mass entities, e.g.\ 
foam or water), {\srt flight\_ent}\/ (indices representing flights,
e.g.\ BA737), etc. {\srt flight\_ent}\/ has one subsort for each
\topl constant that denotes a flight (e.g.\ {\srt ba737}, {\srt
  uk160}\/), plus one sort ({\srt flight\_ent\_var}\/) whose indices
represent \topl variables that denote flights. The other
children-sorts of {\srt non\_temp\_ent}\/ have similar subsorts.

{\srt temp\_ent}\/ has subsorts like {\srt minute\_ent}\/ (indices
representing particular minutes, e.g.\ the 5:00pm minute of 1/1/91),
{\srt day\_ent}\/ (indices representing particular days), etc. {\srt
  minute\_ent}\/ has one subsort for each \topl constant that denotes
a particular minute (e.g.\ {\srt 5:00pm1/1/91}\/), plus one sort
({\srt minute\_ent\_var}\/) whose indices represent \topl variables
that denote particular minutes. The other children-sorts of {\srt
  temp\_ent}\/ have similar subsorts.
The indices of {\srt other\_temp\_ent\_var}\/
\index{othertempentvar@{\srt other\_temp\_ent\_var}\/ (\hpsg sort,
  represents some time-denoting \topl variables)}
(figure \ref{ind_hierarchy}) represent \topl variables that denote
temporal-entities which do not correspond to sister-sorts of {\srt
  other\_temp\_ent\_var}\/ ({\srt minute\_ent}, {\srt day\_ent},
etc.).  This is needed because not all \topl variables denote
particular minutes, days, months, etc. In \pref{lentr:3}, for example,
$e^v$ denotes a past period that covers exactly a taxiing of UK160 to
gate 1 (from start to completion). The taxiing may have started at
5:00pm on 1/1/95, and it may have been completed at 5:05pm on the same
day. In that case, $e^v$ denotes a period that is neither a
minute-period, nor a day-period, nor a month-period, etc.
\begin{examps}
\item $\past[e^v, \culm[taxiing\_to(occr^v,uk160, gate1)]]$ \label{lentr:3}
\end{examps}
{\srt occr\_var}\/ contains indices that represent \topl
variables used as occurrence identifiers (section
\ref{occurrence_ids}).
Indices of sorts that represent \topl constants (e.g.\ {\srt foam}, 
{\srt 5:00pm1/1/91}, {\srt Jun93}\/ in figure \ref{ind_hierarchy}) have
their {\feat tvar}\/
\index{tvar@{\feat tvar} (\hpsg feature, shows if an index represent a \topl variable or not)} 
set to $-$. Indices of sorts that represent \topl
variables (e.g.\ {\srt flight\_ent\_var}, {\srt minute\_ent\_var},
{\srt other\_temp\_ent\_var}, {\srt occr\_var}\/) have their {\feat
tvar}\/ set to $+$. There is also a special sort {\srt now}\/ (not
shown in figure \ref{ind_hierarchy}) that is used to represent the
\topl expression $now^*$ 
\index{now2@{\srt now}\/ (\hpsg sort, represents \topl's $now^*$)}
(section \ref{ntense_op}). 

The sorts of figure \ref{operator_sorts} mirror the definitions of
\topl's operators. For example, the {\srt ntense}\/ sort reflects that
fact that the first argument of an \ntense operator must be $now^*$
or a variable ({\feat tvar}~$+$) denoting a period ({\srt temp\_ent}),
while the second argument must be a yes/no formula ({\srt psoa}). (The
{\srt sem\_num}\/
\index{semnum@{\srt sem\_num}\/ (\hpsg sort, represents numbers)}
sort in {\srt for\_op} is a child-sort of {\srt
  non\_temp\_ent}, with subsorts that represent the numbers 1, 2, 3,
etc. The {\srt compl\_partng}\/ and {\srt gappy\_partng}\/ sorts in
{\srt for\_op}\/ and {\srt part}\/ are discussed below.)

The hierarchy under {\srt ind}\/ is domain-dependent. For example, in
an application where the database contains information about a
company, the subsorts of {\srt non\_temp\_ent}\/ would correspond to
departments, managers, etc.\ I assume, however, that in all
application domains, {\srt ind} would have the children-sorts {\srt
  temp\_ent}, {\srt non\_temp\_ent}, {\srt occr\_var}, {\srt
  gappy\_partng} (to be discussed below), and possibly more. I also
assume that the subsorts of {\srt partng} (see below) and {\srt
  temp\_ent}\/ would have the general form of figure
\ref{ind_hierarchy}, though they would have to be adjusted to reflect
the partitionings and temporal entities used in the particular
application.

I now turn to the {\srt partng}\/ 
\index{partng2@{\srt partng}\/ (\hpsg sort, represents \topl partitioning names)}
sort of figure \ref{ind_hierarchy},
which has the subsorts {\srt compl\_partng}\/ 
\index{complpartng@{\srt compl\_partng}\/ (\hpsg sort, represents
  \topl complete partitioning names)}
and {\srt gappy\_partng}\/ 
\index{gappypartng@{\srt gappy\_partng}\/ (\hpsg sort, represents
  \topl gappy part.\ names and some terms)}
(these three sorts do not exist in
\cite{Pollard2}). For each \topl complete or gappy partitioning name
(e.g.\ $minute^c$, $day^c$, $\text{\textit{5:00pm}}^g$, $monday^g$)
there is a leaf-subsort of {\srt compl\_partng}\/ or {\srt
  gappy\_partng}\/ respectively that represents that name. (The
leaf-subsorts of {\srt gappy\_partng} are also used to represent some
\topl terms; this is discussed below.) In figure \ref{ind_hierarchy},
the sorts {\srt 5:00pm}, {\srt 9:00am}, etc.\ are grouped under {\srt
  minute\_gappy}\/ to reflect the fact that the corresponding
partitionings contain minute-periods. (I assume here that these
partitioning names denote the obvious partitionings.) Similarly, {\srt
  monday}, {\srt tuesday}, etc.\ are grouped under {\srt day\_gappy}\/
to reflect the fact that the corresponding partitionings contain
day-periods. Section \ref{habituals} provides examples where sorts
like {\srt minute\_gappy} and {\srt day\_gappy} prove useful.

Apart from gappy partitioning names, the subsorts of {\srt
  gappy\_partng}\/ are also used to represent \topl terms that denote
generic representatives of partitionings (section \ref{hab_problems}).
(To allow the subsorts of {\srt gappy\_partng}\/ to represent \topl
terms, {\srt gappy\_partng}\/ is not only a subsort of {\srt partng},
but also of {\srt ind}; see figure \ref{ind_hierarchy}.) For example,
\pref{lentr:6} (that expresses the habitual reading of \pref{lentr:5})
is represented as \pref{lentr:7}. In this case, the subsort {\srt
  5:00pm}\/ of {\srt gappy\_partng}\/ represents the \topl constant
\textit{5:00pm}.
\avmoptions{active}
\begin{examps}
\item BA737 departs (habitually) at 5:00pm \label{lentr:5} 
\item $\pres[hab\_departs\_at(ba737, \text{\textit{5:00pm}})]$ \label{lentr:6}
\item \begin{avm}
      \osort{pres}{
      [main\_psoa & \sort{hab\_departs\_at}{
                          [arg1 & \sort{ba737}{
                                  [tvar $-$]} \\
                           arg2 & \sort{5:00pm}{
                                  [tvar $-$]}]}]}
      \end{avm} 
      \label{lentr:7}
\end{examps}
In contrast, \pref{lentr:9} (that expresses the
non-habitual reading of \pref{lentr:8}) is represented as 
\pref{lentr:10}. In this case, the subsort {\srt
5:00pm}\/ of {\srt gappy\_partng}\/ represents the \topl gappy
partitioning name $\text{\textit{5:00pm}}^g$. (It cannot represent a
\topl term, because \topl terms cannot be used as first arguments of
\partop operators.) The \avmbox{1}s in
\pref{lentr:10} mean that the values of {\feat part\_var} and {\feat
  time\_spec} must be token-identical, i.e.\ they must represent the
same \topl variable.
\avmoptions{active}
\begin{examps}
\item BA737 departed (actually) at 5:00pm. \label{lentr:8}
\item $\partop[\text{\textit{5:00pm}}^g, fv^v] \land 
       \at[fv^v, \past[e^v, depart(ba737)]]$ \label{lentr:9}
\item \begin{avm}
      \osort{and}{
      [conjunct1 & \osort{part}{
                   [partng    & \sort{5:00pm}{
                                [tvar $-$]} \\
                    part\_var & \sort{minute\_ent\_var}{
                                [tvar $+$]}@1 ]} \\
       conjunct2 & \osort{at\_op}{
                   [time\_spec & @1 \\
                    main\_psoa & \osort{past}{
                                 [et\_handle & \sort{temp\_ent}{
                                               [tvar $+$]} \\
                                  main\_psoa & \sort{depart}{
                                               [arg1 & \sort{ba737}{
                                                       [tvar $-$]} ]}]}]}]}
      \end{avm}
      \label{lentr:10}
\end{examps}

The {\srt minute\_gappy}\/ and {\srt gappy\_var}\/ sorts of figure
\ref{ind_hierarchy} are used only to represent \topl variables that
denote generic representatives of unknown {\srt minute\_gappy}\/ or
{\srt day\_gappy}\/ partitionings.  The $t^v$ variable of
\pref{lentr:16}, for example, denotes the generic representative of an
unknown {\srt minute\_gappy}\/ partitioning. (If BA737 departs
habitually at 5:00pm, $t^v$ denotes the generic representative of the
$\text{\textit{5:00pm}}^g$ partitioning, the same generic
representative that the \textit{5:00pm} constant of \pref{lentr:6}
denotes.) The $\pres[hab\_departs\_at(ba737, t^v)]$ part of
\pref{lentr:16} is represented as \pref{lentr:17}. (The
feature-structure representation of quantifiers will be discussed in
section \ref{TOP_FS_WH}.)
\begin{examps}
\item When does BA737 (habitually) depart? \label{lentr:15}
\item $?t^v \; \pres[hab\_departs\_at(ba737, t^v)]$ \label{lentr:16}
\item \begin{avm}
      \osort{pres}{
      [main\_psoa & \sort{hab\_departs\_at}{
                          [arg1 & \sort{ba737}{
                                  [tvar $-$]} \\
                           arg2 & \sort{minute\_gappy\_var}{
                                  [tvar $+$]} ]} ]}
      \end{avm}
      \label{lentr:17}
\end{examps}
The indices of sorts like {\srt minute\_gappy\_var}\/ and {\srt
  day\_gappy\_var}\/ have their {\feat tvar} 
\index{tvar@{\feat tvar} (\hpsg feature, shows if an index represent a \topl variable or not)} 
set to $+$. The indices of all other leaf-subsorts of {\srt
  gappy\_partng}\/ (e.g.\ {\srt 5:00pm}, {\srt monday}\/) have their
{\feat tvar} set to $-$.


\section{Representing TOP quantifiers in HPSG}
\label{TOP_FS_WH}

\topl yes/no formulae are represented in the \hpsg version of this
thesis as feature-structures of sort {\srt psoa} (figure
\ref{psoa_fig}). To represent \topl wh-formulae (formulae with
interrogative or interrogative-maximal quantifiers) additional
feature-structure sorts are needed. I discuss these below. 

Feature structures of sort {\srt quant}\/ represent unresolved
quantifiers (quantifiers whose scope is not known yet). They have
two features: {\feat det} and {\feat restind} (restricted index), as
shown in \pref{nps:5}. The {\feat det} feature shows the type of the
quantifier. In this thesis, {\feat det} can have the values {\srt
  exists}\/ (existential quantifier), {\srt interrog}\/
\index{interrog5@{\srt interrog}\/ (\hpsg sort, represents \topl
  interrogative quantifiers)}
(interrogative quantifier), and {\srt interrog\_mxl}\/
\index{interrogmxl5@{\srt interrog\_mxl}\/ (\hpsg sort, represents \topl
  interrogative-maximal quantifiers)}
(interrogative-maximal quantifier). (Apart from the values of {\feat
  det}, {\srt quant}\/ is as in \cite{Pollard2}.)
\begin{examps}
\item 
\avmoptions{active}
\begin{avm}
\sort{quant}{
[det     & exists $\lor$ interrog $\lor$ interrog\_mxl \\
 restind & \osort{nom\_obj}{
           [index & \sort{ind}{
                    [tvar & $+$]} \\
            restr & set\(psoa\)]}]}
\end{avm}
\label{nps:5}
\end{examps}
The values of {\feat restind} are feature structures of sort {\srt
  nom\_obj}\/ (nominal object).\footnote{In \cite{Pollard2}, {\srt
    nom\_obj}\/ has the subsorts {\srt npro}\/ (non-pronoun) and {\srt
    pron}\/ (pronoun). These subsorts are not used in this thesis.}
These have the features {\feat index} (whose values are of sort {\srt
  ind}\/) and {\feat restr} (whose values are sets of {\srt psoa}s). 
When a {\srt nom\_obj}\/ feature structure is the value of {\feat
  restind}, the {\feat index} corresponds to the \topl variable being
quantified, and the {\feat restr} corresponds to the restriction of
the quantifier. (If the {\feat restr} set contains more than one {\srt
  psoa}s, the {\srt psoa}-elements of the set are treated as forming a
conjunction.) For example, \pref{nps:6} represents \pref{nps:7}.
\begin{examps}
\item
\avmoptions{active}
\begin{avm}
\sort{quant}{
[det     & interrog \\
 restind & \osort{nom\_obj}{
           [index & \sort{ind}{
                    [tvar & $+$]}@1 \\
            restr & \{\sort{flight}{
                      [arg1 & @1]} \}]}]}
\end{avm}
\label{nps:6}
\item $?f^v \; flight(f^v)$ \label{nps:7}
\end{examps}
Although \topl does not use explicit existential quantifiers
(universal quantification is not supported, and \topl variables can be
thought of as existentially quantified), the \hpsg version of this
thesis employs explicit existential quantifiers ({\srt quant}s whose
{\feat det} is {\srt exists}\/) for compatibility with
\cite{Pollard2}.  These explicit existential quantifiers are removed
when extracting \topl formulae from signs (this is discussed in
section \ref{extraction_hpsg} below).


\section{Extracting TOP formulae from HPSG signs}
\label{extraction_hpsg}

The parser maps each question to a sign. (Multiple signs are generated
when the parser understands a question to be ambiguous.) For example,
\qit{Which inspector was at gate 2?} is mapped to \pref{nps:14b}
(exactly how \pref{nps:14b} is generated will become clearer in the
following sections; see also the comments about \ntense{s} in section
\ref{non_pred_nps} below).
\begin{examps}
\setbox\avmboxa=\hbox{\begin{avm}
\sort{inspector}{
[arg1 & @1]}
\end{avm}}
\avmoptions{active,center}
\item
\begin{avm}
[\avmspan{phon \; \<\fval which, inspector, was, at, gate2\>} \\
 synsem|loc & [cat & [head  & \osort{verb}{
                              [vform & fin \\
                               aux   & $+$]} \\
                      aspect & lex\_state \\
                      spr    & \<\> \\
                      subj   & \<\>  \\
                      comps  & \<\>] \\
               cont & \osort{past}{
                      [et\_handle & \osort{temp\_ent}{
                                    [tvar & $+$]} \\
                       main\_psoa & \osort{located\_at}{
                                    [arg1 & @1 \\
                                     arg2 & gate2]}]}] \\
 \avmspan{qstore \; \{\sort{quant}{
                      [det & interrog \\
                       restind & \osort{nom\_obj}{
                                  [index & \sort{person\_ent}{
                                           [tvar & $+$]}@1 \\
                                   restr & \{\box\avmboxa\}
                                  ]}
                      ]}\}
         }
]
\end{avm}
\label{nps:14b}
\end{examps}
Apart from the features that were discussed in section
\ref{HPSG_basics}, signs also have the feature {\feat qstore}, whose
values are sets of {\srt quant}s (section \ref{TOP_FS_WH}). The {\feat
  cont} value of signs that correspond to questions is of sort {\srt
  psoa}\/, i.e.\ it represents a \topl yes/no formula. In the \hpsg
version of this thesis, the {\feat qstore} value represents
quantifiers that must be ``inserted'' in front of the formula of
{\feat cont}. In the prototype \nlitdb (to be discussed in chapter
\ref{implementation}), there is an ``extractor'' of \topl formulae
that examines the {\feat cont} and {\feat qstore} features of the
question's sign, and generates the corresponding \topl formula. This
is a trivial process, which I discuss only at an abstract level: the
extractor first examines recursively the features and feature values
of {\feat cont}, rewriting them in term notation (in \pref{nps:14b},
this generates \pref{nps:15}); then, for each element of {\feat
  qstore}, the extractor adds a suitable quantifier in front of the
formula of {\feat cont} (in \pref{nps:14b}, this transforms
\pref{nps:15} into \pref{nps:17}).
\begin{examps}
\item $\past[e^v, located\_at(p^v, gate2)]$ \label{nps:15}
\item $?p^v \; inspector(p^v) \land
       \past[e^v, located\_at(p^v, gate2)]$ \label{nps:17}
\end{examps}
In the case of elements of {\feat qstore} that correspond to
existential quantifiers, no explicit existential quantifier is added
to the formula of {\feat cont} (only the expression that corresponds
to the {\feat restr} of the {\srt quant}-element is added). For example,
if the {\feat det} of \pref{nps:14b} were {\srt exists}, \pref{nps:17}
would be \pref{nps:17b}.
\begin{examps}
\item $inspector(p^v) \land
       \past[e^v, located\_at(p^v, gate2)]$ \label{nps:17b}
\end{examps}

The extracted formula then undergoes an additional post-processing
phase (to be discussed in section \ref{post_processing}). This is a
collection of transformations that need to be applied to some of the
extracted formulae. (In \pref{nps:17} and \pref{nps:17b}, the
post-processing has no effect.)


\section{Verb forms} \label{hpsg:verb_forms}

I now present the treatment of the various linguistic constructs,
starting from verb forms (simple present, past continuous, etc.).
(Pollard and Sag do not discuss temporal linguistic mechanisms.)

\subsection{Single-word verb forms} \label{single_word_forms}

Let us first examine the lexical rules that generate signs for
(single-word) non-base verb forms from signs for base forms. The
signs for simple present forms are generated by \pref{vforms:1}.
\begin{examps}
\item \lexrule{Simple Present Lexical Rule:}
\begin{center}
\avmoptions{active}
\begin{avm}
[\avmspan{phon \; \<$\lambda$\>} \\
 synsem|loc & [cat  &  [head   & \osort{verb}{
                                 [vform & bse \\
                                  aux   & $-$]} \\
                        aspect & lex\_state] \\
               cont & @1]]
\end{avm}
\\
$\Downarrow$
\\
\begin{avm}
[\avmspan{phon \; \<\fval morph\($\lambda$, simple\_present\)\>} \\
 synsem|loc & [cat  & [head   & \osort{verb}{
                                [vform & fin \\
                                 aux   & $-$]} \\
                       aspect & lex\_state] \\
               cont & \sort{pres}{
                      [main\_psoa & @1]} ]]
\end{avm}
\end{center}
\label{vforms:1}
\end{examps}
\pref{vforms:1} means that for each lexical sign that matches the
first feature structure (the ``left hand side'', LHS) of the rule, a
new lexical sign should be generated as shown in the second feature
structure (the ``right hand side'', RHS) of the rule. (Following
standard \hpsg notation, I write {\feat synsem$\mid$loc} to refer to
the {\feat loc} feature of the value of {\feat synsem}.) The {\feat
  head}s of the LHS and RHS mean that the original sign must
correspond to the base form of a non-auxiliary verb (auxiliary verbs
are treated separately), and that the resulting sign corresponds to a
finite verb form (a form that does not need to combine with an
auxiliary verb). The {\feat cont} of the new sign is the same as the
{\feat cont} of the original one, except that it contains an
additional \pres operator. Features of the original sign not shown in
the LHS (e.g.\ {\feat subj}, {\feat comps}) have the same
values in the generated sign. \pref{vforms:1} requires the original
sign to correspond to a (lexical) state base form. No simple present
signs are generated for verbs whose base forms are not states. This is
in accordance with the assumption of section \ref{simple_present} that
the simple present can be used only with state verbs.

$morph(\lambda, simple\_present)$ denotes a morphological
transformation that generates the simple present form (e.g.\ 
\qit{contains}) from the base form (e.g.\ \qit{contain}). The
prototype \nlitdb actually employs two different simple present
lexical rules. These generate signs for singular and plural simple
present forms respectively. As mentioned in sections
\ref{ling_not_supported} and \ref{TOP_FS}, plurals are treated
semantically as singulars, and no number-agreement checks are made.
Hence, the two lexical rules differ only in the {\feat phon} values of
the generated signs.

\pref{vforms:2} shows the base form sign of \qit{to contain} in the
airport domain. From \pref{vforms:2}, \pref{vforms:1} generates
\pref{vforms:3}. The {\srt tank\_ent}\/ and {\srt mass\_ent}\/ in 
\pref{vforms:2} and \pref{vforms:3} mean that the
indices introduced by the subject and the object must be of sort {\srt
tank\_ent} and {\srt mass\_ent}\/ respectively ({\srt tank\_ent}\/ is
a sister of {\srt flight\_ent}\/ in figure \ref{ind_hierarchy}). Hence,
the semantically anomalous \qit{Gate 2 contains water.}
(where the subject introduces an index of sort {\srt gate2}, which is
not a subsort of {\srt tank\_ent}\/) would be rejected. All lexical
signs of verb forms have their {\feat qstore} set to $\{\}$. For
simplicity, I do not show the {\feat qstore} feature here.
\begin{examps}
\item 
\avmoptions{active}
\begin{avm}
[\avmspan{phon \; \<\fval contain\>} \\
 synsem|loc & [cat & [head  & \osort{verb}{
                              [vform & bse \\
                               aux   & $-$ ]} \\
                      aspect & lex\_state \\
                      spr    & \<\> \\
                      subj   & \<\feat np[-prd]$_{tank\_ent@1}$\>  \\
                      comps  & \<\feat np[-prd]$_{mass\_ent@2}$\> ]\\
               cont & \sort{contains}{
                      [arg1 & @1 \\
                       arg2 & @2]}]]
\end{avm}
\label{vforms:2}
\item 
\begin{avm}
[\avmspan{phon \; \<\fval contains\>} \\
 synsem|loc & [cat & [head  & \osort{verb}{
                              [vform & fin \\
                               aux   & $-$ ]} \\
                      aspect & lex\_state \\
                      spr    & \<\> \\
                      subj   & \<\feat np[-prd]$_{tank\_ent@1}$\>  \\
                      comps  & \<\feat np[-prd]$_{mass\_ent@2}$\> ]\\
               cont & \osort{pres}{
                      [main\_psoa & \osort{contains}{
                                    [arg1 & @1 \\
                                     arg2 & @2]}]}]]
\end{avm}
\label{vforms:3}
\end{examps}

The simple past signs of culminating activity verbs are generated by
\pref{vforms:4}, shown below. The simple past signs of non-culminating activity
verbs are generated by a lexical rule that is similar to
\pref{vforms:4}, except that it does not introduce a \culm operator in
the resulting sign. 

The signs of past participles (e.g.\ \qit{inspected} in \qit{Who had 
  inspected BA737?}) are generated by two lexical rules
which are similar to the simple past ones. There is a rule for
culminating activity verbs (which introduces a \culm in the past
participle sign), and a rule for non-culminating activity verbs (that
introduces no \culm). Both rules do not introduce \past operators. The
generated signs have their {\feat vform} set to {\fval psp}\/ (past
participle), and the same {\feat aspect} as the base signs, i.e.\ 
their {\feat aspect} is not changed to {\fval cnsq\_state} (consequent
state). The shift to consequent state takes place when the auxiliary
\qit{had} combines with the past participle (this will be discussed in
section \ref{multi_forms}).
\newpage
\begin{examps}
\item 
\lexrule{Simple Past Lexical Rule (Culminating Activity Base Form):}
\avmoptions{active}
\begin{center}
\begin{avm}
[\avmspan{phon \; \<$\lambda$\>} \\
 synsem|loc & [cat  &  [head   & \sort{verb}{
                                 [vform & bse \\
                                  aux   & $-$]} \\
                        aspect & culmact] \\
               cont & @1]]
\end{avm}
\\
$\Downarrow$
\\
\begin{avm}
[\avmspan{phon \; \<\fval morph\($\lambda$, simple\_past\)\>} \\
 synsem|loc & [cat  & [head   & \sort{verb}{
                                [vform & fin \\
                                 aux   & $-$]} \\
                       aspect & culmact] \\
               cont & \osort{past}{
                      [et\_handle & \sort{temp\_ent}{
                                    [tvar $+$]} \\
                       main\_psoa & \sort{culm}{
                                    [main\_psoa & @1]} ]} ]]
\end{avm}
\end{center}
\label{vforms:4}
\end{examps}

The signs for present participles (e.g.\ \qit{servicing} in
\qit{Which company is servicing BA737?}) are generated by
\pref{vforms:10}. The present participle signs are the same
as the base ones, except that their {\feat vform} is {\fval
prp}\/ (present participle), and their {\feat aspect} is {\srt
progressive}\/ (progressive state). 
\begin{examps}
\item
\lexrule{Present Participle Lexical Rule:} 
\avmoptions{active}
\begin{center}
\begin{avm}
[\avmspan{phon \; \<$\lambda$\>} \\
 synsem|loc & [cat  &  [head   & \sort{verb}{
                                 [vform & bse \\
                                  aux   & $-$]} \\
                        aspect & aspect] \\
               cont & @1]]
\end{avm}
\\
$\Downarrow$
\\
\begin{avm}
[\avmspan{phon \; \<\fval morph\($\lambda$, present\_participle\)\>} \\
 synsem|loc & [cat  & [head   & \sort{verb}{
                                [vform & prp \\
                                 aux   & $-$]} \\
                       aspect & progressive] \\
               cont & @1]]
\end{avm}
\end{center}
\label{vforms:10}
\end{examps}

Gerund signs are generated by a lexical rule that is similar to
\pref{vforms:10}, except that the generated signs retain the {\feat
  aspect} of the original ones, and have their {\feat vform} set to
{\srt ger}. In English, there is no morphological distinction between
gerunds and present participles. \hpsg and most traditional grammars
(e.g.\ \cite{Thomson}), however, distinguish between the two. In
\pref{vforms:10x1}, the \qit{inspecting} is the gerund of \qit{to
  inspect}, while in \pref{vforms:10x2}, the \qit{inspecting} is
the present participle.
\begin{examps}
\item J.Adams finished inspecting BA737. \label{vforms:10x1}
\item J.Adams was inspecting BA737. \label{vforms:10x2}
\end{examps}
The fact that gerund signs retain the {\feat aspect} of the base signs
is used in the treatment of \qit{to finish} (section
\ref{special_verbs}). The simple past \qit{finished} receives multiple
signs. (These are generated from corresponding base form signs by the
simple past lexical rules.) \pref{vforms:14} is used when
\qit{finished} combines with a culminating activity verb phrase, and
\pref{vforms:23} when it combines with a state or activity verb
phrase.
\avmoptions{active}
\begin{examps}
\item 
\begin{avm}
[\avmspan{phon \; \<\fval finished\>} \\
 synsem|loc & [cat & [head  & \osort{verb}{
                              [vform & fin \\
                               aux   & $-$ ]} \\
                      aspect & point \\
                      spr    & \<\> \\
                      subj   & \<@1\>  \\
                      comps  & \<\feat 
                                 vp[subj \<@1\>, vform {\fval ger}, 
                                    aspect {\fval culmact}]:@2
                               \> ]\\ 
               cont & \osort{past}{
                      [et\_handle & \sort{temp\_ent}{
                                    [tvar $+$]} \\
                       main\_psoa & \osort{end}{
                                    [main\_psoa & \sort{culm}{
                                                  [main\_psoa & @2]}]}]}]]
\end{avm}
\label{vforms:14}
\item
\begin{avm}
[\avmspan{phon \; \<\fval finished\>} \\
 synsem|loc & [cat & [head  & \osort{verb}{
                              [vform & fin \\
                               aux   & $-$ ]} \\
                      aspect & point \\
                      spr    & \<\> \\
                      subj   & \<@1\>  \\
                      comps  & \<\feat 
                                 vp[subj \<@1\>, vform {\fval ger}, \\
                                    aspect {\fval state $\lor$ activity}]:@2
                               \> ]\\ 
               cont & \osort{past}{
                      [et\_handle & \sort{temp\_ent}{
                                    [tvar $+$]} \\
                       main\_psoa & \sort{end}{
                                    [main\_psoa & @2]}]}]]
\end{avm}
\label{vforms:23}
\end{examps}
In \pref{vforms:14}, the {\feat vp[subj $<$\avmbox{1}$>$, vform {\fval
    ger}, aspect {\fval culmact}]:\avmbox{2}} means that
\qit{finished} requires as its complement a gerund verb phrase (a
gerund that has combined with its complements but not its subject)
whose aspect is culminating activity. The \avmbox{1} of {\feat comps}
points to a description of the required subject of the gerund verb
phrase, and the \avmbox{2} is a pointer to the {\feat cont} value of
the sign of the gerund verb phrase. The two \avmbox{1}s in
\pref{vforms:14} have the effect that \qit{finished} requires as its
subject whatever the gerund verb phrase requires as its subject. The
two \avmbox{2}s cause the sign of \qit{finished} to inherit the {\feat
  cont} value of the sign of the gerund verb phrase, but with
additional \past, \lend, and \culm operators. \pref{vforms:23} is
similar, but it introduces no \culm.

In \pref{vforms:10x1}, the sign of the gerund \qit{inspecting} retains
the {\feat aspect} of the base sign, which in the airport domain is
{\srt culmact}. The sign of the gerund verb phrase \qit{inspecting
  BA737} inherits the {\srt culmact}\/ {\feat aspect} of the gerund
sign (following the aspect principle, to be discussed in section
\ref{hpsg:punc_adv}). Hence, \pref{vforms:14} is used. This causes
\pref{vforms:10x1} to receive a sign whose {\feat cont} represents
\pref{vforms:24x1}, which requires the inspection to have been completed.
\begin{examps}
\item $\past[e^v, \lend[\culm[inspecting(occr^v, jadams, ba737)]]]$
  \label{vforms:24x1} 
\end{examps}
In \pref{vforms:24}, the sign of \qit{circling} inherits the {\srt
  activity}\/ {\feat aspect} of the base sign, causing
\pref{vforms:23} to be used. This leads to \pref{vforms:25}, which
does not require any completion to have been reached.
\begin{examps}
\item BA737 finished circling. \label{vforms:24}
\item $\past[e^v, \lend[circling(ba737)]$ \label{vforms:25}
\end{examps}

There is also a sign of the simple past \qit{finished} for the case
where the gerund verb phrase is a point. In that case, the {\feat
  cont} of the sign of \qit{finished} is identical to the the {\feat
  cont} of the sign of the gerund verb phrase, i.e.\ the
\qit{finished} has no semantic contribution. This is in accordance
with the arrangements of section \ref{special_verbs}.  The signs of
\qit{started}, \qit{stopped}, and \qit{began} are similar, except that
they introduce \lbegin operators instead of \lend ones. Unlike
\qit{finished}, the signs of \qit{stopped} do not introduce \culm
operators when \qit{stopped} combines with culminating activities,
reflecting the fact that there is no need for a completion to have
been reached.

\subsection{Auxiliary verbs and multi-word verb forms} \label{multi_forms}

I now move on to auxiliary verbs and multi-word verb forms (e.g.\ 
\qit{had departed}, \qit{is inspecting}). \pref{vforms:30} shows the
sign of the simple past auxiliary \qit{had}. According to
\pref{vforms:30}, \qit{had} requires as its complement a past
participle verb phrase. The \avmbox{1}s mean that \qit{had}
requires as its subject whatever the past participle verb phrase
requires as its subject. The \avmbox{2}s mean that the {\feat
  main\_psoa} value of the {\srt perf}\/ is the {\feat cont} value of
the sign of the past participle verb phrase.
\begin{examps}
\avmoptions{active}
\item 
\begin{avm}
[\avmspan{phon \; \<\fval had\>} \\
 synsem|loc & [cat & [head  & \osort{verb}{
                              [vform & fin \\
                               aux   & $+$ ]} \\
                      aspect & cnsq\_state \\
                      spr    & \<\> \\
                      subj   & \<@1\>  \\
                      comps  & \<\feat 
                                 vp[subj \<@1\>, vform {\fval psp}]:@2
                               \> ]\\ 
               cont & \osort{past}{
                      [et\_handle & \sort{temp\_ent}{
                                    [tvar $+$]} \\
                       main\_psoa & \osort{perf}{
                                    [et\_handle & \sort{temp\_ent}{
                                                  [tvar $+$]} \\
                                     main\_psoa & @2]}]}]]
\end{avm}
\label{vforms:30}
\end{examps}
In the airport domain, the past participle \qit{departed} receives
multiple signs (for various habitual and non-habitual uses; these
signs are generated from the corresponding base form signs by the
lexical rules of section \ref{single_word_forms}). The sign of
\pref{vforms:32} is used in \pref{vforms:31}. 
\begin{examps}
\avmoptions{active}
\item BA737 had departed. \label{vforms:31}
\item 
\begin{avm}
[\avmspan{phon \; \<\fval departed\>} \\
 synsem|loc & [cat & [head  & \osort{verb}{
                              [vform & psp \\
                               aux   & $-$ ]} \\
                      aspect & point \\
                      spr    & \<\> \\
                      subj   & \<\feat np[-prd]$_{flight\_ent@3}$\>  \\
                      comps  & \<\> ]\\ 
               cont & \sort{actl\_depart}{
                      [arg1 & @3]}]]
\end{avm}
\label{vforms:32}
\end{examps}
According to \pref{vforms:32}, \qit{departed} requires no complements,
i.e.\ it counts as a verb phrase, and can be used as the complement of
\qit{had}. When \qit{had} combines with \qit{departed}, the {\feat
  subj} of \pref{vforms:30} becomes the same as the {\feat subj} of
\pref{vforms:32} (because of the \avmbox{1}s in \pref{vforms:30}), and
the {\feat main\_psoa} of the {\srt perf}\/ in \pref{vforms:30}
becomes the same as the {\feat cont} of \pref{vforms:32} (because of
the \avmbox{2}s in \pref{vforms:30}). The resulting constituent
\qit{had departed} receives \pref{vforms:33}.
\begin{examps}
\avmoptions{active}
\item 
\begin{avm}
[\avmspan{phon \; \<\fval had, departed\>} \\
 synsem|loc & [cat & [head  & \osort{verb}{
                              [vform & fin \\
                               aux   & $+$ ]} \\
                      aspect & cnsq\_state \\
                      spr    & \<\> \\
                      subj   & \<\feat np[-prd]$_{flight\_ent@3}$\>  \\
                      comps  & \<\> ]\\ 
               cont & \osort{past}{
                      [et\_handle & \sort{temp\_ent}{
                                    [tvar $+$]} \\
                       main\_psoa & \osort{perf}{
                                    [et\_handle & \sort{temp\_ent}{
                                                  [tvar $+$]} \\
                                     main\_psoa & \sort{actl\_depart}{
                                                  [arg1 & @3]}]}]}]]
\end{avm}
\label{vforms:33}
\end{examps}
The \hpsg principles (including the semantics and aspect principles
that will be discussed in sections \ref{non_pred_nps} and
\ref{hpsg:punc_adv}) cause \pref{vforms:33} to inherit the {\feat
  head}, {\feat aspect}, {\feat spr}, {\feat subj}, and {\feat cont}
values of \pref{vforms:30}. Notice that this causes the aspect of
\qit{had departed} to become consequent state (\qit{departed} was a
point). As will be discussed in section \ref{hpsg:nouns}, the proper
name \qit{BA737} contributes an index that represents the flight
BA737. When \qit{had departed} combines with its subject \qit{BA737},
the index of \qit{BA737} becomes the {\feat arg1} value of
\pref{vforms:33} (because of the \avmbox{3}s of \pref{vforms:33}).
This causes \pref{vforms:31} to receive a sign whose {\feat cont}
represents \pref{vforms:35}.
\begin{examps}
\item $\past[e1^v, \perf[e2^v, actl\_depart(ba737)]]$ \label{vforms:35}
\end{examps}

As mentioned in sections \ref{present_perfect} and \ref{perf_op},
present perfect forms are treated semantically as simple past forms.
This is why, unlike the sign of \qit{had}, the sign of \qit{has}
(shown in \pref{vforms:36}) does not introduce a \perf operator, and
preserves the aspect of the past participle. This causes \qit{BA737
  has departed.} to receive the same \topl formula as \qit{BA737
  departed.}. 
\begin{examps}
\avmoptions{active}
\item 
\begin{avm}
[\avmspan{phon \; \<\fval has\>} \\
 synsem|loc & [cat & [head  & \osort{verb}{
                              [vform & fin \\
                               aux   & $+$ ]} \\
                      aspect & @1 \\
                      spr    & \<\> \\
                      subj   & \<@2\>  \\
                      comps  & \<\feat 
                                 vp[subj \<@2\>, vform {\fval psp},
                                    aspect @1]:@3
                               \> ]\\ 
               cont & \osort{past}{
                      [et\_handle & \sort{temp\_ent}{
                                    [tvar $+$]} \\
                       main\_psoa & @3]}]]
\end{avm}
\label{vforms:36}
\end{examps}

\qit{Does} receives the sign of \pref{vforms:40.1}, which indicates
that it requires as its complement a base verb phrase. The verb phrase
must be a (lexical) state. (This is in accordance
with the assumption of section \ref{simple_present} that the simple
present can be used only with state verbs.) \pref{vforms:40.1} and the
(habitual) base sign of \pref{vforms:41.1} cause \pref{vforms:41} to
receive \pref{vforms:42}. 
\avmoptions{active}
\begin{examps}
\item
\begin{avm}
[\avmspan{phon \; \<\fval does\>} \\
 synsem|loc & [cat & [head  & \osort{verb}{
                              [vform & fin \\
                               aux   & $+$ ]} \\
                      aspect & lex\_state @1 \\
                      spr    & \<\> \\
                      subj   & \<@2\>  \\
                      comps  & \<\feat 
                              vp[subj \<@2\>, vform {\fval bse}, aspect @1]:@3
                               \> ]\\ 
               cont & \sort{pres}{
                      [main\_psoa & @3]}]]
\end{avm}
\label{vforms:40.1}
\item 
\begin{avm}
[\avmspan{phon \; \<\fval service\>} \\
 synsem|loc & [cat & [head  & \osort{verb}{
                              [vform & bse \\
                               aux   & $-$ ]} \\
                      aspect & lex\_state \\
                      spr    & \<\> \\
                      subj   & \<\feat np[-prd]$_{company\_ent@4}$\>  \\
                      comps  & \<\feat np[-prd]$_{flight\_ent@5}$\> ]\\ 
               cont & \sort{hab\_servicer\_of}{
                      [arg1 & @4 \\
                       arg2 & @5]} ]]
\end{avm}
\label{vforms:41.1}
\item Does Airserve service BA737? \label{vforms:41}
\item 
\begin{avm}
[\avmspan{phon \; \<\fval Does, Airserve, service, BA737\>} \\
 synsem|loc & [cat & [head  & \osort{verb}{
                              [vform & bse \\
                               aux   & $+$ ]} \\
                      aspect & lex\_state \\
                      spr    & \<\> \\
                      subj   & \<\>  \\
                      comps  & \<\> ]\\ 
               cont & \osort{pres}{
                      [main\_psoa & \sort{hab\_servicer\_of}{
                                    [arg1 & airserve \\
                                     arg2 & ba737]}]}]]
\end{avm}
\label{vforms:42}
\end{examps}
In the airport domain, the base form of \qit{to service} receives also
a sign that corresponds to the non-habitual homonym. This is similar
to \pref{vforms:41.1}, but it introduces the predicate functor
$actl\_servicing$, and its {\feat aspect} is {\srt culmact}. This sign
cannot be used in \pref{vforms:41}, because \pref{vforms:40.1}
requires the verb-phrase complement to be a state not a culminating
activity. This correctly predicts that \pref{vforms:41} cannot be
asking if Airserve is actually servicing BA737 at the present moment.

\qit{Did} receives two signs: one for culminating-activity
verb-phrase complements (shown in \pref{vforms:43}), and one for
state, activity, or point verb-phrase complements (this is similar to
\pref{vforms:43}, but introduces no \culm). In both cases, a \past
operator is added. In the case of culminating-activity complements, a
\culm operator is added as well.
\avmoptions{active}
\begin{examps}
\item 
\begin{avm}
[\avmspan{phon \; \<\fval did\>} \\
 synsem|loc & [cat & [head  & \osort{verb}{
                              [vform & fin \\
                               aux   & $+$ ]} \\
                      aspect & culmact @1 \\
                      spr    & \<\> \\
                      subj   & \<@2\>  \\
                      comps  & \<\feat 
                             vp[subj \<@2\>, vform {\fval bse}, aspect @1]:@3
                               \> ]\\ 
               cont & \osort{past}{
                      [et\_handle & \sort{temp\_ent}{
                                    [tvar $+$]} \\
                       main\_psoa & \sort{culm}{
                                    [main\_psoa & @3]}]}]]
\end{avm}
\label{vforms:43}
\end{examps}
The non-habitual sign of \qit{service} and \pref{vforms:43} cause
\pref{vforms:45} to be mapped to \pref{vforms:46}, which requires
Airserve to have actually serviced BA737 in the past. The habitual
sign of \pref{vforms:41.1} and the \qit{did} sign for non-culminating
activity complements cause \pref{vforms:45} to be mapped to
\pref{vforms:46x1}, which requires Airserve to have been a past habitual
servicer of BA737. 
\begin{examps}
\item Did Airserve service BA737? \label{vforms:45}
\item $\past[e^v, \culm[actl\_servicing(occr^v, airserve, ba737)]]$ 
  \label{vforms:46}
\item $\past[e^v, hab\_servicer\_of(airserve, ba737)]$ \label{vforms:46x1}
\end{examps}

The sign for the auxiliary \qit{is} is shown in \pref{vforms:50}. The
present participle \qit{servicing} receives two signs, a non-habitual
one (shown in \pref{vforms:51}) and a habitual one. The latter is
similar to \pref{vforms:51}, but it introduces the functor
$hab\_servicer\_of$, and its {\feat aspect} is {\srt lex\_state}. (The
two present participle signs are generated from the base ones by the
present participle lexical rule of section \ref{single_word_forms}.)
\pref{vforms:50} and \pref{vforms:51} cause \pref{vforms:52} to be
mapped to \pref{vforms:53}, which requires Airserve to be actually
servicing BA737 at the present. \pref{vforms:50} and the habitual
present participle sign cause \pref{vforms:52} to be mapped to
\pref{vforms:53x1}, which requires Airserve to be the current habitual
servicer of BA737. 
\avmoptions{active}
\begin{examps}
\item
\begin{avm}
[\avmspan{phon \; \<\fval is\>} \\
 synsem|loc & [cat & [head  & \osort{verb}{
                              [vform & fin \\
                               aux   & $+$ ]} \\
                      aspect & progressive \\
                      spr    & \<\> \\
                      subj   & \<@1\>  \\
                      comps  & \<\feat 
                                 vp[subj \<@1\>, vform {\fval prp}]:@2
                               \> ]\\ 
               cont & \sort{pres}{
                      [main\_psoa & @2]}]]
\end{avm}
\label{vforms:50}
\item 
\begin{avm}
[\avmspan{phon \; \<\fval servicing\>} \\
 synsem|loc & [cat & [head  & \osort{verb}{
                              [vform & prp \\
                               aux   & $-$ ]} \\
                      aspect & culmact \\
                      spr    & \<\> \\
                      subj   & \<\feat np[-prd]$_{company\_ent@1}$\>  \\
                      comps  & \<\feat np[-prd]$_{flight\_ent@2}$\> ]\\ 
               cont & \sort{actl\_servicing}{
                      [arg1 & occr\_var \\
                       arg2 & @1 \\
                       arg3 & @2]} ]]

\end{avm}
\label{vforms:51}
\item Airserve is servicing BA737. \label{vforms:52}
\item $\pres[actl\_servicing(occr^v, airserve, ba737)]$
  \label{vforms:53}
\item $\pres[hab\_servicer\_of(airserve, ba737)]$ \label{vforms:53x1}
\end{examps}

The sign for the auxiliary \qit{was} is similar to \pref{vforms:50},
except that it introduces a \past operator instead of a \pres one.


\section{Predicative and non-predicative prepositions} \label{hpsg:pps}

\avmoptions{active}

Following Pollard and Sag (\cite{Pollard1}, p.65), prepositions
receive separate signs for their predicative and non-predicative uses.
In sentences like \pref{pps:3} and \pref{pps:4}, where the
prepositions introduce complements of \qit{to be}, the prepositions
are said to be predicative. In \pref{pps:1} and \pref{pps:2}, where
they introduce complements of other verbs, the prepositions are
non-predicative.
\begin{examps}
\item BA737 is at gate 2. \label{pps:3} 
\item BA737 was on runway 3. \label{pps:4}
\item BA737 (habitually) arrives at gate 2. \label{pps:1}
\item BA737 landed on runway 3. \label{pps:2}
\end{examps}
Predicative prepositions introduce their own \topl predicates, while
non-predicative prepositions have no semantic contribution.

\subsection{Predicative prepositions}

\pref{pps:5} shows the predicative sign of \qit{at}. (The predicative
signs of other prepositions are similar.) The {\feat prd}~$+$ shows
that the sign is predicative. ({\feat prd} is also used to distinguish
predicative adjectives and nouns; this will be discussed in sections
\ref{hpsg:nouns} and \ref{hpsg:adjectives}.) {\feat pform} reflects
the preposition to which the sign corresponds. Signs for prepositional
phrases inherit the {\feat pform} of the preposition's sign. This is
useful in verbs that require prepositional phrases introduced by
particular prepositions.
\begin{examps}
\item 
\begin{avm}
[\avmspan{phon \; \<\fval at\>} \\
 synsem|loc & [cat & [head  & \osort{prep}{
                              [pform & at \\
                               prd   & $+$]} \\
                      spr    & \<\> \\
                      subj   & \<\feat np[-prd]$_{@1}$\>  \\
                      comps  & \<\feat np[-prd]$_{@2}$\> ]\\ 
               cont & \sort{located\_at}{
                      [arg1 & @1 \\
                       arg2 & non\_temp\_ent@2 ]}]]
\end{avm}
\label{pps:5}
\end{examps}
According to \pref{pps:5}, \qit{at} requires a (non-predicative)
noun-phrase (\qit{BA737} in \pref{pps:3}) as its subject, and another
one (\qit{gate 2} in \pref{pps:3}) as its complement. As will be
discussed in section \ref{hpsg:nouns}, \qit{BA737} and \qit{gate 2}
contribute indices that represent the corresponding world entities.
The \avmbox{2} of \pref{pps:5} denotes the index of \qit{gate 2}.
\pref{pps:5} causes \qit{at gate 2} to receive \pref{pps:6}.
\begin{examps}
\item 
\begin{avm}
[\avmspan{phon \; \<\fval at, gate2\>} \\
 synsem|loc & [cat & [head  & \osort{prep}{
                              [pform & at \\
                               prd   & $+$]} \\
                      spr    & \<\> \\
                      subj   & \<\feat np[-prd]$_{@1}$\>  \\
                      comps  & \<\> ]\\ 
               cont & \sort{located\_at}{
                      [arg1 & @1 \\
                       arg2 & gate2 ]}]]
\end{avm}
\label{pps:6}
\end{examps}
Apart from \pref{vforms:50} (which is used when \qit{is}
combines with a present-participle complement), \qit{is} also receives
\pref{pps:7} (which is used when \qit{is} combines with
predicative prepositional-phrases). 
\begin{examps}
\item
\begin{avm}
[\avmspan{phon \; \<\fval is\>} \\
 synsem|loc & [cat & [head  & \osort{verb}{
                              [vform & fin \\
                               aux   & $+$ ]} \\
                      aspect & lex\_state \\
                      spr    & \<\> \\
                      subj   & \<@3\>  \\
                      comps  & \<\feat pp[subj \<@3\>, prd $+$]:@4 \> ]\\ 
               cont & \sort{pres}{
                      [main\_psoa & @4]}]]
\end{avm}
\label{pps:7}
\end{examps}
According to \pref{pps:7}, \qit{is} requires as its complement a predicative
prepositional phrase (a predicative preposition that has combined with
its complements but not its subject), like the \qit{at gate 2} of
\pref{pps:6}. \pref{pps:6} and \pref{pps:7} cause \pref{pps:3} to
receive \pref{pps:10}. 
\begin{examps}
\item
\begin{avm}
[\avmspan{phon \; \<\fval BA737, is, at, gate2\>} \\
 synsem|loc & [cat & [head  & \osort{verb}{
                              [vform & fin \\
                               aux   & $+$]} \\
                      aspect & lex\_state \\
                      spr    & \<\> \\
                      subj   & \<\>  \\
                      comps  & \<\>] \\ 
               cont & \osort{pres}{
                      [main\_psoa & \osort{located\_at}{
                                    [arg1 & ba737 \\
                                     arg2 & gate2]}]}]]
\end{avm}
\label{pps:10}
\end{examps}

Like \qit{is}, \qit{was} receives two signs: one for
present-participle complements (as in \qit{BA737 was circling.}), and
one for predicative prepositional-phrase complements (as in
\pref{pps:4}). These are similar to the signs of \qit{was}, but they
introduce \past operators rather than \pres ones.

\subsection{Non-predicative prepositions}

The non-predicative sign of \qit{at} is shown in \pref{pps:12}. (The
non-predicative signs of other prepositions are similar.) The
\avmbox{1} is a pointer to the {\feat cont} value of the sign that
corresponds to the noun-phrase complement of \qit{at}. Notice that in
this case the \qit{at} has no semantic contribution (the \qit{at} sign
simply copies the {\feat cont} of the noun-phrase complement).
\begin{examps}
\item 
\begin{avm}
[\avmspan{phon \; \<\fval at\>} \\
 synsem|loc & [cat & [head  & \osort{prep}{
                              [pform & at \\
                               prd   & $-$]} \\
                      spr    & \<\> \\
                      subj   & \<\>  \\
                      comps  & \<\feat np[-prd]:@1\> ]\\ 
               cont & @1]]
\end{avm}
\label{pps:12}
\end{examps}
\pref{pps:12} and the habitual sign of \qit{arrives} of \pref{pps:13}
cause \pref{pps:1} to receive \pref{pps:16}. 
\begin{examps}
\item 
\begin{avm}
[\avmspan{phon \; \<\fval arrives\>} \\
 synsem|loc & [cat & [head  & \osort{verb}{
                              [vform & fin \\
                               aux   & $-$ ]} \\
                      aspect & lex\_state \\
                      spr    & \<\> \\
                      subj   & \<\feat np[-prd]$_{flight\_ent@1}$\>  \\
                      comps  & \<\feat
                            pp[-prd, pform {\fval at}]$_{gate\_ent@2}$
                               \> ]\\
               cont & \osort{pres}{
                      [main\_psoa & \osort{hab\_arrive\_at}{
                                    [arg1 & @1 \\
                                     arg2 & @2]}]}]]
\end{avm}
\label{pps:13}
\item 
\avmoptions{active}
\begin{avm}
[\avmspan{phon \; \<\fval BA737, arrives, at, gate2\>} \\
 synsem|loc & [cat & [head  & \osort{verb}{
                              [vform & fin \\
                               aux   & $-$ ]} \\
                      aspect & lex\_state \\
                      spr    & \<\> \\
                      subj   & \<\>  \\
                      comps  & \<\> ]\\
               cont & \osort{pres}{
                      [main\_psoa & \osort{hab\_arrive\_at}{
                                    [arg1 & ba737 \\
                                     arg2 & gate2]}]}]]
\end{avm}
\label{pps:16}
\end{examps}

The (predicative and non-predicative) prepositional signs
of this section are not used when prepositions introduce temporal
adverbials (e.g.\ \qit{BA737 departed at 5:00pm.}). There are
additional prepositional signs for these cases (see section
\ref{hpsg:pupe_adv} below).


\section{Nouns} \label{hpsg:nouns} 

\avmoptions{active}

Like prepositions, nouns receive different signs for their predicative
and non-predicative uses. Nouns used in noun-phrase complements of 
\qit{to be} (more precisely, the lexical heads of such
noun-phrase complements), like the \qit{president} of \pref{nps:3},
are \emph{predicative}. The corresponding noun phrases (e.g.\ 
\qit{the president} of \pref{nps:3}) are also said to be
predicative. In all other cases (e.g.\ \qit{the president} of
\pref{nps:1}), the nouns and noun phrases are \emph{non-predicative}.
\begin{examps}
\item J.Adams is the president. \label{nps:3}
\item The president was at gate 2. \label{nps:1}
\end{examps}

\subsection{Non-predicative nouns} 
\label{non_pred_nps}

Let us first examine non-predicative nouns. \pref{nps:2} shows the
sign of \qit{president} that would be used in \pref{nps:1}. The {\feat
  prd} value shows that the sign corresponds to a non-predicative use
of the noun. The {\feat spr} value means that the noun requires as its
specifier a determiner (e.g.\ \qit{a}, \qit{the}).
\begin{examps}
\item 
\setbox\avmboxa=\hbox{\begin{avm}
\sort{ntense}{
[et\_handle & \osort{temp\_ent}{
              [tvar $+$]} $\lor$ now \\
 main\_psoa & \osort{president}{
              [arg1 & @1]}]}
\end{avm}}
\avmoptions{active,center}
\begin{avm}
[\avmspan{phon \; \<\fval president\>} \\
 synsem|loc & [cat & [head  & \osort{noun}{
                              [prd & $-$]} \\
                      spr    & \<[loc|cat|head & {\fval det}] \> \\
                      subj   & \<\>  \\
                      comps  & \<\> ]\\ 
               cont & \osort{nom\_obj}{
                      [index & person\_ent@1 \\
                       restr &  \{\box\avmboxa\}]}] \\
 \avmspan{qstore \; \{\}} ]
\end{avm}
\label{nps:2}
\end{examps}
The {\feat cont} values of signs that correspond to non-predicative
nouns are of sort {\srt nom\_obj}\/ (section \ref{TOP_FS_WH}). The
{\feat index} value stands for the world entity described by the noun,
and the {\feat restr} value represents \topl expressions that are
introduced by the noun.

\pref{nps:4} shows the sign of \qit{the} that is used in \pref{nps:1}.
(In this thesis, \qit{the} is treated semantically as \qit{a}. This is
of course an over-simplification.)
\begin{examps}
\setbox\avmboxa=\hbox{\begin{avm}
\osort{det}{
[spec|loc & [cat & [head & \osort{noun}{
                           [prd & $-$]} \\
                    spr & \<\_\> \\
                    subj & \<\> \\
                    comps & \<\> ] \\
             cont & @2]]}
\end{avm}}
\avmoptions{active,center}
\item
\begin{avm}
[\avmspan{phon \; \<\fval the\>} \\
 synsem|loc & [cat & [head   & \box\avmboxa \\
                      spr    & \<\> \\
                      subj   & \<\>  \\
                      comps  & \<\> ]\\ 
               cont & \osort{quant}{
                      [det     & exists \\
                       restind & [index & [tvar $+$]]@2 
                      ]}@3 
               ] \\
 \avmspan{qstore \; \{@3\}} ]
\end{avm}
\label{nps:4}
\end{examps}
The {\feat spec} feature of \pref{nps:4} means that \qit{the} must be
used as the specifier of a non-predicative $\bar{N}$, i.e.\ as the
specifier of a non-predicative noun that has combined with its
complements and that requires a specifier.  The \avmbox{3}s of
\pref{nps:4} cause an existential quantifier to be inserted into the
quantifier store, and the \avmbox{2}s cause the {\feat restind} of
that quantifier to be unified with the {\feat cont} of the $\bar{N}$'s
sign.

According to \pref{nps:2}, \qit{president} is non-predicative, it does not
need to combine with any complements, and it requires a specifier.
Hence, it satisfies the {\feat spec} restrictions of \pref{nps:4}, and
\qit{the} can be used as the specifier of \qit{president}.  When
\qit{the} combines with \qit{president}, the {\feat restind} of
\pref{nps:4} is unified with the {\feat cont} of \pref{nps:2} (because of
the \avmbox{2}s in \pref{nps:4}), and the {\feat qstore} of
\pref{nps:4} becomes \pref{nps:7.1} (because of the \avmbox{3}s in
\pref{nps:4}). The resulting noun phrase receives \pref{nps:8}.
\begin{examps}
\item 
\setbox\avmboxa=\hbox{\begin{avm}
\sort{ntense}{
[et\_handle & \osort{temp\_ent}{
              [tvar $+$]} $\lor$ now \\
 main\_psoa & \osort{president}{
              [arg1 & @1]}]}
\end{avm}}
\avmoptions{active,center}
\begin{avm}
\{\sort{quant}{
  [det     & exists \\
   restind & [index & \sort{person\_ent}{
                      [tvar & $+$]}@1 \\
              restr & \{\box\avmboxa\}
             ]@2
  ]}@3\}
\end{avm}
\label{nps:7.1}
\item 
\setbox\avmboxa=\hbox{\begin{avm}
\sort{ntense}{
[et\_handle & \osort{temp\_ent}{
              [tvar $+$]} $\lor$ now \\
 main\_psoa & \osort{president}{
              [arg1 & @1]}]}
\end{avm}}
\begin{avm}
[\avmspan{phon \; \<\fval the, president\>} \\
 synsem|loc & [cat & [head  & \osort{noun}{
                              [prd & $-$]} \\
                      spr    & \<\> \\
                      subj   & \<\>  \\
                      comps  & \<\> ]\\ 
               cont & \osort{nom\_obj}{
                      [index & \sort{person\_ent}{
                               [tvar & $+$]}@1 \\
                       restr &  \{\box\avmboxa\}]}@2] \\
 \avmspan{qstore \; \{\sort{quant}{
                      [det & exists \\
                       restind & @2]}\}}]
\end{avm}
\label{nps:8}
\end{examps}
According to the head feature principle (section
\ref{schemata_principles}), \pref{nps:8} inherits the {\feat head} of
\pref{nps:2} (which is the sign of the ``head daughter'' in this
case). The propagation of {\feat cont} and {\feat qstore} 
is controlled by the semantics principle, which in this thesis has the
simplified form of \pref{nps:9}. (\pref{nps:9} uses the terminology of
\cite{Pollard2}. I explain below what \pref{nps:9} means for readers
not familiar with \cite{Pollard2}.)
\begin{examps}
\item
\principle{Semantics Principle (simplified version of this thesis):}\\
In a headed phrase, (a) the {\feat qstore} value is the union of the
{\feat qstore} values of the daughters, and (b) the {\feat
synsem$\mid$loc$\mid$cont} value is token-identical with that of
the semantic head. (In a headed phrase, the \emph{semantic head} is
the {\feat adjunct-daughter} if any, and the {\feat head-daughter}
otherwise.)
\label{nps:9}
\end{examps}
Part (a) means that the {\feat qstore} of each (non-lexical) syntactic
constituent is the union of the {\feat qstore}s of its
subconstituents. Part (b) means that each syntactic constituent
inherits the {\feat cont} of its head-daughter (the noun in
noun-phrases, the verb in verb phrases, the preposition in
prepositional phrases), except for cases where the head-daughter
combines with an adjunct-daughter (a modifier). In the latter case,
the mother syntactic constituent inherits the {\feat cont} of the
adjunct-daughter. (This will be discussed further in section
\ref{hpsg:pupe_adv}.) Readers familiar with \cite{Pollard2} will have
noticed that \pref{nps:9} does not allow quantifiers to be unstored
from {\feat qstore}. Apart from this, \pref{nps:9} is the same as in
\cite{Pollard2}.

\pref{nps:9} causes the {\feat qstore} of \pref{nps:8} to become the
union of the {\feat qstore}s of \pref{nps:2} (the empty set) and
\pref{nps:4} (which has become \pref{nps:7.1}). Since \qit{the
  president} involves no adjuncts, the ``semantic head'' is the
``head-daughter'' (i.e.\ \qit{president}), and \pref{nps:8} inherits
the {\feat cont} of \pref{nps:2} (which is now the {\feat restind} of
\pref{nps:7.1}).

The \qit{gate 2} of \pref{nps:1} is treated as a one-word proper name.
(In the prototype \nlitdb, the user has to type \qit{terminal 2} as a
single word; the same is true for \qit{J.Adams} of \pref{nps:3}. This
will be discussed in section \ref{preprocessor}.) Proper names are
mapped to signs whose {\feat cont} is a {\srt nom\_obj}\/ with
an empty-set {\feat restr}.\footnote{In \cite{Pollard2}, the signs of
  proper names involve {\srt naming}\/ relations, and {\feat context}
  and {\feat background} features. These are not used in this thesis.}
\qit{Gate 2}, for example, receives \pref{nps:10}. 
\begin{examps}
\item 
\avmoptions{active}
\begin{avm}
[\avmspan{phon \; \<\fval gate2\>} \\
 synsem|loc & [cat & [head  & \osort{noun}{
                              [prd & $-$]} \\
                      spr    & \<\> \\
                      subj   & \<\>  \\
                      comps  & \<\> ]\\ 
               cont & \osort{nom\_obj}{
                      [index & gate2 \\
                       restr &  \{\}]}] \\
 \avmspan{qstore \; \{\}} ]
\end{avm}
\label{nps:10}
\end{examps}
The predicative sign of \qit{at} of \pref{pps:5}, the predicative sign
of \qit{was} (which is similar to \pref{pps:7}, except that it
introduces a \past), and \pref{nps:10} cause the \qit{was at gate 2}
of \pref{nps:1} to receive \pref{nps:13}.
\begin{examps}
\item
\begin{avm}
[\avmspan{phon \; \<\fval was, at, gate2\>} \\
 synsem|loc & [cat & [head  & \osort{verb}{
                              [vform & fin \\
                               aux   & $+$ ]} \\
                      aspect & lex\_state \\
                      spr    & \<\> \\
                      subj   & \<\feat np[-prd]$_{@1}$\>  \\
                      comps  & \<\> ]\\ 
               cont & \osort{past}{
                      [et\_handle & \osort{temp\_ent}{
                                    [tvar & $+$]} \\
                       main\_psoa & \osort{located\_at}{
                                    [arg1 & @1 \\
                                     arg2 & gate2]}]}] \\
 \avmspan{qstore \; \{\}}]
\end{avm}
\label{nps:13}
\end{examps}
When \qit{was at gate 2} combines with \qit{the president},
\pref{nps:1} receives \pref{nps:14}. According to the semantics
principle, the {\feat qstore} of \pref{nps:14} is the union of the
{\feat qstore}s of \pref{nps:13} and \pref{nps:8}, and the {\feat
  cont} of \pref{nps:14} is the same as the {\feat cont} of
\pref{nps:13}.
\begin{examps}
\setbox\avmboxa=\hbox{\begin{avm}
\sort{ntense}{
[et\_handle & \osort{temp\_ent}{
              [tvar $+$]} $\lor$ now \\
 main\_psoa & \osort{president}{
              [arg1 & @1]}
]}
\end{avm}}
\avmoptions{active,center}
\item
\begin{avm}
[\avmspan{phon \; \<\fval the, president, was, at, gate2\>} \\
 synsem|loc & [cat & [head  & \osort{verb}{
                              [vform & fin \\
                               aux   & $+$]} \\
                      aspect & lex\_state \\
                      spr    & \<\> \\
                      subj   & \<\>  \\
                      comps  & \<\>] \\
               cont & \osort{past}{
                      [et\_handle & \osort{temp\_ent}{
                                    [tvar & $+$]} \\
                       main\_psoa & \osort{located\_at}{
                                    [arg1 & @1 \\
                                     arg2 & gate2]}]}] \\
 \avmspan{qstore \; \{[det & exists \\
                       restind & \osort{nom\_obj}{
                                  [index & \sort{person\_ent}{
                                           [tvar & $+$]}@1 \\
                                   restr & \{\box\avmboxa\}
                                  ]}
                      ]\}
         }
]
\end{avm}
\label{nps:14}
\end{examps}
\pref{nps:18} is then extracted from \pref{nps:14}, as discussed in
section \ref{extraction_hpsg}. Whenever an \ntense operator is
encountered during the extraction of the \topl formulae, if there is
no definite information showing that the first argument of the \ntense
should be $now^*$, the first argument is taken to be a variable.
\pref{nps:14}, for example, shows that the first argument of the
\ntense could be either a \topl variable or $now^*$. Hence, in
\pref{nps:18} the first argument of the \ntense has become a variable
($t^v$). During the post-processing phase (section
\ref{post_processing} below), the \ntense of \pref{nps:18} would give
rise to two separate formulae: one where the first argument of the
\ntense has been replaced by $now^*$ (current president), and one
where the first argument of the \ntense has been replaced by the $e^v$
of the \past operator (president when at gate 2).  In contrast, if the
sign shows that the first argument of the \ntense is definitely
$now^*$, the first argument of the \ntense in the extracted formula is
$now^*$, and the post-processing has no effect on this argument.
\begin{examps}
\item $\ntense[t^v, president(p^v)] \land \past[e^v, located\_at(p^v, gate2)]$ 
\label{nps:18}
\end{examps}

It is possible to force a (non-predicative) noun to be interpreted as
referring always to the speech time, or always to the time of the verb
tense. (This also applies to the non-predicative adjectives of section
\ref{hpsg:adjectives} below.) To force a noun to refer always to the
speech time, one sets the {\feat et\_handle} of the {\srt ntense}\/ in
the noun's sign to simply {\srt now}\/ (instead of allowing it to be
either {\srt now}\/ or a variable-representing index as in
\pref{nps:2}). This way, the {\feat et\_handle} of the {\srt ntense}\/
in \pref{nps:14} would be {\srt now}. \pref{nps:18} would contain
$now^*$ instead of $t^v$ (because in this case the sign shows that the
first argument of the \ntense should definitely be $now^*$), and the
post-processing mechanism would have no effect.

To force a noun to refer always to the time of the verb tense, one
simply omits the \ntense from the noun's sign. This would cause the
formula extracted from the sign of \pref{nps:1} to be
\pref{nps:25}.
\begin{examps}
\item $president(p^v) \land \past[e^v, located\_at(p^v, gate2)]$
\label{nps:25} 
\end{examps}
The semantics of \topl's conjunction (section \ref{denotation}) and of
the \past operator (section \ref{past_op}) require $president(p^v)$
and $located\_at(p^v, gate2)$ to be true at the same (past) event
time. Hence, \pref{nps:25} expresses the reading where the person at
gate 2 was the president of that time.

There are however, two complications when (non-predicative) noun signs
do not introduce \ntense{s}. (These also apply to adjective signs, to
be discussed in section \ref{hpsg:adjectives}.)  First, a past perfect
sentence like \pref{nps:26} receives only \pref{nps:27}, which
requires $president(p^v)$ to be true at the event time pointed to by
$e1^v$ (the ``reference time'', which is required to fall within
1/1/95). That is, \qit{the president} is taken to refer to somebody
who was the president on 1/1/95, and who may not have been the president
during the visit. 
\begin{examps}
\item The president had visited Rome on 1/1/95. \label{nps:26}
\item $president(p^v) \land
 \at[\text{\textit{1/1/95}}, \past[e1^v, \perf[e2^v, visiting(p^v, rome)]]]$
 \label{nps:27}
\end{examps}
In contrast, if the sign of \qit{president} introduces an
\ntense, the formula extracted from the sign of \pref{nps:26} is
\pref{nps:28}. The post-processing generates three different formulae
from \pref{nps:28}. These correspond to readings where \qit{president}
refers to the time of the visit ($t^v$ replaced by $e2^v$), the
reference time ($t^v$ replaced by $e1^v$, equivalent to
\pref{nps:27}), or the speech time ($t^v$ replaced by $now^*$). 
\begin{examps}
\item $\ntense[t^v, president(p^v)] \land$ \\
 $\at[\text{\textit{1/1/95}}, \past[e1^v, \perf[e2^v, visiting(p^v, rome)]]]$
 \label{nps:28}
\end{examps}

The second complication is that (non-predicative) nouns that do not
introduce \ntense{s} are taken to refer to the time of the \emph{main
  clause's} tense, even if the nouns appear in subordinate clauses
(subordinate clauses will be discussed in section
\ref{hpsg:subordinates}). For example, if \qit{president} does not
introduce an \ntense, \pref{nps:31.1} is mapped to \pref{nps:31.2}.
The semantics of \pref{nps:31.2} requires the visitor to have been
president during the building of terminal 2 (the visitor is not
required to have been president during the visit to terminal 3).
\begin{examps}
\item Housecorp built terminal 2 before the president visited terminal 3. 
   \label{nps:31.1}
\item $\begin{aligned}[t]
        president(p^v) \land  \before[&\past[e1^v, visiting(p^v, term3)],\\
                & \past[e2^v, \culm[building(housecorp, term2)]]]
       \end{aligned}$
   \label{nps:31.2}
\end{examps}
In contrast, if
\qit{president} introduces an \ntense, the post-processing (section
\ref{post_processing} below) generates 
three readings, where \qit{president}
refers to the speech time, the time of the building, or the time of
the visit. 

\medskip

The non-predicative signs of nouns like \qit{day} or \qit{summer},
that refer to members of partitionings (section \ref{top_model}) are
similar to the non-predicative signs of ``ordinary'' nouns like
\qit{president}, except that they introduce \partop operators, and
they do not introduce \ntense{s}. \pref{nps:32}, for example, shows
the non-predicative sign of \qit{day}. (The {\srt day}\/ and {\srt
  day\_ent\_var}\/ sorts are as in figure \vref{ind_hierarchy}.)
\begin{examps}
\item 
\setbox\avmboxa=\hbox{\begin{avm}
\sort{part}{
[partng    & day \\
 part\_var & @1]}
\end{avm}}
\avmoptions{active,center}
\begin{avm}
[\avmspan{phon \; \<\fval day\>} \\
 synsem|loc & [cat & [head  & \osort{noun}{
                              [prd & $-$]} \\
                      spr    & \<[loc|cat|head & {\fval det}] \> \\
                      subj   & \<\>  \\
                      comps  & \<\> ]\\ 
               cont & \osort{nom\_obj}{
                      [index & day\_ent\_var@1 \\
                       restr &  \{\box\avmboxa\}]}] \\
 \avmspan{qstore \; \{\}} ]
\end{avm}
\label{nps:32}
\end{examps}
Names of months and days (e.g.\ \qit{Monday}, \qit{January}) that can
be used both with and without determiners (e.g.\ \qit{on a Monday},
\qit{on Monday}) receive two non-predicative signs each: one that
requires a determiner, and one that does not.  Finally, proper names
that refer to particular time-periods (e.g.\ the year-name \qit{1991},
the date \qit{25/10/95}) receive non-predicative signs that are
similar to those of ``normal'' proper names (e.g.\ \qit{gate 2}),
except that their {\feat index} values are subsorts of {\srt
  temp\_ent}\/ rather than {\srt non\_temp\_ent}. I demonstrate in
following sections how the signs of temporal nouns and proper names
(e.g.\ \qit{day}, \qit{25/10/95}) are used to form the signs of
temporal adverbials (e.g.\ \qit{for two days}, \qit{before 25/10/95}).

\subsection{Predicative nouns} \label{pred_nps}

I now turn to predicative nouns, like the \qit{president} of
\pref{nps:3}. \pref{nps:41} shows the predicative sign of
\qit{president}. Unlike non-predicative noun-signs, whose {\feat cont}
values are of sort {\srt nom\_obj}, the {\feat cont} values of
predicative noun-signs are of sort {\srt psoa}. The {\srt president}\/
in \pref{nps:41} is a subsort of {\srt psoa}.
\begin{examps}
\item 
\avmoptions{active}
\begin{avm}
[\avmspan{phon \; \<\fval president\>} \\
 synsem|loc & [cat & [head  & \osort{noun}{
                              [prd & $+$]} \\
                      spr    & \<[loc|cat|head & {\fval det}]\> \\
                      subj   & \<\feat np[-prd]$_{@1}$\>  \\
                      comps  & \<\> ]\\ 
               cont & \sort{president}{
                      [arg1 & person\_ent@1]}] \\
 \avmspan{qstore \; \{\}} ]
\end{avm}
\label{nps:41}
\end{examps}
Unlike non-predicative nouns that do not require subjects (e.g.\ 
\pref{nps:2}), predicative nouns do require subjects. In
\pref{nps:41}, \qit{president} requires a non-predicative noun phrase
as its subject. The \avmbox{1} denotes the index of that noun phrase.

In the \hpsg version of this thesis, the predicative signs of nouns
are generated automatically from the non-predicative ones by
\pref{nps:42}.\footnote{Apart from the $remove\_ntense$, \pref{nps:42}
  is essentially the same as Borsley's ``predicative NP lexical
  rule'', discussed in the footnote of p.360 of \cite{Pollard2}.}
\begin{examps}
\item \lexrule{Predicative Nouns Lexical Rule:} 
\avmoptions{active}
\begin{center}
\begin{avm}
[synsem|loc & [cat & [head  & \osort{noun}{
                              [prd & $-$]} \\
                      subj   & \<\>]\\ 
               cont & \osort{nom\_obj}{
                      [index & @1 \\
                       restr & \{@2\}]}]]
\end{avm}
\\
$\Downarrow$
\\
\begin{avm}
[synsem|loc & [cat & [head & \osort{noun}{
                             [prd & $+$]} \\
                      subj & \<\feat np[-prd]$_{@1}$\>]\\ 
               cont & remove\_ntense\(@2\)]]
\end{avm}
\end{center}
\label{nps:42}
\end{examps}
The $remove\_ntense($\avmbox{2}$)$ in \pref{nps:42} means that if
\avmbox{2} (the single element of the {\feat restr} set of the
non-predicative sign) is of sort {\srt ntense}, then the {\feat cont}
of the predicative sign should be the {\feat main\_psoa} of \avmbox{2}
(see also \pref{nps:2}). Otherwise, the {\feat cont} of the
predicative sign should be \avmbox{2}.  In other words, if the
non-predicative sign introduces an \ntense, the \ntense is removed in
the predicative sign. This is related to the observation in section
\ref{noun_anaphora}, that noun phrases that are complements of \qit{to
  be} always refer to the time of the verb tense. For example,
\pref{nps:43} means that J.Adams was the president in 1992, not at the
speech time. \pref{nps:43} is represented correctly by \pref{nps:44}
which contains no \ntense{s}.
\begin{examps}
\item J.Adams was the president in 1992. \label{nps:43}
\item $\at[1992, \past[e^v, president(j\_adams)]]$ \label{nps:44}
\end{examps}
\topl predicates introduced by predicative nouns (e.g.\ 
$president(j\_adams)$ in \pref{nps:44}) end up within the operators of
the tenses of \qit{to be} (e.g.\ the \past of \pref{nps:44}). This
requires the predicates to hold at the times of the tenses.

As with previous lexical rules, features not shown in \pref{nps:42}
(e.g.\ {\feat spr}, {\feat comps}) have the same values in both the
original and the generated signs.  For example, \pref{nps:42}
generates \pref{nps:41} from \pref{nps:2}.

In this thesis, determiners also receive different signs for their
uses in predicative and non-predicative noun phrases. (Pollard and Sag
do not provide much information on determiners of predicative noun
phrases. The footnote of p.360 of \cite{Pollard2}, however, seems to
acknowledge that determiners of predicative noun phrases have to be
treated differently from determiners of non-predicative noun phrases.)
For example, apart from \pref{nps:4}, \qit{the} is also given
\pref{nps:45}. The {\feat spec} of \pref{nps:45} shows that
\pref{nps:45} can only be used with predicative nouns (cf.\ 
\pref{nps:4}).  Unlike determiners of non-predicative noun phrases,
determiners of predicative noun-phrases have no semantic contribution
(the {\feat synsem$\mid$loc$\mid$cont} of \pref{nps:45} is simply a
copy of the {\feat cont} of the noun, and no quantifier is introduced
in {\feat qstore}; cf.\ \pref{nps:4}).
\begin{examps}
\setbox\avmboxa=\hbox{\begin{avm}
\osort{det}{
[spec|loc & [cat & [head & \osort{noun}{
                           [prd & $+$]} \\
                    spr & \<\_\> \\
                    subj & \<\_\> \\
                    comps & \<\> ] \\
             cont & @2]]}
\end{avm}}
\avmoptions{active,center}
\item
\begin{avm}
[\avmspan{phon \; \<\fval the\>} \\
 synsem|loc & [cat & [head   & \box\avmboxa \\
                      spr    & \<\> \\
                      subj   & \<\>  \\
                      comps  & \<\> ]\\ 
               cont & @2 
               ] \\
 \avmspan{qstore \; \{\}} ]
\end{avm}
\label{nps:45}
\end{examps}

In \pref{nps:3}, when \qit{the} combines with \qit{president}, the
resulting noun phrase receives \pref{nps:46}. (\hpsg's principles,
including the semantics principle of \pref{nps:9}, cause \pref{nps:46}
to inherit the {\feat head}, {\feat subj}, and {\feat cont} of
\pref{nps:41}.)
\begin{examps}
\item 
\avmoptions{active}
\begin{avm}
[\avmspan{phon \; \<\fval the, president\>} \\
 synsem|loc & [cat & [head  & \osort{noun}{
                              [prd & $+$]} \\
                      spr    & \<\> \\
                      subj   & \<\feat np[-prd]$_{@1}$\>  \\
                      comps  & \<\> ]\\ 
               cont & \sort{president}{
                      [arg1 & person\_ent@1]}] \\
 \avmspan{qstore \; \{\}} ]
\end{avm}
\label{nps:46}
\end{examps}

Apart from \pref{vforms:50} and \pref{pps:7}, \qit{is} also receives
\pref{nps:47}, which allows the complement of \qit{is} to be a
predicative noun phrase. (There is also a sign of \qit{is} for
adjectival complements, as in \qit{Runway 2 is closed.}; this will be
discussed in section \ref{hpsg:adjectives}. \qit{Was} receives similar
signs.) The \avmbox{4}s in \pref{nps:47} denote the {\feat cont} of
the predicative noun-phrase.
\begin{examps}
\item
\begin{avm}
[\avmspan{phon \; \<\fval is\>} \\
 synsem|loc & [cat & [head  & \osort{verb}{
                              [vform & fin \\
                               aux   & $+$ ]} \\
                      aspect & lex\_state \\
                      spr    & \<\> \\
                      subj   & \<@3\>  \\
                      comps  & \<\feat np[subj \<@3\>, prd $+$]:@4 \> ]\\ 
               cont & \osort{pres}{
                      [main\_psoa & @4]}] \\
 \avmspan{qstore \; \{\}}]
\end{avm}
\label{nps:47}
\end{examps}

\pref{nps:47} and \pref{nps:46} cause the \qit{is the president} of
\pref{nps:3} to receive \pref{nps:48}. Finally, when \qit{is the
  president} combines with \qit{J.Adams}, \pref{nps:3} receives a sign
with an empty {\feat qstore}, whose {\feat cont} represents
\pref{nps:50}.
\begin{examps}
\item
\begin{avm}
[\avmspan{phon \; \<\fval is, the, president\>} \\
 synsem|loc & [cat & [head  & \osort{verb}{
                              [vform & fin \\
                               aux   & $+$ ]} \\
                      aspect & lex\_state \\
                      spr    & \<\> \\
                      subj   & \<\feat np[-prd]$_{@1}$\>  \\
                      comps  & \<\> ]\\ 
               cont & \osort{pres}{
                      [main\_psoa & \osort{president}{
                                    [arg1 & person\_ent@1]}]}] \\
 \avmspan{qstore \; \{\}}]
\end{avm}
\label{nps:48}
\item $\pres[president(j\_adams)]$ \label{nps:50}
\end{examps}

There are currently two complications with predicative noun phrases. The
first is that in the non-predicative signs of proper names like
\qit{gate2}, the value of {\feat restr} is the empty set (see
\pref{nps:10}). Hence, \pref{nps:42} does not generate the
corresponding predicative signs, because the non-predicative signs do
not match the {\feat restr} description of the LHS of \pref{nps:42}
(which requires the {\feat restr} value to be a one-element set). This
causes \pref{nps:51} to be rejected, because there is no predicative
sign for \qit{J.Adams}.
\begin{examps}
\item The inspector is J.Adams.  \label{nps:51}
\end{examps}
One way to solve this problem is to employ the additional rule of
\pref{nps:52}.
\begin{examps}
\item \lexrule{Additional Predicative Nouns Lexical Rule:} 
\avmoptions{active}
\begin{center}
\begin{avm}
[synsem|loc & [cat & [head  & \osort{noun}{
                              [prd & $-$]} \\
                      subj   & \<\>]\\ 
               cont & \osort{nom\_obj}{
                      [index & @1 \\
                       restr & \{\}]}]]
\end{avm}
\\
$\Downarrow$
\\
\begin{avm}
[synsem|loc & [cat & [head & \osort{noun}{
                             [prd & $+$]} \\
                      subj & \<\feat np[-prd]$_{@2}$\>]\\ 
               cont & \sort{identity}{
                      [arg1 & @1 \\
                       arg2 & @2]}]]
\end{avm}
\end{center}
\label{nps:52}
\end{examps}
This would generate \pref{nps:53} from the non-predicative sign of
\qit{J.Adams} (which is similar to \pref{nps:10}). \pref{nps:47} and
\pref{nps:53} would cause \pref{nps:51} to be mapped to \pref{nps:54}.
(I assume here that the non-predicative \qit{inspector} does not
introduce an \ntense.)
\begin{examps}
\item 
\avmoptions{active}
\begin{avm}
[\avmspan{phon \; \<\fval J.Adams\>} \\
 synsem|loc & [cat & [head  & \osort{noun}{
                              [prd & $+$]} \\
                      spr    & \<\> \\
                      subj   & \<\feat np[-prd]$_{@2}$\>  \\
                      comps  & \<\> ]\\ 
               cont & \sort{identity}{
                      [arg1 & j\_adams \\
                       arg2 & @2]}] \\
 \avmspan{qstore \; \{\}} ]
\end{avm}
\label{nps:53}
\item $inspector(insp^v) \land 
       \pres[identity(j\_adams, insp^v)]$ \label{nps:54}
\end{examps}
$identity(\tau_1,\tau_2)$ is intended to be true at event times
where its two arguments denote the same entity. This calls for a
special domain-independent semantics for $identity(\tau_1,\tau_2)$. I
have not explored this issue any further, however, and \pref{nps:52}
is not used in the prototype \nlitdb.

A second complication is that the non-predicative sign of \qit{Monday}
(which is similar to \pref{nps:32}) and the treatment of predicative
noun phrases above lead to an attempt to map \pref{nps:55} to
\pref{nps:56}.
\begin{examps}
\item 23/10/95 was a Monday. \label{nps:55}
\item $\past[e^v, \partop[monday^g, \text{\textit{23/10/95}}]]$ \label{nps:56}
\end{examps}
\pref{nps:56} is problematic for two reasons. (a) The past tense of
\pref{nps:55} is in effect ignored, because the denotation of
$\partop[\sigma, \beta]$ does not depend on $lt$, which is what the
\past operator affects (see sections \ref{denotation} and
\ref{past_op}). Hence, the implication of \pref{nps:55} that 23/10/95
is a past day is not captured. This problem could be solved by adding
the constraint $g(\beta) \subper lt$ in the semantics of
$\partop[\sigma, \beta]$ (section \ref{denotation}). (b)
\pref{nps:56} violates the syntax of \topl (section \ref{top_syntax}),
which does not allow the second argument of a \partop operator to be a
constant. This problem could be solved by modifying \topl to allow the
second argument of \partop to be a constant.


\section{Adjectives} \label{hpsg:adjectives}

Following Pollard and Sag (\cite{Pollard1}, pp.\ 64 -- 65), adjectives
also receive different signs for their predicative and non-predicative
uses. When used as complements of \qit{to be} (e.g.\ \qit{closed} in
\pref{adj:1}) adjectives are said to be predicative. In all other
cases (e.g.\ \qit{closed} in \pref{adj:2}), adjectives are
non-predicative. (\pref{adj:1} is actually ambiguous. The \qit{closed}
may be a predicative adjective, or the passive form of \qit{to close}.
As noted in section \ref{ling_not_supported}, however, passives are
ignored in this thesis. Hence, I ignore the passive reading of
\pref{adj:1}.)
\begin{examps}
\item Runway 2 was closed. \label{adj:1}
\item BA737 landed on a closed runway. \label{adj:2}
\end{examps}

In the airport domain, the predicative sign of the adjective
\qit{closed} is \pref{adj:3}.
\begin{examps}
\item 
\avmoptions{active}
\begin{avm}
[\avmspan{phon \; \<\fval closed\>} \\
 synsem|loc & [cat & [head  & \osort{adj}{
                              [prd & $+$]} \\
                      spr    & \<\> \\
                      subj   & \<\feat np[-prd]$_{@1}$\>  \\
                      comps  & \<\> ]\\ 
               cont & \sort{closed}{
                      [arg1 & \(gate\_ent $\lor$ runway\_ent\)@1]}] \\
 \avmspan{qstore \; \{\}} ]
\end{avm}
\label{adj:3}
\end{examps}
As noted in section \ref{hpsg:nouns}, \qit{is} and \qit{was}
receive four signs each. One for progressive forms (see
\pref{vforms:50}), one for prepositional phrase complements (see
\pref{pps:7}), one for noun-phrase complements (see \pref{nps:47}),
and one for adjectival complements (\pref{adj:4} below). \pref{adj:4}
and \pref{adj:3} cause \pref{adj:1} to be mapped to \pref{adj:6}.
\begin{examps}
\avmoptions{active,center}
\setbox\avmboxa=\hbox{\begin{avm}
[loc & [cat & [head & \osort{adj}{
                      [prd $+$]} \\
               subj & \<@3\> \\
               comps & \<\>] \\
        cont & @2]]
\end{avm}}
\item
\begin{avm}
[\avmspan{phon \; \<\fval was\>} \\
 synsem|loc & [cat & [head  & \osort{verb}{
                              [vform & fin \\
                               aux   & $+$ ]} \\
                      aspect & lex\_state \\
                      spr    & \<\> \\
                      subj   & \<@3\>  \\
                      comps  & \<\box\avmboxa\> ]\\ 
               cont & \osort{past}{
                      [et\_handle & \osort{temp\_ent}{
                                    [tvar & $+$]} \\
                       main\_psoa & @2]}] \\
 \avmspan{qstore \; \{\}}]
\end{avm}
\label{adj:4}
\item $\past[e^v, closed(runway2)]$ \label{adj:6}
\end{examps}

\pref{adj:7} shows the non-predicative sign of \qit{closed}. The
\qit{closed} in \pref{adj:2} is a modifier (adjunct) of \qit{runway}.
The {\feat mod} in \pref{adj:7} refers to the {\feat synsem} of the
sign of the noun that the adjective modifies. The {\feat
  synsem$\mid$loc$\mid$cont} of \pref{adj:7} is the same as the one of
the noun-sign, except that an \ntense is added to the {\feat restr} of
the noun-sign (i.e.\ to the set denoted by \avmbox{2}). The additional
\ntense requires the entity described by the noun (the entity
represented by \avmbox{1}) to be closed at some unspecified time. The
{\feat index} of the noun's sign is also required to represent a gate
or runway.
\begin{examps}
\item 
\avmoptions{active,center}
\setbox\avmboxa=\hbox{\begin{avm}
\sort{ntense}{
[et\_handle & \osort{temp\_ent}{
              [tvar $+$]} $\lor$ now \\
 main\_psoa & \osort{closed}{
              [arg1 & @1]}]}
\end{avm}}
\setbox\avmboxb=\hbox{\begin{avm}
[cat & [head  & noun \\
        spr   & \<\_\> \\
        comps & \<\>] \\
 cont & \osort{nom\_obj}{
        [index & @1 \\
         restr & @2]}]
\end{avm}}
\begin{avm}
[\avmspan{phon \; \<\fval closed\>} \\
 synsem|loc & [cat & [head  & \osort{adj}{
                              [\avmspan{\feat prd \; $-$} \\
                               mod|loc & \box\avmboxb]} \\
                      spr    & \<\>  \\
                      subj   & \<\>  \\
                      comps  & \<\> ]\\ 
               cont & \osort{nom\_obj}{
                      [index & \(gate\_ent $\lor$ runway\_ent\)@1 \\
                       restr & @2 $\union$ \{\box\avmboxa\}]}] \\
 \avmspan{qstore \; \{\}} ]
\end{avm}
\label{adj:7}
\end{examps}

In the airport domain, the non-predicative sign of \qit{runway} is
\pref{adj:8}. (I assume that \qit{runway} does not introduce an
\ntense.)
\begin{examps}
\item 
\avmoptions{active,center}
\setbox\avmboxa=\hbox{\begin{avm}
\sort{runway}{
[arg1 & @1]}
\end{avm}}
\begin{avm}
[\avmspan{phon \; \<\fval runway\>} \\
 synsem|loc & [cat & [head  & \osort{noun}{
                              [prd & $-$]} \\
                      spr    & \<[loc|cat|head & {\fval det}] \> \\
                      subj   & \<\>  \\
                      comps  & \<\> ]\\ 
               cont & \osort{nom\_obj}{
                      [index & runway\_ent@1 \\
                       restr &  \{\box\avmboxa\}]}] \\
 \avmspan{qstore \; \{\}} ]
\end{avm}
\label{adj:8}
\end{examps}

In \pref{adj:2}, \qit{closed} combines with \qit{runway} according to
\hpsg's head-adjunct schema (see \cite{Pollard2}). 
\qit{Closed runway} receives the sign of \pref{adj:9},
where \avmbox{3} is the set of \pref{adj:9.2}. (Sets of {\srt psoa}s
are treated as conjunctions.)
\begin{examps}
\avmoptions{active}
\item 
\begin{avm}
[\avmspan{phon \; \<\fval closed, runway\>} \\
 synsem|loc & [cat & [head  & \osort{noun}{
                              [prd & $-$]} \\
                      spr    & \<[loc|cat|head & {\fval det}] \> \\
                      subj   & \<\>  \\
                      comps  & \<\> ]\\ 
               cont & \osort{nom\_obj}{
                      [index & runway\_ent@1 \\
                       restr & @3]}] \\
 \avmspan{qstore \; \{\}} ]
\end{avm}
\label{adj:9}
\item 
\begin{avm}
\{
\sort{runway}{
[arg1 & @1]},
\osort{ntense}{
[et\_handle & \osort{temp\_ent}{
              [tvar $+$]} $\lor$ now \\
 main\_psoa & \osort{closed}{
              [arg1 @1]}]} 
\}@3
\end{avm}
\label{adj:9.2}
\end{examps}
The principles of \hpsg cause \pref{adj:9} to inherit the {\feat head}
and {\feat spr} of \pref{adj:8}. \pref{adj:9} inherits the {\feat
  cont} of \pref{adj:7} according to the semantics principle of
\pref{nps:9} (in this case, the ``semantic head'' is the adjunct
\qit{closed}).  \pref{adj:9}, the sign of \qit{landed} (which is the
same as \pref{lentr:1}, except that it also introduces \past and \culm
operators), and the non-predicative sign of \qit{on} (which is similar
to \pref{pps:12}), cause \pref{adj:2} to be mapped to the
\pref{adj:10}. During the post-processing (section
\ref{post_processing} below), \pref{adj:10} gives rise to two
different formulae, one where $t^v$ is replaced by $now^*$ (currently
closed runway), and one where $t^v$ is replaced by $e^v$ (closed
during the landing).
\begin{examps}
\item $runway(r^v) \land  \ntense[t^v, closed(r^v)] \; \land$\\
      $\past[e^v, \culm[landing\_on(occr^v, ba737, r^v)]]$ \label{adj:10}
\end{examps}

An additional sign is needed for each non-predicative adjective
to allow sentences like \pref{adj:13}, where a non-predicative
adjective (\qit{closed}) combines with a predicative noun (\qit{runway}).
\begin{examps}
\item Runway 2 is a closed runway. \label{adj:13}
\end{examps}
\pref{adj:7} cannot be used in \pref{adj:13}, because here
\qit{runway} is predicative, and hence the {\feat cont} of its sign is
a {\srt psoa}\/ (the predicative sign of \qit{runway} is similar to
\pref{nps:41}). In contrast, \pref{adj:7} assumes that the {\feat
  cont} of the noun is a {\srt nom\_obj}. One has to use the
additional sign of \pref{adj:14}.\footnote{It is unclear how
  \pref{adj:14} could be written in the \hpsg version of
  \cite{Pollard2}. In \cite{Pollard2}, the {\srt and}\/ sort does not
  exist, and conjunctions of {\srt psoa}s can only be expressed using
  sets of {\srt psoa}s, as in \pref{adj:9.2}. In \pref{adj:14},
  however, the value of {\feat sysnsem$\mid$loc$\mid$cont} cannot be a
  set of {\srt psoa}s, because {\feat cont} accepts only values whose
  sort is {\srt psoa}, {\srt nom\_obj}, or {\srt quant}.} Using
\pref{adj:14}, \pref{adj:13} is mapped to \pref{adj:15},
which requires runway 2 to be closed at the speech time. 
\begin{examps}
\item 
\avmoptions{active,center}
\setbox\avmboxa=\hbox{\begin{avm}
\sort{ntense}{
[et\_handle & \osort{temp\_ent}{
              [tvar $+$]} $\lor$ now \\
 main\_psoa & ]}
\end{avm}}
\setbox\avmboxb=\hbox{\begin{avm}
[cat & [head  & \osort{noun}{
                [prd \; $+$]} \\
        spr   & \<\_\> \\
        subj  & \<\feat np[-prd]$_{@1}$\> \\
        comps & \<\>] \\
 cont & @2]
\end{avm}}
\begin{avm}
[\avmspan{phon \; \<\fval closed\>} \\
 synsem|loc & [cat & [head  & \osort{adj}{
                              [\avmspan{\feat prd \; $-$} \\
                               mod|loc & \box\avmboxb]} \\
                      spr    & \<\>  \\
                      subj   & \<\>  \\
                      comps  & \<\> ]\\ 
               cont & \osort{and}{
                      [conjunct1 & @2 \\
                       conjunct2 & \sort{closed}{
                                   [arg1 & @1]}]}] \\
 \avmspan{qstore \; \{\}} ]
\end{avm}
\label{adj:14}
\item $\pres[runway(runway2) \land closed(runway2)]$ \label{adj:15}
\end{examps}

As discussed in section \ref{temporal_adjectives}, temporal adjectives
(e.g.\ \qit{former}, \qit{annual}) are not considered in this thesis.
The prototype \nlitdb allows only non-predicative uses of the temporal
adjective \qit{current} (as in \pref{adj:20}), by mapping
\qit{current} to a sign that sets the first argument of the noun's
\ntense to $now^*$. (This does not allow \qit{current} to be used with
nouns that do not introduce \ntense{s}; see section
\ref{non_pred_nps}.)
\begin{examps}
\item The current president was at terminal 2. \label{adj:20} 
\end{examps}


\section{Temporal adverbials} \label{hpsg:pupe_adv}

I now discuss temporal adverbials, starting from punctual adverbials
(section \ref{point_adverbials}). 

\subsection{Punctual adverbials} \label{hpsg:punc_adv}

Apart from \pref{pps:5} and \pref{pps:12} (which are used in sentences
like \qit{BA737 is at gate 2.} and \qit{BA737 (habitually) arrives at
  gate 2.}), \qit{at} also receives signs that are used when it
introduces punctual adverbials, as in \pref{pupe:1}. \pref{pupe:2}
shows one of these signs.
\begin{examps}
\item Tank 2 was empty at 5:00pm. \label{pupe:1}
\avmoptions{active, center}
\setbox\avmboxa=\hbox{\begin{avm}
\osort{prep}{
[\avmspan{prd \; $-$} \\
 \avmspan{mod \; \feat s[vform {\fval fin}]:@2 $\lor$ 
                 \feat vp[vform {\fval psp}]:@2} \\
 mod|loc|cat|aspect & {\fval state $\lor$ activity} \\
                    & {\fval $\lor$ point}]}
\end{avm}}
\item
\begin{avm}
[\avmspan{phon \; \<\fval at\>} \\
 synsem|loc & [cat & [head   & \box\avmboxa \\
                      spr    & \<\>  \\
                      subj   & \<\>  \\
                      comps  & \<\feat np[-prd]$_{minute\_ent@1}$\> \\
                      aspect & point]\\ 
               cont & \osort{at\_op}{
                      [time\_spec & @1 \\
                       main\_psoa & @2]}] \\
 \avmspan{qstore \; \{\}} ]
\end{avm}
\label{pupe:2}
\end{examps}
The {\feat mod} feature refers to the {\feat synsem} of the sign of
the constituent modified by \qit{at}. {\feat s[vform {\fval
    fin}]:\avmbox{2}} is an abbreviation for a finite sentence (a
finite verb form that has combined with its subject and complements).
The \avmbox{2} refers to the {\feat cont} of the sign of the finite
sentence. Similarly, {\feat vp[vform {\fval psp}]:\avmbox{2}} stands
for a past participle verb phrase (a past participle that has combined
with its complements but not its subject). The {\feat mod} of
\pref{pupe:2} means that \pref{pupe:2} can be used when \qit{at}
modifies finite sentences or past participle verb phrases, whose
aspect is state, activity, or point. Generally, in this thesis
temporal adverbials (punctual adverbials, period adverbials, duration
adverbials) and temporal subordinate clauses (to be discussed in
section \ref{hpsg:subordinates}) are allowed to modify only finite
sentences and past participle verb phrases.

\pref{pupe:2} and the sign of \qit{5:00pm} (shown in \pref{pupe:3})
cause \qit{at 5:00pm} to receive \pref{pupe:4} (\qit{5:00pm} acts as
the noun-phrase complement of \qit{at}).
\begin{examps}
\avmoptions{active,center}
\setbox\avmboxa=\hbox{\begin{avm}
\sort{part}{
[partng    & 5:00pm \\
 part\_var & @1]}
\end{avm}}
\item
\begin{avm}
[\avmspan{phon \; \<\fval 5:00pm\>} \\
 synsem|loc & [cat & [head  & \osort{noun}{
                              [prd & $-$]} \\
                      spr    & \<\> \\
                      subj   & \<\>  \\
                      comps  & \<\> ]\\ 
               cont & \osort{nom\_obj}{
                      [index & minute\_ent\_var@1 \\
                       restr &  \{\box\avmboxa\}]}@3 ] \\
 \avmspan{qstore \; \{[det     & exists \\
                       restind & @3]\}}]
\end{avm}
\label{pupe:3}
\item 
\setbox\avmboxa=\hbox{\begin{avm}
\osort{prep}{
[\avmspan{prd \; $-$} \\
 \avmspan{mod \; \feat s[vform {\fval fin}]:@2 $\lor$ 
       \feat vp[vform {\fval psp}]:@2} \\
 mod|loc|cat|aspect & {\fval state $\lor$ activity}\\
                    & {\fval $\lor$ point}]}
\end{avm}}
\setbox\avmboxb=\hbox{\begin{avm}
\{
\sort{part}{
[partng    & 5:00pm \\
 part\_var & @1]}
\}
\end{avm}}
\begin{avm}
[\avmspan{phon \; \<\fval at, 5:00pm\>} \\
 synsem|loc & [cat & [head   & \box\avmboxa \\
                      spr    & \<\>  \\
                      subj   & \<\>  \\
                      comps  & \<\> \\
                      aspect & point]\\ 
               cont & \osort{at\_op}{
                      [time\_spec & @1 \\
                       main\_psoa & @2]}] \\
 \avmspan{qstore \; \{[det     & exists \\
                       restind & [index & minute\_ent\_var@1 \\
                                  restr & \box\avmboxb]]\} } ]
\end{avm}
\label{pupe:4}
\end{examps}
According to \hpsg's head feature principle (section
\ref{schemata_principles}), \pref{pupe:4} inherits the {\feat head} of
\pref{pupe:2} (\qit{at} is the ``head-daughter'' of \qit{at 5:00pm},
and \qit{5:00pm} is the ``complement-daughter''). Following the
semantics principle of \pref{nps:9}, the {\feat qstore} of
\pref{pupe:4} is the union of the {\feat qstore}s of \pref{pupe:2} and
\pref{pupe:3}, and the {\feat cont} of \pref{pupe:4} is the same as
the {\feat cont} of \pref{pupe:2} (in this case, the ``semantic head''
is the head-daughter, i.e.\ \qit{at}).

The propagation of {\feat aspect} is controlled by \pref{pupe:5}, a
new principle of this thesis. (As with \pref{nps:9}, in \pref{pupe:5} I use the
terminology of \cite{Pollard2}.) 
\begin{examps}
\item \principle{Aspect Principle:}
\index{aspect@{\feat aspect} (\hpsg feature)} \\
In a headed-phrase, the {\feat synsem$\mid$loc$\mid$cat$\mid$aspect} value is
token-identical with that of the semantic head. (In a headed phrase,
the \emph{semantic head} is the {\feat adjunct-daughter} if any, and
the {\feat head-daughter} otherwise.)
\label{pupe:5}
\end{examps}
\pref{pupe:5} means that each syntactic constituent inherits the
{\feat aspect} of its head-daughter (the noun in noun phrases, the
verb in verb phrases, the preposition in prepositional phrases),
except for cases where the head-daughter combines with an
adjunct-daughter (a modifier). In the latter case, the mother
syntactic constituent inherits the {\feat cont} of the
adjunct-daughter. \pref{pupe:5} causes \pref{pupe:4} to inherit the
{\feat aspect} value of the semantic head \qit{at}.

The \qit{tank 2 was empty} of \pref{pupe:1} receives \pref{pupe:6}.
\begin{examps}
\avmoptions{active}
\item
\begin{avm}
[\avmspan{phon \; \<\fval tank2, was, empty\>} \\
 synsem|loc & [cat & [head  & \osort{verb}{
                              [vform & fin \\
                               aux   & $+$]} \\
                      aspect & lex\_state \\
                      spr    & \<\> \\
                      subj   & \<\>  \\
                      comps  & \<\>] \\
               cont & \osort{past}{
                      [et\_handle & \osort{temp\_ent}{
                                    [tvar & $+$]} \\
                        main\_psoa & \osort{empty}{
                                     [arg1 & tank2]}]}] \\
 \avmspan{qstore \; \{\}}]
\end{avm}
\label{pupe:6}
\setbox\avmboxa=\hbox{\begin{avm}
\osort{past}{
[et\_handle & \osort{temp\_ent}{
              [tvar & $+$]} \\
 main\_psoa & \osort{empty}{
              [arg1 & tank2]}]}
\end{avm}}
\setbox\avmboxb=\hbox{\begin{avm}
\{
\sort{part}{
[partng    & 5:00pm \\
 part\_var & @1]}
\}
\end{avm}}
\end{examps}
When \qit{tank 2 was empty} combines with \qit{at 5:00pm},
\pref{pupe:1} receives \pref{pupe:7}. In this case, \qit{tank 2 was
  empty} is the head-daughter, and \qit{at 5:00pm} is an
adjunct-daughter (a modifier). Hence, according to \pref{pupe:5},
\pref{pupe:7} inherits the {\feat aspect} of \pref{pupe:4} (i.e.\ 
{\srt point}\/; in contrast, the {\feat aspect} of \pref{pupe:6} was
{\srt lex\_state}.)  This is in accordance with the arrangements of
section \ref{point_adverbials}, whereby punctual adverbials trigger an
aspectual shift to point.
\begin{examps}
\item
\setbox\avmboxa=\hbox{\begin{avm}
\osort{past}{
[et\_handle & \osort{temp\_ent}{
              [tvar & $+$]} \\
 main\_psoa & \osort{empty}{
              [arg1 & tank2]}]}
\end{avm}}
\setbox\avmboxb=\hbox{\begin{avm}
\{
\sort{part}{
[partng    & 5:00pm \\
 part\_var & @1]}
\}
\end{avm}}
\begin{avm}
[\avmspan{phon \; \<\fval tank2, was, empty, at, 5:00pm\>} \\
 synsem|loc & [cat & [head  & \osort{verb}{
                              [vform & fin \\
                               aux   & $+$]} \\
                      aspect & point \\
                      spr    & \<\> \\
                      subj   & \<\>  \\
                      comps  & \<\>] \\
               cont & \osort{at\_op}{
                      [time\_spec & @1 \\
                       main\_psoa & \box\avmboxa]}] \\
 \avmspan{qstore \; \{[det     & exists \\
                       restind & [index & minute\_ent\_var@1 \\
                                  restr & \box\avmboxb]]\}}]
\end{avm}
\label{pupe:7}
\end{examps}
According to the semantics principle, \pref{pupe:7} also inherits the
{\feat cont} of \pref{pupe:4} (the sign of the modifier), and the
{\feat qstore} of \pref{pupe:7} is the union of the {\feat qstore}s of
\pref{pupe:4} and \pref{pupe:6}. Finally, according to the head
feature principle (section \ref{schemata_principles}), \pref{pupe:7}
inherits the {\feat head} of \pref{pupe:6} (the sign of the
head-daughter). The {\feat qstore} and {\feat cont} of \pref{pupe:7}
represent \pref{pupe:7.1}.
\begin{examps}
\item $\partop[\text{\textit{5:00pm}}^g, fv^v] \land 
       \at[fv^v, \past[e^v, empty(tank2)]]$ \label{pupe:7.1}
\end{examps}

The reader may wonder why temporal adverbials (e.g.\ \qit{at 5:00pm}
in \pref{pupe:1}) are taken to modify whole finite sentences
(\qit{tank 2 was empty}), rather than finite verb phrases (\qit{was
  empty}). The latter approach leads to problems in questions like
\qit{Was tank 2 empty at 5:00pm?}, where \qit{was} combines in one
step with both its subject \qit{tank 2} and its complement
\qit{empty}, following the head-subject-complement schema of
\cite{Pollard2}. In this case, there is no verb phrase constituent
(verb that has combined with its complements but not its subject) to
be modified by \qit{at 5:00pm}.

Apart from finite sentences, temporal adverbials are also allowed to
modify past participle verb phrases (see the {\feat mod} of
\pref{pupe:2}). This is needed in past perfect sentences like
\pref{pupe:9}.
\begin{examps}
\item BA737 had entered sector 2 at 5:00pm. \label{pupe:9}
\end{examps}
As discussed in section \ref{past_perfect}, \pref{pupe:9} has two
readings: one where the entrance occurs at 5:00pm, and one where
5:00pm is a ``reference time'', a time where the entrance has already
occurred. The two readings are expressed by \pref{pupe:10} and
\pref{pupe:11} respectively (see also section \ref{perf_op}).
\pref{pupe:9} is taken to be syntactically ambiguous with two possible
parses, sketched in \pref{pupe:12} and \pref{pupe:13}. These give rise
to \pref{pupe:10} and \pref{pupe:11} respectively.
\begin{examps}
\item BA737 had [[entered sector 2] at 5:00pm]. \label{pupe:12}
\item $\partop[\text{\textit{5:00pm}}^g, fv^v] \land 
       \past[e1^v, \perf[e2^v, \at[fv^v, enter(ba737, sector2)]]]$
   \label{pupe:10}
\item {[}BA737 had [entered sector 2]] at 5:00pm. \label{pupe:13}
\item $\partop[\text{\textit{5:00pm}}^g, fv^v] \land 
       \at[fv^v, \past[e1^v, \perf[e2^v, enter(ba737, sector2)]]]$
   \label{pupe:11}
\end{examps}

One complication of this approach is that it generates two equivalent
formulae for the present perfect \pref{pupe:17}, shown in
\pref{pupe:18} and \pref{pupe:19}. (\qit{Has} does not
introduce a \perf; see \pref{vforms:36}.) These correspond to the
parses of \pref{pupe:17a} and \pref{pupe:17b} respectively.
\begin{examps}
\item BA737 has entered sector 2 at 5:00pm. \label{pupe:17}
\item {[}BA737 has [entered sector 2]] at 5:00pm. \label{pupe:17a}
\item $\partop[\text{\textit{5:00pm}}^g, fv^v] \land
       \at[fv^v, \past[e^v, enter(ba737, sector2)]]$ \label{pupe:18}
\item BA737 has [[entered sector 2] at 5:00pm]. \label{pupe:17b}
\item $\partop[\text{\textit{5:00pm}}^g, fv^v] \land
       \past[e^v, \at[fv^v, enter(ba737, sector2)]]$ \label{pupe:19}
\end{examps}
In the prototype \nlitdb, the sign of \qit{has} is slightly more
complex than \pref{vforms:36}. It requires the {\feat cont} 
of the verb-phrase complement of \qit{has} to be of sort {\srt
  predicate}. This blocks \pref{pupe:17b} and \pref{pupe:19},
because in \pref{pupe:17b} the \qit{at 5:00pm} causes the {\feat cont}
of \qit{entered sector 2 at 5:00pm} to become of sort {\srt
  at\_op}\/ (it inserts an \at operator), which is not a subsort
of {\srt predicate}. 

\pref{pupe:2} corresponds to the interjacent meaning of punctual
adverbials, which according to table \vref{punctual_adverbials_table}
is possible only with states and activities. \pref{pupe:2} also covers
cases where punctual adverbials combine with points. There are also
other \qit{at} signs, that are similar to \pref{pupe:2} but that
introduce additional \lbegin or \lend operators. These correspond to
the inchoative (with activities and culminating activities) and
terminal (with culminating activities) meanings of punctual
adverbials.

\subsection{Period adverbials} \label{hpsg:per_advs}

I now turn to period adverbials (section \ref{period_adverbials}).
\pref{pupe:29} shows one of the signs of \qit{on} that are used when
\qit{on} introduces period adverbials.
\begin{examps}
\avmoptions{active, center}
\item 
\setbox\avmboxa=\hbox{\begin{avm}
\osort{prep}{
[prd & $-$ \\
 mod & \feat s[vform {\fval fin}]:@2 $\lor$ 
       \feat vp[vform {\fval psp}]:@2 \\
 \avmspan{mod|loc|cat|aspect \; {\fval culmact}}]}
\end{avm}}
\begin{avm}
[\avmspan{phon \; \<\fval on\>} \\
 synsem|loc & [cat & [head   & \box\avmboxa \\
                      spr    & \<\>  \\
                      subj   & \<\>  \\
                      comps  & \<\feat np[-prd]$_{day\_ent@1}$\> \\
                      aspect & point]\\ 
               cont & \osort{at\_op}{
                      [time\_spec & @1 \\
                       main\_psoa & \osort{end}{
                                    [main\_psoa & @2]}]}] \\
 \avmspan{qstore \; \{\}} ]
\end{avm}
\label{pupe:29}
\end{examps}
\pref{pupe:29}, which can be used only when the \qit{on~\dots}
adverbial modifies a culminating activity, corresponds to the reading
where the situation of the culminating activity simply reaches its
completion within the adverbial's period (table
\vref{period_adverbials_table}). (\pref{pupe:29} causes the aspectual
class of the culminating activity to become point. This agrees with
table \ref{period_adverbials_table}.) For example, \pref{pupe:29}
causes \pref{pupe:31} to be mapped to \pref{pupe:32}. (I assume here
that \qit{to repair} is classified as culminating activity verb.)
Intuitively, \pref{pupe:32} requires a past period $e^v$ to exist,
such that $e^v$ covers a whole repair of engine 2 by J.Adams (from
start to completion), and the end-point of $e^v$ falls within some
Monday. That is, the repair must have been completed on Monday, but it
may have started before Monday.
\begin{examps}
\item J.Adams repaired engine 2 on Monday. \label{pupe:31}
\item $\partop[monday^g, m^v] \; \land$\\
      $\at[m^v, \lend[\past[e^v, [\culm[repairing(occr^v, j\_adams, eng2)]]]]$
   \label{pupe:32}
\end{examps}
There is also an \qit{on} sign that is similar to \pref{pupe:29}, but
that does not introduce an \lend operator, preserves the {\feat
  aspect} of the modified expression, and can be used when
\qit{on~\dots} adverbials modify expressions from all four aspectual
classes. This sign causes \pref{pupe:31} to be mapped to
\pref{pupe:33} (the prototype \nlitdb would generate both
\pref{pupe:32} and \pref{pupe:33}). \pref{pupe:33} corresponds to the
reading where the repair must have both started and been completed
within a (the same) Monday. The \qit{on} sign that does not introduce
an \lend also gives rise to appropriate formulae when
\qit{on~\dots} adverbials modify state, activity, or point
expressions.
\begin{examps}
\item $\partop[monday^g, m^v] \; \land$\\ 
      $\at[m^v, \past[e^v, [\culm[repairing(occr^v, j\_adams, eng2)]]]$
   \label{pupe:33}
\end{examps}

Both \pref{pupe:29} and the \qit{on} sign that does not introduce an
\lend require the noun-phrase complement of \qit{on} to introduce an
index of sort {\srt day\_ent}. The signs of \qit{1/1/91} and
\qit{Monday} introduce indices of sorts {\srt 1/1/91}\/ and {\srt
  day\_ent\_var}\/ respectively, which are subsorts of {\srt
  day\_ent}\/ (see figure \vref{ind_hierarchy}). Hence, \qit{1/1/91}
and \qit{Monday} are legitimate complements of \qit{on} in period
adverbials. In contrast, \qit{5:00pm} introduces an index of sort
{\srt minute\_ent\_var}\/ (see \pref{pupe:3}), which is not a subsort
of {\srt day\_ent}. Hence, \pref{pupe:40} is correctly rejected.
\begin{examps}
\item \bad Tank 2 was empty on 5:00pm. \label{pupe:40}
\end{examps}

The signs of other prepositions that introduce period adverbials
(e.g.\ \qit{\underline{in} 1991}, \qit{\underline{before} 29/10/95},
\qit{\underline{after} 5:00pm}) and the signs of \qit{yesterday} and
\qit{today} are similar to the signs of \qit{on}, except that
\qit{before} and \qit{after} introduce \before and \after operators
instead of \at{s}. Also, \qit{before} is given only one sign, that does
not introduce an \lend (there is no \qit{before} sign for
culminating activities analogous to \pref{pupe:29}, that introduces an
\lend). This is related to comments in section \ref{period_adverbials}
that in the case of \qit{before~\dots} adverbials, requiring the
situation of a culminating activity to simply reach its completion
before some time (reading with \lend) is equivalent to requiring the
situation to both start and reach its completion before that time
(reading without \lend).

\subsection{Duration adverbials}  \label{duration_adverbials}

The treatment of \qit{for~\dots} duration adverbials is rather ad hoc
from a syntax point of view. In an adverbial like \qit{for two days},
both \qit{two} and \qit{days} are taken to be complements of
\qit{for}, instead of treating \qit{two} as the determiner of
\qit{days}, and \qit{two days} as a noun-phrase complement of
\qit{for}. 

Number-words like \qit{one}, \qit{two}, \qit{three}, etc.\ are mapped
to signs of the form of \pref{dadv:4}. Their {\feat restr}s are empty,
and their indices represent the corresponding numbers. (The {\srt 2}\/
of \pref{dadv:4} is a subsort of {\srt sem\_num}\/; see section
\ref{more_ind}.)
\begin{examps}
\item 
\avmoptions{active}
\begin{avm}
[\avmspan{phon \; \<\fval two\>} \\
 synsem|loc & [cat & [head  & \sort{det}{
                              [spec & none]}\\
                      spr    & \<\> \\
                      subj   & \<\>  \\
                      comps  & \<\> ]\\ 
               cont & \osort{nom\_obj}{
                      [index & 2 \\
                       restr & \{\}]}] \\
 \avmspan{qstore \; \{\}} ]
\end{avm}
\label{dadv:4}
\end{examps}
Although words like \qit{one}, \qit{two}, \qit{three}, etc.\ are
classified as determiners (the {\feat head} of \pref{dadv:4} is
of sort {\srt det}), the {\srt none}\/ value of their {\feat spec}
does not allow them to be used as determiners of 
any noun. (Determiners combining with nouns are the specifiers
of the nouns. The {\srt none}\/ means that the word of the sign cannot
be the specifier of any constituent, and hence cannot be used as
the determiner of any noun.)

\pref{dadv:1} shows the sign of \qit{for} that is used in duration
adverbials (for typesetting reasons, I show the feature structures
that correspond to \avmbox{5} and \avmbox{6} separately, in
\pref{dadv:2} and \pref{dadv:3} respectively).
\begin{examps}
\avmoptions{active, center}
\item
\setbox\avmboxa=\hbox{\begin{avm}
\osort{prep}{
[\avmspan{prd \; $-$} \\
 \avmspan{mod \; \feat s[vform {\fval fin}]:@4 $\lor$ 
                 \feat vp[vform {\fval psp}]:@4} \\
 mod|loc|cat|aspect & \({\fval lex\_state} \\
                    & $\lor$ {\fval progressive} \\
                    & $\lor$ {\fval activity}\)@1]}
\end{avm}}
\begin{avm}
[\avmspan{phon \; \<\fval for\>} \\
 synsem|loc & [cat & [head   & \box\avmboxa \\
                      spr    & \<\>  \\
                      subj   & \<\>  \\
                      comps  & \<@5, @6\> \\
                      aspect & @1]\\ 
               cont & \osort{for\_op}{
                      [dur\_unit  & @2 \\
                       duration   & @3 \\
                       main\_psoa & @4]}] \\
 \avmspan{qstore \; \{\}} ]
\end{avm}
\label{dadv:1}
\item
\begin{avm}
[loc & [cat|head & det \\
        cont|index & sem\_num@3]]@5
\end{avm}
\label{dadv:2}
\item
\begin{avm}
[loc & [cat & [head & noun \\
               spr  & \<\_\> \\
               subj & \<\> \\
               comps & \<\>] \\
        cont|restr & \{\sort{part}{
                       [partng & compl\_partng@2]}\}]]@6
\end{avm}
\label{dadv:3}
\end{examps}
The {\feat comps} of \pref{dadv:1} means that \qit{for} requires two
complements: a determiner that introduces a number-denoting ({\srt
  sem\_num}\/) index (like the \qit{two} of \pref{dadv:4}), and a noun
that introduces a \partop operator whose first argument is a complete
partitioning name (like the \qit{day} of \pref{nps:32}).  In
\pref{dadv:6}, \qit{for two days} receives \pref{dadv:7}. (As already
mentioned, no number-agreement checks are made, and plural nouns are
treated semantically as singular ones. Apart from {\feat phon},
the sign of \qit{days} is the same as \pref{nps:32}.)
\begin{examps}
\item Tank 2 was empty for two days. \label{dadv:6}
\item
\avmoptions{active, center}
\setbox\avmboxa=\hbox{\begin{avm}
\osort{prep}{
[\avmspan{prd \; $-$} \\
 \avmspan{mod \; \feat s[vform {\fval fin}]:@4 $\lor$ 
                 \feat vp[vform {\fval psp}]:@4} \\
 mod|loc|cat|aspect & \({\fval lex\_state}\\
                    & $\lor$ {\fval progressive} \\
                    & $\lor$ {\fval activity}\)@1]}
\end{avm}}
\begin{avm}
[\avmspan{phon \; \<\fval for, two, days\>} \\
 synsem|loc & [cat & [head   & \box\avmboxa \\
                      spr    & \<\>  \\
                      subj   & \<\>  \\
                      comps  & \<\> \\
                      aspect & @1]\\ 
               cont & \osort{for\_op}{
                      [dur\_unit  & day \\
                       duration   & 2 \\
                       main\_psoa & @4]}] \\
 \avmspan{qstore \; \{\}} ]
\end{avm}
\label{dadv:7}
\end{examps}
When \qit{tank 2 was empty} combines with its temporal-adverbial
modifier \qit{for two days}, the \avmbox{4} of \pref{dadv:7} becomes a
feature structure that represents the \topl formula for \qit{tank 2
  was empty}, i.e.\ \pref{dadv:8}. According to the semantics
principle of \pref{nps:9}, the sign of \pref{dadv:6} inherits the
{\feat cont} of \pref{dadv:7} (where \avmbox{4} now represents
\pref{dadv:8}). Hence, \pref{dadv:6} is mapped to \pref{dadv:9}.
\begin{examps}
\item $\past[e^v, empty(tank2)]$ \label{dadv:8}
\item $\for[day^c, 2, \past[e^v, empty(tank2)]]$ \label{dadv:9}
\end{examps}
Following table \vref{for_adverbials_table}, \pref{dadv:1} does not
allow \qit{for~\dots} adverbials to modify point expressions (the
{\feat mod$\mid$loc$\mid$cat$\mid$aspect} of \pref{dadv:1} cannot be
{\srt point}\/). It also does not allow \qit{for~\dots} adverbials to
modify consequent states. If \qit{for~\dots} adverbials were allowed
to modify consequent states, \pref{dadv:10} would receive
\pref{dadv:11} and \pref{dadv:12}.
\begin{examps}
\item BA737 had circled for two hours. \label{dadv:10}
\item $\past[e1^v, \perf[e2^v, \for[hour^c, 2, circling(ba737)]]]$ 
   \label{dadv:11}
\item $\for[hour^c, 2, \past[e1^v, \perf[e2^v, circling(ba737)]]]$
   \label{dadv:12}
\end{examps}
\pref{dadv:11} corresponds to the parse of \pref{dadv:10} where
\qit{for two hours} modifies the past participle \qit{circled} before
\qit{circled} combines with \qit{had}. In that case, the
\qit{for~\dots} adverbial modifies an activity, because past
participles retain the aspectual class of the base form (\qit{to
  circle} is an activity verb in the airport domain). \pref{dadv:12}
corresponds to the parse where \qit{for two hours} modifies the whole
sentence \qit{BA737 had circled}. In that case, the \qit{for~\dots}
adverbial modifies a consequent state, because the \qit{had} has
caused the aspectual class of \qit{BA737 had circled} to become
consequent state. By not allowing \qit{for~\dots} adverbials to modify
consequent states, \pref{dadv:12} is blocked. This is needed, because
in \pref{dadv:12} two hours is the duration of a period (pointed to by
$e1^v$) that follows a period (pointed to by $e2^v$) where BA737 was
circling. This reading is never possible when \qit{for~\dots}
adverbials are used in past perfect sentences. The \qit{for~\dots}
adverbial of \pref{dadv:10} can only specify the duration of the
circling (a reading captured by \pref{dadv:11}). (A similar
observation is made on p.~587 of \cite{Kamp1993}.)

The present treatment of \qit{for~\dots} duration adverbials causes
\pref{dadv:13} to receive \pref{dadv:14}. \pref{dadv:14} does not
capture correctly the meaning of \pref{dadv:13}, because it requires
the taxiing to have been completed, i.e.\ BA737 to have reached gate
2. In contrast, as discussed in section \ref{for_adverbials}, the
\qit{for~\dots} adverbial of \pref{dadv:13} cancels the normal
implication of \qit{BA737 taxied to gate 2.} that the taxiing was
completed. The post-processing (section \ref{post_processing} below)
removes the \culm of \pref{dadv:14}, generating a formula that does
not require the taxiing to have been completed.
\begin{examps}
\item BA737 taxied to gate 2 for five minutes. \label{dadv:13}
\item $\for[minute^c, 5, \past[e^v, \culm[taxiing\_to(ba737, gate2)]]]$
   \label{dadv:14}
\end{examps}

Duration adverbials introduced by \qit{in} (e.g.\ \pref{dadv:15}) are
treated by mapping \qit{in} to a sign that is the same as
\pref{dadv:1}, except that it allows the adverbial to modify
only culminating activities. (The framework of this
thesis does not allow \qit{in~\dots} duration adverbials to
modify states, activities, or points; see section \ref{in_adverbials}.)
\begin{examps}
\item BA737 taxied to gate 2 in five minutes. \label{dadv:15}
\end{examps}
This causes \pref{dadv:15} to be mapped to \pref{dadv:14}, which
correctly requires the taxiing to have been completed, and the
duration of the taxiing (from start to completion) to be 
five minutes. (In this case, the post-processing does not remove the \culm.)


\section{Temporal complements of habituals} \label{habituals}

Let us now examine more closely the status of temporal prepositional-phrases,
like \qit{at 5:00pm} and \qit{on Monday} in \pref{hab:1} -- \pref{hab:4}.
\begin{examps}
\item BA737 departed at 5:00pm. \label{hab:1}
\item BA737 departs at 5:00pm. \label{hab:2}
\item J.Adams inspected gate 2 on Monday. \label{hab:3}
\item J.Adams inspects gate 2 on Monday. \label{hab:4}
\end{examps}
\pref{hab:1} has both a habitual and a non-habitual reading. Under the
non-habitual reading, it refers to an actual departure that took place
at 5:00pm. Under the habitual reading, it means that BA737 had the
habit of departing at 5:00pm (this reading is easier to accept if an
adverbial like \qit{in 1992} is added). In \pref{hab:2}, only the
habitual reading is possible, i.e.\ BA737 currently has the habit of
departing at 5:00pm. (A scheduled-to-happen reading is also possible,
but as discussed in section \ref{simple_present} this is ignored in
this thesis.) Similar comments apply to \pref{hab:3} and \pref{hab:4}.

To account for the habitual and non-habitual readings of \qit{to
  depart} in \pref{hab:1} and \pref{hab:2}, the base form of \qit{to
  depart} is given the signs of \pref{hab:7} and \pref{hab:8}. These
correspond to what chapter \ref{linguistic_data} called informally the
habitual and non-habitual homonyms of \qit{to depart}. \pref{hab:7}
classifies the habitual homonym as (lexical) state, while \pref{hab:8}
classifies the non-habitual homonym as point (this agrees with table
\vref{airport_verbs}). According to \pref{hab:7}, the habitual homonym
requires an \qit{at~\dots} prepositional phrase that specifies the
habitual departure time (this is discussed further below). In
contrast, the non-habitual homonym of \pref{hab:8} requires no
complement.
\avmoptions{active}
\begin{examps}
\item 
\begin{avm}
[\avmspan{phon \; \<\fval depart\>} \\
 synsem & [loc & [cat & [head  & \osort{verb}{
                                 [vform & bse \\
                                  aux   & $-$ ]} \\
                         aspect & lex\_state \\
                         spr    & \<\> \\
                         subj   & \< \feat np[-prd]$_{flight\_ent@1}$ \>  \\
                         comps  & \< 
                       \feat pp[-prd, pform {\fval at}]$_{minute\_gappy@2}$
                                  \> ]\\
                  cont & \sort{hab\_departs\_at}{
                         [arg1 & @1 \\
                          arg2 & @2]} ]] \\
 \avmspan{qstore \; \{\}}]
\end{avm}
\label{hab:7}
\item 
\begin{avm}
[\avmspan{phon \; \<\fval depart\>} \\
 synsem & [loc & [cat & [head  & \osort{verb}{
                                 [vform & bse \\
                                  aux   & $-$ ]} \\
                         aspect & point \\
                         spr    & \<\> \\
                         subj   & \< \feat np[-prd]$_{flight\_ent@1}$ \>  \\
                         comps  & \<\> ]\\
                  cont & \sort{actl\_depart}{
                         [arg1 & @1]} ]] \\
 \avmspan{qstore \; \{\}}]
\end{avm}
\label{hab:8}
\end{examps}
In the airport domain, there are actually two habitual signs for
\qit{to depart}, one where \qit{to depart} requires an \qit{at~\dots}
prepositional-phrase complement (as in \pref{hab:7}), and one where
\qit{to depart} requires a \qit{from~\dots} prepositional-phrase
complement (this is needed in \pref{hab:8.1}). There are also two
non-habitual signs of \qit{to depart}, one where \qit{to depart}
requires no complement (as in \pref{hab:8}), and one where \qit{to
  depart} requires a \qit{from~\dots} prepositional-phrase complement
(needed in \pref{hab:8.2}). For simplicity, here I ignore these extra
signs.
\begin{examps}
\item BA737 (habitually) departs from gate 2. \label{hab:8.1}
\item BA737 (actually) departed from gate 2. \label{hab:8.2}
\end{examps}

\pref{hab:7}, \pref{hab:8}, and the simple-past lexical rules of
section \ref{single_word_forms} give rise to two signs (a habitual and
a non-habitual one) for the simple past \qit{departed}. These are the
same as \pref{hab:7} and \pref{hab:8}, except that they contain
additional \past operators. In contrast, the simple-present lexical
rule of section \ref{single_word_forms} generates only one sign for the
simple present \qit{departs}. This is the same as \pref{hab:7}, except
that it contains an additional \pres operator. No simple-present sign is
generated from \pref{hab:8}, because the simple-present lexical rule
requires the aspect of the base sign to be state.

The non-habitual simple-past sign of \qit{departed}, the \qit{at} sign
of \pref{pupe:2}, and the \qit{5:00pm} sign of \pref{pupe:3}, cause
\pref{hab:1} to be mapped to \pref{hab:13}, which expresses the
non-habitual reading of \pref{hab:1}. In this case, \qit{at
5:00pm} is treated as a temporal-adverbial modifier of
\qit{BA737 departed}, as discussed in section \ref{hpsg:punc_adv}.
\begin{examps}
\item $\partop[\text{\textit{5:00pm}}, fv^v] \land
       \at[fv^v, \past[e^v, act\_depart(ba737)]]$ \label{hab:13}
\end{examps}

In the habitual reading of \pref{hab:1}, where the habitual sign of
\qit{departed} (derived from \pref{hab:7}) is used, \qit{at 5:00pm} is
treated as a prepositional-phrase complement of \qit{departed}. In
this case, the sign of \qit{at} that introduces non-predicative
prepositional-phrase complements (i.e.\ \pref{pps:12}) is used. The
intention is to map \pref{hab:1} to \pref{hab:14}, where
\textit{5:00pm} is a constant acting as a ``generic representative''
of 5:00pm minutes (section \ref{hab_problems}).
\begin{examps}
\item $\past[e^v, hab\_departs\_at(ba737, \text{\textit{5:00pm}})]$
   \label{hab:14}
\end{examps}
The problem is that in this case the \qit{5:00pm} sign of
\pref{pupe:3} cannot be used, because it inserts a \partop operator in
{\feat qstore}. The semantics principle would cause this \partop
operator to be inherited by the sign of the overall \pref{hab:1}, and
thus the \partop operator would appear in the resulting formula. In
contrast, \pref{hab:14} (the intended formula for \pref{hab:1})
contains no \partop operators. To solve this problem, one has to allow
an extra sign for \qit{5:00pm}, shown in \pref{hab:15}, which does not
introduce a \partop. Similarly, an extra \qit{Monday} sign is needed
in \pref{hab:3}. (The fact that these extra signs have to be
introduced is admittedly inelegant. This is caused by the fact that
\qit{at 5:00pm} is treated differently in \pref{hab:13} and
\pref{hab:14}; see also the discussion in section \ref{hab_problems}.)
The \qit{5:00pm} sign of \pref{hab:15}, the \qit{at} sign that is used
when \qit{at} introduces non-predicative prepositional-phrase
complements (i.e.\ \pref{pps:12}), and the habitual \qit{departed}
sign (derived from \pref{hab:7}) cause \pref{hab:1} to be mapped to
\pref{hab:14}.
\avmoptions{active}
\begin{examps}
\item
\begin{avm}
[\avmspan{phon \; \<\fval 5:00pm\>} \\
 synsem|loc & [cat & [head  & \osort{noun}{
                              [prd & $-$]} \\
                      spr    & \<\> \\
                      subj   & \<\>  \\
                      comps  & \<\> ]\\ 
               cont & \osort{nom\_obj}{
                      [index & 5:00pm \\
                       restr &  \{\}]} ] \\
 \avmspan{qstore \; \{\}}]
\end{avm}
\label{hab:15}
\end{examps}
The habitual \qit{departed} sign (which derives from \pref{hab:7}) requires
the index of the prepositional-phrase complement to be of sort {\srt
  minute\_gappy}. As wanted, this does not allow the \qit{5:00pm} sign
of \pref{pupe:3} (the one that introduces a \partop) to be used in the
prepositional-phrase complement of the habitual \qit{departed},
because if \pref{pupe:3} is used, the index of the prepositional
phrase will be of sort {\srt minute\_ent\_var}, which is not a subsort
of {\srt minute\_gappy}\/ (see figure \vref{ind_hierarchy}). In
contrast, \pref{hab:15} introduces an index of sort {\srt 5:00pm},
which is a subsort of {\srt minute\_gappy}, and hence that sign can be
used in the complement of the habitual \qit{departed}.

The treatment of the simple present \pref{hab:2} is similar. In this
case, the habitual simple present sign (that is derived from
\pref{hab:7}) is used, and \pref{hab:2} is mapped to \pref{hab:16}.
No \topl formula is generated for the (impossible)
non-habitual reading of \pref{hab:2}, because there is no non-habitual
sign for the simple present \qit{departs} (see comments above about
the simple present lexical rule). 
\begin{examps}
\item $\pres[hab\_departs\_at(ba737, \text{\textit{5:00pm}})]$
   \label{hab:16}
\end{examps}


\section{Fronted temporal modifiers} \label{fronted}

As discussed in section \ref{hpsg:pupe_adv}, in this thesis
temporal-adverbial modifiers (e.g.\ \qit{at 5:00pm} in \pref{front:1}
-- \pref{front:2}, \qit{on Monday} in \pref{front:3} --
\pref{front:4}) can modify either whole finite sentences or past
participle verb phrases.
\begin{examps}
\item BA737 entered sector 2 at 5:00pm. \label{front:1}
\item At 5:00pm BA737 entered sector 2. \label{front:2}
\item Tank 2 was empty on Monday. \label{front:3}
\item On Monday tank 2 was empty. \label{front:4}
\end{examps}
In \hpsg, the order in which a modifier and the constituent to which
the modifier attaches can appear in a sentence is controlled by the
``constituent-ordering principle'' (\textsc{Cop}). This is a general
(and not fully developed) principle that controls the order in which
the various constituents can appear in a sentence (see chapter 7 of
\cite{Pollard1}). This thesis uses an over-simplified version of
\textsc{Cop}, that places no restriction on the order between temporal
modifiers and modified constituents when the modified constituents are
sentences. This allows \qit{at 5:00pm} to either follow \qit{BA737
  entered sector 2} (as in \pref{front:1}), or to precede it (as in
\pref{front:2}). Similarly, \qit{on Monday} may either follow
\qit{tank 2 was empty} (as in \pref{front:3}), or precede it (as in
\pref{front:4}).\footnote{An alternative approach is to allow temporal
  modifiers to participate in unbounded dependency constructions; see
  pp.~176 -- 181 of \cite{Pollard2}.} When temporal modifiers attach
to past-participle verb phrases, however, I require the modifiers to
follow the verb phrases, as in \pref{front:6}. This rules out
unacceptable sentences like \pref{front:5}, where \qit{at 5:00pm}
precedes the \qit{entered sector 2}.\footnote{Constituent-ordering
  restrictions are enforced in the \ale grammar of the prototype
  \nlitdb in a rather ad hoc manner, which involves partitioning the
  {\srt synsem}\/ sort into {\srt pre\_mod\_synsem}\/ and {\srt
    post\_mod\_synsem}, and using feature structures from the two
  subsorts as values of {\feat mod} to signal that the
  modifier can only precede or follow the modified constituent. This
  idea was borrowed from a grammar written by Suresh Manandhar.}
\begin{examps}
\item BA737 had [[entered sector 2] at 5:00pm]. \label{front:6}
\item \bad BA737 had [at 5:00pm [entered sector 2]]. \label{front:5}
\end{examps}
This approach causes \pref{front:7} to receive only \pref{front:8},
because in \pref{front:7} \qit{at 5:00pm} can modify only the whole
\qit{BA737 had entered sector 2} (it cannot modify just \qit{entered
  sector 2} because of the intervening \qit{BA737 had}). In
\pref{front:8}, 5:00pm is a reference time, a time where the entrance
had already occurred.  In contrast, \pref{front:8.5} receives both
\pref{front:8} and \pref{front:11}, because in that case \qit{at
  5:00pm} can modify either the whole \qit{BA737 had entered sector 2}
or only \qit{entered sector 2}. In \pref{front:11}, 5:00pm is the time
where the entrance occurred.
\begin{examps}
\item At 5:00pm [BA737 had entered sector 2]. \label{front:7}
\item $\partop[\text{\textit{5:00pm}}, fv^v] \land
       \at[fv^v, \past[e1^v, \perf[e2^v, enter(ba737, sector2)]]]$
   \label{front:8}
\item BA737 had entered sector 2 at 5:00pm. \label{front:8.5}
\item $\partop[\text{\textit{5:00pm}}, fv^v] \land
       \past[e1^v, \perf[e2^v, \at[fv^v, enter(ba737, sector2)]]]$
   \label{front:11}
\end{examps}
The fact that \pref{front:7} does not receive \pref{front:11} does not
seem to be a disadvantage, because in \pref{front:7} the reading where
\qit{at 5:00pm} specifies the time of the entrance seems unlikely (or
at least much more unlikely than in \pref{front:8.5}). 


\section{Temporal subordinate clauses}
\label{hpsg:subordinates}

I now discuss temporal subordinate clauses (section
\ref{subordinate_clauses}), focusing on \qit{while~\dots} clauses. The
treatment of \qit{before~\dots} and \qit{after~\dots} clauses is very
similar. 

As with period adverbials, \qit{while~\dots} clauses are treated as
temporal modifiers of finite sentences or past participle verb
phrases. As with prepositions introducing period adverbials,
\qit{while} is given two signs. The first one, shown in \pref{subs:1},
introduces an \lend operator, causes an aspectual shift to point, and
can be used only with culminating activity main clauses (\pref{subs:1}
is similar to \pref{pupe:29}.) The second one is the same as
\pref{subs:1}, except that it does not introduce an \lend, it
preserves the aspectual class of the main clause, and it can be used
with main clauses of any aspectual class.  In both cases, \qit{while}
requires as its complement a finite sentence whose aspect must not be
consequent state (this agrees with table \vref{while_clauses_table},
which does not allow the aspectual class of the \qit{while}-clause to
be consequent state).
\begin{examps}
\avmoptions{active, center}
\item 
\setbox\avmboxa=\hbox{\begin{avm}
[mod & \feat s[vform {\fval fin}]:@2 $\lor$ 
       \feat vp[vform {\fval psp}]:@2 \\
 \avmspan{mod|loc|cat|aspect \; {\fval culmact}}]
\end{avm}}
\begin{avm}
[\avmspan{phon \; \<\fval while\>} \\
 synsem|loc & [cat & [head   & \box\avmboxa \\
                      spr    & \<\>  \\
                      subj   & \<\>  \\
                      comps  & \<\feat s[vform & {\fval fin}\\
                                         aspect & $\neg$cnsq\_state]:@1\> \\
                      aspect & point]\\ 
               cont & \osort{at\_op}{
                      [time\_spec & @1 \\
                       main\_psoa & \osort{end}{
                                    [main\_psoa & @2]}]}] \\
 \avmspan{qstore \; \{\}} ]
\end{avm}
\label{subs:1}
\end{examps}
The \avmbox{1} in \pref{subs:1} denotes the {\feat cont} of the sign
of the complement of \qit{while} (the subordinate clause). The two
\qit{while} signs cause \pref{subs:3} to receive \pref{subs:4} and
\pref{subs:5}. (\qit{To land} is a culminating activity verb in the
airport domain). \pref{subs:4} requires the landing to have simply been
completed during the inspection, while \pref{subs:5} requires
the landing to have both started and been completed during the
inspection.
\begin{examps}
\item UK160 landed while J.Adams was inspecting BA737. \label{subs:3}
\item $\begin{aligned}[t]
       \at[&\past[e1^v, inspecting(j\_adams, ba737)], \\
           &\lend[\past[e2^v, \culm[landing(occr^v, uk160)]]]]
       \end{aligned}$ 
   \label{subs:4}
\item $\begin{aligned}[t]
       \at[&\past[e1^v, inspecting(j\_adams, ba737)], \\
           &\past[e2^v, \culm[landing(occr^v, uk160)]]]
       \end{aligned}$
   \label{subs:5}
\end{examps}

Since \qit{while~\dots} clauses are treated as temporal modifiers, the
ordering arrangements of section \ref{fronted} apply to
\qit{while~\dots} clauses as well. Hence, \qit{while~\dots} clauses
can either precede or follow finite sentences (e.g.\ \pref{subs:8.1},
\pref{subs:9}).
\begin{examps}
\item UK160 arrived while J.Adams was inspecting BA737. \label{subs:8.1}
\item While J.Adams was inspecting BA737, UK160 arrived. \label{subs:9}
\end{examps}

One problem with the present treatment of \qit{while~\dots} clauses is
that it maps \pref{subs:10} to \pref{subs:11}, which requires the
inspection to have been completed. This does not agree with table
\vref{while_clauses_table}, according to which any requirement that
the situation of a culminating activity sentence must have been
reached is cancelled when the sentence is used as a \qit{while~\dots}
clause. To overcome this problem, the post-processing (section
\ref{post_processing} below) removes any \culm operators that are within
first arguments of \at operators. This removes the \culm of
\pref{subs:11}, generating a formula that no longer requires the
inspection to have been completed.
\begin{examps}
\item UK160 departed while J.Adams inspected BA737. \label{subs:10}
\item $\begin{aligned}[t]
       \at[&\past[e1^v, \culm[inspecting(j\_adams, ba737)]], \\
           &\past[e2^v, [actl\_depart(uk160)]]]
       \end{aligned}$
   \label{subs:11}
\end{examps}


\section{Interrogatives} \label{unb_dep}

So far, this chapter has considered mainly assertions (e.g.\ 
\pref{unb:1}). (The reader is reminded that assertions are treated as
yes/no questions; e.g.\ \pref{unb:1} is treated as \pref{unb:3}.) I
now explain how the \hpsg version of this thesis copes with questions
(e.g.\ \pref{unb:3} -- \pref{unb:8}).
\begin{examps}
\item Tank 2 was empty. \label{unb:1}
\item Was tank 2 empty? \label{unb:3}
\item Did J.Adams inspect BA737? \label{unb:4}
\item Which tank was empty? \label{unb:5}
\item Who inspected BA737? \label{unb:6}
\item What did J.Adams inspect? \label{unb:7}
\item When did J.Adams inspect BA737? \label{unb:8}
\end{examps}

Yes/no questions (e.g.\ \pref{unb:3}, \pref{unb:4}) constitute no
particular problem. \hpsg's schemata allow auxiliary verbs to be used
in sentence-initial positions, and cause \pref{unb:3} to receive the
same formula (shown in \pref{unb:9.1}) as \pref{unb:9}. In both cases,
the same lexical signs are used. Similar comments apply to
\pref{unb:4} and \pref{unb:10}, which are mapped to \pref{unb:11}.
\begin{examps}
\item Tank 2 was empty. \label{unb:9}
\item $\past[e^v, empty(tank2)]$ \label{unb:9.1}
\item J.Adams did inspect BA737. \label{unb:10}
\item $\past[e^v, \culm[inspecting(occr^v, j\_adams, ba737)]]$
   \label{unb:11}
\end{examps}

The interrogative \qit{which} is treated syntactically as a determiner
of (non-predicative) noun phrases. The sign of \qit{which} is the same
as the sign of \qit{the} of \pref{nps:4}, except that it introduces an
interrogative quantifier rather than an existential one. For example,
\pref{unb:5} is analysed syntactically in the same way as
\pref{unb:12} (punctuation is ignored). However, the formula of
\pref{unb:5} (shown in \pref{unb:14}) contains an additional
interrogative quantifier (cf.\ the formula of \pref{unb:12}, shown in
\pref{unb:13}). (I assume here that \qit{tank} does not introduce an
\ntense. The \qit{a} of \pref{unb:12} introduces an existential
quantifier which is removed during the extraction of \pref{unb:13}
from the sign of \pref{unb:12}, as discussed in section
\ref{extraction_hpsg}.)
\begin{examps}
\item A tank was empty. \label{unb:12}
\item $tank(tk^v) \land \past[e^v, empty(tk^v)]$ \label{unb:13}
\item $?tk^v \; tank(tk^v) \land \past[e^v, empty(tk^v)]$ \label{unb:14}
\end{examps}

The interrogative \qit{who} is treated syntactically as a
non-predicative noun-phrase. Its sign, shown in \pref{unb:15},
introduces an interrogative quantifier. 
\begin{examps}
\avmoptions{active}
\item
\begin{avm}
[\avmspan{phon \; \<\fval who\>} \\
 synsem|loc & [cat & [head  & \osort{noun}{
                              [prd & $-$]} \\
                      spr    & \<\> \\
                      subj   & \<\>  \\
                      comps  & \<\> ]\\ 
               cont & \osort{nom\_obj}{
                      [index & \sort{person\_ent}{
                               [tvar & $+$]} \\
                       restr &  \{\}]}@1] \\
 \avmspan{qstore \; \{\sort{quant}{
                      [det & interrog \\
                       restind & @1]}\}}]
\end{avm}
\label{unb:15}
\end{examps}
\pref{unb:6} is analysed syntactically in the same way as
\pref{unb:16}. The sign of \qit{who}, however, gives rise to an
interrogative quantifier in the formula of \pref{unb:6} (shown in
\pref{unb:18}), which is not present in the formula of \pref{unb:16}
(shown in \pref{unb:17}). The interrogative \qit{what} is treated
similarly.
\begin{examps}
\item J.Adams inspected BA737. \label{unb:16}
\item $\past[\culm[inspecting(occr^v, j\_adams, ba737)]]$ \label{unb:17}
\item $?wh^v \; \past[\culm[inspecting(occr^v, wh^v, ba737)]]$ 
   \label{unb:18}
\end{examps}

The \hpsg version of this thesis admits questions like \pref{unb:19},
which are unacceptable in most contexts. \pref{unb:19} is licensed by the
same syntactic analysis that allows \pref{unb:20}, and receives the same
formula as \pref{unb:21}. 
\begin{examps}
\item \odd Did J.Adams inspect which flight? \label{unb:19}
\item Did J.Adams inspect a flight? \label{unb:20}
\item Which flight did J.Adams inspect? \label{unb:21}
\end{examps}
Questions like \pref{unb:21}, where the interrogative refers to the
object of the verb, are treated using \hpsg's unbounded-dependencies
mechanisms (more precisely, using the {\feat slash} feature; see
chapter 4 of \cite{Pollard2}).\footnote{Pollard and Sag also reserve a
  {\feat que} feature, which is supposed to be used in the treatment
  of interrogatives. They provide virtually no information on the role
  of {\feat que}, however, pointing to \cite{Ginzburg1992} where
  {\feat que} is used in a general theory of interrogatives.
  Ginzburg's theory is intended to address issues well beyond the
  scope of this thesis (e.g.\ the relation between a question and the
  facts that can be said to \emph{resolve} that question; see also
  \cite{Ginzburg1995}, \cite{Ginzburg1995b}). {\feat que} is not used
  in this thesis.} Roughly speaking, \pref{unb:21} is analysed as
being a form of \pref{unb:19}, where the object \qit{which flight} has
moved to the beginning of the question. \hpsg's unbounded-dependencies
mechanisms will not be discussed here (see \cite{Pollard2}; the
prototype \nlitdb uses the traceless analysis of unbounded
dependencies, presented in chapter 9 of \cite{Pollard2}).

The present treatment of interrogatives allows questions with multiple
interrogatives, like \pref{unb:24} which receives \pref{unb:24.1}.
(\pref{unb:24} is parsed in the same way as \pref{unb:25}.)
Unfortunately, it also allows ungrammatical questions like
\pref{unb:26}, which is treated as a version of \pref{unb:24} where
the \qit{what} complement has moved to the beginning of the sentence.
(\pref{unb:26} receives \pref{unb:24}.)
\begin{examps}
\item Who inspected what. \label{unb:24}
\item $?w1^v \; ?w2^v \; \past[e^v, \culm[inspecting(occr^v, w1^v, w2^v)]]$
   \label{unb:24.1}
\item J.Adams inspected BA737. \label{unb:25}
\item \bad What who inspected. \label{unb:26}
\end{examps}

The interrogative \qit{when} of \pref{unb:27} is treated as a
temporal-adverbial modifier of finite sentences. \pref{unb:28} shows
the sign of \qit{when} that is used in \pref{unb:27}. \pref{unb:28}
causes \pref{unb:27} to receive \pref{unb:27.1}.
\avmoptions{active, center}
\begin{examps}
\item When was tank 2 empty? \label{unb:27}
\item $?_{mxl}w^v \; \past[e^v, empty(tank2)]$ \label{unb:27.1}
\item 
\setbox\avmboxa=\hbox{\begin{avm}
[mod & \feat s[vform {\fval fin}]:@1 \\
 \avmspan{mod|loc|cat|aspect \; @2}]
\end{avm}}
\setbox\avmboxb=\hbox{\begin{avm}
[det & interrog\_mxl \\
 restind & [index & \sort{temp\_ent}{
                    [tvar & $+$]} \\
            restr & \{\}]]
\end{avm}}
\begin{avm}
[\avmspan{phon \; \<\fval when\>} \\
 synsem|loc & [cat & [head   & \box\avmboxa \\
                      spr    & \<\>  \\
                      subj   & \<\>  \\
                      comps  & \<\> \\
                      aspect & @2]\\ 
               cont & @1] \\
 \avmspan{qstore \; \{\box\avmboxb\}} ]
\end{avm}
\label{unb:28}
\end{examps}
\pref{unb:28} introduces interrogative-maximal
quantifiers whose variables ($w^v$ in \pref{unb:27.1}) do not appear
elsewhere in the formula. The post-processing
(to be discussed in section \ref{post_processing}) replaces the
variables of interrogative-maximal quantifiers by
variables that appear as first arguments of \past or \perf
operators. In \pref{unb:27.1}, this would replace $w^v$
by $e^v$, generating a formula that asks for the maximal past
periods where tank 2 was empty.

There is also a second sign for the interrogative \qit{when} (shown in
\pref{unb:32}), that is used in habitual questions like \pref{unb:29}.
In \pref{unb:29}, \qit{when} is taken to play the same role as 
\qit{at 5:00pm} in \pref{unb:30}, i.e.\ it is treated as the
prepositional-phrase complement of the habitual \qit{depart} (see
section \ref{habituals}), which has moved to the beginning of the
sentence via the unbounded-dependencies mechanisms.
\avmoptions{active, center}
\begin{examps}
\item When does BA737 depart (habitually)? \label{unb:29}
\item Does BA737 depart (habitually) at 5:00pm? \label{unb:30}
\item 
\setbox\avmboxa=\hbox{\begin{avm}
[det & interrog \\
 restind & @1]
\end{avm}}
\begin{avm}
[\avmspan{phon \; \<\fval when\>} \\
 synsem|loc & [cat & [head  & \osort{prep}{
                              [prd   & $-$]} \\
                      spr    & \<\> \\
                      subj   & \<\>  \\
                      comps  & \<\> ]\\ 
               cont & \osort{nom\_obj}{
                      [index & \sort{gappy\_partng}{
                               [tvar & $+$]} \\
                       restr & \{\}]}@1] \\
 \avmspan{qstore \; \{\box\avmboxa\}}]
\end{avm}
\label{unb:32}
\end{examps}

In the simple past \pref{unb:33}, both the (state) habitual homonym of
\qit{to depart} (that of \pref{hab:7}, which requires a prepositional
phrase complement) and the (point) non-habitual homonym (that of
\pref{hab:8}, which requires no complement) can be used. Hence,
\qit{when} can be either a prepositional-phrase complement of the
habitual \qit{depart} (using \pref{unb:32}), or a temporal modifier of
the non-habitual sentence \qit{did BA737 depart} (using
\pref{unb:28}). This gives rise to \pref{unb:34} and \pref{unb:35},
which correspond to the habitual and non-habitual readings of
\pref{unb:33} (the $w^v$ of \pref{unb:35} would be replaced by $e^v$
during the post-processing).
\begin{examps}
\item When did BA737 depart? \label{unb:33}
\item $?w^v \; \past[e^v, hab\_departs\_at(ba737, w^v)]$ \label{unb:34}
\item $?_{mxl}w^v \; \past[e^v, actl\_depart(ba737)]$ \label{unb:35}
\end{examps}


\section{Multiple temporal modifiers} \label{hpsg:mult_mods}

The framework of this thesis currently runs into several problems
in sentences with multiple temporal modifiers. This section discusses these
problems. 

\paragraph{Both preceding and trailing temporal modifiers:}
Temporal modifiers are allowed to either
precede or follow finite sentences (section
\ref{fronted}). When a finite sentence is modified by both a
preceding and a trailing temporal modifier (as in \pref{mults:1}),
two parses are generated: one where the trailing modifier
attaches first to the sentence (as in \pref{mults:2}), and one where
the preceding modifier attaches first (as in \pref{mults:4}). In most
cases, this generates two semantically equivalent formulae
(\pref{mults:3} and \pref{mults:5} in the case of \pref{mults:1}). A
mechanism is needed to eliminate one of the two formulae.  
\begin{examps}
\item Yesterday BA737 was at gate 2 for two hours. \label{mults:1}
\item Yesterday [[BA737 was at gate 2] for two hours.] \label{mults:2}
\item $\at[yesterday, \for[hour^c, 2, \past[e^v, located\_at(ba737, gate2)]]]$
   \label{mults:3}
\item {[}Yesterday [BA737 was at gate 2]] for two hours. \label{mults:4}
\item $\for[hour^c, 2, \at[yesterday, \past[e^v, located\_at(ba737, gate2)]]]$
   \label{mults:5}
\end{examps}

\paragraph{Multiple temporal modifiers and anaphora:}
Another problem is that a question like \pref{mults:10} is mapped to
\pref{mults:11}. (I assume here that \qit{flight} does not introduce
an \ntense.) The problem with \pref{mults:11} is that it does not
require $fv^v$ to be the particular 5:00pm-minute of 2/11/95.
\pref{mults:11} requires the flight to have arrived on 2/11/95 and
after an arbitrary 5:00pm-minute (e.g.\ the 5:00pm-minute of 1/11/95).
In effect, this causes the \qit{after 5:00pm} to be ignored.
\begin{examps}
\item Which flight arrived after 5:00pm on 2/11/95? \label{mults:10}
\item $?fl^v \; flight(fl^v) \land \partop[\text{\textit{5:00pm}}^g, fv^v]
       \land$\\
      $\at[\text{\textit{2/11/95}}, \after[fv^v, \past[e^v, arrive(fl^v)]]]$
   \label{mults:11}
\end{examps}
This problem seems related to the need for temporal anaphora
resolution mechanisms (section \ref{temporal_anaphora}). In
\pref{mults:16}, for example, the user most probably has a particular
(contextually-salient) 5:00pm-minute in mind, and an anaphora
resolution mechanism is needed to determine that minute. A similar
mechanism could be responsible for reasoning that in \pref{mults:10}
the most obvious contextually salient 5:00pm-minute is that of
2/11/95.
\begin{examps}
\item Which tanks were empty before/at/after 5:00pm? \label{mults:16}
\end{examps}

\paragraph{Culminating activity with both punctual and period adverbial:}
A further problem appears when a culminating activity is modified by
both a punctual and a period adverbial.\footnote{The problems of this
  section that involve period adverbials also arise when temporal
  subordinate clauses are used instead of period adverbials.} The
problem is that, unlike what one would expect, \pref{mults:18} and
\pref{mults:19} do not receive equivalent \topl formulae. (I assume
here that \qit{to repair} is classified as culminating activity verb.)
\begin{examps}
\item J.Adams repaired fault 2 at 5:00pm on 2/11/95. \label{mults:18}
\item J.Adams repaired fault 2 on 2/11/95 at 5:00pm. \label{mults:19}
\end{examps}
In \pref{mults:18}, the punctual adverbial \qit{at 5:00pm} modifies
the culminating activity sentence \qit{J.Adams repaired fault 2}. The
punctual adverbial causes \qit{J.Adams repaired fault 2 at 5:00pm} to
become a point (see table \vref{punctual_adverbials_table}). Two
formulae are generated: one that requires the repair to have started
at 5:00pm, and one that requires the repair to have been completed at
5:00pm.  \qit{On 2/11/95} then modifies the point expression
\qit{J.Adams repaired fault 2 at 5:00pm}. This leads to
\pref{mults:22} and \pref{mults:23}. In \pref{mults:22} the repair
\emph{starts} at the 5:00pm-minute of 2/11/95, while in
\pref{mults:23} the repair is \emph{completed} at the 5:00pm-minute of
2/11/95. (The first reading is easier to accept in \qit{J.Adams
  inspected BA737 at 5:00pm on 2/11/95}.)
\begin{examps}
\item $\partop[\text{\textit{5:00pm}}^g, fv^v] \land 
       \at[\text{\textit{2/11/95}},$\\
 $\at[fv^v, \lbegin[\past[e^v, \culm[repairing(occr^v, j\_adams, fault2)]]]]]$
\label{mults:22}
\item $\partop[\text{\textit{5:00pm}}^g, fv^v] \land 
       \at[\text{\textit{2/11/95}},$\\
 $\at[fv^v, \lend[\past[e^v, \culm[repairing(occr^v, j\_adams, fault2)]]]]]$
\label{mults:23}
\end{examps}
(A digression: this example also demonstrates why punctual adverbials
are taken to trigger an aspectual shift to point; see section
\ref{point_adverbials}. Without this shift, the aspectual class of
\qit{J.Adams repaired fault 2 at 5:00pm} would be culminating
activity, and the \qit{on} signs of section \ref{hpsg:per_advs} would
lead to the additional formulae of \pref{mults:22f} and
\pref{mults:23f}. These are equivalent to \pref{mults:22} and
\pref{mults:23} respectively.)
\begin{examps}
\item $\partop[\text{\textit{5:00pm}}^g, fv^v] \land 
       \at[\text{\textit{2/11/95}}, $\\
 $\lend[\at[fv^v, \lbegin[\past[e^v, 
    \culm[repairing(occr^v, j\_adams, fault2)]]]]]]$
\label{mults:22f}
\item $\partop[\text{\textit{5:00pm}}^g, fv^v] \land 
       \at[\text{\textit{2/11/95}},$\\
 $\lend[\at[fv^v, \lend[\past[e^v, 
   \culm[repairing(occr^v, j\_adams, fault2)]]]]]]$
\label{mults:23f}
\end{examps}

In \pref{mults:19}, \qit{J.Adams repaired fault 2} is
first modified by the period adverbial \qit{on 2/11/95}. Two formulae
(shown in \pref{mults:24} and \pref{mults:25}) are
generated. \pref{mults:24} requires the repair to simply reach its
completion on 2/11/95, while \pref{mults:25} requires the repair to
both start and reach its completion on 2/11/95. In the first case (where
\pref{mults:24} is generated), the aspectual class of \qit{J.Adams repaired
fault 2 on 2/11/95} becomes point, while in the other case the
aspectual class remains culminating activity (see also table
\vref{period_adverbials_table}). 
\begin{examps}
\item $\at[\text{\textit{2/11/95}}, 
    \lend[\past[e^v, \culm[repairing(occr^v, j\_adams, fault2)]]]]$
\label{mults:24}
\item $\at[\text{\textit{2/11/95}},
    \past[e^v, \culm[repairing(occr^v, j\_adams, fault2)]]]$
\label{mults:25}
\end{examps}
In the case of \pref{mults:24}, where the aspectual class of \qit{J.Adams
repaired fault 2 on 2/11/95} is point,
the signs of section \ref{hpsg:punc_adv} lead to \pref{mults:26},
while in the case of \pref{mults:25}, they lead to \pref{mults:27} and \pref{mults:28}.
\begin{examps}
\item $\partop[\text{\textit{5:00pm}}^g, fv^v] \land \at[fv^v,$\\
      $\at[\text{\textit{2/11/95}}, 
       \lend[\past[e^v, \culm[repairing(occr^v, j\_adams, fault2)]]]]]$
\label{mults:26}
\item $\partop[\text{\textit{5:00pm}}^g] \land \at[fv^v,$\\
      $\lbegin[
       \at[\text{\textit{2/11/95}}, 
       \past[e^v, \culm[repairing(occr^v, j\_adams, fault2)]]]]]$
\label{mults:27}
\item $\partop[\text{\textit{5:00pm}}^g] \land \at[fv^v,$\\
      $\lend[
       \at[\text{\textit{2/11/95}}, 
       \past[e^v, \culm[repairing(occr^v, j\_adams, fault2)]]]]]$
\label{mults:28}
\end{examps}
Hence, \pref{mults:18} receives two formulae (\pref{mults:22} and
\pref{mults:23}), while \pref{mults:19} receives three
(\pref{mults:26} -- \pref{mults:28}). \pref{mults:26} is equivalent to
\pref{mults:23}. They both require the repair to reach its completion
within the 5:00pm-minute of 2/11/95. Unlike what one might expect,
however, \pref{mults:27} is not equivalent to \pref{mults:22}.
\pref{mults:27} requires a past period that covers exactly the whole
repair (from start to completion) to fall within 2/11/95, and the
beginning of that period to fall within some 5:00pm-minute. This means
that the repair must start at the 5:00pm-minute of 2/11/95 (as in
\pref{mults:22}), but it also means that the repair must reach its
completion within 2/11/95 (this is not a requirement in
\pref{mults:22}). Also, unlike what one might expect, \pref{mults:28}
is not equivalent to \pref{mults:23} and \pref{mults:26}.
\pref{mults:28} requires the repair to reach its completion within the
5:00pm-minute of 2/11/95 (as in \pref{mults:23} and \pref{mults:26}),
but it also requires the repair to start within 2/11/95 (which is not
a requirement in \pref{mults:23} and \pref{mults:26}).

The differences in the number and semantics of the generated formulae
in \pref{mults:18} and \pref{mults:19} lead to differences in the
behaviour of the \nlitdb that are difficult to explain to the user.  A
tentative solution is to adopt some mechanism that would reorder the
temporal modifiers, so that the punctual adverbial attaches before the
period one. This would reverse the order of \qit{on 2/11/95} and
\qit{at 5:00pm} in \pref{mults:19}, and would cause \pref{mults:19} to
be treated in the same way as \pref{mults:18} (i.e.\ to be mapped to
\pref{mults:22} and \pref{mults:23}; these seem to capture the most
natural readings of \pref{mults:18} and \pref{mults:19}).

\paragraph{Culminating activity and multiple period adverbials:}
A further problem is that a sentence like \pref{mults:30}, where a
culminating activity is modified by two period adverbials, receives
three formulae, shown in \pref{mults:32} -- \pref{mults:31}. It
turns out that \pref{mults:33} is equivalent to \pref{mults:31}, and
hence one of the two should be eliminated.
\begin{examps}
\item J.Adams repaired fault 2 in June in 1992. \label{mults:30}
\item $\partop[june^g, j^v] \land \at[1992,$\\
   $\at[j^v, \lend[\past[e^v, \culm[repairing(occr^v, j\_adams, fault2)]]]]]$
   \label{mults:32}
\item $\partop[june^g, j^v] \land \at[1992,$\\
   $\lend[\at[j^v, \past[e^v, \culm[repairing(occr^v, j\_adams, fault2)]]]]]$
   \label{mults:33}
\item $\partop[june^g, j^v] \land \at[1992,$\\
   $\at[j^v, \past[e^v, \culm[repairing(occr^v, j\_adams, fault2)]]]]$
   \label{mults:31}
\end{examps}
A period adverbial combining with a culminating activity can either
insert an \lend operator and cause an aspectual shift to point, or
insert no \lend and leave the aspectual class unchanged (see section
\ref{hpsg:per_advs}). In the case where \pref{mults:32} is generated,
\qit{in June} inserts an \lend and changes the aspectual class to
point. This does not allow \qit{in 1992} (which attaches after \qit{in
  June}) to insert an \lend, because period adverbials combining with
points are not allowed to insert \lend{s} (the \qit{on} sign of
\pref{pupe:29} cannot be used with points). In the cases where
\pref{mults:33} or \pref{mults:31} are generated, \qit{in June} does
not insert an \lend, and the aspectual class remains culminating
activity. \qit{In 1992} can then insert an \lend (as in
\pref{mults:33}) or not (as in \pref{mults:31}). \pref{mults:31}
requires the whole repair to be located within a June and 1992 (i.e.\ 
within the June of 1992). \pref{mults:32} is weaker: it requires only
the completion point of the repair to be located within the June of
1992. Finally, \pref{mults:33} requires the whole of the repair to be
located within a June, and the completion point of the repair to fall
within 1992. This is equivalent to requiring the whole of the repair
to fall within the June of 1992, i.e.\ \pref{mults:33} is equivalent
to \pref{mults:31}, and one of the two should be eliminated.


\section{Post-processing}  \label{post_processing}

The parsing maps each English question to an \hpsg sign (or multiple
signs, if the parser understands the question to be ambiguous). From
that sign, a \topl formula is extracted as discussed in section
\ref{extraction_hpsg}. The extracted formula then undergoes an
additional post-processing phase. This is a collection of minor
transformations, discussed below, that cannot be carried out easily
during the parsing.

\paragraph{Removing Culms:} \pref{post:2} shows the \topl formula that
is extracted from the sign of \pref{post:1}. As discussed in section
\ref{duration_adverbials}, \pref{post:2} does not represent correctly
\pref{post:1}, because \pref{post:2} requires the taxiing to have been
completed. In contrast, as discussed in section \ref{for_adverbials},
the \qit{for~\dots} adverbial of \pref{post:1} cancels the normal
implication of \qit{BA737 taxied to gate 2} that the taxiing must have
been completed. To express correctly \pref{post:1}, the \culm of
\pref{post:2} has to be removed.
\begin{examps}
\item BA737 taxied to gate 2 for five minutes. \label{post:1}
\item $\for[minute^c, 5, \past[e^v, \culm[taxiing\_to(ba737, gate2)]]]$
   \label{post:2}
\end{examps}
A first solution would be to remove during the post-processing any
\culm operator that is within the scope of a \for operator. The
problem with this approach is that duration \qit{in~\dots} adverbials
also introduce \for operators (see section \ref{duration_adverbials}), but
unlike \qit{for~\dots} adverbials they do not cancel the implication
that the completion must have been reached. For example, the formula
extracted from the sign of \pref{post:5} is \pref{post:2}. In this
case, \pref{post:2} is a correct rendering of \pref{post:5} (because
\pref{post:5} \emph{does} imply that BA737 reached gate 2), and hence
the \culm operator should not be removed. To overcome this problem, the
prototype \nlitdb attaches to each \for operator a flag showing
whether it was introduced by a \qit{for~\dots} or an \qit{in~\dots}
adverbial. Only \for operators introduced by \qit{for~\dots}
adverbials cause \culm operators within their scope to be removed.
\begin{examps}
\item BA737 taxied to gate 2 in five minutes. \label{post:5}
\end{examps}

The post-processing also removes any \culm operator from within the
first argument of an \at operator. As explained in section
\ref{hpsg:subordinates}, this is needed to express correctly
\qit{while~\dots} clauses.

\paragraph{$\mathbf{?_{mxl}}$ quantifiers:} As noted in section \ref{unb_dep},
before the post-processing the variables of interrogative-maximal
quantifiers introduced by \qit{when} do not occur elsewhere in their
formulae. For example, \pref{post:9} and \pref{post:6} are 
extracted from the signs of \pref{post:8} and \pref{post:6}. In both
formulae, $w^v$ occurs only immediately after the $?_{mxl}$. 
\begin{examps}
\item When was J.Adams a manager? \label{post:8}
\item $?_{mxl}w^v \; \past[e^v, manager(j\_adams)]$ \label{post:9}
\item When while BA737 was circling was runway 2 open? \label{post:6}
\item $?_{mxl}w^v \; \at[\past[e1^v, circling(ba737)], 
                         \past[e2^v, open(runway2)]]$ 
   \label{post:7}
\end{examps}
During the post-processing, the variables of interrogative-maximal
quantifiers are replaced by variables that appear as first arguments
of \past or \perf operators, excluding \past and \perf operators that
are within first arguments of \at, \before, or \after operators. In
\pref{post:9}, this causes $w^v$ to be replaced by $e^v$. The
resulting formula asks for the maximal past periods where J.Adams was
a manager. Similarly, the $w^v$ of \pref{post:7} is replaced by
$e2^v$. The resulting formula asks for the maximal past periods
$e2^v$, such that runway 2 was open at $e2^v$, and $e2^v$ is a
subperiod of a period $e1^v$ where BA737 was circling. In
\pref{post:7}, $w^v$ cannot be replaced by $e1^v$, because
$\past[e1^v, circling(ba737)]$ is within the first argument of an \at.

\past and \perf operators located within first arguments of \at,
\before, or \after operators are excluded, to avoid interpreting
\qit{when} as referring to the time where the situation of a
subordinate clause held (formulae that express subordinate clauses
end-up within first arguments of \at, \before, or \after operators).
The interrogative \qit{when} always refers to the situation of the
main clause. For example, \pref{post:6} cannot be asking for maximal
periods where BA737 was circling that subsume periods where runway 2
was open (this would be the meaning of \pref{post:7} if $w^v$ were
replaced by $e1^v$).

When the main clause is in the past perfect, this arrangement allows
the variable of $?_{mxl}$ to be replaced by either the first argument
of the main-clause's \past operator, or the first argument of the
main-clause's \perf operator. \pref{post:11}, for example, shows the
formula extracted from the sign of \pref{post:10}. The
post-processing generates two formulae: one where $w^v$ is replaced by
$e1^v$, and one where $w^v$ is replaced by $e2^v$. The first one asks
for what section \ref{point_adverbials} called the ``consequent
period'' of the inspection (the period from the end of the inspection
to the end of time). The second one asks for the time of the actual
inspection.
\begin{examps}
\item When had J.Adams inspected BA737? \label{post:10}
\item $?_{mxl}w^v \; \past[e1^v, \perf[e2^v, 
               \culm[inspecting(occr^v, j\_adams, ba737)]]]$
   \label{post:11}
\end{examps}

\paragraph{Ntense operators:} As noted in section \ref{non_pred_nps},
when extracting \topl formulae from signs, if an \ntense operator is
encountered and the sign contains no definite indication that the
first argument of the \ntense should be $now^*$, in the extracted
formula the first argument of the \ntense becomes a variable. That
variable does not occur elsewhere in the extracted formula. Assuming,
for example, that the (non-predicative) \qit{queen} introduces an
\ntense, the formula extracted from the sign of \pref{post:14} is
\pref{post:15}. The $t^v$ of the \ntense does not occur elsewhere in
\pref{post:15}.
\begin{examps}
\item The queen was in Rome. \label{post:14}
\item $\ntense[t^v, queen(q^v)] \land
       \past[e1^v, located\_at(q^v, rome)]$ \label{post:15}
\end{examps}
During the post-processing, variables appearing as first arguments of
\ntense{s} give rise to multiple formulae, where the first arguments
of the \ntense{s} are replaced by $now^*$ or by first arguments of
\past or \perf operators. In \pref{post:15}, for example, the
post-processing generates two formulae: one where $t^v$ is replaced by
$now^*$ (queen at the speech time), and one where $t^v$ is replaced by $e^v$
(queen when in Rome).

In \pref{post:17} (the formula extracted from the sign of
\pref{post:16}), there is no \past or \perf operator, and hence $t^v$
can only become $now^*$. This captures the fact that the \qit{queen}
in \pref{post:16} most probably refers to the queen of the speech
time.
\begin{examps}
\item The queen is in Rome. \label{post:16}
\item $\ntense[t^v, queen(q^v)] \land 
       \pres[located\_at(q^v, gate2)]$ \label{post:17}
\end{examps}
In \pref{post:19} (the formula extracted from the sign of
\pref{post:18}), the post-processing leads to three formulae, where
$t^v$ is replaced by $now^*$ (queen at speech time), $e2^v$ (queen 
during the visit), or $e1^v$ (queen at a ``reference time'' after
the visit).
\begin{examps}
\item The queen had visited Rome. \label{post:18}
\item $\ntense[t^v, queen(q^v)] \land 
       \past[e1^v, \perf[e2^v, visiting(q^v, rome)]]$ \label{post:19}
\end{examps}


\section{Summary}

This chapter has shown how \hpsg can be used to translate English
questions directed to a \nlitdb to appropriate \topl formulae. During
the parsing, each question receives one or more \hpsg signs, from
which \topl formulae are extracted. The extracted formulae then
undergo an additional post-processing phase, which leads to formulae
that capture the semantics of the original English questions.

Several modifications were made to \hpsg. The main modifications were:
(i) \hpsg features and sorts that are intended to account for
phenomena not examined in this thesis (e.g.\ pronouns, relative
clauses, number agreement) were dropped. (ii) The quantifier storage
mechanism of \hpsg was replaced by a more primitive one, that does not
allow quantifiers to be unstored during the parsing; the semantics
principle was modified accordingly. (iii) An {\feat aspect} feature
was added, along with a principle that controls its propagation. (iv)
The possible values of {\feat cont} and {\feat qstore} were modified,
to represent \topl expressions rather than situation-theory
constructs. (v) A hierarchy of world-entity types was mounted under
the {\srt ind}\/ sort; this is used to disambiguate sentences, and to
block semantically ill-formed ones. (vi) New lexical signs and lexical
rules were introduced to cope with temporal linguistic mechanisms
(verb tenses, temporal adverbials, temporal subordinate clauses, etc.).
Apart from these modifications, the \hpsg version of this thesis
follows closely \cite{Pollard2}.



\chapter{From TOP to TSQL2} \label{tdb_chapter}

\proverb{Time is money.}


\section{Introduction} 

This chapter describes the translation from \topl to \tsql. The
discussion starts with an introduction to \tsql and the version of the
relational model on which \tsql is based. This thesis adopts some
modifications to \tsql. These are described next, along with some
minor alterations in the \topl definition of chapter
\ref{TOP_chapter}. The translation from \topl to \tsql requires
\topl's model to be linked to the database; this is explained
next. The translation is carried out by a set of rules. I explore
formally the properties that these rules must possess for the
translation to be correct, and I describe the intuitions behind the
design of the rules. An illustration of how some of the rules work is
also given. The full set of the translation rules, along with a proof
that they possess the necessary properties, is given in appendix
\ref{trans_proofs}. The chapter ends with a discussion of related work
and reflections on how the generated \tsql code could be optimised.


\section{An introduction to TSQL2} \label{TSQL2_intro}

This section introduces \tsql and the version of the relational model
on which \tsql is based. Some definitions that are not part of the
\tsql documentation are also given; these will be used in following
sections. I note that although \cite{TSQL2book} defines \tsql's
syntax rigorously, the semantics of the language is defined very
informally, with parts of the semantics left to the intuition of the
reader. There are also some inconsistencies in the \tsql definition
(several of these were pointed out in \cite{Androutsopoulos1995b}).

\subsection{The traditional relational model} \label{relational}

As explained in section \ref{tdbs_general}, the traditional relational
model stores information in relations, which can be thought of as
tables. For example, $salaries$ below is a relation showing the
current salaries of a company's employees. $salaries$ has two
\emph{attributes} (intuitively, columns), $employee$ and $salary$. The
\emph{tuples of the relation} are intuitively the rows of the table
($salaries$ has three tuples). 
\adbtable{2}{|l|l|}{$salaries$}
{$employee$ & $salary$ }
{$J.Adams$ & $17000$ \\
 $T.Smith$ & $19000$ \\
 $G.Papas$ & $14500$
}
I adopt a set-theoretic definition of relations (see section 2.3 of
\cite{Ullman} for alternative approaches).  A set of attributes 
${\cal D}_A$ 
\index{da@${\cal D}_A$ (set of all attributes)}
is assumed (e.g.\ $employee$ and $salary$ are elements of ${\cal
D}_A$). A \emph{relation schema} is an ordered tuple of one or more
attributes (e.g.\ $\tup{employee, salary}$). A set of \emph{domains}
${\cal D}_D = \{D_1, D_2, \dots, D_{n_D}\}$ 
\index{dd@${\cal D}_D$ (set of all domains)}
is also assumed. Each element $D_i$ of ${\cal D}_D$ is itself a
set. For example, $D_1$ may contain all strings, $D_2$ all positive
integers, etc. Each attribute (element of ${\cal D}_A$) is assigned a
domain (element of ${\cal D}_D$). $D(A)$ 
\index{d()@$D(A)$ (domain of the attribute $A$)}
denotes the domain of attribute $A$. $D$ 
\index{d@$D$ (universal domain)}
on its own refers to the \emph{universal domain}, the union of all $D_i
\in {\cal D}_D$. 

A \emph{relation} over a relation schema $R = \tup{A_1, A_2, \dots,
A_n}$ is a subset of $D(A_1) \times D(A_2) \times \dots \times
D(A_n)$, where $\times$ denotes the cartesian product, and $D(A_1)$,
$D(A_2)$,~\dots, $D(A_n)$ are the domains of the attributes $A_1$,
$A_2$,~\dots, $A_n$ respectively. That is, a relation over $R$ is a
set of tuples of the form $\tup{v_1, v_2, \dots, v_n}$, where $v_1 \in
D(A_1)$, $v_2 \in D(A_2)$, \dots, $v_n \in D(A_n)$. In each tuple
$\tup{v_1, v_2, \dots, v_n}$, $v_1$ is the \emph{attribute value} of
$A_1$, $v_2$ is the attribute value of $A_2$, etc. The universal
domain $D$ is the set of all possible attribute values.  Assuming, for
example, that $employee, salary \in {\cal D}_A$, that $D_1$ and $D_2$
are as in the previous paragraph, and that $employee$ and $salary$ are
assigned $D_1$ and $D_2$, $r$ below is a relation over $\tup{employee,
salary}$. ($r$ is a mathematical representation of $salaries$ above.)
On its own, ``relation'' will be used to refer to a relation over any
relation schema.
\[
r = \{\tup{J.Adams, 17000}, \tup{T.Smith, 19000}, \tup{G.Papas, 14500}\} 
\]
The \emph{arity} of a relation over $R$ is the number of attributes in
$R$ (e.g.\ the arity of $r$ is 2). The \emph{cardinality} of a relation is
the number of tuples it contains (the cardinality of $r$ is
3). A relational \emph{database} is a set of relations (more elaborate
definitions are possible, but this is sufficient for our purposes).

I assume that every element of $D$ (universal domain) denotes an
object in the modelled world. (``Object in the world'' is used here
with a very loose meaning, that covers qualifications of employees,
salaries, etc.) \objsdb 
\index{objsdb@\objsdb (\bcdm's world objects)}
is the set of all the world objects that are each denoted by a single
element of $D$. (Some world objects may be represented in the database
as collections of elements of $D$, e.g.\ as whole tuples. \objsdb
contains only world objects that are denoted by \emph{single} elements
of $D$.) I also assume that a function $f_D : D \mapsto \objsdb$
\index{fd@$f_D()$ (maps attribute values to world objects)} 
is available, that maps each element $v$ of $D$ to the world object
denoted by $v$. $f_D$ reflects the semantics assigned to the attribute
values by the people who use the database. In practice, an element of
$D$ may denote different world objects when used as the value of
different attributes. For example, $15700$ may denote a salary when
used as the value of $salary$, and a part of an engine when used as
the value of an attribute $part\_no$. Hence, the value of $f_D$ should
also depend on the attribute where the element of $D$ is used, i.e.\
it should be a function $f_D : D \times {\cal D}_A \mapsto
\objsdb$. For simplicity, I overlook this detail.

I also assume that $f_D$ is 1-1 (injective), i.e.\ that every element
of $D$ denotes a different world object. In practice, $f_D$ may not be
1-1: the database may use two different attribute values (e.g.\ $dpt3$
and $sales\_dpt$) to refer to the same world object. The \topl to
\tsql translation could be formulated without assuming that $f_D$ is
1-1. This assumption, however, bypasses uninteresting details. By the
definition of \objsdb, any element of \objsdb is a world object
denoted by some element of $D$. That is, for every $o \in \objsdb$,
there is a $v \in D$, such that $f_D(v) = o$, i.e.\ $f_D$ is also
surjective. Since $f_D$ is both 1-1 and surjective, the inverse
mapping \fdi is a function, and \fdi is also 1-1 and surjective.
 
\subsection{TSQL2's model of time} \label{tsql2_time}

Like \topl, \tsql assumes that time is discrete, linear, and bounded.
In effect, \tsql models time as consisting of
\emph{chronons}. Chronons are the shortest representable units of
time, and correspond to \topl's time-points.\footnote{\tsql
distinguishes between \emph{valid-time chronons},
\emph{transaction-time chronons}, and \emph{bitemporal chronons}
(pairs each comprising a valid-time and a transaction-time chronon;
see chapter 10 of \cite{TSQL2book}). As noted in section
\ref{tdbs_general}, transaction-time is ignored in this thesis. Hence,
transaction-time and bitemporal chronons are not used, and ``chronon''
refers to valid-time chronons.} Depending on the \tsql implementation,
a chronon may represent a nanosecond, a day, or a whole century. Let
us call the (implementation-specific) set of chronons \chrons. 
\index{chrons@\chrons (set of all chronons)}
Although not stated explicitly, it is clear from the discussion in
chapter 6 of \cite{TSQL2book} that $\chrons \not= \emptyset$, that
chronons are ordered by a binary precedence relation (let us call it
$\prec^{db}$), and that $\tup{\chrons, \prec^{db}}$ has the properties
of transitivity, irreflexivity, linearity, left and right boundedness,
and discreteness (section \ref{temporal_ontology}).

I define periods over $\tup{\chrons, \prec^{db}}$ in the same way as
periods over $\tup{\pts, \prec}$ (section \ref{temporal_ontology}). A
period over $\tup{\chrons, \prec^{db}}$ is a non-empty and convex set
of chronons. An instantaneous period over $\tup{\chrons, \prec^{db}}$
is a set that contains a single chronon. 
$\periods_{\tup{\chrons, \prec^{db}}}$ 
\index{periods2@$\periods_{\tup{\chrons, \prec^{db}}}$ (set of all
periods over $\tup{\chrons, \prec^{db}}$}
and $\instants_{\tup{\chrons, \prec^{db}}}$ 
\index{instants2@$\instants_{\tup{\chrons, \prec^{db}}}$ (set of all instantaneous periods over $\tup{\chrons, \prec^{db}}$)}
are the sets of all periods and all instantaneous periods over
$\tup{\chrons, \prec^{db}}$ respectively. In section
\ref{resulting_model}, I set the point structure $\tup{\pts, \prec}$
of \topl's model to $\tup{\chrons, \prec^{db}}$. Hence,
$\periods_{\tup{\pts, \prec}}$ and $\instants_{\tup{\pts, \prec}}$
become $\periods_{\tup{\chrons, \prec^{db}}}$ and
$\instants_{\tup{\chrons, \prec^{db}}}$. As in 
chapter \ref{TOP_chapter}, I write \periods 
\index{periods@$\periods$ (set of all periods)} 
and \instants 
\index{instants@$\instants$ (set of all instantaneous periods)}
to refer to these sets, and $\periods^*$ 
\index{periods*@$\periods^*$ ($\periods \union \emptyset$)}
to refer to $\periods \union \{\emptyset\}$.  

A \emph{temporal element} over $\tup{\chrons, \prec^{db}}$ is a
non-empty (but not necessarily convex) set of
chronons. $\telems_{\tup{\chrons, \prec^{db}}}$ 
\index{telems2@$\telems_{\tup{\chrons, \prec^{db}}}$ (set of all temporal elements over $\tup{\chrons, \prec^{db}}$)}
(or simply \telems)
\index{telems@\telems (set of all temporal elements)}
is the set of all temporal elements over $\tup{\chrons,
\prec^{db}}$. Obviously, $\periods \subseteq \telems$. For every $l
\in \telems$, $mxlpers(l)$ 
\index{mxlpers@$mxlpers()$ (maximal periods of a set or temporal element)}
is the set of the \emph{maximal periods} of $l$, defined as follows: 
\begin{align*}
mxlpers(l) \defeq  \{p \subseteq l \mid & \; p \in \periods \text{ and for no }
                      p' \in \periods \text{ is it true that } \\
                    & \; p' \subseteq l \text{ and } p \propsubper p'\}
\end{align*}
The $mxlpers$ symbol is overloaded. When $l \in \telems$, $mxlpers(l)$
is defined as above. When $S$ is a set of periods, $mxlpers(S)$ is defined
as in section \ref{temporal_ontology}. 

\tsql supports multiple \emph{granularities}. These correspond to
\topl complete partitionings. A granularity can be
thought of as a set of periods over $\tup{\chrons, \prec^{db}}$
(called \emph{granules}), such that no two periods overlap, and the
union of all the periods is \chrons. A lattice is used to capture
relations between granularities (e.g.\ a year-granule contains twelve 
month-granules, etc; see chapter 19 of \cite{TSQL2book}). \instants,
also called the \emph{granularity of chronons}, is the finest
available granularity. 

\tsql allows periods and temporal elements to be specified at any
granularity. For example, one may specify that the first day of a
period is 25/11/95, and the last day is 28/11/95. If the granularity
of chronons is finer than the granularity of days, the exact chronons
within 25/11/95 and 28/11/95 where the period starts and ends are
unknown. Similarly, if a temporal element is specified at a
granularity coarser than \instants, the exact chronon-boundaries of
its maximal periods are unknown.\footnote{To bypass this problem, in
\cite{Androutsopoulos1995b} periods and temporal elements are defined
as sets of granules (of any granularity) rather than sets of
chronons.} These are examples of \emph{indeterminate temporal
information} (see chapter 18 of \cite{TSQL2book}). Information of this
kind is ignored in this thesis. I assume that all periods and temporal
elements are specified at the granularity of chronons, and that we
know exactly which chronons are or are not included in periods and
temporal elements. Granularities other than \instants will be used 
only to express durations (see below). 

Finally, \tsql uses the term \emph{interval} to refer to a
duration (see comments in section \ref{top_intro}). An interval is a
number of consecutive granules of some particular granularity (e.g.\
two day-granules, five minute-granules). 

\subsection{BCDM} \label{bcdm}
 
As noted in section \ref{tdbs_general}, numerous temporal versions of
the relational model have been proposed. \tsql is based on a version
called \bcdm.  Apart from the relations of the traditional relational
model (section \ref{relational}), which are called \emph{snapshot
relations} in \tsql, \bcdm provides \emph{valid-time relations},
\emph{transaction-time relations}, and \emph{bitemporal
relations}. Transaction-time and bitemporal relations are not used in
this thesis (see chapter 10 of \cite{TSQL2book}). Valid-time relations
are similar to snapshot relations, except that they have a special
extra attribute (the \emph{implicit attribute}) that shows when the
information of each tuple was/is/will be true.

A special domain $D_T \in {\cal D}_D$
\index{dt@$D_T$ (set of all attribute values that denote temporal elements)}
is assumed, whose elements denote the elements of \telems (temporal
elements). For every $v_t \in D_T$, $f_D(v_t) \in
\telems$; and for every $l \in \telems$, $\fdi(l) \in D_T$. $D_T$ is
the domain of the implicit attribute. Since $D_T \in {\cal
D}_D$, $D_T \subseteq D$ ($D$ is the union of all
the domains in ${\cal D}_D$). The assumptions of section
\ref{relational} about $f_D$ still hold: I assume that $f_D$
is an injective and surjective function from $D$ (which now includes
$D_T$) to \objsdb. Since the elements of $D_T$ denote all the elements
of \telems, $D_T \subseteq D$, and \objsdb contains all the objects
denoted by elements of $D$, it must be the case that $\telems
\subseteq \objsdb$. Then, the fact that $\periods \subseteq \telems$
(section \ref{tsql2_time}) implies that $\periods \subseteq
\objsdb$.

A \emph{valid-time relation} $r$ over a relation-schema $R = \tup{A_1,
A_2, \dots, A_n}$ is a subset of $D(A_1) \times D(A_2) \times \dots
\times D(A_n) \times D_T$, where $D(A_1)$, $D(A_2)$,~\dots,
$D(A_n)$ are the domains of $A_1$, $A_2$,~\dots, $A_n$. $A_1$,
$A_2$,~\dots, $A_n$ are the \emph{explicit attributes} of $r$.  I use
the notation $\tup{v_1, v_2, \dots, v_n; v_t}$ to refer to tuples of
valid-time relations. If $r$ is as above and $\tup{v_1, v_2, \dots,
v_n; v_t} \in r$, then $v_1 \in D(A_1)$, $v_2 \in D(A_2)$,~\dots,
$v_n \in D(A_n)$, and $v_t \in D_T$. $v_1$, $v_2$,~\dots, $v_n$ are
the \emph{values of the explicit attributes}, while $v_t$ is the
\emph{value of the implicit attribute} and the \emph{time-stamp} of
the tuple. In snapshot relations, all attributes count as explicit.
In the rest of this thesis, ``valid-time relation'' on its own 
refers to a valid-time relation over any relation-schema. 

\tsql actually distinguishes between \emph{state valid-time relations}
and \emph{event valid-time relations} (see chapter 16 of
\cite{TSQL2book}). These are intended to model situations that have
duration or are instantaneous respectively. This distinction seems
particularly interesting, because it appears to capture some facets of
the aspectual taxonomy of chapter
\ref{linguistic_data}. Unfortunately, it is also one of the most
unclear and problematically defined features of \tsql. The time-stamps
of state and event valid-time relations are supposed to denote
``temporal elements'' and ``instant sets'' respectively. ``Temporal
elements'' are said to be unions of periods, while ``instant sets''
simply sets of chronons (see p.314 of \cite{TSQL2book}). This
distinction between ``temporal elements'' and ``instant sets'' is
problematic. A union of periods is a union of convex sets of chronons,
i.e.\ simply a set of chronons. (The union of two convex sets of
chronons is not necessarily convex.) Hence, one cannot distinguish
between unions of periods and sets of chronons (see also section 2 of
\cite{Androutsopoulos1995b}). In section 3.3 of
\cite{Androutsopoulos1995b} we also argue that \tsql does not allow
specifying whether a computed valid-time relation should be state or
event. Given these problems, I chose to drop the distinction between
state and event valid-time relations. I assume that the time-stamps of
all valid-time relations denote temporal elements, with temporal
elements being sets of chronons.

For example, assuming that the domains of $employee$ and $salary$ are
as in section \ref{relational}, $val\_salaries$ below is a valid-time
relation over $\tup{employee, salary}$, shown in its tabular form (the
double vertical line separates the explicit attributes from the
implicit one). According to chapter 10 of \cite{TSQL2book}, the
elements of $D_T$ are non-atomic. Each element $v_t$ of $D_T$ is in
turn a set, whose elements denote the chronons that belong to the
temporal element represented by $v_t$.
\adbtable{3}{|l|l||l|}{$val\_salaries$}
{$employee$ & $salary$ &}
{$J.Adams$ & $17000$ & $\{c^1_1, c^1_2, c^1_3, \dots, c^1_{n_1}\}$ \\
 $J.Adams$ & $18000$ & $\{c^2_1, c^2_2, c^2_3, \dots, c^2_{n_2}\}$ \\
 $J.Adams$ & $18500$ & $\{c^3_1, c^3_2, c^3_3, \dots, c^3_{n_3}\}$ \\
 $T.Smith$ & $19000$ & $\{c^4_1, c^4_2, c^4_3, \dots, c^4_{n_4}\}$ \\
 $T.Smith$ & $21000$ & $\{c^5_1, c^5_2, c^5_3, \dots, c^5_{n_5}\}$
}
For example, $c^1_1, c^1_2, c^1_3, \dots, c^1_{n_1}$ in the first
tuple for J.Adams above represent all the chronons where the salary of
J.Adams was/is/will be 17000. $\{c^1_1, c^1_2, c^1_3, \dots,
c^1_{n_1}\}$ is an element of $D_T$. For simplicity, when depicting
valid-time relations I often show (in an informal manner) the temporal
elements denoted by the time-stamps rather the time-stamps
themselves. $val\_salaries$ would be shown as below, meaning that the
time-stamp of the first tuple represents a temporal element of two
maximal periods, 1/1/92 to 12/6/92 and 8/5/94 to 30/10/94. (I assume
here that chronons correspond to days. $now$ refers to the current
chronon.)
\begin{examps}
\item \label{tlang:4}
\dbtable{3}{|l|l||l|}{$val\_salaries$}
{$employee$ & $salary$ &}
{$J.Adams$ & $17000$ & $[1/1/92, \; 12/6/92] \union [8/5/94, \; 30/10/94]$ \\
 $J.Adams$ & $18000$ & $[13/6/92, \; 7/5/94] \union [31/10/94, \; now]$ \\
 $T.Smith$ & $21000$ & $[15/6/92, \; now]$ 
}
\end{examps}
Two tuples $\tup{v_1^1, \dots, v_n^1; v_t^1}$ and $\tup{v^2_1, \dots,
v_n^2; v_t^2}$ are \emph{value-equivalent} iff if $v^1_1 = v^2_1$,
\dots, $v^1_n = v^2_n$. A valid-time relation is \emph{coalesced} iff
it contains no value-equivalent tuples. \bcdm requires all valid-time
relations to be coalesced (see p.188 of \cite{TSQL2book}). For
example, \pref{bcdm:1} is not allowed (its first and third tuples are
value-equivalent). In this thesis, this \bcdm restriction is dropped,
and \pref{bcdm:1} is allowed.
\begin{examps}
\item \label{bcdm:1}
\dbtablec{|l|l||l|}
{$employee$ & $salary$ &}
{$J.Adams$ & $17000$ & $[1/1/92, \; 12/6/92]$ \\
 $J.Adams$ & $18000$ & $[13/6/92, \; 7/5/94]$ \\
 $J.Adams$ & $17000$ & $[8/5/94, \; 30/10/94]$ \\
 $J.Adams$ & $18000$ & $[31/10/94, \; now]$ \\
 $T.Smith$ & $21000$ & $[15/6/92, \; now]$ 
}
\end{examps}
By the definition of $D_T$, the elements of $D_T$ denote all the
elements of $\telems$ (temporal elements). Since $\periods \subseteq
\telems$, some of the elements of $D_T$ denote periods. $D_P$
\index{dp@$D_P$ (set of all attribute values that denote periods)}
 is the subset of all elements of $D_T$ that denote
periods.\footnote{\cite{TSQL2book} seems to adopt a different
approach, where $D_P \intersect D_T = \emptyset$.} I also assume that
there is a special value $\vempty \in D$,
\index{ve@$\vempty$ (attribute value denoting $\emptyset$)}
that is used to denote the empty set (of chronons). For example, a
\tsql expression that computes the intersection of two non-overlapping
periods evaluates to \vempty.\footnote{Table 8.3 of \cite{TSQL2book}
implies that \vempty is the special ``null'' value. In \sqll, null has
several roles. Here, I assume that there is a special value \vempty
whose only role is to denote the empty set.} I use $D_P^*$ 
\index{dp*@$D_P^*$ ($D_P \union \emptyset$)}
to refer to $D_P \union \{\vempty\}$.

The following notation will prove useful:
\begin{itemize}

\item \vrel 
\index{vrelp@\vrel (set of all valid time relations time-stamped by elements of $D_P$)} 
is the set of all valid-time relations whose time-stamps
are all elements of $D_P$ (all the time-stamps denote periods). 

\item \cvrel 
\index{nvrelp@\cvrel (``normalised'' elements of \vrel)}
is the set of all the (intuitively, ``normalised'') relations $r \in
\vrel$ with the following property: if $\tup{v_1, \dots, v_n; v^1_t}
\in r$, $\tup{v_1, \dots, v_n; v^2_t} \in r$, and $f_D(v^1_t) \union
f_D(v^2_t) \in \periods$, then $v^1_t = v^2_t$. This definition
ensures that in any $r \in \cvrel$, there is no pair of different
value-equivalent tuples whose time-stamps $v^1_t$ and $v^2_t$ denote 
overlapping or adjacent periods (because if the periods of $v^1_t$ and
$v^2_t$ overlap or they are adjacent, their union is also a period,
and then it must be true that $v^1_t = v^2_t$, i.e.\ the
value-equivalent tuples are not different). 

\item \srel 
\index{srel@\srel (set of all snapshot relations)}
is the set of all snapshot relations. 

\item For every $n \in \{1,2,3,\dots\}$, $\vrel(n)$ 
\index{vrelpn@$\vrel(n)$ (relations in \vrel with $n$ explicit attributes)} 
contains all the relations of \vrel that have $n$ explicit
attributes. Similarly, $\cvrel(n)$ 
\index{nvrelpn@$\cvrel(n)$ (set of all relations in \vrel with $n$ explicit attributes)}
and $\srel(n)$ 
\index{sreln@$\srel(n)$ (set of all snapshot relations of $n$ attributes)}
contain all the relations of \cvrel and \srel respectively that have
$n$ explicit attributes. 

\end{itemize}

To simplify the proofs in the rest of this chapter, I include the
empty relation in all $\vrel(n)$, $\cvrel(n)$, $\srel(n)$, for $n=
1,2,3,\dots$. 

\subsection{The TSQL2 language} \label{tsql2_lang}

This section is an introduction to the features of \tsql that
are used in this thesis. 

\subsubsection*{SELECT statements}

As noted in section \ref{tdbs_general}, \tsql is an extension of
\sqlnt. Roughly speaking, \sqlnt queries (e.g.\ \ref{tlang:1}) consist
of three clauses: a \sql{SELECT},
\index{select@\sql{SELECT} (\tsql keyword, introduces a \tsql query)}
a \sql{FROM}, 
\index{from@\sql{FROM} (\tsql keyword, shows the relations on which a \sql{SELECT} operates)}
and a \sql{WHERE} 
\index{where@\sql{WHERE} (\tsql keyword, introduces restrictions)}
clause. (The term \emph{\sql{SELECT} statement} will
be used to refer to the whole of a \sqlnt or \tsql query.)
\begin{examps}
\item 
\index{as@\sql{AS} (\tsql keyword, introduces correlation names)}
\index{and@\sql{AND} (\tsql's conjunction)}
\label{tlang:1}
\select{SELECT DISTINCT sal.salary \\
        FROM salaries AS sal, managers AS mgr \\
        WHERE mgr.manager = 'J.Adams' AND sal.employee = mgr.managed}
\end{examps}
Assuming that $salaries$ and $managers$ are as below, \pref{tlang:1}
generates the third relation below.
\begin{examps}
\item[]
\dbtable{2}{|l|l|}{$salaries$}
{$employee$ & $salary$ }
{$J.Adams$  & $17000$ \\
 $T.Smith$  & $18000$ \\
 $G.Papas$  & $14500$ \\
 $B.Hunter$ & $17000$ \\
 $K.Kofen$  & $16000$
}
\ \ 
\dbtable{2}{|l|l|}{$managers$}
{$manager$ & $managed$ }
{$J.Adams$ & $G.Papas$ \\
 $J.Adams$ & $B.Hunter$ \\
 $J.Adams$ & $J.Adams$ \\
 $T.Smith$ & $K.Kofen$ \\
 $T.Smith$ & $T.Smith$
}
\ \ 
\dbtable{1}{|l|}{$(result)$}
{$salary$}
{$17000$ \\
 $14500$
}
\end{examps}
\pref{tlang:1} generates a snapshot one-attribute relation that
contains the salaries of all employees managed by J.Adams. The
\sql{FROM} clause of \pref{tlang:1} shows that the query operates on
the $salaries$ and $managers$ relations. \sql{sal} and \sql{mgr} are 
\emph{correlation names}.
They can be thought of as tuple-variables ranging over the tuples of
$salaries$ and $managers$ respectively. The (optional) \sql{WHERE}
clause imposes restrictions on the possible combinations of
tuple-values of \sql{sal} and \sql{mgr}. In every combination, the
$manager$ value of \sql{mgr} must be $J.Adams$, and the $managed$
value of \sql{mgr} must be the same as the $employee$ value of
\sql{sal}. For example, $\tup{J.Adams, G.Papas}$ and $\tup{G.Papas,
14500}$ is an acceptable combination of
\sql{mgr} and \sql{sal} values respectively, while $\tup{J.Adams,
G.Papas}$ and $\tup{B.Hunter, 17000}$ is not.

In \sqlnt (and \tsql), correlation names are optional, and relation
names can be used to refer to attribute values. In \pref{tlang:1}, for
example, one could omit \sql{AS mgr}, and replace \sql{mgr.manager} and
\sql{mgr.managed} by \sql{managers.manager} and
\sql{managers.managed}. To simplify the definitions of section
\ref{additional_tsql2} below, I treat correlation names as 
mandatory, and I do not allow relation names to be used to refer to
attribute values.

The \sql{SELECT} clause specifies the contents of the resulting
relation. In \pref{tlang:1}, it specifies that the resulting relation
should have only one attribute, $salary$, and that for each acceptable
combination of \sql{sal} and \sql{mgr} values, the corresponding tuple
of the resulting relation should contain the $salary$ value of
\sql{sal}'s tuple. The \sql{DISTINCT} 
\index{distinct@\sql{DISTINCT} (\tsql keyword, removes duplicate tuples)}
in \pref{tlang:1} causes duplicates of tuples to be removed from the
resulting relation. Without the \sql{DISTINCT} duplicates are not
removed. The result of \pref{tlang:1} would contain two identical
tuples $\tup{17000}$, deriving from the tuples for J.Adams and
B.Hunter in $salaries$. This is against the set-theoretic definition
of relations of sections \ref{relational} and \ref{bcdm} (relations were
defined to be \emph{sets} of tuples, and hence cannot contain
duplicates.)  To ensure that relations contain no duplicates, in this
thesis \sql{SELECT} statements always have a
\sql{DISTINCT} in their \sql{SELECT} clauses.

\tsql allows \sql{SELECT} statements to operate 
on valid-time relations as well. A \sql{SNAPSHOT} 
\index{snapshot@\sql{SNAPSHOT} (\tsql keyword, signals that a snapshot relation is to be created)}
keyword in the \sql{SELECT} statement indicates that the resulting
relation is snapshot. When the resulting relation is valid-time, an
additional \sql{VALID} clause is present. In the latter case, the
\sql{SELECT} clause specifies the values of the explicit attributes of
the resulting relation, while the
\sql{VALID} clause specifies the time-stamps of the resulting tuples.
Assuming, for example, that $val\_salaries$ is as in 
\pref{tlang:4}, \pref{tlang:7} returns \pref{tlang:8}.
\begin{examps}
\item \label{tlang:7}
\sql{SELECT DISTINCT sal.employee, sal.salary \\
     VALID PERIOD(BEGIN(VALID(sal)), END(VALID(sal))) \\
     FROM val\_salaries AS sal}
\item \label{tlang:8}
\dbtablec{|l|l||l|}
{$employee$ & $salary$ &}
{$J.Adams$ & $17000$ & $[1/1/92, \; 30/10/94]$ \\
 $J.Adams$ & $18000$ & $[13/6/92, \; now]$ \\
 $T.Smith$ & $21000$ & $[15/6/92, \; now]$ 
}
\end{examps}
The \sql{VALID} 
\index{valid@\sql{VALID} (\tsql keyword, refers to time-stamps of tuples)}
keyword is used both to start a \sql{VALID}-clause (a clause that
specifies the time-stamps of the resulting relation) and to refer to
the time-stamp of the tuple-value of a correlation name. In
\pref{tlang:7}, \sql{VALID(sal)} refers to the time-stamp of
\sql{sal}'s tuple (i.e.\ to the time-stamp of a tuple from
$val\_salaries$). \sql{BEGIN(VALID(sal))} 
\index{begin2@\sql{BEGIN} (\tsql keyword, returns the start-point of a temporal element)}
refers to the first chronon of the temporal element represented by
that time-stamp, and \sql{END(VALID(sal))} 
\index{end2@\sql{END} (\tsql keyword, returns the end-point of a temporal element)}
to the last chronon of that temporal
element.\footnote{Section 30.5 of \cite{TSQL2book} allows 
\sql{BEGIN} and \sql{END} to be used only with periods. I see no
reason for this limitation. I allow \sql{BEGIN} and \sql{END} to be
used with any temporal element.}  The 
\sql{PERIOD}
\index{period@\sql{PERIOD} (\tsql keyword, constructs periods or introduces period literals)}
function generates a period that starts at the
chronon of its argument, and ends at the chronon of its second
argument.  Hence, each time-stamp of \pref{tlang:8} represents a
period that starts/ends at the earliest/latest chronon of the temporal
element of the corresponding time-stamp of $val\_salaries$.

\subsubsection*{Literals}

\tsql provides \sql{PERIOD} 
\index{period@\sql{PERIOD} (\tsql keyword, constructs periods or introduces period literals)}
literals, \sql{INTERVAL} 
\index{interval@\sql{INTERVAL} (\tsql keyword, returns intervals or introduces interval literals)}
literals, and \sql{TIMESTAMP} 
\index{timestamp@\sql{TIMESTAMP} (\tsql keyword, introduces chronon-denoting literals)}
literals (the use of ``\sql{TIMESTAMP}'' in this case is unfortunate;
these literals specify time-points, not time-stamps of valid-time
relations, which denote temporal-elements). For example,
\sql{PERIOD '[March 3, 1995 - March 20, 1995]'} is a literal that
specifies a period at the granularity of days. If 
chronons are finer than days, the assumption in
\tsql is that the exact chronons within March 3 and March 20 where the period
starts and ends are unknown (section \ref{tsql2_time}). In this
thesis, \sql{PERIOD} literals that refer to granularities other than
that of chronons are abbreviations for literals that refer to the
granularity of chronons. The denoted period contains all the chronons
that fall within the granules specified by the literal. For example,
if chronons correspond to minutes,
\sql{PERIOD '[March 3, 1995 - March 20, 1995]'} is an
abbreviation for \sql{PERIOD '[00:00 March 3, 1995 - 23:59 March 20,
1995]'}.

\tsql supports multiple calendars (e.g.\ Gregorian, Julian, lunar
calendar; see chapter 7 of \cite{TSQL2book}). The strings that
can appear between the quotes of \sql{PERIOD} literals (e.g.\
\sql{'[March 3, 1995 - March 20, 1995]'}, \sql{'(3/4/95 - 20/4/95]'})
depend on the available calendars and the selected formatting
options (see chapter 7 of \cite{TSQL2book}). The convention seems to
be that the boundaries are separated by a dash, and that the first and last
characters of the quoted string are square or round brackets,
depending on whether the boundaries are to be included or not. 
I also assume that \sql{PERIOD 'today'} can be used (provided
that chronons are at least as fine as days) to refer to the
period that covers all the chronons of the present
day. (There are other \tsql expressions that can be used to refer to
the current day, but I would have to discuss \tsql
granularity-conversion commands to explain these.
Assuming that \sql{PERIOD 'today'} is available allows
me to avoid these commands.)

\sql{TIMESTAMP} literals specify chronons. Only the following special
\sql{TIMESTAMP} literals are used in this thesis: \sql{TIMESTAMP
'beginning'}, \sql{TIMESTAMP 'forever'}, \sql{TIMESTAMP 'now'}. These
refer to the beginning of time, the end of time, and the present chronon.

An example of an \sql{INTERVAL} literal is \sql{INTERVAL '5' DAY},
which specifies a duration of five consecutive day-granules.  The
available granularities depend on the calendars that are active. The
granularities of years, months, days, hours, minutes, and seconds are
supported by default.  Intervals can also be used to shift periods or
chronons towards the past or the future. For example, \sql{PERIOD
'[1991 - 1995]' + INTERVAL '1' YEAR} is the same as \sql{PERIOD '[1992
- 1996]'}. If chronons correspond to minutes, \sql{PERIOD(TIMESTAMP
'beginning', TIMESTAMP 'now' - INTERVAL '1' MINUTE)} specifies the
period that covers all the chronons from the beginning of time up to
(but not including) the current chronon.

\subsubsection*{Other TSQL2 functions and predicates}

The \sql{INTERSECT}
\index{intersect@\sql{INTERSECT} (\tsql keyword, computes the intersection of two sets of chronons)} 
function computes the intersection of two sets of
chronons.\footnote{Section 8.3.3 of \cite{TSQL2book} requires both
arguments of \sql{INTERSECT} to denote periods, but section 30.14
allows the arguments of \sql{INTERSECT} to denote temporal elements. I
follow the latter. I also allow the arguments of \sql{INTERSECT} to
denote the empty set.} For example, \sql{INTERSECT(PERIOD '[May 1,
1995 - May 10, 1995]', PERIOD '[May 3, 1995 - May 15, 1995]')} is the
same as \sql{PERIOD '[May 3, 1995 - May 10, 1995]'}. If the intersection is
the empty set, \sql{INTERSECT} returns \vempty (section
\ref{bcdm}).

The \sql{CONTAINS} 
\index{contains@\sql{CONTAINS} (\tsql keyword, checks if a chronon belongs to a set of chronons)}
predicate checks if a chronon belongs to a 
set of chronons. For example, if $val\_salaries$ is as in
\pref{tlang:4}, \pref{tlang:9} generates a snapshot relation showing
the current salary of each employee. \sql{CONTAINS} can also be used
to check if a set of chronons is a subset of another set of
chronons.\footnote{Table 8.7 in section 8.3.6 and additional syntax
rule 3 in section 32.4 of \cite{TSQL2book} allow the arguments of
\sql{CONTAINS} to denote periods but not generally temporal
elements. Table 32.1 in section 32.4 of \cite{TSQL2book}, however,
allows the arguments of \sql{CONTAINS} to denote temporal elements. I
follow the latter. I also allow the arguments of \sql{CONTAINS} to
denote the empty set. The same comments apply in the case of
\sql{PRECEDES}.}
\begin{examps}
\item \label{tlang:9}
\select{SELECT DISTINCT SNAPSHOT sal.employee, sal.salary \\
        FROM val\_salaries AS sal \\
        WHERE VALID(sal) CONTAINS TIMESTAMP 'now'}
\end{examps}
The \sql{PRECEDES} 
\index{precedes@\sql{PRECEDES} (\tsql keyword, checks temporal precedence)}
predicate checks if a chronon or set of chronons strictly precedes
another chronon or set of chronons.  Section 8.3.6 of
\cite{TSQL2book} specifies the semantics of \sql{PRECEDES} only in
cases where its arguments are chronons or periods. I assume that
$expr_1$ \sql{PRECEDES} $expr_2$ is true, iff the chronon of $expr_1$
(if $expr_1$ specifies a single chronon) or all the chronons of
$expr_1$ (if $expr_1$ specifies a set of chronons) strictly precede
the chronon of $expr_2$ (if $expr_2$ specifies a single chronon) or
all the chronons of $expr_2$ (if $expr_2$ specifies a set of
chronons). For example,
\sql{PERIOD '[1/6/95 - 21/6/95]' PRECEDES PERIOD '[24/6/95 -
30/6/95]'} is true, but \sql{PERIOD '[1/6/95 - 21/6/95]' PRECEDES
PERIOD '[19/6/95 - 30/6/95]'} is not.

\subsubsection*{Embedded SELECT statements}

\tsql (and \sqlnt) allow embedded \sql{SELECT} statements to be used
in the \sql{FROM} clause, in the same way that relation names are
used (e.g.\ \pref{tlang:10}). 
\begin{examps}
\item \label{tlang:10}
\select{SELECT DISTINCT SNAPSHOT sal2.salary \\
        FROM (\select{SELECT DISTINCT sal1.salary \\
                      VALID VALID(sal1) \\
                      FROM val\_salaries AS sal1} \\
        \ \ \ \ \ ) AS sal2 \\
        WHERE sal2.salary > 17500}
\end{examps}
Assuming that $val\_salaries$ is as in \pref{tlang:4}, the embedded
\sql{SELECT} statement above simply drops the $employee$ attribute of 
$val\_salaries$, generating \pref{tlang:11}. \sql{sal2} ranges over
the tuples of \pref{tlang:11}. \pref{tlang:10} generates a relation
that is the same as \pref{tlang:11}, except that tuples whose $salary$
values are not greater than 17500 are dropped.
\begin{examps}
\item \label{tlang:11}
\dbtablec{|l||l|}
{$salary$ &}
{$17000$ & $[1/1/92, \; 12/6/92] \union [8/5/94, \; 30/10/94]$ \\
 $18000$ & $[13/6/92, \; 7/5/94] \union [31/10/94, \; now]$ \\
 $21000$ & $[15/6/92, \; now]$ 
}
\end{examps}

\subsubsection*{Partitioning units} 

In \tsql, relation names and embedded \sql{SELECT} statements in the
\sql{FROM} clause can be followed by \emph{partitioning
units}.\footnote{Section 30.3 of \cite{TSQL2book} allows relation
names but not embedded \sql{SELECT} statements to be followed by partitioning
units in \sql{FROM} clauses. \cite{Snodgrass1994d} (queries
Q.1.2.2, Q.1.2.5, Q.1.7.6), however, shows \sql{SELECT} statements
embedded in \sql{FROM} clauses and followed by partitioning units. I
follow \cite{Snodgrass1994d}.} 
\tsql currently provides two partitioning units: \sql{(PERIOD)} and
\sql{(INSTANT)}
\index{instant@\sql{(INSTANT)} (\tsql partitioning unit)}
(see section 30.3 and chapter 12 of \cite{TSQL2book}). \sql{(INSTANT)}
is not used in this thesis.  Previous \tsql 
versions (e.g.\ the September 1994 version
of chapter 12 of \cite{TSQL2book}) provided an additional
\sql{(ELEMENT)}. For reasons explained below, \sql{(ELEMENT)} is still
used in this thesis.

\sql{(ELEMENT)}
\index{element@\sql{(ELEMENT)} (\tsql partitioning unit)}
merges value-equivalent tuples.\footnote{The semantics of
\sql{(ELEMENT)} was never clear. The discussion here reflects my
understanding of the September 1994 \tsql documentation, and the
semantics that is assigned to \sql{(ELEMENT)} in this thesis.} For
example, if $rel1$ is the relation of \pref{pus:1}, \pref{pus:2}
generates the coalesced relation of
\pref{pus:3}. 
\begin{examps}
\item \label{pus:1}
\dbtable{3}{|l|l||l|}{$rel1$}
{$employee$ & $salary$ &}
{$J.Adams$ & $17000$ & $[1986, \; 1988]$ \\
 $J.Adams$ & $17000$ & $[1987, \; 1990]$ \\
 $J.Adams$ & $17000$ & $[1992, \; 1994]$ \\
 $G.Papas$ & $14500$ & $[1988, \; 1990]$ \\
 $G.Papas$ & $14500$ & $[1990, \; 1992]$ 
}
\item \label{pus:2}
\select{SELECT DISTINCT r1.employee, r1.salary \\
        VALID VALID(r1) \\
        FROM rel1(ELEMENT) AS r1}
\item \label{pus:3}
\dbtablec{|l|l||l|}
{$employee$ & $salary$ &}
{$J.Adams$ & $17000$ & $[1986, \; 1990] \union [1992, \; 1994]$ \\
 $G.Papas$ & $14500$ & $[1988, \; 1992]$
}
\end{examps}
The effect of \sql{(ELEMENT)} on a valid-time relation $r$ is captured
by the $coalesce$ function:
\index{coalesce@$coalesce()$ (effect of \sql{(ELEMENT)})}
\[
\begin{aligned}
coalesce(r) \defeq
\{&\tup{v_1, \dots, v_n; v_t} \mid 
   \tup{v_1, \dots, v_n; v_t'} \in r \text{ and} \\
  &f_D(v_t) = \bigcup_{\tup{v_1, \dots, v_n; v_t''} \in r}f_D(v_t'') \} \\
\end{aligned}
\]
\sql{(ELEMENT)} has no effect on already coalesced valid-time relations.
Hence, in the \bcdm version of
\cite{TSQL2book}, where all valid-time relations are 
coalesced, \sql{(ELEMENT)} is redundant (and this is probably why it
was dropped). In this thesis, valid-time relations are not necessarily
coalesced (section \ref{bcdm}), and \sql{(ELEMENT)} plays an important role. 

\sql{(PERIOD)} 
\index{period2@\sql{(PERIOD)} (\tsql partitioning unit)}
intuitively breaks each tuple of a valid-time relation into 
value-equivalent tuples, each corresponding to a maximal period of the
temporal element of the original time-stamp. Assuming, for example,
that $rel2$ is the relation of \pref{pus:3}, \pref{pus:4} generates
\pref{pus:5}.
\begin{examps}
\item \label{pus:4}
\select{SELECT DISTINCT r2.employee, r2.salary \\
        VALID VALID(r2) \\
        FROM rel2(PERIOD) AS r2}
\item \label{pus:5}
\dbtablec{|l|l||l|}
{$employee$ & $salary$ &}
{$J.Adams$ & $17000$ & $[1986, \; 1990]$ \\
 $J.Adams$ & $17000$ & $[1992, \; 1994]$ \\
 $G.Papas$ & $14500$ & $[1988, \; 1992]$ 
}
\end{examps}
As the example shows, \sql{(PERIOD)} may generate non-coalesced
relations. This is mysterious in the \bcdm version of
\cite{TSQL2book}, where non-coalesced valid-time
relations are not allowed. The assumption seems to be that
although non-coalesced valid-time relations are not allowed, during
the execution of \sql{SELECT} statements temporary non-coalesced
valid-time relations may be generated. Any resulting
valid-time relations, however, are coalesced automatically at the end of
the statement's execution. \pref{pus:5} would be coalesced
automatically at the end of the execution of \pref{pus:4} (cancelling,
in this particular example, the effect of \sql{(PERIOD)}).
In this thesis, no automatic coalescing takes place, and the result of
\pref{pus:4} is \pref{pus:5}.

To preserve the spirit of \sql{(PERIOD)} in the \bcdm version of this
thesis where valid-time relations are not necessarily coalesced, I 
assume that \sql{(PERIOD)} operates on a coalesced copy of the original
relation. Intuitively, \sql{(PERIOD)} first causes \pref{pus:1} to
become \pref{pus:3}, and then generates \pref{pus:5}.
The effect of \sql{(PERIOD)} on a
valid-time relation $r$ is captured by the $pcoalesce$ function:
\index{pcoalesce@$pcoalesce()$ (effect of \sql{(PERIOD)})}
\[
\begin{aligned}
pcoalesce(r) \defeq
\{&\tup{v_1, \dots, v_n; v_t} \mid 
  \tup{v_1, \dots, v_n; v_t'} \in coalesce(r) \text { and} \\
  &f_D(v_t) \in mxlpers(f_D(v_t'))\}
\end{aligned}
\]
 

\section{Modifications of TSQL2} \label{TSQL2_mods}

This thesis adopts some modifications of \tsql. Some of the
modifications were mentioned in section \ref{TSQL2_intro}. The main of
those were: 
\begin{itemize}
\item The requirement that all valid-time relations must be
coalesced was dropped. 
\item The distinction between state and event valid-time relations was
abandoned. 
\item \sql{(ELEMENT)} was re-introduced.
\item The semantics of \sql{(PERIOD)} was
enhanced, to reflect the fact that in this thesis
valid-time relations are not necessarily coalesced. 
\item All periods and temporal
elements are specified at the granularity of chronons. Literals
referring to other granularities are used as abbreviations for
literals that refer to the granularity of chronons.
\end{itemize}
This section describes the remaining \tsql modifications of this
thesis. 

\subsection{Referring to attributes by number} \label{by_num}

In \tsql (and \sqlnt) explicit attributes are referred to by their
names. In \pref{tlang:1b}, for example, \sql{sal.salary} refers to the
$salary$ attribute of $val\_salaries$.
\begin{examps}
\item \label{tlang:1b}
\select{SELECT DISTINCT sal.salary \\
        VALID VALID(sal) \\
        FROM val\_salaries AS sal}
\end{examps}
In the \tsql version of this thesis, explicit attributes are referred
to by number, with numbers corresponding to the order in which the
attributes appear in the relation schema (section
\ref{relational}). For example, if the relation schema of
$val\_salaries$ is $\tup{employee, salary}$, $employee$ is the first
explicit attribute and $salary$ the second one.
\pref{tlang:1c} would be used instead of \pref{tlang:1b}. To refer to
the implicit attribute, one still uses \sql{VALID} (e.g.\ \sql{VALID(sal)}).
\begin{examps}
\item \label{tlang:1c}
\select{SELECT DISTINCT sal.2 \\
        VALID VALID(sal) \\ FROM salaries AS sal}
\end{examps}
Referring to explicit attributes by number simplifies the \topl to
\tsql translation, because this way there is no need to keep track of
the attribute names of the various relations.

\subsection{(SUBPERIOD) and (NOSUBPERIOD)} \label{new_pus}

Two new partitioning units, \sql{(SUBPERIOD)} and \sql{(NOSUBPERIOD)},
were introduced for the purposes of this thesis. \sql{(SUBPERIOD)} is
designed to be used with relations from \vrel (section \ref{bcdm}).
The effect of \sql{(SUBPERIOD)} \index{subperiod2@\sql{(SUBPERIOD)}
  (\tsql partitioning unit)} on a relation $r$ is captured by the
$subperiod$ function: \index{subperiod@$subperiod()$ (effect of
  \sql{(SUBPERIOD)})}
\[
subperiod(r) \defeq
\{ \tup{v_1, \dots, v_n; v_t} \mid
   \tup{v_1, \dots, v_n; v_t'} \in r \text{ and }
   f_D(v_t) \subper f_D(v_t') \}
\]
For each tuple $\tup{v_1, \dots, v_n; v_t'} \in r$, the resulting
relation contains many value-equivalent tuples of the form $\tup{v_1,
\dots, v_n; v_t}$, one for each period $f_D(v_t)$ that is a subperiod of
$f_D(v_t')$. Assuming, for example, that chronons correspond to years, and
that $rel$ is the relation of \pref{subper:0}, \pref{subper:1}
returns the relation of \pref{subper:2}.
\begin{examps}
\item \label{subper:0}
\dbtableb{|l|l||l|}
{$J.Adams$ & $17000$ & $[1992, \; 1993]$ \\
 $G.Papas$ & $14500$ & $[1988, \; 1990]$ \\
 $G.Papas$ & $14500$ & $[1990, \; 1991]$
}
\item \label{subper:1}
\sql{SELECT DISTINCT r.1, r.2 \\
     VALID VALID(r) \\
     FROM rel(SUBPERIOD) AS r}
\item \label{subper:2}
\dbtableb{|l|l||l|}
{$J.Adams$ & $17000$ & $[1992, \; 1993]$ \\
 $J.Adams$ & $17000$ & $[1992, \; 1992]$ \\
 $J.Adams$ & $17000$ & $[1993, \; 1993]$ \\
           &         &                 \\
 $G.Papas$ & $14500$ & $[1988, \; 1990]$ \\
 $G.Papas$ & $14500$ & $[1988, \; 1988]$ \\
 $G.Papas$ & $14500$ & $[1988, \; 1989]$ \\
 $G.Papas$ & $14500$ & $[1989, \; 1989]$ \\
 $G.Papas$ & $14500$ & $[1989, \; 1990]$ \\
 $G.Papas$ & $14500$ & $[1990, \; 1990]$ \\
           &         &                 \\
 $G.Papas$ & $14500$ & $[1990, \; 1991]$ \\
 $G.Papas$ & $14500$ & $[1991, \; 1991]$ 
}
\end{examps}
The first three tuples of \pref{subper:2} correspond to the
first tuple of \pref{subper:0}. The following six tuples correspond
to the first tuple of $G.Papas$ in \pref{subper:0}. The remaining
tuples of \pref{subper:2} derive from the second tuple of $G.Papas$
in \pref{subper:0} (the tuple for the subperiod $[1990, \;
1990]$ has already been included in \pref{subper:2}).
Notice that \sql{(SUBPERIOD)} does not coalesce the original relation
before generating the result (this is why
there is no tuple for G.Papas time-stamped by $[1988, \; 1991]$ in
\pref{subper:2}). 

Obviously, the cardinality of the resulting relations can be very
large (especially if chronons are very fine, e.g.\ seconds).  The
cardinality, however, is never infinite (assuming that the cardinality
of the original relation is finite): given that time is discrete,
linear, and bounded, any period $p$ is a finite set of chronons, and
there is at most a finite number of periods (convex sets of chronons)
that are subperiods (subsets) of $p$\/; hence, for any tuple in the
original relation whose time-stamp represents a period $p$, there will
be at most a finite number of tuples in the resulting relation whose
time-stamps represent subperiods of $p$. It remains, of course, to be
examined if \sql{(SUBPERIOD)} can be supported efficiently in 
\dbms{s}. It is obviously very inefficient to store (or print)
individually all the tuples of the resulting relation. A more
space-efficient encoding of the resulting relation is needed. I have
not explored this issue. 

Roughly speaking, \sql{(SUBPERIOD)} is needed because during the
\topl to \tsql translation every \topl formula
is mapped to a valid-time relation whose time-stamps
denote the event-time periods where the formula is true. Some (but not
all) formulae are homogeneous (section
\ref{denotation}). For these formulae we need to ensure that
if the valid-time relation contains a tuple for 
an event-time $et$, it also contains tuples for 
all the subperiods of $et$. This will become clearer in section
\ref{trans_rules}. 

\sql{(NOSUBPERIOD)}
\index{nosubperiod2@\sql{(NOSUBPERIOD)} (\tsql partitioning unit)}
is roughly speaking used when the effect of
\sql{(SUBPERIOD)} needs to be cancelled.
\sql{(NOSUBPERIOD)} is designed to be used with 
relations from \vrel. It eliminates any tuple $\tup{v_1, \dots, v_n;
v_t}$, for which there is a value-equivalent tuple $\tup{v_1, \dots,
v_n; v_t'}$, such that $f_D(v_t) \propsubper f_D(v_t')$.  
The effect of \sql{(NOSUBPERIOD)} on a valid-time relation $r$
is captured by the $nosubperiod$ function:
\index{nosubperiod@$nosubperiod()$ (effect of \sql{(NOSUBPERIOD)})}
\[
\begin{aligned}
nosubperiod(r) \defeq
\{ \tup{v_1, \dots, v_n; v_t} \in r \mid 
   &\text{ there is no } \tup{v_1, \dots, v_n; v_t'} \in r \\
   &\text{ such that } f_D(v_t) \propsubper f_D(v_t') \}
\end{aligned}
\]
Applying \sql{(NOSUBPERIOD)} to \pref{subper:2} generates
\pref{subper:3}. 
\begin{examps}
\item \label{subper:3}
\dbtableb{|l|l||l|}
{$J.Adams$ & $17000$ & $[1992, \; 1993]$ \\
 $G.Papas$ & $14500$ & $[1988, \; 1990]$ \\
 $G.Papas$ & $14500$ & $[1990, \; 1991]$
}
\end{examps}

Although \sql{(SUBPERIOD)} and \sql{(NOSUBPERIOD)} are designed to be
used (and in practice will always be used) with relations from \vrel,
I allow \sql{(SUBPERIOD)} and \sql{(NOSUBPERIOD)} to be used with any
valid-time relation. In the proofs of appendix
\ref{trans_proofs}, this saves me having to prove that the
original relation is an element of \vrel whenever \sql{(SUBPERIOD)}
and \sql{(NOSUBPERIOD)} are used.

\subsection{Calendric relations} \label{calrels}

As mentioned in section \ref{tsql2_lang}, \tsql supports multiple
calendars. Roughly speaking, a \tsql calendar describes a system that
people use to measure time (Gregorian calendar, Julian calendar,
etc.). \tsql calendars also specify the meanings of strings within
the quotes of temporal literals, and the available
granularities. According to section 3.2 of \cite{TSQL2book}, \tsql
calendars are defined by the database administrator, the \dbms vendor,
or third parties. In this thesis, I assume that \tsql calendars can
also provide \emph{calendric relations}. Calendric relations behave
like ordinary relations in the database, except that they are
defined by the creator of the \tsql calendar, and cannot be updated.

The exact purpose and contents of each calendric relation are left to
the calendar creator. I assume, however, that a calendric relation
provides information about the time-measuring system of the
corresponding \tsql calendar.\footnote{Future work could establish a
more systematic link between calendric relations and 
\tsql calendars. For example, calendric relations could be required to
reflect (as a minimum) the lattice that shows how the granularities of
the calendar relate to each other (section \ref{tsql2_time}).} The
Gregorian \tsql calendar could, for example, provide the calendric
valid-time relation $gregorian$ below. (I assume here that chronons are
finer than minutes.)
\adbtable{7}{|c|c|c|c|c|c||c|}{$gregorian$}
{$year$ & $month$ & $dnum$ & $dname$ & $hour$ & $minute$ &}
{
\ \dots & \ \dots & \ \dots & \ \dots & \ \dots & \ \dots & \ \dots \\
$1994$ & $Sept$ & $4$ & $Sun$ & $00$ & $00$ & $\{c_{n_1}, \dots, c_{n_2}\}$ \\
$1994$ & $Sept$ & $4$ & $Sun$ & $00$ & $01$ & $\{c_{n_3}, \dots, c_{n_4}\}$ \\
\ \dots & \ \dots & \ \dots & \ \dots & \ \dots & \ \dots & \ \dots \\
$1995$ & $Dec$  & $5$ & $Tue$ & $21$ & $35$ & $\{c_{n_5}, \dots, c_{n_6}\}$ \\
\ \dots & \ \dots & \ \dots & \ \dots & \ \dots & \ \dots & \ \dots 
}
The relation above means that the first minute (00:00) of September
4th 1994 (which was a Sunday) covers exactly the period that starts at
the chronon $c_{n_1}$ and ends at the chronon $c_{n_2}$. Similarly,
the period that starts at $c_{n_3}$ and ends at $c_{n_4}$ is the
second minute (00:01) of September 4th 1994. Of course, the
cardinality of $gregorian$ is very large, though not infinite (time in
\tsql is bounded, and hence there is at most a finite number of
minute-granules). It is important, however, to realise that although
$gregorian$ behaves like a normal relation in the database, it does
not need to be physically present in the database. Its tuples could be
computed dynamically, whenever they are needed, using some algorithm
specified by the \tsql calendar. Other calendric relations may list
the periods that correspond to seasons (spring-periods,
summer-periods, etc.), special days (e.g.\ Easter days), etc. 

Calendric relations like $gregorian$ can be used to construct
relations that represent the periods of partitionings. \pref{calrels:6}, for
example, constructs a one-attribute snapshot relation, that contains
all the time-stamps of $gregorian$ that correspond to
21:36-minutes. The resulting relation represents all the
periods of the partitioning of 21:36-minutes. 
\begin{examps}
\item \select{SELECT DISTINCT SNAPSHOT VALID(greg) \\
              FROM gregorian AS greg \\
              WHERE greg.5 = 21 AND greg.6 = 36}
\label{calrels:6}
\end{examps}
Similarly, \pref{calrels:5} generates a one-attribute snapshot
relation that represents the periods of the partitioning of
Sunday-periods. The embedded \sql{SELECT} statement generates a
valid-time relation of one explicit attribute (whose value is
$\mathit{Sun}$ in all tuples). The time-stamps of this relation are
all the time-stamps of $gregorian$ that correspond to
Sundays (there are many tuples for each Sunday). The
\sql{(PERIOD)} coalesces tuples that correspond to the same Sunday,
leading to a single period-denoting time-stamp for each Sunday. These
time-stamps become the attribute values of the 
relation generated by the overall \pref{calrels:5}.
\begin{examps}
\item \select{SELECT DISTINCT SNAPSHOT VALID(greg2) \\
              FROM (\select{SELECT DISTINCT greg1.4 \\
                            VALID VALID(greg1) \\
                            FROM gregorian AS greg1 \\
                            WHERE greg1.4 = 'Sun'} \\
              \ \ \ \ \ )(PERIOD) AS greg2}
\label{calrels:5}
\end{examps}

In \cite{Androutsopoulos1995b} we argue that calendric relations
constitute a generally useful addition to \tsql, and that unless
appropriate calendric relations are available, it
is not possible to formulate \tsql queries for questions involving
existential or universal quantification or counts over day-names,
month names, season-names, etc.\ (e.g.\ \pref{calrels:1} --
\pref{calrels:3}). 
\begin{examps}
\item Which technicians were at some site on a Sunday?
\label{calrels:1}   
\item Which technician was at Glasgow Central on every Monday in 1994?
\label{calrels:2} 
\item On how many Sundays was J.Adams at Glasgow Central in 1994?
\label{calrels:3} 
\end{examps}

\subsection{The INTERVAL function} \label{interv_fun}

\index{interval@\sql{INTERVAL} (\tsql keyword, returns intervals or introduces interval literals)} 

\tsql provides a function \sql{INTERVAL} that accepts a 
period-denoting expression as its argument, and returns an interval
reflecting the duration of the period. The assumption seems to be that
the resulting interval is specified at whatever granularity the period
is specified. For example, \sql{INTERVAL(PERIOD '[1/12/95 -
3/12/95]')} is the same as \sql{INTERVAL '3' DAY}.  In this thesis, all
periods are specified at the granularity of chronons, and if chronons
correspond to minutes, \sql{PERIOD '[1/12/95 - 3/12/95]'} is an
abbreviation for \sql{PERIOD '[00:00 1/12/95 - 23:59 3/12/95]'}
(sections \ref{tsql2_time} and \ref{tsql2_lang}). Hence, the results of
\sql{INTERVAL} are always specified at the granularity of chronons.
When translating from \topl to \tsql, however, there are cases where
we want the results of \sql{INTERVAL} to be specified at other granularities.

This could be achieved by converting the results of 
\sql{INTERVAL} to the desired granularities. The \tsql mechanisms for
converting intervals from one granularity to another, however, are
very obscure (see section 19.4.6 of \cite{TSQL2book}). To avoid these
mechanisms, I introduce an additional version of the \sql{INTERVAL}
function. If $expr_1$ is a \tsql expression that specifies a period
$p$, and $expr_2$ is the \tsql name (e.g.\
\sql{DAY}, \sql{MONTH}) of a granularity $G$, then
\sql{INTERVAL(}$expr_1$, $expr_2$\sql{)} specifies an interval of $n$
granules (periods) of $G$, where $n$ is as follows. If there
are $k$ consecutive granules $g_1, g_2, g_3, \dots, g_k$ in $G$ such
that $g_1 \union g_2 \union g_3 \union \dots \union g_k = p$, then $n
= k$. Otherwise, $n = 0$. For example,
\sql{INTERVAL(PERIOD '[May 5, 1995 - May 6, 1995]', DAY)} is
the same as \sql{INTERVAL '2' DAY}, because the period covers exactly
2 consecutive day-granules. Similarly, \sql{INTERVAL(PERIOD '[May 1,
1995 - June 30, 1995]', MONTH)} is the same as \sql{INTERVAL '2'
MONTH}, because the period covers exactly two consecutive
month-granules. In contrast, \sql{INTERVAL(PERIOD '[May 1, 1995 - June
15, 1995]', MONTH)} is the same as \sql{INTERVAL '0' MONTH} (zero
duration), because there is no union of consecutive month-granules
that covers exactly the period of \sql{PERIOD '[May 1, 1995 -
June 15, 1995]'}.

\subsection{Correlation names used in the same FROM clause where they
are defined} \label{same_FROM}

The syntax of \tsql (and \sqlnt) does not allow a correlation name to
be used in a \sql{SELECT} statement that is embedded in the same
\sql{FROM} clause that defines the correlation name. For example,
\pref{sfrom:1} is not allowed, because the embedded \sql{SELECT}
statement uses \sql{r1}, which is defined by the same \sql{FROM}
clause that contains the embedded \sql{SELECT} statement.
\begin{examps}
\item \label{sfrom:1}
\select{SELECT \dots \\
        VALID VALID(r1) \\
        FROM rel1 AS r1, \\
        \ \ \ \ \ (\select{SELECT \dots \\
                           VALID VALID(r2) \\
                           FROM rel2 AS r2 \\
                           WHERE VALID(r1) CONTAINS VALID(r2)} \\
        \ \ \ \ \ ) AS r3 \\
        WHERE \dots}
\end{examps}
By \emph{definition of a correlation name} $\alpha$, I mean the expression
``\sql{AS $\alpha$}'' that associates $\alpha$ with a relation. For
example, in \pref{sfrom:1} the definition of \sql{r1} is the ``\sql{AS
r1}''.\footnote{In \sql{SELECT} statements that contain other embedded
\sql{SELECT} statements, multiple definitions of the same correlation
name may be present (there are rules that determine the scope of each
definition). We do not need to worry about such cases, however,
because the generated \tsql code of this chapter never contains
multiple definitions of the same correlation name.} A correlation
name $\alpha$ is \emph{defined by a \sql{FROM} clause} $\xi$, if $\xi$
contains the definition of $\alpha$, and this definition is not within
a \sql{SELECT} statement which is embedded in $\xi$. For example, in
\pref{sfrom:1} the \sql{r2} is defined by the ``\sql{FROM rel2 AS
r2}'' clause, not by the ``\sql{FROM rel1 AS r1, (\dots) AS r3}''
clause.

In this thesis, I allow a correlation name to be used in a
\sql{SELECT} statement that is embedded in the same \sql{FROM} clause
that defines the correlation name, provided that the definition of
the correlation name precedes the embedded \sql{SELECT} statement.
\pref{sfrom:1} is acceptable, because the definition of \sql{r1}
precedes the embedded \sql{SELECT} statement where \sql{r1} is
used. In contrast, \pref{sfrom:2} is not acceptable, 
because the definition of \sql{r1} follows the
embedded \sql{SELECT} statement where \sql{r1} is used.
\begin{examps}
\item \label{sfrom:2}
\select{SELECT \dots \\
        VALID VALID(r1) \\
        FROM (\select{SELECT \dots \\
                      VALID VALID(r2) \\
                      FROM rel2 AS r2 \\
                      WHERE VALID(r1) CONTAINS VALID(r2)} \\
        \ \ \ \ \ ) AS r3, \\
        \ \ \ \ \ rel1 AS r1 \\
        WHERE \dots}
\end{examps}
The intended semantics of statements like \pref{sfrom:1}
should be easy to see: when evaluating the embedded \sql{SELECT}
statement, \sql{VALID(r1)} should represent the time-stamp of a tuple
from $rel1$. The restriction that the definition of the correlation
name must precede the embedded \sql{SELECT} is imposed to make this
modification easier to implement.

The modification of this section is used in the \topl to \tsql
translation rules for $\at[\phi_1,
\phi_2]$, $\before[\phi_1, \phi_2]$, and $\after[\phi_1, \phi_2]$
(section \ref{trans_rules} below and appendix
\ref{trans_proofs}).

\subsection{Equality checks and different domains} \label{eq_checks}

Using the equality predicate (\sql{=}) with expressions that refer to
values from different domains often causes the \tsql (or \sqlnt)
interpreter to report an error. If, for example, the domain of the
first explicit attribute of $rel$ is the set of all integers, 
\sql{r.1} in \pref{eqs:1} stands for an integer. \tsql (and \sqlnt)
does not allow integers to be compared to strings (e.g.\
``J.Adams''). Consequently, \pref{eqs:1} would be rejected, and an
error message would be generated.
\begin{examps}
\item \select{SELECT DISTINCT SNAPSHOT r.2 \\ 
              FROM rel AS r \\
              WHERE r.1 = 'J.Adams'} \label{eqs:1}
\end{examps}
In other cases (e.g.\ if a real number is compared to an integer),
type conversions take place before the comparison. To by-pass
uninteresting details, in this thesis I assume that no type
conversions occur when ``\sql{=}'' is used. The equality predicate is
satisfied iff both of its arguments refer to the same element of $D$
(universal domain). No error occurs if the arguments refer to values
from different domains. In the example of \pref{eqs:1}, \sql{r.1 =
'J.Adams'} is not satisfied, because \sql{r.1} refers to an integer in
$D$, \sql{'J.Adams'} to a string in $D$, and integers are different from
strings. Consequently, in the \tsql version of this thesis
\pref{eqs:1} generates the empty relation (no errors occur). 

\subsection{Other minor changes}

\tsql does not allow partitioning units to follow
\sql{SELECT} statements that are not embedded into other \sql{SELECT}
statements. For example, \pref{pus:10} on its own is not acceptable.
\begin{examps}
\item \label{pus:10}
\sql{(}\select{SELECT DISTINCT r1.1, r1.2 \\
               VALID VALID(r1) \\
               FROM rel AS r1}\\
\sql{)(PERIOD)}
\end{examps}
\sql{SELECT} statements like \pref{pus:10} can be easily made
acceptable by embedding them into another
\sql{SELECT} statement (e.g.\ \pref{pus:11}).
\begin{examps}
\item \label{pus:11}
\select{SELECT DISTINCT r2.1, r2.2 \\
        VALID VALID(r2) \\
        FROM (\select{SELECT DISTINCT r1.1, r1.2 \\
                      VALID VALID(r1) \\
                      FROM rel AS r1} \\
        \ \ \ \ \ )(PERIOD) AS r2}
\end{examps}
For simplicity, I allow stand-alone statements like
\pref{pus:10}. I assume that \pref{pus:10}
generates the same relation as \pref{pus:11}. I also allow stand-alone
\sql{SELECT} statements enclosed in brackets (e.g.\ \pref{pus:12}). 
I assume that the enclosing brackets are simply ignored. 
\begin{examps}
\item \label{pus:12}
\sql{(}\select{SELECT DISTINCT r1.1, r1.2 \\
               VALID VALID(r1) \\
               FROM rel AS r1}\\
\sql{)}
\end{examps}


\section{Additional TSQL2 terminology} \label{additional_tsql2}

This section defines some additional terminology, that is used to formulate
and prove the correctness of the \topl to \tsql translation.

\paragraph{Column reference:} A \emph{column
reference} is an expression of the form $\alpha.i$ or
\sql{VALID(}$\alpha$\sql{)}, where $\alpha$ is a correlation name and
$i \in \{1,2,3,\dots\}$ (e.g.\ \sql{sal.2}, \sql{VALID(sal)}). 

\paragraph{Binding context:} A \sql{SELECT} statement $\Sigma$ is a
\emph{binding context} for a column reference $\alpha.i$ or
\sql{VALID(}$\alpha$\sql{)} iff:
\begin{itemize}  

\item the column reference is part of $\Sigma$, 

\item $\alpha$ is defined (in the sense of section \ref{same_FROM}) by
the topmost \sql{FROM} clause of $\Sigma$, and

\item the column reference is not in the topmost \sql{FROM} clause of
$\Sigma$; or it is in the topmost \sql{FROM} clause of $\Sigma$, but
the definition of $\alpha$ precedes the column reference.

\end{itemize}

By \emph{topmost \sql{FROM} clause of $\Sigma$} I mean the (single)
\sql{FROM} clause of $\Sigma$ that does not appear in any \sql{SELECT}
statement embedded in $\Sigma$ (e.g.\ the topmost \sql{FROM} clause of
\pref{fcn:1} is the ``\sql{FROM tab1 AS r1, (\dots) AS r3}''). We will
often have to distinguish between individual \emph{occurrences} of column
references. For example, \pref{fcn:2} is a binding context for the
occurrence of \sql{VALID(r1)} in the \sql{VALID} clause, because that
occurrence is part of \pref{fcn:2}, \sql{r1} is defined by the topmost
\sql{FROM} clause of \pref{fcn:2}, and the occurrence of
\sql{VALID(r1)} is not in the topmost \sql{FROM} clause of
\pref{fcn:2}. \pref{fcn:2}, however, is \emph{not} a binding context
for the occurrence of \sql{VALID(r1)} in the embedded \sql{SELECT}
statement of \pref{fcn:2}, because that occurrence is in the topmost
\sql{FROM} clause, and it does not follow the definition of \sql{r1}.
\begin{examps}
\item \label{fcn:2}
\select{SELECT DISTINCT r1.1, r3.2 \\
        VALID VALID(r1) \\
        FROM (\select{SELECT DISTINCT SNAPSHOT r2.1, r2.2 \\
                           FROM tab2 AS r2 \\
                           WHERE VALID(r2) CONTAINS VALID(r1)} \\
        \ \ \ \ \ ) AS r3, \\
        \ \ \ \ \ tab1 AS r1 \\
        WHERE r1.1 = 'J.Adams'}
\end{examps}
In contrast, \pref{fcn:1} \emph{is} a binding context for the
\sql{VALID(r1)} in the embedded \sql{SELECT}, because the definition
of \sql{r1} precedes that occurrence of \sql{VALID(r1)}. 
\begin{examps}

\item \label{fcn:1}
\select{SELECT DISTINCT r1.1, r3.2 \\
        VALID VALID(r1) \\
        FROM tab1 AS r1, \\
        \ \ \ \ \ (\select{SELECT DISTINCT SNAPSHOT r2.1, r2.2 \\
                           FROM tab2 AS r2 \\
                           WHERE VALID(r2) CONTAINS VALID(r1)} \\
        \ \ \ \ \ ) AS r3 \\
        WHERE r1.1 = 'J.Adams'}
\end{examps}
In both \pref{fcn:2} and \pref{fcn:1}, the overall \sql{SELECT}
statement is not a binding context for
\sql{r2.1}, \sql{r2.2}, and \sql{VALID(r2)}, because \sql{r2} is not
defined by the topmost \sql{FROM} clause of the overall \sql{SELECT}
statement. The embedded \sql{SELECT} statement of \pref{fcn:2} and
\pref{fcn:1}, however, \emph{is} a binding context for \sql{r2.1},
\sql{r2.2}, and \sql{VALID(r2)}. 

\paragraph{Free column reference:} A column reference $\alpha.i$ or
\sql{VALID($\alpha$)} is a \emph{free column reference} in a \tsql
expression $\xi$, iff:
\begin{itemize}
\item the column reference is part of $\xi$, and
\item there is no \sql{SELECT} statement in $\xi$ (possibly being the
whole $\xi$) that is a binding context for the column reference.
\end{itemize}
The \sql{VALID(r1)} in the embedded \sql{SELECT} statement of
\pref{fcn:2} is free in \pref{fcn:2}, because there is no
binding context for that occurrence in \pref{fcn:2}. In contrast, the
\sql{VALID(r1)} in the \sql{VALID} clause of \pref{fcn:2} is not free
in \pref{fcn:2}, because \pref{fcn:2} is a binding context for that
occurrence. The \sql{VALID(r2)} of \pref{fcn:2} is not free in
\pref{fcn:2}, because the embedded \sql{SELECT}
statement is a binding context for \sql{VALID(r2)}.

A correlation name $\alpha$ \emph{has a free column reference in} a
\tsql expression $\xi$, iff there is a free column reference $\alpha.i$ or
\sql{VALID($\alpha$)} in $\xi$. For every \tsql expression $\xi$,
$\fcn(\xi)$ 
\index{fcn@$\fcn(\xi)$ (set of all correlation names with free column references in $\xi$)}
is the set of all correlation names that have a free column reference
in $\xi$. For example, if $\xi$ is \pref{fcn:2}, $\fcn(\xi) =
\{$\sql{r1}$\}$ (the \sql{VALID(r1)} of the embedded
\sql{SELECT} statement is free in \pref{fcn:2}).

There must be no free column references in the overall \sql{SELECT}
statements that are submitted to the \tsql (or \sqlnt) interpreter
(though there may be free column references in their embedded
\sql{SELECT} statements). Hence, it is important to prove that there
are no free column references in the overall \sql{SELECT} statements
generated by the \topl to \tsql translation. 

\paragraph{Value expression:} 
In \tsql (and \sqlnt), \emph{value expression} refers to expressions
that normally evaluate to elements of $D$ (universal domain). (The
meaning of ``normally'' will be explained in following paragraphs.)
For example, \sql{'J.Adams'}, \sql{VALID(sal)}, and
\sql{INTERSECT(PERIOD '[1993 - 1995]', PERIOD '[1994 - 1996]')} are
all value expressions.

\paragraph{Assignment to correlation names:} An \emph{assignment to
correlation names} is a function $g^{db}$ 
\index{gdb@$g^{db}()$, $(g^{db})^{\alpha}_{\tup{v_1, v_2, \dots}}()$ (assignment to correlation names)} 
that maps every \tsql correlation name to a possible tuple of a
snapshot or valid-time relation. $G^{db}$
\index{Gdb@$G^{db}$ (set of all assignments to correlation names)}
is the set of all assignments to correlation names 

If $\alpha$ is a (particular) correlation name, $\tup{v_1, v_2,
\dots}$ is a (particular) tuple of a snapshot or valid-time relation, 
and $g^{db} \in G^{db}$, $(g^{db})^{\alpha}_{\tup{v_1, v_2, \dots}}$ 
\index{gdb@$g^{db}()$, $(g^{db})^{\alpha}_{\tup{v_1, v_2, \dots}}()$ (assignment to correlation names)} 
is the same as $g^{db}$, except that it assigns $\tup{v_1, v_2,
\dots}$ to $\alpha$. (For every other correlation name, the values of
$g^{db}$ and $(g^{db})^{\alpha}_{\tup{v_1, v_2, \dots}}$ are
identical.)

\paragraph{eval:} 
\index{eval@$eval()$ (evaluates \tsql expressions)}
For every \tsql \sql{SELECT} statement
or value expression $\xi$, and every $st \in \chrons$ and $g^{db} \in
G^{db}$, $eval(st, \xi, g^{db})$ is the relation (if $\xi$ is a
\sql{SELECT} statement) or the element of $D$ (if $\xi$ is a value
expression) that is generated when the \tsql interpreter evaluates
$\xi$ in the following way:
\begin{itemize}
\item $st$ is taken to be the current chronon.

\item Every free column reference of the form 
$\alpha.i$ is treated as a value expression that evaluates to $v_i$,
where $v_i$ is the $i$-th attribute value in the tuple
$g^{db}(\alpha)$.

\item Every free column reference of the form \sql{VALID($\alpha$)} is
treated as a value expression that evaluates to $v_t$, where $v_t$ is
the time-stamp of $g^{db}(\alpha)$. 

\end{itemize}
If $\xi$ cannot be evaluated in this way (e.g.\ $\xi$ contains a free
column reference of the form $\alpha.4$, and $g^{db}(\alpha) =
\tup{v_1, v_2, v_3}$), $eval(st, \xi, g^{db})$ returns the special
value $error$.
\index{error@$error$ (signals evaluation error)}
(I assume that $error \not\in D$.)  A value expression $\xi$
\emph{normally} (but not always) evaluates to an element of $D$,
because when errors arise $eval(st, \xi, g^{db}) = error \not\in
D$. If, however, $eval(st, \xi, g^{db}) \not= error$, $eval(st, \xi,
g^{db}) \in D$.

Strictly speaking, $eval$ should also have as its argument the
database against which $\xi$ is evaluated. For simplicity, I
overlook this detail. Finally, if $\fcn(\xi) = \emptyset$ ($\xi$
contains no free column references), $eval(st, \xi, g^{db})$ does not
depend on $g^{db}$. In this case, I write simply $eval(st, \xi)$.


\section{Modifications in TOP and additional TOP terminology} \label{TOP_mods}

In the formulae generated by the English to \topl translation, each  
$\partop[\sigma, \beta]$ is conjoined with a subformula
that is (or contains another subformula) of the form $\at[\beta,
\phi]$, $\before[\beta, \phi]$, or $\after[\beta, \phi]$ ($\sigma \in \parts$, $\phi \in \ynforms$, $\beta \in
\vars$, and the $\beta$ of \partop is the same as that of
\at, \before, or \after). For example, \pref{tmods:1} and
\pref{tmods:3} are mapped to \pref{tmods:2} and \pref{tmods:4}.
Also the reading of \pref{tmods:5} where Monday is the time when the
tank was empty (rather than a reference time; section
\ref{past_perfect}) is mapped to \pref{tmods:6}.
\begin{examps}
\item Tank 2 was empty on a Monday. \label{tmods:1}
\item $\partop[monday^g, mon^v] \land \at[mon^v, \past[e^v, empty(tank2)]]$
   \label{tmods:2}
\item On which Monday was tank 2 empty? \label{tmods:3}
\item $?mon^v \; \partop[monday^g, mon^v] \land 
       \at[mon^v, \past[e^v, empty(tank2)]]$ \label{tmods:4}
\item Tank 2 had been empty on a Monday. \label{tmods:5}
\item $\partop[monday^g, mon^v] \land 
       \past[e1^v, \perf[e2^v, \at[mon^v, empty(tank2)]]]$ \label{tmods:6}
\end{examps}
In this chapter, I use a slightly different version of \topl, where the
\partop is merged with the corresponding \at,
\before, or \after. For example, \pref{tmods:2}, \pref{tmods:4}, and
\pref{tmods:6} become \pref{tmods:7}, \pref{tmods:8}, and \pref{tmods:9}
respectively.
\begin{examps}
\item $\at[monday^g, mon^v, \past[e^v, empty(tank2)]]$ \label{tmods:7}
\item $?mon^v \; \at[monday^g, mon^v, \past[e^v, empty(tank2)]]$
  \label{tmods:8}
\item $\past[e1^v, \perf[e2^v, \at[monday^g, mon^v, empty(tank2)]]]$
  \label{tmods:9}
\end{examps}
The semantics of $\at[\sigma, \beta, \phi]$, $\before[\sigma, \beta,
\phi]$, and $\after[\sigma, \beta, \phi]$ follow ($f$ is
$\fgparts$ if $\sigma \in \gparts$, and $\fcparts$ if $\sigma
\in \cparts$.)
\begin{itemize}
\item $\denot{st,et,lt,g}{\at[\sigma, \beta, \phi]} = T$ iff 
   $g(\beta) \in f(\sigma)$ and 
   $\denot{st, et, lt \intersect g(\beta), g}{\phi} = T$. 
\item $\denot{st,et,lt,g}{\before[\sigma, \beta, \phi]} = T$ iff 
   $g(\beta) \in f(\sigma)$ and 
   $\denot{st, et, lt \intersect [t_{first}, minpt(\denot{g}{\beta})), g}
    {\phi} = T$. 
\item $\denot{st,et,lt,g}{\after[\sigma, \beta, \phi]} = T$ iff 
   $g(\beta) \in f(\sigma)$ and 
   $\denot{st, et, lt \intersect (maxpt(\denot{g}{\beta}), t_{last}], g}
    {\phi} = T$. 
\end{itemize}
In the \topl version of this chapter, $\partop[\sigma, \beta]$,
$\at[\beta, \phi]$, $\before[\beta, \phi]$, and $\after[\beta, \phi]$
($\beta \in \vars$) are no longer yes/no formulae. 
$\at[\kappa, \phi]$, $\before[\kappa, \phi]$, and $\after[\kappa,
\phi]$ ($\kappa \in \cons$), however, are still
yes/no formulae.

The \topl version of chapter \ref{TOP_chapter} is more convenient for
the English to \topl mapping, while the version of this chapter
simplifies the \topl to \tsql translation.  In the prototype \nlitdb,
there is a converter between the module that translates from English
to \topl and the \topl to \tsql translator. The module that translates
from English to \topl maps \pref{tmods:1}, \pref{tmods:3}, and
\pref{tmods:5} to \pref{tmods:2}, \pref{tmods:4}, and \pref{tmods:6}
respectively. The converter turns \pref{tmods:2}, \pref{tmods:4}, and
\pref{tmods:6} into \pref{tmods:7}, \pref{tmods:8}, and
\pref{tmods:9}, which are then passed to the \topl to \tsql
translator.

The reader is reminded that the $\partop[\sigma, \beta,
\nu_{ord}]$ version of \partop is not used
in the translation from English to \topl (section
\ref{TOP_FS}). Hence, only the $\partop[\sigma, \beta]$ form of
\partop is possible in formulae generated by the
English to \topl translation. In the \topl version of this
chapter, \partop operators of this form are merged with \at, \before,
or \after operators. Therefore, no \partop operators occur in the
formulae that are passed to the \topl to \tsql translator. 
As with the \at, \before, and \after of chapter
\ref{TOP_chapter} (section \ref{top_syntax}), in every $\at[\sigma, \beta,
\phi]$, $\before[\sigma, \beta, \phi]$, and $\after[\sigma, \beta,
\phi]$, I require $\beta$ not to occur within
$\phi$. This is needed to prove the correctness of the
\topl to \tsql translation.

To avoid complications in the \topl to \tsql translation, I require
that in any $\at[\kappa, \phi]$, $\before[\kappa, \phi]$, or
$\after[\kappa, \phi]$ ($\kappa \in \cons$, $\phi \in \ynforms$) that
is passed to the \topl to \tsql translator, $\fcons(\kappa) \in
\periods$. (The definitions of section \ref{at_before_after_op} are
more liberal: they allow $\fcons(\kappa)$ not to belong to \periods,
though if $\fcons(\kappa) \not\in \periods$, the denotation of
$\at[\kappa, \phi]$, $\before[\kappa, \phi]$, or $\after[\kappa,
\phi]$ is always $F$.) In practice,  formulae generated by the English
to \topl mapping never violate this constraint.

For every $\phi \in \ynforms$, $\corn{\phi}$ 
\index{'`@$\corn{}$ (corners)}
(pronounced ``corners $\phi$'') is the tuple $\tup{\tau_1, \tau_2,
\tau_3, \dots, \tau_n}$, where $\tau_1,
\dots, \tau_n$ are all the constants that are used as arguments
of predicates in $\phi$, and all the variables that occur in $\phi$,
in the same order (from left to right) they appear in $\phi$. If a
constant occurs more than once as a predicate argument in $\phi$, or
if a variable occurs more than once in $\phi$, there are multiple
$\tau_i$s in $\corn{\phi}$ for that constant or variable. If $\corn{\phi} =
\tup{\tau_1, \tau_2, \tau_3, \dots,
\tau_n}$, the \emph{length} of $\corn{\phi}$ is 
$n$. For example, if:
\[
\phi = \ntense[t^v, woman(p^v)] \land 
       \at[1991, \past[e^v, manager\_of(p^v, sales)]]
\]
then $\corn{\phi} = \tup{t^v, p^v, e^v, p^v, sales}$, and the
length of $\corn{\phi}$ is 5.


\section{Linking the TOP model to the database} \label{linking_model}

As discussed in section \ref{denotation}, the answer to an English
question submitted at $st$ must report the denotation
$\denot{M,st}{\phi}$ of the corresponding \topl formula $\phi$.
$\denot{M,st}{\phi}$ follows from the semantics of \topl, provided
that the model $M$, which intuitively provides all the necessary
information about the modelled world, has been defined. In a \nlidb,
the only source of information about the world is the
database.\footnote{This is not entirely true in the framework of this
thesis, as there is also a type-hierarchy of world-entities in the
\hpsg grammar (section \ref{HPSG_basics}).} Hence, $M$ has to be
defined in terms of the information in the database. This mainly
involves defining  \fcons, \fpfuns, \fculms, \fcparts, and \fgparts
(which are parts of $M$) in terms of database concepts. 

\begin{figure}
\hrule
\begin{center}
\medskip
\includegraphics[scale=.6]{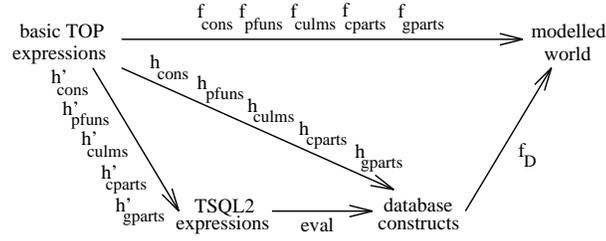}
\caption{Paths from basic \topl expressions to the modelled world}
\label{link_paths_fig}
\end{center}
\hrule
\end{figure}

\fcons, \fpfuns, \fculms, \fcparts, and \fgparts show how certain basic
\topl expressions (constants, predicates, and partitioning names)
relate to the modelled world. These functions will be defined in terms
of the functions \hcons, \hpfuns, \hculms, \hcparts, and
\hgparts (to be discussed in section \ref{h_funs}), and $f_D$ (section
\ref{relational}). Roughly speaking, the $h$ functions map basic \topl
expressions to database constructs (attribute values or relations),
and $f_D$ maps the attribute values of these constructs to world
objects (figure \ref{link_paths_fig}). \hcons, \hpfuns, \hculms,
\hcparts, and \hgparts will in turn be defined in terms of the
functions \hconsp, \hpfunsp, \hculmsp, \hcpartsp, and \hgpartsp (to be
discussed in section \ref{via_TSQL2}), and $eval$ (section
\ref{additional_tsql2}). The $h'$ functions map basic \topl
expressions to \tsql expressions, and $eval$ maps \tsql expressions to
database constructs.

After defining the $h'$ functions, one could compute
$\denot{M,st}{\phi}$ using a reasoning system, that would contain rules
encoding the semantics of \topl, and that would use the path basic
\topl expressions $\rightarrow$ \tsql expressions $\rightarrow$
database constructs $\rightarrow$ modelled world (figure
\ref{link_paths_fig}) to compute any necessary values of \fcons,
\fpfuns, \fculms, \fcparts, and \fgparts. That is, only basic \topl
expressions would be translated into \tsql, and the \dbms would be
used only to evaluate the \tsql translations of these expressions. The
rest of the processing to compute $\denot{M,st}{\phi}$
would be carried out by the reasoning system.

\begin{figure}
\hrule
\begin{center}
\medskip
\includegraphics[scale=.6]{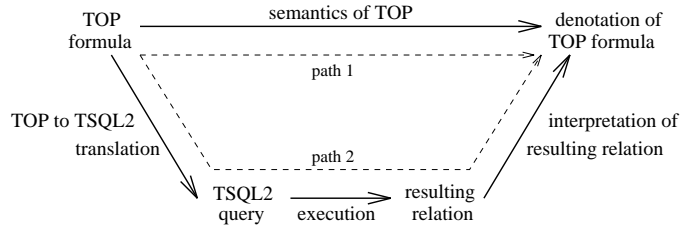}
\caption{Paths from TOP formulae to their denotations}
\label{trans_paths_fig}
\end{center}
\hrule
\end{figure}

This thesis adopts an alternative approach that exploits the
capabilities of the \dbms to a larger extent, and that requires no
reasoning system. Based on the $h'$ functions (that map only basic
\topl expressions to \tsql expressions), a method to translate
\emph{any} \topl formula into \tsql will be developed. Each \topl
formula $\phi$ will be mapped to a single \tsql query
(figure \ref{trans_paths_fig}). This will be executed by the \dbms,
generating a relation that represents (via an interpretation function)
$\denot{M,st}{\phi}$. It will be proven formally that this approach
generates indeed $\denot{M,st}{\phi}$ (i.e.\, that paths 1 and 2
of figure \ref{trans_paths_fig} lead to the same result).

There is one further complication: the values of \fcons,
\fpfuns, \fculms, \fcparts, and \fgparts will ultimately be obtained
by evaluating \tsql expressions returned by \hconsp, \hpfunsp,
\hculmsp, \hcpartsp, and \hgpartsp. A \tsql expression, however, may
generate different results when evaluated at different times (e.g.\ a
\sql{SELECT} statement may return different results after a
database relation on which the statement operates has been
updated). This causes the values of \fcons,
\fpfuns, \fculms, \fcparts, and \fgparts to become sensitive to the
time where the \tsql expressions of the $h'$ functions are
evaluated. We want this time to be $st$, so that the \tsql expressions
of the $h'$ functions will operate on the information that is in the
database when the question is submitted, and so that a \tsql literal
like \sql{PERIOD 'today'} (section \ref{tsql2_lang}) in the
expressions of the $h'$ functions will be correctly taken to refer to
the day that contains $st$. To accommodate this, \fcons,
\fpfuns, \fculms, \fcparts, and \fgparts must be made sensitive to
$st$:

\fcons becomes a function $\pts \mapsto (\cons \mapsto \objs)$ instead
of $\cons \mapsto \objs$. This allows the world objects that are
assigned to \topl constants via \fcons to be different at different
$st$s. Similarly, \fpfuns is now a function over \pts. For every $st
\in \pts$, $\fpfuns(st)$ is in turn a function that maps each pair
$\tup{\pi,n}$, where $\pi \in \pfuns$ and $n \in \{1,2,3,\dots\}$, to
another function $(\objs)^n \mapsto pow(\periods)$ (cf.\ the
definition of \fpfuns in section \ref{top_model}). The definitions of
\fculms, \fcparts, and \fgparts are modified accordingly.
Whatever restrictions applied to \fcons, \fpfuns, \fculms, \fcparts,
and \fgparts, now apply to $\fcons(st)$, $\fpfuns(st)$, $\fculms(st)$,
$\fcparts(st)$, and $\fgparts(st)$, for every $st \in \chrons$. Also,
wherever \fcons, \fpfuns, \fculms, \fgparts, \fcparts were used in the
semantics of \topl, $\fcons(st)$, $\fpfuns(st)$, $\fculms(st)$, and
$\fgparts(st)$ should now be used.  The \topl model also becomes
sensitive to $st$, and is now defined as follows:
\[
M(st) =
\tup{\tup{\pts, \prec}, \objs, 
     \fcons(st), \fpfuns(st), \fculms(st), \fgparts(st), \fcparts(st)}
\]
Intuitively, $M(st)$ reflects the history of the world as recorded in
the database at $st$. (If the database supports both valid and
transaction time, $M(st)$ reflects the ``beliefs'' of the database at
$st$; see section \ref{tdbs_general}.) The answer to an English
question submitted at $st$ must now report the denotation
$\denot{M(st),st}{\phi}$ of the corresponding \topl formula $\phi$. 


\section{The $h$ functions} \label{h_funs}

I first discuss \hcons, \hpfuns, \hculms, \hcparts, and \hgparts, the
functions that -- roughly speaking -- map basic \topl expressions to
database constructs. As with \fcons, \fpfuns, \fculms, \fcparts,
and \fgparts, the values of \hcons, \hpfuns, \hculms, \hcparts, and
\hgparts will ultimately be obtained by evaluating \tsql expressions
at $st$. The results of these evaluations can be different at different
$st$s, and hence the definitions of the $h$ functions must be
sensitive to $st$. 

\paragraph{$\mathbf{h_{cons}}$:} 
\index{hcons@$\hcons()$ (\topl constants to attribute values)}
\hcons is a function $\pts \mapsto (\cons \mapsto D)$. For every $st
\in \pts$, $\hcons(st)$ is in turn a function that maps each \topl
constant to an attribute value that represents the same
world-entity. For example, $\hcons(st)$ could map the \topl constant
$sales\_department$ to the string attribute value $Sales \;
Department$, and the constant $\mathit{today}$ to the element
of $D_P$ ($D_P \subseteq D$) which denotes the day-period that
contains $st$. 

\paragraph{$\mathbf{h_{pfuns}}$:}
\index{hpfuns@$\hpfuns()$ (predicates to relations showing maximal periods of situations)}
\hpfuns is a function over \pts. For every $st \in \pts$, $\hpfuns(st)$
is in turn a function over 
$\pfuns \times \{1,2,3,\dots\}$, such that for every $\pi \in \pfuns$
and $n \in \{1,2,3,\dots\}$, $\hpfuns(st)(\pi, n) \in
\cvrel(n)$ (section \ref{bcdm}). $\hpfuns(st)$ is intended to map every
\topl predicate of functor $\pi$ and arity $n$ to a relation that
shows for which arguments of the predicate and at which maximal
periods the situation represented by the predicate is true, according
to the ``beliefs'' of the database at $st$. For example, if
$circling(ba737)$ represents the situation where BA737 is circling,
and according to the ``beliefs'' of the database at $st$, $p$ is a maximal period where BA737 was/is/will be circling,
$\hpfuns(st)(circling, 1)$ must contain a tuple $\tup{v;v_t}$, where
$f_D(v) = \fcons(ba737)$ ($v$ denotes the flight BA737), and $f_D(v_t)
= p$. Similarly, if $\hpfuns(st)(circling, 1)$ contains a tuple
$\tup{v;v_t}$, where $f_D(v) =
\fcons(ba737)$ and $f_D(v_t) = p$, $p$ is a maximal
period where BA737 was/is/will be circling, according to the
``beliefs'' of the database at $st$.

\paragraph{$\mathbf{h_{culms}}$:} 
\index{hculms@$\hculms()$ (predicates to relations showing if situations reach their climaxes)}
\hculms is a function over \pts. For every $st \in \pts$,
$\hculms(st)$ is in turn a function over
$\pfuns \times \{1,2,3,\dots\}$, such that for every
$\pi \in \pfuns$ and $n \in \{1,2,3,\dots\}$, $\hculms(st)(\pi, n) \in
\srel(n)$. Intuitively, \hculms plays the same role as \fculms 
(section \ref{top_model}). In practice, \hculms is consulted only for
predicates that describe situations with inherent
climaxes. $\hculms(st)$ maps each \topl predicate of functor $\pi$ and arity
$n$ to a relation that shows for which predicate arguments the situation of 
the predicate reaches its climax at the latest time-point where the
situation is ongoing, according to the ``beliefs'' of the database at $st$. If, for example, $inspecting(j\_adams, ba737)$
represents the situation where J.Adams is inspecting BA737, 
$\hpfuns(st)(inspecting, 2)$ is a relation in $\cvrel(2)$ and
$\hculms(st)(inspecting, 2)$ a relation in $\srel(2)$. If, according to the
``beliefs'' of the database at $st$, the
maximal periods where J.Adams was/is/will be inspecting BA737 are $p_1,
p_2, \dots, p_j$, $\hpfuns(st)(inspecting, 2)$ contains the tuples
$\tup{v_1, v_2; v^1_t}$, $\tup{v_1, v_2; v^2_t}$, \dots,
$\tup{v_1, v_2; v^j_t}$, where $f_D(v_1) = \fcons(j\_adams)$,
$f_D(v_2) = \fcons(ba737)$, and $f_D(v^1_t) = p_1$, $f_D(v^2_t) =
p_2$, \dots, $f_D(v^j_t) = p_j$. Let us assume that $p$ is the latest
maximal period among $p_1, \dots, p_j$. $\hculms(st)(inspecting, 2)$ 
contains $\tup{v_1, v_2}$ iff according to the ``beliefs'' of the
database at $st$, the inspection of BA737 by J.Adams reaches its
completion at the end of $p$.

\paragraph{$\mathbf{h_{gparts}}$:} 
\index{hgparts@$\hgparts()$ (gappy part.\ names to relations representing gappy partitionings)}
\hgparts is a function over \pts. For every $st \in \pts$,
$\hgparts(st)$ is in turn a function that maps
every element of \gparts to an $r \in \srel(1)$, such that the set $S
= \{ f_D(v) \mid \tup{v} \in r \}$ is a gappy partitioning.
$\hgparts(st)$ is intended to map each \topl gappy partitioning
name $\sigma_g$ to a one-attribute snapshot relation $r$, whose
attribute values represent the periods of the gappy partitioning $S$
that is assigned to $\sigma_g$. For example, $\hgparts(st)$ could map
$monday^g$ to a one-attribute snapshot relation whose attribute values
denote all the Monday-periods. 

As with the other $h$ functions, the
values of \hgparts will ultimately be obtained by evaluating \tsql
expressions at $st$ (see section \ref{via_TSQL2} below). The
results of these evaluations can in principle be different at
different $st$s, and this is why \hgparts is defined to be sensitive
to $st$. In practice, however, the \tsql expressions that are
evaluated to obtain the values of \hgparts will be insensitive to their
evaluation time, and hence the values of \hgparts will not
depend on $st$. Similar comments apply to \hcparts below. 

\paragraph{$\mathbf{h_{cparts}}$:} 
\index{hcparts@$\hcparts()$ (compl.\ part.\ names to relations representing compl.\ partitionings)}
\hcparts is a function over \pts. For every $st \in \pts$,
$\hcparts(st)$ is in turn a function that maps
every element of \cparts to an $r \in \srel(1)$, such that the set $S
= \{f_D(v) \mid \tup{v} \in r \}$ is a complete partitioning.
$\hcparts(st)$ is intended to map each \topl complete partitioning name
$\sigma_c$ to a one-attribute snapshot relation $r$, whose attribute
values represent the periods of the complete partitioning $S$ that is
assigned to $\sigma_c$. For example, $\hcparts(st)$ could map $day^c$ to a
one-attribute snapshot relation whose attribute values denote all the
day-periods.


\section{The TOP model in terms of database concepts}
\label{resulting_model} 

The \topl model (see section \ref{top_model} and the revisions of
section \ref{linking_model}) can now be defined in terms
of database concepts as follows. 

\paragraph{Point structure:} $\tup{\pts, \prec} \defeq 
\tup{\chrons, \prec^{chrons}}$ \\
As mentioned in section \ref{tsql2_time}, $\chrons \not= \emptyset$,
and $\tup{\chrons, \prec^{chrons}}$ has the properties of
transitivity, irreflexivity, linearity, left and right boundedness,
and discreteness. Hence, $\tup{\chrons, \prec^{chrons}}$ qualifies as
a point structure for \topl (section \ref{temporal_ontology}). 

Since $\tup{\pts, \prec} =
\tup{\chrons, \prec^{chrons}}$, $\periods_{\tup{\pts, \prec}} =
\periods_{\tup{\chrons, \prec^{chrons}}}$, and $\instants_{\tup{\pts,
\prec}} = \instants_{\tup{\chrons, \prec^{chrons}}}$. I write
simply \periods and \instants to refer to these sets.

\paragraph{$\mathbf{OBJS}$:} $\objs \defeq \objsdb$ \\
Since $\periods \subseteq \objsdb$ (section \ref{bcdm}) and $\objs =
\objs^{db}$, $\periods \subseteq \objs$, as
required by section \ref{top_model}.

\paragraph{$\mathbf{f_{cons}}$:} 
\index{fcons@$\fcons()$ (maps \topl constants to world objects)}
For every $st \in \pts$ and $\kappa \in \cons$, I define
$\fcons(st)(\kappa) \defeq f_D(\hcons(st)(\kappa))$.
Since $\hcons(st)$ is a function $\cons \mapsto D$, and $f_D$ is a
function $D \mapsto \objsdb$, and $\objs = \objsdb$, 
$\fcons(st)$ is a function $\cons \mapsto \objs$, as required by section
\ref{top_model} and the revisions of section \ref{linking_model}. 

\paragraph{$\mathbf{f_{pfuns}}$:} 
\index{fpfuns@$\fpfuns()$ (returns the maximal periods where predicates hold)}
According to section \ref{top_model} and the revisions of section
\ref{linking_model}, for every $st \in \pts$, $\fpfuns(st)$ must be a
function:  
\[ \pfuns \times \{1,2,3,\dots\} \mapsto 
   ((\objs)^n \mapsto \pow(\periods))
\]
That is, for every $\pi \in \pfuns$, every $n \in \{1,2,3,\dots\}$,
and every $o_1, \dots, o_n \in \objs$, $\fpfuns(st)(\pi,n)(o_1, \dots,
o_n)$ must be a set of periods. I define $\fpfuns(st)(\pi,n)(o_1, \dots,
o_n)$ as follows:
\[
\fpfuns(st)(\pi, n)(o_1, \dots, o_n) \defeq
\{ f_D(v_t) \mid 
   \tup{\fdi(o_1), \dots, \fdi(o_n); v_t} \in \hpfuns(st)(\pi, n) \}
\]
The restrictions of section \ref{h_funs} guarantee that
$\hpfuns(st)(\pi,n) \in \cvrel(n)$, which implies that for every
$\tup{\fdi(o_1), \dots, \fdi(o_n); v_t} \in \hpfuns(st)(\pi, n)$,
$f_D(v_t) \in \periods$. Hence, $\fpfuns(st)(\pi,n)(o_1, \dots, o_n)$ is a
set of periods as wanted. 

As discussed in section \ref{h_funs}, if $\pi(\tau_1,
\dots, \tau_n)$ represents some situation, and $\tau_1, \dots, \tau_n$
denote $o_1, \dots, o_n$, then $\hpfuns(st)(\pi, n)$ contains $\tup{\fdi(o_1),
\dots, \fdi(o_n); v_t}$ iff $f_D(v_t)$ is a maximal period
where the situation of $\pi(\tau_1, \dots, \tau_n)$ holds.
$\fpfuns(st)(\pi, n)(o_1, \dots, o_n)$ is supposed to be the set of
the maximal periods where the situation of $\pi(\tau_1, \dots, \tau_n)$
holds. The definition of \fpfuns above achieves this.

According to section \ref{top_model} and the revisions of section
\ref{linking_model}, it must also be the case that:
\[
\text{if } p_1, p_2 \in \fpfuns(st)(\pi, n)(o_1,\dots,o_n) \text{ and }
p_1 \union p_2 \in \periods, \text{ then } p_1 = p_2
\]
\fpfuns, as defined above, has this property. The proof follows.
Let us assume that $p_1$ and $p_2$ are as above, but $p_1 \not=
p_2$. As discussed above, the assumption that $p_1, p_2 \in
\fpfuns(st)(\pi, n)(o_1,\dots,o_n)$ implies that $p_1, p_2 \in
\periods$. 

Let $v_t^1 = \fdi(p_1)$ and $v_t^2 = \fdi(p_2)$ (i.e.\ $p_1 =
f_D(v_t^1)$ and $p_2 = f_D(v_t^2)$). Since, $p_1 \not= p_2$ and \fdi
is 1-1 (section \ref{relational}), $\fdi(p_1) \not= \fdi(p_2)$, i.e.
$v_t^1 \not= v_t^2$. The definition of $\fpfuns(st)(\pi,
n)(o_1,\dots,o_n)$, the assumptions that $p_1, p_2 \in
\fpfuns(st)(\pi, n)(o_1,\dots,o_n)$ and that $p_1 \union p_2 \in
\periods$, and the fact that $p_1 = f_D(v_t^1)$ and $p_2 = f_D(v_t^2)$
imply that $\hpfuns(st)(\pi,n)$ contains the value-equivalent tuples
$\tup{\fdi(o_1), \dots, \fdi(o_n); v_t^1}$ and $\tup{\fdi(o_1), \dots,
  \fdi(o_n); v_t^2}$, where $f_D(v_t^1) \union f_D(v_t^2) \in
\periods$. This conclusion, the fact that $\hpfuns(st)(\pi,n) \in
\cvrel(n)$ (see previous paragraphs), and the definition of
$\cvrel(n)$ (section \ref{bcdm}) imply that $v_t^1 = v^t_2$, which is
against the hypothesis. Hence, it cannot be the case that $p_1 \not=
p_2$, i.e.\ $p_1 = p_2$. \qed

\paragraph{$\mathbf{f_{culms}}$:} 
\index{fculms@$\fculms()$ (shows if the situation of a predicate reaches its climax)}
According to section \ref{top_model} and the revisions of section
\ref{linking_model}, for every $st \in \pts$, $\fculms(st)$ must be a
function:  
\[ \pfuns \times \{1,2,3,\dots\} \mapsto
   ((\objs)^n \mapsto \{T, F\})
\]
For every $\pi \in \pfuns$, $n \in \{1,2,3,\dots\}$, and 
$o_1, \dots, o_n \in \objs$, I define:
\[
\fculms(\pi, n)(o_1, \dots, o_n) \defeq
\begin{cases}
T, & \text{if } \tup{\fdi(o_1), \dots, \fdi(o_n)} \in \hculms(st)(\pi, n) \\
F, & \text{otherwise}
\end{cases}
\]
The restrictions of section \ref{h_funs}, guarantee that $\hculms(st)(\pi,
n) \in \srel(n)$. As discussed in section \ref{h_funs}, if a
predicate $\pi(\tau_1, \dots, \tau_n)$ represents some situation with
an inherent climax, and $\tau_1$, \dots, $\tau_n$
denote $o_1$, \dots, $o_n$, then $\hculms(st)(\pi, n)$
contains $\tup{\fdi(o_1), \dots, \fdi(o_n)}$ iff the situation
reaches its climax at the end of the latest maximal period where
the situation is ongoing. $\fculms(st)(\pi, n)(o_1, \dots, o_n)$
is supposed to be $T$ iff the situation of $\pi(\tau_1, \dots,
\tau_n)$ reaches its climax at the end of the latest maximal
period where it is ongoing. The definition of \fculms above achieves
this. 

\paragraph{$\mathbf{f_{gparts}}$:} 
\index{fgparts@$\fgparts()$ (assigns gappy partitionings to elements of \gparts)}
For every $st \in \pts$ and $\sigma_g \in \gparts$,
$\fgparts(st)(\sigma_g) \defeq \{f_D(v) \mid 
\tup{v} \in \hgparts(st)(\sigma_g)\}$. The restrictions on \hgparts
of section \ref{h_funs} guarantee that $\fgparts(st)(\sigma_g)$ is always
a gappy partitioning, as required by section \ref{top_model} and the
revisions of section \ref{linking_model}. 

\paragraph{$\mathbf{f_{cparts}}$:} 
\index{fcparts@$\fcparts()$ (assigns complete partitionings to elements of \cparts)}
For every $st \in \pts$ and $\sigma_c \in \cparts$,
$\fcparts(st)(\sigma_c) \defeq
   \{f_D(v) \mid \tup{v} \in \hcparts(st)(\sigma_c)\}$.
The restrictions on \hcparts of section \ref{h_funs} guarantee that
$\fcparts(st)(\sigma_c)$ is always a complete
partitioning, as required by section \ref{top_model} and the
revisions of section \ref{linking_model}. 


\section{The $h'$ functions} \label{via_TSQL2} 

I now discuss $\hconsp$, $\hpfunsp$, $\hculmsp$,
$\hgpartsp$, and $\hcpartsp$, the functions that map basic \topl
expressions (constants, predicates, etc.) to \tsql expressions. I
assume that these functions are defined by the configurer of 
the \nlitdb (section \ref{domain_config}). 

\paragraph{$\mathbf{h_{cons}'}$:} 
\index{hconsp@$\hconsp()$ (similar to \hcons but returns \tsql expressions)}
$\hconsp$ maps every \topl constant $\kappa$ to a \tsql value
expression $\xi$, such that $\fcn(\xi) = \emptyset$, and for every $st
\in \chrons$, $eval(st, \xi) \in D$. (The latter guarantees that
$eval(st, \xi) \not= error$.) $\xi$ is intended to represent the same world
object as $\kappa$. For example, $\hconsp$ could map the \topl
constant $sales\_department$ to the \tsql value expression \sql{'Sales
Department'}, and the \topl constant $yesterday$ to \sql{PERIOD
'today' - INTERVAL '1' DAY}. In practice, the values of \hconsp need
to be defined only for \topl constants that are used in the particular
application domain. The values of \hconsp for other constants are not
used, and can be chosen arbitrarily. Similar comments apply to \hpfunsp,
\hculmsp, \hgpartsp, and \hcpartsp.

$\hcons$ is defined in terms of $\hconsp$. For every
$st \in \chrons$ and $\kappa \in \cons$:
\index{hcons@$\hcons()$ (\topl constants to attribute values)}
\[
\hcons(st)(\kappa) \defeq eval(st, \hconsp(\kappa))
\]
The restrictions above guarantee that $eval(st, \hconsp(\kappa)) \in
D$. Hence, $\hcons(st)$ is a function $\cons \mapsto D$, as required by
section \ref{h_funs}.
 
\paragraph{$\mathbf{h_{pfuns}'}$:} 
\index{hpfunsp@$\hpfunsp()$ (similar to \hpfuns but returns \tsql expressions)}
$\hpfunsp$ is a function that maps every $\pi \in \pfuns$ and $n \in
\{1,2,3,\dots\}$ to a \tsql \sql{SELECT} statement $\Sigma$, such that
$\fcn(\Sigma) = \emptyset$, and for every $st \in
\chrons$, $eval(st, \Sigma) \in \cvrel(n)$.
$\hpfunsp(\pi, n)$ is intended to be a \tsql \sql{SELECT} statement
that generates the relation to which $\hpfuns(st)$ maps $\pi$ and $n$
(the relation that shows for which arguments and at which maximal
periods the situation described by $\pi(\tau_1, \dots,
\tau_n)$ is true).

\hpfuns is defined in terms of $\hpfunsp$. For
every $st \in \chrons$, $\pi \in \pfuns$, and $n \in \{1,2,3,\dots\}$:
\index{hpfuns@$\hpfuns()$ (predicates to relations showing maximal periods of situations)}
\[
\hpfuns(st)(\pi, n) \defeq eval(st, \hpfunsp(\pi, n))
\]
The restrictions on $\hpfunsp$ above guarantee that $eval(st,
\hpfunsp(\pi, n)) \in \cvrel(n)$. Hence, $\hpfuns(st)(\pi, n) \in
\cvrel(n)$, as required by section \ref{h_funs}. 

Let us assume, for example, that $manager(\tau)$ means that $\tau$ is
a manager, and that $manager\_of$ is the relation of $\cvrel(2)$ in
\pref{hpfuns:99a} that shows the maximal periods where somebody is the
manager of a department. (To save space, I often omit the names of the
explicit attributes. These are not needed, since explicit attributes
are referred to by number.)
\begin{examps}
\item \label{hpfuns:99a}
\dbtableb{|l|l||l|}
{$J.Adams$ & $sales$     & $[1/5/93, \; 31/12/94]$ \\
 $J.Adams$ & $personnel$ & $[1/1/95, \; 31/3/95]$ \\
 $J.Adams$ & $research$  & $[5/9/95, \; 31/12/95]$ \\
 $T.Smith$ & $sales$     & $[1/1/95, \; 7/5/95]$ \\
 \ \dots   & \ \dots     & \ \dots
}
\end{examps}
$\hpfunsp(manager, 1)$ could be defined to be 
\pref{hpfuns:2}, which generates \pref{hpfuns:3} (\pref{hpfuns:3} is
an element of $\cvrel(1)$, as required by the definition of
$\hpfunsp$). The embedded \sql{SELECT} statement of  \pref{hpfuns:2}
discards the second explicit attribute of $manager\_of$. The \sql{(PERIOD)}
coalesces tuples that correspond to the same employees (e.g.\ the
three periods for J.Adams), generating one tuple for each maximal
period. 
\begin{examps}
\item \select{SELECT DISTINCT mgr2.1 \\
              VALID VALID(mgr2) \\
              FROM (\select{SELECT DISTINCT mgr1.1 \\   
                            VALID VALID(mgr1) \\
                            FROM manager\_of AS mgr1} \\
              \ \ \ \ \ )(PERIOD) AS mgr2}
   \label{hpfuns:2}
\item 
\dbtableb{|l||l|}
{$J.Adams$ & $[1/5/93, \; 31/3/95]$ \\
 $J.Adams$ & $[5/9/95, \; 31/12/95]$ \\
 $T.Smith$ & $[1/1/95, \; 7/5/95]$ \\
 \ \dots   & \ \dots   
}
\label{hpfuns:3}
\end{examps}

\paragraph{$\mathbf{h_{culms}'}$:} 
\index{hculmsp@$\hculmsp()$ (similar to \hculmsp but returns \topl expressions)}
\hculmsp is a function that maps 
every $\pi \in \pfuns$ and $n \in \{1,2,3,\dots\}$ to a \tsql
\sql{SELECT} statement $\Sigma$, such that $\fcn(\Sigma) = \emptyset$,
and for every $st \in \chrons$, $eval(st, \Sigma) \in \srel(n)$.
$\hculmsp(\pi, n)$ is intended to be a \tsql \sql{SELECT} statement
that generates the relation to which $\hculms(st)$ maps $\pi$ and $n$
(the relation that shows for which arguments of 
$\pi(\tau_1, \dots, \tau_n)$ the situation of the predicate
reaches its climax at the end of the latest maximal period where
it is ongoing).

\hculms is defined in terms of $\hculmsp$. For
every $st \in \chrons$, $\pi \in \pfuns$, and $n \in \{1,2,3,\dots\}$:
\index{hculms@$\hculms()$ (predicates to relations showing if situations reach their climaxes)}
\[
\hculms(st)(\pi, n) \defeq eval(st, \hculmsp(\pi, n))
\]
The restrictions on $\hculmsp$ above guarantee that $eval(st,
\hculmsp(\pi, n)) \in \srel(n)$. Hence, for every $\pi \in \pfuns$
and $n \in \{1,2,3,\dots\}$, $\hculms(st)(\pi, n) \in \srel(n)$, as
required by section \ref{h_funs}. 

In the airport application, for example, $inspecting(\tau_1,
\tau_2, \tau_3)$ means that an occurrence $\tau_1$ of an inspection of
$\tau_3$ by $\tau_2$ is ongoing. $inspections$ is a relation of the
following form:
\adbtable{5}{|l|l|l|l||l|}{$inspections$}
{$code$ & $inspector$ & $inspected$ & $status$ &}
{$i158$ & $J.Adams$ & $UK160$ & $complete$ & 
      $[9\text{:}00am \; 1/5/95 - 9\text{:}45am \; 1/5/95]$ \\
 &&&& $\;\; \union \; 
       [10\text{:}10am \; 1/5/95 - 10\text{:}25am \; 1/5/95]$ \\
 $i160$ & $J.Adams$ & $UK160$ & $incomplete$  & 
      $[11\text{:}00pm \; 2/7/95 - 1\text{:}00am \; 3/7/95]$ \\
 &&&& $\;\; \union \; 
       [6\text{:}00am \; 3/7/95 - 6\text{:}20am \; 3/7/95]$ \\
 $i205$ & $T.Smith$ & $BA737$ & $complete$  & 
      $[8\text{:}00am \; 16/11/95 - 8\text{:}20am \; 16/11/95]$ \\
 $i214$ & $T.Smith$ & $BA737$ & $incomplete$  & 
      $[8\text{:}10am \; 14/2/96 - now]$
}
The first tuple above shows that J.Adams started to inspect UK160 at
9:00am on 1/5/95, and continued the inspection up to 9:45am. He
resumed the inspection at 10:10am, and completed the inspection at
10:25am on the same day. $status$ shows whether
or not the inspection reaches its completion at the last time-point of
the time-stamp. In the first tuple, its value is $complete$,
signaling that the inspection was completed at 10:25am on 1/5/95. The
inspection of the second tuple was ongoing from 11:00pm on 2/7/95 to
1:00am on 3/7/95, and from 6:00am to 6:20am on 3/7/95. It did not
reach its completion at 6:20am on 3/7/95 (perhaps it was aborted for
ever). The inspection of the last tuple started at 8:10am on 14/2/96
and is still ongoing. Each inspection is assigned a unique inspection
code, stored as the value of the $code$ attribute. The
inspection codes are useful to distinguish, for example, J.Adams'
inspection of UK160 on 1/5/95 from that on 2-3/7/95 (section
\ref{occurrence_ids}). $\hpfunsp(inspecting, 3)$ and
$\hculmsp(inspecting, 3)$ are defined to be \pref{hpfuns:4} and
\pref{hpfuns:5} respectively.
\begin{examps}
\item \select{SELECT DISTINCT insp.1, insp.2, insp.3 \\
              VALID VALID(insp) \\
              FROM inspections(PERIOD) AS insp} \label{hpfuns:4}
\item \select{SELECT DISTINCT SNAPSHOT inspcmpl.1, inspcmpl.2, 
                                       inspcmpl.3\\ 
              FROM inspections AS inspcmpl \\
              WHERE inspcmpl.4 = 'complete'}\label{hpfuns:5}
\end{examps}
This causes $\hpfuns(st)(inspecting, 2)$ and $\hculms(st)(inspecting, 2)$
to be \pref{hpfuns:6} and \pref{hpfuns:7} respectively. 
\begin{examps}
\item 
\dbtableb{|l|l|l||l|}
{$i158$ & $J.Adams$ & $UK160$ &  
      $[9\text{:}00am \; 1/5/95, \; 9\text{:}45am \; 1/5/95]$ \\
 $i158$ & $J.Adams$ & $UK160$ & 
      $[10\text{:}10am \; 1/5/95, \; 10\text{:}25am \; 1/5/95]$ \\
 $i160$ & $J.Adams$ & $UK160$ & 
      $[11\text{:}00pm \; 2/7/95, \; 1\text{:}00am \; 3/7/95]$ \\
 $i160$ & $J.Adams$ & $UK160$ & 
      $[6\text{:}00am \; 3/7/95, \; 6\text{:}20am \; 3/7/95]$ \\
 $i205$ & $T.Smith$ & $BA737$ & 
      $[8\text{:}00am \; 16/11/95, \; 8\text{:}20am \; 16/11/95]$ \\
 $i214$ & $T.Smith$ & $BA737$ & 
      $[8\text{:}10am \; 14/2/96, \; now]$
} \label{hpfuns:6}
\item 
\dbtableb{|l|l|l|}
{$i158$ & $J.Adams$ & $UK160$ \\
 $i205$ & $T.Smith$ & $BA737$ 
} \label{hpfuns:7}
\end{examps}

\paragraph{$\mathbf{h_{gparts}'}$:} 
\index{hgpartsp@$\hgpartsp()$ (similar to \hgparts but returns \tsql expressions)}
\hgpartsp is a function that maps
every \topl gappy partitioning name $\sigma_g$ to a \tsql \sql{SELECT}
statement $\Sigma$, such that $\fcn(\Sigma) = \emptyset$, and for
every $st \in \chrons$, it is true that $eval(st, \Sigma) \in
\srel(1)$ and $\{f_D(v) \mid \tup{v} \in eval(st, \Sigma)\}$ is a
gappy partitioning. $\hgpartsp(\sigma_g)$ is intended to generate the
relation to which $\hgparts(st)$ maps $\sigma_g$ (the relation
that represents the members of the gappy partitioning).
Assuming, for example, that the $gregorian$ calendric relation of
section \ref{calrels} is available, $\hgpartsp(sunday^g)$ could be 
\pref{calrels:5} of page \pageref{calrels:5}. 

\hgparts is defined in terms of \hgpartsp. For every
$st \in \chrons$ and $\sigma_g \in \gparts$:
\index{hgparts@$\hgparts()$ (gappy part.\ names to relations representing gappy partitionings)}
\[
\hgparts(st)(\sigma_g) \defeq eval(st, \hgpartsp(\sigma_g))
\]
The restrictions on \hgpartsp and the definition of $\hgparts(st)$
above satisfy the requirements on \hgparts of section \ref{h_funs}.

\paragraph{$\mathbf{h_{cparts}'}$:} 
\index{hcpartsp@$\hcpartsp()$ (similar to \hcparts but returns \tsql expressions)}
I assume that for each complete partitioning used in the \topl
formulae, there is a corresponding \tsql granularity (section
\ref{tsql2_time}). \hcpartsp is a function that maps each \topl
complete partitioning name to an ordered pair $\tup{\gamma, \Sigma}$,
where $\gamma$ is the name of the corresponding \tsql granularity, and
$\Sigma$ is a \sql{SELECT} statement that returns a relation representing the
periods of the partitioning. More precisely, it must be the case that
$\fcn(\Sigma) = \emptyset$, and for every $st \in \chrons$, $eval(st,
\Sigma) \in \srel(1)$ and $\{f_D(v) \mid
\tup{v} \in eval(st, \Sigma)\}$ is a complete partitioning. 
For example, if the $gregorian$ relation of section
\ref{calrels} is available, \hcpartsp could map $day^c$ to 
$\langle$\sql{DAY}$,\Sigma\rangle$, where $\Sigma$ is
\pref{hpfuns:8.87}. \pref{hpfuns:8.87} returns a one-attribute
snapshot relation whose attribute values denote all the day-periods.
\begin{examps}
\item \label{hpfuns:8.87}
\select{SELECT DISTINCT SNAPSHOT VALID(greg2) \\
        FROM (\select{SELECT DISTINCT greg1.4 \\
                      VALID VALID(greg1) \\
                      FROM gregorian AS greg1} \\
        \ \ \ \ \ )(PERIOD) AS greg2}
\end{examps}

\hcparts is defined in terms of \hcpartsp. For every
$st \in \chrons$ and $\sigma_c \in \cparts$, if $\hcpartsp(\sigma_c) =
\tup{\gamma, \Sigma}$, then:
\index{hcparts@$\hcparts()$ (compl.\ part.\ names to relations representing compl.\ partitionings)}
\[
\hcparts(st)(\sigma_c) = eval(st, \Sigma)
\]
The restrictions on \hcpartsp and the
definition of $\hcparts(st)$ above satisfy the requirements on \hcparts of
section \ref{h_funs}. The $\gamma$ is used in the translation rule
for $\for[\sigma_c, \nu_{qty}, \phi]$ (appendix \ref{trans_proofs}).


\section{Formulation of the translation problem} \label{formulation}

Let us now specify formally what we want the 
\topl to \tsql translation to achieve. I first define $interp$
(interpretation of a resulting relation). For
every $\phi \in \forms$ and every relation $r$\/:
\index{interp@$interp()$ (interpretation of resulting relation)}
\begin{equation}
interp(r, \phi) \defeq
\begin{cases}
T, \text{ if } \phi \in \ynforms \text{ and } 
    r \not= \emptyset \\
F, \text{ if } \phi \in \ynforms \text{ and } 
    r = \emptyset \\
\{\tup{f_D(v_1), \dots, f_D(v_n)} \mid
  \tup{v_1, \dots, v_n} \in r \}, \\
 \text{\ \ \ \  if } \phi \in \whforms
\end{cases}
\label{formulation:2}
\end{equation}
Intuitively, if $\phi$ was translated to a \sql{SELECT} statement that
generated $r$, $interp(r, \phi)$ shows how to interpret $r$. 
If $\phi \in \ynforms$ (yes/no English question) and $r \not= \emptyset$,
the answer should be affirmative. If $\phi \in \ynforms$ and $r =
\emptyset$, the answer should be negative. Otherwise, if $\phi \in
\whforms$ (the English question contains interrogatives, e.g.\
\qit{Who~\dots?}, \qit{When~\dots?}), the answer should report all the
tuples of world objects $\tup{f_D(v_1), \dots, f_D(v_n)}$ represented
by tuples $\tup{v_1, \dots, v_n} \in r$.

A translation function $tr$ is needed, that
maps every $\phi \in \forms$ to a \tsql \sql{SELECT}
statement $\mathit{tr(\phi)}$, 
\index{tr@$tr()$ (\topl to \tsql mapping)}
such that for every $st \in \pts$, \pref{formulation:1sq} and
\pref{formulation:1} hold. 
\begin{gather}
\fcn(tr(\phi)) = \emptyset \label{formulation:1sq} \\ 
interp(eval(st, tr(\phi)), \phi) = \denot{M(st), st}{\phi}
\label{formulation:1}
\end{gather}
$M(st)$ must be as in section \ref{linking_model}.
As discussed in section \ref{denotation}, each (reading of
an) English question is mapped to a \topl formula $\phi$.
The answer must report $\denot{M(st), st}{\phi}$. If $\mathit{tr}$ satisfies
\pref{formulation:1}, $\denot{M(st), st}{\phi}$ can be computed
as $interp(eval(st, tr(\phi)), \phi)$, by letting the \dbms
execute $tr(\phi)$ (i.e.\ compute $eval(st, tr(\phi))$).

$\mathit{tr}$ will be defined in terms of an auxiliary
function $\mathit{trans}$. $\mathit{trans}$ is a function of two
arguments:
\index{trans@$trans()$ (auxiliary \topl to \tsql mapping)}
\[
trans(\phi, \lambda) = \Sigma
\]
where $\phi \in \forms$, $\lambda$ is a \tsql value expression, and
$\Sigma$ a \tsql \sql{SELECT} statement. A set of ``translation
rules'' (to be discussed in section \ref{trans_rules}) specifies the
$\Sigma$-values of $\mathit{trans}$. In practice, $\lambda$ always
represents a period. Intuitively, $\lambda$ corresponds to \topl's $lt$. When
$trans$ is first invoked (by calling $tr$, discussed below) to
translate a formula $\phi$, $\lambda$ is set to \sql{PERIOD(TIMESTAMP
'beginning', TIMESTAMP 'forever')} to reflect the fact that \topl's
$lt$ is initially set to \pts (see the definition of
$\denot{M,st}{\phi}$ in section \ref{denotation}). $trans$ may call
itself recursively to translate subformulae of $\phi$ (this will
become clearer in following sections). When calling $trans$
recursively, $\lambda$ may represent a period that does not cover the
whole time-axis, to reflect the fact that already encountered \topl
operators may have narrowed $lt$. 

I define $\mathit{tr}$ as follows:
\begin{equation}
tr(\phi) \defeq trans(\phi, \linit)
\label{formulation:3}
\end{equation}
where $\linit \defeq$ \sql{PERIOD (TIMESTAMP 'beginning', TIMESTAMP
'forever')}. Obviously, \linit contains no correlation names, and
hence $\fcn(\linit) = \emptyset$. This implies that $eval(st, \linit,
g^{db})$ does not depend on $g^{db}$. \linit evaluates to the element
of $D_P$ that represents the period that covers the whole time-axis,
i.e.\ for every $st \in \pts$, it is true that $eval(st, \linit) \in D_P$ and
$f_D(eval(st, \linit)) = \pts$. Therefore, lemma \ref{linit_lemma}
holds.

\begin{lemma}
\label{linit_lemma}
{\rm $\fcn(\linit) = \emptyset$, and for every $st \in \pts$,
$eval(st, \linit) \in D_P$ and $f_{D}(eval(st,\linit)) = \pts$.}
\end{lemma}

Using \pref{formulation:3}, \pref{formulation:1sq} and
\pref{formulation:1} become \pref{formulation:6} and
\pref{formulation:4} respectively. The translation rules (that specify
the values of $\mathit{trans}$ for each $\phi$ and $\lambda$) must be
defined so that for every $\phi \in \forms$ and $st \in \pts$,
\pref{formulation:6} and \pref{formulation:4} hold.
\begin{gather}
\fcn(trans(\phi, \linit)) = \emptyset \label{formulation:6} \\
interp(eval(st, trans(\phi, \linit)), \phi) = \denot{M(st), st}{\phi}
\label{formulation:4}
\end{gather}

Appendix \ref{trans_proofs} proves that theorems \ref{wh_theorem} and
\ref{yn_theorem} hold for the translation rules of this thesis. 

\begin{theorem}
\label{wh_theorem}
{\rm If $\phi \in \whforms$, $st \in \pts$, $trans(\phi,
\linit) = \Sigma$, and the total number of interrogative and
interrogative-maximal quantifiers in $\phi$ is $n$, then:
\begin{enumerate}
\item $\fcn(\Sigma) = \emptyset$ 
\item $eval(st, \Sigma) \in \srel(n)$
\item $\{\tup{f_D(v_1), \dots, f_D(v_n)} \mid
       \tup{v_1, \dots, v_n} \in eval(st, \Sigma)\} 
       = \denot{M(st), st}{\phi}$
\end{enumerate}
} 
\end{theorem}
That is, the translation $\Sigma$ of $\phi$ contains no free column
references, and it evaluates to a snapshot relation of $n$ attributes,
whose tuples represent $\denot{M(st), st}{\phi}$.

\begin{theorem}
\label{yn_theorem}
{\rm If $\phi \in \ynforms$, $st \in \pts$, $\lambda$ is a \tsql
expression, $g^{db} \in G^{db}$, $eval(st,
\lambda, g^{db}) \in D_P^*$, $\corn{\phi} = \tup{\tau_1, \dots,
\tau_n}$, and $\Sigma = trans(\phi, \lambda)$, then:
\begin{enumerate}
\item $\fcn(\Sigma) \subseteq \fcn(\lambda)$
\item $eval(st, \Sigma, g^{db}) \in \vrel(n)$
\item $\tup{v_1, \dots, v_n; v_t} \in 
       eval(st, \Sigma, g^{db})$ iff for some $g \in G$: \\
   $\denot{M(st), g}{\tau_1} = f_D(v_1)$, \dots, 
   $\denot{M(st), g}{\tau_n} = f_D(v_n)$, and \\
   $\denot{M(st), st, f_D(v_t), f_D(eval(st, \lambda, g^{db})), g}{\phi} = T$
\end{enumerate}
} 
\end{theorem}

$\tau_1, \tau_2, \dots, \tau_n$ are all the constants in
predicate argument positions and all the variables in $\phi$ (section
\ref{TOP_mods}). Clause 3 intuitively means that the tuples of
$eval(st, \Sigma, g^{db})$ represent all the possible combinations of
values of $\tau_1, \dots, \tau_n$ and event times $et$, such that
$\denot{M(st), st, et, lt, g}{\phi} = T$, where $lt$ is the element of
$\periods^*$ represented by $\lambda$.

I now prove that theorems \ref{wh_theorem} and \ref{yn_theorem} imply that
\pref{formulation:6} and \pref{formulation:4} hold for every $st \in
\pts$ and $\phi \in \forms$, i.e.\ that $trans$ has the desired properties.

\textbf{Proof of \pref{formulation:6}:} Let $st \in
\pts$ and $\phi \in \forms$. We need to show that
\pref{formulation:6} holds. Since $\forms = \whforms
\union \ynforms$, the hypothesis that $\phi \in \forms$ implies that
$\phi \in \whforms$ or $\phi \in \ynforms$. In both cases
\pref{formulation:6} holds: 
\begin{itemize}
\item If $\phi \in \whforms$, then by theorem \ref{wh_theorem},
$\fcn(trans(\phi, \linit)) = \emptyset$, i.e.\ \pref{formulation:6}
holds.
\item If $\phi \in \ynforms$, then by theorem \ref{yn_theorem} and
lemma \ref{linit_lemma}, the following holds, which implies that
\pref{formulation:6} also holds. 
\[
\fcn(trans(\phi, \linit)) \subseteq \fcn(\linit) = \emptyset
\]
\end{itemize}

\textbf{Proof of \pref{formulation:4}:} Let $st \in
\pts$ and $\phi \in \forms$. Again, it will either be the case
that $\phi \in \whforms$ or $\phi \in \ynforms$. 

If $\phi \in \whforms$, then by theorem \ref{wh_theorem} the following
is true:
\[
\{\tup{f_D(v_1), \dots, f_D(v_n)} \mid
       \tup{v_1, \dots, v_n} \in eval(st, trans(\phi, \linit))\} 
       = \denot{M(st), st}{\phi}
\]
The definition of $interp$, the hypothesis that $\phi \in \whforms$,
and the equation above imply \pref{formulation:4}.

It remains to prove \pref{formulation:4} for 
$\phi \in \ynforms$. Let $\corn{\phi} = \tup{\tau_1, \dots,\tau_n}$.
By lemma \ref{linit_lemma}, for every $g^{db} \in G^{db}$, $eval(st,
\linit, g^{db}) = eval(st, \linit) \in D_P$ and $f_D(eval(st,
\linit)) = \pts$. Also, \pref{formulation:6} (proven above) implies
that $eval(st, trans(\phi, \linit), g^{db})$ does not depend on
$g^{db}$. Then, from theorem \ref{yn_theorem} we get
\pref{formulation:10} and \pref{formulation:12}. 
\begin{examples}
\item \label{formulation:10}
$eval(st, trans(\phi, \linit)) \in \vrel(n)$
\item \label{formulation:12}
  $\tup{v_1, \dots, v_n; v_t} \in eval(st, trans(\phi, \linit))$ iff for
  some $g \in G$: \\
  $\denot{M(st), g}{\tau_1} = f_D(v_1), \dots, 
   \denot{M(st), g}{\tau_n} = f_D(v_n)$, and \\
  $\denot{M(st), st, f_D(v_t), \pts, g}{\phi} = T$
  \notag
\end{examples}
The hypothesis that $\phi \in \ynforms$ and the definition of
$\mathit{interp}$ imply that the left-hand side of
\pref{formulation:4} has the following values:
\[
\begin{cases}
T, & \text{if } eval(st, trans(\phi, \linit)) \not= \emptyset \\
F, & \text{if } eval(st, trans(\phi, \linit)) = \emptyset
\end{cases}
\]
The hypothesis that $\phi \in \ynforms$ and the definition of
$\denot{M(st), st}{\phi}$ (section \ref{denotation}) imply that the 
right-hand side of \pref{formulation:4} has the following values:
\[
\begin{cases}
T, & \text{if for some } g \in G \text{ and } et \in \periods, \;
    \denot{M(st), st, et, \pts, g}{\phi} = T \\
F, & \text{otherwise}
\end{cases}
\]
Hence, to prove \pref{formulation:4} it is enough to prove
\pref{formulation:15}. 
\begin{examples}
\item \label{formulation:15}
$eval(st, trans(\phi, \linit)) \not= \emptyset$ iff \\
for some $g \in G$ and $et  \in \periods$,
$\denot{M(st), st, et, \pts, g}{\phi} = T$
\end{examples}

I first prove the forward direction of \pref{formulation:15}. If
it is true that $eval(st, trans(\phi, \linit)) \not= \emptyset$, 
by \pref{formulation:10} $eval(st, trans(\phi, \linit))$ contains at least
a tuple of the form $\tup{v_1, \dots, v_n; v_t}$, i.e.\
\pref{formulation:20} is true.
\begin{equation}
\label{formulation:20}
\tup{v_1, \dots, v_n; v_t} \in eval(st, trans(\phi, \linit))
\end{equation}
\pref{formulation:20} and \pref{formulation:12} imply that for some $g
\in G$, \pref{formulation:21} holds.
\begin{equation}
\label{formulation:21}
\denot{M(st), st, f_D(v_t), \pts, g}{\phi} = T
\end{equation}
\pref{formulation:10} and \pref{formulation:20} imply that $v_t$ is
the time-stamp of a tuple in a relation of \vrel, which implies that
$f_D(v_t) \in \periods$. Let $et = f_D(v_t)$. Then,
\pref{formulation:21} becomes \pref{formulation:22}, where $g \in G$ 
and $et = f_D(v_t) \in \periods$. The forward direction of
\pref{formulation:15} has been proven.
\begin{equation}
\label{formulation:22}
\denot{M(st), st, et, \pts, g}{\phi} = T
\end{equation}

I now prove the backwards direction of \pref{formulation:15}. I 
assume that $g \in G$, $et \in \periods$, and $\denot{M(st),
st, et, \pts, g}{\phi} = T$. Let $v_t = \fdi(et)$, which implies that
$et = f_D(v_t)$. Then \pref{formulation:23} holds. 
\begin{equation}
\label{formulation:23}
\denot{M(st), st, f_D(v_t), \pts, g}{\phi} = T
\end{equation}
Let $v_1 = \fdi(\denot{M(st),g}{\tau_1})$, \dots, $v_n =
\fdi(\denot{M(st),g}{\tau_n})$. This implies that
\pref{formulation:24} also holds.
\begin{equation}
\label{formulation:24}
\denot{M(st), g}{\tau_1} = f_D(v_1), \; \dots, \; \denot{M(st),
g}{\tau_n} = f_D(v_n) 
\end{equation}
\pref{formulation:24}, \pref{formulation:23}, the hypothesis that $g
\in G$, and \pref{formulation:12} imply \pref{formulation:25}, which
in turn implies that $eval(st, trans(\phi, \linit)) \not= \emptyset$. The
backwards direction of \pref{formulation:15} has been proven.
\begin{equation}
\label{formulation:25}
\tup{v_1, \dots, v_n; v_t} \in eval(st, trans(\phi, \linit))
\end{equation}

This concludes the proof of \pref{formulation:4}. I have proven that
$trans$ satisfies \pref{formulation:6} and \pref{formulation:4} for
every $\phi \in \forms$ and $st \in \pts$, i.e.\ that $trans$ has all the 
desired properties. 


\section{The translation rules} \label{trans_rules}

The values (\sql{SELECT} statements) of $trans$ are specified by a set
of ``translation rules''. These rules are of two kinds: (a) base
(non-recursive) rules that specify $trans(\phi, \lambda)$ when $\phi$
is an atomic formula or a formula of the form $\culm[\pi(\tau_1,
\dots, \tau_n)]$; and (b) recursive rules that specify 
$trans(\phi, \lambda)$ in all other cases, by recursively calling
other translation rules to translate subformulae of $\phi$. In this
section, I attempt to convey the intuitions behind the design of the
translation rules, and to illustrate the functionality of some
representative rules. 

In the case of a yes/no formula $\phi$, the aim is for the resulting
\sql{SELECT} statement to return a relation of $\vrel(n)$ that shows all
the combinations of event-times $et$ and values of $\tau_1,
\dots, \tau_n$ ($\tup{\tau_1, \dots, \tau_n} = \corn{\phi}$)
for which $\phi$ is satisfied. More precisely, the tuples of the
relation must represent all the combinations of event times $et$ and
world objects assigned (by $\fcons(st)$ and some variable assignment
$g$) to $\tau_1, \dots, \tau_n$, for which $\denot{M(st), st, et, lt,
g}{\phi} = T$, where $lt$ is the element of $\periods^*$ represented
by $\lambda$.  In each tuple $\tup{v_1, \dots, v_n; v_t}$, $v_t$
represents $et$, while $v_1, \dots, v_n$ represent the world objects
of $\tau_1, \dots, \tau_n$. For example, the rule for predicates is as
follows:

\textbf{Translation rule for predicates:} \\
$trans(\pi(\tau_1, \dots, \tau_n), \lambda) \defeq$\\
\sql{(}\select{SELECT DISTINCT $\alpha.1$, $\alpha.2$, \dots, $\alpha.n$ \\
               VALID VALID($\alpha$) \\
               FROM ($\hpfunsp(\pi, n)$)(SUBPERIOD) AS $\alpha$ \\
               WHERE \dots \\
               \ \ AND \dots \\
               \ \ \vdots \\
               \ \ AND \dots \\
               \ \ AND $\lambda$ CONTAINS VALID($\alpha$))} 

where the ``\dots''s in the \sql{WHERE} clause stand for all the
strings in $S_1 \union S_2$, and: 
\begin{gather*}
S_1 = 
\{\text{``}\alpha.i = \hconsp(\tau_i)\text{''} \mid
  i \in \{1,2,3,\dots,n\} \text{ and } \tau_i \in \cons\} \\
S_2 = 
\{\text{``}\alpha.i = \alpha.j\text{''} \mid
  i,j \in \{1,2,3,\dots,n\}, \; i < j, \; \tau_i = \tau_j, \text{ and }
  \tau_i, \tau_j \in \vars\}
\end{gather*}

I assume that whenever the
translation rule is invoked, a new correlation name $\alpha$ is used,
that is obtained by calling a 
\emph{generator of correlation names}. Whenever called,
the generator returns a new correlation name that has never
been generated before. I assume that the correlation names
of the generator are of some distinctive form (e.g.\
\sql{t1}, \sql{t2}, \sql{t3},~\dots), and that the correlation
names in the \sql{SELECT} statements returned by \hpfunsp,
\hculmsp, \hcpartsp, and \hgpartsp are not of this
distinctive form. I also assume that some mechanism is in
place to ensure that no correlation name of the distinctive form of
the generator can be used before it has been generated. 
The use of the generator means that $\mathit{trans}$ is strictly
speaking not a pure function, since the same $\pi$ and $\tau_1, \dots,
\tau_n$ lead to slightly different \sql{SELECT} statements whenever
$trans(\pi(\tau_1, \dots, \tau_n), \lambda)$ is computed: each time
the resulting statement contains a different $\alpha$ (similar
comments apply to other translation
rules). There are ways to make $\mathit{trans}$ a pure function, but
these complicate the translation rules and
the proof of their correctness, without offering any practical
advantage.

Let us consider, for example, the predicate $inspecting(i158,
j\_adams, uk160)$. According to section \ref{denotation},
$\denot{M(st), st, et, lt, g}{inspecting(i158, j\_adams, uk160)} = T$ iff $et \subper lt$ and $et \subper p$,
where:
\[
p \in \fpfuns(st)(inspecting, 3)(\denot{M(st),g}{i158},
\denot{M(st),g}{j\_adams}, \denot{M(st),g}{uk160})
\]
Let us assume that $\hpfunsp(inspecting, 3)$ and
$\hpfuns(st)(inspecting, 3)$ are \pref{hpfuns:4} and \pref{hpfuns:6}
respectively (p.~\pageref{hpfuns:4}), that $i158$, $j\_adams$, and
$uk160$ correspond to the obvious attribute values of \pref{hpfuns:6},
and that $\lambda$ is \sql{PERIOD '[9:00am 1/5/95 - 9:30pm 1/5/95]'}.
$lt$ is the period represented by $\lambda$. By the definition of
\fpfuns of section \ref{resulting_model}:
\[
\fpfuns(st)(inspecting,3)(\denot{M(st),g}{i158},
\denot{M(st),g}{j\_adams}, \denot{M(st),g}{uk160}) = \{p_1, p_2\}
\]
where $p_1$ and $p_2$ are the periods of the first two
tuples of \pref{hpfuns:6}. The
denotation of $inspecting(i158, j\_adams, uk160)$ is $T$ for all
the $et$s that are subperiods of $p_1$ or $p_2$ and also
subperiods of $lt$. 

The translation rule above maps $inspecting(i158, j\_adams,
uk160)$ to \pref{trans:1}, where $\hpfunsp(inspecting, 3)$ is the
\sql{SELECT} statement of \pref{hpfuns:4} (that returns \pref{hpfuns:6}).
\begin{examps}
\item \label{trans:1}
\sql{(}\select{SELECT DISTINCT t1.1, t1.2, t1.3 \\
               VALID VALID(t1) \\
               FROM ($\hpfunsp(inspecting, 3)$)(SUBPERIOD) AS t1 \\
               WHERE t1.1 = 'i158' \\
               \ \ AND t1.2 = 'J.Adams' \\
               \ \ AND t1.3 = 'UK160' \\
               \ \ AND PERIOD '[9:00am 1/5/95 - 9:30pm 1/5/95]' 
                       CONTAINS VALID(t1))} 
\end{examps}
\pref{trans:1} returns \pref{trans:2}, where the time-stamps
correspond to all the subperiods of $p_1$ and $p_2$ ($p_1$ and $p_2$
are the periods of the first two time-stamps of
\pref{hpfuns:6}) that are also subperiods of $lt$ (the period
represented by $\lambda$). 
\begin{examps}
\item 
\dbtableb{|l|l|l||l|}
{$i158$ & $J.Adams$ & $UK160$ & $[9\text{:}00am \; 1/5/95, \;
9\text{:}30pm \; 1/5/95]$ \\
 $i158$ & $J.Adams$ & $UK160$ & $[9\text{:}10am \; 1/5/95, \; 9\text{:}15pm \; 1/5/95]$ \\
 $i158$ & $J.Adams$ & $UK160$ & $[9\text{:}20am \; 1/5/95, \; 9\text{:}25pm \; 1/5/95]$ \\
 \ \dots & \ \dots & \ \dots & \ \dots 
}
\label{trans:2}
\end{examps}
In other words, the time-stamps of \pref{trans:2} represent correctly
all the $et$s where the denotation of $inspecting(i158, j\_adams,
uk160)$ is $T$. In this example, all the predicate arguments are
constants. Hence, there can be no variation in the values of the
arguments, and the values of the explicit attributes in \pref{trans:2}
are the same in all the tuples. When some of the predicate arguments
are variables, the values of the corresponding explicit attributes are
not necessarily fixed. 

The $S_2$ constraints in the \sql{WHERE} clause of the translation
rule are needed when the predicate contains the same
variable in more than one argument positions. In those cases, 
$S_2$ requires the attributes that correspond to the
argument positions where the variable appears to have the same
values. $S_2$ contains redundant
constraints when some variable appears in more than two
argument positions. For example, in $\pi(\beta,
\beta, \beta)$ ($\beta \in \vars$), $S_2$ requires the tuples
$\tup{v_1, v_2, v_3; v_t}$ of the resulting relation to satisfy:
$v_1 = v_2$, $v_1 = v_3$, and $v_2 = v_3$. The third
constraint is redundant, because it follows from the others. The
prototype \nlitdb employs a slightly more complex
definition of $S_2$ that does not generate the third
constraint. Similar comments apply to the rule for
$\culm[\pi(\tau_1, \dots, \tau_n)]$ below, and the rules
for conjunction, $\at[\phi_1, \phi_2]$, $\before[\phi_1, \phi_2]$, and
$\after[\phi_1, \phi_2]$ (appendix \ref{trans_proofs}).

\textbf{Translation rule for $\culm[\pi(\tau_1, \dots, \tau_n)]$:}\\
$trans(\culm[\pi(\tau_1, \dots, \tau_n)], \lambda) \defeq$\\
\sql{(}\select{SELECT DISTINCT 
                      $\alpha_1.1$, $\alpha_1.2$, \dots, $\alpha_1.n$ \\
               VALID PERIOD(BEGIN(VALID($\alpha_1$)),
                            END(VALID($\alpha_1$))) \\
               FROM ($\hpfunsp(\pi, n)$)(ELEMENT) AS $\alpha_1$, \\
               \ \ \ \ \ ($\hculmsp(\pi, n)$) AS $\alpha_2$ \\
               WHERE $\alpha_1.1 = \alpha_2.1$ \\
               \ \ AND $\alpha_1.2 = \alpha_2.2$ \\
               \ \ \ \ \vdots \\
               \ \ AND $\alpha_1.n = \alpha_2.n$ \\ 
               \ \ AND \dots \\
               \ \ \ \ \vdots \\
               \ \ AND \dots \\
               \ \ AND $\lambda$ CONTAINS 
                       PERIOD(BEGIN(VALID($\alpha_1$)),
                              END(VALID($\alpha_1$)))}

Whenever the rule is used, $\alpha_1$ and $\alpha_2$ are
two new different correlation names, obtained by calling the
correlation names generator after $\lambda$ has been supplied. The
``\dots'' in the \sql{WHERE} clause stand for all the strings in
$S_1 \union S_2$, where $S_1$ and $S_2$ are as in the translation rule
for predicates, except that $\alpha$ is now $\alpha_1$. 

The rule for $\culm[\pi(\tau_1, \dots, \tau_n)]$ is
similar to that for $\pi(\tau_1, \dots, \tau_n)$. The resulting
\sql{SELECT} statement returns an element of $\vrel(n)$ that shows 
the $et$s and the values of the predicate arguments for which the
denotation of $\culm[\pi(\tau_1, \dots, \tau_n)]$ is $T$. In the case
of $\culm[\pi(\tau_1, \dots, \tau_n)]$, however, the generated
relation contains only tuples $\tup{v_1, \dots, v_n; v_t}$, for which
$\tup{v_1, \dots, v_n}$ appears in $\hculms(st)(\pi, n)$ (the relation
returned by $\hculmsp(\pi, n)$). That is, the situation of
$\pi(\tau_1, \dots, \tau_n)$ must reach its climax at the latest
time-point where it is ongoing. Also, $\hpfuns(st)(\pi, n)$
(the relation returned by $\hpfunsp(\pi,n)$) is coalesced using 
\sql{(ELEMENT)}. This causes all tuples of $\hpfuns(st)(\pi, n)$
that refer to the same situation to be merged into one tuple,
time-stamped by a temporal element that is the union of all the
periods where the situation is ongoing. Let us refer to this coalesced
version of $\hpfuns(st)(\pi, n)$ as $r$. $\alpha_1$ ranges
over the tuples of $r$, while $\alpha_2$ over the tuples of
$\hculms(st)(\pi,n)$. The relation returned by
$trans(\culm[\pi(\tau_1, \dots, \tau_n)], \lambda)$ contains all
tuples $\tup{v_1, \dots, v_n; v_t}$, such that $\tup{v_1, \dots, v_n;
v_t'} \in r$, $v_t$ represents the period that starts at the beginning
of the temporal element of $v_t'$ and ends at the end of the temporal
element of $v_t'$, $\tup{v_1, \dots, v_n} \in \hculms(st)(\pi, n)$,
and $v_t$'s period (i.e.\ $et$) is a subperiod of $\lambda$'s period
(i.e.\ $lt$). $S_1$ and $S_2$ play the same role as in the translation
rule for predicates.

Let us assume that $\hpfunsp(inspecting,3)$ and
$\hculmsp(inspecting,3)$ are \pref{hpfuns:4} and \pref{hpfuns:5}
respectively, that $\hpfuns(st)(inspecting,3)$ and
$\hculms(st)(inspecting,3)$ are \pref{hpfuns:6} and
\pref{hpfuns:7}, and that $\lambda =$ \sql{PERIOD '[1/5/95 -
18/11/95]'}. The translation rule above maps $\culm[inspecting(occr^v,
person^v, flight^v)]$ to \pref{trans:5}. 
\begin{examps}
\item \label{trans:5}
\sql{(}\select{SELECT DISTINCT t1.1, t1.2, t1.3 \\
               VALID PERIOD(BEGIN(VALID(t1)),
                            END(VALID(t1))) \\
               FROM ($\hpfunsp(inspecting,3)$)(ELEMENT) AS t1, \\
               \ \ \ \ \ ($\hculmsp(inspecting,3)$) AS t2 \\
               WHERE t1.1 = t2.1 \\ 
               \ \ AND t1.2 = t2.2 \\
               \ \ AND t1.3 = t2.3 \\
               \ \ AND PERIOD '[1/5/95 - 18/11/95]' CONTAINS \\
               \ \ \ \ \ \ PERIOD(BEGIN(VALID(t1)),
                           END(VALID(t1))))}
\end{examps}
\pref{trans:5} returns \pref{trans:6}. There is (correctly) no tuple
for inspection $i160$: the semantics of \culm (section
\ref{culm_op}) requires the inspection to reach its completion at the
latest time-point where it is ongoing; according to \pref{hpfuns:7},
this is not the case for $i160$. There is also (correctly) no tuple
for $i214$: the semantics of \culm
requires $et$ (the time of the inspection) to be a
subperiod of $lt$ ($\lambda$'s period), but $i214$ does not occur
within $lt$. Finally, \pref{trans:6} does not
contain tuples for the subperiods of [9:00am 1/5/95, 10:25am
1/5/95] and [8:00am 16/11/95, 8:20am 16/11/95]. This is in accordance
with the semantics of \culm, that allows $\culm[inspecting(occr^v, j\_adams,
ba737)]$ to be true only at $et$s that cover entire inspections
(from start to completion).
\begin{examps}
\item 
\dbtableb{|l|l|l||l|}
{$i158$ & $J.Adams$ & $UK160$ & $[9\text{:}00am \; 1/5/95, \; 
10\text{:}25am \; 1/5/95]$ \\
 $i205$ & $T.Smith$ & $BA737$ & $[8\text{:}00am \; 16/11/95, \;
8\text{:}20am \; 16/11/95]$
}
\label{trans:6}
\end{examps}

All the other translation rules for yes/no formulae are recursive. For
example, $\past[\beta, \phi']$ is translated using the following:

\textbf{Translation rule for $\past[\beta, \phi']$:}\\
\label{past_trans_discuss}
$trans(\past[\beta, \phi'], \lambda) \defeq$\\
\sql{(}\select{SELECT DISTINCT VALID($\alpha$), 
                      $\alpha$.1, $\alpha$.2, \dots, $\alpha$.$n$ \\
               VALID VALID($\alpha$) \\
               FROM $trans(\phi', \lambda')$ AS $\alpha$)}

$\lambda'$ is the expression \sql{INTERSECT($\lambda$, PERIOD(TIMESTAMP
'beginning', TIMESTAMP 'now' - INTERVAL '1' $\chi$))}, $\chi$
stands for the \tsql name of the granularity of chronons (e.g.\
\sql{DAY}), and $n$ is the length of $\corn{\phi'}$. Whenever the rule
is used, $\alpha$ is a new correlation name obtained by calling the
correlation names generator. 

The rule for
$\past[\beta, \phi']$ calls recursively $\mathit{trans}$ to translate
$\phi'$. $\phi'$ is translated with
respect to $\lambda'$, which represents the intersection of
the period of the original $\lambda$ with the period that covers all
the time up to (but not including) the present chronon. This
reflects the semantics of \past (section \ref{past_op}),
that narrows $lt$ to $lt
\intersect [t_{first}, st)$.  The relation returned by
$trans(\past[\beta, \phi'], \lambda)$ is the same as that of 
$trans(\phi', \lambda')$, except that the relation of
$trans(\past[\beta, \phi'], \lambda)$ contains an additional explicit
attribute, that corresponds to the $\beta$ of $\past[\beta,\phi']$. The
values of that attribute are the same as the corresponding time-stamps
(that represent $et$). This reflects the semantics of
$\past[\beta, \phi']$, that requires the value of $\beta$ to be $et$.
As a further example, $\at[\kappa, \phi']$ ($\kappa \in \cons$) is
translated using the following:

\textbf{Translation rule for $\at[\kappa, \phi']$:}\\
$trans(\at[\kappa, \phi'], \lambda) \defeq trans(\phi', \lambda')$,
where $\lambda'$ is \sql{INTERSECT($\lambda$,
$\hconsp(\kappa)$)}.

The translation of $\at[\kappa, \phi']$ is the same as the
translation of $\phi'$, but $\phi'$ is translated with respect to
$\lambda'$, which represents the intersection of $\lambda$'s period 
with that of $\kappa$. This reflects the fact
that in $\at[\kappa, \phi']$, the \at narrows $lt$ to the
intersection of the original $lt$ with $\kappa$'s period. 
There are separate translation rules for 
$\at[\sigma_c, \beta, \phi']$, $\at[\sigma_g, \beta, \phi']$, and
$\at[\phi_1, \phi_2]$ ($\sigma_c \in \cparts$, $\sigma_g \in
\gparts$, and $\phi', \phi_1, \phi_2 \in \ynforms$). 

The complete set of translation rules for yes/no formulae is given in
appendix \ref{trans_proofs}, along with a formal proof that
$trans(\phi, \lambda)$ satisfies theorem \ref{yn_theorem}. Theorem
\ref{yn_theorem} is proven by induction on the syntactic complexity of
$\phi$. I first prove that theorem \ref{yn_theorem} holds if $\phi$ is
a predicate or $\culm[\pi(\tau_1, \dots, \tau_n)]$. For all other
$\phi \in \ynforms$, $\phi$ is non-atomic. In those cases, I prove
that theorem \ref{yn_theorem} holds if it holds for the
subformulae of $\phi$. 

Let us now consider wh-formulae. These have 
the form $?\beta_1 \; ?\beta_2 \; ?\beta_3 \dots \; ?\beta_k
\; \phi'$ or $?_{mxl}\beta_1 \; ?\beta_2 \; ?\beta_3 \;
\dots \; ?\beta_k \; \phi'$, where $\phi' \in \ynforms$ (section
\ref{top_syntax}). The first case is covered by the following rule.
(The rules for wh-formulae define $trans(\phi, \lambda)$
only for $\lambda = \linit$. The values of $\mathit{trans}$ for $\phi
\in \whforms$ and $\lambda \not= \linit$ are not used anywhere and can
be chosen arbitrarily. Intuitively, for $\phi \in \whforms$ the goal is
to define $trans(\phi, \lambda)$ so that it satisfies theorem
\ref{wh_theorem}. That theorem is indifferent to the values of
$\mathit{trans}$ for $\lambda \not= \linit$.)
 
\textbf{Translation rule for $?\beta_1 \; ?\beta_2 \; ?\beta_3 \dots \;
?\beta_k \; \phi'$:} \\
$trans(?\beta_1 \; ?\beta_2 \; ?\beta_3 \dots \;
?\beta_k \; \phi', \linit) \defeq$ \\
\sql{(}\select{SELECT DISTINCT SNAPSHOT $\alpha.\omega_1$,
                      $\alpha.\omega_2$, \dots, $\alpha.\omega_k$ \\
               FROM $trans(\phi', \linit)$ AS $\alpha$)} 

Whenever the rule is used, $\alpha$ is a new correlation
name, obtained by calling the correlation names generator. Assuming
that $\corn{\phi'} = \tup{\tau_1, \dots,
\tau_n}$, for every $i \in \{1,2,3, \dots, \kappa\}$:
\[
\omega_i = min(\{j \mid 
                 j \in \{1,2,3,\dots,n\} \text{ and } \tau_j = \beta_j\})
\]
That is, the first position (from left to right) where $\beta_i$
appears in $\tup{\tau_1, \dots, \tau_n}$ is the $\omega_i$-th one. 
Intuitively, we want $?\beta_1 \; ?\beta_2 \; ?\beta_3 \dots \;
?\beta_k \; \phi'$ to be translated to a \sql{SELECT} statement
that returns a snapshot relation, whose tuples represent
$\denot{M(st), st}{?\beta_1 \; ?\beta_2 \; ?\beta_3 \dots \; ?\beta_k
\; \phi'}$. According to section
\ref{denotation}, $\denot{M(st), st}{?\beta_1 \; ?\beta_2 \; ?\beta_3
\dots \; ?\beta_k \; \phi'}$ is the set of all tuples that represent
combinations of values assigned to $\beta_1, \dots,
\beta_k$ by some $g \in G$, such that for some $et \in \periods$,
$\denot{M(st), st, et, \pts, g}{\phi'} = T$.

By theorem \ref{yn_theorem}, the relation returned
by $trans(\phi', \linit)$ (see the translation rule) is a valid-time
relation, whose tuples show all the possible combinations of $et$s and
values assigned (by $\fcons(st)$ and some $g \in G$) to $\tau_1, \dots,
\tau_n$, for which $\denot{M(st), st, et, \pts, g}{\phi'} = T$. The
syntax of \topl (section \ref{top_syntax}) guarantees that $\beta_1, \dots,
\beta_k$ appear within $\phi'$. This in turn guarantees that $\beta_1,
\dots, \beta_k$ appear among $\tau_1, \dots, \tau_n$, i.e.\ the
relation of $trans(\phi',\linit)$ contains attributes for
$\beta_1,\dots,\beta_k$. To find
all the possible combinations of values of $\beta_1, \dots, \beta_k$
for which (for some $et$) $\denot{M(st), st, et, \pts, g}{\phi'} = T$,
we simply need to pick (to ``project'' in relational terms) from the
relation of $trans(\phi', 
\linit)$ the attributes that correspond to
$\beta_1, \dots, 
\beta_k$.  For $i \in \{1,2,3,\dots,k\}$, $\beta_i$ may appear
more than once in $\phi'$. In this case, the relation of $trans(\phi',
\linit)$ contains more than one attributes for $\beta_i$ (these
attributes have the same values in each tuple). We only need to
project one of the attributes that correspond to $\beta_i$. The
translation rule projects only the first one; this is the
$\omega_i$-th attribute of $trans(\phi', \linit)$, the attribute that
corresponds to the first (from left to right) $\tau_j$ in
$\tup{\tau_1, \dots,
\tau_n}$ that is equal to $\beta_i$.

Let us consider, for example, the following wh-formula (\qit{Who
inspected what?}):
\begin{equation}
\label{trans:10.1}
?w1^v \; ?w2^v \; \past[e^v, \culm[inspecting(occr^v, w1^v, w2^v)]]
\end{equation}
Here, $\phi' = \past[e^v, \culm[inspecting(occr^v, w1^v,
w2^v)]]$ and $\corn{\phi'} = \tup{e^v, occr^v, w1^v, w2^v}$.  Let us
assume that $trans(\phi', \linit)$ returns
\pref{trans:10}. \pref{trans:10} shows all the possible combinations
of $et$s and values that can be assigned by some $g
\in G$ to $e^v$, $occr^v$, $w1^v$, and $w2^v$, such that
$\denot{M(st), st, et, \pts, g}{\phi'} = T$. In every tuple, the
time-stamp is the same as the value of the first explicit attribute,
because the semantics of \past requires the value of $e^v$
(represented by the first explicit attribute) to be $et$ (represented
by the time-stamp). To save space, I omit the time-stamps of
\pref{trans:10}.
\begin{examps}
\item \label{trans:10}
\dbtableb{|l|l|l|l||l|}
{$[9\text{:}00am \; 1/5/95, \; 3\text{:}00pm \; 1/5/95]$ & $i158$ & $J.Adams$ & $UK160$ & \dots \\
 $[10\text{:}00am \; 4/5/95, \; 11\text{:}30am \; 4/5/95]$ & $i165$ & $J.Adams$ & $BA737$ & \dots \\
 $[7\text{:}00am \; 16/11/95, \; 7\text{:}30am \; 16/11/95]$ & $i204$ & $T.Smith$ & $UK160$ & \dots
}
\end{examps}
To generate the snapshot relation that represents $\denot{M(st),
st}{?w1^v \; ?w2^v \; \phi'}$, i.e.\ the relation that shows the
combinations of values of $w1^v$ and $w2^v$ for
which (for some $et$ and $g$) $\denot{M(st), st, et, \pts, g}{\past[e^v,
\culm[inspecting(occr^v, w1^v, w2^v)]]} = T$, we simply need to
project the explicit attributes of \pref{trans:10} that correspond to
$w1^v$ and $w2^v$. The first positions where $w1^v$ and $w2^v$ appear
in $\corn{\phi'} = \tup{e^v, occr^v, w1^v, w2^v}$ are the
third and fourth (i.e.\ $\omega_1 = 3$ and $\omega_2 =
4$). Hence, we need to project the third and fourth explicit
attributes of \pref{trans:10}. The translation rule for $?\beta_1 \;
\dots \; ?\beta_k \; \phi'$ maps \pref{trans:10.1} to \pref{trans:11},
which achieves exactly that (it returns \pref{trans:12}).
\begin{examps}
\item \label{trans:11}
\sql{(}\select{SELECT DISTINCT SNAPSHOT t1.3, t1.4 \\
               FROM $trans(\past[e^v, \culm[inspecting(occr^v, w1^v, w2^v)]],
                           \linit)$ AS t1)}
\item \label{trans:12}
\dbtableb{|l|l|}
{$J.Adams$ & $UK160$ \\
 $J.Adams$ & $BA737$ \\
 $T.Smith$ & $UK160$ 
}
\end{examps}

Wh-formulae of the form $?_{mxl}\beta_1 \; ?\beta_2 \; ?\beta_3 \;
\dots \; ?\beta_k \; \phi'$ ($\phi' \in \ynforms$) are
translated using the following:

\textbf{Translation rule for $?_{mxl}\beta_1 \; ?\beta_2 \; ?\beta_3 \dots \;
?\beta_k \; \phi'$:} \\
$trans(?_{mxl}\beta_1 \; ?\beta_2 \; ?\beta_3 \dots \;
?\beta_k \; \phi', \linit) \defeq$ \\
\sql{(}\select{SELECT DISTINCT SNAPSHOT VALID($\alpha_2$), 
                      $\alpha_2$.2,
                      $\alpha_2$.3, \dots, $\alpha_2$.$k$ \\
               FROM (\select{SELECT DISTINCT 'dummy',
                                    $\alpha_1.\omega_2$,
                                    $\alpha_1.\omega_3$, \dots, 
                                    $\alpha_1.\omega_k$ \\
                             VALID $\alpha_1.\omega_1$ \\
                             FROM $trans(\phi', \linit)
                                  $ AS $\alpha_1$}\\
               \ \ \ \ \ )(NOSUBPERIOD) AS $\alpha_2$)}

Whenever the rule is used, $\alpha_1$ and $\alpha_2$ are
two different new correlation names, obtained by calling the
correlation names generator. Assuming that $\corn{\phi'} =
\tup{\tau_1, \dots, \tau_n}$, $\omega_1, \dots, \omega_k$ are as
in the rule for $?\beta_1 \; \dots \; ?\beta_k \; \phi$.
That is, the first position (from left to right) where $\beta_i$
appears in $\tup{\tau_1, \dots, \tau_n}$ is the $\omega_i$-th one.

Let us consider, for example, \pref{trans:14} (\qit{What circled when.}).
\begin{equation}
?_{mxl}e^v \; ?w^v \; \past[e^v, circling(w^v)] \label{trans:14}
\end{equation}
Let us also assume that $trans(\past[e^v, circling(w^v)], \linit)$
returns \pref{trans:15}. In this case, $\phi' = \past[e^v,
circling(w^v)]$ and $\corn{\phi'} = \tup{e^v, w^v}$. \pref{trans:15}
shows all the combinations of $et$s and values of $e^v$ and $w^v$, for
which the denotation of $\past[e^v, circling(w^v)]$ is $T$. In each
tuple, the value of the first explicit attribute (that corresponds to
$e^v$) is the same as the time-stamp, because the semantics of \past
requires the value of $e^v$ to be the same as $et$ (represented by the
time-stamp). To save space, I omit the time-stamps.
\begin{examps}
\item \label{trans:15}
\dbtableb{|l|l||l|}
{
 $[5\text{:}02pm \; 22/11/95, \; 5\text{:}17pm \; 22/11/95]$ & $BA737$ & \ \dots \\
 $[5\text{:}05pm \; 22/11/95, \; 5\text{:}15pm \; 22/11/95]$ & $BA737$ & \ \dots \\
 $[5\text{:}07pm \; 22/11/95, \; 5\text{:}13pm \; 22/11/95]$ & $BA737$ & \ \dots \\
 \ \dots & \ \dots & \ \dots \\
 $[4\text{:}57pm \; 23/11/95, \; 5\text{:}08pm \; 23/11/95]$ & $BA737$ & \ \dots \\
 $[4\text{:}59pm \; 23/11/95, \; 5\text{:}06pm \; 23/11/95]$ & $BA737$ & \ \dots \\
 $[5\text{:}01pm \; 23/11/95, \; 5\text{:}04pm \; 23/11/95]$ & $BA737$ & \ \dots \\
 \ \dots & \ \dots & \ \dots \\
 $[8\text{:}07am \; 22/11/95, \; 8\text{:}19am \; 22/11/95]$ & $UK160$ & \ \dots \\
 $[8\text{:}08am \; 22/11/95, \; 8\text{:}12am \; 22/11/95]$ & $UK160$ & \ \dots \\
 $[8\text{:}09am \; 22/11/95, \; 8\text{:}10am \; 22/11/95]$ & $UK160$ & \ \dots \\
 \ \dots & \ \dots & \ \dots
}
\end{examps}
BA737 was circling from 5:02pm to 5:17pm on 22/11/95, and from 4:57pm
to 5:08pm on 23/11/95. UK160 was circling from 8:07am to 8:19am on
22/11/95. \pref{trans:15} also contains tuples for the subperiods of
these periods, because $circling(w^v)$ (like all \topl predicates) is
homogeneous (section \ref{denotation}). $\past[e^v, circling(w^v)]$ is
true at all these subperiods that end before the present chronon.

In our example, the embedded \sql{SELECT} statement of $trans(?_{mxl}\beta_1 \;
?\beta_2 \; ?\beta_3 \dots \; ?\beta_k \; \phi', \linit)$ is:
\begin{examps} 
\item \label{trans:16}
\sql{(}\select{SELECT DISTINCT 'dummy', t1.2 \\
               VALID t1.1 \\
               FROM $trans(\past[e^v, circling(w^v)], \linit)$ AS t1)}
\end{examps}
\pref{trans:16} generates \pref{trans:17}, where
the time-stamps are the values of the first explicit attribute of
\pref{trans:15} (i.e.\ they correspond to $e^v$). The \sql{'dummy'} in
the embedded \sql{SELECT} statement (\pref{trans:16} in our example)
means that the first explicit attribute of that statement's resulting
relation should have the string ``$dummy$'' as its value in all
tuples. This is needed when $k = 1$. If, for example, \pref{trans:14}
were $?_{mxl}e^v \; \past[e^v, circling(ba737)]$, without the
\sql{'dummy'} the \sql{SELECT} clause of \pref{trans:16} would contain
nothing after \sql{DISTINCT} (this is not allowed in \tsql).
\begin{examps}
\item \label{trans:17}
\dbtableb{|l|l||l|}
{
 $dummy$ & $BA737$ & $[5\text{:}02pm \; 22/11/95, \; 5\text{:}17pm \; 22/11/95]$ \\
 $dummy$ & $BA737$ & $[5\text{:}05pm \; 22/11/95, \; 5\text{:}15pm \; 22/11/95]$ \\
 $dummy$ & $BA737$ & $[5\text{:}07pm \; 22/11/95, \; 5\text{:}13pm \; 22/11/95]$ \\
 \ \dots & \ \dots & \ \dots \\
 $dummy$ & $BA737$ & $[4\text{:}57pm \; 23/11/95, \; 5\text{:}08pm \; 23/11/95]$ \\
 $dummy$ & $BA737$ & $[4\text{:}59pm \; 23/11/95, \; 5\text{:}06pm \; 23/11/95]$ \\
 $dummy$ & $BA737$ & $[4\text{:}59pm \; 23/11/95, \; 5\text{:}06pm \; 23/11/95]$ \\
 \ \dots & \ \dots & \ \dots \\
 $dummy$ & $UK160$ & $[8\text{:}07am \; 22/11/95, \; 8\text{:}19am \; 22/11/95]$ \\
 $dummy$ & $UK160$ & $[8\text{:}08am \; 22/11/95, \; 8\text{:}12am \; 22/11/95]$ \\
 $dummy$ & $UK160$ & $[8\text{:}09am \; 22/11/95, \; 8\text{:}10am \; 22/11/95]$ \\
 \ \dots & \ \dots & \ \dots
}
\end{examps}
The \sql{(NOSUBPERIOD)} of the translation rule removes from
\pref{trans:17} any tuples that do not correspond to maximal
periods. That is \pref{trans:17} becomes \pref{trans:18}.
\begin{examps}
\item \label{trans:18}
\dbtableb{|l|l||l|}
{
 $dummy$ & $BA737$ & $[5\text{:}02pm \; 22/11/95, \; 5\text{:}17pm \; 22/11/95]$ \\
 $dummy$ & $BA737$ & $[4\text{:}57pm \; 23/11/95, \; 5\text{:}08pm \; 23/11/95]$ \\
 $dummy$ & $UK160$ & $[8\text{:}07am \; 22/11/95, \; 8\text{:}19am \; 22/11/95]$
}
\end{examps}
The overall \pref{trans:14} is mapped to \pref{trans:19}, which
generates \pref{trans:20}. \pref{trans:20}
represents the denotation of \pref{trans:14} w.r.t.\
$M(st)$ and $st$ (pairs of maximal circling periods and the
corresponding flights).
\begin{examps} 
\item \label{trans:19}
\sql{(}\select{SELECT DISTINCT SNAPSHOT VALID(t2), t2.2 \\
               FROM (\select{SELECT DISTINCT 'dummy', t1.2 \\
                     VALID t1.1 \\
                     FROM $trans(\past[e^v, circling(w^v)], 
                                 \linit)$ AS t1} \\
               \ \ \ \ \ )(NOSUBPERIOD) AS t2)}
\item \label{trans:20}
\dbtableb{|l|l|}
{
 $[5\text{:}02pm \; 22/11/95, \; 5\text{:}17pm \; 22/11/95]$ & $BA737$ \\
 $[4\text{:}57pm \; 23/11/95, \; 5\text{:}08pm \; 23/11/95]$ & $BA737$ \\
 $[8\text{:}07am \; 22/11/95, \; 8\text{:}19am \; 22/11/95]$ & $UK160$
}
\end{examps}

Appendix \ref{trans_proofs} proves that the translation rules for
wh-formulae satisfy theorem \ref{wh_theorem}. 


\section{Optimising the generated TSQL2 code} \label{tsql2_opt}

The generated \tsql code is often verbose. There are usually ways in
which it could be shortened and still return the same results.
Figure \ref{optimise_code}, for example, shows the code that is
generated by the translation of \pref{opt:1}, if chronons correspond
to minutes. (\pref{opt:1} expresses 
the reading of \qit{Who inspected UK160 yesterday?} where the
inspection must have both started and been completed on the 
previous day.) 
\begin{eqnarray}
&&?w^v \; \at[yesterday, \past[e^v, \culm[inspecting(occr^v, w^v, uk160)]]]
\label{opt:1}
\end{eqnarray}

\begin{figure}
\hrule
\medskip
{\small
\begin{verbatim}
(SELECT DISTINCT SNAPSHOT t4.3
 FROM (SELECT DISTINCT VALID(t3), t3.1, t3.2, t3.3
       VALID VALID(t3)
       FROM (SELECT DISTINCT t1.1, t1.2, t1.3
             VALID PERIOD(BEGIN(VALID(t1)), END(VALID(t1)))
             FROM (SELECT DISTINCT insp.1, insp.2, insp.3
                   VALID VALID(insp)
                   FROM inspections(PERIOD) AS insp)(ELEMENT) AS t1,
                  (SELECT DISTINCT SNAPSHOT inspcmpl.1, inspcmpl.2, inspcmpl.3
                   FROM inspections AS inspcmpl
                   WHERE inspcmpl.4 = 'complete') AS t2
             WHERE t1.1 = t2.1 AND t1.2 = t2.2
               AND t1.3 = t2.3 AND t1.3 = 'UK160'
               AND INTERSECT(
                    INTERSECT(
                      PERIOD(TIMESTAMP 'beginning', TIMESTAMP 'forever'),
                      PERIOD 'today' - INTERVAL '1' DAY),
                    PERIOD(TIMESTAMP 'beginning', 
                           TIMESTAMP 'now' - INTERVAL '1' MINUTE))
                   CONTAINS PERIOD(BEGIN(VALID(t1)), END(VALID(t1)))
            ) AS t3
      ) AS t4)
\end{verbatim}
}
\vspace*{-5mm}
\caption{Example of generated \tsql code}
\label{optimise_code}
\medskip
\hrule
\end{figure}

I assume here that $\hpfunsp(inspecting, 3)$ and $\hculmsp(inspecting,
3)$ are \pref{hpfuns:4} and \pref{hpfuns:5} respectively. The embedded
\sql{SELECT} statements of figure \ref{optimise_code} that are
associated with \sql{t1} and \sql{t2} are \pref{hpfuns:4} and
\pref{hpfuns:5}. The embedded \sql{SELECT} statement that is
associated with \sql{t3} corresponds to $\culm[inspecting(occr^v, w^v,
uk160)]$ (see the rule for $\culm[\pi(\tau_1, \dots, \tau_n)]$ in section
\ref{trans_rules}). It generates a relation whose explicit
attributes show all the combinations of codes,
inspectors, and inspected objects that correspond to complete
inspections. The time-stamps of this relation represent periods that
cover whole inspections (from start to completion). The last
constraint in the \sql{WHERE} clause (the one with
\sql{CONTAINS}) admits only tuples whose time-stamps (whole 
inspections) are subperiods of $lt$. The two nested \sql{INTERSECT}s
before \sql{CONTAINS} represent $lt$. The \at has narrowed $lt$ to the
intersection of its original value (whole time-axis) with the previous
day (\sql{PERIOD 'today' - INTERVAL '1' DAY)}). The \past has 
narrowed $lt$ further to the intersection with 
$[t_{first}, st)$ (\sql{PERIOD(TIMESTAMP 'beginning', TIMESTAMP
'now' - INTERVAL '1' MINUTE)}).

The embedded \sql{SELECT} statement that is associated with \sql{t4}
is generated by the translation rule for $\past[\beta, \phi']$
(section \ref{trans_rules}). It returns the same relation as the
statement that is associated with \sql{t3}, except that the relation
of \sql{t4}'s statement has an additional explicit attribute that
corresponds to the first argument of \past. In
each tuple, the value of this extra attribute is the same as the
time-stamp ($et$). The topmost 
\sql{SELECT} clause projects only the third explicit attribute of the
relation returned by \sql{t4}'s statement (this attribute corresponds
to $w^v$ of \pref{opt:1}).

The code of figure \ref{optimise_code} could be shortened in several
ways. \sql{t4}'s statement, for example, simply adds an extra attribute for the
first argument of \past. In this particular case, this extra
attribute is not used, because
\pref{opt:1} contains no interrogative quantifier for the first 
argument of \past. Hence, \sql{t4}'s statement could be replaced by
\sql{t3}'s (the topmost \sql{SELECT} clause would have to become
\sql{SELECT DISTINCT SNAPSHOT t3.2}). One could also drop the
top-level \sql{SELECT} statement, and replace the \sql{SELECT} clause
of \sql{t3}'s statement with \sql{SELECT DISTINCT SNAPSHOT
  t1.2}. Furthermore, the intersection of the whole time-axis
(\sql{PERIOD(TIMESTAMP 'beginning', TIMESTAMP 'forever')}) with any
period $p$ is simply $p$.  Hence, the second \sql{INTERSECT(\dots,
  \dots)} could be replaced by its second argument.  The resulting
code is shown in figure \ref{optimise_code2}. Further simplifications
are possible. 

\begin{figure}
\hrule
\medskip
{\small
\begin{verbatim}
(SELECT DISTINCT SNAPSHOT t1.2
 FROM (SELECT DISTINCT insp.1, insp.2, insp.3
       VALID VALID(insp)
       FROM inspections(PERIOD) AS insp)(ELEMENT) AS t1,
      (SELECT SNAPSHOT inspcmpl.1, inspcmpl.2, inspcmpl.3
       FROM inspections AS inspcmpl
       WHERE inspcmpl.4 = 'complete') AS t2
 WHERE t1.1 = t2.1 AND t1.2 = t2.2
   AND t1.3 = t2.3 AND t1.3 = 'UK160'
   AND INTERSECT(PERIOD 'today' - INTERVAL '1' DAY,
                 PERIOD(TIMESTAMP 'beginning',
                        TIMESTAMP 'now' - INTERVAL '1' MINUTE))
       CONTAINS PERIOD(BEGIN(VALID(t1)), END(VALID(t1)))
\end{verbatim}
}
\vspace*{-5mm}
\caption{Shortened \tsql code}
\label{optimise_code2}
\medskip
\hrule
\end{figure}

Most \dbms{s} employ optimisation techniques. A commercial \dbms
supporting \tsql would probably be able to carry out at least some of
the above simplifications. Hence, the reader may wonder why should the
\nlitdb attempt to optimise the \tsql code, rather than delegate the
optimisation to the \dbms. First, as mentioned in section
\ref{tdbs_general}, only a prototype \dbms currently supports
\tsql. Full-scale \tsql \dbms{s} with optimisers may not appear in the
near future. Second, long database language queries (like the ones
generated by the framework of this thesis) can often confuse generic
\dbms optimisers, causing them to produce inefficient code. Hence,
shortening the \tsql code before submitting it to the \dbms is again
important. It would be interesting to examine if optimisations like
the ones discussed above could be automated, and integrated into the
framework of this thesis as an additional layer between the \topl to
\tsql translator and the \dbms. I have not explored this issue. 


\section{Related work}

Various mappings from forms of logic to and from relational algebra
(e.g.\ \cite{Ullman}, \cite{VanGelder1991}), from logic programming
languages to \sqll (e.g.\ \cite{Lucas1988}, \cite{Draxler1992}), and
from logic formulae generated by \nlidb{s} to \sqll (\cite{Lowden1},
\cite{Androutsopoulos}, \cite{Androutsopoulos3}, \cite{Rayner93}) have
been discussed in the past. The mapping which is most relevant to the
\topl to \tsql translation of this chapter is that of \cite{Boehlen1996}.

Boehlen et al.\ study the relation between \tsql and an extended
version of first order predicate logic (henceforth called \sul), that
provides the additional temporal operators $\prevop$ (previous),
$\nextop$ (next), $\since$, and $\until$. \sul is point-based, in the
sense that \sul formulae are evaluated with respect to single
time-points. \sul assumes that time is discrete. Roughly speaking,
$\prevop \phi$
\index{.@$\prevop$ (\sul operator, previous time-point)}
is true at a time-point $t$ iff $\phi$ is true at the
time-point immediately before $t$. Similarly, $\nextop \phi$
\index{..@$\nextop$ (\sul operator, next time-point)}
is true at $t$ iff $\phi$ is true at the time-point immediately after
$t$. $\phi_1 \; \since \; \phi_2$ 
\index{since@$\since$ (\sul operator)}
is true at $t$ iff there is some
$t'$ before $t$, such that $\phi_2$ is true at $t'$, and for every
$t''$ between $t'$ and $t$, $\phi_1$ is true at $t''$. Similarly,
$\phi_1 \; \until \; \phi_2$ 
\index{until@$\until$ (\sul operator)}
is true at $t$ iff there is some $t'$ after $t$, such that $\phi_2$ is
true at $t'$, and for every $t''$ between $t$ and $t'$, $\phi_1$ is
true at $t''$. Various other operators are also defined, but these are
all definable in terms of $\prevop$, $\nextop$, $\since$, and
$\until$. For example, $\tlpast \phi$ 
\index{<>@$\tlpast$ (\sul's past operator)}
is equivalent to $true \; \since \; \phi$ ($true$ is a special formula
that is true at all time-points). In effect, $\tlpast \phi$ is  
true at $t$ if there is a $t'$ before $t$, and $\phi$ is true at $t'$.
For example, \pref{sul:1a} and \pref{sul:2a} can be expressed as
\pref{sul:1} and \pref{sul:2} respectively.
\begin{examps}
\item BA737 departed (at some past time). \label{sul:1a} 
\item $\tlpast depart(ba737)$ \label{sul:1}
\item Tank 2 has been empty (all the time) since BA737 departed. \label{sul:2a}
\item $empty(tank2) \; \since \; depart(ba737)$ \label{sul:2}
\end{examps}

Boehlen et al.\ provide rules that translate from \sul to
\tsql. (They also show how to translate from a fragment of
\tsql back to \sul, but this direction is irrelevant here.) The
underlying ideas are very similar to those of this chapter. Roughly
speaking, there are non-recursive rules for atomic formulae, and
recursive rules for non-atomic formulae. For example, the translation
rule for $\phi_1 \; \since \; \phi_2$ calls recursively the
translation algorithm to translate $\phi_1$ and $\phi_2$. The result
is a \sql{SELECT} statement, that contains two embedded \sql{SELECT}
statements corresponding to $\phi_1$ and $\phi_2$. Devising
rules to map from \sul to \tsql is much easier
than in the case of \topl, mainly because \sul formulae are evaluated
with respect to only one time-parameter (\topl formulae are
evaluated with respect to three parameters, $st$, $et$, and $lt$),
\sul is point-based (\topl is period-based; section \ref{top_intro}), and
\sul provides only four temporal operators whose semantics are very
simple (the \topl version of this chapter has eleven
operators, whose semantics are more complex). Consequently, proving
the correctness of the \sul to \tsql mapping is much simpler than in
the case of \topl.

It has to be stressed, however, that \topl and \sul were designed for
very different purposes. \sul is interesting from a theoretical
temporal-logic point of view. Roughly speaking, it has been proven
that whatever can be expressed in traditional first-order
predicate logic with a temporal precedence connective by treating time
as an extra predicate argument (e.g.\ \pref{tlogi:2} of page
\pageref{tlogi:2}) can also be expressed in first-order predicate 
logic enhanced with only a $\since$ and an $\until$ operator, subject
to some continuity conditions (the reverse is not true; see chapter
II.2 of \cite{VanBenthem}). The mapping from \sul to \tsql (and the
reverse mapping from a fragment of \tsql to \sul) is part of a study
of the relative expressiveness of \sul and \tsql.  The existence of a
mapping from \sul to \tsql shows that \tsql is at least as expressive
as \sul. (The reverse is not true. Full \tsql is more expressive than
\sul; see \cite{Boehlen1996}.)

In contrast, \topl was not designed to study expressiveness issues,
but to facilitate the mapping from (a fragment of) English to logical
form. Chapter \ref{English_to_TOP} showed how to translate
systematically from a non-trivial fragment of English temporal questions into
\topl. No such systematic translation has been shown to exist in the
case of \sul, and it is not at all obvious how temporal English
questions (e.g.\ containing progressive and perfect tenses, temporal
adverbials, temporal subordinate clauses) could be mapped
systematically to appropriate \sul formulae.

Although the study of expressiveness issues is not a goal of this
thesis, I note that the \topl to \tsql translation of this chapter
implies that \tsql is at least as expressive as \topl (every \topl
formula can be mapped to an appropriate \tsql query). The reverse is not
true: it is easy to think of \tsql queries (e.g.\ queries
that report cardinalities of sets) that cannot be expressed in (the
current version of) \topl. Finally, neither
\topl nor \sul can be said to be more expressive than the other, as
there are English sentences that can be expressed in \sul but not 
\topl, and vice-versa. For example, the \sul formula \pref{sul:2}
expresses \pref{sul:2a}, a sentence that cannot be expressed in
\topl. Also, the \topl formula \pref{sul:11} expresses
\pref{sul:10}. There does not seem to be any way to express
\pref{sul:10} in \sul.
\begin{examps}
\item Tank 2 was empty for two hours. \label{sul:10}
\item $\for[hour^c, 2, \past[e^v, empty(tank2)]]$ \label{sul:11}
\end{examps}


\section{Summary} 

\tsql is an extension of \sqlnt that provides special
facilities for manipulating temporal information. Some modifications
of \tsql were adopted in this chapter. Some of these are minor, and
were introduced to bypass uninteresting details (e.g.\ referring to
explicit attributes by number) or obscure points in the \tsql definition
(e.g.\ the new version of the \sql{INTERVAL} function). Other
modifications are more significant, and were introduced to facilitate the
\topl to \tsql translation (e.g.\ \sql{(SUBPERIOD)} and
\sql{(NOSUBPERIOD)}). One of these more significant modifications
(calendric relations) is generally useful. Some minor modifications of
\topl were also adopted in this chapter. 

A method to translate from \topl to \tsql was
framed. Each \topl formula $\phi$ is mapped to a \tsql query. This is
executed by the \dbms, generating a relation that represents (via an
interpretation function) $\denot{M(st),st}{\phi}$. Before the
translation method can be used, the configurer of the \nlitdb must
specify some functions (\hconsp, \hpfunsp, \hculmsp, \hgpartsp,
\hcpartsp) that link certain basic \topl expressions to  
\tsql expressions. The \topl to \tsql translation 
is then carried out by a set of translation rules. The rules have to
satisfy two theorems (\ref{wh_theorem} and \ref{yn_theorem}) for the
translation to be correct (i.e.\ for the \tsql query to generate a
relation that represents $\denot{M(st),st}{\phi}$). An informal
description of the functionality of some of the rules was given. The
full set of the translation rules, along with a proof that they
satisfy theorems \ref{wh_theorem} and \ref{yn_theorem}, is given in
appendix \ref{trans_proofs}. Further work could explore how to
optimise the generated \tsql code.

The \topl to \tsql translation is in principle similar to the \sul to \tsql
translation of \cite{Boehlen1996}. \topl and \sul, however, were designed for
very different purposes, and the \sul to \tsql translation is 
much simpler than the \topl to \tsql one. 



\chapter{The prototype NLITDB} \label{implementation}

\proverb{Time works wonders.}


\section{Introduction}

This chapter discusses the architecture of the prototype \nlitdb,
provides some information on how the modules of the system were
implemented, and explains which modules would have to be added if the
prototype \nlitdb were to be used in real-life applications. A
description of the hypothetical airport database is also given,
followed by sample questions from the airport domain and the
corresponding output of the \nlitdb. The chapter ends with information
on the speed of the system.


\section{Architecture of the prototype NLITDB} \label{prototype_arch}

\begin{figure}
\hrule
\medskip
\begin{center}
\includegraphics[scale=.6]{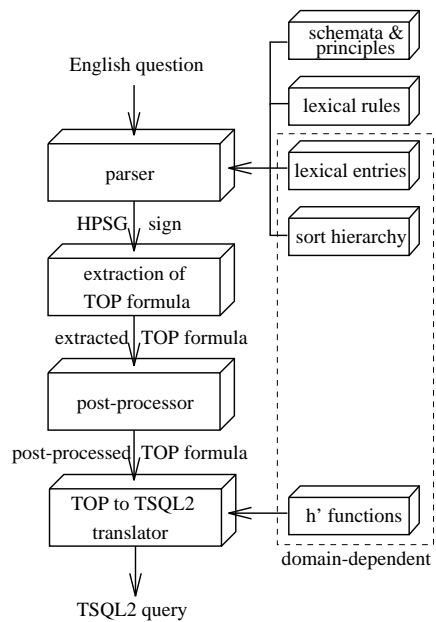}
\caption{Architecture of the prototype NLITDB}
\label{simple_arch_fig}
\end{center}
\hrule
\end{figure}

Figure \ref{simple_arch_fig} shows the architecture of the prototype
\nlitdb. Each English question is first parsed using the \hpsg grammar
of chapter \ref{English_to_TOP}, generating an \hpsg sign. Multiple
signs are generated for questions that the parser understands to be
ambiguous. A \topl formula is then extracted from each sign, as
discussed in section \ref{extraction_hpsg}. Each extracted formula
subsequently undergoes the post-processing of section
\ref{post_processing}. (The post-processor also converts the formulae
from the \topl version of chapters \ref{TOP_chapter} and
\ref{English_to_TOP} to the version of \ref{tdb_chapter}; see section
\ref{TOP_mods}.) As discussed in section \ref{post_processing}, the
post-processing sometimes generates multiple formulae from the same
original formula.

Each one of the formulae that are generated at the end of the
post-processing captures what the \nlitdb understands to be a possible
reading of the English question. Many fully-fledged \nlidb{s} use
preference measures to guess which reading among the possible ones the
user had in mind, or generate ``unambiguous'' English paraphrases of
the possible readings, asking the user to select one (see
\cite{Alshawi}, \cite{Alshawi2}, \cite{DeRoeck1986} and
\cite{Lowden1986}). No such mechanism is currently present in the
prototype \nlitdb. All the formulae that are generated at the end of
the post-processing are translated into \tsql. The \nlitdb prints all
the resulting \tsql queries along with the corresponding \topl
formulae. The \tsql queries would be executed by the \dbms to
retrieve the information requested by the user. As mentioned in
section \ref{tdbs_general}, however, the prototype \nlitdb has not
been linked to a \dbms. Hence, the \tsql queries are currently not
executed, and no answers are produced.

The following sections provide more information about the grammar and
parser, the module that extracts \topl formulae from \hpsg signs, the
post-processor, and the \topl to \tsql translator.


\section{The grammar and parser} \label{parser}

The \hpsg version of chapter \ref{English_to_TOP} was coded in the
formalism of \ale \cite{Carpenter1992} \cite{Carpenter1994}, building
on previous \ale encodings of \hpsg fragments by Gerald Penn, Bob
Carpenter, Suresh Manandhar, and Claire Grover.\footnote{The prototype
  \nlitdb was implemented using \ale version 2.0.2 and Sicstus Prolog
  version 2.1.9. Chris Brew provided additional \ale code for
  displaying feature structures. The software of the prototype
  \nlitdb, including the \ale grammar, is available from
  \texttt{http://www.dai.ed.ac.uk/groups/nlp/NLP\_home\_page.html}. An
  earlier version of the prototype \nlitdb was implemented using the
  \textsc{Hpsg-Pl} and \textsc{Pleuk} systems \cite{Popowich}
  \cite{Calder}.} \ale can be thought of as a grammar-development
environment. It provides a chart parser (which is the one used in the
prototype \nlitdb; see \cite{Gazdar1989} for an introduction to chart
parsers) and a formalism that can be used to write
unification-grammars based on feature structures.

Coding the \hpsg version of chapter \ref{English_to_TOP} in \ale's
formalism proved straight-forward. \ale's formalism allows one to specify
grammar rules, definite constraints (these are similar to Prolog
rules, except that predicate arguments are feature structures),
lexical entries, lexical rules, and a hierarchy of sorts of feature
structures. The schemata and principles of the \hpsg version of
chapter \ref{English_to_TOP} were coded using \ale grammar rules and
definite constraints. \ale's lexical entries, lexical rules, and sort
hierarchy were used to code \hpsg's lexical signs, lexical rules, and
sort hierarchy respectively.

The \ale grammar rules and definite constraints that encode the \hpsg
schemata and principles are domain-independent, i.e.\ they require no
modifications when the \nlitdb is configured for a new application
domain. The lexical rules of the prototype \nlitdb are also intended
to be domain-independent, though their morphology parts need to be
extended (e.g.\ more information about the morphology of irregular
verbs is needed) before they can be used in arbitrary application
domains. The lexical entries of the system that correspond to
determiners (e.g.\ \qit{a}, \qit{some}), auxiliary verbs,
interrogative words (e.g.\ \qit{who}, \qit{when}), prepositions,
temporal subordinators (e.g.\ \qit{while}, \qit{before}), month names,
day names, etc.\ (e.g.\ \qit{January}, \qit{Monday}) are also
domain-independent. The person configuring the \nlitdb, however, needs
to provide lexical entries for the nouns, adjectives, and
(non-auxiliary) verbs that are used in the particular application
domain (e.g.\ \qit{flight}, \qit{open}, \qit{to land}).  The largest
part of the \nlitdb's sort hierarchy is also domain independent. Two
parts of it need to be modified when the system is configured for a
new domain: the hierarchy of world entities that is mounted under
{\srt ind}\/ (section \ref{more_ind} and figure
\vref{ind_hierarchy}), and the subsorts of {\srt predicate}\/ that
correspond to \topl predicates used in the domain (section
\ref{TOP_FS} and figure \vref{psoa_fig}). As will be discussed in
section \ref{modules_to_add}, tools could be added to help the
configurer modify the domain-dependent modules.


\section{The extractor of TOP formulae and the post-processor}
\label{extraction_impl} 

The module that extracts \topl formulae from \hpsg signs actually
generates \topl formulae written as Prolog terms. For example, it
generates \pref{extr:5} instead of \pref{extr:3}. The correspondence
between the two notations should be obvious. The Prolog-like notation
of \pref{extr:5} is also used in the formulae that are passed to the
\topl to \tsql translator, and in the output of the \nlitdb (see
section \ref{samples} below). 
\begin{examps}
\item $?x1^v \; \ntense[x3^v, president(x1^v)] \land 
       \past[x2^v, located\_at(x1^v, terminal2)]$ \label{extr:3}
\item \texttt{\small interrog(x1\^{}v, and(\hspace*{-2mm}\begin{tabular}[t]{l}
                        ntense(x3\^{}v, president(x1\^{}v)),\\
                        past(x2\^{}v, located\_at(x1\^{}v, terminal2))))
                        \end{tabular}} 
      \label{extr:5} 
\end{examps}
The extractor of the \topl formulae is implemented using Prolog rules
and \ale definite constraints (Prolog-like rules whose
predicate-arguments are feature structures). Although the
functionality of the extractor's code is simple, the code itself is
rather complicated (it has to manipulate the internal data structures
that \ale uses to represent feature structures) and will not be
discussed.

As mentioned in section \ref{prototype_arch}, the post-processor of
figure \ref{simple_arch_fig} implements the post-processing phase of
section \ref{post_processing}. The post-processor also eliminates
\partop operators by merging them with the corresponding \at, \before,
or \after operators, to convert the formulae into the \topl version of
the \topl to \tsql translator (section \ref{TOP_mods}). The
post-processor's code, which is written in Prolog, presents no
particular interest and will not be discussed.


\section{The TOP to TSQL2 translator} \label{translator_module}

Implementing in Prolog the \topl to \tsql mapping of chapter
\ref{tdb_chapter} proved easy. The code of the \topl to \tsql
translator of figure \ref{simple_arch_fig} is basically a collection
of Prolog rules for the predicate \texttt{trans}. Each one of these
rules implements one of the translation rules of section
\ref{trans_rules} and appendix \ref{trans_proofs}. For example, the
following Prolog rule implements the translation rule for
$\past[\beta, \phi']$ (p.~\pageref{past_trans_discuss}). (I omit some
uninteresting details of the actual Prolog rule.)

\ssmall
\begin{verbatim}
trans(past(_^v, PhiPrime), Lambda, Sigma):-
  chronons(Chronon),
  multiappend([
    "INTERSECT(", Lambda, ", ", 
               "PERIOD(TIMESTAMP 'beginning', ",
                      "TIMESTAMP 'now' - INTERVAL '1' ", Chronon, "))"
    ], LambdaPrime),
  trans(PhiPrime, LambdaPrime, SigmaPrime), 
  new_cn(Alpha), 
  corners(PhiPrime, CList), 
  length(CList, N), 
  generate_select_list(Alpha, N, SelectList),
  multiappend([
    "(SELECT DISTINCT VALID(", Alpha, "), ", SelectList, 
     "VALID VALID(", Alpha, ")", 
     "FROM ", SigmaPrime, " AS ", Alpha, ")", 
    ], Sigma).
\end{verbatim}

\dnorm The first argument of \texttt{trans} is the \topl formula to be
translated (in the notation of \pref{extr:5}). \texttt{Lambda} is a
string standing for the $\lambda$ argument of the $trans$ function of
section \ref{formulation} (initially \texttt{"PERIOD(TIMESTAMP
'beginning', TIMESTAMP 'forever')"}). The generated \tsql code is
returned as a string in \texttt{Sigma}.

The \texttt{chronons(Chronon)} causes \texttt{Chronon} to become a
string holding the \tsql name of the granularity of chronons (e.g.\
\sql{"MINUTE"}). The \texttt{chronons} predicate is supplied by the
configurer of the \nlitdb, along with Prolog predicates that define
the $h'$ functions of section \ref{via_TSQL2}. For example, the
following predicate defines $\hpfunsp(inspecting,3)$ to be the
\sql{SELECT} statement of \pref{hpfuns:4} on page
\pageref{hpfuns:4}. The \texttt{chronons} predicate and the predicates
that define the $h'$ functions are the only domain-dependent parts of
the \topl to \tsql translator.

\ssmall
\begin{verbatim}
h_prime_pfuns_map(inspecting, 3, 
  ["SELECT DISTINCT insp.1, insp.2, insp.3", 
   "VALID VALID(insp)",
   "FROM inspections(PERIOD) AS insp"]).
\end{verbatim}

\dnorm The first \texttt{multiappend} in the $trans$ rule above generates the
$\lambda'$ string of the translation rule for $\past[\beta, \phi']$
(p.~\pageref{past_trans_discuss}). It concatenates the string-elements
of the list provided as first argument to \texttt{multiappend}, and
the resulting string ($\lambda'$) is returned in
\texttt{LambdaPrime}. As in the translation rule for $\past[\beta,
\phi']$, the translation mapping is then invoked recursively to
translate $\phi'$ (\texttt{PhiPrime}). The result of this translation
is stored in \texttt{SigmaPrime}.

\texttt{new\_cn(Alpha)} returns in \texttt{Alpha} a string 
holding a new correlation name (\texttt{new\_cn} 
implements the correlation names generator of section
\ref{trans_rules}). The \texttt{corners(PhiPrime, CList)} causes 
\texttt{CList} to become $\corn{\phi'}$, and \texttt{length(CList,
N)} returns in \texttt{N} the length of $\corn{\phi'}$. The
\texttt{generate\_select\_list(Alpha, N, SelectList)} returns in 
\texttt{SelectList} a string of the form \texttt{Alpha.1, Alpha.2, \dots,
Alpha.N}. Finally, the second \texttt{multiappend} returns in
\texttt{Sigma} a string that holds the overall \tsql code.


\section{Modules to be added} \label{modules_to_add}

\begin{figure}
\hrule
\medskip
\begin{center}
\includegraphics[scale=.6]{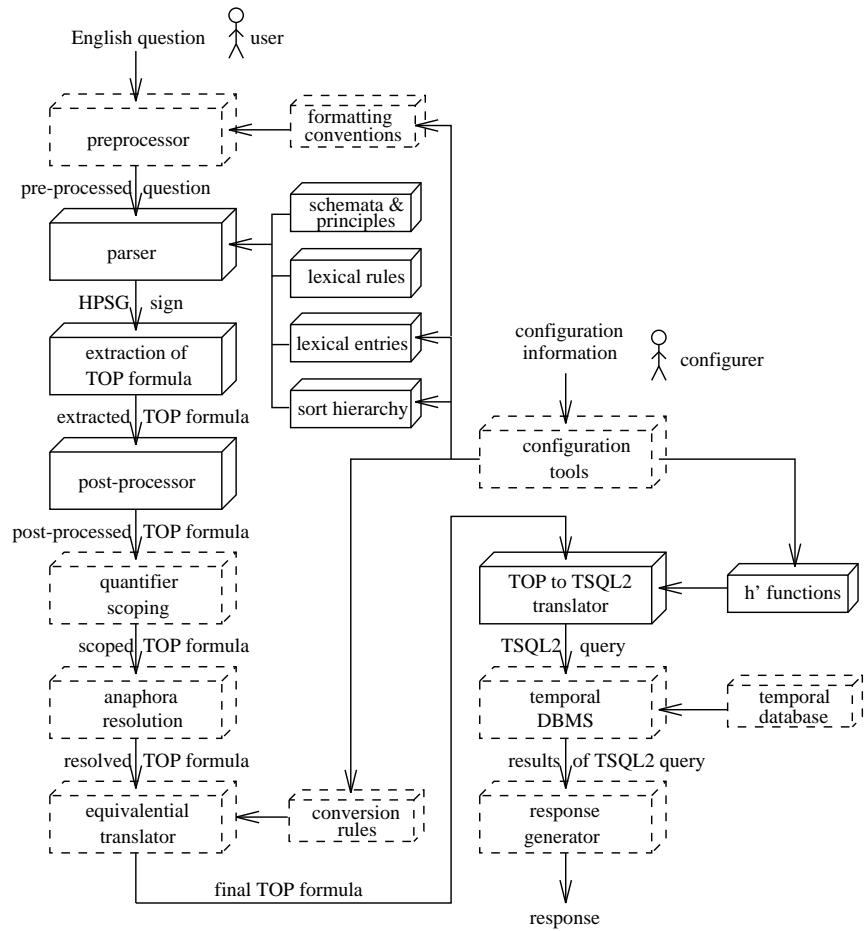}
\caption{Extended architecture of the prototype NLITDB}
\label{arch_fig}
\end{center}
\hrule
\end{figure}

The prototype \nlitdb is intended to demonstrate that the mappings
from English to \topl and from \topl to \tsql are
implementable. Consequently, the architecture of the prototype \nlitdb
is minimal. Several modules, to be sketched in the following sections,
would have to be added if the system were to be used in real-life
applications. Figure \ref{arch_fig} shows how these modules would fit
into the existing system architecture. (Modules drawn with dashed
lines are currently not present.)


\subsection{Preprocessor} \label{preprocessor}

The \ale parser requires its input sentence to by provided as a Prolog
list of symbols (e.g.\ \pref{prepro:2}). 
\begin{examps}
\item \texttt{[was,ba737,circling,at,pm5\_00]} \label{prepro:2}
\end{examps}
As there is essentially no interface between the user and the \ale
parser in the prototype \nlitdb, English questions have to be
typed in this form. This does not allow words to start with capital
letters or numbers, or to contain characters like ``\texttt{/}'' and
``\texttt{:}'' (Prolog symbols cannot contain these characters and
must start with lower case letters). To bypass these constraints,
proper names, dates, and times currently need to be typed in unnatural
formats (e.g.\ ``\texttt{london}'', ``\texttt{d1\_5\_92}'',
``\texttt{pm5\_00}'', ``\texttt{y1991}'' instead of
``\texttt{London}'', ``\texttt{1/5/92}'' ``\texttt{5:00pm}'',
``\texttt{1991}'').

A preprocessing module is needed, that would allow English questions
to be typed in more natural formats (e.g.\ \pref{prepro:4}), and would
transform the questions into the format required by the parser (e.g.\
\pref{prepro:2}).
\begin{examps}
\item Was BA737 circling at 5:00pm? \label{prepro:4}
\end{examps}

Similar preprocessing modules are used in several natural language
front-ends (e.g.\ \textsc{Cle} \cite{Alshawi} and \textsc{Masque}
\cite{Lindop}). These modules typically also merge parts of the input
sentence that need to be processed as single words. For example, the
lexicon of the airport domain has a single lexical entry for \qit{gate
2}. The preprocessor would merge the two words of \qit{gate 2} in
\pref{prepro:7}, generating \pref{prepro:8}. (Currently, \qit{gate 2}
has to be typed as a single word.)
\begin{examps}
\item Which flights departed from gate 2 yesterday? \label{prepro:7}
\item \texttt{[which,flights,departed,from,gate2,yesterday]} \label{prepro:8}
\end{examps}
The preprocessing modules typically also handle proper names that
cannot be included in the lexicon because they are too many, or
because they are not known when creating the lexicon. In a large
airport, for example, there would be hundreds of flight names
(\qit{BA737}, \qit{UK1751}, etc.). Having a different lexical entry
for each flight name is impractical, as it would require hundreds of
entries to be added into the lexicon. Also, new flights (and hence
flight names) are created occasionally, which means that the
lexicon would have to be updated whenever a new flight is
created. Instead, the lexicon could contain entries for a small
number of pseudo-flight names (e.g.\ \qit{flight\_name1},
\qit{flight\_name2}, \dots, \qit{flight\_nameN}; N is the maximum
number of flight names that may appear in a question, e.g.\
5). Each one of these lexical entries would map a
pseudo-flight name to a \topl constant (e.g.\ $flight1$, $flight2$,
\dots, $flightN$).\footnote{In the \hpsg grammar of chapter
\ref{English_to_TOP}, these constants would be represented using sorts
like {\srt flight1}, {\srt flight2}, \dots, {\srt flightN}, which
would be daughters of {\srt flight\_ent}\/ and sisters of {\srt
flight\_ent\_var}\/ in figure \vref{ind_hierarchy}.} The
preprocessor would use domain-dependent formatting conventions to identify
flight names in the English question (e.g.\ that any word that starts
with two or three capital letters and is followed by three or four
digits is a flight name). Each flight name in the question
would be replaced by a pseudo-flight name. For example, the
preprocessor would turn \pref{prepro:4.1} into \pref{prepro:4.2}.
\begin{examps}
\item Did BA737 depart before UK160 started to land? \label{prepro:4.1} 
\item \texttt{[did,flight\_name1,depart,before,flight\_name2,started,to,land]}
   \label{prepro:4.2} 
\end{examps}
\pref{prepro:4.2} would then be parsed, giving rise to
\pref{prepro:4.3} ($flight1$ and $flight2$ are \topl constants
introduced by the lexical entries of \qit{flight\_name1} and
\qit{flight\_name2}). An extra step would be added to the
post-processing phase of section \ref{post_processing}, to substitute
$flight1$ and $flight2$ with \topl constants that reflect the original
flight names. For example, the preprocessor could pass to the
post-processor the original flight names (\qit{BA737} and
\qit{UK160}), and the post-processor could replace $flight1$ and
$flight2$ by the original flight names in lower case. This would cause
\pref{prepro:4.3} to become \pref{prepro:4.4}. Similar problems arise
in the case of dates, times, numbers, etc.
\begin{examps}
\item $\before[\past[e1^v, \lbegin[landing(flight2)]], 
               \past[e2^v, depart(flight1)]]$ \label{prepro:4.3}
\item $\before[\past[e1^v, \lbegin[landing(uk160)]], 
               \past[e2^v, depart(ba737)]]$ \label{prepro:4.4}
\end{examps}
No preprocessing mechanism is currently present in the prototype
\nlitdb. The lexicon contains (for demonstration purposes) only a few
entries for particular (not pseudo-) flight names, times, dates, and
numbers (e.g.\ \qit{BA737}, \qit{9:00am}). For example, there is no
entry for \qit{9:10am}. This causes the parsing of \qit{Which tanks
  were empty at 9:10am?} to fail. In contrast the parsing of
\qit{Which tanks were empty at 9:00am?} succeeds, because there
\emph{is} a lexical entry for \qit{9:00am}.


\subsection{Quantifier scoping} \label{quantif_scoping}

When both words that introduce existential quantification (e.g.\
\qit{a}, \qit{some}) and words that introduce universal
quantification (e.g.\ \qit{every}, \qit{each}) are allowed, it is
often difficult to decide which quantifiers should be given scope over
which other quantifiers. For example, \pref{qsco:1} has two possible
readings. Ignoring the temporal information of \pref{qsco:1},
these readings would be expressed in the traditional first-order
predicate logic (\textsc{Fopl}) using formulae like \pref{qsco:2} and
\pref{qsco:3}.
\begin{examps}
\item A guard inspected every gate. \label{qsco:1}
\item $\exists x \; (guard(x) \land 
                     \forall y \; (gate(y) \rightarrow inspect(x, y)))$ 
   \label{qsco:2}
\item $\forall y \; (gate(y) \rightarrow 
                    \exists x \; (guard(x) \land inspect(x, y)))$
   \label{qsco:3}
\end{examps}
In \pref{qsco:2}, the existential quantifier (introduced by \qit{a})
is given wider scope over the universal one (introduced by
\qit{every}). According to \pref{qsco:2}, all the gates were inspected
by the same guard. In contrast, in \pref{qsco:3} where the universal
quantifier is given wider scope over the existential one, each gate
was inspected by a possibly different guard.

In \pref{qsco:1}, both scopings seem possible (at least in the absence
of previous context). In many cases, however, one of the possible
scopings is the preferred one, and there are heuristics to determine
this scoping (see chapter 8 of \cite{Alshawi}). For example, universal
quantifiers introduced by \qit{each} tend to have wider scope over
existential quantifiers (e.g. if the \qit{every} of \pref{qsco:1} is
replaced by \qit{each}, the scoping of \pref{qsco:3} becomes more
likely than that of \pref{qsco:2}).

In this thesis, words that introduce universal quantifiers were
deliberately excluded from the linguistic coverage (section
\ref{ling_not_supported}). This leaves only existential quantification
and by-passes the quantifier scoping problem, because if all
quantifiers are existential ones, the relative scoping of the
quantifiers does not matter. (In \topl, existential quantification is
expressed using free variables. There are also interrogative and
interrogative-maximal quantifiers, but these are in effect existential
quantifiers that have the additional side-effect of including the
values of their variables in the answer.)

If the linguistic coverage of the prototype \nlitdb were to be
extended to support words that introduce universal quantification, an
additional scoping module would have to be added (figure
\ref{arch_fig}). The input to that module would be an
``underspecified'' \topl formula (see the discussion in chapter 2 of
\cite{Alshawi}), a formula that would not specify the exact scope of
each quantifier. In a \textsc{Fopl}-like formalism, an underspecified
formula for \pref{qsco:1} could look like \pref{qsco:10}.
\begin{examps}
\item $inspect((\exists x \; guard(x)), (\forall y \; gate(y)))$
  \label{qsco:10}
\end{examps}
The scoping module would generate all the possible scopings, and
determine the most probable ones, producing formulae where the scope
of each quantifier is explicit (e.g.\ \pref{qsco:2} or \pref{qsco:3}).
Alternatively, one could attempt to use the the \hpsg quantifier
scoping mechanism (see \cite{Pollard2} and relevant comments in
section \ref{TOP_FS}) to reason about the possible scopings
during the parsing. That mechanism, however, is a not yet fully
developed part of \hpsg.


\subsection{Anaphora resolution} \label{anaphora_module}

As discussed in sections \ref{no_issues} and \ref{temporal_anaphora},
nominal anaphora (e.g.\ \qit{she}, \qit{his salary}) and most cases of
temporal anaphora (e.g.\ \qit{in January}, tense anaphora) are
currently not supported. An anaphora resolution module would be needed
if phenomena of this kind were to be supported. As in the case of
quantifier scoping, I envisage a module that would accept
``underspecified'' \topl formulae, formulae that would not name
explicitly the entities or times to which anaphoric expressions refer
(figure \ref{arch_fig}). The module would determine the most probable
referents of these expressions, using a discourse model that would
contain descriptions of previously mentioned entities and times,
information showing in which segments of the previous discourse the
entities or times were mentioned, etc.\ (see \cite{Barros1994} for a
description of a similar module). The output of this module would be a
formula that names explicitly the referents of anaphoric expressions.


\subsection{Equivalential translator} \label{equiv_translat}

The translation from \topl to \tsql of chapter \ref{tdb_chapter}
assumes that each pair $\tup{\pi,n}$ of a \topl predicate functor
$\pi$ and an arity $n$ can be mapped to a valid-time relation (stored
directly in the database, or computed from information in the
database) that shows the event times where $\pi(\tau_1, \dots,
\tau_n)$ is true, for all the possible world entities denoted by the
\topl terms $\tau_1, \dots, \tau_n$. The configurer of the \nlitdb
specifies this mapping when defining \hpfunsp (section
\ref{via_TSQL2}). Although the assumption that each $\tup{\pi,n}$ can
be mapped to a suitable relation is valid in most situations, there
are cases where this assumption does not hold. The ``doctor on board''
problem \cite{Rayner93} is a well-known example of such a case.

Let us imagine a database that contains only the following coalesced valid-time
relation, that shows the times when a (any) doctor was on board each ship of
a fleet. 
\adbtable{2}{|l||l|}{$doctor\_on\_board$}
{$ship$ & }
{$Vincent$ & $[8\text{:}30am \; 22/1/96 - 
               11\text{:}45am \; 22/1/96]$ \\
           & $\;\;\union \; [3\text{:}10pm \; 23/1/96 -
                         5\text{:}50pm \; 23/1/96]$ \\
           & $\;\; \union \; [9\text{:}20am \; 24/1/96 -
                         2\text{:}10pm \; 24/1/96]$ \\
 $Invincible$ & $[8\text{:}20am \; 22/1/96 - 
                 10\text{:}15am \; 22/1/96]$ \\
              & $\;\; \union \; [1\text{:}25pm \; 23/1/96 -
                            3\text{:}50pm \; 23/1/96]$ \\
 \; \dots & \; \dots
}
Let us also consider a question like \pref{doct:1}, which would be
mapped to the \topl formula \pref{doct:2}. I assume here that
\qit{doctor} and \qit{ship} introduce predicates of the form
$doctor(\tau_1)$ and $ship(\tau_2)$, and that the predicative
preposition \qit{on} introduces a predicate of the form
$located\_on(\tau_3, \tau_4)$ ($\tau_1, \dots, \tau_4 \in \terms$).
For simplicity, I assume that \qit{doctor} and \qit{ship} do not
introduce \ntense operators (section \ref{non_pred_nps}).
\begin{examps}
\item Is there a doctor on some ship? \label{doct:1}
\item $doctor(d^v) \land ship(s^v) \land \pres[located\_on(d^v,
   s^v)]$ \label{doct:2} 
\end{examps}
To apply the \topl to \tsql translation method of chapter
\ref{tdb_chapter}, one needs to map $\tup{doctor,1}$ to a valid-time
relation (computed from information in the database) that shows the
event times where $doctor(\tau_1)$ is true, i.e.\ when the entity
denoted by $\tau_1$ was a doctor. Unfortunately, the database (which
contains only $doctor\_on\_board$) does not show when particular
entities were doctors, and hence such a relation cannot be computed.
In the same manner, $\tup{ship,1}$ has to be mapped to a relation that
shows the ships that existed at each time. This relation cannot be
computed: $doctor\_on\_board$ does not list all the ships that existed
at each time; it shows only ships that had a doctor on board at each
time. Similarly, $\tup{located\_on,2}$ has to be mapped to a relation
that shows when $located\_on(\tau_3,\tau_4)$ is true, i.e.\ when the
entity denoted by $\tau_3$ was on the entity denoted by $\tau_4$.
Again, this relation cannot be computed. If, for example, $\tau_3$
denotes a doctor (e.g.\ Dr.\ Adams) and $\tau_4$ denotes Vincent,
there is no way to find out when that particular doctor was on
Vincent: $doctor\_on\_board$ shows only when some (any) doctor was on
each ship; it does not show when particular doctors (e.g.\ Dr.\ Adams)
were on each ship. Hence, the translation method of chapter
\ref{tdb_chapter} cannot be used.

It should be easy to see, however, that \pref{doct:2} is equivalent to
\pref{doct:3}, if $doctor\_on\_ship(\tau_5)$ is true at event times
where the entity denoted by $\tau_5$ is a ship, and a doctor of that
time is on that ship. What is interesting about \pref{doct:3} is that
there \emph{is} enough information in the database to map
$\tup{doctor\_on\_ship,1}$ to a relation that shows the event times
where $doctor\_on\_ship(\tau_5)$ holds. Roughly speaking, one simply
needs to map $\tup{doctor\_on\_ship,1}$  to the
$doctor\_on\_board$ relation. 
Hence, the \topl to \tsql translation method of chapter
\ref{tdb_chapter} \emph{can} be applied to \pref{doct:3},
and the answer to \pref{doct:1} can be found by evaluating the
resulting \tsql code. 
\begin{examps}
\item $\pres[doctor\_on\_ship(s^v)]$ \label{doct:3}
\end{examps}
The problem is that \pref{doct:1} cannot be mapped directly to
\pref{doct:3}: the English to \topl mapping of chapter
\ref{English_to_TOP} generates \pref{doct:2}. We need to
convert \pref{doct:2} (whose predicates are introduced by the lexical
entries of nouns, prepositions, etc.) to \pref{doct:3} (whose
predicates are chosen to be mappable to relations computed from
information in the database). An ``equivalential translator''
similar to the ``abductive equivalential translator'' of
\cite{Rayner93} and \cite{Alshawi2} could be used to carry out this
conversion. Roughly speaking, this would be an
inference module that would use domain-dependent conversion rules, like
\pref{doct:4} which allows any formula of the form $doctor(\tau_1)
\land ship(\tau_2) \land \pres[located\_on(\tau_1,\tau_2)]$ ($\tau_1,
\tau_2 \in \terms$) to be replaced by
$\pres[doctor\_on\_ship(\tau_2)]$. \pref{doct:4} would license the
conversion of \pref{doct:2} into \pref{doct:3}.
\begin{examps}
\item $doctor(\tau_1) \land ship(\tau_2) \land
       \pres[located\_on(\tau_1, \tau_2)] \equiv
       \pres[doctor\_on\_ship(\tau_2)]$
       \label{doct:4} 
\end{examps}
There would be two kinds of pairs $\tup{\pi,n}$ ($\pi$ is a predicate
functor and $n$ an arity): pairs that are mapped to relations, and
pairs for which this mapping is impossible (the value of \hpfunsp
would be undefined for the latter). The formula generated after the
scoping and anaphora resolution would be passed to the equivalential
translator (figure \ref{arch_fig}). If all the predicate functors and
arities in the formula are among the pairs that are mapped to
relations, the equivalential translator would have no effect.
Otherwise, the equivalential translator would attempt to convert the
formula into another one that contains only predicate functors and
arities that are mapped to relations (an error would be reported if
the conversion is impossible). The new formula would then be passed to
the \topl to \tsql translator.


\subsection{Response generator} \label{response_generator}

The execution of the \tsql code produces the information that is
needed to answer the user's question. A response generator is needed
to report this information to the user. In the simplest case, if the
question is a yes/no one, the response generator would simply print a
\qit{yes} or \qit{no}, depending on whether or not the \tsql code
retrieved at least one tuple (section \ref{formulation}).  Otherwise,
the response generator would print the tuples retrieved by the \tsql
code.

Ideally, the response generator would also attempt to provide
\emph{cooperative responses} (section \ref{no_issues}; see also
section \ref{to_do} below). In \pref{respgen:1}, for example, if BA737 is at
gate 4, the response generator would produce \pref{respgen:2} rather
than a simple \qit{no}. That is, it would report the answer
to \pref{respgen:1} along with the answer to \pref{respgen:3}.
\begin{examps}
\item Is BA737 at gate 2? \label{respgen:1}
\item \sys{No, BA737 is at gate 4.} \label{respgen:2}
\item Which gate is BA737 at? \label{respgen:3}
\end{examps}
In that case, the architecture of the \nlitdb would have to be more
elaborate than that of figure \ref{arch_fig}, as the response
generator would have to submit questions (e.g.\ \pref{respgen:3}) on
its own, in order to collect the additional information that is needed
for the cooperative responses.


\subsection{Configuration tools}

As already noted, there are several parts of the prototype \nlitdb
that need to be modified whenever the \nlitdb is configured for a new
application. Most large-scale \nlidb{s} provide tools that automate
these modifications, ideally allowing people that are not aware of the
details of the \nlitdb's code to configure the system (see section 6
of \cite{Androutsopoulos1995}, and chapter 11 of \cite{Alshawi}). A
similar tool is needed in the prototype \nlitdb of this thesis. Figure
\ref{arch_fig} shows how this tool would fit into the \nlitdb's
architecture.\footnote{Some of the heuristics of the quantifier
  scoping module and parts of the anaphora resolution module may in
  practice be also domain-dependent. In that case, parts of these
  modules would also have to be modified during the configuration. For
  simplicity, this is not shown in figure \ref{arch_fig}.}


\section{The airport database} 

This section provides more information about the hypothetical airport
database, for which the prototype \nlitdb was configured. 

\begin{figure}
\hrule
\medskip
\begin{center}
$\begin{array}{l}
gates(gate, availability) \\
runways(runway, availability) \\
queues(queue, runway) \\
servicers(servicer) \\
inspectors(inspector) \\
sectors(sector) \\
flights(flight) \\
tanks(tank, content) \\
norm\_departures(flight, norm\_dep\_time, norm\_dep\_gate) \\
norm\_arrivals(flight, norm\_arr\_time, norm\_arr\_gate) \\
norm\_servicer(flight, servicer) \\
flight\_locations(flight, location) \\
circling(flight) \\
inspections(code, inspector, inspected, status) \\
services(code, servicer, flight, status) \\
boardings(code, flight, gate, status) \\
landings(code, flight, runway, status) \\
\mathit{takeoffs}(code, flight, runway, status) \\
taxiings(code, flight, origin, destination, status)
\end{array}$
\caption[Relations of the airport database]{Relations of the airport database}
\label{db_relations}
\end{center}
\hrule
\end{figure}

The airport database contains nineteen relations, all valid-time and
coalesced (section \ref{bcdm}). Figure \ref{db_relations} shows the
names and explicit attributes of the relations. For simplicity, I
assume that the values of all the explicit attributes are strings. I
also assume that chronons correspond to minutes, and that the
$gregorian$ calendric relation of section \ref{calrels} is available.
The $runways$ relation has the following form:
\adbtable{3}{|l|l||l|}{$runways$}
{$runway$ & $\mathit{availability}$ & }
{$runway1$ & $open$   & $[8\text{:}00am \; 1/1/96, \; 7\text{:}30pm \; 3/1/96]$ \\
         &          & $\;\; \union \; 
                       [6\text{:}00am \; 4/1/96, \; 2\text{:}05pm \; 8/1/96] 
                       \; \union \; \dots$ \\
 $runway1$ & $closed$ & $[7\text{:}31pm \; 3/1/96, \; 5\text{:}59am \; 4/1/96]$ \\
         &          & $\;\; \union \; 
                      [2\text{:}06pm \; 8/1/96, \; 5\text{:}45pm \; 8/1/96]$ \\
 $runway2$ & $open$   & $[5\text{:}00am \; 1/1/96, \; 9\text{:}30pm \; 9/1/96]
                         \; \union \; \dots$ \\
 $runway2$ & $closed$ & $[9\text{:}31pm \; 9/1/96, \;  
                        10\text{:}59am \; 10/1/96] \; \union \; \dots$ 
}
The $\mathit{availability}$ values are always $open$ or $closed$.
There are two tuples for each runway: one showing the times when the
runway was open, and one showing the times when it was closed. If a
runway did not exist at some time (e.g.\ a runway may not have been
constructed yet at that time), both tuples of that runway exclude this
time from their time-stamps.  The $gates$ relation is similar. Its
$availability$ values are always $open$ or $closed$, and there are two
tuples for each gate, showing the times when the gate was open
(available) or closed (unavailable) respectively.

Runways that are used for landings or take-offs have queues, where
flights wait until they are given permission to enter the runway. The
$queues$ relation lists the names of the queues that exist at various
times, along with the runways where the queues lead to. The
$servicers$ relation shows the names of the servicing companies that
existed at any time. The $inspectors$, $sectors$, and $flights$
relations are similar. The $tanks$ relation shows the contents
($water$, $foam$, etc., or $empty$ if the tank was empty) of each tank
at every time where the tank existed.

Each outgoing flight is assigned a normal departure time and gate (see
also section \ref{aspect_examples}). The $norm\_departures$ relation
shows these times and gates. For example, if $norm\_departures$ were
as follows, this would mean that from 9:00am on 1/1/92 to 5:30pm on
31/11/95 BA737 normally departed each day from gate 2 at 2:05pm. (For
simplicity, I assume that all flights are daily.) At 5:31pm on
31/11/95, the normal departure time of BA737 was changed to 2:20pm,
while the normal departure gate remained gate 2. No further change to the
normal departure time or gate of BA737 was made since then.
\adbtable{4}{|l|c|c||l|}{$norm\_departures$}
{$flight$ & $norm\_dep\_time$ & $norm\_dep\_gate$ &}
{$BA737$ & $2\text{:}05pm$  & $gate2$ & 
   $[9\text{:}00am \; 1/1/92, \; 5\text{:}30pm \; 31/11/95]$ \\
 $BA737$ & $2\text{:}20pm$  & $gate2$ & 
   $[5\text{:}31pm \; 31/11/95, \; now]$
}
Similarly, each incoming flight is assigned a normal arrival time and
a gate, listed in $norm\_arrivals$. Flights are also assigned normal
servicers, servicing companies that over a period of time normally
service the flights whenever they arrive or depart. This information
is stored in $norm\_servicer$. The
$flight\_locations$ relation shows the location of each flight over
the time. Possible $location$ values are the names of airspace
sectors, gates, runways, or queues of runways. The $circling$ relation
shows the flights that were circling at each time.

As discussed in section \ref{aspect_examples}, flights, gates, and
runways are occasionally inspected. The $inspections$ relation was
discussed in section \ref{via_TSQL2}. It shows the inspection code,
inspector, inspected object, status (completed or not), and time of
each inspection. The $services$, $boardings$, $landings$,
$\mathit{takeoffs}$, and $taxiings$ relations are very similar. They
provide information about actual services, boardings, landings,
take-offs, and taxiings from one location ($origin$) to another
($destination$). Each service, boarding, landing, take-off, or taxiing
is assigned a unique code, stored in the $code$ attribute. The
$status$ attribute shows if the climax is reached at the latest
time-point of the time-stamp. The values of the $origin$ and
$destination$ attributes of $taxiings$ are names of gates, runways,
and queues.

Apart from relations, a database would in practice also contain
\emph{integrity constraints} (see \cite{Gertz1995} and
\cite{Wijsen1995}). There would be, for example, a constraint saying
that if the $circling$ relation shows a flight as circling at some
time, the $flights$ relation must show that flight as existing at the
same time. I do not discuss integrity constraints, as they are not
directly relevant to this thesis.


\section{Sample questions and output} \label{samples}

This section presents sample questions from the airport domain, along
with the corresponding output of the prototype \nlitdb. The questions
are chosen to demonstrate that the \nlitdb behaves according to the
specifications of the previous chapters. The questions are \emph{not}
intended to be (and are probably not) a representative sample of
questions that a real user might want to submit in the airport domain
(see comments about Wizard of Oz experiments in section \ref{wizard}
below). The user submits questions using the \texttt{nli} Prolog
predicate:

\ssmall
\begin{verbatim}
| ?- nli([which,flight,left,sector3,at,pm5_00,yesterday]).
\end{verbatim}

\dnorm
The system parses the question and reports the generated \hpsg sign.

\ssmall
\begin{verbatim}
HPSG Sign:

(phrase,
 qstore:(ne_set_quant,
         elt:(det:exists,
              restind:(index:(_10148,
                              minute_ent,
                              tvar:plus),
                       restr:(ne_set_psoa,
                              elt:(part,
                                   part_var:_10148,
                                   partng:pm5_00),
                              elts:e_set))),
         elts:(ne_set_quant,
               elt:(det:interrog,
                    restind:(index:(_10486,
                                    flight_ent,
                                    tvar:plus),
                             restr:(ne_set_psoa,
                                    elt:(flight,
                                         arg1:_10486),
                                    elts:e_set))),
               elts:e_set)),
 synsem:(synsem,
         loc:(cat:(aspect:point,
                   comps:e_list,
                   head:(aux:minus,
                         inv:minus,
                         mod:none,
                         prd:minus,
                         vform:fin),
                   spr:e_list,
                   subj:e_list),
              cont:(at_op,
                    main_psoa:(at_op,
                               main_psoa:(past,
                                          et_handle:(temp_ent,
                                                     tvar:plus),
                                          main_psoa:(leave_something,
                                                     arg1:_10486,
                                                     arg2:sector3)),
                               time_spec:_10148),
                    time_spec:yesterday)),
         nonloc:(inherited:slash:e_set,
                 to_bind:slash:e_set)))
\end{verbatim}

\dnorm The sign above is written in \ale's notation. The sign is of
sort {\srt phrase}\/ (it corresponds to a phrase rather than a single
word), and it has the features {\feat qstore} and {\feat synsem}. The
{\feat qstore} value represents a non-empty set of quantifiers ({\srt
  ne\_set\_quant}). Its {\feat elt} feature describes the first
element of that set, which is an existential quantifier. The
quantifier ranges over a \topl variable, represented by an \hpsg index
of sort {\srt minute\_ent} (see figure \vref{ind_hierarchy}) whose
{\feat tvar} is $+$ (the index represents a \topl variable rather than
a constant). The {\feat elt} value represents the \topl-like
expression $\exists \, x2^v \; \partop[pm5\_00^g, x2^v]$. The Prolog
variable \texttt{\_10148} is a pointer to the index of the quantifier,
i.e.\ it plays the same role as the boxed numbers (e.g.\ \avmbox{1},
\avmbox{2}) in the \hpsg formalism of chapter \ref{English_to_TOP}.
The {\feat elts} value describes the rest of the set of quantifiers,
using in turn an {\feat elt} feature (second element of the overall
set), and an {\feat elts} feature (remainder of the set, in this case
the empty set). The second element of the overall set represents the
\topl expression $?x1^v \; flight(x1^v)$. In the airport application,
the lexical entries of non-predicative nouns do not introduce \ntense
operators (this generates appropriate readings in most cases; see the
discussion in section \ref{non_pred_nps}). This is why no \ntense
operator is present in the second quantifier of the sign. (The effect
of \ntense{s} can still be seen in the airport application in the
case of non-predicative adjectives, that do introduce \ntense{s}.)

The features of the {\feat synsem} value are as in chapter
\ref{English_to_TOP}. The {\feat cont} value represents the \topl
expression $\at[yesterday, \at[x2^v, \past[x3^v,
leave\_something(x1^v, sector3)]]]$. The extractor of section
\ref{extraction_impl} extracts a \topl formula from the sign, and
prints it as a Prolog term.

\ssmall
\begin{verbatim}
TOP formula extracted from HPSG sign:

interrog(x1^v,
         and(part(pm5_00^g, x2^v),
             and(flight(x1^v),
                 at(yesterday,
                    at(x2^v,
                       past(x3^v,
                            leave_something(x1^v, sector3)))))))
\end{verbatim}

\dnorm
The Prolog term above stands for:
\begin{examps}
\item[] $?x1^v \;
  \partop[pm5\_00^g, x2^v] \land flight(x1^v) \; \land$ \\
  $\at[yesterday,\at[x2^v, \past[x3^v, leave\_something(x1^v, sector3)]]]$
\end{examps}
The extracted formula then goes through the post-processor of section
\ref{extraction_impl}. The post-processor
eliminates the \partop operator, adding the $pm5\_00^g$ as an
extra argument to the corresponding \at operator:

\ssmall
\begin{verbatim}
TOP formula after post-processing:

interrog(x1^v,
         and(flight(x1^v),
             at(yesterday,
                at(pm5_00^g, x2^v,
                   past(x3^v,
                        leave_something(x1^v, sector3))))))
\end{verbatim}

\dnorm
The Prolog term above stands for:
\begin{examps}
\item $?x1^v \; flight(x1^v)  \land \at[yesterday,$ \\
   $\at[pm5\_00^g, x2^v, \past[x3^v, leave\_something(x1^v, sector3)]]]$
\label{log:1}
\end{examps}

\doublespace
The post-processed formula is then translated into \tsql:

\ssmall
\begin{verbatim}
TSQL2 query:

(SELECT DISTINCT SNAPSHOT t8.1
 FROM (SELECT DISTINCT t6.1, t7.1, t7.2, t7.3, t7.4
       VALID VALID(t6)
       FROM (SELECT DISTINCT t1.1
             VALID VALID(t1)
             FROM (SELECT DISTINCT fl.1
                   VALID VALID(fl)
                   FROM flights(PERIOD) AS fl
                  )(SUBPERIOD) AS t1
             WHERE PERIOD(TIMESTAMP 'beginning', TIMESTAMP 'forever')
                   CONTAINS VALID(t1)
            ) AS t6,
            (SELECT DISTINCT t2.1, t5.1, t5.2, t5.3
             VALID VALID(t5)
             FROM (SELECT DISTINCT SNAPSHOT VALID(cp2)
                   FROM gregorian AS cp2
                   WHERE cp2.5 = '17' AND cp2.6 = '00'
                  ) AS t2, 
                  (SELECT DISTINCT VALID(t4), t4.1, t4.2
                   VALID VALID(t4)
                   FROM (SELECT DISTINCT t3.1, t3.2
                         VALID VALID(t3)
                         FROM (SELECT DISTINCT flocs.1, flocs.2
                               VALID PERIOD(END(VALID(flocs)), 
                                            END(VALID(flocs)))
                               FROM flight_locations(PERIOD) AS flocs
                              )(SUBPERIOD) AS t3
                         WHERE t3.2 = 'sector3'
                           AND INTERSECT(
                                 INTERSECT(t2.1, 
                                   INTERSECT(
                                      PERIOD(TIMESTAMP 'beginning', 
                                             TIMESTAMP 'forever'), 
                                      PERIOD 'today' - INTERVAL '1' DAY)), 
                                 PERIOD(TIMESTAMP 'beginning', 
                                        TIMESTAMP 'now' - 
                                          INTERVAL '1' MINUTE))
                               CONTAINS VALID(t3)
                        ) AS t4
                  ) AS t5
            ) AS t7
       WHERE t6.1 = t7.3
         AND VALID(t6) = VALID(t7)
      ) AS t8
)
\end{verbatim}

\dnorm The ``\sql{SELECT DISTINCT fl.1}~\dots \sql{FROM
  flights(PERIOD) AS fl}'' that starts at the sixth line of the \tsql
code is the \sql{SELECT} statement to which \hpfunsp maps predicates
of the form $flight(\tau_1)$. This statement returns a relation that
shows the flights that existed at each time. The embedded \sql{SELECT}
statement that is associated with the correlation name \sql{t6} is the
result of applying the translation rule for predicates (section
\ref{trans_rules}) to the $flight(x1^v)$ of \pref{log:1}. The
``\sql{WHERE PERIOD(TIMESTAMP 'beginning', TIMESTAMP 'forever')
  CONTAINS VALID(t1)}'' corresponds to the restriction that $et$ must
fall within $lt$. (At this point, no constraint has been imposed on
$lt$, and hence $lt$ covers the whole time-axis.) This \sql{WHERE}
clause has no effect and could be removed during an optimisation phase
(section \ref{tsql2_opt}).

The ``\sql{SELECT DISTINCT flocs.1}~\dots
\sql{flight\_locations(PERIOD) AS flocs}'' that starts at the 23rd
line of the \tsql code is the \sql{SELECT} statement to which \hpfunsp
maps predicates of the form $leave\_something(\tau_1, \tau_2)$. This
statement generates a relation that for each flight and location,
shows the end-points of maximal periods where the flight was at that
location. The embedded \sql{SELECT} statement that is associated with
\sql{t4} is the result of applying the translation rule for predicates
to the $leave\_something(x1^v, sector3)$ of \pref{log:1}.
\sql{VALID(t3)} is the leaving-time, which has to fall within $lt$.
The three nested \sql{INTERSECT}s represent constraints that have been
imposed on $lt$: the \past operator requires $lt$ to be a subperiod of
$[p_{first}, st)$ (i.e.\ a subperiod of \sql{TIMESTAMP 'beginning',
  TIMESTAMP 'now' - INTERVAL '1' MINUTE}), the $\at[pm5\_00^g, \dots]$
requires $lt$ to be a subperiod of a 5:00pm-period (\sql{t2.1} ranges
over 5:00pm-periods), and the $\at[yesterday,\dots]$ requires the
localisation time to be a subperiod of the previous day (\sql{PERIOD
'today' - INTERVAL '1' DAY}).

The \sql{SELECT} statement that is associated with \sql{t5} is
generated by the translation rule for \past (section
\ref{trans_rules}), and the \sql{SELECT} statement that is associated
with \sql{t7} is introduced by the translation rule for $\at[\sigma_g,
\beta, \phi']$ (section \ref{atsg_rule}). (The $\at[yesterday, \dots]$
of \pref{log:1} does not introduce its own \sql{SELECT} statement, it
only restricts $lt$; see the translation rule for $\at[\kappa, \phi']$
in section \ref{trans_rules}.)  The \sql{SELECT} statement that is
associated with \sql{t8} is introduced by the translation rule for
conjunction (section \ref{conj_rule}). It requires the attribute
values that correspond to the $x1^v$ arguments of $flight(x1^v)$ and
$leave\_something(x1^v, sector3)$, and the event times where the two
predicates are true to be identical. Finally, the top-level
\sql{SELECT} statement is introduced by the translation rule for
$?\beta_1 \; ?\beta_2 \; ?\beta_3 \; \dots \; ?\beta_k \; \phi'$ (section
\ref{wh1_rule}). It returns a snapshot relation that
contains the attribute values that correspond to $x1^v$ (the flights).

No further comments need to be made about the generated \hpsg signs
and \tsql queries. To save space, I do not show these in the rest of
the examples. I also do not show the \topl formulae before the
post-processing, unless some point needs to be made about them.

As noted in section \ref{progressives}, no attempt is made to block
progressive forms of state verbs. The progressive forms of these verbs
are taken to have the same meanings as the corresponding
non-progressive ones. This causes the two questions below to receive
the same \topl formula.

\ssmall
\begin{verbatim}
| ?- nli([which,tanks,contain,water]).

TOP formula after post-processing:

interrog(x1^v,
         and(tank(x1^v),
             pres(contains(x1^v, water))))

| ?- nli([which,tanks,are,containing,water]).

TOP formula after post-processing:

[same formula as above]
\end{verbatim}

\dnorm There are two lexical entries for the base form of \qit{to
service}, one for the habitual homonym, and one for the non-habitual
one. The habitual entry introduces the predicate functor
$hab\_servicer\_of$ and classifies the base form as state. The
non-habitual entry introduces the functor
$actl\_servicing$ and classifies the base form as culminating
activity. The simple present lexical rule (section
\ref{single_word_forms}) generates a simple present lexical entry for
only the habitual homonym (whose base form is state). Hence, the
\qit{services} below is treated as the simple present of the habitual
homonym (not as the simple present of the non-habitual homonym), and
only a formula that contains the $hab\_servicer\_of$ functor is
generated. This captures the fact that the question
can only have a habitual meaning (it cannot refer to a servicer that is
actually servicing BA737 at the present; the reader is reminded that
the scheduled-to-happen reading of the simple present is ignored in
this thesis -- see section \ref{simple_present}). 

\ssmall
\begin{verbatim}
| ?- nli([which,servicer,services,ba737]).

TOP formula after post-processing:

interrog(x1^v,
         and(servicer(x1^v),
             pres(hab_servicer_of(x1^v, ba737))))
\end{verbatim}

\dnorm In contrast, the present participle lexical rule (section
\ref{single_word_forms}) generates progressive entries for both the
non-habitual (culminating activity base form) and the habitual (state
base form) homonyms. This causes the question below to receive two
parses, one where the \qit{is servicing} is the present continuous of
the non-habitual homonym, and one where it is the
present continuous of the habitual homonym. This gives
rise to two formulae, one involving the $actl\_servicing$
functor (the servicer must be servicing BA737 at the present),
and one involving the $hab\_servicer\_of$ functor (the servicer must
be the current normal servicer of BA737). (The $x2^v$ in the first formula is
an occurrence identifier; see section \ref{occurrence_ids}.) The
habitual reading of the second formula seems rather unlikely in this
case.

\ssmall
\begin{verbatim}
| ?- nli([which,servicer,is,servicing,ba737]).

TOP formula after post-processing:

interrog(x1^v,
         and(servicer(x1^v),
             pres(actl_servicing(x2^v, x1^v, ba737))))

TOP formula after post-processing:

interrog(x1^v,
         and(servicer(x1^v),
             pres(hab_servicer_of(x1^v, ba737))))
\end{verbatim}

\dnorm There are also different lexical entries for the actual \qit{to
  depart} and the habitual \qit{to depart} (at some time). The
habitual entry introduces the functor $hab\_dep\_time$, requires an
\qit{at~\dots} complement, and classifies the base form as state.  The
non-habitual entry introduces the functor $actl\_depart$, requires no
complement, and classifies the base form as point. When \qit{BA737
  departed at 5:00pm.} is taken to involve the habitual homonym,
\qit{at 5:00pm} is treated as the complement that specifies the
habitual departure time (the second argument of
$hab\_dep\_time(\tau_1, \tau_2)$). When the sentence is taken to
involve the non-habitual homonym, \qit{at 5:00pm} is treated as a
temporal modifier, and it introduces an \at operator (section
\ref{hpsg:punc_adv}). In the following question, this analysis leads
to two formulae: one where each reported flight must have actually
departed at 5:00pm at least once in 1993, and one where the habitual
departure time of each reported flight must have been 5:00pm some time
in 1993. The second reading seems the preferred one in this example.

\ssmall
\begin{verbatim}
| ?- nli([which,flights,departed,at,pm5_00,in,y1993]).

TOP formula after post-processing:

interrog(x1^v,
         and(flight(x1^v),
             at(y1993,
                at(pm5_00^g, x2^v,
                   past(x3^v,
                        actl_depart(x1^v))))))

TOP formula after post-processing:

interrog(x1^v,
         and(flight(x1^v),
             at(y1993,
                past(x2^v,
                     hab_dep_time(x1^v, pm5_00)))))
\end{verbatim}

\dnorm The first question below receives no parse, because \qit{to
  circle} is classified as activity verb (there is no habitual state
homonym in this case), and the simple present lexical rule does not
generate simple present lexical entries for activity verbs. In
contrast, the present participle lexical rule does generate
progressive entries for activity verbs. This causes the second
question below to be mapped to the formula one would expect. The
failure to parse the first question is justified, in the sense that
the question seems to be asking about flights that have some circling
habit, and the \nlitdb has no access to information on circling
habits. A more cooperative response, however, is needed to
explain this to the user.

\ssmall
\begin{verbatim}
| ?- nli([does,ba737,circle]).

  **No (more) parses.

| ?- nli([is,ba737,circling]).

TOP formula after post-processing:

pres(circling(ba737))
\end{verbatim}

\dnorm Following the arrangements of section \ref{hpsg:per_advs}, in
the following question where a culminating activity combines with a
period adverbial, two formulae are generated: one where the inspection
must have simply been completed on 1/5/92, and one where the whole
inspection (from start to completion) must have been carried out on
1/5/92. The first reading seems unlikely in this example, though as
discussed in section \ref{period_adverbials}, there are sentences
where the first reading is the intended one.

\ssmall
\begin{verbatim}
| ?- nli([who,inspected,uk160,on,d1_5_92]).

TOP formula after post-processing:

interrog(x1^v,
         at(d1_5_92,
            end(past(x2^v,
                     culm(inspecting(x3^v, x1^v, uk160))))))

TOP formula after post-processing:

interrog(x1^v,
         at(d1_5_92,
            past(x2^v,
                 culm(inspecting(x3^v, x1^v, uk160)))))
\end{verbatim}

\dnorm

In the following question, the punctual adverbial \qit{at 5:00pm}
combines with a culminating activity. According to section
\ref{point_adverbials}, two readings arise: one where the taxiing
starts at 5:00pm, and one where it finishes at 5:00pm. In both cases,
the punctual adverbial causes the aspect of \qit{which flight taxied
  to gate 2 at 5:00pm} to become point. That point sentence then
combines with the period adverbial \qit{yesterday}.  According to
section \ref{period_adverbials}, the instantaneous situation of the
point phrase (the start or end of the taxiing) must occur within the
period of the adverbial. This analysis leads to two formulae: one
where the taxiing starts at 5:00pm on the previous day, and one where
the taxiing finishes at 5:00pm on the previous day. These formulae
capture the most likely readings of the question. Unfortunately, if
the order of \qit{at 5:00pm} and \qit{yesterday} is reversed, the
generated formulae are not equivalent to the ones below (see the
discussion in section \ref{hpsg:mult_mods})

\ssmall
\begin{verbatim}
| ?- nli([which,flight,taxied,to,gate2,at,pm5_00,yesterday]).

TOP formula after post-processing:

interrog(x1^v,
         and(flight(x1^v),
             at(yesterday,
                at(pm5_00^g, x2^v,
                   end(past(x3^v,
                            culm(taxiing_to(x4^v, x1^v, gate2))))))))
\end{verbatim}

\newpage
\begin{verbatim}
TOP formula after post-processing:

interrog(x1^v,
         and(flight(x1^v),
             at(yesterday,
                at(pm5_00^g, x2^v,
                   begin(past(x3^v,
                              culm(taxiing_to(x4^v, x1^v, gate2))))))))
\end{verbatim}

\dnorm In the sentence below (which is treated as a yes/no question),
the treatment of past perfects and punctual adverbials of section
\ref{hpsg:punc_adv} allows \qit{at 5:00pm} to modify either the verb
phrase \qit{left gate 2}, or the entire \qit{BA737 had left gate 2}.
This gives rise to two \topl formulae: one where 5:00pm is the time at
which BA737 left gate 2, and one where 5:00pm is a reference time at
which BA737 had already left gate 2. The two formulae capture the two
most likely readings of the sentence.

\ssmall
\begin{verbatim}
| ?- nli([ba737,had,left,gate2,at,pm5_00]).

TOP formula after post-processing:

past(x2^v,
     perf(x3^v,
          at(pm5_00^g, x1^v,
             leave_something(ba737, gate2))))

TOP formula after post-processing:

at(pm5_00^g, x1^v,
   past(x2^v,
        perf(x3^v,
             leave_something(ba737, gate2))))
\end{verbatim}

\dnorm Similarly, in the following question, the \qit{at 5:00pm} is
allowed to modify either the verb phrase \qit{taken off}, or the
entire \qit{BA737 had taken off}. In the first case, the verb phrase
still has the aspectual class of the base form, i.e.\ culminating
activity. According to section \ref{point_adverbials}, 5:00pm is the
time where the taking off was completed or started. These two readings
are captured by the first and second formulae below. (The second
reading seems unlikely in this example.) In the case where \qit{at
  5:00pm} modifies the entire \qit{BA737 had taken off}, the \qit{had}
has already caused the aspect of \qit{BA737 had taken off} to become
(consequent) state. According to section \ref{point_adverbials}, in
that case 5:00pm is simply a time-point where the situation of the
sentence (having departed) holds.  This reading is captured by the
third formula.

\ssmall
\begin{verbatim}
| ?- nli([ba737,had,taken,off,at,pm5_00]).

TOP formula after post-processing:

past(x2^v,
     perf(x3^v,
          at(pm5_00^g, x1^v,
             end(culm(taking_off(x4^v, ba737))))))

TOP formula after post-processing:

past(x2^v,
     perf(x3^v,
          at(pm5_00^g, x1^v,
             begin(culm(taking_off(x4^v, ba737))))))

TOP formula after post-processing:

at(pm5_00^g, x1^v,
   past(x2^v,
        perf(x3^v,
             culm(taking_off(x4^v, ba737)))))
\end{verbatim}

\dnorm The first question below receives the formula one would expect.
As discussed in section \ref{hpsg:mult_mods}, in the second question
below the grammar of chapter \ref{English_to_TOP} allows two parses:
one where \qit{yesterday} attaches to \qit{BA737 was circling}, and
one where \qit{yesterday} attaches to \qit{BA737 was circling for two
  hours}. These two parses give rise to two different but logically
equivalent formulae.

\ssmall
\begin{verbatim}
| ?- nli([ba737,was,circling,for,two,hours,yesterday]).

TOP formula after post-processing:

at(yesterday,
   for(hour^c, 2,
       past(x1^v,
            circling(ba737))))

| ?- nli([yesterday,ba737,was,circling,for,two,hours]).

TOP formula after post-processing:

for(hour^c, 2,
    at(yesterday,
       past(x1^v,
            circling(ba737))))

TOP formula after post-processing:

at(yesterday,
   for(hour^c, 2,
       past(x1^v,
            circling(ba737))))
\end{verbatim}

\dnorm The following example reveals a problem in the current
treatment of temporal modifiers. The \hpsg version of this thesis
(section \ref{hpsg:pupe_adv}) allows temporal modifiers to attach only
to finite sentences (finite verb forms that have already combined with
their subjects and complements) or past participle verb phrases (past
participles that have combined with all their complements but not
their subjects). In both cases, the temporal modifier attaches after
the verb has combined with all its complements. English temporal
modifiers typically appear either at the beginning or the end of the
sentence (not between the verb and its complements), and hence
requiring temporal modifiers to attach after the verb has combined with
its complements is in most cases not a problem. However, in the
following question (which most native English speakers find acceptable) the
temporal modifier (\qit{for two hours}) is between the verb
(\qit{queued}) and its complement (\qit{for runway2}). Therefore,
the temporal modifier cannot attach to the verb after the verb
has combined with its complement, and the system fails to parse
the sentence. (In contrast, \qit{UK160 queued for runway 2 for two
  hours.}, where the temporal modifier follows the complement, is parsed
without problems.)

\ssmall
\begin{verbatim}
| ?- nli([uk160,queued,for,two,hours,for,runway2]).

  **No (more) parses.
\end{verbatim}

\dnorm As explained in section \ref{post_processing}, the
post-processing removes \culm operators that are within \for operators
introduced by \qit{for~\dots} adverbials. This is demonstrated in the
following example. The \qit{for~\dots} adverbial introduces a
\texttt{for\_remove\_culm} pseudo-operator, which can be thought of as
a \for operator with a flag attached to it, that signals that \culm{s}
within the \for operator must be removed. The post-processor
removes the \culm, and replaces the
\texttt{for\_remove\_culm} with an ordinary \for operator.

\ssmall
\begin{verbatim}
| ?- nli([which,flight,boarded,for,two,hours]).

TOP formula extracted from HPSG sign:

interrog(x1^v,
         and(flight(x1^v),
             for_remove_culm(hour^c, 2,
                             past(x2^v,
                                  culm(boarding(x3^v, x1^v))))))

TOP formula after post-processing:

interrog(x1^v,
         and(flight(x1^v),
             for(hour^c, 2,
                 past(x2^v,
                      boarding(x3^v, x1^v)))))
\end{verbatim}

\dnorm Duration \qit{in~\dots} adverbials introduce \for operators
that carry no flag to remove enclosed \culm{s}. In the
following question, this leads to a formula that (correctly)
requires the boarding to have been completed.

\ssmall
\begin{verbatim}
| ?- nli([which,flight,boarded,in,two,hours]).

TOP formula after post-processing:

interrog(x1^v,
         and(flight(x1^v),
             for(hour^c, 2,
                 past(x2^v,
                      culm(boarding(x3^v, x1^v))))))
\end{verbatim}

\dnorm As explained in section \ref{present_perfect}, the present
perfect is treated in exactly the same way as the simple past. This
causes the two questions below to receive the same formula. 

\ssmall
\begin{verbatim}
| ?- nli([which,flight,has,been,at,gate2,for,two,hours]).

TOP formula after post-processing:

interrog(x1^v,
         and(flight(x1^v),
             for(hour^c, 2,
                 past(x2^v,
                      located_at(x1^v, gate2)))))

| ?- nli([which,flight,was,at,gate2,for,two,hours]).

TOP formula after post-processing:

[same formula as above]
\end{verbatim}

\dnorm
As discussed in section \ref{special_verbs}, when \qit{finished}
combines with a culminating activity, the situation must have reached
its completion. In contrast, when \qit{stopped} combines with a
culminating activity, the situation must have simply stopped, without
necessarily reaching its completion. This difference is captured in
the two formulae below by the existence or absence of a \culm.

\ssmall
\begin{verbatim}
| ?- nli([j_adams,finished,inspecting,uk160,at,pm5_00]).

TOP formula after post-processing:

at(pm5_00^g, x1^v,
   past(x2^v,
        end(culm(inspecting(x3^v, j_adams, uk160)))))

| ?- nli([j_adams,stopped,inspecting,uk160,at,pm5_00]).

TOP formula after post-processing:

at(pm5_00^g, x1^v,
   past(x2^v,
        end(inspecting(x3^v, j_adams, uk160))))
\end{verbatim}

\dnorm In the airport domain, non-predicative adjectives (like
\qit{closed} below) introduce \ntense operators. In the question
below, the formula that is extracted from the \hpsg sign contains an
\ntense whose first argument is a variable. As explained in section
\ref{post_processing}, this leads to two different formulae after the
post-processing, one where \qit{closed} refers to the present, and one
where \qit{closed} refers to the time of the verb tense.

\ssmall
\begin{verbatim}
| ?- nli([was,any,flight,on,a,closed,runway,yesterday]).

TOP formula extracted from HPSG sign:

and(flight(x1^v),
    and(and(ntense(x2^v, closed(x3^v)),
            runway(x3^v)),
        at(yesterday,
           past(x4^v,
                located_at(x1^v, x3^v)))))

  **Post processing of TOP formula generated 2 different formulae.

TOP formula after post-processing:

and(flight(x1^v),
    and(and(ntense(now, closed(x3^v)),
            runway(x3^v)),
        at(yesterday,
           past(x4^v,
                located_at(x1^v, x3^v)))))

TOP formula after post-processing:

and(flight(x1^v),
    and(and(ntense(x4^v, closed(x3^v)),
            runway(x3^v)),
        at(yesterday,
           past(x4^v,
                located_at(x1^v, x3^v)))))
\end{verbatim}

\dnorm In the following question, the \qit{currently} clarifies that
\qit{closed} refers to the present. The \ntense in the formula
extracted from the \hpsg sign has $now^*$ as its first
argument. The post-processing has no effect.

\ssmall
\begin{verbatim}
| ?- nli([was,any,flight,on,a,currently,closed,runway,yesterday]).

TOP formula extracted from HPSG sign:

and(flight(x1^v),
    and(and(ntense(now, closed(x2^v)),
            runway(x2^v)),
        at(yesterday,
           past(x3^v,
                located_at(x1^v, x2^v)))))
\end{verbatim}

\dnorm In the following question, the verb tense refers to the
present, and hence \qit{closed} can only refer to a currently closed
runway. The post-processor generates only one formula,
where the first argument of \ntense is $now^*$.

\ssmall
\begin{verbatim}
| ?- nli([is,any,flight,on,a,closed,runway]).

TOP formula extracted from HPSG sign:

and(flight(x1^v),
    and(and(ntense(x2^v, closed(x3^v)),
            runway(x3^v)),
        pres(located_at(x1^v, x3^v))))

TOP formula after post-processing:

and(flight(x1^v),
    and(and(ntense(now, closed(x3^v)),
            runway(x3^v)),
        pres(located_at(x1^v, x3^v))))
\end{verbatim}

\dnorm Predicative adjectives do not introduce \ntense{s} (section
\ref{hpsg:adjectives}), and \topl predicates introduced by these
adjectives always end up within the operator(s) of the verb
tense. This captures the fact that predicative adjectives always refer
to the time of the verb tense.

\ssmall
\begin{verbatim}
| ?- nli([was,gate2,open,on,monday]).

TOP formula after post-processing:

at(monday^g, x1^v
   past(x2^v,
        open(gate2)))
\end{verbatim}

\dnorm
For reasons explained in section \ref{pred_nps}, the system fails to
parse sentences that contain proper names or names of days, months,
etc.\ when these are used as predicative noun phrases (e.g.\ the first
two questions below). Other predicative noun phrases pose no problem
(e.g.\ the third question below).

\ssmall
\begin{verbatim}
| ?- nli([d1_1_91,was,a,monday]).

  **No (more) parses.

| ?- nli([ba737,is,uk160]).

  **No (more) parses.

| ?- nli([ba737,is,a,flight]).

TOP formula after post-processing:

pres(flight(ba737))
\end{verbatim}

\dnorm
Multiple interrogative words can be handled, as demonstrated below.

\ssmall
\begin{verbatim}
| ?- nli([which,flight,is,at,which,gate]).

TOP formula after post-processing:

interrog(x1^v,
         interrog(x2^v,
                  and(gate(x1^v),
                      and(flight(x2^v),
                          pres(located_at(x2^v, x1^v))))))
\end{verbatim}

\dnorm
In the first question below, the grammar of chapter
\ref{English_to_TOP} allows \qit{yesterday} to attach to either 
\qit{BA737 was circling} or to the whole \qit{did any flight leave a
gate while BA737 was circling}. Two \hpsg signs are generated as a
result of this, from which two different but logically equivalent
formulae are extracted. In contrast, in the second question below, the
\qit{yesterday} cannot attach to \qit{BA737 was circling}, because of
the intervening \qit{while} (\qit{while BA737 was circling} is treated
as an adverbial, and \qit{yesterday} cannot attach to another
adverbial). Consequently, only one formula is generated.

\ssmall
\begin{verbatim}
| ?- nli([did,any,flight,leave,a,gate,while,ba737,was,circling,yesterday]).

TOP formula after post-processing:

and(flight(x1^v),
    and(gate(x2^v),
        at(at(yesterday,
              past(x3^v,
                   circling(ba737))),
           past(x4^v,
                leave_something(x1^v, x2^v)))))

TOP formula after post-processing:

and(flight(x1^v),
    and(gate(x2^v),
        at(yesterday,
           at(past(x3^v,
                   circling(ba737)),
              past(x4^v,
                   leave_something(x1^v, x2^v))))))

| ?- nli([did,any,flight,leave,a,gate,yesterday,while,ba737,was,circling]).

TOP formula after post-processing:

and(flight(x1^v),
    and(gate(x2^v),
        at(past(x3^v,
                circling(ba737)),
           at(yesterday,
              past(x5^v,
                   leave_something(x1^v, x2^v))))))
\end{verbatim}

\dnorm In the questions below, the subordinate clause is a
(progressive) state. According to section \ref{before_after_clauses},
in the first question the flights must have arrived before a
time-point where BA737 started to board (\qit{to arrive} is a point
verb in the airport domain). In the second question, section
\ref{before_after_clauses} allows two readings: the flights must
have arrived after a time-point where BA737 started or stopped
boarding. The generated formulae capture these readings.

\ssmall
\begin{verbatim}
| ?- nli([which,flights,arrived,before,ba737,was,boarding]).

TOP formula after post-processing:

interrog(x1^v,
         and(flight(x1^v),
             before(past(x2^v,
                         boarding(x3^v, ba737)),
                    past(x4^v,
                         actl_arrive(x1^v)))))

| ?- nli([which,flights,arrived,after,ba737,was,boarding]).

TOP formula after post-processing:

interrog(x1^v,
         and(flight(x1^v),
             after(begin(past(x2^v,
                              boarding(x3^v, ba737))),
                   past(x4^v,
                        actl_arrive(x1^v)))))

TOP formula after post-processing:

interrog(x1^v,
         and(flight(x1^v),
             after(past(x2^v,
                        boarding(x3^v, ba737)),
                   past(x4^v,
                        actl_arrive(x1^v)))))

\end{verbatim}

\dnorm Below, the subordinate clause is a culminating activity. In the
first question, according to section \ref{before_after_clauses} the
flights must have arrived before a time-point where BA737
finished or started to board. In the second question,
the flights must have arrived after a time-point where BA737
finished boarding. These readings are captured by the generated
formulae.

\ssmall
\begin{verbatim}
| ?- nli([which,flights,arrived,before,ba737,boarded]).

TOP formula after post-processing:

interrog(x1^v,
         and(flight(x1^v),
             before(end(past(x2^v,
                             culm(boarding(x3^v, ba737)))),
                    past(x4^v,
                         actl_arrive(x1^v)))))

TOP formula after post-processing:

interrog(x1^v,
         and(flight(x1^v),
             before(past(x2^v,
                         culm(boarding(x3^v, ba737))),
                    past(x4^v,
                         actl_arrive(x1^v)))))

| ?- nli([which,flights,arrived,after,ba737,boarded]).

TOP formula after post-processing:

interrog(x1^v,
         and(flight(x1^v),
             after(past(x2^v,
                        culm(boarding(x3^v, ba737))),
                   past(x4^v,
                        actl_arrive(x1^v)))))

\end{verbatim}

\dnorm In the next two questions, the subordinate clause is a
consequent state. According to section \ref{before_after_clauses}, in
the first question the flights must have arrived before the situation
of the subordinate clause (having boarded) began, i.e.\ before BA737
finished boarding. In the second question, the flights must have
arrived after the situation of the subordinate clause (having boarded)
began, i.e.\ after BA737 finished boarding. These readings are
captured by the generated \topl formulae. 

\ssmall
\begin{verbatim}
| ?- nli([which,flights,arrived,before,ba737,had,boarded]).


TOP formula after post-processing:

interrog(x1^v,
         and(flight(x1^v),
             before(past(x2^v,
                         perf(x3^v,
                              culm(boarding(x4^v,
                                            ba737)))),
                    past(x5^v,
                         actl_arrive(x1^v))))))


| ?- nli([which,flights,arrived,after,ba737,had,boarded]).

TOP formula after post-processing:

interrog(x1^v,
         and(flight(x1^v),
             after(begin(past(x2^v,
                              perf(x3^v,
                                   culm(boarding(x4^v,
                                                 ba737))))),
                   past(x5^v,
                        actl_arrive(x1^v))))))
\end{verbatim}

\dnorm The question below combines a
\qit{when} interrogative and a \qit{while~\dots} clause. The generated
formula asks for maximal past circling-periods of BA737 that fall
within maximal past periods where UK160 was located at gate 2.

\ssmall
\begin{verbatim}
| ?- nli([when,while,uk160,was,at,gate2,was,ba737,circling]).

TOP formula after post-processing:

interrog_mxl(x3^v,
             at(past(x2^v,
                     located_at(uk160, gate2)),
                past(x3^v,
                     circling(ba737))))
\end{verbatim}

\dnorm Finally, the question below receives two formulae: the first
one asks for times of past actual departures; the second one asks for
past normal departure times. (The latter reading is easier to accept
if an adverbial like \qit{in 1992} is attached.) In the second
question, only a formula for the habitual reading is generated,
because the simple present lexical rule (section
\ref{single_word_forms}) does not generate a simple present lexical
entry for the non-habitual \qit{to depart} (which is a point verb).

\ssmall
\begin{verbatim}
| ?- nli([when,did,ba737,depart]).

TOP formula after post-processing:

interrog_mxl(x2^v,
             past(x2^v,
                  actl_depart(ba737)))

TOP formula after post-processing:

interrog(x1^v,
         past(x2^v,
              hab_dep_time(ba737, x1^v)))

| ?- nli([when,does,ba737,depart]).

TOP formula after post-processing:

interrog(x1^v,
         pres(hab_dep_time(ba737, x1^v)))
\end{verbatim}

\dnorm


\section{Speed issues}

As already noted, the prototype \nlitdb was developed simply to
demonstrate that the mappings from English to \topl and from \topl to
\tsql are implementable. Execution speed was not a priority, and the
\nlitdb code is by no means optimised for fast execution. On a lightly
loaded Sun \textsc{Sparc}station 5, single-clause questions with
single parses are typically mapped to \tsql queries in about 15--30
seconds. Longer questions with subordinate clauses and multiple parses
usually take 1--2 minutes to process. (These times include the
printing of all the \hpsg signs, \topl formulae, and \tsql queries.)
The system's speed seems acceptable for a research prototype, but it is
unsatisfactory for real-life applications. 

Whenever a modification is made in the software, the code of the
affected modules has to be recompiled. This takes only a few seconds
in the case of modules that are written in Prolog (the post-processor
and the \topl to \tsql translator), but it is very time-consuming in
the case of modules that are written in \ale's formalism (the
components of the \hpsg grammar and the extractor of \topl formulae).
This becomes particularly annoying when experimenting with the
grammar, as in many cases after modifying the grammar all its
components (sort hierarchy, lexical rules, etc.) have to be
recompiled, and this recompilation takes approximately 8 minutes on
the above machine.


\section{Summary}

The framework of this thesis was tested by developing a prototype
\nlitdb, implemented using Prolog and \ale.  The prototype was
configured for the hypothetical airport application. A number of
sample questions were used to demonstrate that the system behaves
according to the specifications of the previous chapters. The
architecture of the prototype is currently minimal. A preprocessor,
mechanisms for quantifier-scoping and anaphora resolution, an
equivalential translator, a response generator, and configuration
tools would have to be added if the system were to be used in
real-life applications. Execution speed would also have to be
improved.



\chapter{Comparison with Previous Work on NLITDBs} \label{comp_chapt}

\proverb{Other times other manners.}

This chapter begins with a discussion of previous work on
\nlitdb{s}. The discussion identifies six problems from which previous
proposals on \nlitdb{s} suffer. I then examine if the
framework of this thesis overcomes these problems. 


\section{Previous work on NLITDBs} \label{previous_nlitdbs}

This section discusses previous work on \nlitdb{s}. Clifford's work,
which is the most significant and directly relevant to this thesis, is
presented first.


\subsection{Clifford} \label{Clifford_prev}

Clifford \cite{Clifford} defined a temporal version of the relational
model. He also showed how a fragment of English questions involving
time can be mapped systematically to logical expressions whose
semantics are defined in terms of a database structured according to
his model.\footnote{Parts of \cite{Clifford} can be found in
  \cite{Clifford4}, \cite{Clifford5}, and \cite{Clifford3}. The
  database model of this section is that of \cite{Clifford}. A
  previous version of this model appears in \cite{Clifford2}.}
Clifford's approach is notable in that both the semantics of the
English fragment and of the temporal database are defined within
a common model-theoretic framework, based on Montague semantics
\cite{Dowty}.

Clifford extended the syntactic coverage of Montague's \ptq grammar,
to allow past, present, and future verb forms, some temporal connectives
and adverbials (e.g.\ \qit{while}, \qit{during}, \qit{in 1978},
\qit{yesterday}), and questions. \pref{cliff:2} -- \pref{cliff:7} are
all within Clifford's syntactic coverage. (Assertions like
\pref{cliff:4} are treated as yes/no questions.)
\begin{examps}
\item Is it the case that Peter earned 25K in 1978? \label{cliff:2}
\item Does Rachel manage an employee such that he earned 30K? \label{cliff:3}
\item John worked before Mary worked. \label{cliff:4}
\item Who manages which employees? \label{cliff:6}
\item When did Liz manage Peter? \label{cliff:7}
\end{examps}
Clifford does not allow progressive verb forms. He also claims that no
distinction between progressive and non-progressive forms is necessary
in the context of \nlitdb{s} (see p.12 of \cite{Clifford4}).
According to Clifford's view, \pref{cliff:1a} can be treated in
exactly the same manner as \pref{cliff:1b}.  This ignores the fact
that \pref{cliff:1b} most probably refers to a company that habitually
or normally services BA737, or to a company that will service BA737
according to some plan, not to a company that is actually servicing
BA737 at the present. In contrast, \pref{cliff:1a} most probably
refers to a company that is actually servicing BA737 at the present,
or to a company that is going to service BA737. Therefore, the \nlitdb
should not treat the two questions as identical, if its responses are
to be appropriate to the meanings users have in mind.
\begin{examps}
\item Which company is servicing flight BA737? \label{cliff:1a}
\item Which company services flight BA737? \label{cliff:1b}
\end{examps}
Clifford also does not discuss perfect tenses (present perfect, past
perfect, etc.), which do not seem to be allowed in his framework.
Finally, he employs no aspectual taxonomy (this will be discussed in
section \ref{sem_assess}).

Following the Montague tradition, Clifford employs an intensional
higher order language (called \ils) to represent the meanings of
English questions. There is a set of syntactic rules that determine the
syntactic structure of each sentence, and a set of semantic rules
that map syntactic structures to expressions of \ils. For example,
\pref{cliff:7} is mapped to the \ils expression of \pref{cliff:8}.
\begin{examps}
\item $\begin{aligned}[t]
       \lambda i_1 [[i_1 < i] \land \exists y [&EMP'_*(i_1)(Peter) \land \\
                                               &MGR'(i_1)(y) \land 
                                                y(i_1) = Liz \land
                                                AS\_1(Peter, y)]]
       \end{aligned}$
   \label{cliff:8}
\end{examps}
Roughly speaking, \pref{cliff:8} has the following meaning.
$EMP'_*(i_1)(Peter)$ means that Peter must be an employee at the
time-point $i_1$. $MGR'(i_1)(y)$ means that $y$ must be a partial
function from time-points to managers (an \emph{intension} in Montague
semantics terminology) which is defined for (at least) the time-point
$i_1$. $AS\_1(Peter, y)$ requires $y$ to represent the history of
Peter's managers (i.e.\ the value $y(i_1)$ of $y$ at each time-point
$i_1$ must be the manager of Peter at that time-point). The $y(i_1) =
Liz$ requires the manager of Peter at $i_1$ to be Liz. Finally, $i$ is
the present time-point, and $i_1 < i$ means that $i_1$ must precede
$i$. \pref{cliff:8} requires all time-points $i_1$ to be
reported, such that $i_1$ precedes the present time-point, Peter is an
employee at $i_1$, and Peter's manager at $i_1$ is Liz.

The following (from \cite{Clifford}) is a relation in 
Clifford's database model (called \hrdm\ -- Historical Relational
Database Model).

\singlespace
\begin{center}
{\small
\begin{tabular}{|l|l|l|l|l|}
\hline
\multicolumn{5}{|l|}{$emprel$} \\
\hline \hline
$EMP$ & $MGR$ & $DEPT$ & $SAL$ & $lifespan$ \\
\hline
&&&& \\
$Peter$ & $ \left[ \begin{array}{l}
                 S2 \rightarrow Elsie\\
                 S3 \rightarrow Liz
                 \end{array} \right] $ 
      & $ \left[ \begin{array}{l}
                 S2 \rightarrow Hardware\\
                 S3 \rightarrow Linen
                 \end{array} \right] $
      & $ \left[ \begin{array}{l}
                 S2 \rightarrow 30K \\
                 S3 \rightarrow 35K
                 \end{array} \right] $
      & $\{S2,S3\}$\\
&&&& \\
\hline
&&&& \\
$Liz$ & $ \left[ \begin{array}{l}
                 S2 \rightarrow Elsie\\
                 S3 \rightarrow Liz
                 \end{array} \right] $
      & $ \left[ \begin{array}{l}
                 S2 \rightarrow Toy\\
                 S3 \rightarrow Hardware
                 \end{array} \right] $
      & $ \left[ \begin{array}{l}
                 S2 \rightarrow 35K \\
                 S3 \rightarrow 50K
                 \end{array} \right] $
      & $\{S2,S3\}$\\
&&&& \\
\hline
&&&& \\
$Elsie$ & $ \left[ \begin{array}{l}
                 S1 \rightarrow Elsie\\
                 S2 \rightarrow Elsie
                 \end{array} \right] $ 
      & $ \left[ \begin{array}{l}
                 S1 \rightarrow Toy\\
                 S2 \rightarrow Toy
                 \end{array} \right] $
      & $ \left[ \begin{array}{l}
                 S1 \rightarrow 50K \\
                 S2 \rightarrow 50K
                 \end{array} \right] $
      & $\{S1,S2\}$\\
&&&& \\
\hline
\end{tabular}
} 
\end{center}

\doublespace The $\mathit{lifespan}$ of each tuple shows the
time-points (``states'' in Clifford's terminology) for which the tuple
carries information. In \hrdm, attribute values are not necessarily
atomic. They can also be sets of time-point denoting symbols (as in
the case of $\mathit{lifespan}$), or partial functions from time-point
denoting symbols to atomic values. The relation above means that at
the time-point $S2$ the manager of Peter was Elsie, and that at $S3$
the manager of Peter was Liz. \hrdm uses additional time-stamps to
cope with schema-evolution (section \ref{no_issues}). I do not discuss
these here.

Clifford shows how the semantics of \ils expressions can be defined in
terms of an \hrdm database (e.g.\ how the semantics of \pref{cliff:8}
can be defined in terms of information in $emprel$). He also defines
an algebra for \hrdm, similar to the relational algebra of the
traditional relational model \cite{Ullman}. (Relational algebra is a
theoretical database query language. Most \dbms{s} do not support it
directly.  \dbms users typically specify their requests in more
user-friendly languages, like \sqll. \dbms{s}, however, often use
relational algebra internally, to represent operations that need to be
carried out to satisfy the users' requests.)  The answer to
\pref{cliff:7} can be found using \pref{cliff:9}, which is an
expression in Clifford's algebra.
\begin{examps}
\item $\mathit{\omega(\sigma\text{-}WHEN_{EMP = Peter, MGR = Liz}(emprel))}$
   \label{cliff:9}
\end{examps}
$\sigma\text{-}WHEN_{EMP = Peter, MGR = Liz}(emprel)$
\index{swhen@$\sigma\text{-}WHEN$ (operator of Clifford's algebra)}
generates a single-tuple relation (shown below as
$emprel2$) that carries the information of Peter's tuple from
$emprel$, restricted to when his manager was Liz. The $\omega$
\index{o@$\omega$ (operator of Clifford's algebra)}
operator returns a set of time-point denoting symbols, that represents
all the time-points for which there is information in the
relation-argument of $\omega$. In our example, \pref{cliff:9} 
returns $\{S3\}$.

\singlespace
\begin{center}
{\small
\begin{tabular}{|l|l|l|l|l|}
\hline
\multicolumn{5}{|l|}{$emprel2$} \\
\hline \hline
$EMP$ & $MGR$ & $DEPT$ & $SAL$ & $lifespan$ \\
\hline
&&&& \\
$Peter$ & $[S3 \rightarrow Liz] $ 
        & $[S3 \rightarrow Linen]$
        & $[S3 \rightarrow 35K] $
        & $\{S3\}$\\
&&&& \\
\hline
\end{tabular}
} 
\end{center}

\doublespace
Clifford outlines an algorithm for mapping \ils expressions to
appropriate algebraic expressions (e.g.\ mapping \pref{cliff:8} to
\pref{cliff:9}; see p.170 of \cite{Clifford}). The description of this
algorithm, however, is very sketchy and informal. 

According to Clifford
(\cite{Clifford4}, p.16), a parser for his version of the \ptq grammar
(that presumably also maps English questions to \ils
expressions) has been developed. Clifford, however, does not provide any
information on whether or not a translator from \ils to his algebra
was ever implemented (as noted above, this mapping is not
fully defined), and there is no indication that Clifford's framework
was ever used to implement an actual \nlitdb.


\subsection{Bruce} \label{Bruce_prev}

Bruce's \textsc{Chronos} \cite{Bruce1972} is probably the first
natural language question-answering system that attempted to address
specifically time-related issues. \textsc{Chronos} is not really 
an interface to a stand-alone database system. When invoked, it has no
information about the world. The user ``teaches'' \textsc{Chronos}
various facts (using statements like \pref{Bruce:1} and
\pref{Bruce:2}), which are stored internally as expressions of a
Lisp-like representation language. Questions about the stored facts
can then be asked (e.g.\ \pref{Bruce:3}, \pref{Bruce:4}).
\begin{examps}
\item The American war for independence began in 1775. \label{Bruce:1}
\item The articles of confederation period was from 1777 to 1789.
  \label{Bruce:2}
\item Does the American war for independence coincide with the time
  from 1775 to 1781? \label{Bruce:3}
\item Did the time of the American war for independence overlap the
  articles of confederation period? \label{Bruce:4}
\end{examps}
Bruce defines formally a model of time, and explores how relations
between time-segments of that model can represent the semantics of
some English temporal mechanisms (mainly verb tenses). Bruce's
time-model and temporal relations seem to underlie \textsc{Chronos}'
Lisp-like representation language.  Bruce, however, provides no
information about the representation language itself. With the
exception of verb tenses, there is very little information on the
linguistic coverage of the system and the linguistic assumptions on
which the system is based, and scarcely any information on the mapping
from English to representation language.  (The discussion in
\cite{Bruce1972} suggests that the latter mapping may be based on
simplistic pattern-matching techniques.) Finally, Bruce does not
discuss exactly how the stored facts are used to answer questions like
\pref{Bruce:3} and \pref{Bruce:4}.


\subsection{De, Pan, and Whinston}

De, Pan, and Whinston \cite{De} \cite{De2} describe a
question-answering system that can handle a fragment of English
questions involving time. The ``temporal database'' in this case is a
rather ad hoc collection of facts and inference rules (that can be
used to infer new information from the facts), rather than a
principled database built on a well-defined database model. Both the
grammar of the linguistic processor and the facts and rules of the
database are specified in ``equational logic'' (a kind of
logic-programming language). There is no clear intermediate
representation language, and it is very difficult to distinguish the
part of the system that is responsible for the linguistic processing
from the part of the system that is responsible for retrieving
information from the ``database''. De et al.\ consider this an
advantage, but it clearly sacrifices modularity and portability.  For
example, it is very hard to see which parts of the software would have
to be modified if the natural language processor were to be used with
a commercial \dbms.

The system of De et al.\ does not seem to be based on any clear
linguistic analysis. There is also very little information in
\cite{De} and \cite{De2} on exactly which temporal linguistic
mechanisms are supported, and which semantics are assigned to these
mechanisms. Furthermore, no aspectual classes are used (see related
comments in section \ref{sem_assess}).


\subsection{Moens} \label{Moens_prev}

Moens' work on temporal linguistic phenomena \cite{Moens}
\cite{Moens2} has been highly influential in the area of tense and
aspect theories (some ideas from Moens' work were mentioned in chapter
\ref{linguistic_data}). In the last part of \cite{Moens} (see also
\cite{Moens3}), Moens develops a simplistic \nlitdb. This has a very
limited linguistic coverage, and is mainly intended to illustrate
Moens' tense and aspect theory, rather than to constitute a detailed
exploration of  issues related to \nlitdb{s}.

As in the case of Bruce and De et al., Moens' ``database'' is not a
stand-alone system built according to an established (e.g.\ 
relational) database model. Instead, it is a collection of Prolog
facts of particular forms, that record information according to an
idiosyncratic and unclearly defined database model. Apart from purely
temporal information (that shows when various events took place),
Moens' database model also stores information about \emph{episodes}.
According to Moens, an episode is a sequence of ``contingently''
related events. Moens uses the term ``contingency'' in a rather vague
manner: in some cases it denotes a consequence relation (event A was a
consequence of event B); in other cases it is used to refer to events
that constitute steps towards the satisfaction of a common goal. The
intention is, for example, to be able to store an event where John
writes chapter 1 of his thesis together with an event where John
writes chapter 2 of his thesis as constituting parts of an episode
where John writes his thesis. Some episodes may be parts of larger
episodes (e.g.\ the episode where John writes his thesis could be part
of a larger episode where John earns his PhD).

Moens claims that episodic information of this kind is necessary if
certain time-related linguistic mechanisms (e.g.\ \qit{when~\dots}
clauses, present perfect) are to be handled appropriately. Although I
agree that episodic information seems to play an important role in how
people perceive temporal information, it is often difficult to see how
Moens' episodic information (especially when events in an episode are
linked with consequence relations) can be used in a practical \nlitdb
(e.g.\ in section \ref{present_perfect}, I discussed common claims
that the English present perfect involves a consequence relation, and
I explained why an analysis of the present perfect that posits a
consequence relation is impractical in \nlitdb{s}). By assuming that
the database contains episodic information, one also moves away from
current proposals in temporal databases, that do not consider
information of this kind. For these reasons, I chose not to assume
that the database provides episodic information. As was demonstrated
in the previous chapters, even in the absence of such information
reasonable responses can be generated in a large number of cases.

Moen's database model is also interesting in that it provides some
support for \emph{imprecise temporal information}. One may know, for
example, that two events A and B occurred, and that B was a
consequence of A, without knowing the precise times where A and B
occurred. Information of this kind can be stored in Moens' database,
because in his model events are not necessarily associated with times.
One can store events A and B as a sequence of contingently related
events (here contingency would have its consequence meaning) without
assigning them specific times. (If, however, there is no contingency
relation between the two events and their exact times are unknown,
Moens' model does not allow the relative order of A and B to be
stored.) Although there has been research on imprecise temporal
information in databases (e.g.\ \cite{Brusoni1995},
\cite{Koubarakis1995}), most of the work on temporal databases assumes
that events are assigned specific times. To remain compatible with
this work, I adopted the same assumption.

Moens' system uses a subset of Prolog as its meaning representation
language. English questions are translated into expressions of this
subset using a \textsc{Dcg} grammar \cite{Pereira1980}, and there are
Prolog rules that evaluate the resulting expressions against the
database. Moens provides no information about the \textsc{Dcg}
grammar. Also, the definition of the meaning representation language
is unclear. It is difficult to see exactly which Prolog expressions
are part of the representation language, and the semantics of the
language is defined in a rather informal way (by listing Prolog code
that evaluates some of the possible expressions of the representation
language against the database).


\subsection{Spenceley} \label{Spenceley_prev}

Spenceley \cite{Spenceley1989} developed a version of the
\textsc{Masque} natural language front-end \cite{Auxerre2} that can
cope with certain kinds of imperatives and temporal questions. The
front-end was used to interface to a Prolog database that modelled a
blocks-world similar to that of Winograd's
\textsc{Shrdlu} \cite{Winograd1973}. The dialogue in \pref{Spenc:1} --
\pref{Spenc:5.5} illustrates the capabilities of Spenceley's
system. The user can type imperatives, like \pref{Spenc:1} and
\pref{Spenc:2}, that cause the database to be updated
to reflect the new state of the blocks-world. At any point, questions
like \pref{Spenc:3} and \pref{Spenc:5} can be issued, to ask about
previous actions or about the current state of the world.
\begin{examps}
\item Take Cube1. \label{Spenc:1}
\item Put Cube1 on Cube2. \label{Spenc:2}
\item What was put on Cube2? \label{Spenc:3}
\item \sys{Cube1.} \label{Spenc:4}
\item Is Cube2 on Cube1? \label{Spenc:5}
\item \sys{No.} \label{Spenc:5.5}
\end{examps}
A simplistic aspectual taxonomy is adopted, that distinguishes between
\emph{states} and \emph{actions} (the latter containing Vendler's
activities, accomplishments, and achievements; see section
\ref{asp_taxes}). The linguistic coverage is severely
restricted. For example, the user can ask about past actions
and present states (e.g.\ \pref{Spenc:3}, \pref{Spenc:5}), but not
about past states (\pref{Spenc:6} is rejected). Only
\qit{while~\dots}, \qit{before~\dots}, and \qit{after~\dots}
subordinate clauses can be used to specify past times, and subordinate
clauses can refer only to actions, not states (e.g.\ \pref{Spenc:7} is
allowed, but \pref{Spenc:8} is not). Temporal adverbials, like \qit{at
  5:00pm} in \pref{Spenc:9}, are not supported. Spenceley also
attempts to provide some support for \emph{tense anaphora} (section
\ref{temporal_anaphora}), but her tense anaphora mechanism is very
rudimentary.
\begin{examps}
\item \rej Where was Cube1? \label{Spenc:6}
\item What was taken before Cube1 was put on Cube2? \label{Spenc:7}
\item \rej What was taken before Cube1 was on Cube2? \label{Spenc:8}
\item \rej What was taken at 5:00pm? \label{Spenc:9}
\end{examps}
The English requests are parsed using an ``extraposition grammar''
\cite{Pereira}, and they are translated into a subset of Prolog that
acts as a meaning representation language.\footnote{The syntax and
  semantics of a similar Prolog subset, that is used as the meaning
  representation language of another version of \textsc{Masque}, are
  defined in \cite{Androutsopoulos}.} The resulting Prolog expressions
are then executed by the Prolog interpreter to update the database or
to retrieve the requested information. The ``database'' is a
collection of ad hoc Prolog facts (and in that respect similar to the
``databases'' of Bruce, De et al., and Moens). It stores information
about past actions, but not states (this is probably why questions
like \pref{Spenc:8} are not allowed). Also, the database records
temporal relations between actions (which action followed which
action, which action happened during some other action), but not the
specific times where the actions happened. Hence, there is no
information in the database to answer questions like \pref{Spenc:9},
that require the specific times where the actions happened to be
known.


\subsection{Brown} 

Brown \cite{Brown1994} describes a question-answering system that can
handle some temporal linguistic phenomena. As in Bruce's system,
the user first ``teaches'' the system various facts (e.g.\ \qit{Pedro
  is beating Chiquita.}), and he/she can then ask questions about
these facts (e.g.\ \qit{Is he beating her?}). Brown's system is
interesting in that it is based on Discourse Representation Theory
(\drt), a theory in which tense and aspect have received
particular attention \cite{Kamp1993}. Brown's system, however, seems
to implement the tense and aspect mechanisms of \drt to a very limited
extent. Brown shows only how simple present, simple past, present
continuous, and past continuous verb forms can be handled. Other tenses,
temporal adverbials, temporal subordinate clauses, etc.\ do not seem
to be supported.

Brown's system transforms the English sentences into \drt discourse
representation structures, using a grammar written in an extended
\textsc{Dcg} version \cite{Covington1993}. Brown provides very little
information about this grammar. The relation of Brown's grammar to
that sketched in \cite{Kamp1993} is also unclear. The discourse
representation structures are then translated into Prolog facts (this
turns out to be relatively straight-forward). As in Moens' and
Spenceley's systems, the ``database'' is a collection of Prolog facts,
rather than a principled stand-alone system.


\subsection{Other related work}  \label{other_related_prev}

Hafner \cite{Hafner} considers the inability of existing \nlidb{s} to
handle questions involving time a major weakness. Observing that there
is no consensus among database researchers on how the notion of time
should be supported in databases (this was true when \cite{Hafner} was
written), Hafner concludes that \nlidb designers who wish their
systems to handle questions involving time cannot look to the
underlying \dbms for special temporal support. She therefore proposes
a temporal reasoning model (consisting of a temporal ontology, a
Prolog-like representation language, and inference rules written in
Prolog), intended to be incorporated into a hypothetical \nlidb to
compensate for the lack of temporal support from the \dbms. Hafner,
however, does not describe exactly how her reasoning model would be
embedded into a \nlidb (e.g.\ how the semantics of verb tenses,
temporal adverbials, etc.\ could be captured in her representation
language, how English questions could be translated systematically
into her representation language, and exactly how her inference rules
would interact with the \dbms). Also, although when \cite{Hafner} was
written it was true that there was no consensus among temporal
database researchers, and that the \nlidb designer could not expect
special temporal support from the \dbms, this is (at least to some
extent) not true at the present. A temporal database query language
(\tsql) that was designed by a committee comprising most leading
temporal database researchers now exists, and a prototype \dbms
(\textsc{TimeDB}; section \ref{tdbs_general}) that supports \tsql has
already appeared. Instead of including into the \nlitdb a
temporal reasoning module (as sketched by Hafner), in this thesis I
assumed that a \dbms supporting \tsql is available, and I exploited
\tsql's temporal facilities.

Mays \cite{Mays1986} defines a modal logic which can be used to reason
about possible or necessary states of the world (what may or will
become true, what was or could have been true; see also the discussion
on modal questions in section \ref{no_issues}). Mays envisages a
reasoning module based on his logic that would be used when a \nlidb
attempts to generate cooperative responses. In \pref{Mays:2}, for
example, the system has offered to monitor the database, and to inform
the user when Kitty Hawk reaches Norfolk. In order to avoid responses
like \pref{Mays:4}, the system must be able to reason that the
distance between the two cities will never change. Mays, however, does
not discuss exactly how that reasoning module would be embedded into a
\nlidb (e.g.\ how English questions would be mapped to expressions of
his logic, and how the reasoning module would interact with the
database).
\begin{examps}
\item Is the Kitty Hawk in Norfolk? \label{Mays:1}
\item \sys{No, shall I let you know when she is?} \label{Mays:2}
\item Is New York less than 50 miles from Philadelphia? \label{Mays:3}
\item \sys{No, shall I let you know when it is?} \label{Mays:4}
\end{examps}
Hinrichs \cite{Hinrichs} proposes methods to address some time-related
linguistic phenomena, reporting on experience from a natural language
understanding system that, among other things, allows the user to
access time-dependent information stored in a database. Although
Hinrichs' methods are interesting (some of them were discussed in
section \ref{noun_anaphora}), \cite{Hinrichs} provides little
information on the actual natural language understanding system, and
essentially no information on the underlying \dbms and how the
intermediate representation language expressions are evaluated against
the database. There is also no indication that any aspectual taxonomy
is used, and the system uses a version of Montague's \ptq grammar (see
related comments in section \ref{eval_Grammar} below). 

Finally, in \textsc{Cle} (a generic natural language front-end
\cite{Alshawi}) verb tenses introduce into the logical expressions
temporal operators, and variables that are intended to represent
states or events. The semantics of these operators and variables,
however, are left undefined. In \textsc{Clare} (roughly speaking, a
\nlidb based on \textsc{Cle}; see \cite{Alshawi2}) the temporal
operators are dropped, and verb tenses are expressed using
predications over event or state variables. The precise semantic
status of these variables remains obscure. Both \cite{Alshawi} and
\cite{Alshawi2} do not discuss temporal linguistic phenomena in any
detail.


\section{Assessment} \label{evaluation}

It follows from the discussion in section \ref{previous_nlitdbs} that
previous approaches to \nlitdb{s} suffer from one or more of
the following: (i) they ignore important English temporal mechanisms,
or assign to them over-simplified semantics (e.g.\ Clifford,
Spenceley, Brown), (ii) they lack clearly defined meaning representation
languages (e.g.\ Bruce, De et al., Moens), (iii) they do not provide
complete descriptions of the mappings from natural language to meaning
representation language (e.g.\ Bruce, Moens, Brown), or (iv) from meaning
representation language to database language (e.g.\ Clifford), (v)
they adopt idiosyncratic and often not well-defined database models or
languages (e.g.\ Bruce, De et al., Moens, Spenceley, Brown), (vi) they do not
demonstrate that their ideas are implementable (e.g.\ Clifford,
Hafner, Mayes). In this section I assess the work of this thesis with
respect to (i) -- (vi), comparing mainly to Clifford's work, which
constitutes the most significant previous exploration of \nlitdb{s}.


\subsection{English temporal mechanisms and their semantics} \label{sem_assess}

In section \ref{Clifford_prev}, I criticised Clifford's lack of
aspectual taxonomy. It should be clear from the discussion in chapter
\ref{linguistic_data} that the distinction between aspectual classes
pertains to the semantics of most temporal linguistic mechanisms, and
that without an aspectual taxonomy important semantic distinctions
cannot be captured (e.g.\ the fact that the simple past of a
culminating activity verb normally implies that the climax was
reached, while the simple past of a point, state, or activity verb
carries no such implication; the fact that an \qit{at~\dots} adverbial
typically has an inchoative or terminal meaning with a culminating
activity, but an interjacent meaning with a state, etc.) The aspectual
taxonomy of this thesis allowed me to capture many distinctions of
this kind, which cannot be accounted for in Clifford's framework.
Generally, this thesis examined the semantics of English temporal
mechanisms at a much more detailed level compared to Clifford's work.
Particular care was also taken to explain clearly which temporal
linguistic mechanisms this thesis attempts to support, which
simplifications were introduced in the semantics of these mechanisms,
and which phenomena remain to be considered (see table
\vref{coverage_table} for a summary). This information is difficult to
obtain in the case of Clifford's work.

In terms of syntactic coverage of time-related phenomena, the grammar
of this thesis is similar to Clifford's. Both grammars, for example,
support only three kinds of temporal subordinate clauses:
\qit{while~\dots}, \qit{before~\dots}, and
\qit{after~\dots} clauses. Clifford's grammar allows simple-future
verb forms (these are not supported by the grammar of this
thesis), but it does not allow progressive or perfect forms
(which are partially supported by the grammar of this thesis). 
The two grammars allow similar temporal adverbials (e.g.\
\qit{in~1991}, \qit{before 3/5/90}, \qit{yesterday}), though there are
adverbials that are supported by Clifford's grammar but not by the
grammar of this thesis (e.g.\ \qit{never}, \qit{always}), and
adverbials that are supported by the grammar of this thesis but not by
Clifford's (e.g.\ \qit{for five hours}, \qit{in two days}). Both
grammars support yes/no questions, \qit{Who/What/Which~\dots?} and
\qit{When~\dots?} questions, multiple interrogatives
(e.g.\ \qit{Who inspected what on 1/1/91?}), and assertions (which are
treated as yes/no questions). The reader is reminded, however, that
Clifford assigns to temporal linguistic mechanisms semantics which
are typically much shallower than the semantics of this thesis.

Although the framework of this thesis can cope with an interesting set
of temporal linguistic phenomena, there are still many English temporal
mechanisms that are not covered (e.g.\ \qit{since~\dots} adverbials,
\qit{when~\dots} clauses, tense anaphora). Hence,
the criticism about previous approaches, that important
temporal linguistic mechanisms are not supported, applies to the
work of this thesis as well. (It also applies to Clifford's framework, where
most of these mechanisms are also not covered.) I claim, however, that
the temporal mechanisms that are currently supported are assigned
sufficiently elaborate semantics, to the extent that the other criticism
about previous approaches, that they use over-simplified semantics, does
not apply to the work of this thesis. I hope that further work on the
framework of this thesis will extend its coverage of temporal
phenomena (see section \ref{to_do} below).


\subsection{Intermediate representation language}

From the discussion in section \ref{previous_nlitdbs}, it follows that
some previous proposals on \nlitdb{s} (e.g.\ Bruce, De et al., Moens) use
meaning representation languages that are not clearly defined.
(Clifford's work does not suffer from this problem; his \ils language
is defined rigorously.) This is a severe problem. Without a detailed
description of the syntax of the representation language, it is very
difficult to design a mapping from the representation language to a
new database language (one may want to use the linguistic front-end
with a new \dbms that supports another database language), and to
check that existing mappings to database languages cover all the
possible expressions of the representation language. Also, without a
rigorously defined semantics of the representation language, it is
difficult to see the exact semantics that the linguistic front-end
assigns to natural language expressions, and it is impossible to prove
formally that the mapping from representation language to database
language preserves the semantics of the representation language
expressions. This pitfall was avoided in this thesis: both
the syntax and the semantics of \topl are completely and formally
defined (chapter \ref{TOP_chapter}).


\subsection{Mapping from English to representation language} 
\label{eval_Grammar}

In section \ref{previous_nlitdbs}, I noted that some previous \nlitdb
proposals (e.g.\ Bruce, Moens) provide very little or no information
on the mapping from English to meaning representation language.
(Again, this criticism does not apply to Clifford's work; his mapping
from English to \ils is well-documented.) In this thesis, this pitfall
was avoided: I adopted \hpsg, a well-documented and currently
widely-used grammar theory, and I explained in detail (in chapter
\ref{English_to_TOP}) all the modifications that were introduced to
\hpsg, and how \hpsg is used to map from English to \topl.  I consider
the fact that this thesis adopts \hpsg to be an improvement over
Clifford's framework, which is based on Montague's ageing \ptq
grammar, and certainly a major improvement over other previous \nlitdb
proposals (e.g.\ Bruce, Spenceley, De et al., Moens) that employ ad
hoc grammars which are not built on any principled grammar theory.


\subsection{Mapping from representation language to database language}

As mentioned in section \ref{Clifford_prev}, Clifford outlines an
algorithm for translating from \ils (his intermediate representation
language) to a version of relational algebra. This algorithm, however,
is described in a very sketchy manner, and there is no proof that the
algorithm is correct (i.e.\ that the generated algebraic expressions
preserve the semantics of the \ils expressions). In contrast, the
\topl to \tsql mapping of this thesis is defined rigorously, and I
have proven formally that it generates appropriate \tsql queries
(chapter \ref{tdb_chapter} and appendix \ref{trans_proofs}).


\subsection{Temporal database model and language}

Several previous proposals on \nlitdb{s} (e.g.\ De et al., Spenceley,
Moens) adopt temporal database models and languages that are
idiosyncratic (not based on established database models and languages)
and often not well-defined. Although Clifford's database model and
algebra are well-defined temporal versions of the traditional
relational database model and algebra, they constitute just one of
numerous similar proposals in temporal databases, and it is unlikely
that \dbms{s} supporting Clifford's model and algebra will ever
appear. This thesis adopted \tsql and its underlying \bcdm model. As
already noted, \tsql was designed by a committee comprising most
leading temporal database researchers, and hence it has much better
chances of being supported by forthcoming temporal \dbms{s}, or at
least of influencing the models and languages that these \dbms{s} will
support. As mentioned in section \ref{tdbs_general}, a prototype \dbms
for a version of \tsql has already appeared. Although I had to
introduce some modifications to \tsql and \bcdm (and hence the
database language and model of this thesis diverge from the
committee's proposal), these modifications are relatively few and
well-documented (chapter \ref{tdb_chapter}).


\subsection{Implementation}

As mentioned in section \ref{Clifford_prev}, although a parser for
Clifford's \ptq version has been implemented, there is no indication
that a translator from \ils to his relational algebra was ever
constructed, or that his framework was ever used to build an actual
\nlitdb. (Similar comments apply to the work of Hafner and Mays of
section \ref{other_related_prev}.) In contrast, the framework of this
thesis was used to implement a prototype \nlitdb. Although several
modules need to be added to the prototype \nlitdb (section
\ref{modules_to_add}), the existence of this prototype constitutes an
improvement over Clifford's work. Unfortunately, the \nlitdb of this
thesis still suffers from the fact that it has never been linked to a
\dbms (section \ref{prototype_arch}). I hope that this will be
achieved in future (see section \ref{to_do} below).


\section{Summary}

In terms of syntactic coverage of temporal linguistic mechanisms, the
framework of this thesis is similar to Clifford's. The semantics that
Clifford assigns to these mechanisms, however, are much shallower than
those of this thesis. In both frameworks, there are several
time-related phenomena that remain to be covered. Unlike some of the
previous \nlitdb proposals, the intermediate representation language
of this thesis (\topl) is defined rigorously, and the mapping from
English to \topl is fully documented. Unlike Clifford's and other
previous proposals, this thesis adopts a temporal database model and
language (\tsql) that were designed by a committee comprising most
leading temporal database researchers, and that are more likely to be
supported by (or at least influencing) forthcoming temporal \dbms{s}.
The mapping from \topl to \tsql is fully defined and formally proven.
In contrast, Clifford's corresponding mapping is specified in a
sketchy way, with no proof of its correctness. Also, unlike Clifford's
and other previous proposals, the framework of this thesis was used to
implement a prototype \nlitdb.  The implementation of this thesis
still suffers from the fact that the prototype \nlitdb has not been
linked to a \dbms. I hope, however, that this will be achieved in
future.


\chapter{Conclusions} \label{conclusions_chapt}

\proverb{Times change and we with time.}


\section{Summary of this thesis}

This thesis has proposed a principled framework for constructing
natural language interfaces to temporal databases (\nlitdb{s}). This
framework consists of:
\begin{itemize}
\item a formal meaning representation language (\topl), used to represent the
  semantics of English questions involving time,
\item an \hpsg version that maps a wide range of English temporal
  questions to appropriate \topl expressions,
\item a set of translation rules that turn \topl expressions into
  suitable \tsql queries. 
\end{itemize}
The framework of this thesis is principled, in the sense that it is
clearly defined and based on current ideas from tense and aspect
theories, grammar theories, temporal logics, and temporal databases.
To demonstrate that it is also workable, it was employed to construct a
prototype \nlitdb, implemented using \ale and Prolog.

Although several issues remain to be addressed (these are discussed in
section \ref{to_do} below), the work of this thesis constitutes an
improvement over previous work on \nlitdb{s}, in that: (i) the
semantics of English temporal mechanisms are generally examined at a
more detailed level, (ii) the meaning representation language is
completely and formally defined, (iii) the mapping from English to
meaning representation language is well-documented and based on a
widely-used grammar theory, (iv) a temporal database language and
model that were designed by a committee comprising most leading
temporal database researchers are adopted, (v) the mapping from
meaning representation language to database language is clearly
defined and formally proven, (vi) it was demonstrated that the
theoretical framework of this thesis is implementable, by constructing
a prototype \nlitdb on which more elaborate systems can be based.


\section{Further work} \label{to_do} 

There are several ways in which the work of this thesis could be
extended:

\paragraph{Extending the linguistic coverage:} 
\label{wizard}
In section \ref{evaluation}, I noted that although the framework of
this thesis can handle an interesting set of temporal linguistic
mechanisms, there are still many time-related linguistic phenomena
that are not supported (see table \vref{coverage_table}). One could
explore how some of these phenomena could be handled. The temporal
anaphoric phenomena of section \ref{temporal_anaphora} are among those
that seem most interesting to investigate: several researchers have
examined temporal anaphoric phenomena, e.g.\ \cite{Partee1984},
\cite{Hinrichs1986}, \cite{Webber1988}, \cite{Eberle1989}, and it
would be interesting to explore the applicability of their proposals
to \nlitdb{s}. A Wizard of Oz experiment could also be carried out to
determine which temporal phenomena most urgently need to be added to
the linguistic coverage, and to collect sample questions that could be
used as a test suite for \nlitdb{s} \cite{King1996}. (In a Wizard of
Oz experiment, users interact through terminals with a person that
pretends to be a natural language front-end; see \cite{Diaper1986}.)

\paragraph{Cooperative responses:} In section \ref{no_issues}, I noted
that the framework of this thesis provides no mechanism for
cooperative responses. It became evident during the work of this
thesis that such a mechanism is particularly important in \nlitdb{s}
and should be added (cases where cooperative responses are needed were
encountered in sections \ref{simple_past}, \ref{progressives},
\ref{special_verbs}, \ref{period_adverbials}, \ref{while_clauses},
\ref{before_after_clauses}, \ref{at_before_after_op}, and
\ref{samples}). To use an example from section \ref{simple_past},
\pref{coop:2} is assigned a \topl formula that requires BA737 to have
\emph{reached} gate 2 for the answer to be affirmative. This causes a
negative response to be generated if BA737 was taxiing to gate 2 but
never reached it. While a simple negative response is strictly
speaking correct, it is hardly satisfactory in this case. A more
cooperative response like \pref{coop:4} is needed.
\begin{examps}
\item Did BA737 taxi to gate 2? \label{coop:2}
\item \sys{BA737 was taxiing to gate 2 but never reached it.} \label{coop:4}
\end{examps}
In other cases, the use of certain English expressions reveals a
misunderstanding of how situations are modelled in the database and
the \nlitdb. In \pref{coop:3}, for example, the \qit{for~\dots}
adverbial shows that the user considers departures to have durations
(perhaps because he/she considers the boarding part of the departure;
see section \ref{point_criterion}). In the airport application,
however, departures are treated as instantaneous (they include only
the time-points where the flights leave the gates), and \qit{to taxi}
is classified as a point verb. The \qit{for~\dots} adverbial combines
with a point expression, which is not allowed in the framework of this
thesis (see table \vref{for_adverbials_table}). This causes
\pref{coop:3} to be rejected without any explanation to the user. It
would be better if a message like \pref{coop:3a} could be generated.
\begin{examps}
\item Which flight was departing for twenty minutes? \label{coop:3}
\item \sys{Departures of flights are modelled as instantaneous.}
  \label{coop:3a} 
\end{examps}

\paragraph{Paraphrases:} As explained in section \ref{prototype_arch},
a mechanism is needed to generate English paraphrases of possible
readings in cases where the \nlitdb understands a question to be ambiguous.

\paragraph{Optimising the TSQL2 queries:} As discussed in section
\ref{tsql2_opt}, there are ways in which the generated \tsql queries
could be optimised before submitting them to the \dbms. One could 
examine exactly how these optimisations would be carried out.

\paragraph{Additional modules in the prototype NLITDB:} 
Section \ref{modules_to_add} identified several modules that would
have to be added to the prototype \nlitdb if this were to be used in
real-life applications: a preprocessor, modules to handle quantifier
scoping and anaphora resolution, an equivalential translator, and a
response generator. Adding a preprocessor and a simplistic response
generator (as described at the beginning of section
\ref{response_generator}) should be easy, though developing a response
generator that would produce cooperative responses is more complicated
(see the discussion above and section \ref{response_generator}). It
should also be possible to add an equivalential translator without introducing
major revisions in the work of this thesis.  In contrast, adding
modules to handle quantifier scoping and anaphora requires extending
first the theoretical framework of this thesis: one has to modify
\topl to represent universal quantification, unresolved quantifiers,
and unresolved anaphoric expressions (sections \ref{quantif_scoping}
and \ref{anaphora_module}), and to decide how to determine the scopes
or referents of unresolved quantifiers and anaphoric expressions.

\paragraph{Linking to a DBMS:} As explained in sections
\ref{tdbs_general} and \ref{contribution}, a prototype \dbms
(\textsc{TimeDb}) that supports a version of \tsql was released
recently, but the prototype \nlitdb of this thesis has not been linked
to that system (or any other \dbms). Obviously, it would be
particularly interesting to connect the \nlitdb of this thesis to
\textsc{TimeDb}. This requires bridging the differences between the
versions of \tsql that the two systems adopt (section \ref{contribution}).

\paragraph{Embedding ideas from this thesis into existing NLIDBs:}
Finally, one could explore if ideas from this thesis can be used in
existing natural language front-ends.  In section
\ref{other_related_prev}, for example, I noted that \textsc{Cle}'s
formulae contain temporal operators whose semantics are undefined. One
could examine if \topl operators (whose semantics are formally
defined) could be used instead. Ideas from the \topl to \tsql mapping
of chapter \ref{tdb_chapter} could then be employed to translate the
resulting \textsc{Cle} formulae into \tsql.



    \newpage                      
    \singlespace                             
    \addcontentsline{toc}{chapter}{Bibliography}
    \bibliography{biblio}                    

    \newpage                                
    \addcontentsline{toc}{chapter}{Appendix}
    \appendix   

\chapter{Translation rules and proofs for chapter \ref{tdb_chapter}}
\label{trans_proofs}

\section{Introduction}

This appendix contains the full set of \topl to \tsql translation
rules, and the proofs of theorems \ref{wh_theorem} and
\ref{yn_theorem} (see chapter \ref{tdb_chapter}). As noted in section
\ref{formulation}, the translation rules specify the values of
$trans(\phi, \lambda)$ for every \topl formula $\phi$ and \tsql value
expression $\lambda$. (In practice, $\lambda$ always represents a
period.) There are two kinds of translation rules: (a) base
(non-recursive) rules that define the values of $trans(\phi, \lambda)$
when $\phi$ is an atomic formula or a formula of the form
$\culm[\pi(\tau_1, \dots, \tau_n)]$; and (b) recursive rules that
define the values of $trans(\phi, \lambda)$ in all other cases (where
$\phi$ is non-atomic), by recursively calling other translation rules
to translate subformulae of $\phi$.

Section \ref{yn_rules} lists the translation rules for yes/no formulae
$\phi$. According to section \ref{formulation}, these rules have to
satisfy theorem \ref{yn_theorem}. Theorem \ref{yn_theorem} is proven
by induction on the syntactic complexity of $\phi$. I first prove that
theorem \ref{yn_theorem} holds if $\phi$ is a predicate $\pi(\tau_1,
\dots, \tau_n)$ or a formula of the form $\culm[\pi(\tau_1, \dots,
\tau_n)]$. For all other yes/no formulae $\phi$, $\phi$ is non-atomic.
In those cases, I prove that theorem \ref{yn_theorem} holds if it
holds for all the subformulae of $\phi$. Each translation rule of
section \ref{yn_rules} is followed by the corresponding part of
theorem \ref{yn_theorem}'s proof. For example, the translation rule
that specifies the values of $trans(\phi, \lambda)$ when $\phi =
\pres[\phi']$ is followed by the proof that theorem \ref{yn_theorem}
holds for $\phi = \pres[\phi']$ if it holds for $\phi = \phi'$.

Section \ref{wh_rules} lists the translation rules for wh-formulae.
These rules have to satisfy theorem \ref{wh_theorem} (see section
\ref{formulation}). There are two translation rules for wh-formulae,
that correspond to the case where $\phi \in \whforms_1$ or $\phi \in
\whforms_2$ (see section \ref{top_syntax}). Each rule is followed by a
proof that theorem \ref{wh_theorem} holds if $\phi \in \whforms_1$ or
$\phi \in \whforms_2$ respectively.

In the rest of this appendix, I use the term \emph{\tsql expression}
to refer to any piece of \tsql code. In 
contrast, the term \emph{\tsql value expression} is used to refer to a
piece of \tsql code that normally evaluates to an element of $D$ (section
\ref{additional_tsql2}). 

\section{Lemmata}

The following lemmata will prove useful in the following sections.

\begin{lemma}
\label{fcn_lemma}
{\rm If $\xi$ is a \tsql expression, and $\xi_1, \xi_2, \xi_3, \dots,
\xi_k$ are substrings of $\xi$, and any free column reference
in $\xi$ is situated within $\xi_1$ or $\xi_2$ or $\xi_3$ or \dots or
$\xi_k$, then:
\[
\fcn(\xi) \subseteq \fcn(\xi_1) \union \fcn(\xi_2) \union \fcn(\xi_3)
\union \dots \union \fcn(\xi_k)
\]
} 
\end{lemma}

\textbf{Proof:} Let us assume that $\alpha \in \fcn(\xi)$, i.e.\
$\alpha$ is a correlation name that has a free column reference
$\zeta$ in $\xi$. We need to show that $\alpha \in \fcn(\xi_1) \union
\dots \union \fcn(\xi_k)$. 

Since $\zeta$ is a free column reference in $\xi$, according to the
hypothesis for some $i \in \{1,2,3,\dots,k\}$, $\zeta$ is situated
within $\xi_i$. There is no binding context for $\zeta$ in $\xi_i$,
because since $\xi_i$ is part of $\xi$, if there is a binding context
for $\zeta$ in $\xi_i$, then there is also a (the same) binding
context for $\zeta$ in $\xi$; this would imply that $\zeta$ is not a
free column reference in $\xi$, which is against the hypothesis.

Since there is no binding context for $\zeta$ in $\xi_i$, by
definition $\zeta$ is a free column reference in $\xi_i$. This means
that $\alpha$ has a free column reference in $\xi_i$, i.e.\ $\alpha
\in \fcn(\xi_i)$. Then, $\alpha \in \fcn(\xi_1) \union \dots \union
\fcn(\xi_k)$. \qed

\begin{lemma}
\label{l2}
{\rm If $st \in \pts$, $g \in G$, and for every $i \in
  \{1,2,3,\dots,n\}$, $\tau_i \in \terms$, $v_i \in D$, and
  \pref{l2:a} -- \pref{l2:b} hold, then \pref{l2:c} also holds.
\begin{gather}
\text{If } \tau_i \in \vars, 
   \text{ then } g(\tau_i) = f_D(v_i) \label{l2:a} \\
\text{If } \tau_i \in \cons, 
   \text{ then } v_i = \hcons(st)(\tau_i) \label{l2:b} \\
\denot{M(st),g}{\tau_i} = f_D(v_i),\label{l2:c}
\end{gather}
} 
\end{lemma}

\textbf{Proof:} Since $\terms = \cons \union
\vars$, $\tau_i \in \cons$ or $\tau_i \in \vars$:
\begin{itemize}
\item $\tau_i \in \cons$: By the definition of \fcons in
  section \ref{resulting_model}, \pref{r1.37} holds. \pref{l2:b} and
  \pref{r1.37} imply \pref{r1.38}.
  \begin{gather}
    \fcons(st)(\tau_i) = f_D(\hcons(st)(\tau_i)) \label{r1.37} \\
    \fcons(st)(\tau_i) = f_D(v_i) \label{r1.38}
  \end{gather}
  The semantics of \topl imply
  that $\fcons(st)(\tau_i) = \denot{M(st),g}{\tau_i}$. Hence,
  $\denot{M(st), g}{\tau_i} = f_D(v_i)$. 

\item $\tau_i \in \vars$: By the semantics of
  \topl and \pref{l2:a}, $\denot{M(st),g}{\tau_i} = g(\tau_i) = f_D(v_i)$. 
\end{itemize}
Hence, in both cases $\denot{M(st), g}{\tau_i} = f_D(v_i)$. \qed

\begin{lemma}
\label{l3}
{\rm If $st \in \pts$, $g \in G$, $g' \in G$, $\phi \in \ynforms$,
  $\tup{\tau_1, \tau_2, \tau_3, \dots, \tau_n} = \corn{\phi}$, $v_1,
  v_2, v_3, \dots, v_n \in D$, \pref{l3:a} holds, and for every
  variable $\beta$ of $\phi$, $g(\beta) = g'(\beta)$, then \pref{l3:c}
  also holds.
\begin{gather}
\denot{M(st),g'}{\tau_1} = f_D(v_1), \; \dots \; , 
   \denot{M(st),g'}{\tau_n} = f_D(v_n) \label{l3:a} \\
\denot{M(st),g}{\tau_1} = f_D(v_1), \; \dots \; , 
   \denot{M(st),g}{\tau_n} = f_D(v_n) \label{l3:c}
\end{gather}
} 
\end{lemma}

\textbf{Proof:} The definition of $\corn{\phi}$ implies that for every
$i \in \{1,2,3,\dots,n\}$, $\tau_i \in \terms$. Hence, $\tau_i \in
\cons$ or $\tau_i \in \vars$: 
\begin{itemize}
\item $\tau_i \in \cons$: The semantics of \topl implies that
  $\denot{M(st),g}{\tau_i} = \fcons(st)(\tau_i) =
  \denot{M(st),g'}{\tau_i}$.

\item $\tau_i \in \vars$: By the definition of $\corn{\phi}$, $\tau_i$
  is a variable in $\phi$. Then, $g(\tau_i) = g'(\tau_i)$, because $g$
  and $g'$ assign the same values to all the variables of $\phi$.
  $\tau_i$ . The semantics of \topl imply that
  $\denot{M(st),g}{\tau_i} = g(\tau_i)$ and $\denot{M(st),g'}{\tau_i}
  = g'(\tau_i)$. Then, since $g(\tau_i) = g'(\tau_i)$,
  $\denot{M(st),g}{\tau_i} = \denot{M(st),g'}{\tau_i}$.
\end{itemize}
Hence, for every $i \in \{1,2,3,\dots,n\}$, $\denot{M(st),g}{\tau_i} =
\denot{M(st),g'}{\tau_i}$. This conclusion and \pref{l3:a} imply
\pref{l3:c}. \qed

\begin{lemma}
\label{l7}
{\rm If $g,g' \in G$, $\phi \in \ynforms$, $\corn{\phi} = \tup{\tau_1,
    \dots, \tau_n}$, and \pref{l7:a} holds, then for every variable
    $\beta$ of $\phi$, $g(\beta) = g'(\beta)$.
\begin{equation}
\denot{M(st),g}{\tau_1} = \denot{M(st),g'}{\tau_1}, \; \dots \;
\denot{M(st),g}{\tau_n} = \denot{M(st),g'}{\tau_n} 
\label{l7:a}
\end{equation}
} 
\end{lemma}

\textbf{Proof:}

From the definition of $\corn{\phi}$, for any variable $\beta$ of
$\phi$, there is an $i \in \{1,2,3,\dots,n\}$, such that $\tau_i =
\beta$. According to the semantics of \topl:
\begin{eqnarray}
\denot{M(st),g}{\tau_i} &=& g(\tau_i) \label{r13.407} \\
\denot{M(st),g'}{\tau_i} &=& g'(\tau_i) \label{r13.408} 
\end{eqnarray}
\pref{l7:a}, \pref{r13.407}, and \pref{r13.408} imply
that $g(\tau_i) = g'(\tau_i)$, i.e.\ $g(\beta) = g'(\beta)$. In
other words, for every variable $\beta$ of $\phi'$,
$g$ and $g'$ assign the same value to $\beta$. \qed

\begin{lemma}
\label{l4}
{\rm If $\tau_1, \tau_2, \tau_3, \dots, \tau_n \in \terms$, $v_1, v_2,
  v_3, \dots, v_n \in D$, \pref{l4:a} holds, and the mapping $g: \vars
  \mapsto \objs$ is as in \pref{l4:b} ($o$\/ is a particular element
  of \objs, chosen arbitrarily), then $g \in G$.
\begin{gather}
\mbox{if } i,j \in \{1,2,3,\dots,n\}, \; i \not= j, \; \tau_i = \tau_j, 
   \label{l4:a} \\
\;\;\;\; \text{and } \tau_i,\tau_j \in \vars, \mbox{ then } v_i = v_j 
   \nonumber \\
g(\beta) \defeq
   \left\{
   \begin{array}{l}
   f_D(v_i), \mbox{ if for some } i \in \{1,2,3,\dots,n\},
      \; \beta = \tau_i  \\
   o, \mbox{ otherwise}
   \end{array}
   \right. \label{l4:b}
\end{gather}
} 
\end{lemma}

\textbf{Proof:} $g$ is a function. To show this, I need to prove that
for each $\beta \in \vars$, $g(\beta)$ is uniquely defined. There is
only one case where $g(\beta)$ may not be uniquely defined: there may
be two different $i_1,i_2 \in \{1,2,3,\dots,n\}$, with $\beta =
\tau_{i_1} = \tau_{i_2}$. In this case I need to show that
$f_D(v_{i_1}) = f_D(v_{i_2})$. I will show that $v_{i_1} = v_{i_2}$,
which implies $f_D(v_{i_1}) = f_D(v_{i_2})$. The proof follows:

If $i_1,i_2 \in \{1,2,3,\dots,n\}$, with $\beta = \tau_{i_1} =
\tau_{i_2} \in \vars$, and $i_1 \not= i_2$, then let $i$ be the
smaller of $i_1$ and $i_2$, and $j$ the greater of $i_1$ and $i_2$.
By \pref{l4:a}, $v_i = v_j$.  This implies that $v_{i_1} =
v_{i_2}$. Hence, $g(\beta)$ is uniquely defined, and $g$ is a
function. Since $g$ also maps from \vars to \objs, $g \in G$. \qed

\begin{lemma}
\label{l5}
{\rm If $\tau^1_1, \tau^1_2, \dots, \tau^1_{n_1}, \tau^2_1, \dots,
  \tau^2_{n_2} \in \terms$, $v^1_1, v^1_2, \dots, v^1_{n_1}, v^2_1,
  \dots, v^2_{n_2} \in D$, $st \in \pts$, $g_1, g_2 \in G$,
  \pref{l5:a} -- \pref{l5:c} hold, and the mapping $g: \vars \mapsto
  \objs$ is as in \pref{l5:d} ($o$ is a particular element of \objs,
  chosen arbitrarily), then $g \in G$.
\begin{gather}
\mbox{if } i \in \{1,2,3,\dots,n_1\}, \; j \in
   \{1,2,3,\dots,n_2\}, \; \tau^1_i, \tau^2_j \in \vars, \label{l5:a} \\
\;\;\;\; \mbox{and } \tau^1_i = \tau^2_j, 
         \mbox{ then } v^1_i = v^2_j \nonumber \\
\denot{M(st),g_1}{\tau^1_1} = f_D(v^1_1), \dots, 
   \denot{M(st),g_1}{\tau^1_{n_1}} = f_D(v^1_{n_1}) \label{l5:b} \\
\denot{M(st),g_2}{\tau^2_1} = f_D(v^2_1), \dots, 
   \denot{M(st),g_2}{\tau^2_{n_2}} = f_D(v^2_{n_2}) \label{l5:c} \\
g(\beta) \defeq
   \left\{
   \begin{array}{l}
   g_1(\beta), \mbox{ if for some } i \in \{1,2,3,\dots,n_1\},
      \; \beta = \tau^1_i  \\
   g_2(\beta), \mbox{ if for some } j \in \{1,2,3,\dots,n_2\},
      \; \beta = \tau^2_j  \\
   o, \mbox{ otherwise}
   \end{array}
   \right. \label{l5:d}  
\end{gather}
} 
\end{lemma}

\textbf{Proof:} $g$ is a function. To show this, I need to prove that
for each $\beta \in \vars$, $g(\beta)$ is uniquely defined. There is
only one case where $g(\beta)$ may not be uniquely defined: there may
be both an $i \in \{1,2,3,\dots,n_1\}$ and a $j \in
\{1,2,3,\dots,n_2\}$, with $\beta = \tau^1_i = \tau^2_j \in \vars$. In
this case, I need to prove that $g_1(\beta) = g_2(\beta)$, i.e.\ that
$g_1(\tau^1_i) = g_2(\tau^2_j)$.  ($g_1(\tau^1_i)$ and $g_2(\tau^2_j)$
are uniquely defined, because $g_1,g_2 \in G$, which implies that
$g_1,g_2$ are functions.)

Let us assume that $i \in \{1,2,3,\dots,n_1\}$, $j \in
\{1,2,3,\dots,n_2\}$, and that $\beta = \tau^1_i = \tau^2_j \in
\vars$. Since $\tau^1_i, \tau^2_j \in \vars$, the semantics
of \topl implies that:
\begin{gather}
g_1(\tau^1_i) = \denot{M(st),g_1}{\tau^1_i} \label{r9.32} \\
g_2(\tau^2_j) = \denot{M(st),g_2}{\tau^2_j} \label{r9.33}
\end{gather}
\pref{r9.32} and \pref{l5:b} imply \pref{r9.34}, while \pref{r9.33}
and \pref{l5:c} imply \pref{r9.35}. 
\begin{gather}
  g_1(\tau^1_i) = f_D(v^1_i)  \label{r9.34} \\
  g_2(\tau^2_j) = f_D(v^2_j)  \label{r9.35}
\end{gather}
Since $i \in \{1,2,3,\dots,n_1\}$, $j \in \{1,2,3,\dots,n_2\}$,
$\tau^1_i,\tau^2_j \in \vars$, and $\tau^1_i = \tau^2_j$, \pref{l5:a}
implies that $v^1_i = v^2_j$.  This, along with \pref{r9.34} and
\pref{r9.35} imply that $g_1(\tau^1_i) = g_2(\tau^2_j)$. Hence,
$g(\beta)$ is uniquely defined, and $g$ is a function. Since $g$ also
maps from $\vars$ to $\objs$, $g \in G$. \qed

\begin{lemma}
\label{l6}
{\rm If $v \in D_P$, $\beta \in \vars$, $g' \in G$, and $g =
  (g')^{\beta}_{f_D(v)}$, then $g \in G$.
} 
\end{lemma}

\textbf{Proof:} To show that $g \in G$, it is enough to show that
$f_D(v) \in \objs$. $f_D(v) \in \periods$, because $v \in D_P$. Since
$\periods \subseteq \objs$ (sections \ref{top_model} and
\ref{resulting_model}), $f_D(v) \in \objs$.  Hence, $g \in G$. \qed


\section{Translation rules for yes/no formulae and proof of theorem
\ref{yn_theorem}} \label{yn_rules}


\subsection{$\pi(\tau_1, \dots, \tau_n)$} \label{pred_trans_section}

\subsubsection*{Translation rule}

If $\pi \in \pfuns$, $\tau_1, \dots, \tau_n \in \terms$, and
$\lambda$ is a \tsql value expression, then:

$trans(\pi(\tau_1, \dots, \tau_n), \lambda) \defeq$\\
\sql{(}\select{SELECT DISTINCT $\alpha.1$, $\alpha.2$, \dots, $\alpha.n$ \\
               VALID VALID($\alpha$) \\
               FROM ($\hpfunsp(\pi, n)$)(SUBPERIOD) AS $\alpha$ \\
               WHERE \dots \\
               \ \ AND \dots \\
               \ \ \vdots \\
               \ \ AND \dots \\
               \ \ AND $\lambda$ CONTAINS VALID($\alpha$))} 

Each time the translation rule is used, $\alpha$ is a new correlation
name, obtained by calling the correlation names
generator after $\lambda$ has been supplied. The ``\dots''s in
the \sql{WHERE} clause correspond to all the strings in $S_1
\union S_2$, where: 
\begin{gather*}
S_1 = 
\{\text{``}\alpha.i = \hconsp(\tau_i)\text{''} \mid
  i \in \{1,2,3,\dots,n\} \text{ and } \tau_i \in \cons\} \\
S_2 = 
\{\text{``}\alpha.i = \alpha.j\text{''} \mid
  i,j \in \{1,2,3,\dots,n\}, \; i < j, \; \tau_i = \tau_j, \text{ and }
  \tau_i, \tau_j \in \vars\}
\end{gather*}

\subsubsection*{Proof that theorem \ref{yn_theorem} holds for $\phi =
\pi(\tau_1, \dots, \tau_n)$}

I assume that $\pi \in \pfuns$ and $\tau_1, \dots, \tau_n \in
\terms$. By the syntax of \topl, this implies that $\pi(\tau_1, \dots,
\tau_n) \in \ynforms$. I also assume that $st \in
\pts$, $\lambda$ is a \tsql value expression, $g^{db} \in
G^{db}$, $eval(st, \lambda, g^{db}) \in D_P^*$, and $\Sigma =
trans(\pi(\tau_1, \dots, \tau_n), \lambda)$. By the definition of
$\corn{\dots}$, $\corn{\pi(\tau_1, \dots, \tau_n)} = \tup{\tau_1,
\dots, \tau_n}$. I need to show that the three clauses of theorem
\ref{yn_theorem} hold. 

\subsubsection*{Proof of clause 1}

The $\alpha.1$, $\alpha.2$, \dots, $\alpha.n$ in the \sql{SELECT}
clause of $\Sigma$ and the \sql{VALID($\alpha$)} in the \sql{VALID}
and \sql{WHERE} clauses are not free column references in $\Sigma$,
because $\Sigma$ is a binding context for all of them. For the same
reason, any column references of the form $\alpha.i$ and $\alpha.j$
(deriving from $S_1$ and $S_2$) in the \sql{WHERE} clause are not free
column references in $\Sigma$. The only remaining parts of $\Sigma$
where column references (and hence free column references) may occur
are the $\hpfunsp(\pi,n)$ of the \sql{FROM} clause, the $\lambda$ of
the \sql{WHERE} clause, and the $\hconsp(\tau_{i_1})$,
$\hconsp(\tau_{i_2})$, $\hconsp(\tau_{i_3})$, \dots,
$\hconsp(\tau_{i_m})$ of the \sql{WHERE} clause
($\hconsp(\tau_{i_1})$, \dots, $\hconsp(\tau_{i_m})$ derive from
$S_1$; $\tau_{i_1}, \dots, \tau_{i_m}$ are all the \topl constants
among $\tau_1, \dots, \tau_n$). By lemma \ref{fcn_lemma}, this implies
that:
\begin{align}
\label{pred:1}
\fcn(\Sigma) \subseteq 
 & \; \fcn(\hpfunsp(\pi,n)) \union \fcn(\lambda) \union \\
 & \; \fcn(\hconsp(\tau_{i_1})) \union \dots \union
      \fcn(\hconsp(\tau_{i_m})) \notag
\end{align}
According to section \ref{via_TSQL2}, for every $\kappa \in \cons$,
$\fcn(\hconsp(\kappa)) = \emptyset$. Since $\tau_{i_1}, \dots,
\tau_{i_m} \in \cons$ (see comments above), 
$\fcn(\hconsp(\tau_{i_1})) = \emptyset$, \dots,
$\fcn(\hconsp(\tau_{i_m})) = \emptyset$. According to section
\ref{via_TSQL2}, it is also true that for every $\pi \in
\pfuns$ and $n \in \{1,2,3,\dots\}$, $\fcn(\hpfunsp(\pi,n)) =
\emptyset$. Hence, \pref{pred:1} becomes $\fcn(\Sigma) \subseteq
\fcn(\lambda)$. Clause 1 has been proven. 

\subsubsection*{Proof of clause 2}

According to section \ref{via_TSQL2}, $\hpfunsp(\pi, n)$ is a \tsql
\sql{SELECT} statement, $\fcn(\hpfunsp(\pi, n)) = \emptyset$, and
$eval(st, \hpfunsp(\pi, n)) \in \cvrel(n)$. The $\alpha$ of $\Sigma$
ranges over the tuples of the relation $subperiod(eval(st,
\hpfunsp(\pi, n)))$.  From the definition of $subperiod$ (section
\ref{new_pus}), it is easy to see that $subperiod(eval(st,
\hpfunsp(\pi, n)))$ is a valid-time relation that has the same number
of explicit attributes as $eval(st, \hpfunsp(\pi, n))$, i.e.\ $n$.
Hence, $\alpha$ ranges over tuples $\tup{v_1, \dots, v_n; v_t} \in
subperiod(eval(st,\hpfunsp(\pi, n)))$.

The $\alpha$ of $\Sigma$ is generated by calling the correlation names
generator after $\lambda$ has been supplied. Hence, $\alpha$ cannot
appear in $\lambda$. Since $\alpha$ does not appear in $\lambda$, for
every tuple $\tup{v_1, \dots, v_n; v_t}$:
\begin{equation}
\label{pred:3}
eval(st, \lambda, (g^{db})^\alpha_{\tup{v_1, \dots, v_n;v_t}}) =
eval(st, \lambda, g^{db})
\end{equation}

The reader should now be able to see from the translation rule that
\pref{r1.4} holds. Intuitively, $\tup{v_1, \dots, v_n; v_t}$ is the
tuple of $subperiod(eval(st, \hpfunsp(\pi, n)))$ to which $\alpha$
refers. The last line of \pref{r1.4} corresponds to the \sql{CONTAINS}
constraint in the \sql{WHERE} clause of $\Sigma$. I should have used
$eval(st, \lambda, (g^{db})^\alpha_{\tup{v_1, \dots, v_n;v_t}})$
instead of $eval(st, \lambda, g^{db})$, to capture the fact that if
there is any free column reference of $\alpha$ in $\lambda$, this has
to be taken to refer to the $\tup{v_1, \dots, v_n; v_t}$ tuple to
which $\alpha$ refers. By \pref{pred:3}, however, $eval(st, \lambda,
(g^{db})^\alpha_{\tup{v_1, \dots, v_n;v_t}})$ is the same as $eval(st,
\lambda, g^{db})$. The second and third lines of \pref{r1.4}
correspond to the restrictions of $S_1$ and $S_2$ (see also the
comments about the equality predicate in section \ref{eq_checks}). I
do not include in the arguments of $eval(st, \hconsp(\tau_i))$ the
assignment to the correlation names, because according to section
\ref{via_TSQL2}, for $\tau_i \in \cons$, $\fcn(\hconsp(\tau_i)) =
\emptyset$.
\begin{eqnarray}
\label{r1.4}
&& eval(st, \Sigma, g^{db})  = \{ \tup{v_1,\dots,v_n;v_t} \in
  subperiod(eval(st,\hpfunsp(\pi,n))) \mid \\
&& \mbox{if } i \in \{1,2,3,\dots,n\} \mbox{ and } \tau_i \in \cons, 
   \mbox{ then } v_i = eval(st, \hconsp(\tau_i)),  
\nonumber \\
&& \mbox{if } i,j \in \{1,2,3,\dots,n\}, \; i < j, \; \tau_i = \tau_j, \mbox{
  and } \tau_i,\tau_j \in \vars, \mbox{ then } v_i = v_j, 
\nonumber \\
&& f_D(v_t) \subseteq f_D(eval(st,\lambda,g^{db})) \}
\nonumber 
\end{eqnarray}

According to the definition of $subperiod$ (section \ref{new_pus}),
\pref{r1.22pp} holds iff \pref{pred:5} holds for some $v'_t$. By 
the definition of \hpfuns of section \ref{via_TSQL2}, \pref{pred:5} is
in turn equivalent to \pref{r1.5}.
\begin{gather}
 \label{r1.22pp}
 \tup{v_1,\dots,v_n;v_t} \in subperiod(eval(st,\hpfunsp(\pi, n)))  \\
 \label{pred:5}
 \tup{v_1,\dots,v_n;v_t'} \in eval(st,\hpfunsp(\pi,n)) \mbox{ and }
 f_D(v_t) \subper f_D(v_t') \\
 \label{r1.5}
 \tup{v_1,\dots,v_n;v_t'}  \in \hpfuns(st)(\pi,n) \mbox{ and }  
 f_D(v_t) \subper f_D(v_t')
\end{gather}

Using the fact that \pref{r1.22pp} holds iff 
\pref{r1.5} holds for some $v_t'$, and the fact that for $\tau_i \in \cons$, 
$eval(st, \hconsp(\tau_i)) = \hcons(st)(\tau_i)$ (section
\ref{via_TSQL2}), \pref{r1.4} becomes:
\begin{eqnarray}
\label{r1.6}
&& eval(st,\Sigma, g^{db}) = \{ \tup{v_1,\dots,v_n;v_t} \mid \text{
   for some } v_t', \\ 
&& \tup{v_1,\dots,v_n;v_t'} \in
   \hpfuns(st)(\pi,n), \; f_D(v_t) \subper f_D(v_t')
\nonumber \\
&& \mbox{if } i \in \{1,2,3,\dots,n\} \mbox{ and } \tau_i \in \cons, 
   \mbox{ then } v_i = \hcons(st)(\tau_i),  
\nonumber \\
&& \mbox{if } i,j \in \{1,2,3,\dots,n\}, \; i < j, \; \tau_i = \tau_j, \mbox{
  and } \tau_i,\tau_j \in \vars, \mbox{ then } v_i = v_j, 
\nonumber \\
&& f_D(v_t) \subseteq f_D(eval(st,\lambda,g^{db})) \}
\nonumber 
\end{eqnarray}
For every $\tup{v_1, \dots, v_n; v_t} \in eval(st, \Sigma, g^{db})$,
the $f_D(v_t) \subper f_D(v_t')$ in the second line of \pref{r1.6}
implies that $f_D(v_t) \in \periods$, which in turn implies that $v_t
\in D_P$. That is, all the time-stamps of $eval(st, \Sigma, g^{db})$
are elements of $D_P$. \pref{r1.6} also implies that $eval(st, \Sigma,
g^{db})$ is a valid-time relation of $n$ explicit attributes. Hence,
$eval(st,\Sigma, g^{db}) \in \vrel(n)$, and clause 2 has been proven.

\subsubsection*{Proof of clause 3}

Using the definition of $\denot{M(st),st,et,lt,g}{\pi(\tau_1, \dots,
\tau_n)}$ (section \ref{denotation}), clause 3 becomes:
\begin{eqnarray}
&& \tup{v_1,\dots,v_n;v_t} \in eval(st,\Sigma,g^{db})
   \mbox{ iff for some } g \text{ and } p_{mxl} \text{:} \nonumber \\ 
&& g \in G \label{r1.44} \\
&& \denot{M(st), g}{\tau_1} = f_D(v_1), \dots, \denot{M(st),g}{\tau_n} =
   f_D(v_n) \label{r1.25} \label{r1.47} \\
&& p_{mxl} \in \fpfuns(st)(\pi, n)(\denot{M(st),g}{\tau_1},\dots,
   \denot{M(st),g}{\tau_n}) \label{r1.24} \label{r1.49} \\
&& f_D(v_t) \subper p_{mxl} \label{r1.27} \label{r1.50} \\
&& f_D(v_t) \subper f_D(eval(st, \lambda, g^{db})) \label{r1.28} \label{r1.51}
\end{eqnarray}

I first show that the forward direction of clause 3 holds. I assume 
that $\tup{v_1,\dots,v_n;v_t} \in eval(st,\Sigma,g^{db})$. Then, \pref{r1.6}
implies that for some $v_t'$:
\begin{eqnarray}
&& \tup{v_1,\dots,v_n;v_t'} \in \hpfuns(st)(\pi,n) \label{r1.7}
 \label{r1.7n} \\ 
&& f_D(v_t) \subper f_D(v_t') \label{r1.8} \label{r1.9n} \\
&& \mbox{if } i \in \{1,2,3,\dots,n\}, \mbox{ and } \tau_i \in \cons, 
   \mbox{ then } v_i = h_{cons}(st)(\tau_i) \label{r1.10} \label{r1.10n}\\
&& \mbox{if } i,j \in \{1,2,3,\dots,n\}, \; i < j, \; \tau_i = \tau_j, \mbox{
  and } \tau_i,\tau_j \in \vars, \mbox{ then } v_i = v_j  
    \label{r1.11} \label{r1.11n} \\
&& f_D(v_t) \subseteq f_D(eval(st, \lambda, g^{db})) \label{r1.9} \label{r1.8n}
\end{eqnarray}
To prove the forward direction of clause 3, I must prove 
that for some $g$ and $p_{mxl}$, \pref{r1.44} -- \pref{r1.51} hold. 
I define the mapping $g : \vars \mapsto \objs$ as follows:
\[ g(\beta) \defeq
   \left\{
   \begin{array}{l}
   f_D(v_i), \mbox{ if for some } i \in \{1,2,3,\dots,n\},
      \; \beta = \tau_i  \\
   o, \mbox{ otherwise}
   \end{array}
   \right. 
\]
where $o$\/ is a particular element of \objs, chosen arbitrarily.
\pref{r1.44} follows from lemma \ref{l4}, the definition of $g$, and
\pref{r1.11}. I set $p_{mxl}$ as in \pref{r1.31}, and show that
\pref{r1.25} -- \pref{r1.28} also hold. \pref{r1.25} follows from
lemma \ref{l2}, \pref{r1.10}, and the definition of $g$.
\begin{eqnarray}
p_{mxl} &=& f_D(v_t') \label{r1.31}
\end{eqnarray}
I now prove \pref{r1.24}. \pref{r1.25} (proven above) implies
\pref{r1.39b}. \pref{r1.7} and \pref{r1.39b} imply \pref{r1.40}.
\pref{r1.40} and the definition of \fpfuns of section
\ref{resulting_model} imply \pref{pred:6}. \pref{pred:6} and
\pref{r1.31} imply \pref{r1.24}.
\begin{eqnarray}
&&  v_1 = \fdi(\denot{M(st),g}{\tau_1}), \dots, 
    v_n = \fdi(\denot{M(st),g}{\tau_n}) \label{r1.39b} \\
&&  \tup{\fdi(\denot{M(st),g}{\tau_1}), \dots, 
       \fdi(\denot{M(st),g}{\tau_n});v_t'} \; \in \hpfuns(st)(\pi,n)
  \label{r1.40} \\
&&  f_D(v_t') \in \fpfuns(st)(\pi,n)(\denot{M(st),g}{\tau_1},\dots,
   \denot{M(st),g}{\tau_n}) \label{pred:6}
\end{eqnarray}

\pref{r1.27} follows from \pref{r1.8} and \pref{r1.31}. I now prove
\pref{r1.28}. \pref{r1.8} implies that $f_D(v_t) \in \periods$. From
the hypothesis, $eval(st, \lambda, g^{db}) \in D_P^*$, which implies
that $f_D(eval(st, \lambda, g^{db}))$ is a period or the empty set.
$f_D(eval(st, \lambda, g^{db}))$ cannot be the empty set, because
according to \pref{r1.9} $f_D(v_t)$ (which is a period and therefore a
non-empty set) is a subset of $f_D(eval(st, \lambda, g^{db}))$. Hence,
$f_D(eval(st, \lambda, g^{db})) \in \periods$. \pref{r1.9} and the
fact that both $f_D(v_t)$ and $f_D(eval(st, \lambda, g^{db}))$ are
periods imply \pref{r1.28}. The forward direction of clause 3 has
been proven.

\bigskip

I now prove the backwards direction of clause 3. I assume that
\pref{r1.44} -- \pref{r1.51} hold for some $g$ and $p_{mxl}$.  I need
to show that $\tup{v_1,\dots,v_n;v_t} \in eval(st,\Sigma,g^{db})$.
According to \pref{r1.6}, it is enough to prove that for some $v_t'$,
\pref{r1.7n} -- \pref{r1.8n} hold. I set $v_t' = \fdi(p_{mxl})$, which
implies \pref{pred:21}.
\begin{gather}
p_{mxl} = f_D(v_t') \label{pred:21}
\end{gather}

I first prove \pref{r1.7n}. \pref{r1.49}, \pref{pred:21}, and the
definition of \fpfuns of section \ref{resulting_model} imply
\pref{r1.51b}. \pref{r1.47} implies \pref{r1.52}. \pref{r1.51b} and
\pref{r1.52} imply \pref{r1.7n}.
\begin{gather}
\tup{\fdi(\denot{M(st),g}{\tau_1}), \dots, 
        \fdi(\denot{M(st),g}{\tau_n});v_t'} \in \hpfuns(st)(\pi,n)
   \label{r1.51b} \\
  \fdi(\denot{M(st),g}{\tau_1}) = v_1, \dots, 
  \fdi(\denot{M(st),g}{\tau_n}) = v_n   \label{r1.52}
\end{gather}

\pref{r1.9n} follows from \pref{r1.50} and \pref{pred:21}.
I now prove \pref{r1.10n}. If $i \in \{1,2,3,\dots,n\}$ and $\tau_i
\in \cons$, the semantics of \topl implies \pref{r1.54}, and
\pref{r1.47} implies \pref{r1.55}. 
\begin{gather}
  \denot{M(st),g}{\tau_i} = \fcons(st)(\tau_i) \label{r1.54} \\
  \denot{M(st),g}{\tau_i} = f_D(v_i) \label{r1.55}
\end{gather}
\pref{r1.54} and \pref{r1.55} imply that $f_D(v_i) =
\fcons(st)(\tau_i)$, which in turn implies \pref{r1.57}. The
definition of \fcons of section \ref{resulting_model} implies
\pref{r1.58}. \pref{r1.57} and \pref{r1.58} imply that $v_i =
\hcons(st)(\tau_i)$. \pref{r1.10n} has been proven.
\begin{gather}
  v_i = \fdi(\fcons(st)(\tau_i)) \label{r1.57} \\
  \fcons(st)(\tau_i) = f_D(\hcons(st)(\tau_i)) \label{r1.58}
\end{gather}

I now prove \pref{r1.11n}. If $i,j \in \{1,2,3,\dots,n\}$ and $\tau_i
= \tau_j$, then $\denot{M(st),g}{\tau_i} = \denot{M(st),g}{\tau_j}$.
Then, \pref{r1.47} implies that $f_D(v_i) = f_D(v_j)$, which in turn
implies that $\fdi(f_D(v_i)) = \fdi(f_D(v_j))$, i.e.\ $v_i = v_j$.
\pref{r1.11n} has been proven.  \pref{r1.8n} follows from
\pref{r1.51}. The backwards direction of clause 3 has been proven.


\subsection{$Culm[\pi(\tau_1, \dots, \tau_n)]$}

\subsubsection*{Translation rule}

If $\pi \in \pfuns$, $\tau_1, \dots, \tau_n \in \terms$, and
$\lambda$ is a \tsql value expression, then:

$trans(\culm[\pi(\tau_1, \dots, \tau_n)], \lambda) \defeq$\\
\sql{(}\select{SELECT DISTINCT 
                      $\alpha_1.1$, $\alpha_1.2$, \dots, $\alpha_1.n$ \\
               VALID PERIOD(BEGIN(VALID($\alpha_1$)),
                            END(VALID($\alpha_1$))) \\
               FROM ($\hpfunsp(\pi, n)$)(ELEMENT) AS $\alpha_1$, \\
               \ \ \ \ \ ($\hculmsp(\pi, n)$) AS $\alpha_2$ \\
               WHERE $\alpha_1.1 = \alpha_2.1$ \\
               \ \ AND $\alpha_1.2 = \alpha_2.2$ \\
               \ \ \ \ \vdots \\
               \ \ AND $\alpha_1.n = \alpha_2.n$ \\ 
               \ \ AND \dots \\
               \ \ \ \ \vdots \\
               \ \ AND \dots \\
               \ \ AND $\lambda$ CONTAINS 
                           PERIOD(BEGIN(VALID($\alpha_1$)),
                           END(VALID($\alpha_1$)))}

Each time the translation rule is used, $\alpha_1$ and $\alpha_2$ are
two new different correlation names, obtained by calling the
correlation names generator after $\lambda$ has been supplied. The
``\dots''s in the \sql{WHERE} clause correspond to all the strings in
$S_1 \union S_2$, where $S_1$ and $S_2$ are as in section
\ref{pred_trans_section}, except that $\alpha$ is now $\alpha_1$. 

\subsubsection*{Proof that theorem \ref{yn_theorem} holds for $\phi =
\culm[\pi(\tau_1, \dots, \tau_n)]$}

I assume that $\pi \in \pfuns$ and $\tau_1, \dots, \tau_n \in
\terms$. By the syntax of \topl, this implies that $\culm[\pi(\tau_1, \dots,
\tau_n)] \in \ynforms$. I also assume that $st \in
\pts$, that $\lambda$ is a \tsql value expression, $g^{db} \in
G^{db}$, $eval(st, \lambda, g^{db}) \in D_P^*$, and that $\Sigma =
trans(\culm[\pi(\tau_1, \dots, \tau_n)], \lambda)$. By the definition of
$\corn{\dots}$, $\corn{\culm[\pi(\tau_1, \dots, \tau_n)]} = \tup{\tau_1,
\dots, \tau_n}$. I need to show that the three clauses of theorem
\ref{yn_theorem} hold. 

\subsubsection*{Proof of clause 1}

The $\alpha_1.1$, $\alpha_1.2$,\dots, $\alpha_1.n$ in the \sql{SELECT}
clause of $\Sigma$, and the four \sql{VALID($\alpha_1$)} in the
\sql{VALID} and \sql{WHERE} clauses are not free column references in
$\Sigma$, because $\Sigma$ is a binding context for all of them. For
the same reason, all the column references of the form $\alpha.i$ ($i
\in \{1,2,3,\dots,n\}$) in the \sql{WHERE} clause of $\Sigma$ are not
free column references in $\Sigma$. The only remaining parts of
$\Sigma$ where column references (and hence free column references)
may occur are the $\hpfunsp(\pi,n)$ and the $\hculmsp(\pi,n)$ of the
\sql{FROM} clause, the $\lambda$ of the \sql{WHERE} clause, and the
$\hconsp(\tau_{i_1})$, $\hconsp(\tau_{i_2})$, $\hconsp(\tau_{i_3})$,
\dots, $\hconsp(\tau_{i_m})$ of the \sql{WHERE} clause
($\hconsp(\tau_{i_1})$, \dots, $\hconsp(\tau_{i_m})$ derive from
$S_1$; $\tau_{i_1}, \dots, \tau_{i_m}$ are all the \topl constants
among $\tau_1, \dots, \tau_n$). By lemma \ref{fcn_lemma}, this implies
that:
\begin{gather}
\label{culmr:1}
\fcn(\Sigma) \subseteq 
 \fcn(\hpfunsp(\pi,n)) \union  \fcn(\hculmsp(\pi,n)) \union \\
 \fcn(\lambda) \union
 \fcn(\hconsp(\tau_{i_1})) \union \dots \union
 \fcn(\hconsp(\tau_{i_m})) \nonumber
\end{gather}
By section \ref{via_TSQL2}, for every $\kappa \in \cons$,
$\fcn(\hconsp(\kappa)) = \emptyset$. Since $\tau_{i_1}, \dots,
\tau_{i_m} \in \cons$ (see comments above), $\fcn(\hconsp(\tau_{i_1}))
= \emptyset$, \dots, $\fcn(\hconsp(\tau_{i_m})) = \emptyset$.
According to section \ref{via_TSQL2}, it is also true that for every
$\pi \in \pfuns$ and $n \in \{1,2,3,\dots\}$, $\fcn(\hpfunsp(\pi,n)) =
\emptyset$ and $\fcn(\hculmsp(\pi,n)) = \emptyset$. Hence,
\pref{culmr:1} becomes $\fcn(\Sigma) \subseteq \fcn(\lambda)$. Clause
1 has been proven.

\subsubsection*{Proof of clause 2}

According to section \ref{via_TSQL2}, $\hpfunsp(\pi, n)$ and
$\hculmsp(\pi,n)$ are \tsql \sql{SELECT} statements,
$\fcn(\hpfunsp(\pi, n)) = \fcn(\hculmsp(\pi,n)) = \emptyset$,
$eval(st, \hpfunsp(\pi, n)) \in \cvrel(n)$, and $eval(st,
\hculmsp(\pi,n)) \in \srel(n)$. The $\alpha_1$ of $\Sigma$ ranges over
the tuples of the relation $coalesce(eval(st, \hpfunsp(\pi, n)))$ (see
section \ref{tsql2_lang}). From the definition of $coalesce$, it is
easy to see that $coalesce(eval(st, \hpfunsp(\pi, n)))$ is a
valid-time relation that has the same number of explicit attributes as
$eval(st, \hpfunsp(\pi, n))$, i.e.\ $n$. Hence, $\alpha_1$ ranges over
tuples $\tup{v_1, \dots, v_n; v_t'} \in coalesce(eval(st,\hpfunsp(\pi,
n)))$. The $\alpha_2$ of $\Sigma$ ranges over the tuples of $eval(st,
\hculmsp(\pi,n))$. Since $eval(st, \hculmsp(\pi,n)) \in \srel(n)$,
$\alpha_2$ ranges over tuples $\tup{v_1, \dots, v_n} \in eval(st,
\hculmsp(\pi,n))$.

The $\alpha_1$ and $\alpha_2$ of $\Sigma$ are generated by calling the
correlation names generator after $\lambda$ has been supplied.  Hence,
$\alpha_1$ and $\alpha_2$ cannot appear in $\lambda$. The fact that
$\alpha_1$ and $\alpha_2$ do not appear in $\lambda$ means that for
every $v_1, \dots, v_n \in D$ and every $v_t' \in D_T$:
\begin{equation}
\label{culmr:3}
eval(st, \lambda, ((g^{db})^{\alpha_1}_{\tup{v_1, \dots,
v_n;v_t'}})^{\alpha_2}_{\tup{v_1, \dots, v_n}}) =
eval(st, \lambda, g^{db})
\end{equation}

The reader should now be able to see from the translation rule that
\pref{r8.4} holds. Intuitively, $\tup{v_1, \dots, v_n; v_t'}$ is the
tuple of $coalesce(eval(st, \hpfunsp(\pi, n)))$ to which $\alpha_1$
refers, and $\tup{v_1, \dots, v_n}$ is the tuple of $eval(st,
\hculmsp(\pi,n))$ to which $\alpha_2$ refers. The last line of
\pref{r8.4} corresponds to the \sql{CONTAINS} constraint in the
\sql{WHERE} clause of $\Sigma$. I should have used 
$eval(st, \lambda, ((g^{db})^{\alpha_1}_{\tup{v_1,
\dots, v_n;v_t'}})^{\alpha_2}_{\tup{v_1, \dots, v_n}})$ instead of
$eval(st, \lambda, g^{db})$, to capture the 
fact that if there are any free column references of $\alpha_1$ or
$\alpha_2$ in $\lambda$, these have to be taken to refer to the
$\tup{v_1, \dots, v_n; v_t'}$ or $\tup{v_1, \dots, v_n}$ tuples to
which $\alpha_1$ and $\alpha_2$ refer respectively. By \pref{culmr:3}, however,
$eval(st, \lambda, ((g^{db})^{\alpha_1}_{\tup{v_1,
\dots, v_n;v_t'}})^{\alpha_2}_{\tup{v_1, \dots, v_n}})$ is the same as
$eval(st, \lambda, g^{db})$.  The fifth and sixth lines of 
\pref{r8.4} correspond to the restrictions of $S_1$ and $S_2$ (see
also the comments about the equality predicate in section
\ref{eq_checks}). I do not include in the arguments of $eval(st,
\hconsp(\tau_i))$ the assignment to the correlation names,
because according to section \ref{via_TSQL2}, for $\tau_i \in \cons$,
$\fcn(\hconsp(\tau_i)) = \emptyset$. 
\begin{eqnarray}
&& eval(st,\Sigma, g^{db}) = \{ \tup{v_1,\dots,v_n;v_t} \mid 
\text{ for some } v_t',
\label{r8.4} \\ 
&& \tup{v_1,\dots,v_n;v_t'} \in coalesce(eval(st,\hpfunsp(\pi,n))), 
\nonumber \\
&& \tup{v_1,\dots,v_n} \in eval(st,\hculmsp(\pi,n)), 
\nonumber \\
&& f_D(v_t) = [minpt(f_D(v_t')), maxpt(f_D(v_t'))],
\nonumber \\
&& \mbox{if } i \in \{1,2,3,\dots,n\} \mbox{ and } \tau_i \in \cons, 
   \mbox{ then } v_i = eval(st,\hconsp(\tau_i)),  
\nonumber \\
&& \mbox{if } i,j \in \{1,2,3,\dots,n\}, \; i < j, \; \tau_i = \tau_j, \mbox{
  and } \tau_i,\tau_j \in \vars, \mbox{ then } v_i = v_j, 
\nonumber \\
&& f_D(v_t) \subseteq f_D(eval(st, \lambda, g^{db})) \} 
\nonumber 
\end{eqnarray}
According to section \ref{via_TSQL2}:
\begin{eqnarray}
&& eval(st,\hpfunsp(\pi,n)) = \hpfuns(st)(\pi,n)  \label{r8.410} \\
&& eval(st,\hculmsp(\pi,n)) = \hculms(st)(\pi,n)  \label{r8.420} \\
&& \mbox{for } \tau_i \in \cons, \;\; eval(st, \hconsp(\tau_i)) =
   \hcons(st)(\tau_i) \label{r8.430}
\end{eqnarray}
Using \pref{r8.410}, \pref{r8.420}, \pref{r8.430}, and the
definition of $coalesce$ of section \ref{tsql2_lang}, 
\pref{r8.4} becomes:
\begin{eqnarray}
&& eval(st,\Sigma, g^{db}) = \{ \tup{v_1,\dots,v_n;v_t} \mid 
\text{ for some } v_t', v_t'',
\label{r8.440} \\ 
&& \tup{v_1,\dots,v_n;v_t''} \in \hpfuns(st)(\pi,n),
\nonumber \\
&& f_D(v_t') =
 \bigcup_{\tup{v_1,\dots,v_n;v_t'''} \; \in \; 
          \hpfuns(st)(\pi,n)}f_D(v_t''')
\nonumber \\
&& \tup{v_1,\dots,v_n} \in \hculms(st)(\pi,n), 
\nonumber \\
&& f_D(v_t) = [minpt(f_D(v_t')), maxpt(f_D(v_t'))],
\nonumber \\
&& \mbox{if } i \in \{1,2,3,\dots,n\} \mbox{ and } \tau_i \in \cons, 
   \mbox{ then } v_i = \hcons(st)(\tau_i),  
\nonumber \\
&& \mbox{if } i,j \in \{1,2,3,\dots,n\}, \; i < j, \; \tau_i = \tau_j, \mbox{
  and } \tau_i,\tau_j \in \vars, \mbox{ then } v_i = v_j, 
\nonumber \\
&& f_D(v_t) \subseteq f_D(eval(st, \lambda, g^{db})) \} 
\nonumber 
\end{eqnarray}
According to \pref{r8.440}, for every tuple $\tup{v_1, \dots, v_n;
  v_t} \in eval(st, \Sigma, g^{db})$, there is a $v_t'$, such that
$f_D(v_t')$ is the union of all the temporal elements $f_D(v_t''')$
represented by time-stamps of tuples $\tup{v_1, \dots, v_n;v_t'''} \in
\hpfuns(st)(\pi,n)$. ($f_D(v_t')$ is not the empty set, because by the
second line of \pref{r8.440}, there is at least one tuple $\tup{v_1,
  \dots, v_n;v_t''} \in \hpfuns(st)(\pi,n)$). $f_D(v_t)$ is the period
$[minpt(f_D(v_t')), maxpt(f_D(v_t'))]$. This implies that $v_t \in
D_P$. We have concluded that for every tuple in $eval(st, \Sigma,
g^{db})$, the time-stamp $v_t$ is an element of $D_P$. \pref{r8.440}
also implies that $eval(st, \Sigma, g^{db})$ is a valid-time relation
of $n$ explicit attributes. Hence, $eval(st, \Sigma, g^{db}) \in
\vrel(n)$. Clause 2 has been proven.

\subsubsection*{Proof of clause 3}

Using the definition of $\denot{M(st),st,et,lt,g}{\culm[\pi(\tau_1, \dots,
\tau_n)]}$ (section \ref{culm_op}), clause 3 becomes:
\begin{eqnarray}
&& \tup{v_1,\dots,v_n;v_t} \in  eval(st,\Sigma,g^{db}) 
\mbox{ iff for some } g \text{ and } S \text{:} \nonumber \\
&& g \in G \label{r8.530} \label{r8.900} \\
&& \denot{M(st),g}{\tau_1} =
   f_D(v_1), \; \dots, \; \denot{M(st),g}{\tau_n} =
   f_D(v_n) \label{r8.540} \label{r8.902} \\
&& f_D(v_t) \subper f_D(eval(st, \lambda, g^{db}))
   \label{r8.560} \label{r8.904} \\
&& \fculms(st)(\pi,n)(\denot{M(st),g}{\tau_1},\dots,
   \denot{M(st),g}{\tau_n})= T 
   \label{r8.580} \label{r8.906} \\
&& S \not= \emptyset \label{culmr:50} \label{culmr:60} \\
&& f_D(v_t) = [minpt(S), maxpt(S)] \label{r8.590} \label{r8.907} \\
&& S = \bigcup_{p \; \in \; \fpfuns(st)(\pi,n)(\denot{M(st),g}{\tau_1}, \dots, 
\denot{M(st),g}{\tau_n})}p \label{r8.600} \label{r8.908} 
\end{eqnarray}

I first show that the forward direction of clause 3 holds. I assume that
$\tup{v_1, \dots, v_n; v_t} \in eval(st, \Sigma, g^{db})$. I need to
show that for some $g$ and $S$, \pref{r8.530} -- \pref{r8.600} hold. 
The assumption that $\tup{v_1, \dots, v_n;v_t} \in eval(st, \Sigma,
g^{db})$ and \pref{r8.440} imply that for some $v_t'$ and $v_t''$:
\begin{eqnarray}
&& \tup{v_1,\dots,v_n;v_t''} \in \hpfuns(st)(\pi,n) 
   \label{r8.460} \label{r8.910} \\
&& f_D(v_t') =
 \bigcup_{\tup{v_1,\dots,v_n;v_t'''} \; \in \; 
          \hpfuns(st)(\pi,n)}f_D(v_t''') \label{culmr:10} \label{culmr:10b} \\
&& \tup{v_1,\dots,v_n} \in \hculms(st)(\pi,n) \label{r8.470} \label{r8.920} \\
&& f_D(v_t) = [minpt(f_D(v_t')), maxpt(f_D(v_t'))] \label{r8.480.1}
 \label{r8.940} \\ 
&& \mbox{if } i \in \{1,2,3,\dots,n\} \mbox{ and } \tau_i \in \cons, 
   \mbox{ then } v_i = \hcons(st)(\tau_i) \label{r8.490} \label{r8.960}\\
&& \mbox{if } i,j \in \{1,2,3,\dots,n\}, \; i < j, \; \tau_i = \tau_j, \mbox{
  and } \tau_i,\tau_j \in \vars, \mbox{ then } v_i = v_j
  \label{r8.500} \label{r8.970}\\
&& f_D(v_t) \subseteq f_D(eval(st, \lambda, g^{db})) \label{r8.510}
   \label{r8.980} 
\end{eqnarray}
I define the mapping $g : \vars \mapsto \objs$
as follows:
\[ g(\beta) \defeq
   \left\{
   \begin{array}{l}
   f_D(v_i), \mbox{ if for some } i \in \{1,2,3,\dots,n\},
      \; \beta = \tau_i  \\
   o, \mbox{ otherwise}
   \end{array}
   \right. 
\]
where $o$\/ is a particular element of \objs, chosen arbitrarily.
\pref{r8.530} follows from lemma \ref{l4}, \pref{r8.500}, and the
definition of $g$.  I set $S$ as in \pref{culmr:11}, and show that
\pref{r8.540} -- \pref{r8.600} also hold. \pref{r8.540} follows from
lemma \ref{l2}, \pref{r8.490}, and the definition of $g$.
\begin{equation}
\label{culmr:11}
S = f_D(v_t')
\end{equation}

I now prove \pref{r8.580}. \pref{r8.540} (proven above) implies
\pref{r8.801}. \pref{r8.801} and \pref{r8.470} imply \pref{r8.802}.
\pref{r8.802} and the definition of \fculms of section
\ref{resulting_model} imply \pref{r8.580}.
\begin{gather}
  \fdi(\denot{M(st),g}{\tau_1}) = v_1, \dots, 
  \fdi(\denot{M(st),g}{\tau_n}) = v_n  \label{r8.801} \\
  \tup{\fdi(\denot{M(st),g}{\tau_1}), \dots,
   \fdi(\denot{M(st),g}{\tau_n})} \in \hculms(st)(\pi,n) \label{r8.802}
\end{gather}

I now prove \pref{culmr:50}. \pref{culmr:10} and \pref{culmr:11} imply
\pref{culmr:51}. \pref{r8.460} implies that there is at least one tuple
$\tup{v_1, \dots, v_n; v_t''}$ in $\hpfuns(st)(\pi,n)$. Then, by
\pref{culmr:51}, $S$ contains at least the chronons of $f_D(v_t'')$,
and therefore $S \not= \emptyset$. \pref{culmr:50} has been proven.
\begin{equation}
\label{culmr:51}
S = 
 \bigcup_{\tup{v_1, \dots, v_n; v_t'''} \; \in \; 
          \hpfuns(st)(\pi,n)}f_D(v_t''')
\end{equation}

I now prove \pref{r8.600}. Using \pref{r8.801}, \pref{culmr:51}
becomes \pref{culmr:13}. By the definition of \fpfuns of section
\ref{resulting_model}, \pref{culmr:14} is equivalent to
\pref{culmr:15}. Then, \pref{culmr:13} can be written as
\pref{culmr:16}. By replacing $f_D(v_t''')$ with $p$, \pref{culmr:16}
becomes \pref{r8.600}.
\begin{eqnarray}
&& S = 
 \bigcup_{\tup{\fdi(\denot{M(st),g}{\tau_1}), \dots, 
               \fdi(\denot{M(st),g}{\tau_n}); v_t'''} \; \in \; 
          \hpfuns(st)(\pi,n)}f_D(v_t''')
   \label{culmr:13} \\
&& \tup{\fdi(\denot{M(st),g}{\tau_1}), \dots,
   \fdi(\denot{M(st),g}{\tau_n}); v_t'''} \in \hpfuns(st)(\pi,n)
   \label{culmr:14} \\
&& f_D(v_t''') \in \fpfuns(\pi,n)(\denot{M(st),g}{\tau_1}, \dots, 
                                  \denot{M(st),g}{\tau_n}) \label{culmr:15} \\
&& S = 
 \bigcup_{f_D(v_t''') \in \fpfuns(\pi,n)(\denot{M(st),g}{\tau_1}, \dots, 
          \denot{M(st),g}{\tau_n})}f_D(v_t''') \label{culmr:16}
\end{eqnarray}
\pref{r8.590} follows from \pref{culmr:11} and \pref{r8.480.1}. It
remains to prove \pref{r8.560}. \pref{culmr:50} and \pref{r8.590}
(both proven above) imply that $f_D(v_t) \in \periods$. From the
hypothesis, $eval(st, \lambda, g^{db}) \in D_P^*$, which implies that
$f_D(eval(st, \lambda, g^{db}))$ is a period or the empty set.
$f_D(eval(st, \lambda, g^{db}))$ cannot be the empty set, because
according to \pref{r8.510} $f_D(v_t)$ (which is a period and therefore
a non-empty set) is a subset of $f_D(eval(st, \lambda, g^{db}))$.
Hence, $f_D(eval(st, \lambda, g^{db})) \in \periods$. \pref{r8.510}
and the fact that both $f_D(v_t)$ and $f_D(eval(st, \lambda, g^{db}))$
are periods imply \pref{r8.560}. The forward direction of clause 3 has
been proven.
 
\bigskip

I now prove the backwards direction of clause 3. I assume that
\pref{r8.900} -- \pref{r8.908} hold. I need to prove that
$\tup{v_1,\dots,v_n;v_t} \in eval(st,\Sigma,g^{db})$. According to
\pref{r8.440}, it is enough to prove that for some $v_t'$ and $v_t''$,
\pref{r8.910} -- \pref{r8.980} hold. \pref{r8.980} follows from
\pref{r8.904}.

I now prove \pref{r8.960}. If $i \in \{1,2,3,\dots,n\}$ and $\tau_i
\in \cons$, then the semantics of \topl implies \pref{r8.more.54}, and
\pref{r8.902} implies \pref{r8.more.55}. 
\begin{gather}
  \denot{M(st),g}{\tau_i} = \fcons(st)(\tau_i) \label{r8.more.54} \\
  \denot{M(st),g}{\tau_i} = f_D(v_i) \label{r8.more.55}
\end{gather}
\pref{r8.more.54} and \pref{r8.more.55} imply that $f_D(v_i) =
\fcons(st)(\tau_i)$, which in turn implies \pref{r8.more.57}. The
definition of \fcons of section \ref{resulting_model} implies
\pref{r8.more.58}. \pref{r8.more.57} and \pref{r8.more.58} imply that
$v_i = \hcons(st)(\tau_i)$. \pref{r8.960} has been proven.
\begin{gather}
  v_i = \fdi(\fcons(st)(\tau_i)) \label{r8.more.57} \\
  \fcons(st)(\tau_i) = f_D(\hcons(st)(\tau_i))  \label{r8.more.58}
\end{gather}

I now prove \pref{r8.970}. If $i,j \in \{1,2,3,\dots,n\}$ and $\tau_i
= \tau_j$, then $\denot{M(st),g}{\tau_i} = \denot{M(st),g}{\tau_j}$.
Then, from \pref{r8.902}, $f_D(v_i) = f_D(v_j)$, which implies
that $\fdi(f_D(v_i)) = \fdi(f_D(v_j))$, i.e.\ $v_i = v_j$.
\pref{r8.970} has been proven.

I now prove \pref{r8.920}. \pref{r8.906} and the definition of
\fculms of section \ref{resulting_model} imply \pref{r8.1000}. \pref{r8.902} implies \pref{r8.1001}. \pref{r8.1000} and \pref{r8.1001} imply \pref{r8.920}.
\begin{eqnarray}
&& \tup{\fdi(\denot{M(st),g}{\tau_1}), \dots, 
   \fdi(\denot{M(st),g}{\tau_n})} \in \hculms(st)(\pi,n) \label{r8.1000} \\
&& \fdi(\denot{M(st),g}{\tau_1}) = v_1, \dots, 
   \fdi(\denot{M(st),g}{\tau_n}) = v_n  \label{r8.1001} 
\end{eqnarray}
 
It remains to prove that \pref{r8.910}, \pref{culmr:10b}, and
\pref{r8.940} hold for some $v_t', v_t''$. I start with 
\pref{r8.910}. According to section \ref{resulting_model},
$\fpfuns(st)(\pi,n)(\denot{M(st),g}{\tau_1}, \dots,
\denot{M(st),g}{\tau_n})$ is a set of periods. By 
\pref{r8.908}, $S$ is the union of all these periods.
According to \pref{culmr:60}, $S$ is not the empty set, which implies
that there is at least one period $p'$ such that:
\begin{equation}
\label{culmr:70}
p' \in \fpfuns(st)(\pi,n)(\denot{M(st),g}{\tau_1}, \dots, 
\denot{M(st),g}{\tau_n})
\end{equation}
Let $v_t'' = \fdi(p')$, which implies that $p' = f_D(v_t'')$. Then, from
\pref{culmr:70} we get:
\begin{equation}
\label{culmr:61}
f_D(v_t'') \in \fpfuns(st)(\pi,n)(\denot{M(st),g}{\tau_1}, \dots,
\denot{M(st),g}{\tau_n})
\end{equation}
\pref{culmr:61} and the definition of \fpfuns of section
\ref{resulting_model} imply \pref{culmr:62}. \pref{culmr:62} and
\pref{r8.1001} imply \pref{r8.910}.
\begin{eqnarray}
&& \tup{\fdi(\denot{M(st),g}{\tau_1}), \dots,
   \fdi(\denot{M(st),g}{\tau_n}); v_t''} \in \hpfuns(st)(\pi,n)
   \label{culmr:62}
\end{eqnarray}

From the discussion above, $S$ is a (non-empty) union of periods. This
implies that $S$ is a temporal element, which in turn implies that
there is a $v_t' \in D_T$, such that \pref{culmr:63} holds.
\pref{r8.940} follows from \pref{r8.907} and \pref{culmr:63}.
\begin{equation}
\label{culmr:63}
f_D(v_t') = S
\end{equation}
I now prove \pref{culmr:10b}. \pref{r8.908} and \pref{culmr:63} imply
\pref{culmr:64}. 
\begin{equation}
\label{culmr:64}
f_D(v_t') = 
\bigcup_{p \; \in \; \fpfuns(st)(\pi,n)(\denot{M(st),g}{\tau_1}, \dots, 
\denot{M(st),g}{\tau_n})}p 
\end{equation}
According to the discussion above,
$\fpfuns(st)(\pi,n)(\denot{M(st),g}{\tau_1}, \dots, \denot{M(st),g}{\tau_n})$
is a set of periods. Therefore, for every $p$ in that set, there is a
$v_t''' \in D_P$, such that $f_D(v_t''') = p$. Hence, \pref{culmr:64}
can be written as \pref{culmr:66}. 
\begin{equation}
\label{culmr:66}
f_D(v_t') = 
\bigcup_{f_D(v_t''') \; \in \; 
\fpfuns(st)(\pi,n)(\denot{M(st),g}{\tau_1}, \dots, 
\denot{M(st),g}{\tau_n})}f_D(v_t''') 
\end{equation}
By the definition of \fpfuns of section \ref{resulting_model},
\pref{culmr:67} is equivalent to \pref{culmr:68}. By \pref{r8.1001},
\pref{culmr:68} is in turn equivalent to \pref{culmr:69}. \pref{culmr:66} and
the fact that \pref{culmr:67} is equivalent to \pref{culmr:69} imply
\pref{culmr:10b}. The backwards direction of clause 3 has been proven.
\begin{eqnarray}
\label{culmr:67}
&& f_D(v_t''') \in \fpfuns(st)(\pi,n)(\denot{M(st),g}{\tau_1}, \dots, 
   \denot{M(st),g}{\tau_n}) \\
\label{culmr:68}
&& \tup{\fdi(\denot{M(st),g}{\tau_1}), \dots, 
      \fdi(\denot{M(st),g}{\tau_n}); v_t'''} \in \hpfuns(st)(\pi,n) \\
\label{culmr:69}
&& \tup{v_1, \dots, v_n; v_t'''} \in \hpfuns(st)(\pi,n)
\end{eqnarray}


\subsection{$\phi_1 \land \phi_2$} \label{conj_rule}

\subsubsection*{Translation rule}

If $\phi_1, \phi_2 \in \ynforms$ and $\lambda$ is a \tsql value
expression, then: 

$trans(\phi_1 \land \phi_2, \lambda) \defeq$\\
\sql{(}\select{SELECT DISTINCT
                  $\alpha_1.1$, $\alpha_1.2$, \dots, $\alpha_1.n_1$, 
                  $\alpha_2.1$, $\alpha_2.2$, \dots, $\alpha_2.n_2$\\
               VALID VALID($\alpha_1$) \\
               FROM $trans(\phi_1,\lambda)$ AS $\alpha_1$,
                    $trans(\phi_2,\lambda)$ AS $\alpha_2$ \\
               WHERE \dots \\
               \ \ AND \dots \\
               \ \ \vdots \\
               \ \ AND \dots \\
               \ \ AND VALID($\alpha_1$) = VALID($\alpha_2$))} 

$n_1$ and $n_2$ are the lengths of $\corn{\phi_1}$ and $\corn{\phi_2}$
respectively. Each time the translation rule is used, $\alpha_1$ and
$\alpha_2$ are two new different correlation names, obtained by calling the
correlation names generator after $trans(\phi_1, \lambda)$ and
$trans(\phi_2, \lambda)$ have been computed. Assuming that $\corn{\phi_1} =
\tup{\tau^1_1, \dots, \tau^1_{n_1}}$ and $\corn{\phi_2} =
\tup{\tau^2_1, \dots, \tau^2_{n_2}}$, the ``\dots''s in the \sql{WHERE} clause
are all the strings in $S$:
\[
\begin{aligned}
S = \{&\mbox{``}\alpha_1.i = \alpha_2.j\mbox{''} \mid
 i \in \{1,2,3,\dots,n_1\}, \; j \in \{1,2,3,\dots,n_2\}, \\
      &\tau^1_i = \tau^2_j, \mbox{ and } \tau^1_i, \tau^2_j \in \vars \} 
\end{aligned}
\]

\subsubsection*{Proof that theorem \ref{yn_theorem} holds for $\phi =
\phi_1 \land \phi_2$, if it holds for $\phi = \phi_1$ and $\phi = \phi_2$}

I assume that $\phi_1, \phi_2 \in \ynforms$. By the syntax of \topl,
this implies that $\phi_1 \land \phi_2 \in \ynforms$. I also assume
that $st \in \pts$, $\lambda$ is a \tsql value expression, $g^{db} \in
G^{db}$, $eval(st,\lambda, g^{db}) \in D_P^*$, $\corn{\phi_1} =
\tup{\tau_1^1, \dots, \tau_{n_1}^1}$, $\corn{\phi_2} = \tup{\tau_1^2,
  \dots, \tau_{n_2}^2}$, and $\Sigma = trans(\phi_1 \land \phi_2,
\lambda)$. From the definition of $\corn{\dots}$, it should be easy to
see that $\corn{\phi_1 \land \phi_2} =\tup{\tau_1^1, \dots,
  \tau_{n_1}^1, \tau_1^2, \dots, \tau_{n_2}^2}$. Finally, I assume
that theorem \ref{yn_theorem} holds for $\phi = \phi_1$ and $\phi =
\phi_2$. I need to show that:
\begin{enumerate}
\item $\fcn(\Sigma) \subseteq \fcn(\lambda)$

\item $eval(st,\Sigma,g^{db}) \in \vrel(n_1 + n_2)$

\item $\tup{v^1_1,\dots,v^1_{n_1},v^2_1,\dots,v^2_{n_2};v_t} \in 
  eval(st,\Sigma,g^{db})$ iff for some $g \in G$:\\
  $\denot{M(st),g}{\tau^1_1} = f_D(v^1_1)$, \dots, 
  $\denot{M(st),g}{\tau^1_{n_1}} = f_D(v^1_{n_1})$,\\ 
  $\denot{M(st),g}{\tau^2_1} = f_D(v^2_1)$, \dots, 
  $\denot{M(st),g}{\tau^2_{n_2}} = f_D(v^2_{n_2})$,\\
  $\denot{M(st),st,f_D(v_t),f_D(eval(st,\lambda,g^{db})),g}
         {\phi_1 \land \phi_2} = T$.
\end{enumerate}

Let $\Sigma_1$ and $\Sigma_2$ be the embedded
\sql{SELECT} statements in the \sql{FROM} clause of $\Sigma$
(i.e.\ $\Sigma_1 = trans(\phi_1, \lambda)$ and $\Sigma_2
= trans(\phi_2, \lambda)$). From the hypothesis, $\phi_1, \phi_2 \in
\ynforms$, $st \in \pts$, $\corn{\phi_1} = \tup{\tau_1^1, \dots,
\tau_{n_1}^1}$, $\corn{\phi_2} = \tup{\tau_1^2, \dots,
\tau_{n_2}^2}$, $g^{db} \in G^{db}$, and $eval(st, \lambda,
g^{db}) \in D_P^*$. Then, from theorem \ref{yn_theorem} for $\phi =
\phi_1$ (according to the hypothesis, theorem \ref{yn_theorem} holds
for $\phi = \phi_1$ and $\phi = \phi_2$) we get: 
\begin{enumerate}
\item[$1^1$.] $\fcn(\Sigma_1) \subseteq \fcn(\lambda)$

\item[$2^1$.] $eval(st,\Sigma_1,g^{db}) \in \vrel(n_1)$

\item[$3^1$.] $\tup{v^1_1,\dots,v^1_{n_1};v_t} \in
  eval(st,\Sigma_1,g^{db})$ iff for some $g_1 \in G$:\\
  $\denot{M(st),g_1}{\tau^1_1} = f_D(v^1_1)$, \dots, 
  $\denot{M(st),g_1}{\tau^1_{n_1}} = f_D(v^1_{n_1})$,\\ 
  $\denot{M(st),st,f_D(v_t),f_D(eval(st, \lambda, g^{db})),g_1}{\phi_1} = T$.
\end{enumerate}
and from theorem \ref{yn_theorem} for $\phi =
\phi_2$:
\begin{enumerate}
\item[$1^2$.] $\fcn(\Sigma_2) \subseteq \fcn(\lambda)$

\item[$2^2$.] $eval(st,\Sigma_2,g^{db}) \in \vrel(n_2)$

\item[$3^2$.] $\tup{v^2_1,\dots,v^2_{n_2};v_t} \in
  eval(st,\Sigma_2,g^{db})$ iff for some $g_2 \in G$:\\
  $\denot{M(st),g_2}{\tau^2_1} = f_D(v^2_1)$, \dots, 
  $\denot{M(st),g_2}{\tau^2_{n_2}} = f_D(v^2_{n_2})$,\\ 
  $\denot{M(st),st,f_D(v_t),f_D(eval(st, \lambda, g^{db})),g_2}{\phi_2} = T$.
\end{enumerate}

\subsubsection*{Proof of clause 1}

The two \sql{VALID($\alpha_1$)} in the \sql{VALID} and \sql{WHERE}
clauses of $\Sigma$, and the \sql{VALID($\alpha_2$)} in the
\sql{WHERE} clause of $\Sigma$ are not free column references in
$\Sigma$, because $\Sigma$ is a binding context for all of them. The
$\alpha_1.1$, \dots, $\alpha_1.n_1$, $\alpha_2.1$, \dots,
$\alpha_2.n_2$ in the \sql{SELECT} clause of $\Sigma$, and any column
references of the form $\alpha_1.i$ or $\alpha_2.j$ in the \sql{WHERE}
clause of $\Sigma$ ($i \in \{1,2,3,\dots,n_1\}$, $j \in
\{1,2,3,\dots,n_2\}$; these column references derive from $S$) are not
free column references in $\Sigma$ for the same reason. $\Sigma$
contains no other column references (and hence no other free column
references), apart from those that possibly appear within $\Sigma_1$
or $\Sigma_2$. By lemma \ref{fcn_lemma}, this implies \pref{conjr:1}.
\pref{conjr:1}, clause $1^1$, and clause $1^2$ imply clause 1.
\begin{equation}
\label{conjr:1}
\fcn(\Sigma) \subseteq \fcn(\Sigma_1) \union \fcn(\Sigma_2)
\end{equation}

\subsubsection*{Proof of clause 2}

When computing $eval(st, \Sigma, g^{db})$, $\alpha_1$ and $\alpha_2$
range over the tuples of $eval(st,\Sigma_1,g^{db})$ and
$eval(st,\Sigma_2,g^{db})$ respectively.  Clauses $2^1$ and $2^2$
imply that $eval(st,\Sigma_1,g^{db})$ and $eval(st,\Sigma_2,g^{db})$
are valid-time relations of $n_1$ and $n_2$ explicit attributes
respectively. Hence, $\alpha_1$ ranges over tuples $\tup{v^1_1, \dots,
  v^1_{n_1};v^1_t} \in eval(st,\Sigma_1,g^{db})$, and $\alpha_2$
ranges over tuples $\tup{v^2_1, \dots, v^2_{n_2};v^2_t} \in
eval(st,\Sigma_2,g^{db})$.

$\alpha_1$ is generated after computing $\Sigma_2 = trans(\phi_2,
\lambda)$ (see the translation rule). Hence, $\alpha_1$ cannot appear
in $\Sigma_2$. Since $\alpha_1$ does not appear in $\Sigma_2$, for
every tuple $\tup{v_1^1, \dots, v_n^2; v_t^1}$:
\begin{equation}
\label{conjr:3}
eval(st, \Sigma_2, (g^{db})^{\alpha_1}_{\tup{v^1_1, \dots, v^1_n;v^1_t}}) =
eval(st, \Sigma_2, g^{db})
\end{equation}

It should now be easy to see from the translation rule that
\pref{r9.3} holds. $\tup{v^1_1,\dots,v^1_{n_1};v_t}$ is
the tuple of $eval(st,\Sigma_1,g^{db})$ that corresponds to
$\alpha_1$, while $\tup{v^2_1,\dots,v^2_{n_2};v_t}$ is the tuple of
$eval(st,\Sigma_2,g^{db})$ that corresponds to $\alpha_2$. The last
constraint in the \sql{WHERE} clause of $\Sigma$ requires the
time-stamps of the two tuples to be identical. The last constraint in
\pref{r9.3} corresponds to the constraints in the \sql{WHERE} clause
of $\Sigma$ that derive from $S$ (see also the comments about the
equality predicate in section \ref{eq_checks}). I should have used
$eval(st, \Sigma_2, (g^{db})^{\alpha_1}_{\tup{v^1_1,
\dots, v^1_{n_1};v_t}})$ instead of $eval(st, \Sigma_2, g^{db})$, to
capture the fact that if there is any free column reference of $\alpha_1$ in
$\Sigma_2$, this has to be taken to refer to the $\tup{v^1_1, \dots,
v^1_{n_1}; v_t}$ tuple to which $\alpha_1$ refers. By \pref{conjr:3},
however, $eval(st, \Sigma_2, (g^{db})^{\alpha_1}_{\tup{v^1_1,
\dots, v^1_{n_1};v_t}})$ is the same as $eval(st, \Sigma_2, g^{db})$. 
\begin{eqnarray}
&& eval(st,\Sigma,g^{db}) = 
   \{ \tup{v^1_1,\dots,v^1_{n_1},v^2_1,\dots,v^2_{n_2};v_t} \mid
\label{r9.3} \\
&&    \tup{v^1_1,\dots,v^1_{n_1};v_t} \in eval(st,\Sigma_1,g^{db}), 
\nonumber \\
&&    \tup{v^2_1,\dots,v^2_{n_2};v_t} \in eval(st,\Sigma_2,g^{db}),
\mbox{ and} \nonumber \\
&& \mbox{if } i \in \{1,2,3,\dots,n_1\}, \; j \in
   \{1,2,3,\dots,n_2\}, \; \tau^1_i, \tau^2_j \in \vars, 
   \mbox{ and } \tau^1_i = \tau^2_j,
\nonumber \\
&& \mbox{then } v^1_i = v^2_j \}
\nonumber
\end{eqnarray}
According to \pref{r9.3}, in every tuple of $eval(st, \Sigma,
g^{db})$, the time-stamp $v_t$ is also the time-stamp of a tuple in
$eval(st, \Sigma_1, g^{db})$. By clause $1^1$, $eval(st, \Sigma_1,
g^{db}) \in \vrel(n_1)$, which implies that $v_t \in D_P$. Hence, all
the time-stamps of $eval(st, \Sigma, g^{db})$ are elements of $D_P$.
\pref{r9.3} also implies that $eval(st, \Sigma, g^{db})$ is a
valid-time relation of $n_1 + n_2$ explicit attributes. Hence,
$eval(st,\Sigma,g^{db}) \in \vrel(n_1 + n_2)$. Clause 2 has been
proven.

\subsubsection*{Proof of clause 3}

Using the definition of $\denot{M(st),st,et,lt,g}{\phi_1 \land
\phi_2}$ (section \ref{denotation}), clause 3 becomes: 
\begin{eqnarray}
&& \tup{v^1_1,\dots,v^1_{n_1},v^2_1,\dots,v^2_{n_2};v_t} \in
  eval(st,\Sigma,g^{db}) \mbox{ iff for some } g: \nonumber \\
&& g \in G \label{r9.5} \label{r9.98} \\
&& \denot{M(st),g}{\tau^1_1} = f_D(v^1_1), \dots, 
   \denot{M(st),g}{\tau^1_{n_1}} = f_D(v^1_{n_1}) \label{r9.7} \label{r9.100}\\
&& \denot{M(st),g}{\tau^2_1} = f_D(v^2_1), \dots, 
   \denot{M(st),g}{\tau^2_{n_2}} = f_D(v^2_{n_2}) \label{r9.8} \label{r9.101}\\
&& \denot{M(st),st,f_D(v_t),f_D(eval(st, \lambda, g^{db})),g}{\phi_1} = T
   \label{r9.9} \label{r9.103}\\
&& \denot{M(st),st,f_D(v_t),f_D(eval(st, \lambda, g^{db})),g}{\phi_2} = T
   \label{r9.10} \label{r9.104} 
\end{eqnarray}

I first prove the forward direction of clause 3. I assume
that \pref{r9.4} holds. I need to prove that for some $g$, \pref{r9.5}
-- \pref{r9.10} also hold. 
\begin{equation}
  \label{r9.4}
  \tup{v^1_1,\dots,v^1_{n_1},v^2_1,\dots,v^2_{n_2};v_t} \in
  eval(st,\Sigma,g^{db}) 
\end{equation}
\pref{r9.4} and \pref{r9.3} imply that:
\begin{eqnarray}
&& \tup{v^1_1,\dots,v^1_{n_1};v_t} \in eval(st,\Sigma_1,g^{db})
   \label{r9.11} \label{r9.105} \\ 
&& \tup{v^2_1,\dots,v^2_{n_2};v_t} \in eval(st,\Sigma_2,g^{db}) 
   \label{r9.12} \label{r9.106} \\
&& \mbox{if } i \in \{1,2,3,\dots,n_1\}, \; j \in
   \{1,2,3,\dots,n_2\}, \; \tau^1_i, \tau^2_j \in \vars, 
   \mbox{ and } \tau^1_i = \tau^2_j,
   \label{r9.13} \label{r9.107} \\
&& \;\;\;\; \mbox{then } v^1_i = v^2_j
\nonumber 
\end{eqnarray}
\pref{r9.11} and clause $3^1$ imply that for some $g_1$:
\begin{eqnarray}
&& g_1 \in G \label{r9.14} \\
&& \denot{M(st),g_1}{\tau^1_1} = f_D(v^1_1), \dots, 
   \denot{M(st),g_1}{\tau^1_{n_1}} = f_D(v^1_{n_1}) \label{r9.16} \\
&& \denot{M(st),st,f_D(v_t),f_D(eval(st, \lambda,
g^{db})),g_1}{\phi_1} = T \label{r9.18.1} 
\end{eqnarray}
Similarly, \pref{r9.12} and clause $3^2$ imply that for
some $g_2$:
\begin{eqnarray}
&& g_2 \in G \label{r9.18} \\
&& \denot{M(st),g_2}{\tau^2_1} = f_D(v^2_1), \dots, 
   \denot{M(st),g_2}{\tau^2_{n_2}} = f_D(v^2_{n_2}) \label{r9.20} \\
&& \denot{M(st),st,f_D(v_t),f_D(eval(st, \lambda, g^{db})),g_2}{\phi_2} = T
\label{r9.22} 
\end{eqnarray}

I define the mapping $g : \vars \mapsto \objs$ as
follows:
\[ g(\beta) \defeq
   \left\{
   \begin{array}{l}
   g_1(\beta), \mbox{ if for some } i \in \{1,2,3,\dots,n_1\},
      \; \beta = \tau^1_i  \\
   g_2(\beta), \mbox{ if for some } j \in \{1,2,3,\dots,n_2\},
      \; \beta = \tau^2_j  \\
   o, \mbox{ otherwise}
   \end{array}
   \right. 
\]
where $o$ is a particular element of \objs, chosen arbitrarily. 
\pref{r9.5} follows from lemma \ref{l5}, the definition of $g$,
\pref{r9.13}, \pref{r9.16}, and \pref{r9.20}. 

The definition of $g$ implies that $g$ and $g_1$ assign the same
values to all the variables among $\tau^1_1, \dots, \tau^1_{n_1}$. The
assumption that $\corn{\phi_1} = \tup{\tau^1_1, \dots, \tau^1_{n_1}}$
and the definition of $\corn{\phi_1}$ imply that all the variables of
$\phi_1$ are among $\tau^1_1, \dots, \tau^1_{n_1}$. Hence, $g$ and
$g_1$ assign the same values to all the variables in $\phi_1$.
\pref{r9.7} follows from lemma \ref{l3}, the assumption that
$\tup{\tau^1_1, \dots, \tau^1_{n_1}} = \corn{\phi_1}$, \pref{r9.16},
and the fact that $g$ and $g_1$ assign the same values to all the
variables of $\phi_1$. The proof of \pref{r9.8} is very similar.

I now prove \pref{r9.9}. The fact that $g$ and $g_1$ assign the same
values to all the variables of $\phi_1$ implies \pref{r9.50}.
\pref{r9.50} and \pref{r9.18.1} imply \pref{r9.9}. The proof of
\pref{r9.10} is very similar. This concludes the proof of the forward
direction of clause 3.
\begin{eqnarray}
&& \denot{M(st),st,f_D(v_t),f_D(eval(st, \lambda, g^{db})),g}{\phi_1} = 
    \denot{M(st),st,f_D(v_t),f_D(eval(st, \lambda, g^{db})),g_1}{\phi_1}
    \label{r9.50}
\end{eqnarray}

\bigskip

I now prove the backwards direction of clause 3. I assume that for
some $g$, \pref{r9.98} -- \pref{r9.104} hold.  I need to prove that
$\tup{v^1_1,\dots,v^1_{n_1},v^2_1,\dots,v^2_{n_2};v_t} \in
eval(st,\Sigma,g^{db})$. According to \pref{r9.3}, it is enough to
prove \pref{r9.105} -- \pref{r9.107}.  Clause $3^1$, \pref{r9.98},
\pref{r9.100}, and \pref{r9.103} imply \pref{r9.105}.  Clause $3^2$,
\pref{r9.98}, \pref{r9.101}, and \pref{r9.104} imply \pref{r9.106}.

It remains to prove \pref{r9.107}. Let us assume that $i \in
\{1,2,3,\dots,n_1\}$, $j \in \{1,2,3,\dots,n_2\}$, $\tau^1_i, \tau^2_j
\in \vars$, and $\tau^1_i = \tau^2_j$. \pref{r9.100} and \pref{r9.101}
imply that $f_D(v^1_i) = f_D(v^2_j)$. This in turn implies that
$\fdi(f_D(v^1_i)) = \fdi(f_D(v^2_j))$, i.e.\ that $v_i^1 = v_j^2$.
\pref{r9.107} and the backwards direction of clause 3 have been
proven.


\subsection{$Pres[\phi']$}

\subsubsection*{Translation rule}

If $\phi' \in \ynforms$ and $\lambda$ is a \tsql value expression, then:

$trans(\pres[\phi'], \lambda) \defeq$\\
\sql{(}\select{SELECT DISTINCT $\alpha$.1, $\alpha$.2, \dots, $\alpha$.$n$ \\
               VALID VALID($\alpha$) \\
               FROM $trans(\phi', \lambda)$ AS $\alpha$ \\
               WHERE VALID($\alpha$) CONTAINS TIMESTAMP 'now')}

$n$ is the length of $\corn{\phi'}$. Each time the translation rule is
used, $\alpha$ is a new correlation name, obtained by calling the
correlation names generator.

\subsubsection*{Proof that theorem \ref{yn_theorem} holds for $\phi =
Pres[\phi']$, if it holds for $\phi = \phi'$}

I assume that $\phi' \in \ynforms$. By the syntax of
\topl, this implies that $\pres[\phi'] \in \ynforms$. I also
assume that $st \in \pts$, $\lambda$ is a \tsql 
value expression, $g^{db} \in G^{db}$, $eval(st,\lambda, g^{db})
\in D_P^*$, $\corn{\phi'} = \tup{\tau_1, \tau_2, \tau_3, \dots,
\tau_n}$, and
$\Sigma = trans(\pres[\phi'], \lambda)$. From the definition of
$\corn{\dots}$, it should be easy to see that $\corn{\pres[\phi']} =
\corn{\phi'} = \tup{\tau_1, \dots, \tau_n}$. Finally, I assume that theorem
\ref{yn_theorem} holds for $\phi = \phi'$. I need to show that:
\begin{enumerate}
\item $\fcn(\Sigma) \subseteq \fcn(\lambda)$
\item $eval(st, \Sigma, g^{db}) \in \vrel(n)$
\item $\tup{v_1, \dots, v_n; v_t} \in 
       eval(st, \Sigma, g^{db})$ iff for some $g \in G$: \\
   $\denot{M(st), g}{\tau_1} = f_D(v_1)$, \dots, 
   $\denot{M(st), g}{\tau_n} = f_D(v_n)$, and \\
   $\denot{M(st), st, f_D(v_t), f_D(eval(st, \lambda, g^{db})), g}{
   \pres[\phi']} = T$
\end{enumerate}

Let $\Sigma'$ be the embedded \sql{SELECT} statement in the
\sql{FROM} clause of $\Sigma$, i.e.\ $\Sigma' = trans(\phi',
\lambda)$. From the hypothesis, $\phi' \in \ynforms$, $st \in \pts$,
$\corn{\phi'} = \tup{\tau_1, \dots, \tau_n}$, $g^{db} \in
G^{db}$, and $eval(st, \lambda, g^{db}) \in D_P^*$.  Then,
from theorem \ref{yn_theorem} for $\phi = \phi'$ (according to the
hypothesis, theorem \ref{yn_theorem} holds for $\phi = \phi'$) we get:
\begin{enumerate}
\item[$1'.$] $\fcn(\Sigma') \subseteq \fcn(\lambda)$
\item[$2'.$] $eval(st, \Sigma', g^{db}) \in \vrel(n)$
\item[$3'.$] $\tup{v_1, \dots, v_n; v_t} \in 
       eval(st, \Sigma', g^{db})$ iff for some $g \in G$: \\
   $\denot{M(st), g}{\tau_1} = f_D(v_1)$, \dots, 
   $\denot{M(st), g}{\tau_n} = f_D(v_n)$, and \\
   $\denot{M(st), st, f_D(v_t), f_D(eval(st, \lambda, g^{db})), g}{
   \phi'} = T$
\end{enumerate}

\subsubsection*{Proof of clause 1}

The two \sql{VALID($\alpha$)} in the \sql{VALID} and \sql{WHERE}
clauses of $\Sigma$ are not free column references in $\Sigma$,
because $\Sigma$ is a binding context for both of them. The
$\alpha.1$, \dots, $\alpha.n$ in the \sql{SELECT} clause of $\Sigma$
are not free column references in $\Sigma$ for the same reason.
$\Sigma$ contains no other column references (and hence no other free
column references), apart from those that possibly appear within
$\Sigma'$. By lemma \ref{fcn_lemma}, this implies \pref{presr:1}.
\pref{presr:1} and clause $1'$ imply clause 1.
\begin{equation}
\label{presr:1}
\fcn(\Sigma) \subseteq \fcn(\Sigma')
\end{equation}

\subsubsection*{Proof of clause 2}

Given that $eval(st, \Sigma', g^{db}) \in \vrel(n)$ (from clause
$2'$), it should be easy to see from the translation rule that:
\begin{eqnarray}
&&  eval(st,\Sigma,g^{db}) = \{\tup{v_1,\dots,v_n;v_t} \in
    eval(st,\Sigma',g^{db}) \mid st \in f_D(v_t)\}
    \label{r2.8}
\end{eqnarray}
By \pref{r2.8}, any time-stamp $v_t$ in $eval(st,\Sigma,g^{db})$ is
also a time-stamp in $eval(st, \Sigma', g^{db})$. This implies that
$v_t \in D_P$, because $eval(st, \Sigma', g^{db}) \in \vrel(n)$.
\pref{r2.8} also implies that $eval(st, \Sigma, g^{db})$ is a
valid-time relation of $n$ explicit attributes. Therefore, $eval(st,
\Sigma, g^{db}) \in \vrel(n)$. Clause 2 has been proven.

\subsubsection*{Proof of clause 3}

Using the definition of $\denot{M(st),st,et,lt,g}{\pres[\phi']}$
(section \ref{pres_op}), clause 3 becomes:
\begin{eqnarray}
&& \tup{v_1,\dots,v_n;v_t} \in eval(st,\Sigma,g^{db}) 
\mbox{ iff for some } g \in G:
\label{r2.7} \\
&& \denot{M(st),g}{\tau_1} =
   f_D(v_1), \; \dots, \; \denot{M(st),g}{\tau_n} = f_D(v_n), \nonumber\\
&& st \in f_D(v_t), \mbox{ and }
   \denot{M(st),st,f_D(v_t),f_D(eval(st, \lambda, g^{db})),g}{\phi'} = T
\nonumber 
\end{eqnarray}

I first prove the forward direction of \pref{r2.7}. If
$\tup{v_1,\dots,v_n;v_t} \in eval(st,\Sigma,g^{db})$, then according to
\pref{r2.8}:
\begin{eqnarray}
  && \tup{v_1,\dots,v_n;v_t} \in eval(st,\Sigma',g^{db}) \label{r2.9} \\
  && st \in f_D(v_t) \label{r2.10}
\end{eqnarray}
\pref{r2.9}, clause $3'$, and \pref{r2.10} imply that for some $g \in G$:
\begin{eqnarray}
&& \denot{M(st),g}{\tau_1} =
   f_D(v_1), \dots, \denot{M(st),g}{\tau_n} =
   f_D(v_n) \label{r2.21} \\
&& st \in f_D(v_t), \text{ and } \denot{M(st),st,f_D(v_t),f_D(eval(st,
\lambda, g^{db})),g}{\phi'} = T \nonumber
\end{eqnarray}
The forward direction of \pref{r2.7} has been proven. I now prove the
backwards direction of \pref{r2.7}. If for some $g \in G$,
\pref{r2.21} holds, then according to clause $3'$, \pref{r2.20} is
true. Also \pref{r2.21} implies \pref{r2.22}. \pref{r2.20},
\pref{r2.22}, and \pref{r2.8} imply that $\tup{v_1,\dots,v_n;v_t} \in
eval(st,\Sigma,g^{db})$. The backwards direction of \pref{r2.7} has
been proven.
\begin{eqnarray}
&& \tup{v_1,\dots,v_n;v_t} \in eval(st,\Sigma',g^{db})  \label{r2.20} \\
&& st \in f_D(v_t) \label{r2.22} 
\end{eqnarray}


\subsection{$Past[\beta, \phi']$}

\subsubsection*{Translation rule}

If $\beta \in \vars$, $\phi' \in \ynforms$, and $\lambda$ is a \tsql
value expression, then:

$trans(\past[\beta, \phi'], \lambda) \defeq$\\
\sql{(}\select{SELECT DISTINCT VALID($\alpha$), 
                      $\alpha$.1, $\alpha$.2, \dots, $\alpha$.$n$ \\
               VALID VALID($\alpha$) \\
               FROM $trans(\phi', \lambda')$ AS $\alpha$)}

$\lambda'$ is the expression 
\sql{INTERSECT($\lambda$, PERIOD(TIMESTAMP
'beginning', TIMESTAMP 'now' - INTERVAL '1' $\chi$))}, 
$\chi$ is the \tsql name of the granularity of chronons
(e.g.\ \sql{DAY} if chronons correspond to days), and $n$ is the length of
$\corn{\phi'}$. Each time the translation rule is used, $\alpha$ is a
new correlation name, obtained by calling the correlation names generator.

\subsubsection*{Proof that theorem \ref{yn_theorem} holds for $\phi =
Past[\beta, \phi']$, if it holds for $\phi = \phi'$}

I assume that $\beta \in \vars$ and $\phi' \in \ynforms$. By the
syntax of \topl, this implies that $\past[\beta, \phi'] \in \ynforms$.
I also assume that $st \in \pts$, $\lambda$ is a \tsql value
expression, $\lambda'$ is as in the translation rule, $g^{db} \in
G^{db}$, $eval(st,\lambda, g^{db}) \in D_P^*$, $\corn{\phi'} =
\tup{\tau_1, \tau_2, \tau_3, \dots, \tau_n}$, and $\Sigma =
trans(\past[\beta, \phi'], \lambda)$. From the definition of
$\corn{\dots}$, it should be easy to see that $\corn{\past[\beta,
  \phi']} =\tup{\beta, \tau_1, \dots, \tau_n}$. Finally, I assume that
theorem \ref{yn_theorem} holds for $\phi = \phi'$. I need to show
that:
\begin{enumerate}
\item $\fcn(\Sigma) \subseteq \fcn(\lambda)$
\item $eval(st, \Sigma, g^{db}) \in \vrel(n+1)$
\item $\tup{v, v_1, \dots, v_n; v_t} \in 
       eval(st, \Sigma, g^{db})$ iff for some $g \in G$: \\
   $\denot{M(st), g}{\beta} = f_D(v)$, $\denot{M(st), g}{\tau_1} =
   f_D(v_1)$, \dots, $\denot{M(st), g}{\tau_n} = f_D(v_n)$, and \\
   $\denot{M(st), st, f_D(v_t), f_D(eval(st, \lambda, g^{db})), g}{
   \past[\beta, \phi']} = T$
\end{enumerate}

The definition of $\lambda'$ implies that any free column reference in
$\lambda'$ is situated within the $\lambda$ of $\lambda'$. By lemma
\ref{fcn_lemma}, this implies \pref{pastr:2}. Also, the syntax of
\tsql and the fact that $\lambda$ is a value expression imply that
$\lambda'$ is a value expression as well.
\begin{equation}
\label{pastr:2}
\fcn(\lambda') \subseteq \fcn(\lambda)
\end{equation}
The assumption that $eval(st, \lambda, g^{db}) \in D_P^*$ and the
definition of $\lambda'$ imply \pref{pastr:10} and that $eval(st,
\lambda', g^{db}) \in D^*_P$.
\begin{equation}
\label{pastr:10}
f_D(eval(st,\lambda', g^{db})) = f_D(eval(st, \lambda, g^{db}))
 \intersect [t_{first}, st)
\end{equation}

Let $\Sigma'$ be the embedded \sql{SELECT} statement in the
\sql{FROM} clause of $\Sigma$, i.e.\ $\Sigma' = trans(\phi',
\lambda')$. From the hypothesis, $\phi' \in \ynforms$, $st \in \pts$,
$\corn{\phi'} = \tup{\tau_1, \dots, \tau_n}$, and $g^{db} \in
G^{db}$. From the discussion above, $\lambda'$ is a value expression,
and $eval(st, \lambda', g^{db}) \in D_P^*$.  Then, from theorem
\ref{yn_theorem} for $\phi = \phi'$ (according to the hypothesis,
theorem \ref{yn_theorem} holds for $\phi = \phi'$):
\begin{enumerate}
\item[$1'.$] $\fcn(\Sigma') \subseteq \fcn(\lambda')$
\item[$2'.$] $eval(st, \Sigma', g^{db}) \in \vrel(n)$
\item[$3'.$] $\tup{v_1, \dots, v_n; v_t} \in 
       eval(st, \Sigma', g^{db})$ iff for some $g' \in G$: \\
   $\denot{M(st), g'}{\tau_1} =
   f_D(v_1)$, \dots, $\denot{M(st), g'}{\tau_n} = f_D(v_n)$, and \\
   $\denot{M(st), st, f_D(v_t), f_D(eval(st, \lambda', g^{db})), g'}{
   \phi'} = T$
\end{enumerate}

\subsubsection*{Proof of clause 1}

The two \sql{VALID($\alpha$)} in the \sql{SELECT} and \sql{VALID}
clauses of $\Sigma$ are not free column references in $\Sigma$,
because $\Sigma$ is a binding context for both of them. The
$\alpha.1$, \dots, $\alpha.n$ in the \sql{SELECT} clause of $\Sigma$
are not free column references in $\Sigma$ for the same reason.
$\Sigma$ contains no other column references (and hence no other free
column references), apart from those that possibly appear within
$\Sigma'$. By lemma \ref{fcn_lemma}, this implies \pref{pastr:1}.
\pref{pastr:1}, clause $1'$, and \pref{pastr:2} imply clause 1.
\begin{equation}
\label{pastr:1}
\fcn(\Sigma) \subseteq \fcn(\Sigma')
\end{equation}

\subsubsection*{Proof of clause 2}

According to clause $2'$, $eval(st, \Sigma', g^{db}) \in
\vrel(n)$. Then, from the translation rule, it should be easy to see
that \pref{pastr:3} holds. 
\begin{equation}
\label{pastr:3} \label{r3.210}
\begin{aligned}[t]
eval(st, \Sigma, g^{db}) = 
\{& \tup{v_t, v_1, \dots, v_n; v_t} \mid \\
  & \tup{v_1, \dots, v_n; v_t} \in eval(st, \Sigma', g^{db})\}
\end{aligned}
\end{equation}
\pref{pastr:3} implies that all the time-stamps $v_t$ of $eval(st,
\Sigma, g^{db})$ are also time-stamps of $eval(st, \Sigma', g^{db})$.
This implies that $v_t \in D_P$, because $eval(st, \Sigma', g^{db})
\in \vrel(n)$. \pref{pastr:3} also implies that $eval(st, \Sigma,
g^{db})$ is a valid-time relation of $n+1$ explicit attributes.
Therefore, $eval(st, \Sigma, g^{db}) \in \vrel(n+1)$. Clause 2 has
been proven.

\subsubsection*{Proof of clause 3}

Using the definition of $\denot{st,et,lt,g}{\past[\beta,\phi']}$
(section \ref{past_op}) and the fact that $\denot{M(st),g}{\beta} =
g(\beta)$, clause 3 becomes:
\begin{eqnarray}
&& \tup{v,v_1,\dots,v_n;v_t} \in eval(st,\Sigma,g^{db}) \mbox{ iff
   for some } g: \nonumber \\
&& g \in G \label{r3.201} \label{r3.501} \\
&& g(\beta) = f_D(v) \label{r3.204} \label{r3.504}\\
&& \denot{M(st),g}{\tau_1} = f_D(v_1), \; \dots, \; \denot{M(st),g}{\tau_n} =
   f_D(v_n) \label{r3.205} \label{r3.505} \\
&& g(\beta) = f_D(v_t) \label{r3.207} \label{r3.507} \\
&& \denot{st,f_D(v_t),f_D(eval(st, \lambda, g^{db})) 
       \intersect [t_{first}, st),g}{\phi'} = T \label{r3.208} \label{r3.509}
\end{eqnarray}

I first prove the forward direction of clause 3. I assume
that $\tup{v,v_1,\dots,v_n;v_t} \in eval(st,\Sigma,g^{db})$. I need to
prove that for some $g$, \pref{r3.201} -- \pref{r3.208} hold.
The assumption that $\tup{v,v_1,\dots,v_n;v_t} \in
eval(st,\Sigma,g^{db})$ and \pref{r3.210} imply that:
\begin{eqnarray}
&& v = v_t \label{r3.211} \label{r3.511} \\
&& \tup{v_1,\dots,v_n;v_t} \in eval(st,\Sigma',g^{db}) 
   \label{r3.212} \label{r3.510}
\end{eqnarray}

\pref{r3.212} and clause $3'$ imply that for some $g'$:
\begin{eqnarray}
&& g' \in G \label{r3.213}\\
&& \denot{M(st),g'}{\tau_1} = f_D(v_1), \; \dots, \; \denot{M(st),g'}{\tau_n} =
   f_D(v_n) \label{r3.216} \\
&& \denot{M(st), st, f_D(v_t), f_D(eval(st, \lambda', g^{db})), g'}{
   \phi'} = T \label{r3.218}
\end{eqnarray}

Let $g = (g')^\beta_{f_D(v_t)}$. Clause 2 (proven above) and the
assumption that $\tup{v,v_1,\dots,v_n;v_t} \in eval(st, \Sigma,
g^{db})$ imply that $v_t \in D_P$, a fact that along with lemma
\ref{l6}, the assumption that $\beta \in \vars$, \pref{r3.213}, and
the definition of $g$ imply \pref{r3.201}. \pref{r3.207} follows from
the definition of $g$. \pref{r3.204} follows from \pref{r3.207} and
\pref{r3.211}.

\pref{r3.218} and \pref{pastr:10} imply \pref{pastr:11}
\begin{equation}
\label{pastr:11}
 \denot{M(st), st, f_D(v_t), 
        f_D(eval(st, \lambda, g^{db})) \intersect [t_{first}, st), g'}
       {\phi'} = T
\end{equation}
The syntax of \topl (section \ref{top_syntax}) and the fact that
$\past[\beta,\phi'] \in \ynforms$, imply that $\beta$ does not occur
in $\phi'$. This and the definition of $g$ imply \pref{pastr:15}.
\pref{pastr:11} and \pref{pastr:15} imply \pref{r3.208}.
\begin{equation}
\label{pastr:15}
\begin{aligned}[t]
& \denot{M(st), st, f_D(v_t),f_D(eval(st, \lambda, g^{db})) \intersect
   [t_{first}, st), g'}{\phi'} =  \\
& \denot{M(st), st, f_D(v_t),f_D(eval(st, \lambda, g^{db})) \intersect
   [t_{first}, st), (g')^\beta_{f_D(v_t)}}{\phi'} = \\
& \denot{M(st), st, f_D(v_t),f_D(eval(st, \lambda, g^{db})) \intersect
   [t_{first}, st), g}{\phi'}
\end{aligned}
\end{equation}

\pref{r3.205} follows from lemma \ref{l3}, the assumption that
$\corn{\phi'} = \tup{\tau_1, \dots, \tau_n}$, \pref{r3.216}, and the
fact that $g$ and $g'$ assign the same values to all variables,
possibly apart from $\beta$, which does not occur in $\phi'$. The
forward direction of clause 3 has been proven.

\bigskip

I now prove the backwards direction of clause 3. I assume that
\pref{r3.501} -- \pref{r3.509} hold.  I need to prove that
$\tup{v,v_1,\dots,v_n;v_t} \in eval(st,\Sigma,g^{db})$. According to
\pref{r3.210}, it is enough to prove \pref{r3.511} and \pref{r3.510}.
According to clause $3'$, in order to prove \pref{r3.510}, it is
enough to prove \pref{r3.512} -- \pref{r3.517}. \pref{r3.512} and
\pref{r3.515} are the same as \pref{r3.501} and \pref{r3.505}, which
were assumed to hold. \pref{r3.517} follows from \pref{r3.509} and
\pref{pastr:10}.
\begin{eqnarray}
&& g \in G \label{r3.512}\\
&& \denot{M(st),g}{\tau_1} = f_D(v_1), \dots, \denot{M(st),g}{\tau_n} =
   f_D(v_n) \label{r3.515} \\
&& \denot{M(st), st, f_D(v_t), f_D(eval(st, \lambda', g^{db})), g}{
   \phi'} = T \label{r3.517}
\end{eqnarray}

It remains to prove \pref{r3.511}.  From \pref{r3.504} and
\pref{r3.507} we get $f_D(v_t) = f_D(v)$, which implies that
$\fdi(f_D(v_t)) = \fdi(f_D(v))$, i.e.\ $v = v_t$. Hence, \pref{r3.511}
holds. The backwards direction of clause 3 has been proven.


\subsection{$Perf[\beta, \phi']$} \label{perf_app_rule}

\subsubsection*{Translation rule}

If $\beta \in \vars$, $\phi' \in \ynforms$, and $\lambda$ is a \tsql
value expression, then:

$trans(\perf[\beta, \phi'], \lambda) \defeq$\\
\sql{(}\select{SELECT DISTINCT VALID($\alpha$), 
                      $\alpha$.1, $\alpha$.2, \dots, $\alpha$.$n$ \\
               VALID INTERSECT($\lambda$, 
                     PERIOD(END(VALID($\alpha$)) + INTERVAL '1' $\chi$, \\
               \ \ \ \ \ \ \ \ \ \ \ \ \ \ \ \ \ \ \ \ \ \ \ \ \ \ 
                            TIMESTAMP 'forever')) \\
               FROM $trans(\phi', \linit)$ AS $\alpha$} \\
\sql{)(SUBPERIOD)}

$\linit$ is as in section \ref{formulation}, $\chi$ is the \tsql name
of the granularity of chronons, and $n$ is the length of
$\corn{\phi'}$. Each time the translation rule is used, $\alpha$ is a
new correlation name, obtained by calling the correlation names
generator after $\lambda$ has been supplied.

\subsubsection*{Proof that theorem \ref{yn_theorem} holds for $\phi =
Perf[\beta, \phi']$, if it holds for $\phi = \phi'$}

I assume that $\beta \in \vars$ and $\phi' \in \ynforms$. By the
syntax of \topl, this implies that $\perf[\beta, \phi'] \in \ynforms$.
I also assume that $st \in \pts$, $\lambda$ is a \tsql value
expression, $\linit$ is as in section \ref{formulation}, $g^{db} \in
G^{db}$, $eval(st,\lambda, g^{db}) \in D_P^*$, $\corn{\phi'} =
\tup{\tau_1, \tau_2, \tau_3, \dots, \tau_n}$, and $\Sigma =
trans(\perf[\beta, \phi'], \lambda)$. From the definition of
$\corn{\dots}$, it should be easy to see that $\corn{\perf[\beta,
  \phi']} =\tup{\beta, \tau_1, \dots, \tau_n}$. Finally, I assume that
theorem \ref{yn_theorem} holds for $\phi = \phi'$. I need to show
that:
\begin{enumerate}
\item $\fcn(\Sigma) \subseteq \fcn(\lambda)$
\item $eval(st, \Sigma, g^{db}) \in \vrel(n+1)$
\item $\tup{v, v_1, \dots, v_n; v_t} \in 
       eval(st, \Sigma, g^{db})$ iff for some $g \in G$: \\
   $\denot{M(st), g}{\beta} = f_D(v)$, $\denot{M(st), g}{\tau_1} =
   v_1$, \dots, $\denot{M(st), g}{\tau_n} = v_n$, and \\
   $\denot{M(st), st, f_D(v_t), f_D(eval(st, \lambda, g^{db})), g}{
   \perf[\beta, \phi']} = T$
\end{enumerate}

According to the syntax of \tsql, $\linit$ is a value expression.
By lemma \ref{linit_lemma}, \pref{perfr:1} -- \pref{perfr:4} hold.
\begin{gather}
\label{perfr:1}
f_D(eval(st, \linit, g^{db})) = \pts \\
\label{perfr:2}
eval(st, \linit, g^{db}) \in D_P^* \\
\label{perfr:4}
\fcn(\linit) = \emptyset
\end{gather}

Let $\Sigma'$ be the \sql{SELECT} statement in the
\sql{FROM} clause of $\Sigma$, i.e.\ $\Sigma' = trans(\phi',
\linit)$. From the hypothesis, $\phi' \in \ynforms$, $st \in \pts$,
$\corn{\phi'} = \tup{\tau_1, \dots, \tau_n}$, and $g^{db} \in
G^{db}$. From the discussion above, $\linit$ is a
\tsql value expression, and $eval(st, \linit, g^{db}) \in D_P^*$.  Then, from
theorem \ref{yn_theorem} for $\phi = \phi'$ (according to the
hypothesis, theorem \ref{yn_theorem} holds for $\phi = \phi'$), and
using \pref{perfr:1} and \pref{perfr:4}, we get:
\begin{enumerate}
\item[$1'.$] $\fcn(\Sigma') \subseteq \emptyset$
\item[$2'.$] $eval(st, \Sigma', g^{db}) \in \vrel(n)$
\item[$3'.$] $\tup{v_1, \dots, v_n; v_t'} \in 
       eval(st, \Sigma', g^{db})$ iff for some $g' \in G$: \\
   $\denot{M(st), g'}{\tau_1} =
   f_D(v_1)$, \dots, $\denot{M(st), g'}{\tau_n} = f_D(v_n)$, and \\
   $\denot{M(st), st, f_D(v_t'), \pts, g'}{\phi'} = T$
\end{enumerate}

\subsubsection*{Proof of clause 1}

The two \sql{VALID($\alpha$)} in the \sql{SELECT} and \sql{VALID}
clauses of $\Sigma$ are not free column references in $\Sigma$,
because $\Sigma$ is a binding context for both of them. The
$\alpha.1$, \dots, $\alpha.n$ in the \sql{SELECT} clause of $\Sigma$
are not free column references in $\Sigma$ for the same reason.
$\Sigma$ contains no other column references (and hence no other free
column references), apart from those that possibly appear within
$\Sigma'$ or the $\lambda$ in the \sql{VALID} clause of $\Sigma$. By
lemma \ref{fcn_lemma}, this implies \pref{perfr:3}.  Clause $1'$ and
\pref{perfr:3} imply clause 1.
\begin{equation}
\label{perfr:3}
\fcn(\Sigma) \subseteq \fcn(\Sigma') \union \fcn(\lambda)
\end{equation}

\subsubsection*{Proof of clause 2}

When computing $eval(st, \Sigma, g^{db})$, the $\alpha$ of $\Sigma$
ranges over the tuples of $eval(st, \Sigma', g^{db})$. By clause $2'$,
this implies that $\alpha$ ranges over tuples of the form $\tup{v_1,
  \dots, v_n; v_t'}$, where $v_t' \in D_P$. $\alpha$ is generated by
calling the correlation names generator after $\lambda$ has been
supplied. Hence, $\alpha$ cannot appear in $\lambda$. The fact that
$\alpha$ does not appear in $\lambda$ means that for every tuple
$\tup{v_1,\dots, v_n; v_t'}$:
\begin{equation}
\label{perfr:5}
eval(st, \lambda, (g^{db})^\alpha_{\tup{v_1, \dots, v_n;v_t'}}) =
eval(st, \lambda, g^{db})
\end{equation}

The reader should now be able to see that \pref{r7.2} holds. Intuitively,
$\tup{v_1, \dots, v_n; v_t'}$ is the tuple to which $\alpha$ refers. 
For each $\alpha$-tuple $\tup{v_1, \dots, v_n; v_t'}$, the
\sql{SELECT} and \sql{VALID} clauses of $\Sigma$ generate a new tuple $\tup{v,
v_1, \dots, v_n; v_t''}$, where $v = v_t'$ (the old time-stamp), and
$v_t''$ (the new time-stamp) represents the intersection of $eval(st,
\lambda, g^{db})$ with the period (or empty set) that begins
immediately after the end of $f_D(v_t')$ and ends at the end of the
time-axis. The restriction $f_D(v_t'') \not= \emptyset$ is needed to
capture the fact that when evaluating $\Sigma$, new tuples of the form
$\tup{v, v_1,\dots, v_n; v_t''}$ are automatically removed from the
resulting relation if $f_D(v_t'') = \emptyset$. (The time-stamps
of valid-time relations must represent temporal elements, which are
non-empty sets of chronons; see section \ref{bcdm}.)
\begin{eqnarray}
\label{r7.2}
&& eval(st,\Sigma,g^{db}) = subperiod(\{\tup{v,v_1,\dots,v_n;v_t''}
\mid \text{ for some } v_t', \\
&& \tup{v_1,\dots,v_n;v_t'} \in eval(st,\Sigma',g^{db}), 
\nonumber \\
&& f_D(v_t'') = f_D(eval(st,\lambda,g^{db}))
\intersect (maxpt(f_D(v_t')),t_{last}], 
\nonumber \\
&& f_D(v_t'') \not= \emptyset, \mbox{ and }  
v = v'_t \}) 
\nonumber
\end{eqnarray}
In \pref{r7.2}, I should have used $eval(st, \lambda,
(g^{db})^\alpha_{\tup{v_1,\dots, v_n;v_t'}})$ instead of $eval(st,
\lambda, g^{db})$, to capture the fact that if there is any free
column reference of $\alpha$ in $\lambda$, this has to be taken to
refer to the $\tup{v_1, \dots, v_n; v_t'}$ tuple to which $\alpha$
refers. By \pref{perfr:5}, however,  $eval(st, \lambda,
(g^{db})^\alpha_{\tup{v_1, \dots, v_n;v_t'}})$ is the same as
$eval(st, \lambda, g^{db})$. 

Using the definition of $subperiod$ (section \ref{new_pus}), 
\pref{r7.2} becomes \pref{r7.3extra}. (The $f_D(v_t) \subper
f_D(v_t'')$ of \pref{r7.3extra} implies that $f_D(v_t'')$ is a period,
i.e.\ a non-empty set. Hence, the constraint $f_D(v_t'')
\not= \emptyset$ of \pref{r7.2} is not needed in \pref{r7.3extra}.)
\begin{eqnarray}
\label{r7.3extra}
&&eval(st,\Sigma,g^{db}) = \{\tup{v,v_1,\dots,v_n;v_t} \mid  
\text{ for some } v_t', v_t'',  \\
&& \tup{v_1,\dots,v_n;v_t'} \in eval(st,\Sigma',g^{db}), 
\nonumber \\
&& f_D(v_t'') = f_D(eval(st,\lambda,g^{db}))
\intersect (maxpt(f_D(v_t')),t_{last}], 
\nonumber \\
&& v = v'_t, \text{ and } f_D(v_t) \subper f_D(v_t'') \}
\nonumber
\end{eqnarray}
\pref{r7.3extra} is equivalent to \pref{r7.3}. 
\begin{eqnarray}
\label{r7.3}
&&eval(st,\Sigma,g^{db}) = \{\tup{v,v_1,\dots,v_n;v_t} \mid  
\text{ for some } v_t', \\
&& \tup{v_1,\dots,v_n;v_t'} \in eval(st,\Sigma',g^{db}), \; v = v_t',
\text{ and} \nonumber \\
&& f_D(v_t) \subper f_D(eval(st,\lambda,g^{db}))
\intersect (maxpt(f_D(v_t')),t_{last}] \}
\nonumber
\end{eqnarray}
For every tuple $\tup{v_1, \dots, v_n; v_t} \in
eval(st,\Sigma,g^{db})$, the $f_D(v_t) \subper
f_D(eval(st,\lambda,g^{db})) \intersect (maxpt(f_D(v_t')),t_{last}]$
in \pref{r7.3} implies that $f_D(v_t)$ is a period, which in turn
implies that $v_t \in D_P$. \pref{r7.3} also implies that
$eval(st,\Sigma,g^{db})$ is a valid-time relation of $n+1$ explicit
attributes. Hence, $eval(st,\Sigma,g^{db}) \in \vrel(n+1)$. Clause 2
has been proven.

\subsubsection*{Proof of clause 3}

Using the definition of $\denot{M(st),st,et,lt,g}{\perf[\beta,\phi']}$
(see section \ref{perf_op}), and the fact that $\denot{M(st),g}{\beta} =
g(\beta)$, clause 3 becomes:
\begin{eqnarray}
&& \tup{v,v_1,\dots,v_n;v_t} \in eval(st,\Sigma,g^{db}) 
   \mbox{ iff for some } g \mbox{ and } et': \nonumber \\
&& g \in G \label{r7.6} \label{r7.200} \\
&& et' \in \periods \label{r7.9} \label{r7.203} \\
&& g(\beta) = f_D(v) \label{r7.10} \label{r7.204} \\
&& \denot{M(st),g}{\tau_1} = f_D(v_1), \; \dots, \; \denot{M(st),g}{\tau_n} =
   f_D(v_n) \label{r7.11} \label{r7.205} \\
&& g(\beta) = et' \label{r7.13} \label{r7.207} \\
&& f_D(v_t) \subper f_D(eval(st, \lambda, g^{db})) \label{r7.14}
   \label{r7.208} \\
&& maxpt(et') \prec minpt(f_D(v_t)) \label{r7.pp2} \label{r7.pp21} \\
&& \denot{M(st),st,et',\pts,g}{\phi'} = T \label{r7.15} \label{r7.209} 
\end{eqnarray}

I first prove the forward direction of clause 3. I assume that
$\tup{v,v_1,\dots,v_n;v_t} \in eval(st, \Sigma, g^{db})$. I need to
prove that for some $g$ and $et'$, \pref{r7.6} -- \pref{r7.15} hold. 
The assumption that $\tup{v,v_1,\dots,v_n;v_t} \in
eval(st,\Sigma,g^{db})$, and \pref{r7.3} imply that for some $v_t'$:
\begin{eqnarray}
&& \tup{v_1,\dots,v_n;v_t'} \in eval(st,\Sigma',g^{db})
   \label{r7.17} \label{r7.210} \\
&& f_D(v_t) \subper f_D(eval(st,\lambda,g^{db}))
   \intersect (maxpt(f_D(v_t')),t_{last}] \label{r7.pp4} \label{r7.pp23}\\
&& v = v_t' \label{r7.16} \label{r7.212}
\end{eqnarray}
I set $et'$ as in \pref{perfr:10} and show that \pref{r7.6} --
\pref{r7.15} hold. 
\begin{equation}
\label{perfr:10}
et' = f_D(v_t')
\end{equation}
\pref{r7.17} and clause $3'$\/ imply that for some $g'$:
\begin{eqnarray}
&& g' \in G \label{r7.20}\\
&& \denot{M(st),g'}{\tau_1} = f_D(v_1), \dots, \denot{M(st),g'}{\tau_n} =
   f_D(v_n) \label{r7.23} \\
&& \denot{M(st),st,f_D(v_t'),\pts,g'}{\phi'} = T \label{r7.25}
\end{eqnarray}

Let $g = (g')^\beta_{f_D(v_t')}$. \pref{r7.17} and clause $2'$ imply
that $v_t' \in D_P$. By lemma \ref{l6}, \pref{r7.6} holds.
\pref{r7.13} follows from the definition of $g$ and \pref{perfr:10}.
The conclusion that $v'_t \in D_P$ implies that $f_D(v_t') \in
\periods$. This and \pref{perfr:10} imply \pref{r7.9}.  The definition
of $g$ and \pref{r7.16} imply \pref{r7.10}.

I now prove \pref{r7.15}. The syntax of \topl (section
\ref{top_syntax}) and the fact that $\perf[\beta,\phi'] \in \ynforms$,
imply that $\beta$ does not occur in $\phi'$. Along with the
definition of $g$, this implies \pref{perfr:11}. \pref{r7.25},
\pref{perfr:10}, and \pref{perfr:11} imply \pref{r7.15}.
\begin{equation}
\label{perfr:11}
\begin{aligned}[t]
& \denot{M(st), st, et', \pts, g'}{\phi'} =  \\
& \denot{M(st), st, et', \pts, (g')^\beta_{f_D(v_t')}}{\phi'} = \\
& \denot{M(st), st, et', \pts, g}{\phi'}
\end{aligned}
\end{equation}

\pref{r7.11} follows from lemma \ref{l3}, the assumption that
$\corn{\phi'} = \tup{\tau_1, \dots, \tau_n}$, \pref{r7.23}, and the
fact that $g$ and $g'$ assign the same values to all variables,
possibly apart from $\beta$, which does not occur in $\phi'$.

I now prove \pref{r7.14}. \pref{r7.pp4} and the definition of
$\subper$ imply \pref{r7.pp5} and \pref{perfr:12.1}. \pref{perfr:12.1}
in turn implies \pref{perfr:12}.
\begin{eqnarray}
&& f_D(v_t) \in \periods \label{r7.pp5} \\
&& f_D(v_t) \subseteq f_D(eval(st,\lambda,g^{db})) \intersect
(maxpt(f_D(v_t')),t_{last}] \label{perfr:12.1} \\
&& f_D(v_t) \subseteq f_D(eval(st,\lambda,g^{db})) \label{perfr:12}
\end{eqnarray}
From the hypothesis, $eval(st,\lambda,g^{db}) \in D_P^*$, which
implies that $f_D(eval(st,\lambda,g^{db}))$ is the empty set or a
period. $f_D(eval(st,\lambda,g^{db}))$ cannot be the empty set,
because according to \pref{perfr:12}, it has $f_D(v_t)$ as its subset,
and by \pref{r7.pp5} $f_D(v_t)$ is a period, i.e.\ a non-empty set.
Hence, \pref{perfr:13} holds. \pref{perfr:12}, \pref{r7.pp5}, and
\pref{perfr:13} imply \pref{r7.14}.
\begin{equation}
\label{perfr:13}
f_D(eval(st,\lambda,g^{db})) \in \periods
\end{equation}

It remains to prove \pref{r7.pp2}. \pref{perfr:12.1} implies
\pref{perfr:20}. $(maxpt(f_D(v_t')),t_{last}]$ is either a period or
the empty set (if $maxpt(f_D(v_t')) = t_{last}$). It cannot be the
empty set, because according to \pref{perfr:20},
$(maxpt(f_D(v_t')),t_{last}]$ has $f_D(v_t)$ as its subset, and by
\pref{r7.pp5} $f_D(v_t)$ is a period, i.e.\ a non-empty set. Hence,
\pref{perfr:21} holds. \pref{perfr:20}, \pref{r7.pp5}, and
\pref{perfr:21} imply \pref{perfr:22}. \pref{r7.pp2} follows from
\pref{perfr:22} and \pref{perfr:10}. The forward direction of
clause 3 has been proven.
\begin{eqnarray}
&& f_D(v_t) \subseteq (maxpt(f_D(v_t')),t_{last}] \label{perfr:20} \\
&& (maxpt(f_D(v_t')),t_{last}] \in \periods \label{perfr:21} \\
&& f_D(v_t) \subper (maxpt(f_D(v_t')),t_{last}] \label{perfr:22}
\end{eqnarray}

\bigskip

I now prove the backwards direction of clause 3. I assume that
\pref{r7.200} -- \pref{r7.209} hold.  I need to prove that
$\tup{v,v_1,\dots,v_n;v_t} \in eval(st,\Sigma,g^{db})$. According to
\pref{r7.3}, it is enough to prove that for some $v'_t$, \pref{r7.210}
-- \pref{r7.212} hold.  I set $v_t'$ to $v$ as required by
\pref{r7.212}, and prove \pref{r7.210} and \pref{r7.pp23}.  According
to clause $3'$, in order to prove \pref{r7.210}, it is enough to prove
that:
\begin{eqnarray}
&& g \in G \label{r7.220}\\
&& \denot{M(st),g}{\tau_1} = f_D(v_1), \dots, \denot{M(st),g}{\tau_n} =
   f_D(v_n) \label{r7.223} \\
&& \denot{M(st), st, f_D(v_t'), \pts, g}{\phi'} = T \label{r7.225}
\end{eqnarray}
\pref{r7.220} and \pref{r7.223} are the same as \pref{r7.200} and
\pref{r7.205}, which were assumed to hold. \pref{r7.212} (that holds),
\pref{r7.204} and \pref{r7.207} imply \pref{perfr:50}. \pref{perfr:50}
and \pref{r7.209} imply \pref{r7.225}. Hence, \pref{r7.210} holds.
\begin{equation}
\label{perfr:50}
f_D(v_t') = et'
\end{equation}
It remains to prove \pref{r7.pp23}. \pref{r7.208} implies
\pref{perfr:51} and \pref{perfr:51.1}.
\begin{gather}
\label{perfr:51}
f_D(v_t) \in \periods \\
\label{perfr:51.1}
f_D(eval(st, \lambda, g^{db})) \in \periods 
\end{gather}
\pref{perfr:51}, \pref{r7.203}, and \pref{r7.pp21} imply
\pref{perfr:52.1} and \pref{perfr:53}. \pref{r7.208} implies
\pref{perfr:54}. \pref{perfr:53} and \pref{perfr:54} imply
\pref{perfr:55}
\begin{gather}
\label{perfr:52.1}
(maxpt(et'), t_{last}] \in \periods \\
\label{perfr:53}
f_D(v_t) \subseteq f_D(eval(st, \lambda, g^{db})) \\
\label{perfr:54}
f_D(v_t) \subseteq (maxpt(et'), t_{last}] \\
\label{perfr:55}
f_D(v_t) \subseteq f_D(eval(st, \lambda, g^{db})) \intersect
(maxpt(et'), t_{last}] 
\end{gather}
According to \pref{perfr:51.1} and \pref{perfr:52.1}, $f_D(eval(st,
\lambda, g^{db})) \intersect (maxpt(et'), t_{last}]$ is the
intersection of two periods. The intersection of two periods is the
empty set if the periods do not overlap, or a period (the overlap of
the two periods) if the periods overlap. $f_D(eval(st, \lambda,
g^{db}))$ and $(maxpt(et'), t_{last}]$ do overlap, because according
to \pref{perfr:55}, their intersection has $f_D(v_t)$ as its subset,
and by \pref{perfr:51} $f_D(v_t)$ is a period, i.e.\ a non-empty set.
Hence, $f_D(eval(st,\lambda, g^{db})) \intersect (maxpt(et'),
t_{last}]$ is a period.  This conclusion, \pref{perfr:51},
\pref{perfr:55}, and \pref{perfr:50} imply \pref{r7.pp23}. The
backwards direction of clause 3 has been proven.


\subsection{$Ntense[\beta, \phi']$} \label{ntbeta_rule}

\subsubsection*{Translation rule}

If $\beta \in \vars$, $\phi' \in \ynforms$, and $\lambda$ is a \tsql
value expression, then:

$trans(\ntense[\beta, \phi'], \lambda) \defeq$\\
\sql{(}\select{SELECT DISTINCT VALID($\alpha$), 
                      $\alpha$.1, $\alpha$.2, \dots, $\alpha$.$n$ \\
               VALID PERIOD(TIMESTAMP 'beginning', TIMESTAMP 'forever') \\
               FROM $trans(\phi', \linit)$ AS $\alpha$}\\
\sql{)(SUBPERIOD)}

$\linit$ is as in section \ref{formulation}, and $n$ is the length of
$\corn{\phi'}$. Each time the translation rule is used, $\alpha$ is a
new correlation name, obtained by calling the correlation
names generator.

\subsubsection*{Proof that theorem \ref{yn_theorem} holds for $\phi =
Ntense[\beta, \phi']$, if it holds for $\phi = \phi'$}

The proof is very similar to that of section \ref{perf_app_rule}. 


\subsection{$Ntense[now^*, \phi']$}

\subsubsection*{Translation rule}

If $\phi' \in \ynforms$ and $\lambda$ is a \tsql
value expression, then:

$trans(\ntense[now^*, \phi'], \lambda) \defeq$\\
\sql{(}\select{SELECT DISTINCT $\alpha$.1, $\alpha$.2, \dots, $\alpha$.$n$ \\
               VALID PERIOD(TIMESTAMP 'beginning', TIMESTAMP 'forever') \\
               FROM $trans(\phi', \linit)$ AS $\alpha$ \\
               WHERE VALID($\alpha$) =
                      PERIOD(TIMESTAMP 'now', TIMESTAMP 'now')}\\
\sql{)(SUBPERIOD)}

$\linit$ is as in section \ref{formulation}, and $n$ is the length of
$\corn{\phi'}$. Each time the translation rule is used, $\alpha$ is a
new correlation name, obtained by calling the correlation
names generator.

\subsubsection*{Proof that theorem \ref{yn_theorem} holds for $\phi =
Ntense[now^*, \phi']$, if it holds for $\phi = \phi'$}

The proof is very similar to that of section \ref{perf_app_rule}.


\subsection{$For[\sigma_c, \nu_{qty}, \phi']$}

\subsubsection*{Translation rule}

If $\sigma_c \in \cparts$, $\nu_{qty} \in \{1,2,3,\dots\}$, $\phi' \in
\ynforms$, and $\lambda$ is a \tsql value expression, then:

$trans(\for[\sigma_c, \nu_{qty}, \phi'], \lambda) \defeq$\\
\sql{(}\select{SELECT DISTINCT $\alpha$.1, $\alpha$.2, \dots, $\alpha$.$n$ \\
               VALID VALID($\alpha$) \\
               FROM $trans(\phi', \lambda)$ AS $\alpha$ \\
               WHERE INTERVAL(VALID($\alpha$), $\gamma$) =
                     INTERVAL '$\nu_{qty}$' $\gamma$)}

$n$ is the length of $\corn{\phi'}$, and $\gamma$ is the first element
of the pair $\hcpartsp(\sigma_c) = \tup{\gamma, \Sigma_c}$ (section
\ref{via_TSQL2}). Each time the translation rule is used, $\alpha$ is
a new correlation name, obtained by calling the correlation names
generator.

\subsubsection*{Proof that theorem \ref{yn_theorem} holds for $\phi =
For[\sigma_c, \nu_{qty}, \phi']$, if it holds for $\phi = \phi'$}

I assume that $\sigma_c \in \cparts$, $\nu_{qty} \in \{1,2,3,\dots\}$,
and $\phi' \in \ynforms$. By the syntax of \topl, this implies that
$\for[\sigma_c,\nu_{qty},\phi'] \in \ynforms$. I also assume that $st
\in \pts$, $\lambda$ is a \tsql value expression, $g^{db} \in G^{db}$,
$eval(st,\lambda, g^{db}) \in D_P^*$, $\corn{\phi'} = \tup{\tau_1,
  \tau_2, \tau_3, \dots, \tau_n}$, $\hcpartsp(\sigma_c) = \tup{\gamma,
  \Sigma_c}$ (as in the translation rule), and $\Sigma =
trans(\for[\sigma_c, \nu_{qty}, \phi'], \lambda)$. From the definition
of $\corn{\dots}$, it should be easy to see that $\corn{\for[\sigma_c,
  \nu_{qty}, \phi']} = \corn{\phi'} = \tup{\tau_1, \dots, \tau_n}$.
Finally, I assume that theorem \ref{yn_theorem} holds for $\phi =
\phi'$. I need to show that:
\begin{enumerate}
\item $\fcn(\Sigma) \subseteq \fcn(\lambda)$
\item $eval(st, \Sigma, g^{db}) \in \vrel(n)$
\item $\tup{v_1, \dots, v_n; v_t} \in 
       eval(st, \Sigma, g^{db})$ iff for some $g \in G$: \\
   $\denot{M(st), g}{\tau_1} = f_D(v_1)$, \dots, $\denot{M(st), g}{\tau_n}
   = f_D(v_n)$, and \\ $\denot{M(st), st, f_D(v_t), f_D(eval(st, \lambda,
   g^{db})), g}{\for[\sigma_c, \nu_{qty}, \phi']} = T$
\end{enumerate}

Let $\Sigma'$ be the embedded \sql{SELECT} statement in the
\sql{FROM} clause of $\Sigma$, i.e.\ $\Sigma' = trans(\phi',
\lambda)$. From the hypothesis, $\phi' \in \ynforms$, $st \in \pts$,
$\corn{\phi'} = \tup{\tau_1, \dots, \tau_n}$, $\lambda$ is a value
expression, $g^{db} \in G^{db}$, and $eval(st, \lambda,
g^{db}) \in D_P^*$. Then, from theorem \ref{yn_theorem} for $\phi =
\phi'$ (according to the hypothesis, theorem \ref{yn_theorem} holds
for $\phi = \phi'$) we get:
\begin{enumerate}
\item[$1'.$] $\fcn(\Sigma') \subseteq \fcn(\lambda)$
\item[$2'.$] $eval(st, \Sigma', g^{db}) \in \vrel(n)$
\item[$3'.$] $\tup{v_1, \dots, v_n; v_t} \in 
       eval(st, \Sigma', g^{db})$ iff for some $g \in G$: \\
   $\denot{M(st), g}{\tau_1} =
   f_D(v_1)$, \dots, $\denot{M(st), g}{\tau_n} = f_D(v_n)$, and \\
   $\denot{M(st), st, f_D(v_t), f_D(eval(st, \lambda, g^{db})), g}{
   \phi'} = T$
\end{enumerate}

\subsubsection*{Proof of clause 1}

The two \sql{VALID($\alpha$)} in the \sql{VALID} and \sql{WHERE}
clauses of $\Sigma$ are not free column references in $\Sigma$,
because $\Sigma$ is a binding context for both of them. The
$\alpha.1$, \dots, $\alpha.n$ in the \sql{SELECT} clause of $\Sigma$
are not free column references in $\Sigma$ for the same reason.
$\Sigma$ contains no other column references (and hence no other free
column references), apart from those that possibly appear within
$\Sigma'$. By lemma \ref{fcn_lemma}, this implies \pref{forr:1}.
\pref{forr:1} and clause $1'$ imply clause 1.
\begin{equation}
\label{forr:1}
\fcn(\Sigma) \subseteq \fcn(\Sigma')
\end{equation}

\subsubsection*{Proof of clause 2}

When computing $eval(st, \Sigma, g^{db})$, the $\alpha$ of $\Sigma$
ranges over the tuples of $eval(st, \Sigma', g^{db})$. By clause $2'$,
$eval(st,\Sigma', g^{db}) \in \vrel(n)$. Hence, $\alpha$ ranges over
tuples of the form $\tup{v_1, \dots, v_n; v_t}$, where $f_D(v_t) \in
\periods$. The \sql{VALID($\alpha$)} in the \sql{WHERE} clause of
$\Sigma$ stands for $f_D(v_t)$. The arrangements of sections
\ref{resulting_model} and \ref{via_TSQL2} imply that $\gamma$ is the
name of the \tsql granularity that corresponds to the complete
partitioning $\fcparts(st)(\sigma_c)$. Then, from the semantics of the
\sql{INTERVAL} function (section \ref{interv_fun}), it should be easy
to see that the constraint \sql{INTERVAL(VALID($\alpha$), $\gamma$) =
  INTERVAL '$\nu_{qty}$' $\gamma$} in the \sql{WHERE} clause of
$\Sigma$ is satisfied iff $f_D(v_t)$ covers exactly $\nu_{qty}$
consecutive periods from $\fcparts(st)(\sigma_c)$. It should now be
easy to see from the translation rule that \pref{r11.3} holds.
\begin{eqnarray}
\label{r11.3}
&&  eval(st,\Sigma,g^{db}) = 
    \{\tup{v_1,\dots,v_n;v_t} \in eval(st,\Sigma',g^{db}) \mid \\
&& \mbox{for some } p_1, p_2, \dots, p_{\nu_{qty}} \in \fcparts(st)(\sigma_c):
\nonumber \\
&& \;\; minpt(p_1) = minpt(f_D(v_t)), \; next(maxpt(p_1)) = minpt(p_2),
\nonumber \\
&& \;\; next(maxpt(p_2)) = minpt(p_3), \dots, 
   next(maxpt(p_{\nu_{qty}-1})) = minpt(p_{\nu_{qty}}),
\nonumber \\
&& \;\; \mbox{and } maxpt(p_{\nu_{qty}}) = maxpt(f_D(v_t)) \}
\nonumber
\end{eqnarray}
Clause $2'$ and \pref{r11.3} imply that for every
$\tup{v_1,\dots,v_n;v_t} \in eval(st, \Sigma, g^{db})$, $v_t \in D_P$.
\pref{r11.3} also implies that $eval(st, \Sigma, g^{db})$ is a
valid-time relation of $n$ explicit attributes. Hence, $eval(st,
\Sigma, g^{db}) \in \vrel(n)$. Clause 2 has been proven.

\subsubsection*{Proof of clause 3}

Using the definition of
$\denot{st,et,lt,g}{\for[\sigma_c,\nu_{qty},\phi']}$ (section
\ref{for_op}), clause 3 becomes:
\begin{eqnarray}
&& \tup{v_1,\dots,v_n;v_t} \in eval(st,\Sigma,g^{db}) 
   \mbox{ iff for some } g \mbox{ and } p_1, p_2, \dots, p_{\nu_{qty}}:
\nonumber \\
&& g \in G \label{r11.4}\\
&& \denot{M(st),g}{\tau_1} = f_D(v_1), \dots, \denot{M(st),g}{\tau_n} =
   f_D(v_n) \label{r11.6} \\
&& \denot{M(st),st,f_D(v_t),f_D(eval(st,\lambda,g^{db})),g}{\phi'} = T
 \label{r11.9} \\
&& p_1, \dots, p_{\nu_{qty}} \in \fcparts(st)(\sigma_c) \label{r11.9.5} \\
&& minpt(p_1) = minpt(f_D(v_t)), \; next(maxpt(p_1)) = minpt(p_2),
\label{r11.8} \\
&& next(maxpt(p_2)) = minpt(p_3),  \dots, next(maxpt(p_{\nu_{qty}-1})) =
   minpt(p_{\nu_{qty}}),
\nonumber \\
&& \mbox{and } maxpt(p_{\nu_{qty}}) = maxpt(f_D(v_t)) 
\nonumber
\end{eqnarray}

I first prove that the forward direction of clause 3 holds. I assume that
$\tup{v_1,\dots,v_n;v_t} \in eval(st, \Sigma,g^{db})$. I need to prove
that for some $g$, $p_1$, \dots, $p_{\nu_{qty}}$, \pref{r11.4} --
\pref{r11.8} hold.
The assumption that $\tup{v_1,\dots,v_n;v_t} \in
eval(st,\Sigma,g^{db})$, and \pref{r11.3} imply that for some $p_1$,
\dots, $p_{\nu_{qty}}$:
\begin{eqnarray}
&& \tup{v_1,\dots,v_n;v_t} \in eval(st,\Sigma',g^{db}) \label{r11.10} \\
&& p_1, \dots, p_{\nu_{qty}} \in \fcparts(st)(\sigma_c) \label{r11.9.1} \\
&& minpt(p_1) = minpt(f_D(v_t)), \; next(maxpt(p_1)) = minpt(p_2),
\label{r11.11} \\
&& next(maxpt(p_2)) = minpt(p_3),  \dots, next(maxpt(p_{\nu_{qty}-1})) =
   minpt(p_{\nu_{qty}}),
\nonumber \\
&& \mbox{and } maxpt(p_{\nu_{qty}}) = maxpt(f_D(v_t)) 
\nonumber
\end{eqnarray}
\pref{r11.10} and clause $3'$ imply that for some $g$:
\begin{eqnarray}
&& g \in G \label{r11.12}\\
&& \denot{M(st),g}{\tau_1} = f_D(v_1), \dots, \denot{M(st),g}{\tau_n} =
   f_D(v_n) \label{r11.14} \\
&& \denot{M(st),st,f_D(v_t),f_D(eval(st, \lambda, g^{db})),g}{\phi'} = T
 \label{r11.16}
\end{eqnarray}
\pref{r11.4}, \pref{r11.6}, \pref{r11.9}, \pref{r11.9.5}, and
\pref{r11.8}, are the same as \pref{r11.12}, \pref{r11.14}, \pref{r11.16},
\pref{r11.9.1} and \pref{r11.11} respectively, which are known to be
true. The forward direction of clause 3 has been proven

\bigskip

I now prove the backwards direction of clause 3. I assume that
\pref{r11.4} -- \pref{r11.8} hold. I need to prove that
$\tup{v_1,\dots,v_n;v_t} \in eval(st,\Sigma,g^{db})$. According to
\pref{r11.3}, it is enough to prove that \pref{r11.10} --
\pref{r11.11} hold. \pref{r11.9.1} and \pref{r11.11} are the same as
\pref{r11.9.5} and \pref{r11.8} respectively, which were assumed to
hold. \pref{r11.10} follows from clause $3'$ and \pref{r11.4},
\pref{r11.6}, and \pref{r11.9}. The backwards direction of clause 3
has been proven.


\subsection{$Begin[\phi']$} \label{beg_rule}

\subsubsection*{Translation rule}

If $\phi' \in \ynforms$ and $\lambda$ is a \tsql value expression, then:

$trans(\lbegin[\phi'], \lambda) \defeq$\\
\sql{(}\select{SELECT DISTINCT $\alpha$.1, $\alpha$.2, \dots, $\alpha$.$n$ \\
               VALID PERIOD(BEGIN(VALID($\alpha$)), BEGIN(VALID($\alpha$))) \\
               FROM $trans(\phi', \linit)$(NOSUBPERIOD) AS $\alpha$ \\
               WHERE $\lambda$ CONTAINS BEGIN(VALID($\alpha$)))}

$n$ is the length of $\corn{\phi'}$, and $\linit$ is as in section
\ref{formulation}. Each time the translation rule is
used, $\alpha$ is a new correlation name, obtained by calling the
correlation names generator after $\lambda$ has been supplied.

\subsubsection*{Proof that theorem \ref{yn_theorem} holds for $\phi =
Begin[\phi']$, if it holds for $\phi = \phi'$}

I assume that $\phi' \in \ynforms$. By the syntax of \topl, this
implies that $\lbegin[\phi'] \in \ynforms$. I also assume that $st \in
\pts$, $\lambda$ is a \tsql value expression, $\linit$ is as in
section \ref{formulation}, $g^{db} \in G^{db}$, $eval(st,\lambda,
g^{db}) \in D_P^*$, $\corn{\phi'} = \tup{\tau_1, \tau_2, \tau_3,
  \dots, \tau_n}$, and $\Sigma = trans(\lbegin[\phi'], \lambda)$. From
the definition of $\corn{\dots}$, it should be easy to see that
$\corn{\lbegin[\phi']} = \corn{\phi'} = \tup{\tau_1, \dots, \tau_n}$.
Finally, I assume that theorem \ref{yn_theorem} holds for $\phi =
\phi'$. I need to show that:
\begin{enumerate}
\item $\fcn(\Sigma) \subseteq \fcn(\lambda)$
\item $eval(st, \Sigma, g^{db}) \in \vrel(n)$
\item $\tup{v_1, \dots, v_n; v_t} \in 
       eval(st, \Sigma, g^{db})$ iff for some $g \in G$: \\
   $\denot{M(st), g}{\tau_1} = f_D(v_1)$, \dots, 
   $\denot{M(st), g}{\tau_n} = f_D(v_n)$, and \\
   $\denot{M(st), st, f_D(v_t), f_D(eval(st, \lambda, g^{db})), g}{
   \lbegin[\phi']} = T$
\end{enumerate}

According to the syntax of \tsql, $\linit$ is a value expression.
By lemma \ref{linit_lemma}, \pref{begr:1} -- \pref{begr:4} hold. 
\begin{gather}
\label{begr:1}
f_D(eval(st, \linit, g^{db})) = \pts \\
\label{begr:2}
eval(st, \linit, g^{db}) \in D_P^* \\
\label{begr:4}
\fcn(\linit) = \emptyset
\end{gather}

Let $\Sigma'$ be the \sql{SELECT} statement in the \sql{FROM} clause
of $\Sigma$, i.e.\ $\Sigma' = trans(\phi', \linit)$. From the
hypothesis, $\phi' \in \ynforms$, $st \in \pts$, $\corn{\phi'} =
\tup{\tau_1, \dots, \tau_n}$, and $g^{db} \in G^{db}$. From the
discussion above, $\linit$ is a \tsql value expression, and $eval(st,
\linit, g^{db}) \in D_P^*$.  Then, from theorem \ref{yn_theorem} for
$\phi = \phi'$ (according to the hypothesis, theorem \ref{yn_theorem}
holds for $\phi = \phi'$), and using \pref{begr:1} and \pref{begr:4},
we get:
\begin{enumerate}
\item[$1'.$] $\fcn(\Sigma') \subseteq \emptyset$
\item[$2'.$] $eval(st, \Sigma', g^{db}) \in \vrel(n)$
\item[$3'.$] $\tup{v_1, \dots, v_n; v_t'} \in 
       eval(st, \Sigma', g^{db})$ iff for some $g \in G$: \\
   $\denot{M(st), g}{\tau_1} =
   f_D(v_1)$, \dots, $\denot{M(st), g}{\tau_n} = f_D(v_n)$, and \\
   $\denot{M(st), st, f_D(v_t'), \pts, g}{\phi'} = T$
\end{enumerate}

\subsubsection*{Proof of clause 1}

The three \sql{VALID($\alpha$)} in the \sql{VALID} and \sql{WHERE}
clauses of $\Sigma$ are not free column references in $\Sigma$,
because $\Sigma$ is a binding context for all of them. The $\alpha.1$,
\dots, $\alpha.n$ in the \sql{SELECT} clause of $\Sigma$ are not free
column references in $\Sigma$ for the same reason. $\Sigma$ contains
no other column references (and hence no other free column
references), apart from those that possibly appear within $\Sigma'$ or
the $\lambda$ in the \sql{WHERE} clause of $\Sigma$. By lemma
\ref{fcn_lemma}, this implies \pref{begr:3}. Clause $1'$ and
\pref{begr:3} imply clause 1.
\begin{equation}
\label{begr:3}
\fcn(\Sigma) \subseteq \fcn(\Sigma') \union \fcn(\lambda)
\end{equation}

\subsubsection*{Proof of clause 2}

When computing $eval(st, \Sigma, g^{db})$, the $\alpha$ of $\Sigma$
ranges over the tuples of the relation $nosubperiod(eval(st, \Sigma',
g^{db}))$. By clause $2'$, $eval(st, \Sigma', g^{db}) \in \vrel(n)$.
From the definition of $nosubperiod$ (section \ref{new_pus}), it
follows that $nosubperiod(eval(st, \Sigma', g^{db}))$ is
also a valid-time relation of $n$ explicit attributes, and that its
time-stamps are also time-stamps of $eval(st, \Sigma', g^{db})$, i.e.\ 
elements of $D_P$.  Hence, $nosubperiod(eval(st, \Sigma', g^{db})) \in
\vrel(n)$.

$\alpha$ is generated by calling the correlation names generator after
$\lambda$ has been supplied. Hence, $\alpha$ cannot appear in
$\lambda$. The fact that $\alpha$ does not appear in $\lambda$ means
that for every tuple $\tup{v_1,\dots, v_n; v_t'}$:
\begin{equation}
\label{begr:5}
eval(st, \lambda, (g^{db})^\alpha_{\tup{v_1, \dots, v_n;v_t'}}) =
eval(st, \lambda, g^{db})
\end{equation}

The reader should now be able to see from the translation rule that
\pref{r13.5} holds. Intuitively, $\tup{v_1, \dots, v_n; v_t'}$ is the
tuple of $nosubperiod(eval(st, \Sigma', g^{db}))$ to which $\alpha$
refers. I should have used $eval(st, \lambda, (g^{db})^{\alpha}_{\tup{v_1,
\dots, v_n;v_t'}})$ instead of $eval(st, \lambda, g^{db})$, to capture the 
fact that if there are any free column references of $\alpha$ in
$\lambda$, these have to be taken to refer to the $\tup{v_1, \dots,
v_n; v_t'}$ tuple to which $\alpha$ refers. By \pref{begr:5}, however,
$eval(st, \lambda, (g^{db})^{\alpha}_{\tup{v_1,\dots, v_n;v_t'}})$ is
the same as $eval(st, \lambda, g^{db})$. 
\begin{eqnarray}
eval(st,\Sigma,g^{db}) &=& \{\tup{v_1,\dots,v_n;v_t} \mid 
\text{ for some } v_t', 
\label{r13.5} \\
&& \tup{v_1,\dots,v_n;v_t'} \in nosubperiod(eval(st,\Sigma',g^{db})),
\nonumber \\
&& f_D(v_t) = \{minpt(f_D(v_t'))\}, \mbox{ and}
\nonumber \\
&& minpt(f_D(v_t')) \in f_D(eval(st, \lambda, g^{db})) \}
\nonumber 
\end{eqnarray}
Using the definition of $nosubperiod$ (section \ref{new_pus}),
\pref{r13.5} becomes \pref{r13.5.5}.
\begin{eqnarray}
eval(st,\Sigma,g^{db}) &=& \{\tup{v_1,\dots,v_n;v_t} \mid 
\text{for some } v_t', 
\label{r13.5.5} \\
&& \tup{v_1,\dots,v_n;v_t'} \in eval(st,\Sigma',g^{db}),
\nonumber \\
&& \mbox{there is no } \tup{v_1,\dots,v_n;v_t''} \in eval(st,\Sigma',g^{db})
\nonumber \\
&& \;\;\;\; \mbox{such that } f_D(v_t') \propsubper f_D(v_t''), 
\nonumber \\
&& f_D(v_t) = \{minpt(f_D(v_t'))\}, \mbox{ and}
\nonumber \\
&& minpt(f_D(v_t')) \in f_D(eval(st, \lambda, g^{db})) \}
\nonumber 
\end{eqnarray}
\pref{r13.5.5} implies that every time-stamp $v_t$ of $eval(st,
\Sigma, g^{db})$ represents an (instantaneous) period that contains
only the earliest chronon of a time-stamp of $eval(st, \Sigma',
g^{db})$. This implies that $v_t \in D_P$. \pref{r13.5.5} also implies
that $eval(st, \Sigma, g^{db})$ is a valid-time relation of $n$
explicit attributes. Hence, $eval(st, \Sigma, g^{db}) \in \vrel(n)$,
and clause 2 has been proven.

\subsubsection*{Proof of clause 3}

Using the definition of $\denot{M(st),st,et,lt,g}{\lbegin[\phi']}$
(section \ref{begin_end_op}), clause 3 becomes:\footnote{The reader is
  reminded that the $mxlpers$ symbol is overloaded. When $l \in
  \telems$, $mxlpers(l)$ is the set of the maximal periods of a
  \emph{temporal element}, and it is defined as in section
  \ref{tsql2_time}. When $S$ is a set of periods, $mxlpers(S)$ is the
  set of the maximal periods of a \emph{set of periods}, and it is
  defined as in section \ref{temporal_ontology}.}
\begin{eqnarray}
&& \tup{v_1,\dots,v_n;v_t} \in eval(st,\Sigma,g^{db}) 
\mbox{ iff for some } g \mbox{ and } et':
\nonumber \\
&& g \in G \label{r13.6} \label{r13.b.6} \\
&& \denot{M(st),g}{\tau_1} = f_D(v_1), \dots, \denot{M(st),g}{\tau_n} =
   f_D(v_n) \label{r13.8} \label{r13.b.8} \\
&& f_D(v_t) \subper f_D(eval(st, \lambda, g^{db})) \label{r13.10}
   \label{r13.b.10} \\
&& f_D(v_t) = \{minpt(et')\} \label{r13.11.3} \label{r13.b.11.3} \\
&& et' \in
   mxlpers(\{e \in \periods \mid \denot{M(st),st,e,\pts,g}{\phi} = T\}) 
   \label{r13.12} \label{r13.b.12}
\end{eqnarray}

I first prove that the forward direction of clause 3 holds. I assume that
$\tup{v_1,\dots,v_n;v_t} \in eval(st, \Sigma,g^{db})$. I need to prove that
for some $g$ and $et'$, \pref{r13.6} -- \pref{r13.12} hold. 
The assumption that $\tup{v_1,\dots,v_n;v_t} \in
eval(st,\Sigma,g^{db})$, and \pref{r13.5.5} imply that for some $v_t'$:
\begin{eqnarray}
&& \tup{v_1,\dots,v_n;v_t'} \in eval(st,\Sigma',g^{db}) 
   \label{r13.13} \label{r13.b.13} \\
&& \mbox{there is no } \tup{v_1,\dots,v_n;v_t''} \in eval(st,\Sigma',g^{db})
   \label{r13.14} \label{r13.b.14} \\
&& \;\;\;\; \mbox{such that } f_D(v_t') \propsubper
   f_D(v_t'') \nonumber \\
&& f_D(v_t) = \{minpt(f_D(v_t'))\} \label{r13.14.1} \label{r13.b.14.1} \\
&& minpt(f_D(v_t')) \in f_D(eval(st, \lambda, g^{db})) 
   \label{r13.15} \label{r13.b.15} 
\end{eqnarray}
\pref{r13.13} and clause $3'$ imply that for some $g$:
\begin{eqnarray}
&& g \in G \label{r13.17}\\
&& \denot{M(st),g}{\tau_1} = f_D(v_1), \dots, \denot{M(st),g}{\tau_n} =
   f_D(v_n) \label{r13.19} \\
&& \denot{M(st),st,f_D(v_t'),\pts,g}{\phi'} = T \label{r13.21}
\end{eqnarray}
I set $et'$ as in \pref{begr:50} and prove that \pref{r13.6} --
\pref{r13.12} hold. \pref{r13.6} and \pref{r13.8} are the same as
\pref{r13.17} and \pref{r13.19}, which are known to hold.
\pref{r13.11.3} follows from \pref{r13.14.1} and \pref{begr:50}.
\begin{equation}
\label{begr:50}
et' = f_D(v_t')
\end{equation}

I now prove \pref{r13.10}. According to \pref{r13.13}, $v_t'$ is a
time-stamp of $eval(st, \Sigma', g^{db})$, and by clause $2'$,
$eval(st, \Sigma', g^{db}) \in \vrel(n)$. This implies that $v_t' \in
D_P$, which in turn implies \pref{begr:51}. By \pref{r13.14.1},
$f_D(v_t)$ is the instantaneous period that contains only the first
chronon of $f_D(v_t')$. Therefore, \pref{begr:52} also holds.
\begin{gather}
f_D(v_t') \in \periods \label{begr:51} \\
f_D(v_t) \in \periods \label{begr:52}
\end{gather}
From the hypothesis, $eval(st, \lambda, g^{db}) \in D_P^*$, which
implies that $f_D(eval(st, \lambda, g^{db}))$ is a period or the empty
set. \pref{r13.15} implies that $f_D(eval(st, \lambda, g^{db}))$ is
not the empty set. Therefore, \pref{begr:53} holds. \pref{r13.15}
implies \pref{begr:54}. \pref{begr:54}, \pref{r13.14.1},
\pref{begr:52}, and \pref{begr:53} imply \pref{r13.10}.
\begin{gather}
f_D(eval(st, \lambda, g^{db})) \in \periods \label{begr:53} \\
\{minpt(f_D(v_t'))\} \subseteq f_D(eval(st, \lambda, g^{db})) \label{begr:54}
\end{gather}
 
It remains to prove \pref{r13.12}. \pref{begr:50}, \pref{begr:51}, and
\pref{r13.21} imply that:
\begin{equation}
  \label{r13.200}
  et' \in \{e \in \periods \mid
  \denot{M(st),st,e,\pts,g}{\phi'} = T \} 
\end{equation}
To prove \pref{r13.12}, I need to prove that there is no 
$et''$ that satisfies both \pref{r13.201} and \pref{r13.206}. 
\begin{gather}
  \label{r13.201}
  et'' \in \{e \in \periods \mid
  \denot{M(st),st,e,\pts,g}{\phi'} = T \} \\
  \label{r13.206}
  et' \propsubper et''
\end{gather}
Let us assume that for some $et''$, \pref{r13.201} and \pref{r13.206}
hold. \pref{r13.201} implies that:
\begin{gather}
\denot{M(st),st,et'',\pts,g}{\phi'} = T \label{r13.202} \\
et'' \in \periods \label{r13.203}
\end{gather}
I set $v_t'' = \fdi(et'')$, which implies \pref{r13.205}.
\pref{r13.17}, \pref{r13.19}, \pref{r13.205}, \pref{r13.202}, and
clause $3'$ imply \pref{r13.205.1}. \pref{r13.206}, \pref{begr:50},
and \pref{r13.205} imply \pref{r13.207}.
\begin{gather}
  et'' = f_D(v_t'') \label{r13.205} \\
  \tup{v_1,\dots,v_n;v_t''} \in eval(st,\Sigma',g^{db}) \label{r13.205.1}\\
  f_D(v_t') \propsubper f_D(v_t'') \label{r13.207}
\end{gather}
\pref{r13.205.1} and \pref{r13.207} are against
\pref{r13.14}. Therefore, there can be no $et''$ that satisfies
\pref{r13.201} and \pref{r13.206}. \pref{r13.12} and the forward
direction of clause 3 have been proven. 

\bigskip

I now prove the backwards direction of clause 3. I assume that
for some $g$ and $et'$, \pref{r13.b.6} -- \pref{r13.b.12} hold.
I need to prove that $\tup{v_1,\dots,v_n;v_t} \in
eval(st,\Sigma,g^{db})$. According to \pref{r13.5.5}, it is enough to 
prove that for some $v_t'$, \pref{r13.b.13} -- \pref{r13.b.15} hold.

\pref{r13.b.12} implies \pref{r13.300}. I set $v_t' = \fdi(et')$,
which implies \pref{r13.300.2}. \pref{r13.b.6}, \pref{r13.b.8},
\pref{r13.300.2}, \pref{r13.300}, and clause $3'$ imply
\pref{r13.b.13}. \pref{r13.b.14.1} follows from \pref{r13.b.11.3} and
\pref{r13.300.2}.
\begin{gather}
  \denot{M(st),st,et',\pts,g}{\phi} = T \label{r13.300} \\
  et' = f_D(v_t') \label{r13.300.2}
\end{gather}

I now prove \pref{r13.b.15}. \pref{r13.b.10} and \pref{r13.b.11.3}
imply \pref{begr:70}. \pref{begr:70} and \pref{r13.300.2} imply
\pref{r13.b.15}.
\begin{equation}
\label{begr:70}
minpt(et') \in f_D(eval(st, \lambda, g^{db}))
\end{equation}
It remains to prove \pref{r13.b.14}. Let us assume that there is a
tuple $\tup{v_1,\dots,v_n;v_t''}$, such that \pref{r13.400} --
\pref{r13.401} hold. 
\begin{gather}
  \tup{v_1,\dots,v_n;v_t''} \in eval(st,\Sigma',g^{db}) \label{r13.400} \\
  f_D(v_t') \propsubper f_D(v_t'') \label{r13.401}
\end{gather}
\pref{r13.400} and clause $3'$ imply that for some $g'$:
\begin{eqnarray}
&& g' \in G \label{r13.402} \\
&& \denot{M(st),g'}{\tau_1} = f_D(v_1), \dots,
   \denot{M(st),g'}{\tau_n} = f_D(v_n) \label{r13.404} \\
&& \denot{M(st),st,f_D(v_t''),\pts,g'}{\phi'} = T \label{r13.406}
\end{eqnarray}
Lemma \ref{l7}, \pref{r13.b.6}, \pref{r13.402}, the assumptions that
$\phi' \in \ynforms$ and $\corn{\phi'} = \tup{\tau_1,\dots,\tau_n}$, 
\pref{r13.b.8}, and \pref{r13.404} imply that for every variable
$\beta$ in $\phi'$, $g(\beta) = g'(\beta)$. Then, \pref{r13.412}
holds. \pref{r13.406} and \pref{r13.412} imply \pref{r13.413}. 
\begin{gather}
  \denot{M(st),st,f_D(v_t''),\pts,g'}{\phi'} =
  \denot{M(st),st,f_D(v_t''),\pts,g}{\phi'}  \label{r13.412} \\
  \denot{M(st),st,f_D(v_t''),\pts,g}{\phi'} = T  \label{r13.413}
\end{gather}
I set $et''$ as in \pref{begr:85}. \pref{r13.401}, \pref{r13.300.2},
and \pref{begr:85} imply \pref{r13.414}. \pref{r13.413} and
\pref{begr:85} imply \pref{begr:97}.
\begin{gather}
  \label{begr:85}
  et'' = f_D(v_t'') \\
  \label{r13.414}
  et' \propsubper et'' \\
  \label{begr:97}
  \denot{M(st),st,et'',\pts,g}{\phi'} = T
\end{gather}
\pref{r13.414}, and \pref{begr:97} are against
\pref{r13.b.12}, because \pref{r13.b.12} and the definition of
$mxlpers$ (see section \ref{temporal_ontology}) imply that there is no
$et'' \in \periods$, such that $et' \propsubper et''$ and
$\denot{M(st),st,et'',\pts,g}{\phi'} = T$. Therefore, there can be no
tuple $\tup{v_1,\dots,v_n;v_t''}$ such that \pref{r13.400} and
\pref{r13.401} hold. \pref{r13.b.14} and the backwards direction of
clause 3 have been proven.


\subsection{$End[\phi']$} 

\subsubsection*{Translation rule}

If $\phi' \in \ynforms$ and $\lambda$ is a \tsql value expression, then:

$trans(\lend[\phi'], \lambda) \defeq$\\
\sql{(}\select{SELECT DISTINCT $\alpha$.1, $\alpha$.2, \dots, $\alpha$.$n$ \\
               VALID PERIOD(END(VALID($\alpha$)), END(VALID($\alpha$))) \\
               FROM $trans(\phi', \linit)$(NOSUBPERIOD) AS $\alpha$ \\
               WHERE $\lambda$ CONTAINS END(VALID($\alpha$)))}

$n$ is the length of $\corn{\phi'}$, and $\linit$ is as in section
\ref{formulation}. Each time the translation rule is
used, $\alpha$ is a new correlation name, obtained by calling the
correlation names generator after $\lambda$ has been supplied.

\subsubsection*{Proof that theorem \ref{yn_theorem} holds for $\phi =
End[\phi']$, if it holds for $\phi = \phi'$}

The proof is very similar to that of section \ref{beg_rule}. 


\subsection{$At[\kappa, \phi']$} \label{atk_rule}

\subsubsection*{Translation rule}

If $\kappa \in \cons$, $\phi' \in \ynforms$, and $\lambda$ is a \tsql
value expression, then:

$trans(\at[\kappa, \phi'], \lambda) \defeq trans(\phi', \lambda')$

where $\lambda'$ is the expression \sql{INTERSECT($\lambda$,
$\hconsp(\kappa)$)}. 

\subsubsection*{Proof that theorem \ref{yn_theorem} holds for $\phi =
At[\kappa, \phi']$, if it holds for $\phi = \phi'$}

I assume that $\kappa \in \cons$ and $\phi' \in \ynforms$. By the
syntax of \topl, this implies that $\at[\kappa, \phi'] \in \ynforms$.
I also assume that $st \in \pts$, $\lambda$ is a \tsql value
expression, $\lambda'$ is as in the translation rule, $g^{db} \in
G^{db}$, $eval(st,\lambda, g^{db}) \in D_P^*$, $\corn{\phi'} =
\tup{\tau_1, \tau_2, \tau_3, \dots, \tau_n}$, and $\Sigma =
trans(\at[\kappa, \phi'], \lambda)$. From the definition of
$\corn{\dots}$, it should be easy to see that $\corn{\at[\kappa,
  \phi']} = \corn{\phi'} = \tup{\tau_1, \dots, \tau_n}$. Finally, I
assume that theorem \ref{yn_theorem} holds for $\phi = \phi'$. I need
to show that:
\begin{enumerate}
\item $\fcn(\Sigma) \subseteq \fcn(\lambda)$
\item $eval(st, \Sigma, g^{db}) \in \vrel(n)$
\item $\tup{v_1, \dots, v_n; v_t} \in 
       eval(st, \Sigma, g^{db})$ iff for some $g \in G$: \\
   $\denot{M(st), g}{\tau_1} = f_D(v_1)$, \dots, 
   $\denot{M(st), g}{\tau_n} = f_D(v_n)$, and \\
   $\denot{M(st), st, f_D(v_t), f_D(eval(st, \lambda, g^{db})), g}{
   \at[\kappa, \phi']} = T$
\end{enumerate}

The restrictions of section \ref{TOP_mods} guarantee \pref{atkr:1}.
By the definitions of \fcons and \hcons of sections
\ref{resulting_model} and \ref{via_TSQL2}, \pref{atkr:0} holds and
\pref{atkr:1} is equivalent to \pref{atkr:2}. (I do not include the
assignment to the correlation names among the arguments of $eval$,
because by section \ref{via_TSQL2}, $\fcn(\hconsp(\kappa)) =
\emptyset$.)
\begin{gather}
\fcons(st)(\kappa) \in \periods \label{atkr:1} \\
\fcons(st)(\kappa) = f_D(eval(st, \hconsp(\kappa))) \label{atkr:0} \\
f_D(eval(st, \hconsp(\kappa))) \in \periods \label{atkr:2}
\end{gather}

By the syntax of \tsql, since $\lambda$ is a value expression,
$\lambda'$ is also a value expression.
From the definition of $\lambda'$ it should be obvious that any free
column reference in $\lambda'$ is situated within the $\lambda$ of 
$\lambda'$. By lemma \ref{fcn_lemma}, this implies \pref{atkr:4}.
\begin{equation}
\label{atkr:4}
\fcn(\lambda') \subseteq \fcn(\lambda)
\end{equation}
The assumption that $eval(st, \lambda, g^{db}) \in D_P^*$,
\pref{atkr:2}, and the definition of $\lambda'$ imply \pref{atkr:3}
and that $eval(st, \lambda',g^{db}) \in D^*_P$. 
\begin{eqnarray}
&& \label{atkr:3}
f_D(eval(st,\lambda', g^{db})) = f_D(eval(st, \lambda, g^{db}))
 \intersect f_D(eval(st, \hconsp(\kappa)))
\end{eqnarray}

The hypothesis that $\Sigma = trans(\at[\kappa, \phi'], \lambda)$ and
the translation rule of this section imply that $\Sigma = trans(\phi,
\lambda')$. From the hypothesis, $\phi' \in \ynforms$, $st \in \pts$,
$\corn{\phi'} = \tup{\tau_1, \dots, \tau_n}$, and $g^{db} \in G^{db}$.
According to the discussion above, $\lambda'$ is a value expression,
and $eval(st, \lambda', g^{db}) \in D_P^*$.  Then, from theorem
\ref{yn_theorem} for $\phi = \phi'$ (according to the hypothesis,
theorem \ref{yn_theorem} holds for $\phi = \phi'$) we get:
\begin{enumerate}
\item[$1'.$] $\fcn(\Sigma) \subseteq \fcn(\lambda')$
\item[$2'.$] $eval(st, \Sigma, g^{db}) \in \vrel(n)$
\item[$3'.$] $\tup{v_1, \dots, v_n; v_t} \in 
       eval(st, \Sigma, g^{db})$ iff for some $g \in G$: \\
   $\denot{M(st), g}{\tau_1} =
   f_D(v_1)$, \dots, $\denot{M(st), g}{\tau_n} = f_D(v_n)$, and \\
   $\denot{M(st), st, f_D(v_t), f_D(eval(st, \lambda', g^{db})), g}{
   \phi'} = T$
\end{enumerate}

\subsubsection*{Proofs of clauses 1 and 2}

Clause 1 follows from clause $1'$ and \pref{atkr:4}. Clause 2 is the
same as clause $2'$, which is known to hold.

\subsubsection*{Proof of clause 3}

I prove clause 3 by proving \pref{atkr:10}. If \pref{atkr:10} holds,
then clause $3'$ is equivalent to clause 3, and since clause $3'$
holds, clause 3 holds too.
\begin{gather}
\label{atkr:10}
\denot{M(st), st, f_D(v_t), f_D(eval(st, \lambda', g^{db})), g}{
   \phi'} = T,
\text{ iff} \\
\nonumber
\denot{M(st), st, f_D(v_t), f_D(eval(st, \lambda, g^{db})), g}{
   \at[\kappa, \phi']} = T 
\end{gather}
From the hypothesis, $\kappa \in \cons$. By the semantics of \topl,
this implies that:
\begin{equation}
\label{atkr:40}
\denot{M(st), g}{\kappa} = \fcons(st)(\kappa)
\end{equation}

I first prove the forward direction of \pref{atkr:10}. I assume that
\pref{atkr:11} holds.
\begin{equation}
\label{atkr:11}
\denot{M(st), st, f_D(v_t), f_D(eval(st, \lambda', g^{db})), g}{
   \phi'} = T
\end{equation}
Using \pref{atkr:3} and \pref{atkr:0}, \pref{atkr:11} becomes \pref{atkr:30}.
\begin{equation}
\label{atkr:30}
\denot{M(st), st, f_D(v_t), f_D(eval(st, \lambda, g^{db}))
 \intersect \fcons(st)(\kappa), g}{\phi'} = T
\end{equation}
Using \pref{atkr:40}, \pref{atkr:30} and \pref{atkr:1} become
\pref{atkr:31} and \pref{atkr:12} respectively. \pref{atkr:31},
\pref{atkr:12}, and the definition of \at (section
\ref{at_before_after_op}) imply \pref{atkr:14}. The forward direction
of \pref{atkr:10} has been proven.
\begin{gather}
\label{atkr:31}
\denot{M(st), st, f_D(v_t), f_D(eval(st, \lambda, g^{db}))
 \intersect \denot{M(st), g}{\kappa}, g}{\phi'} = T \\
\label{atkr:12}
\denot{M(st), g}{\kappa} \in \periods \\
\label{atkr:14}
\denot{M(st), st, f_D(v_t), f_D(eval(st, \lambda, g^{db})), g}{
   \at[\kappa, \phi']} = T 
\end{gather}

I now prove the backwards direction of \pref{atkr:10}. I assume that
\pref{atkr:14} holds. By the definition of \at, this implies that
\pref{atkr:31} holds. Using \pref{atkr:40}, \pref{atkr:31} becomes
\pref{atkr:30}. \pref{atkr:0}, \pref{atkr:3}, and \pref{atkr:30}
imply \pref{atkr:11}.  The backwards direction of \pref{atkr:10} has
been proven.


\subsection{$Before[\kappa, \phi']$}

\subsubsection*{Translation rule}

If $\kappa \in \cons$, $\phi' \in \ynforms$, and $\lambda$ is a \tsql
value expression, then:

$trans(\before[\kappa, \phi'], \lambda) \defeq trans(\phi', \lambda')$

where $\lambda'$ is the expression:

\sql{INTERSECT($\lambda$, 
     PERIOD(TIMESTAMP 'beginning', 
     BEGIN($\hconsp(\kappa)$) - INTERVAL '1' $\chi$))}

and $\chi$ is the \tsql name of the granularity of chronons.

\subsubsection*{Proof that theorem \ref{yn_theorem} holds for $\phi =
Before[\kappa, \phi']$, if it holds for $\phi = \phi'$}

The proof is very similar to that of section \ref{atk_rule}. 


\subsection{$After[\kappa, \phi']$}

\subsubsection*{Translation rule}

If $\kappa \in \cons$, $\phi' \in \ynforms$, and $\lambda$ is a \tsql
value expression, then:

$trans(\after[\kappa, \phi'], \lambda) \defeq trans(\phi', \lambda')$

where $\lambda'$ is the expression:

\sql{INTERSECT($\lambda$,
     PERIOD(END($\hconsp(\kappa)$) + INTERVAL '1' $\chi$, 
            TIMESTAMP 'forever'))}

and $\chi$ is the \tsql name of the granularity of chronons.

\subsubsection*{Proof that theorem \ref{yn_theorem} holds for $\phi =
After[\kappa, \phi']$, if it holds for $\phi = \phi'$}

The proof is very similar to that of section \ref{atk_rule}. 


\subsection{$At[\sigma_g, \beta, \phi']$} \label{atsg_rule}

\subsubsection*{Translation rule}

If $\sigma_g \in \gparts$, $\beta \in \vars$, $\phi' \in \ynforms$,
and $\lambda$ is a \tsql value expression, then:

$trans(\at[\sigma_g, \beta, \phi'], \lambda) \defeq$\\
\sql{(}\select{SELECT DISTINCT $\alpha_1$.1, $\alpha_2$.1, $\alpha_2$.2, \dots,
                               $\alpha_2$.$n$ \\
               VALID VALID($\alpha_2$) \\
               FROM ($\hgpartsp(\sigma_g)$) AS $\alpha_1$, 
                    $trans(\phi', \lambda')$ AS $\alpha_2$)}

$n$ is the length of $\corn{\phi'}$, and $\lambda'$ is the expression 
\sql{INTERSECT($\alpha_1.1$, $\lambda$)}\footnote{$\lambda'$ will end
up being a part of the embedded \sql{SELECT} statement $trans(\phi',
\lambda')$. This means that $\alpha_1$, which is part of $\lambda'$,
will appear within $trans(\phi', \lambda')$, i.e.\ in the same
\sql{FROM} clause that defines $\alpha_1$. This is allowed in the
\tsql version of this thesis; see section \ref{same_FROM}. Similar
comments apply to the translation rules for $\before[\sigma_g, \beta, \phi']$,
$\after[\sigma_g, \beta, \phi']$, $\at[\sigma_c, \beta, \phi']$,
$\before[\sigma_c, \beta, \phi']$, and $\after[\sigma_c, \beta,
\phi']$.} Each time the translation rule is used, $\alpha_1$ and
$\alpha_2$ are two new different correlation names, obtained by
calling the correlation names generator after $\lambda$ has been
supplied.

\subsubsection*{Proof that theorem \ref{yn_theorem} holds for $\phi =
At[\sigma_g, \beta, \phi']$, if it holds for $\phi = \phi'$}

I assume that $\sigma_g \in \gparts$, $\beta \in \vars$, and $\phi'
\in \ynforms$. By the syntax of \topl, this implies that
$\at[\sigma_g, \beta, \phi'] \in \ynforms$. I also assume that $st \in
\pts$, $\lambda$ is a \tsql value expression, $\lambda'$ is as in the
translation rule, $g^{db} \in G^{db}$, $eval(st,\lambda, g^{db}) \in
D_P^*$, $\corn{\phi'} = \tup{\tau_1, \tau_2, \tau_3, \dots, \tau_n}$,
and $\Sigma = trans(\at[\sigma_g, \beta, \phi'], \lambda)$. From the
definition of $\corn{\dots}$, it should be easy to see that
$\corn{\at[\sigma_g, \beta, \phi']} =\tup{\beta, \tau_1, \dots,
  \tau_n}$. Finally, I assume that theorem \ref{yn_theorem} holds for
$\phi = \phi'$. I need to show that:
\begin{enumerate}
\item $\fcn(\Sigma) \subseteq \fcn(\lambda)$
\item $eval(st, \Sigma, g^{db}) \in \vrel(n+1)$
\item $\tup{v, v_1, \dots, v_n; v_t} \in 
       eval(st, \Sigma, g^{db})$ iff for some $g \in G$: \\
   $\denot{M(st), g}{\beta} = f_D(v)$, $\denot{M(st), g}{\tau_1} =
   f_D(v_1)$, \dots, $\denot{M(st), g}{\tau_n} = f_D(v_n)$, and \\
   $\denot{M(st), st, f_D(v_t), f_D(eval(st, \lambda, g^{db})), g}{
   \at[\sigma_g, \beta, \phi']} = T$
\end{enumerate}

From the definition of $\lambda'$ it should be obvious that any free
column reference in $\lambda'$ is either the \sql{$\alpha_1$.1} of
$\lambda'$, or is situated within the $\lambda$ of $\lambda'$. By
lemma \ref{fcn_lemma}, this implies \pref{atg:1}. The only correlation
name that has a free column reference in $\alpha_1.1$ is $\alpha_1$.
Hence, $\fcn(\alpha_1.1) = \{\alpha_1\}$, and \pref{atg:1} becomes
\pref{atg:2}.
\begin{gather}
\fcn(\lambda') \subseteq \fcn(\alpha_1.1) \union \fcn(\lambda) \label{atg:1}\\
\fcn(\lambda') \subseteq \{\alpha_1\} \union \fcn(\lambda) \label{atg:2}
\end{gather}

The $\alpha_1$ of $\Sigma$ is generated by calling the correlation
names generator after $\lambda$ has been supplied.  Hence, $\alpha_1$
cannot appear in $\lambda$. The fact that $\alpha_1$ does not appear
in $\lambda$ means that for every possible tuple $\tup{v}$ of a
one-attribute snapshot relation (i.e.\ for every $v \in D$):
\begin{equation}
\label{atg:3}
eval(st, \lambda, (g^{db})^{\alpha_1}_{\tup{v}}) =
eval(st, \lambda, g^{db})
\end{equation}

The assumption that $eval(st, \lambda, g^{db}) \in
D_P^*$, and the fact that for every $v \in D$, \pref{atg:3} holds imply
that \pref{atg:5} and \pref{atg:6} hold for every $v \in D$.
\begin{gather}
\label{atg:5}
eval(st, \lambda, (g^{db})^{\alpha_1}_{\tup{v}}) \in D_P^* \\
\label{atg:6}
f_D(eval(st, \lambda, (g^{db})^{\alpha_1}_{\tup{v}})) \in \periods^*
\end{gather}

By the syntax of \topl, since $\lambda$ is a value expression,
$\lambda'$ is also a value expression. If $v \in D_P$ ($D_P \subseteq
D$), by \pref{atg:6} (which holds for $v \in D$) and the definition of
$\lambda'$, \pref{atg:7} and \pref{atg:8} hold.
\begin{eqnarray}
&& \label{atg:7}
   f_D(eval(st, \lambda', (g^{db})^{\alpha_1}_{\tup{v}})) =
   f_D(v) \intersect 
   f_D(eval(st, \lambda, (g^{db})^{\alpha_1}_{\tup{v}})) \\
&& \label{atg:8}
   eval(st, \lambda',(g^{db})^{\alpha_1}_{\tup{v}}) \in D_P^*
\end{eqnarray}

Let $\Sigma'$ be the embedded \sql{SELECT} statement in the \sql{FROM}
clause of $\Sigma$ to which $\alpha_2$ refers, i.e.\ $\Sigma' =
trans(\phi', \lambda')$. From the hypothesis, $\phi' \in \ynforms$,
$st \in \pts$, $g^{db} \in G$, and $\corn{\phi'} = \tup{\tau_1, \dots,
  \tau_n}$. From the discussion above, $\lambda'$ is a value
expression. If $v \in D_P$, then $(g^{db})^{\alpha_1}_{\tup{v}} \in
G^{db}$ and (by \pref{atg:8}) $eval(st, \lambda',
(g^{db})^{\alpha_1}_{\tup{v}}) \in D_P^*$. From theorem
\ref{yn_theorem} for $\phi = \phi'$ (according to the hypothesis,
theorem \ref{yn_theorem} holds for $\phi = \phi'$) we get the
following. (The condition $v \in D_P$ does not affect clause $1'$.)
\begin{enumerate}
\item[$1'.$] $\fcn(\Sigma') \subseteq \fcn(\lambda')$
\item[$2'.$] If $v \in D_P$, then $eval(st, \Sigma',
   (g^{db})^{\alpha_1}_{\tup{v}}) \in \vrel(n)$ 
\item[$3'.$] If $v \in D_P$, then:\\
( \\
 $\tup{v_1, \dots, v_n; v_t} \in 
        eval(st, \Sigma', (g^{db})^{\alpha_1}_{\tup{v}})$ 
         iff for some $g' \in G$: \\
 $\denot{M(st), g'}{\tau_1} =
         f_D(v_1)$, \dots, $\denot{M(st), g'}{\tau_n} = f_D(v_n)$, and \\
 $\denot{M(st), st, f_D(v_t), f_D(eval(st, \lambda',
         (g^{db})^{\alpha_1}_{\tup{v}})), g'}{\phi'} = T$ \\
)
\end{enumerate}

\subsubsection*{Proof of clause 1}

The \sql{VALID($\alpha_2$)} in the \sql{VALID} clause of $\Sigma$ and
the $\alpha_1.1$, $\alpha_2.1$, \dots, $\alpha_2.n$ in the
\sql{SELECT} clause of $\Sigma$ are not free column references in
$\Sigma$, because $\Sigma$ is a binding context for all of them.
$\Sigma$ contains no other column references (and hence no other free
column references), apart from those that possibly appear within the
$\hgpartsp(\sigma_g)$ or the $\Sigma'$ in the \sql{FROM} clause of
$\Sigma$. By lemma \ref{fcn_lemma}, this implies \pref{atg:10}. The
definition of $\hgpartsp(\sigma_g)$ of section \ref{via_TSQL2} implies
that $\fcn(\hgpartsp(\sigma_g)) = \emptyset$. Hence, \pref{atg:10}
becomes \pref{atg:11}. \pref{atg:11} and clause $1'$ imply
\pref{atg:12}.  \pref{atg:12} and \pref{atg:2} imply \pref{atg:13}.
\begin{gather}
\label{atg:10}
\fcn(\Sigma) \subseteq \fcn(\hgpartsp(\sigma_g)) \union \fcn(\Sigma') \\
\label{atg:11}
\fcn(\Sigma) \subseteq \fcn(\Sigma') \\
\label{atg:12}
\fcn(\Sigma) \subseteq \fcn(\lambda') \\
\label{atg:13}
\fcn(\Sigma) \subseteq \{\alpha_1\} \union \fcn(\lambda)
\end{gather}

$\alpha_1$ cannot have a free column reference in $\Sigma$.  The proof
follows. Let us assume that there is a free column reference $\zeta$
of $\alpha_1$ (i.e.\ of the form $\alpha_1.k$ or
\sql{VALID($\alpha_1$)}) in $\Sigma$. Then, by the definition of free
column reference, (a) $\zeta$ is part of $\Sigma$. From the
translation rule, it is also the case that (b) $\alpha_1$ is defined
by the topmost \sql{FROM} clause of $\Sigma$.

If $\zeta$ is in the topmost \sql{FROM} clause of $\Sigma$, it is
either in the $\hgpartsp(\sigma_g)$ or in the $\Sigma'$ (the
$trans(\phi', \lambda')$). $\zeta$, however, cannot be in the
$\hgpartsp(\sigma_g)$, because $\alpha_1$ is generated by the
correlation names generator, and it is assumed that the correlation
names of the generator do not appear in the \sql{SELECT} statements
returned by \hgpartsp (see section \ref{trans_rules}).

So, if $\zeta$ is in the topmost \sql{FROM} clause of $\Sigma$, it is
in the $\Sigma'$, i.e.\ after the definition of $\alpha_1$. Hence, (c)
$\zeta$ is either not in the topmost \sql{FROM} clause of $\Sigma$, or
it is in the topmost \sql{FROM} clause of $\Sigma$, but it follows the
definition of $\alpha_1$. (a), (b), and (c) imply that $\Sigma$ is a
binding context for $\zeta$. This implies that $\zeta$ is not a free
column reference in $\Sigma$, which is against the hypothesis. Hence,
$\alpha_1$ cannot have a free column reference in $\Sigma$, i.e.\ 
$\alpha_1 \not\in \fcn(\Sigma)$. This and \pref{atg:13} imply that
$\fcn(\Sigma) \subseteq \fcn(\lambda)$. Clause 1 has been proven.

\subsubsection*{Proof of clause 2}

According to section \ref{via_TSQL2}, $\fcn(\hgpartsp(\sigma_g)) =
\emptyset$, $eval(st, \hgpartsp(\sigma_g)) \in \srel(1)$, and for
every $\tup{v} \in eval(st, \hgpartsp(\sigma_g))$, $v$ represents a
period of a gappy partitioning (hence, $v \in D_P$). By clause $2'$,
it is also the case that if $v \in D_P$, then $eval(st, \Sigma',
(g^{db})^{\alpha_1}_{\tup{v}}) \in \vrel(n)$. It should now be easy to
see from the translation rule that \pref{r15.8} holds.
\begin{eqnarray}
eval(st,\Sigma,g^{db}) &=& \{\tup{v,v_1,\dots,v_n;v_t} \mid 
\label{r15.8} \\
&& \tup{v} \in eval(st, \hgpartsp(\sigma_g)), \text{ and}
\nonumber \\
&& \tup{v_1,\dots,v_n;v_t} \in
eval(st,\Sigma',(g^{db})^{\alpha_1}_{\tup{v}}) \} 
\nonumber
\end{eqnarray} 
$(g^{db})^{\alpha_1}_{\tup{v}}$ is used in the last line of
\pref{r15.8} instead of $g^{db}$, to capture the fact that if there is
any free column reference of $\alpha_1$ in $\Sigma'$, this has to be
taken to refer to the $\tup{v}$ tuple of $eval(st,
\hgpartsp(\sigma_g))$ to which $\alpha_1$ refers. 
 
\pref{r15.8} and the discussion above imply that for every
$\tup{v,v_1, \dots, v_n;v_t} \in eval(st, \Sigma, g^{db})$, $\tup{v_1,
  \dots, v_n; v_t} \in eval(st,\Sigma',(g^{db})^{\alpha_1}_{\tup{v}})
\in \vrel(n)$. This implies that $v_t \in D_P$. \pref{r15.8} also
implies that $eval(st,\Sigma,g^{db})$ is a valid-time relation of
$n+1$ explicit attributes. Hence, $eval(st, \Sigma, g^{db}) \in
\vrel(n+1)$. Clause 2 has been proven.

\subsubsection*{Proof of clause 3}

Using the definition of
$\denot{st,et,lt,g}{\at[\sigma_g,\beta,\phi']}$ (section
\ref{TOP_mods}), and the fact that $\denot{M(st),g}{\beta} =
g(\beta)$, clause 3 becomes:
\begin{eqnarray}
&& \tup{v,v_1,\dots,v_n;v_t} \in eval(st,\Sigma,g^{db}) 
\mbox{ iff for some } g: \nonumber \\
&& g \in G \label{r15.10} \label{r15.b.10} \\
&& g(\beta) = f_D(v) \label{r15.12} \label{r15.b.12}\\
&& \denot{M(st),g}{\tau_1} = f_D(v_1), \; \dots, \; \denot{M(st),g}{\tau_n} =
   f_D(v_n) \label{r15.13} \label{r15.b.13}\\
&& g(\beta) \in \fgparts(st)(\sigma_g) \label{r15.15} \label{r15.b.15}\\
&& \denot{M(st),st,f_D(v_t),f_D(eval(st, \lambda, g^{db})) \intersect
   g(\beta),g}{\phi'} = T \label{r15.18} \label{r15.b.18}
\end{eqnarray}
I first prove the forward direction of clause 3. I assume that
$\tup{v,v_1,\dots,v_n;v_t} \in eval(st,\Sigma,g^{db})$. I need to prove that
for some $g$, \pref{r15.10} -- \pref{r15.18} hold.
The assumption that $\tup{v,v_1,\dots,v_n;v_t} \in
eval(st,\Sigma,g^{db})$, and \pref{r15.8} imply that:
\begin{eqnarray}
&& \tup{v} \in eval(st, \hgpartsp(\sigma_g)) \label{r15.19} \label{r15.b.19}\\
&& \tup{v_1,\dots,v_n;v_t} \in
   eval(st,\Sigma',(g^{db})^{\alpha_1}_{\tup{v}}) 
   \label{r15.21} \label{r15.b.21} 
\end{eqnarray}
\pref{r15.19} and the discussion in the proof of clause 2 imply
\pref{r15.99}. 
\begin{equation}
  \label{r15.99}
  v \in D_P
\end{equation}
\pref{r15.99}, \pref{r15.21}, and clause $3'$ imply that for some
$g'$:
\begin{eqnarray}
&& g' \in G \label{r15.100} \\
&& \denot{M(st),g'}{\tau_1} = f_D(v_1), \dots, \denot{M(st),g'}{\tau_n} =
   f_D(v_n) \label{r15.102} \\
&& \denot{M(st), st, f_D(v_t), f_D(eval(st, \lambda',
   (g^{db})^{\alpha_1}_{\tup{v}})), g'}{\phi'} = T \label{r15.104}
\end{eqnarray}
Let $g = (g')^\beta_{f_D(v)}$. Lemma \ref{l6}, \pref{r15.99}, the
assumption that $\beta \in \vars$, \pref{r15.100}, and the definition
of $g$ imply \pref{r15.10}. \pref{r15.12} follows from the definition
of $g$. 

I now prove \pref{r15.18}. \pref{r15.99}, \pref{atg:7} (which holds
for $v \in D_P$,) and \pref{r15.12} (already proven) imply
\pref{r15.105ex}. \pref{r15.99}, the fact that $D_P \subseteq D$,
\pref{atg:3} (which holds for $v \in D$), and \pref{r15.105ex} imply
\pref{r15.105}. \pref{r15.105} and \pref{r15.104} imply \pref{r15.106}. 
\begin{eqnarray}
&& f_D(eval(st, \lambda',(g^{db})^{\alpha_1}_{\tup{v}})) = 
   g(\beta) \intersect f_D(eval(st, \lambda,(g^{db})^{\alpha_1}_{\tup{v}}))
   \label{r15.105ex} \\
&& f_D(eval(st, \lambda',(g^{db})^{\alpha_1}_{\tup{v}})) = 
   g(\beta) \intersect f_D(eval(st, \lambda,g^{db}))  \label{r15.105} \\
&& \denot{M(st), st, f_D(v_t), f_D(eval(st, \lambda,g^{db})) \intersect
   g(\beta), g'}{\phi'} = T  \label{r15.106}
\end{eqnarray}
The syntax of \topl (section \ref{TOP_mods}) and the fact that
$\at[\sigma_g,\beta,\phi'] \in \ynforms$, imply that $\beta$ does
not occur in $\phi'$. This and the definition of $g$ imply
\pref{r15.107}. \pref{r15.106} and \pref{r15.107} imply \pref{r15.18}.
\begin{equation}
\label{r15.107}
\begin{aligned}[t]
& \denot{M(st), st, f_D(v_t),f_D(eval(st, \lambda, g^{db})) \intersect
   g(\beta), g'}{\phi'} =  \\
& \denot{M(st), st, f_D(v_t),f_D(eval(st, \lambda, g^{db})) \intersect
   g(\beta), (g')^\beta_{f_D(v)}}{\phi'} = \\
& \denot{M(st), st, f_D(v_t),f_D(eval(st, \lambda, g^{db})) \intersect
   g(\beta), g}{\phi'}
\end{aligned}
\end{equation}
\pref{r15.13} follows from lemma \ref{l3}, the assumption that
$\corn{\phi'} = \tup{\tau_1, \dots, \tau_n}$, \pref{r15.102}, and the
fact that $g$ and $g'$ assign the same values to all variables,
possibly apart from $\beta$, which does not occur in $\phi'$.

It remains to prove \pref{r15.15}. The definition of \hgparts of
section \ref{via_TSQL2} and \pref{r15.19} imply \pref{r15.108}.
\pref{r15.108} and the definition of \fgparts of section
\ref{resulting_model} imply \pref{r15.109}. \pref{r15.109} and
\pref{r15.12} (proven above) imply \pref{r15.15}. The forward
direction of clause 3 has been proven.
\begin{gather}
  \label{r15.108}
  \tup{v} \in \hgparts(st)(\sigma_g) \\
  \label{r15.109}
  f_D(v) \in \fgparts(st)(\sigma_g)
\end{gather}

\bigskip

I now prove the backwards direction of clause 3. I assume that
\pref{r15.b.10} -- \pref{r15.b.18} hold.
I need to prove that $\tup{v,v_1,\dots,v_n;v_t} \in
eval(st,\Sigma,g^{db})$. According to \pref{r15.8}, it is enough to
prove that \pref{r15.b.19} and \pref{r15.b.21} hold. 

\pref{r15.b.12} and \pref{r15.b.15} imply \pref{r15.200}.
\pref{r15.200} and the definition of $\fgparts$ of section
\ref{resulting_model} imply \pref{r15.201}. \pref{r15.201} and the
definition of \hgparts of section \ref{via_TSQL2} imply
\pref{r15.b.19}.
\begin{gather}
  \label{r15.200}
  f_D(v) \in \fgparts(st)(\sigma_g)\\
  \label{r15.201}
  \tup{v} \in \hgparts(st)(\sigma_g) 
\end{gather}

It remains to prove \pref{r15.b.21}. \pref{r15.200} implies that
$f_D(v) \in \periods$, which in turn implies \pref{r15.ym.2}.
\pref{r15.ym.2}, \pref{atg:7} (which holds for $v \in D_P$),
\pref{atg:3} (which holds for $v \in D$), and \pref{r15.b.12} imply
\pref{r15.204}. \pref{r15.b.18} and \pref{r15.204} imply \pref{r15.205}. 
\begin{gather}
  \label{r15.ym.2}
  v \in D_P \\
  \label{r15.204}
  f_D(eval(st,\lambda',(g^{db})^{\alpha_1}_{\tup{v}})) = g(\beta) \intersect
  f_D(eval(st, \lambda,g^{db})  \\
  \label{r15.205}
  \denot{M(st), st, f_D(v_t), f_D(eval(st, \lambda',
   (g^{db})^{\alpha_1}_{\tup{v}})), g}{\phi'} = T
\end{gather}
Clause $3'$, \pref{r15.ym.2}, \pref{r15.b.10}, \pref{r15.b.13}, and
\pref{r15.205} imply \pref{r15.b.21}. The backwards direction of
clause 3 has been proven.


\subsection{$\mathit{Before}[\sigma_g, \beta, \phi']$}

\subsubsection*{Translation rule}

If $\sigma_g \in \gparts$, $\beta \in \vars$, $\phi' \in \ynforms$,
and $\lambda$ is a \tsql value expression, then:

$trans(\before[\sigma_g, \beta, \phi'], \lambda) \defeq$\\
\sql{(}\select{SELECT DISTINCT $\alpha_1.1$, $\alpha_2$.1, $\alpha_2$.2, \dots,
                               $\alpha_2$.$n$ \\
               VALID VALID($\alpha_2$) \\
               FROM ($\hgpartsp(\sigma_g)$) AS $\alpha_1$, 
                    $trans(\phi', \lambda')$ AS $\alpha_2$)}

$n$ is the length of $\corn{\phi'}$, $\lambda'$ is the expression 
\sql{INTERSECT(PERIOD(TIMESTAMP 'beginning', BEGIN($\alpha_1.1$) -
INTERVAL '1' $\chi$), $\lambda$)}, and $\chi$ is the \tsql name of the
granularity of chronons. Each time the translation rule is used, $\alpha_1$ and
$\alpha_2$ are two new different correlation names, obtained by 
calling the correlation names generator after $\lambda$ has been
supplied.

\subsubsection*{Proof that theorem \ref{yn_theorem} holds for $\phi =
\mathit{Before}[\sigma_g, \beta, \phi']$, if it holds for $\phi = \phi'$}

The proof is very similar to that of section \ref{atsg_rule}. 


\subsection{$\mathit{After}[\sigma_g, \beta, \phi']$}

\subsubsection*{Translation rule}

If $\sigma_g \in \gparts$, $\beta \in \vars$, $\phi' \in \ynforms$,
and $\lambda$ is a \tsql value expression, then:

$trans(\after[\sigma_g, \beta, \phi'], \lambda) \defeq$\\
\sql{(}\select{SELECT DISTINCT $\alpha_1.1$, $\alpha_2$.1, $\alpha_2$.2, \dots,
                               $\alpha_2$.$n$ \\
               VALID VALID($\alpha_2$) \\
               FROM ($\hgpartsp(\sigma_g)$) AS $\alpha_1$, 
                    $trans(\phi', \lambda')$ AS $\alpha_2$)}

$n$ is the length of $\corn{\phi'}$, $\lambda'$ is the expression 
\sql{INTERSECT(PERIOD(END($\alpha_1.1$) + INTERVAL '1' $\chi$, TIMESTAMP
'forever'), $\lambda$)}, and $\chi$ is the \tsql name of the
granularity of chronons. Each time the translation rule is used, $\alpha_1$ and
$\alpha_2$ are two new different correlation names, obtained by
calling the correlation names generator after $\lambda$ as been supplied.

\subsubsection*{Proof that theorem \ref{yn_theorem} holds for $\phi =
\mathit{After}[\sigma_g, \beta, \phi']$, if it holds for $\phi = \phi'$}

The proof is very similar to that of section \ref{atsg_rule}. 


\subsection{$At[\sigma_c, \beta, \phi']$}

\subsubsection*{Translation rule}

If $\sigma_c \in \cparts$, $\beta \in \vars$, $\phi' \in \ynforms$,
and $\lambda$ is a \tsql value expression, then:

$trans(\at[\sigma_c, \beta, \phi'], \lambda) \defeq$\\
\sql{(}\select{SELECT DISTINCT $\alpha_1.1$, $\alpha_2$.1, $\alpha_2$.2, \dots,
                               $\alpha_2$.$n$ \\
               VALID VALID($\alpha_2$) \\
               FROM ($\Sigma_c$) AS $\alpha_1$, 
                    $trans(\phi', \lambda')$ AS $\alpha_2$)}

$n$ is the length of $\corn{\phi'}$, $\lambda'$ is the expression 
\sql{INTERSECT($\alpha_1.1$, $\lambda$)}, and $\Sigma_c$ is the second
element of the pair $\tup{\gamma, \Sigma_c} = \hcpartsp(\sigma_c)$
(section \ref{via_TSQL2}). Each time the translation rule is used,
$\alpha_1$ and $\alpha_2$ are two new different correlation names,
obtained by calling the correlation names generator after $\lambda$
has been supplied.

\subsubsection*{Proof that theorem \ref{yn_theorem} holds for $\phi =
At[\sigma_c, \beta, \phi']$, if it holds for $\phi = \phi'$}

The proof is very similar to that of section \ref{atsg_rule}. 


\subsection{$\mathit{Before}[\sigma_c, \beta, \phi']$}

\subsubsection*{Translation rule}

If $\sigma_c \in \cparts$, $\beta \in \vars$, $\phi' \in \ynforms$,
and $\lambda$ is a \tsql value expression, then:

$trans(\before[\sigma_c, \beta, \phi'], \lambda) \defeq$\\
\sql{(}\select{SELECT DISTINCT $\alpha_1.1$, $\alpha_2$.1, $\alpha_2$.2, \dots,
                               $\alpha_2$.$n$ \\
               VALID VALID($\alpha_2$) \\
               FROM ($\Sigma_c$) AS $\alpha_1$, 
                    $trans(\phi', \lambda')$ AS $\alpha_2$)}

$n$ is the length of $\corn{\phi'}$, $\lambda'$ is the expression
\sql{INTERSECT(PERIOD(TIMESTAMP 'beginning', BEGIN($\alpha_1.1$) -
INTERVAL '1' $\chi$), $\lambda$)}, $\chi$ is the \tsql name of the
granularity of chronons, and $\Sigma_c$ is the second element of the
pair $\tup{\gamma, \Sigma_c} = \hcpartsp(\sigma_c)$. Each time the
translation rule is used, $\alpha_1$ and $\alpha_2$ are two new
different correlation names, obtained by calling the correlation names
generator after $\lambda$ has been supplied. 

\subsubsection*{Proof that theorem \ref{yn_theorem} holds for $\phi =
\mathit{Before}[\sigma_c, \beta, \phi']$, if it holds for $\phi = \phi'$}

The proof is very similar to that of section \ref{atsg_rule}. 


\subsection{$\mathit{After}[\sigma_c, \beta, \phi']$}

\subsubsection*{Translation rule}

If $\sigma_c \in \cparts$, $\beta \in \vars$, $\phi' \in \ynforms$,
and $\lambda$ is a \tsql value expression, then:

$trans(\after[\sigma_c, \beta, \phi'], \lambda) \defeq$\\
\sql{(}\select{SELECT DISTINCT $\alpha_1.1$, $\alpha_2$.1, $\alpha_2$.2, \dots,
                               $\alpha_2$.$n$ \\
               VALID VALID($\alpha_2$) \\
               FROM ($\Sigma_c$) AS $\alpha_1$, 
                    $trans(\phi', \lambda')$ AS $\alpha_2$)}

$n$ is the length of $\corn{\phi'}$, $\lambda'$ is the expression
\sql{INTERSECT(PERIOD(END($\alpha_1.1$) + INTERVAL '1' $\chi$, TIMESTAMP
'forever'), $\lambda$)}, $\chi$ is the \tsql name of the
granularity of chronons, and $\Sigma_c$ is the second element of the
pair $\tup{\gamma, \Sigma_c} = \hcpartsp(\sigma_c)$. Each time the
translation rule is used, $\alpha_1$ and $\alpha_2$ are two new
different correlation names, obtained by calling the correlation names
generator after $\lambda$ has been supplied. 

\subsubsection*{Proof that theorem \ref{yn_theorem} holds for $\phi =
\mathit{After}[\sigma_c, \beta, \phi']$, if it holds for $\phi = \phi'$}

The proof is very similar to that of section \ref{atsg_rule}. 


\subsection{$At[\phi_1, \phi_2]$} \label{atphi_rule}

\subsubsection*{Translation rule}

If $\phi_1, \phi_2 \in \ynforms$ and $\lambda$ is a \tsql value
expression, then: 

$trans(\at[\phi_1, \phi_2], \lambda) \defeq$\\
\sql{(}\select{SELECT DISTINCT 
                  $\alpha_1$.1, $\alpha_1$.2, \dots, $\alpha_1$.$n_1$
                  $\alpha_2$.1, $\alpha_2$.2, \dots, $\alpha_2$.$n_2$ \\
               VALID VALID($\alpha_2$) \\
               FROM $trans(\phi_1,\linit)$(NOSUBPERIOD) AS $\alpha_1$, \\
               \ \ \ \ \ $trans(\phi_2, \lambda')$ AS $\alpha_2$) \\
               WHERE \dots \\
               \ \ AND \dots \\
               \ \ \vdots \\
               \ \ AND \dots)}

$\linit$ is as in section \ref{formulation}, $\lambda'$ is the expression
\sql{INTERSECT(VALID($\alpha_1$),$\lambda$)}, and $n_1, n_2$ are
the lengths of $\corn{\phi_1}$ and $\corn{\phi_2}$ respectively. Each
time the translation rule is used, $\alpha_1$ and $\alpha_2$ are two
new different correlation names, obtained by calling the correlation
names generator after $\lambda$ has been supplied. $\alpha_1$ is
generated after $trans(\phi_1, \linit)$ has been computed. Assuming
that $\corn{\phi_1} = \tup{\tau^1_1, \tau^1_2, \dots, \tau^1_{n_1}}$ and
$\corn{\phi_2} = \tup{\tau^2_1, \dots, \tau^2_{n_2}}$, the ``\dots''s
in the \sql{WHERE} clause are all the strings in $S$:
\[
\begin{aligned}
S = \{&\mbox{``}\alpha_1.i = \alpha_2.j\mbox{''} \mid
 i \in \{1,2,3,\dots,n_1\}, \; j \in \{1,2,3,\dots,n_2\}, \\
      &\tau^1_i = \tau^2_j, \mbox{ and } \tau^1_i, \tau^2_j \in \vars \} 
\end{aligned}
\]

\subsubsection*{Proof that theorem \ref{yn_theorem} holds for $\phi =
At[\phi_1, \phi_2]$, if it holds for $\phi = \phi_1$ and $\phi = \phi_2$}

I assume that $\phi_1, \phi_2 \in \ynforms$. By the syntax of \topl,
this implies that $\at[\phi_1, \phi_2] \in \ynforms$. I also assume
that $st \in \pts$, $\lambda$ is a \tsql value expression, $\lambda'$
and $\linit$ are as in the translation rule, $g^{db} \in G^{db}$,
$eval(st,\lambda, g^{db}) \in D_P^*$, $\corn{\phi_1} = \tup{\tau^1_1,
  \tau^1_2, \dots, \tau^1_{n_1}}$, $\corn{\phi_2} = \tup{\tau^2_1,
  \tau^2_2, \dots, \tau^2_{n_2}}$, and $\Sigma = trans(\at[\phi_1,
\phi_2], \lambda)$. From the definition of $\corn{\dots}$, it should
be easy to see that:
\[\corn{\at[\phi_1, \phi_2]} = \tup{\tau^1_1, \tau^1_2, \dots,
\tau^1_{n_1}, \tau^2_1, \tau^2_2, \dots, \tau^2_{n_2}}\]
Finally, I assume that theorem \ref{yn_theorem} holds for $\phi = \phi_1$ and
$\phi = \phi_2$. I need to show that: 
\begin{enumerate}
\item $\fcn(\Sigma) \subseteq \fcn(\lambda)$
\item $eval(st, \Sigma, g^{db}) \in \vrel(n_1 + n_2)$
\item $\tup{v^1_1, \dots, v^1_{n_1}, v^2_1, \dots, v^2_{n_2}; v_t} \in 
       eval(st, \Sigma, g^{db})$ iff for some $g \in G$: \\
   $\denot{M(st), g}{\tau^1_1} =
   f_D(v^1_1)$, \dots, $\denot{M(st), g}{\tau^1_{n_1}} = f_D(v^1_{n_1})$, \\
   $\denot{M(st), g}{\tau^2_1} =
   f_D(v^2_1)$, \dots, $\denot{M(st), g}{\tau^2_{n_2}} =
   f_D(v^2_{n_2})$, and \\
   $\denot{M(st), st, f_D(v_t), f_D(eval(st, \lambda, g^{db})), g}{
   \at[\phi_1, \phi_2]} = T$
\end{enumerate}

From the definition of $\lambda'$ it should be obvious that any free
column reference in $\lambda'$ is either the \sql{VALID($\alpha_1$)} of
$\lambda'$, or is situated within the $\lambda$ of 
$\lambda'$. By lemma \ref{fcn_lemma}, this implies that:
\begin{equation}
\label{atphi:1}
\fcn(\lambda') \subseteq \fcn(\text{\sql{VALID}}(\alpha_1)) 
\union \fcn(\lambda)
\end{equation}
The only correlation name that has a free column reference in
\sql{VALID$(\alpha_1)$} is $\alpha_1$. Hence,
$\fcn($\sql{VALID}$(\alpha_1)) = \{\alpha_1\}$, and \pref{atphi:1}
becomes \pref{atphi:2}.  
\begin{equation}
\label{atphi:2}
\fcn(\lambda') \subseteq \{\alpha_1\} \union \fcn(\lambda)
\end{equation}
The $\alpha_1$ of $\Sigma$ is generated by calling the correlation
names generator after $\lambda$ has been supplied.  Hence, $\alpha_1$
cannot appear in $\lambda$. The fact that $\alpha_1$ does not appear in
$\lambda$ implies that for every $v^1_1, v^1_2, \dots, v^1_{n_1} \in
D$ and $v^1_t \in D_T$:
\begin{equation}
\label{atphi:3}
eval(st, \lambda, (g^{db})^{\alpha_1}_{\tup{v^1_1, v^1_2, \dots,
v^1_{n_1}; v^1_t}}) = eval(st, \lambda, g^{db})
\end{equation}

The fact that for every $v^1_1,v^1_2, \dots, v^1_{n_1} \in D$ and
$v^1_t \in D_T$, \pref{atphi:3} holds, and the assumption that
$eval(st, \lambda, g^{db}) \in D_P^*$, imply that \pref{atphi:5} holds
for every $v^1_1, \dots, v^1_{n_1} \in D$ and $v^1_t \in D_T$. 
\begin{gather}
\label{atphi:5}
f_D(eval(st, \lambda, (g^{db})^{\alpha_1}_{\tup{v^1_1, v^1_2, \dots,
v^1_{n_1}; v^1_t}})) \in \periods^* 
\end{gather}
By the syntax of \topl, since $\lambda$ is a value expression,
$\lambda'$ is also a value expression. \pref{atphi:5} and the
definition of $\lambda'$ imply that \pref{atphi:7} and \pref{atphi:8}
hold for $v^1_1, \dots, v^1_{n_1} \in D$ and $v^1_t \in D_P$
($D_P \subseteq D_T$).
\begin{gather}
\label{atphi:7}
f_D(eval(st, \lambda', (g^{db})^{\alpha_1}_{\tup{v^1_1, v^1_2, \dots,
v^1_{n_1}; v^1_t}})) = \\
f_D(v^1_t) \intersect 
f_D(eval(st, \lambda, (g^{db})^{\alpha_1}_{\tup{v^1_1, v^1_2, \dots,
v^1_{n_1}; v^1_t}})) \notag \\
\label{atphi:8}
eval(st, \lambda',(g^{db})^{\alpha_1}_{\tup{v^1_1, v^1_2, \dots,
v^1_{n_1}; v^1_t}}) \in D_P^*
\end{gather}

By the syntax of \tsql, $\linit$ is a value expression, and 
by lemma \ref{linit_lemma}, \pref{atphi:31} -- \pref{atphi:34} hold. 
\begin{gather}
\label{atphi:31}
f_D(eval(st, \linit, g^{db})) = \pts \\
\label{atphi:32}
eval(st, \linit, g^{db}) \in D_P^* \\
\label{atphi:34}
\fcn(\linit) = \emptyset
\end{gather}

Let $\Sigma_1$ be the first embedded \sql{SELECT} statement in the
\sql{FROM} clause of $\Sigma$, i.e.\ $\Sigma_1 = trans(\phi_1,
\lambda)$.  From the hypothesis and the discussion above, $\phi_1 \in
\ynforms$, $st \in \pts$, $\corn{\phi_1} = \tup{\tau^1_1,
  \dots,\tau^1_{n_1}}$, $\linit$ is a \tsql value expression,
$eval(st, \linit, g^{db}) \in D_P^*$, and $g^{db} \in G^{db}$. From
theorem \ref{yn_theorem} for $\phi = \phi_1$ (according to the
hypothesis, theorem \ref{yn_theorem} holds for $\phi = \phi_1$), and
using \pref{atphi:31} and \pref{atphi:34},we get:
\begin{enumerate}
\item[$1^1.$] $\fcn(\Sigma_1) = \emptyset$
\item[$2^1.$] $eval(st, \Sigma_1, g^{db}) \in \vrel(n_1)$ 
\item[$3^1.$] $\tup{v^1_1, \dots, v^1_{n_1}; v^1_t} \in 
   eval(st, \Sigma_1, g^{db})$ iff for some $g_1 \in G$: \\
   $\denot{M(st), g_1}{\tau^1_1} =
   f_D(v^1_1)$, \dots, $\denot{M(st), g_1}{\tau^1_{n_1}} =
   f_D(v^1_{n_1})$, and \\ 
   $\denot{M(st), st, f_D(v_t^1), \pts, g_1}{\phi_1} = T$
\end{enumerate}
Let $\Sigma_2$ be the second embedded \sql{SELECT} statement in the
\sql{FROM} clause of $\Sigma$, i.e.\ $\Sigma_2 = trans(\phi_2,
\lambda')$. From the hypothesis, $\phi_2 \in \ynforms$, $st \in \pts$,
$g^{db} \in G^{db}$, and $\corn{\phi_2} = \tup{\tau^2_1, \dots,
  \tau^2_{n_2}}$. From the discussion above, $\lambda'$ is a \tsql
value expression. If $v^1_1,v^1_2, \dots, v^1_{n_1} \in D$ and $v^1_t
\in D_P$, then $(g^{db})^{\alpha_1}_{\tup{v^1_1, v^1_2, \dots,
    v^1_{n_1}; v^1_t}} \in G^{db}$ and (by \pref{atphi:8}) $eval(st,
\lambda', (g^{db})^{\alpha_1}_{\tup{v^1_1, v^1_2, \dots, v^1_{n_1};
    v^1_t}}) \in D_P^*$.  From theorem \ref{yn_theorem} for $\phi =
\phi_2$ (according to the hypothesis, theorem \ref{yn_theorem} holds
for $\phi = \phi_2$) we get the following. (The conditions
$v^1_1,v^1_2, \dots, v^1_{n_1} \in D$ and $v^1_t \in D_P$ do not
affect clause $1^2$.)
\begin{enumerate}
\item[$1^2.$] $\fcn(\Sigma_2) \subseteq \fcn(\lambda')$
\item[$2^2.$] If $v^1_1,v^1_2, \dots, v^1_{n_1} \in D$ and $v^1_t \in
D_P$, then: \\
$eval(st, \Sigma_2, (g^{db})^{\alpha_1}_{\tup{v^1_1,v^1_2, \dots,
v^1_{n_1}; v^1_t}}) \in \vrel(n_2)$ 
\item[$3^2.$] If $v^1_1,v^1_2, \dots, v^1_{n_1} \in D$ and $v^1_t \in
D_P$, then: \\
( \\
  $\tup{v^2_1, \dots, v^2_{n_2}; v^2_t} \in 
        eval(st, \Sigma_2, (g^{db})^{\alpha_1}_{\tup{v^1_1,v^1_2, \dots,
        v^1_{n_1}; v^1_t}})$ iff for some $g_2 \in G$: \\
  $\denot{M(st), g_2}{\tau^2_1} =
        f_D(v^2_1)$, \dots, $\denot{M(st), g_2}{\tau^2_{n_2}} =
        f_D(v^2_{n_2})$, and \\ 
  $\denot{M(st), st, f_D(v_t^2), f_D(eval(st, \lambda',
        (g^{db})^{\alpha_1}_{\tup{v^1_1,v^1_2, \dots,
        v^1_{n_1}; v^1_t}})), g_2}{\phi_2} = T$ \\
)
\end{enumerate}

\subsubsection*{Proof of clause 1}

The \sql{VALID($\alpha_2$)} in the \sql{VALID} clause of $\Sigma$ and
the $\alpha_1.1$, \dots, $\alpha_1.n_1$, $\alpha_2.1$, \dots,
$\alpha_2.n_2$ in the \sql{SELECT} clause of $\Sigma$ are not free
column references in $\Sigma$, because $\Sigma$ is a binding context
for all of them. Any column references of the form $\alpha_1.i$ or
$\alpha_2.j$ in the \sql{WHERE} clause of $\Sigma$ ($i \in
\{1,2,3,\dots,n_1\}$, $j \in \{1,2,3,\dots,n_2\}$; these column
references derive from $S$) are not free column references in $\Sigma$
for the same reason. $\Sigma$ contains no other column references (and
hence no other free column references), apart from those that possibly
appear within the $\Sigma_1$ and $\Sigma_2$ in the \sql{FROM} clause
of $\Sigma$. (The reader is reminded that $\Sigma_1 = trans(\phi_1,
\linit)$ and $\Sigma_2 = trans(\phi_2, \lambda')$.) By lemma
\ref{fcn_lemma}, this implies \pref{atphi:10}.  \pref{atphi:10},
clause $1^1$, and clause $1^2$ imply \pref{atphi:12}.  \pref{atphi:12}
and \pref{atphi:2} imply \pref{atphi:13}.
\begin{gather}
\label{atphi:10}
\fcn(\Sigma) \subseteq \fcn(\Sigma_1) \union \fcn(\Sigma_2) \\
\label{atphi:12}
\fcn(\Sigma) \subseteq \fcn(\lambda') \\
\label{atphi:13}
\fcn(\Sigma) \subseteq \{\alpha_1\} \union \fcn(\lambda)
\end{gather}

$\alpha_1$ cannot have a free column reference in $\Sigma$. The proof
follows. Let us assume that there is a free column reference $\zeta$
of $\alpha_1$ in $\Sigma$ ($\zeta$ has the form $\alpha_1.k$ or
\sql{VALID($\alpha_1$)}). Then, by the definition of free column
reference, (a) $\zeta$ is part of $\Sigma$. From the
translation rule, it is also the case that (b) $\alpha_1$ (the
correlation name of $\zeta$) is defined by the topmost \sql{FROM}
clause of $\Sigma$. 

If $\zeta$ is in the topmost \sql{FROM} clause of $\Sigma$, it can
only be in the $\Sigma_1$ (the $trans(\phi_1, \linit)$) or the
$\Sigma_2$ (the $trans(\phi_2, \lambda')$) of $\Sigma$. $\zeta$,
however, cannot be in $\Sigma_1$, because $\alpha_1$ (the correlation
name of $\zeta$) is generated after $\Sigma_1$ has been computed (see
the translation rule), and the correlation names of the generator cannot
be used before they have been generated (see section \ref{trans_rules}). 

So, if $\zeta$ is in the topmost \sql{FROM} clause of $\Sigma$,
$\zeta$ has to be in the $\Sigma_2$ of $\Sigma$, i.e.\ after the
definition of $\alpha_1$. Hence, (c) $\zeta$ is either not in the
topmost \sql{FROM} clause of $\Sigma$, or it is in the topmost
\sql{FROM} clause of $\Sigma$, but it follows the definition of
$\alpha_1$. (a), (b), and (c) imply that $\Sigma$ is a binding
context for $\zeta$, which implies that $\zeta$ is not a free column
reference in $\Sigma$. This is against the hypothesis. Hence,
$\alpha_1$ cannot have a free column reference in $\Sigma$, i.e.\
$\alpha_1 \not\in \fcn(\Sigma)$. This and \pref{atphi:13} imply that 
$\fcn(\Sigma) \subseteq \fcn(\lambda)$. Clause 1 has
been proven. 

\subsubsection*{Proof of clause 2}

When computing $eval(st, \Sigma, g^{db})$, the $\alpha_1$ of $\Sigma$
ranges over the tuples of the relation $nosubperiod(eval(st, \Sigma_1,
g^{db}))$.  By clause $2^1$, $eval(st, \Sigma_1, g^{db}) \in
\vrel(n_1)$. The definition of $nosubperiod$ (section \ref{new_pus})
implies that $nosubperiod(eval(st, \Sigma_1, g^{db}))$ is also a
valid-time relation of $n_1$ explicit attributes, and that all its
time-stamps are also time-stamps of $eval(st, \Sigma_1, g^{db})$,
i.e.\ elements of $D_P$. That is, $eval(st, \Sigma_1, g^{db})$
contains tuples of the form $\tup{v^1_1, \dots, v^1_{n_1}; v^1_t}$,
with $v^1_1, \dots, v^1_{n_1} \in D$ and $v^1_t \in D_P$. By clause
$2^2$, it is also the case that if $v^1_1, \dots, v^1_{n_1}\in D$ and
$v^1_t \in D_P$, then $eval(st, \Sigma_2,
(g^{db})^{\alpha_1}_{\tup{v^1_1, \dots, v^1_{n_1}; v^1_t}}) \in
\vrel(n_2)$. It should now be easy to see from the translation rule
that \pref{r16.8xx} holds.
\begin{eqnarray}
&& eval(st,\Sigma,g^{db}) = 
\label{r16.8xx} \\
&&\{\tup{v^1_1,\dots,v^1_{n_1},v^2_1,\dots,v^2_{n_2};v_t} \mid 
\text{ for some } v^1_t, \nonumber \\
&& \tup{v^1_1,\dots,v^1_{n_1};v^1_t} \in nosubperiod(eval(st,\Sigma_1,g^{db})), 
\nonumber \\
&& \tup{v^2_1,\dots,v^2_{n_2};v_t} \in
   eval(st,\Sigma_2,(g^{db})^{\alpha_1}_{\tup{v^1_1,\dots,v^1_{n_1};v^1_t}}),
   \mbox{ and}
\nonumber \\
&& \mbox{if } i \in \{1,2,3,\dots,n_1\}, \; j \in
   \{1,2,3,\dots,n_2\}, \; \tau^1_i, \tau^2_j \in \vars, 
\nonumber \\
&& \;\;\; \mbox{ and } \tau^1_i = \tau^2_j, \mbox{then } v^1_i = v^2_j \}
\nonumber 
\end{eqnarray}
$(g^{db})^{\alpha_1}_{\tup{v^1_1, \dots, v^1_{n_1}; v^1_t}}$ is used
in the fourth line of \pref{r16.8} instead of $g^{db}$, to capture the
fact that if there is any free column reference of $\alpha_1$ in
$\Sigma_2$, this has to be taken to refer to the tuple of
$nosubperiod(eval(st, \Sigma_1, g^{db}))$ to which $\alpha_1$ refers.

Using the definition of $nosubperiod$ (section \ref{new_pus}),
\pref{r16.8xx} becomes \pref{r16.8}. 
\begin{eqnarray}
&& eval(st,\Sigma,g^{db}) = 
\label{r16.8} \\
&&\{\tup{v^1_1,\dots,v^1_{n_1},v^2_1,\dots,v^2_{n_2};v_t} \mid 
\text{ for some } v^1_t, \nonumber \\
&& \tup{v^1_1,\dots,v^1_{n_1};v^1_t} \in eval(st,\Sigma_1,g^{db}), 
\nonumber \\
&& \mbox{there is no } \tup{v^1_1,\dots,v^1_{n_1};{v^1_t}'} \in
eval(st,\Sigma_1,g^{db}) 
\nonumber \\
&& \;\;\;\; \mbox{such that } f_D(v^1_t) \propsubper f_D({v^1_t}'), 
\nonumber \\
&& \tup{v^2_1,\dots,v^2_{n_2};v_t} \in
   eval(st,\Sigma_2,(g^{db})^{\alpha_1}_{\tup{v^1_1,\dots,v^1_{n_1};v^1_t}}),
   \mbox{ and}
\nonumber \\
&& \mbox{if } i \in \{1,2,3,\dots,n_1\}, \; j \in
   \{1,2,3,\dots,n_2\}, \; \tau^1_i, \tau^2_j \in \vars, 
\nonumber \\
&& \;\;\; \mbox{ and } \tau^1_i = \tau^2_j, \mbox{then } v^1_i = v^2_j \}
\nonumber 
\end{eqnarray}

\pref{r16.8} and the discussion above imply that for every
$\tup{v^1_1,\dots,v^1_{n_1},v^2_1,\dots,v^2_{n_2};v_t} \in eval(st,
\Sigma, g^{db})$, $\tup{v^2_1,\dots,v^2_{n_2};v_t} \in
eval(st,\Sigma_2,(g^{db})^{\alpha_1}_{\tup{v^1_1,\dots,v^1_{n_1};v^1_t}})
\in \vrel(n_2)$. This implies that $v_t \in D_P$.  \pref{r16.8} also
implies that $eval(st,\Sigma,g^{db})$ is a valid-time relation of $n_1
+ n_2$ explicit attributes. Hence, $eval(st, \Sigma, g^{db}) \in
\vrel(n_1 + n_2)$. Clause 2 has been proven.

\subsubsection*{Proof of clause 3}

I now prove clause 3. Using the definition of
$\denot{st,et,lt,g}{\at[\phi_1,\phi_2]}$ (section
\ref{at_before_after_op}), clause 3 becomes:
\begin{eqnarray}
&& \tup{v^1_1,\dots,v^1_{n_1},v^2_1,\dots,v^2_{n_2};v_t} 
   \in eval(st,\Sigma,g^{db}) \mbox{ iff for some } g \mbox{ and } et': 
\nonumber \\
&& g \in G \label{r16.10} \label{r16.b.10}\\
&& \denot{M(st),g}{\tau^1_1} = f_D(v^1_1), \; \dots, \; 
   \denot{M(st),g}{\tau^1_{n_1}} = f_D(v^1_{n_1}) 
   \label{r16.13} \label{r16.b.13}\\
&& \denot{M(st),g}{\tau^2_1} = f_D(v^2_1), \; \dots, \; 
   \denot{M(st),g}{\tau^2_{n_2}} = f_D(v^2_{n_2}) 
   \label{r16.13.5} \label{r16.b.13.5}\\
&& et' \in mxlpers(\{e \in \periods \mid
   \denot{M(st),st,e,\pts,g}{\phi_1} = T\}) 
   \label{r16.15} \label{r16.b.15}\\ 
&& \denot{M(st),st,f_D(v_t),f_D(eval(st, \lambda, g^{db}))
   \intersect et',g}{\phi_2} = T \label{r16.16} \label{r16.b.16}
\end{eqnarray}

I first prove the forward direction of clause 3. I assume
$\tup{v^1_1,\dots,v^1_{n_1},v^2_1,\dots,v^2_{n_2};v_t} \in eval(st,
\Sigma,g^{db})$. I need to prove that for some $g$ and $et'$,
\pref{r16.10} -- \pref{r16.16} hold. The assumption that
$\tup{v^1_1,\dots,v^1_{n_1},v^2_1,\dots,v^2_{n_2};v_t} \in
eval(st,\Sigma,g^{db})$ and \pref{r16.8} imply that for some $v^1_t$:
\begin{eqnarray}
&& \tup{v^1_1,\dots,v^1_{n_1};v^1_t} \in eval(st,\Sigma_1,g^{db}) 
   \label{r16.17} \label{r16.b.17}\\
&& \mbox{there is no } \tup{v^1_1,\dots,v^1_{n_1};{v^1_t}'} \in
   eval(st,\Sigma_1,g^{db}) 
   \label{atphi:20} \label{atphi:20b}\\
&& \;\;\;\; \mbox{such that } f_D(v^1_t) \propsubper f_D({v^1_t}'), 
   \nonumber \\
&& \tup{v^2_1,\dots,v^2_{n_2};v_t} \in
   eval(st,\Sigma_2,(g^{db})^{\alpha_1}_{\tup{v^1_1,\dots,v^1_{n_1};v^1_t}})
   \label{r16.18} \label{r16.b.18}\\
&& \mbox{if } i \in \{1,2,3,\dots,n_1\}, \; j \in
   \{1,2,3,\dots,n_2\}, \; \tau^1_i, \tau^2_j \in \vars, 
   \label{r16.19} \label{r16.b.19}\\
&& \;\;\; \mbox{ and } \tau^1_i = \tau^2_j, \mbox{then } v^1_i = v^2_j
\nonumber 
\end{eqnarray}
\pref{r16.17} and clause $3^1$ imply that for some
$g_1$:
\begin{eqnarray}
&& g_1 \in G \label{r16.20} \\
&& \denot{M(st),g_1}{\tau^1_1} = f_D(v^1_1), \; \dots, \; 
   \denot{M(st),g_1}{\tau^1_{n_1}} = f_D(v^1_{n_1}) \label{r16.22} \\
&& \denot{M(st),st,f_D(v_t^1),\pts,g_1}{\phi_1} = T \label{r16.24} 
\end{eqnarray}
\pref{r16.17} and clause $2^1$ imply that:
\begin{equation}
\label{r16.25.1}
v^1_1, \dots, v^1_{n_1} \in D \text{ and }
v^1_t \in D_P
\end{equation}
\pref{r16.18}, \pref{r16.25.1}, and clause $3^2$ imply that for some
$g_2$:
\begin{eqnarray}
&& g_2 \in G \label{r16.26} \\
&& \denot{M(st),g_2}{\tau^2_1} = f_D(v^2_1), \; \dots, \;  
   \denot{M(st),g_2}{\tau^2_{n_2}} = f_D(v^2_{n_2}) \label{r16.28} \\
&& \denot{M(st), st, f_D(v_t), f_D(eval(st, \lambda',
   (g^{db})^{\alpha_1}_{\tup{v^1_1,v^1_2, \dots,
   v^1_{n_1}; v^1_t}})), g_2}{\phi_2} = T
\label{r16.30}
\end{eqnarray}

I define the mapping $g : \vars \mapsto \objs$ as
follows:
\[ g(\beta) \defeq
   \left\{
   \begin{array}{l}
   g_1(\beta), \mbox{ if for some } i \in \{1,2,3,\dots,n_1\},
      \; \beta = \tau^1_i  \\
   g_2(\beta), \mbox{ if for some } j \in \{1,2,3,\dots,n_2\},
      \; \beta = \tau^2_j  \\
   o, \mbox{ otherwise}
   \end{array}
   \right. 
\]
where $o$ is a particular element of \objs, chosen
arbitrarily. \pref{r16.10} follows from lemma \ref{l5}, the definition of
$g$, \pref{r16.19}, \pref{r16.22}, and \pref{r16.28}. 
I set $et'$ as in \pref{atphi:22}, and show that \pref{r16.13} --
\pref{r16.16} hold.
\begin{equation}
\label{atphi:22}
et' = f_D(v^1_t)
\end{equation}

The definition of $g$ implies that for every variable $\beta$ among
$\tau^1_1, \dots, \tau^1_{n_1}$, $g(\beta) = g_1(\beta)$. The
assumption that $\corn{\phi_1} = \tup{\tau^1_1, \dots, \tau^1_{n_1}}$
and the definition of $\corn{\phi_1}$ imply that all the variables of
$\phi_1$ are among $\tau^1_1, \dots, \tau^1_{n_1}$. Hence, for every
variable $\beta$ in $\phi_1$, $g(\beta) = g_1(\beta)$. Similarly, the
definition of $g$ and the assumption that $\corn{\phi_2} =
\tup{\tau^2_1, \dots, \tau^2_{n_2}}$ imply that for every variable
$\beta$ in $\phi_2$, $g(\beta) = g_2(\beta)$. 

\pref{r16.13} follows from lemma \ref{l3}, the
assumption that $\tup{\tau^1_1, \dots, \tau^1_{n_1}} = \corn{\phi_1}$,
\pref{r16.22}, and the fact that $g$ and $g_1$ assign the same values
to all the variables of $\phi_1$. The proof of \pref{r16.13.5} is very
similar.

I now prove \pref{r16.16}. \pref{atphi:3} and \pref{atphi:7} (which
hold for $v^1_1, \dots, v^1_{n_1} \in D$ and $v^1_t \in D_P$) and
\pref{r16.25.1} imply \pref{atphi:21}. \pref{r16.30}, \pref{atphi:21},
\pref{atphi:22}, and the fact that $g$ and $g_2$ assign the same
values to all the variables of $\phi_2$ (see above), imply
\pref{r16.16}.
\begin{eqnarray}
&& \label{atphi:21}
   f_D(eval(st, \lambda',
      (g^{db})^{\alpha_1}_{\tup{v^1_1,v^1_2, \dots, v^1_{n_1}; v^1_t}})) =
   f_D(eval(st, \lambda, g^{db})) \intersect f_D(v^1_t)
\end{eqnarray}
It remains to show \pref{r16.15}. \pref{r16.17},
clause $2^1$, and \pref{atphi:22} imply \pref{atphi:100}. 
\begin{equation}
\label{atphi:100}
et' \in \periods
\end{equation}
\pref{atphi:100}, \pref{atphi:22}, \pref{r16.24}, and the fact that
$g$ and $g_2$ assign the same values to all the variables of $\phi_1$
imply \pref{r16.200}. 
\begin{equation}
  \label{r16.200}
  et' \in \{e \in \periods \mid
  \denot{M(st),st,e,\pts,g}{\phi_1} = T \} 
\end{equation}
To prove \pref{r16.15} it remains to prove that there is no 
$et''$ that satisfies both \pref{r16.201} and \pref{r16.206}. 
\begin{gather}
  \label{r16.201}
  et'' \in \{e \in \periods \mid
  \denot{M(st),st,e,\pts,g}{\phi_1} = T \} \\
  \label{r16.206}
  et' \propsubper et''
\end{gather}
Let us assume that for some $et''$, \pref{r16.201} and \pref{r16.206}
hold. \pref{r16.201} and the fact that $g$ and $g_1$ assign the same
values to the variables of $\phi_1$ imply that:
\begin{eqnarray}
&& \denot{M(st),st,et'',\pts,g_1}{\phi'} = T \label{r16.202} \\
&& et'' \in \periods \label{r16.203}
\end{eqnarray}
I set ${v^1_t}' = \fdi(et'')$, which implies \pref{r16.205}.
\pref{r16.20}, \pref{r16.22}, \pref{r16.205}, \pref{r16.202}, and
clause $3^1$ imply \pref{r16.205.1}. \pref{r16.206}, \pref{atphi:22},
and \pref{r16.205} imply \pref{r16.207}.
\begin{gather}
  \label{r16.205}
  et'' = f_D({v^1_t}') \\
  \label{r16.205.1}
  \tup{v_1,\dots,v_n;{v^1_t}'} \in eval(st,\Sigma_1,g^{db}) \\
  \label{r16.207}
  f_D(v^1_t) \propsubper f_D({v^1_t}')
\end{gather}
\pref{r16.205.1} and \pref{r16.207} are against \pref{atphi:20}.
Therefore, there can be no $et''$ that satisfies \pref{r16.201} and
\pref{r16.206}. \pref{r16.15} and the forward direction of clause 3
have been proven.

\bigskip

I now prove the backwards direction of clause 3. I assume that for
some $g$ and $et'$, \pref{r16.b.10} -- \pref{r16.b.16} hold.  I must
show that $\tup{v^1_1,\dots,v^1_{n_1},v^2_1,\dots,v^2_{n_2};v_t}
\in eval(st,\Sigma,g^{db})$. According to \pref{r16.8}, it is enough
to prove that for some $v^1_t$, \pref{r16.b.17} -- \pref{r16.b.19} hold.

I first prove \pref{r16.b.17}. I set $v^1_t = \fdi(et')$, which
implies \pref{r16.101}. \pref{r16.b.15} implies \pref{atphi:200}.
Clause $3^1$, \pref{r16.b.10}, \pref{r16.b.13}, \pref{r16.101}, and
\pref{atphi:200} imply \pref{r16.b.17}.
\begin{gather}
  \label{r16.101}
  et' = f_D(v^1_t) \\
  \label{atphi:200}
  \denot{M(st),st,et',\pts,g}{\phi_1} = T
\end{gather}

I now prove \pref{r16.b.18}. \pref{r16.b.17} (proven above) and
clause $2^1$ (also proven above) imply that:
\begin{equation}
\label{r16.787.1}
v^1_1,\dots,v^1_{n_1} \in D \text{ and } v^1_t \in D_P
\end{equation}
\pref{atphi:3} and \pref{atphi:7} (which hold for
$v^1_1, \dots, v^1_{n_1} \in D$ and $v^1_t \in D_P$), 
\pref{r16.787.1}, and \pref{r16.101} imply \pref{atphi:21b}. Clause $3^2$, \pref{r16.787.1}, \pref{r16.b.10}, \pref{r16.b.13.5},
\pref{atphi:21b}, and \pref{r16.b.16} imply \pref{r16.b.18}.
\begin{eqnarray}
  && \label{atphi:21b} f_D(eval(st, \lambda',
  (g^{db})^{\alpha_1}_{\tup{v^1_1,v^1_2, \dots, v^1_{n_1}; v^1_t}})) =
  f_D(eval(st, \lambda, g^{db})) \intersect et'
\end{eqnarray}

I now prove \pref{r16.b.19}. Let us assume that $i \in
\{1,2,3,\dots,n_1\}$, $j \in \{1,2,3,\dots,n_2\}$, $\tau^1_i,
\tau^2_j \in \vars$, and $\tau^1_i = \tau^2_j$. From
\pref{r16.b.13} we get \pref{r16.110}, and from \pref{r16.b.13.5} we
get \pref{r16.111}. 
\begin{gather}
  \label{r16.110}
  \denot{M(st),g}{\tau^1_j} = f_D(v^1_i) \\
  \label{r16.111}
  \denot{M(st),g}{\tau^2_j} = f_D(v^2_j)
\end{gather}
\pref{r16.110}, \pref{r16.111}, and the hypothesis that
$\tau^1_i = \tau^2_j$ imply that $f_D(v^1_i) = f_D(v^2_j)$. This in
turn implies that $\fdi(f_D(v^1_i)) = \fdi(f_D(v^2_j))$, 
i.e.\ that $v_i^1 = v_j^2$. \pref{r16.b.19} has been proven. 

It remains to prove \pref{atphi:20b}. Let us assume that there is a
tuple $\tup{v^1_1,\dots,v^1_n;{v^1_t}'}$, such that \pref{r16.400} and
\pref{r16.401} hold.
\begin{gather}
  \label{r16.400}
  \tup{v^1_1,\dots,v^1_n;{v^1_t}'} \in eval(st,\Sigma_1,g^{db}) \\
  \label{r16.401}
  f_D(v^1_t) \propsubper f_D({v^1_t}')
\end{gather}
\pref{r16.400} and clause $3^1$ imply that for some $g'$:
\begin{eqnarray}
&& g' \in G \label{r16.402} \\
&& \denot{M(st),g'}{\tau_1} = f_D(v^1_1), \dots,
   \denot{M(st),g'}{\tau_n} = f_D(v^1_{n_1}) \label{r16.404} \\
&& \denot{M(st),st,f_D({v^1_t}'),\pts,g'}{\phi_1} = T \label{r16.406}
\end{eqnarray}
Lemma \ref{l7}, \pref{r16.b.10}, \pref{r16.402}, the assumptions that
$\phi_1 \in \ynforms$ and $\corn{\phi_1} = \tup{\tau^1_1, \dots,
  \tau^1_{n_1}}$, \pref{r16.b.13}, and \pref{r16.404} imply that $g$
and $g'$ assign the same values to the variables of $\phi_1$. This
implies \pref{r16.412}. \pref{r16.406} and \pref{r16.412} imply
\pref{r16.413}.
\begin{gather}
  \label{r16.412}
  \denot{M(st),st,f_D({v^1_t}'),\pts,g'}{\phi_1} =
  \denot{M(st),st,f_D({v^1_t}'),\pts,g}{\phi_1} \\
  \label{r16.413}
  \denot{M(st),st,f_D({v^1_t}'),\pts,g}{\phi_1} = T
\end{gather}
I set $et''$ as in \pref{atphr:85}. \pref{r16.401}, \pref{r16.101},
and \pref{atphr:85} imply \pref{r16.414}. \pref{r16.413} and
\pref{atphr:85} imply \pref{atphr:97}.
\begin{gather}
  \label{atphr:85}
  et'' = f_D({v^1_t}') \\
  \label{r16.414}
  et' \propsubper et'' \\
  \label{atphr:97}
  \denot{M(st),st,et'',\pts,g}{\phi_1} = T
\end{gather}
\pref{r16.414} and \pref{atphr:97} are against \pref{r16.b.15},
because \pref{r16.b.15} and the definition of $mxlpers$ (section
\ref{temporal_ontology}) imply that there is no $et''$, such that $et'
\propsubper et''$ and $\denot{M(st),st,et'',\pts,g}{\phi_1} = T$.
Therefore, there can be no tuple
$\tup{v^1_1,\dots,v^1_{n_1};{v^1_t}'}$ such that \pref{r16.400} and
\pref{r16.401} hold. \pref{atphi:20b} and the backwards direction of
clause 3 have been proven.


\subsection{$\mathit{Before}[\phi_1, \phi_2]$}

\subsubsection*{Translation rule}

If $\phi_1, \phi_2 \in \ynforms$ and $\lambda$ is a \tsql value
expression, then: 

$trans(\before[\phi_1, \phi_2], \lambda) \defeq$\\
\sql{(}\select{SELECT DISTINCT 
                  $\alpha_1$.1, $\alpha_1$.2, \dots, $\alpha_1$.$n_1$
                  $\alpha_2$.1, $\alpha_2$.2, \dots, $\alpha_2$.$n_2$ \\
               VALID VALID($\alpha_2$) \\
               FROM $trans(\phi_1,\linit)$(NOSUBPERIOD) AS $\alpha_1$, \\
               \ \ \ \ \ $trans(\phi_2, \lambda')$ AS $\alpha_2$) \\
               WHERE \dots \\
               \ \ AND \dots \\
               \ \ \vdots \\
               \ \ AND \dots)}

$\linit$ is as in section \ref{formulation}, $\lambda'$
is the expression \sql{INTERSECT(PERIOD(TIMESTAMP 'beginning',
  BEGIN(VALID($\alpha_1$)) - INTERVAL '1' $\chi$), $\lambda$)}, and
$\chi$ is the \tsql name of the granularity of chronons. $n_1$ and $n_2$
are the lengths of $\corn{\phi_1}$ and $\corn{\phi_2}$ respectively. 
Each time the translation rule is used, $\alpha_1$ and $\alpha_2$ are two
new different correlation names, obtained by calling the correlation
names generator after $\lambda$ has been supplied. $\alpha_1$ is
generated after $trans(\phi_1, \linit)$ has been computed, and before
computing $trans(\phi_2, \lambda')$. Assuming
that $\corn{\phi_1} = \tup{\tau^1_1, \dots, \tau^1_{n_1}}$ and
$\corn{\phi_2} = \tup{\tau^2_1, \dots, \tau^2_{n_2}}$, the ``\dots''s
in the \sql{WHERE} clause are all the strings in the set $S$ of
section \ref{atphi_rule}. 

\subsubsection*{Proof that theorem \ref{yn_theorem} holds for $\phi =
\mathit{Before}[\phi_1, \phi_2]$, if it holds for $\phi = \phi_1$ and
$\phi = \phi_2$} 

The proof is very similar to that of section \ref{atphi_rule}. 


\subsection{$\mathit{After}[\phi_1, \phi_2]$}

\subsubsection*{Translation rule}

If $\phi_1, \phi_2 \in \ynforms$ and $\lambda$ is a \tsql value
expression, then: 

$trans(\after[\phi_1, \phi_2], \lambda) \defeq$\\
\sql{(}\select{SELECT DISTINCT 
                  $\alpha_1$.1, $\alpha_1$.2, \dots, $\alpha_1$.$n_1$
                  $\alpha_2$.1, $\alpha_2$.2, \dots, $\alpha_2$.$n_2$ \\
               VALID VALID($\alpha_2$) \\
               FROM $trans(\phi_1,\linit)$(NOSUBPERIOD) AS $\alpha_1$, \\
               \ \ \ \ \ $trans(\phi_2, \lambda')$ AS $\alpha_2$) \\
               WHERE \dots \\
               \ \ AND \dots \\
               \ \ \vdots \\
               \ \ AND \dots)}

$n_1, n_2$ are the lengths of $\corn{\phi_1}$ and $\corn{\phi_2}$
respectively.  
$\linit$ is as in section \ref{formulation}, $\lambda'$ is the expression
\sql{INTERSECT(PERIOD(END(VALID($\alpha_1$)) + INTERVAL '1' $\chi$, TIMESTAMP
'forever'), $\lambda$)}, and $\chi$ is the \tsql name of the
granularity of chronons. 
Each time the translation rule is used, $\alpha_1$ and $\alpha_2$ are two
new different correlation names, obtained by calling the correlation
names generator after $\lambda$ has been supplied. $\alpha_1$ is
generated after $trans(\phi_1, \linit)$ has been computed, and before
computing $trans(\phi_2, \lambda')$. Assuming
that $\corn{\phi_1} = \tup{\tau^1_1, \dots, \tau^1_{n_1}}$ and
$\corn{\phi_2} = \tup{\tau^2_1, \dots, \tau^2_{n_2}}$, the ``\dots''s
in the \sql{WHERE} clause are all the strings in the set $S$ of
section \ref{atphi_rule}. 

\subsubsection*{Proof that theorem \ref{yn_theorem} holds for $\phi =
\mathit{After}[\phi_1, \phi_2]$, if it holds for $\phi = \phi_1$ and
$\phi = \phi_2$} 

The proof is very similar to that of section \ref{atphi_rule}. 


\section{Translation rules for wh-formulae and proof of theorem
  \ref{wh_theorem}} 
\label{wh_rules} 

This section lists the translation rules for wh-formulae. These rules
have to satisfy theorem \ref{wh_theorem} (see section
\ref{formulation}). There are two translation rules for wh-formulae.
They correspond to the cases where the \topl formula $\phi$ to be
translated belongs to $\whforms_1$ or $\whforms_2$ (see
section \ref{top_syntax}). Each rule is followed by a proof that
theorem \ref{wh_theorem} holds if $\phi \in \whforms_1$ or $\phi \in
\whforms_2$ respectively.

As explained in section \ref{trans_rules}, the translation rules for
wh-formulae define $trans(\phi, \lambda)$ only for $\lambda = \linit$.
The values of $trans(\phi, \lambda)$ for $\phi \in \whforms$ and
$\lambda \not= \linit$ are not used anywhere (they are also not
examined by theorem \ref{wh_theorem}) and can be chosen arbitrarily.

\subsection{$?\beta_1 \; ?\beta_2 \; ?\beta_3 \; \dots \; ?\beta_n \; \phi'$}
\label{wh1_rule}

\subsubsection*{Translation rule}

If $\beta_1, \beta_2, \dots, \beta_n \in \vars$, $\phi' \in \ynforms$,
and $\linit$ is as in section \ref{formulation}, then:

$trans(?\beta_1 \; ?\beta_2 \; ?\beta_3 \; \dots \; ?\beta_n \; \phi',
\linit) \defeq$\\
\sql{(}\select{SELECT DISTINCT SNAPSHOT
                  $\alpha.\omega_1$, $\alpha.\omega_2$, 
                  $\alpha.\omega_3$, \dots, $\alpha.\omega_n$ \\
               FROM $trans(\phi', \linit)$ AS $\alpha$)}

Each time the translation rule is used, $\alpha$ is a new correlation
name, obtained by calling the correlation names
generator. For every $i \in \{1,2,3,\dots,n\}$, 
\[
\omega_i = min(\{j \mid j \in \{1,2,3,\dots,k\} \mbox{ and } 
  \tau_j = \beta_i \})
\]
where $\tup{\tau_1, \tau_2, \tau_3, \dots, \tau_k} =
\corn{\phi'}$. That is, the first position (from left to right) where
$\beta_i$ appears in $\tup{\tau_1, \dots, \tau_k}$ is the $\omega_i$-th one. 

\subsubsection*{Proof that theorem \ref{wh_theorem} holds for $\phi =
?\beta_1 \; ?\beta_2 \; ?\beta_3 \; \dots \; ?\beta_n \; \phi'$}

I assume that $\beta_1, \beta_2, \dots, \beta_n \in \vars$ and $\phi'
\in \ynforms$. By the syntax of \topl, this implies that $?\beta_1 \;
?\beta_2 \; ?\beta_3 \; \dots \; ?\beta_n \; \phi' \in \whforms$. I
also assume that $st \in \pts$, $\linit$ is as in section
\ref{formulation}, $\Sigma = trans(?\beta_1 \; ?\beta_2 \;
?\beta_3 \; \dots \; ?\beta_n \; \phi', \linit)$, and that
$\corn{\phi'} = \tup{\tau_1, \dots, \tau_k}$ (as in the translation
rule). I need to show that:
\begin{enumerate}
\item $\fcn(\Sigma) = \emptyset$

\item $eval(st, \Sigma) \in \srel(n)$

\item $\{\tup{f_D(v_1), \dots, f_D(v_n)} \mid 
         \tup{v_1, \dots, v_n} \in eval(st, \Sigma)\} =$ \\
      $\denot{M(st), st}
        {?\beta_1 \; ?\beta_2 \; ?\beta_3 \; \dots \; ?\beta_n \; \phi'}$
\end{enumerate}

According to the syntax of \tsql, $\linit$ is a value expression. Let
$g^{db}$ be an arbitrary member of $G^{db}$. By lemma
\ref{linit_lemma}, \pref{whr:1} -- \pref{whr:4} hold.
\begin{gather}
\label{whr:1}
f_D(eval(st, \linit, g^{db})) = \pts \\
\label{whr:2}
eval(st, \linit, g^{db}) \in D_P^* \\
\label{whr:4}
\fcn(\linit) = \emptyset
\end{gather}

Let $\Sigma'$ be the \sql{SELECT} statement in the \sql{FROM} clause
of $\Sigma$, i.e.\ $\Sigma' = trans(\phi', \linit)$. From the
hypothesis and the discussion above, $\phi' \in \ynforms$, $st \in
\pts$, $\corn{\phi'} = \tup{\tau_1, \dots, \tau_k}$, $g^{db} \in
G^{db}$, $\linit$ is a \tsql value expression, and $eval(st, \linit,
g^{db}) \in D_P^*$.  Then, from theorem \ref{yn_theorem} (already
proven), and using \pref{whr:1} and \pref{whr:4}, we get:
\begin{enumerate}
\item[$1'.$] $\fcn(\Sigma') = \emptyset$
\item[$2'.$] $eval(st, \Sigma') \in \vrel(k)$
\item[$3'.$] $\tup{v'_1, \dots, v'_k; v'_t} \in 
       eval(st, \Sigma')$ iff for some $g \in G$: \\
   $\denot{M(st), g}{\tau_1} =
   f_D(v'_1)$, \dots, $\denot{M(st), g}{\tau_k} = f_D(v'_k)$, and 
   $\denot{M(st), st, f_D(v'_t), \pts, g}{\phi'} = T$
\end{enumerate}
In clauses $2'$ and $3'$ I have not included $g^{db}$ among the arguments
of $eval(st, \Sigma')$, because according to clause $1'$,
$\fcn(\Sigma') = \emptyset$ (see comments in section
\ref{additional_tsql2}). 

\subsubsection*{Proof of clause 1}

The $\alpha.\omega_1$, \dots, $\alpha.\omega_n$ in the \sql{SELECT}
clause of $\Sigma$ are not free column references in $\Sigma$, because
$\Sigma$ is a binding context for all of them. $\Sigma$ contains no
other column references (and hence no other free column references),
apart from those that possibly appear within $\Sigma'$. By lemma
\ref{fcn_lemma}, this implies \pref{whr:3}. Clause $1'$ and
\pref{whr:3} imply clause 1.
\begin{equation}
\label{whr:3}
\fcn(\Sigma) \subseteq \fcn(\Sigma')
\end{equation}

\subsubsection*{Proof of clause 2}

When computing $eval(st, \Sigma)$, the $\alpha$ of $\Sigma$ ranges
over the tuples of $eval(st, \Sigma')$. (I do not include in the
arguments of $eval(st, \Sigma)$ and $eval(st, \Sigma')$ the
assignment to the correlation names, because according to clauses
1 and $1'$, $\fcn(\Sigma) = \fcn(\Sigma') = \emptyset$.) 
By clause $2'$, $eval(st,\Sigma') \in \vrel(k)$. Hence, $\alpha$
ranges over tuples of the form $\tup{v'_1, \dots, v'_k;
v'_t} \in eval(st, \Sigma')$. 

The syntax of \topl (section \ref{top_syntax}) and the fact that
$?\beta_1 \; \dots \; ?\beta_n \; \phi' \in \whforms$ imply that for
every $i \in \{1,2,3,\dots,n\}$, $\beta_i$ occurs at least once within
$\phi'$. This and the definition of $\corn{\dots}$ imply that
$\beta_i$ occurs at least once within $\corn{\phi'}$, which according
to the hypothesis is $\tup{\tau_1,\dots,\tau_k}$. Hence, for every $i
\in \{1,2,3,\dots,n\}$, the set $\{j \mid j \in \{1,2,3,\dots,k\}
\mbox{ and } \tau_j = \beta_i \}$ in the definition of $\omega_i$ (see
the translation rule) is not empty, and $\omega_i \in
\{1,2,3,\dots,k\}$. Assuming that $\alpha$ refers to a tuple
$\tup{v'_1, \dots, v'_k; v'_t} \in eval(st, \Sigma')$, for every $i
\in \{1,2,3,\dots,n\}$, the $\alpha.\omega_i$ of $\Sigma$ (see the
translation rule) refers to $v'_{\omega_i}$, i.e.\ the $\omega_i$-th
(from left to right) among $v'_1, v'_2, \dots, v'_k$.

It should now be easy to see from the translation rule that:
\begin{eqnarray}
 eval(st,\Sigma) &=& 
  \{\tup{v'_{\omega_1}, v'_{\omega_2}, v'_{\omega_3},
    \dots, v'_{\omega_n}} \mid 
\label{t2.10} \\
&&   \tup{v'_1,v'_2,v'_3,\dots,v'_k;v'_t} \in eval(st,\Sigma'), \mbox{ and}
\nonumber \\
&&   \mbox{for every } i \in \{1,2,3,\dots,n\},
\nonumber \\
&& \;\;\;\;\;\;
       \omega_i = min(\{j \mid j \in \{1,2,3,\dots,k\} \mbox{ and }
                        \tau_j = \beta_i\}) \;\;\;\}
\nonumber 
\end{eqnarray}
\pref{t2.10} implies that $eval(st, \Sigma)$ is a snapshot relation of
$n$ attributes. Clause 2 has been proven.

\subsubsection*{Proof of clause 3} 

By the definitions of $\denot{M(st), st}{\phi}$ and
$\denot{M(st),st,et,lt,g}{\phi}$ for $\phi \in \whforms_1$ (section
\ref{denotation}), clause 3 becomes:
\begin{eqnarray}
&& \{\tup{f_D(v_1), f_D(v_2), \dots, f_D(v_n)} \mid 
     \tup{v_1,v_2,\dots,v_n} \in eval(st,\Sigma) \}
\label{t2.2} \\
&& = \bigcup_{g \in G, et \in \periods} 
     \{\tup{g(\beta_1),g(\beta_2), \dots, g(\beta_n)}
       \mid \denot{M(st),st,et,\pts,g}{\phi'} = T\}
\nonumber 
\end{eqnarray}
I prove \pref{t2.2} by proving that its left-hand side (LHS) is a
subset of its right-hand side (RHS), and that its RHS is a subset of
its LHS. I start with the proof that the LHS of \pref{t2.2} is a
subset of the RHS. I assume that for some $v_1,\dots,v_n \in D$,
$\tup{f_D(v_1),\dots,f_D(v_n)}$ is an element of the LHS of
\pref{t2.2}, i.e.\ that \pref{t2.7} holds.
\begin{equation}
  \label{t2.7}
  \tup{v_1,\dots,v_n} \in eval(st,\Sigma)
\end{equation}
I need to prove that $\tup{f_D(v_1),\dots,f_D(v_n)}$ is an element of
the RHS of \pref{t2.2}, i.e.\ that for some $g$ and $et$:
\begin{eqnarray}
&& g \in G \label{t2.19} \\
&& et \in \periods \label{t2.20} \\
&& f_D(v_1) = g(\beta_1), \dots, f_D(v_n) =
   g(\beta_n) \label{t2.21} \\
&& \denot{M(st),st,et,\pts,g}{\phi'} = T \label{t2.22}
\end{eqnarray}

\pref{t2.7} and \pref{t2.10} imply that for some $v'_1$,
\dots, $v'_k$, $v'_t$:
\begin{eqnarray}
&& \tup{v'_1,\dots,v'_k;v'_t} \in eval(st,\Sigma') \label{t2.11} \\
&& v_1 = v'_{\omega_1}, \; v_2 = v'_{\omega_2}, \; \dots, \; v_n =
   v'_{\omega_n} \label{t2.12} \\
&& \mbox{for every } i \in \{1,2,3,\dots,n\},
\label{t2.13} \\
&& \;\;\;\;\;\; \omega_i = min(\{j \mid j \in \{1,2,3,\dots,k\} \mbox{
  and } \tau_j = \beta_i\}) 
\nonumber 
\end{eqnarray}
\pref{t2.11} and clause $3'$ imply that for some $g$:
\begin{eqnarray}
&& g \in G \label{t2.14} \\
&& \denot{M(st),g}{\tau_1} = f_D(v'_1), \dots, 
   \denot{M(st),g}{\tau_k} = f_D(v'_k) \label{t2.16} \\
&& \denot{M(st),st,f_D(v_t'),\pts,g}{\phi'} = T \label{t2.18} 
\end{eqnarray}
\pref{t2.19} is the same as \pref{t2.14}, which is known to be true. I
set $et$ as in \pref{whr:11}. Then, \pref{t2.22} follows from
\pref{t2.18}. \pref{t2.11} and clause $2'$ imply \pref{whr:10}.
\pref{t2.20} follows from \pref{whr:11} and \pref{whr:10}.
\begin{gather}
\label{whr:11}
et = f_D(v_t') \\
\label{whr:10}
v_t' \in D_P
\end{gather}

It remains to prove \pref{t2.21}. For every $i \in \{1,2,3,\dots,n\}$,
since $\beta_i \in \vars$, according to the semantics of \topl,
\pref{t2.23} holds. \pref{t2.13} implies \pref{t2.23.1} and \pref{t2.23.2}. 
\begin{gather}
  g(\beta_i) = \denot{M(st),g}{\beta_i} \label{t2.23} \\
  \tau_{\omega_i} = \beta_i \label{t2.23.1} \\
  \omega_i \in \{1,2,3,\dots,k\} \label{t2.23.2} 
\end{gather}
\pref{t2.23.1} and \pref{t2.23} imply \pref{t2.23.3}. \pref{t2.23.2}
and \pref{t2.16} imply \pref{t2.23.4}. \pref{t2.23.3} and
\pref{t2.23.4} imply \pref{t2.30}.
\begin{gather}
  \label{t2.23.3}
  g(\beta_i) = \denot{M(st),g}{\tau_{\omega_i}} \\
  \label{t2.23.4}
  \denot{M(st),g}{\tau_{\omega_i}} = f_D(v'_{\omega_i}) \\
  \label{t2.30}
  g(\beta_i) = f_D(v'_{\omega_i})
\end{gather}
\pref{t2.12} and the assumption that $i \in
\{1,2,3,\dots,n\}$, imply \pref{t2.30.1}. \pref{t2.30.1} and
\pref{t2.30} imply \pref{t2.30.1.last}.  
\begin{gather}
  \label{t2.30.1}
  v'_{\omega_i} = v_i \\
  \label{t2.30.1.last}
  g(\beta_i) = f_D(v_i)
\end{gather}
I have proven that for every $i \in
\{1,2,3,\dots,n\}$, $f_D(v_i) = g(\beta_i)$. Therefore, 
\pref{t2.21} holds. The proof that the LHS of \pref{t2.2} is a subset
of the RHS has been completed. 

\bigskip

I now prove that the RHS of \pref{t2.2} is a subset of the LHS. I
assume that for some $g \in G$,
$\tup{g(\beta_1),g(\beta_2),\dots,g(\beta_n)}$ is an element of the
RHS of \pref{t2.2}, i.e.\ that for some $g$ and $et$, \pref{t2.51} --
\pref{t2.53} hold. 
\begin{eqnarray}
&& g \in G \label{t2.51} \\
&& et \in \periods \label{t2.52} \\
&& \denot{M(st),st,et,\pts,g}{\phi'} = T \label{t2.53}
\end{eqnarray}
I need to prove that $\tup{g(\beta_1),g(\beta_2),\dots,g(\beta_n)}$ is
an element of the LHS of \pref{t2.2}, i.e.\ that for some
$v_1,\dots,v_n \in D$:
\begin{eqnarray}
&& g(\beta_1) = f_D(v_1), \; \dots, \; g(\beta_n) =
  f_D(v_n) \label{t2.54} \\
&& \tup{v_1,\dots,v_n} \in eval(st,\Sigma)  \label{t2.55}
\end{eqnarray}
I set $v_1, \dots, v_n$ as in \pref{t2.64}, which implies that
$v_1, \dots, v_n \in D$ and that \pref{t2.54} holds. 
\begin{equation}
  \label{t2.64}
  v_1 = \fdi(g(\beta_1)), \; \dots, \; 
  v_n = \fdi(g(\beta_n))
\end{equation}

It remains to prove \pref{t2.55}.  By \pref{t2.10}, it is enough to
prove that for some $v'_1$, $v'_2$, \dots, $v'_k$, $v_t'$, $\omega_1$,
$\omega_2$, \dots, $\omega_n$:
\begin{eqnarray}
&& \tup{v'_1,\dots,v'_k;v'_t} \in eval(st,\Sigma') \label{t2.56} \\
&& v_1 = v'_{\omega_1}, \; \dots, \; v_n = v'_{\omega_n} \label{t2.57} \\
&& \mbox{for every }i \in \{1,2,3,\dots,n\}, 
\label{t2.58} \\
&& \;\;\;\;\;\; \omega_i = min(\{j \mid
      j \in \{1,2,3,\dots,k\} \mbox{ and } \tau_j = \beta_i\})
\nonumber 
\end{eqnarray}
I set $v'_1$, $v'_2$, \dots, $v'_k$ as in \pref{t2.59}, which implies
\pref{t2.60}. I also set $v'_t = \fdi(et)$, which implies
\pref{t2.61}. Clause $3'$, \pref{t2.51}, \pref{t2.60}, \pref{t2.61},
and \pref{t2.53} imply \pref{t2.56}.
\begin{gather}
  \label{t2.59}
  v'_1 = \fdi(\denot{M(st),g}{\tau_1}), \; \dots, \; 
  v'_k = \fdi(\denot{M(st),g}{\tau_k}) \\
  \label{t2.60}
  \denot{M(st),g}{\tau_1} = f_D(v'_1), \; \dots, \; 
  \denot{M(st),g}{\tau_k} = f_D(v'_k) \\
  \label{t2.61}
  et = f_D(v'_t)
\end{gather}

For every $i \in \{1,2,3,\dots,n\}$, I set $\omega_i$ as in
\pref{t2.58}. It remains to prove \pref{t2.57}.  \pref{t2.64} implies
that for every $i \in \{1,2,3,\dots,n\}$, \pref{t2.65} holds. Since
$\beta_i \in \vars$, from the semantics of \topl we get \pref{t2.66}.
\pref{t2.65} and \pref{t2.66} imply \pref{t2.67}.
\begin{gather}
  \label{t2.65}
  v_i = \fdi(g(\beta_i)) \\
  \label{t2.66}
  g(\beta_i) = \denot{M(st),g}{\beta_i}\\
  \label{t2.67}
  v_i = \fdi(\denot{M(st),g}{\beta_i})
\end{gather}
\pref{t2.58} (which holds; see above) and the assumption that $i \in
\{1,2,3,\dots,n\}$ imply that:
\begin{gather}
\beta_i = \tau_{\omega_i} \label{t2.68} \\
\omega_i \in \{1,2,3,\dots,k\} \label{t2.68.2} 
\end{gather}
Using \pref{t2.68}, \pref{t2.67} becomes \pref{t2.69}. \pref{t2.68.2}
and \pref{t2.59} imply \pref{t2.70}. 
\begin{gather}
  \label{t2.69}
  v_i = \fdi(\denot{M(st),g}{\tau_{\omega_i}}) \\
  \label{t2.70}
  v'_{\omega_i} = \fdi(\denot{M(st),g}{\tau_{\omega_i}})
\end{gather}
\pref{t2.69} and \pref{t2.70} imply that $v_i = v'_{\omega_i}$. I have
proven that for every $i \in \{1,2,3,\dots,n\}$, $v_i =
v'_{\omega_i}$. Hence, \pref{t2.57} holds. The proof that the RHS of
\pref{t2.2} is a subset of the LHS has been completed. 


\subsection{$?_{mxl}\beta_1 \; ?\beta_2 \; ?\beta_3 \; \dots \;
?\beta_n \; \phi'$} 

\subsubsection*{Translation rule}

If $\beta_1, \beta_2, \dots, \beta_n \in \vars$, $\phi' \in \ynforms$,
and $\linit$ is as in section \ref{formulation}, then:

$trans(?_{mxl}\beta_1 \; ?\beta_2 \; ?\beta_3 \dots \;
?\beta_n \; \phi', \linit) \defeq$ \\
\sql{(}\select{SELECT DISTINCT SNAPSHOT VALID($\alpha_2$), 
                      $\alpha_2$.2,
                      $\alpha_2$.3, \dots, $\alpha_2$.$n$ \\
               FROM (\select{SELECT DISTINCT 'dummy',
                                    $\alpha_1.\omega_2$,
                                    $\alpha_1.\omega_3$, \dots, 
                                    $\alpha_1.\omega_n$ \\
                             VALID $\alpha_1.\omega_1$ \\
                             FROM $trans(\phi', \linit)
                                  $ AS $\alpha_1$}\\
               \ \ \ \ \ )(NOSUBPERIOD) AS $\alpha_2$)}

Each time the translation rule is used, $\alpha_1$ and $\alpha_2$ are
two different new correlation names, obtained by calling the
correlation names generator. For every $i \in \{1,2,3,\dots,n\}$, 
\[
\omega_i = min(\{j \mid j \in \{1,2,3,\dots,k\} \mbox{ and } 
  \tau_j = \beta_i \})
\]
where $\tup{\tau_1, \tau_2, \tau_3, \dots, \tau_k} =
\corn{\phi'}$. That is, the first position (from left to right) where
$\beta_i$ appears in $\tup{\tau_1, \dots, \tau_k}$ is the $\omega_i$-th one. 

\subsubsection*{Proof that theorem \ref{wh_theorem} holds for $\phi =
?_{mxl}\beta_1 \; ?\beta_2 \; ?\beta_3 \; \dots \; ?\beta_n \; \phi'$}

I assume that $\beta_1, \beta_2, \dots, \beta_n \in \vars$ and $\phi'
\in \ynforms$. By the syntax of \topl, this implies that $?_{mxl}\beta_1 \;
?\beta_2 \; ?\beta_3 \; \dots \; ?\beta_n \; \phi' \in \whforms$. I
also assume that $st \in \pts$, $\linit$ is as in section
\ref{formulation}, $\Sigma = trans(?_{mxl}\beta_1 \; ?\beta_2 \;
?\beta_3 \; \dots \; ?\beta_n \; \phi', \linit)$, and that
$\corn{\phi'} = \tup{\tau_1, \dots, \tau_k}$ (as in the translation
rule). I need to show that:
\begin{enumerate}
\item $\fcn(\Sigma) = \emptyset$

\item $eval(st, \Sigma) \in \srel(n)$

\item $\{\tup{f_D(v_1), \dots, f_D(v_n)} \mid 
         \tup{v_1, \dots, v_n} \in eval(st, \Sigma)\} =$ \\
      $\denot{M(st), st}
        {?_{mxl}\beta_1 \; ?\beta_2 \; ?\beta_3 \; \dots \; ?\beta_n \; \phi'}$
\end{enumerate}

Let $\Sigma'$ be the embedded \sql{SELECT} statement to which
$\alpha_1$ refers, i.e.\ $\Sigma' = trans(\phi', \linit)$. 
Following exactly the same steps as in section \ref{wh1_rule}, we
arrive at the conclusion that:
\begin{enumerate}
\item[$1'.$] $\fcn(\Sigma') = \emptyset$
\item[$2'.$] $eval(st, \Sigma') \in \vrel(k)$
\item[$3'.$] $\tup{v'_1, \dots, v'_k; v'_t} \in 
       eval(st, \Sigma')$ iff for some $g \in G$: \\
   $\denot{M(st), g}{\tau_1} =
   f_D(v'_1)$, \dots, $\denot{M(st), g}{\tau_k} = f_D(v'_k)$, and 
   $\denot{M(st), st, f_D(v'_t), \pts, g}{\phi'} = T$
\end{enumerate}

\subsubsection*{Proof of clause 1}

The \sql{VALID($\alpha_2$)}, $\alpha_2$\sql{.2}, \dots,
$\alpha_2$\sql{.}$n$ in the \sql{SELECT} clause of $\Sigma$ are not
free column references in $\Sigma$, because $\Sigma$ is a binding
context for all of them. The $\alpha_1.\omega_1$, \dots,
$\alpha_1.\omega_n$ in the \sql{VALID} and the \sql{SELECT} clauses of
the embedded \sql{SELECT} statement to which $\alpha_2$ refers are
also not free column references in $\Sigma$, because the \sql{SELECT}
statement to which $\alpha_2$ refers is a binding context for all of
them. $\Sigma$ contains no other column references (and hence no other
free column references), apart from those
that possibly appear within $\Sigma'$ (the $trans(\phi', \linit)$).
By lemma \ref{fcn_lemma}, this implies \pref{whm:3}. Clause
$1'$ and \pref{whm:3} imply clause 1. 
\begin{equation}
\label{whm:3}
\fcn(\Sigma) \subseteq \fcn(\Sigma')
\end{equation}

\subsubsection*{Proof of clause 2}

When computing $eval(st, \Sigma)$, the $\alpha_1$ of $\Sigma$ ranges
over the tuples of $eval(st, \Sigma')$. (I do not include in the
arguments of $eval(st, \Sigma)$ and $eval(st, \Sigma')$ the
assignment to the correlation names, because according to clauses
1 and $1'$, $\fcn(\Sigma) = \fcn(\Sigma') = \emptyset$.) 
By clause $2'$, $eval(st,\Sigma') \in \vrel(k)$. Hence, $\alpha_1$
ranges over tuples of the form $\tup{v'_1, \dots, v'_k;
v'_t} \in eval(st, \Sigma')$. 

The syntax of \topl (section \ref{top_syntax}) and the fact that
$?_{mxl}\beta_1 \; \dots \; ?\beta_n \; \phi' \in \whforms$ imply that
for every $i \in \{1,2,3,\dots,n\}$, $\beta_i$ occurs at least once
within $\phi'$. This and the definition of $\corn{\dots}$ imply that
$\beta_i$ occurs at least once within $\corn{\phi'}$, which according
to the hypothesis is $\tup{\tau_1,\dots,\tau_k}$. Hence, for every $i
\in \{1,2,3,\dots,n\}$, the set $\{j \mid j \in \{1,2,3,\dots,k\}
\mbox{ and } \tau_j = \beta_i \}$ in the definition of $\omega_i$ (see
the translation rule) is not empty, and $\omega_i \in
\{1,2,3,\dots,k\}$. Assuming that $\alpha_1$ refers to a tuple
$\tup{v'_1, \dots, v'_k; v'_t} \in eval(st, \Sigma')$, for every $i
\in \{1,2,3,\dots,n\}$, the $\alpha_1.\omega_i$ of $\Sigma$ (see the
translation rule) refers to $v'_{\omega_i}$, i.e.\ the $\omega_i$-th
(from left to right) among $v'_1,v'_2, \dots, v'_k$.

The reader should now be able to see that the embedded \sql{SELECT}
statement of the translation rule returns the following relation,
where $\omega_1, \dots, \omega_n$ are as in the translation rule.  (I
assume that the \tsql string \sql{'dummy'} evaluates to the element of
$D$ $dummy$.)\footnote{The reader may wonder if $v'_{\omega_1}$ is
  always a valid time-stamp, i.e.\ if there is any guarantee that
  $v'_{\omega_1} \in D_T$. Intuitively, this is indeed the case,
  because $v'_{\omega_1}$ corresponds to $\beta_1$, and the syntax of
  \topl requires $\beta_1$ to occur at least once within $\phi'$ as
  the first argument of a \past, \perf, \at, \before, \after, or
  \ntense operator. The semantics of \topl requires variables that
  occur at these positions to denote periods, and hence
  $v'_{\omega_1}$ will also denote a period, i.e.\ $v'_{\omega_1} \in
  D_P \subseteq D_T$. This, however, will not be proven formally
  here.}
\begin{eqnarray}
&& eval(st, \text{\select{SELECT DISTINCT 'dummy',
                                    $\alpha_1.\omega_2$,
                                    $\alpha_1.\omega_3$, \dots, 
                                    $\alpha_1.\omega_n$ \\
                             VALID $\alpha_1.\omega_1$ \\
                             FROM $trans(\phi', \linit)
                                  $ AS $\alpha_1)$}} 
\label{whm:500} \\
&&= 
  \{\tup{dummy, v'_{\omega_2}, v'_{\omega_3},
    \dots, v'_{\omega_n}; v'_{\omega_1}} \mid 
  \tup{v'_1,v'_2,v'_3,\dots,v'_k;v'_t} \in eval(st,\Sigma')\}
\nonumber
\end{eqnarray}

Let us use $r(st)$ to refer to the relation of \pref{whm:500}. 
It should be easy to see from the translation rule and the semantics of
\sql{(NOSUBPERIOD)} (section \ref{new_pus}) that:
\begin{equation}
\label{whm:501} 
\begin{aligned}[t]
eval(st, \Sigma) = \{&\tup{v_t, v_2, v_3, \dots, v_n} \mid \\
                     &\tup{v_1, v_2, v_3, \dots, v_n; v_t} 
                     \in nosubperiod(r(st))\}
\end{aligned}
\end{equation}
Using the definition of $nosubperiod$ (section \ref{new_pus}),
\pref{whm:501} becomes \pref{whm:502}.
\begin{eqnarray}
\label{whm:502} 
eval(st, \Sigma) & = & \{\tup{v_t, v_2, v_3, \dots, v_n} \mid 
\tup{v_1, v_2, v_3, \dots, v_n; v_t} \in r(st), \\
&& \; \text{ and if } \tup{v_1, v_2, v_3, \dots, v_n; u_t} \in r(st), 
\nonumber \\
&& \;\;\;\;\; \text{then } f_D(v_t) \not\propsubper f_D(u_t) \}
\nonumber 
\end{eqnarray}
Replacing $r(st)$ by the relation of \pref{whm:500}, \pref{whm:502}
becomes \pref{whm:503}. \pref{whm:503} is equivalent to \pref{whm:504}.
\begin{eqnarray}
\label{whm:503} 
&& eval(st, \Sigma) =  \{\tup{v_t, v_2, v_3, \dots, v_n} \mid  \\
&& v_1 = dummy, \; v_2 = v'_{\omega_2}, \; v_3 = v'_{\omega_3}, \;
   \dots, \; v_n = v'_{\omega_n}, \; v_t = v'_{\omega_1}, 
\nonumber \\
&& \tup{v'_1, v'_2, v'_3, \dots, v'_k; v'_t} \in eval(st, \Sigma'),
\nonumber \\
&& \text{and if } 
v_1 = dummy, \; v_2 = v''_{\omega_2}, \; v_3 = v''_{\omega_3}, \;
   \dots, \; v_n = v''_{\omega_n}, \; u_t = v''_{\omega_1}, \nonumber \\
&& 
\;\;\; \text{and } \tup{v''_1, v''_2, v''_3, \dots, v''_k; v''_t} \in
eval(st, \Sigma'),  
\text{ then } f_D(v_t) \not\propsubper f_D(u_t) \}
\nonumber \\
\label{whm:504} \label{wht2.10}
&& eval(st, \Sigma) =  
\{\tup{v'_{\omega_1}, v'_{\omega_2}, v'_{\omega_3}, \dots, v'_{\omega_n}}
\mid  \\
&& \tup{v'_1, v'_2, v'_3, \dots, v'_k; v'_t} \in eval(st, \Sigma'),
\nonumber \\
&& \text{and if } 
\tup{v''_1, v''_2, v''_3, \dots, v''_k; v''_t} \in eval(st, \Sigma') 
\text{ and } \nonumber \\
&& 
\;\;\; v'_{\omega_2} = v''_{\omega_2}, \; v'_{\omega_3} = v''_{\omega_3}, \;
   \dots, \; v'_{\omega_n} = v''_{\omega_n}, \text{ then }
f_D(v'_{\omega_1}) \not\propsubper f_D(v''_{\omega_1}) \}
\nonumber
\end{eqnarray}
\pref{whm:504} implies that $eval(st, \Sigma)$ is a snapshot relation
of $n$ attributes. Therefore, clause 2 holds.

\subsubsection*{Proof of clause 3} 

By the definitions of $\denot{M(st), st}{\phi}$ and
$\denot{M(st),st,et,lt,g}{\phi}$ for $\phi \in \whforms_2$ (section
\ref{denotation}), clause 3 becomes:
\begin{eqnarray}
\label{wht2.1}
&&
\{\tup{f_D(v_1), f_D(v_2), f_D(v_3), \dots, f_D(v_n)} \mid 
     \tup{v_1,v_2,v_3,\dots,v_n} \in eval(st,\Sigma) \} \\
&&
\begin{aligned}[t]
 = \bigcup_{g \in G, et \in \periods} 
   \{&\tup{g(\beta_1),g(\beta_2), \dots, g(\beta_n)}
      \mid \denot{M(st),st,et,\pts,g}{\phi'} = T, \\
     &\text{and for no } et' \in \periods \text{ and } g' \in G 
      \text{ is it true that} \\
     &\denot{M(st),st,et',\pts,g'}{\phi'} = T, 
      \; g(\beta_1) \propsubper g'(\beta_1), \text{ and } \\
     &g(\beta_2) = g'(\beta_2), \;
      \dots, \; g(\beta_n) = g'(\beta_n) \}
\end{aligned}
\nonumber
\end{eqnarray}
I prove \pref{wht2.1} by proving that its left-hand side (LHS) is a subset
of its right-hand side (RHS), and that its RHS is a subset of its
LHS. I first prove that the LHS is a subset of the 
RHS. I assume that for some $v_1, \dots, v_n \in D$, 
$\tup{f_D(v_1), \dots, f_D(v_n)}$ is an element of the LHS of
\pref{wht2.1}, i.e.\ that \pref{wht2.7} holds.
\begin{equation}
  \label{wht2.7}
  \tup{v_1,\dots,v_n} \in eval(st,\Sigma)
\end{equation}

I need to prove that $\tup{f_D(v_1),\dots,f_D(v_n)}$ is also an
element of the RHS of \pref{wht2.1}, i.e.\ that for some $g$ and $et$:
\begin{eqnarray}
&& g \in G \label{wht2.19} \\
&& et \in \periods \label{wht2.20} \\
&& f_D(v_1) = g(\beta_1), \dots, f_D(v_n) =
   g(\beta_n) \label{wht2.21} \\
&& \denot{M(st),st,et,\pts,g}{\phi'} = T \label{wht2.22} \\
&& \text{for no } et' \in \periods \text{ and } g' \in G \text{ is it
true that} \label{whm:506} \\
&&\;\;\; \denot{M(st),st,et',\pts,g'}{\phi'} = T, \; 
g(\beta_1) \propsubper g'(\beta_1), \text{ and } 
\nonumber \\
&&\;\;\;  g(\beta_2) = g'(\beta_2), \; \dots, \; g(\beta_n) = g'(\beta_n)
\nonumber 
\end{eqnarray}

\pref{wht2.7} and \pref{wht2.10} imply that for some $v'_1$,
\dots, $v'_k$, $v'_t$:
\begin{eqnarray}
&& \tup{v'_1,\dots,v'_k;v'_t} \in eval(st,\Sigma') \label{wht2.11} \\
&& v_1 = v'_{\omega_1}, \; v_2 = v'_{\omega_2}, \; \dots, \; v_n =
   v'_{\omega_n} \label{wht2.12} \\
&& \text{if } 
\tup{v''_1, v''_2, v''_3, \dots, v''_k; v''_t} \in eval(st, \Sigma') 
\text{ and } \label{whm:507} \\
&& 
\;\;\; v'_{\omega_2} = v''_{\omega_2}, \; v'_{\omega_3} = v''_{\omega_3}, \;
   \dots, \; v'_{\omega_n} = v''_{\omega_n}, \text{ then }
f_D(v'_{\omega_1}) \not\propsubper f_D(v''_{\omega_1}) 
\nonumber 
\end{eqnarray}
where $\omega_1, \dots, \omega_n$ are as in the translation rule. 

\pref{wht2.11} and clause $3'$ imply that for some $g$:
\begin{eqnarray}
&& g \in G \label{wht2.14} \\
&& \denot{M(st),g}{\tau_1} = f_D(v'_1), \dots, 
   \denot{M(st),g}{\tau_k} = f_D(v'_k) \label{wht2.16} \\
&& \denot{M(st),st,f_D(v_t'),\pts,g}{\phi'} = T \label{wht2.18} 
\end{eqnarray}
\pref{wht2.19} is the same as \pref{wht2.14}, which is known to be
true. I set $et$ as in \pref{whm:11}. Then, \pref{wht2.22} follows from
\pref{wht2.18}. 
\begin{equation}
\label{whm:11}
et = f_D(v_t')
\end{equation}
\pref{wht2.11} and clause $2'$ imply that $v_t' \in D_P$, which in
turn implies \pref{whm:10}. \pref{wht2.20} follows from \pref{whm:11}
and \pref{whm:10}.
\begin{equation}
\label{whm:10}
f_D(v_t') \in \periods
\end{equation}
I now prove \pref{wht2.21}. For every $i \in \{1,2,3,\dots,n\}$,
since $\beta_i \in \vars$, according to the semantics of \topl:
\begin{equation}
  \label{wht2.23}
  g(\beta_i) = \denot{M(st),g}{\beta_i}
\end{equation}
The definition of $\omega_1, \dots, \omega_n$ in the translation rule
implies that :
\begin{eqnarray}
&& \tau_{\omega_i} = \beta_i \label{wht2.23.1} \\
&& \omega_i \in \{1,2,3,\dots,k\} \label{wht2.23.2} 
\end{eqnarray}
\pref{wht2.23.1} and \pref{wht2.23} imply \pref{wht2.23.3}.
\pref{wht2.23.2} and \pref{wht2.16} imply \pref{wht2.23.4}.
\pref{wht2.23.3} and \pref{wht2.23.4} imply \pref{wht2.30}.
\begin{gather}
  \label{wht2.23.3}
  g(\beta_i) = \denot{M(st),g}{\tau_{\omega_i}} \\
  \label{wht2.23.4}
  \denot{M(st),g}{\tau_{\omega_i}} = f_D(v'_{\omega_i}) \\
  \label{wht2.30}
  g(\beta_i) = f_D(v'_{\omega_i})
\end{gather}
\pref{wht2.12} and the assumption that $i \in \{1,2,3,\dots,n\}$,
imply \pref{wht2.30.1}. \pref{wht2.30.1} and \pref{wht2.30} imply
\pref{wht2.30.1.last}.
\begin{gather}
  \label{wht2.30.1}
  v'_{\omega_i} = v_i \\
  \label{wht2.30.1.last}
  g(\beta_i) = f_D(v_i)
\end{gather}
I have proven that for every $i \in \{1,2,3,\dots,n\}$, $f_D(v_i) =
g(\beta_i)$. Therefore, \pref{wht2.21} holds.

It remains to prove \pref{whm:506}. Let us assume that for some $et'
\in \periods$ and $g' \in G$, \pref{whm:510} and \pref{whm:511} hold. 
\begin{gather}
\label{whm:510}
\denot{M(st),st,et',\pts,g'}{\phi'} = T \\
\label{whm:511}
g(\beta_1) \propsubper g'(\beta_1), \; g(\beta_2) = g'(\beta_2), \;
\dots, \; g(\beta_n) = g'(\beta_n)
\end{gather}
For $j \in \{1,2,3,\dots,k\}$, I set $v''_j =
\fdi(\denot{M(st),g'}{\tau_j})$, which implies \pref{whm:512}. I also
set $v''_t = \fdi(et')$, which implies \pref{whm:513}.
\begin{gather}
\label{whm:512}
\denot{M(st),g'}{\tau_1} = f_D(v''_1), \; \dots, \; 
\denot{M(st),g'}{\tau_k} = f_D(v''_k) \\
\label{whm:513}
et' = f_D(v_t'')
\end{gather}
\pref{whm:510} and \pref{whm:513} imply \pref{whm:528}. Clause $3'$,
the assumption that $g' \in G$, \pref{whm:512}, and \pref{whm:528}
imply \pref{whm:537}.
\begin{gather}
\label{whm:528}
\denot{M(st),st,f_D(v''_t),\pts,g'}{\phi'} = T \\
\label{whm:537}
\tup{v''_1, v''_2, v''_3, \dots, v''_k; v''_t} \in eval(st, \Sigma') 
\end{gather}
The definition of $\omega_1, \dots, \omega_n$ in the translation rules
implies that $\omega_1 \in \{1,2,3,\dots,k\}$ and $\tau_{\omega_1} =
\beta_1$. Then, from \pref{whm:512} we get \pref{whm:516}.
\pref{whm:516} and the fact that $\beta_1 \in \vars$ imply
\pref{whm:517}.
\begin{gather}
\label{whm:516}
f_D(v''_{\omega_1}) = \denot{M(st),g'}{\tau_{\omega_1}} =
\denot{M(st),g'}{\beta_1} \\
\label{whm:517}
f_D(v''_{\omega_1}) = g'(\beta_1)
\end{gather}
\pref{wht2.21} (proven above) and \pref{wht2.12} imply \pref{whm:520}.
\pref{whm:517}, \pref{whm:520}, and \pref{whm:511} imply
\pref{whm:521}.
\begin{gather}
\label{whm:520}
g(\beta_1) = f_D(v'_{\omega_1}) \\
\label{whm:521}
f_D(v'_{\omega_1}) \propsubper f_D(v''_{\omega_1})
\end{gather}
For $l \in \{2,3,\dots,n\}$, the definition of $\omega_1, \dots,
\omega_n$ in the translation rule implies that:
\begin{gather}
\label{whm:523}
\omega_l \in \{1,2,3,\dots,k\} \\
\label{whm:524}
\tau_{\omega_l} = \beta_l
\end{gather}
\pref{whm:512} and \pref{whm:523} imply \pref{whmx:1}. \pref{whmx:1},
\pref{whm:524}, and the fact that $\beta_l \in \vars$ imply
\pref{whmx:2}, which in turn implies \pref{whmx:3}.
\begin{gather}
\label{whmx:1}
\denot{M(st),g'}{\tau_{\omega_l}} = f_D(v''_{\omega_l}) \\
\label{whmx:2}
\denot{M(st),g'}{\beta_l} = g'(\beta_l) = f_D(v''_{\omega_l}) \\
\label{whmx:3}
v''_{\omega_l} = \fdi(g'(\beta_l))
\end{gather}
Since $l \in \{2,3,\dots,n\}$, \pref{whm:511} implies \pref{whmx:4}. \pref{whmx:3} and \pref{whmx:4} imply \pref{whmx:5}. 
\begin{gather}
\label{whmx:4}
g'(\beta_l) = g(\beta_l) \\
\label{whmx:5}
v''_{\omega_l} = \fdi(g(\beta_l))
\end{gather}
\pref{wht2.21} (proven above) and the fact that $l \in
\{2,3,\dots,n\}$ imply that $g(\beta_l) = f_D(v_l)$, which in turn
implies \pref{whmx:6}. \pref{whmx:5} and \pref{whmx:6} imply
\pref{whmx:7}. \pref{whmx:7}, \pref{wht2.12}, and the fact that $l \in
\{2,3,\dots,n\}$ imply \pref{whmx:8}. 
\begin{gather}
\label{whmx:6}
\fdi(g(\beta_l)) = v_l \\
\label{whmx:7}
v''_{\omega_l} = v_l \\
\label{whmx:8}
v_l = v'_{\omega_l}
\end{gather}
\pref{whmx:7} and \pref{whmx:8} imply that $v''_{\omega_l} = v'_{\omega_l}$.
I have proven that for every $l \in \{2,3,4,\dots,n\}$, $v''_{\omega_l} =
v'_{\omega_l}$. Hence, \pref{whm:526} holds. 
\begin{equation}
\label{whm:526}
v''_{\omega_2} = v'_{\omega_2}, \; v''_{\omega_3} = v'_{\omega_3}, 
\; \dots, \; v''_{\omega_n} = v'_{\omega_n}
\end{equation}
\pref{whm:537}, \pref{whm:526}, and \pref{whm:521} are against
\pref{whm:507}. Therefore, the hypothesis that there is an $et'
\in \periods$ and a $g' \in G$, such that \pref{whm:510} and
\pref{whm:511} are satisfied cannot hold. \pref{whm:506} has been proven.  
The proof that the LHS of \pref{wht2.1} is a subset of the RHS has
been completed. 

\bigskip

I now prove that the RHS of \pref{wht2.1} is a subset of the LHS.  I
assume that for some $g \in G$,
$\tup{g(\beta_1),g(\beta_2),\dots,g(\beta_n)}$ is an element of the
RHS of \pref{wht2.1}, i.e.\ that for some $g$ and $et$, \pref{wht2.51}
-- \pref{whm:600} hold. 
\begin{eqnarray}
&& g \in G \label{wht2.51} \\
&& et \in \periods \label{wht2.52} \\
&& \denot{M(st),st,et,\pts,g}{\phi'} = T \label{wht2.53} \\
&& \text{for no } et' \in \periods \text{ and } g' \in G 
\text{ is it true that} \label{whm:600} \\
&&\;\;\; \denot{M(st),st,et',\pts,g'}{\phi'} = T, 
\; g(\beta_1) \propsubper g'(\beta_1), \text{ and } \nonumber \\
&&\;\;\; g(\beta_2) = g'(\beta_2), \;
\dots, \; g(\beta_n) = g'(\beta_n) \nonumber
\end{eqnarray}
I need to prove that $\tup{g(\beta_1),g(\beta_2),\dots,g(\beta_n)}$ is
also an element of the LHS of \pref{wht2.1}, i.e.\ that for some
$v_1,\dots,v_n \in D$:
\begin{eqnarray}
&& g(\beta_1) = f_D(v_1), \; \dots, \; g(\beta_n) =
  f_D(v_n) \label{wht2.54} \\
&& \tup{v_1,\dots,v_n} \in eval(st,\Sigma)  \label{wht2.55}
\end{eqnarray}
I set $v_1, \dots, v_n$ as in \pref{wht2.64}, which implies
\pref{wht2.54}. 
\begin{equation}
  \label{wht2.64}
  v_1 = \fdi(g(\beta_1)), \; \dots, \; 
  v_n = \fdi(g(\beta_n))
\end{equation}
It remains to prove \pref{wht2.55}.  By \pref{wht2.10}, it is enough
to prove that for some $v'_1$, $v'_2$, \dots, $v'_k$, $v'_t$:
\begin{eqnarray}
&& \tup{v'_1,\dots,v'_k;v'_t} \in eval(st,\Sigma') \label{wht2.56} \\
&& v_1 = v'_{\omega_1}, \; \dots, \; v_n = v'_{\omega_n} \label{wht2.57} \\
&& \text{if } 
\tup{v''_1, v''_2, v''_3, \dots, v''_k; v''_t} \in eval(st, \Sigma') 
\text{ and } \label{whm:601}  \\
&& 
\;\;\; v'_{\omega_2} = v''_{\omega_2}, \; v'_{\omega_3} = v''_{\omega_3}, \;
   \dots, \; v'_{\omega_n} = v''_{\omega_n}, \text{ then }
f_D(v'_{\omega_1}) \not\propsubper f_D(v''_{\omega_1}) \nonumber 
\end{eqnarray}
where $\omega_1, \dots, \omega_n$ are as in the translation rule. 

I set $v'_1, \dots, v'_k$ as in \pref{wht2.59}, which implies
\pref{wht2.60}. I also set $v'_t = \fdi(et)$, which implies
\pref{wht2.61}.  Clause $3'$, \pref{wht2.51}, \pref{wht2.60},
\pref{wht2.61}, and \pref{wht2.53} imply \pref{wht2.56}.
\begin{gather}
  \label{wht2.59}
  v'_1 = \fdi(\denot{M(st),g}{\tau_1}), \; \dots, \; 
  v'_k = \fdi(\denot{M(st),g}{\tau_k}) \\
  \label{wht2.60}
  \denot{M(st),g}{\tau_1} = f_D(v'_1), \; \dots, \;
  \denot{M(st),g}{\tau_k} = f_D(v'_k) \\
  \label{wht2.61}
  et = f_D(v'_t)
\end{gather}

I now prove \pref{wht2.57}.  \pref{wht2.64} implies that
\pref{wht2.65} holds for every $i \in \{1,2,3,\dots,n\}$. Since
$\beta_i \in \vars$, from the semantics of \topl we also get
\pref{wht2.66}. \pref{wht2.65} and \pref{wht2.66} imply \pref{wht2.67}. 
\begin{gather}
  \label{wht2.65}
  v_i = \fdi(g(\beta_i)) \\
  \label{wht2.66}
  g(\beta_i) = \denot{M(st),g}{\beta_i} \\
  \label{wht2.67}
  v_i = \fdi(\denot{M(st),g}{\beta_i})
\end{gather}
The definition of $\omega_1, \dots, \omega_n$ in the translation rule
implies that:
\begin{gather}
\beta_i = \tau_{\omega_i} \label{wht2.68} \\
\omega_i \in \{1,2,3,\dots,k\} \label{wht2.68.2} 
\end{gather}
Using \pref{wht2.68}, \pref{wht2.67} becomes \pref{wht2.69}. \pref{wht2.68.2} and \pref{wht2.59} imply \pref{wht2.70}. 
\begin{gather}
  \label{wht2.69}
  v_i = \fdi(\denot{M(st),g}{\tau_{\omega_i}}) \\
  \label{wht2.70}
  v'_{\omega_i} = \fdi(\denot{M(st),g}{\tau_{\omega_i}})
\end{gather}
\pref{wht2.69} and \pref{wht2.70} imply that
$v_i = v'_{\omega_i}$. I have proven that for every $i \in \{1,2,3,\dots,n\}$, 
$v_i = v'_{\omega_i}$. Hence, \pref{wht2.57} holds. 

It remains to prove \pref{whm:601}. Let us assume that:
\begin{gather}
\label{whmxx:1}
\tup{v''_1, v''_2, v''_3, \dots, v''_k; v''_t} \in eval(st, \Sigma') \\
\label{whmxx:7}
v'_{\omega_2} = v''_{\omega_2}, \; v'_{\omega_3} = v''_{\omega_3}, \; \dots, 
\; v'_{\omega_n} = v''_{\omega_n}
\end{gather}
We need to prove that $f_D(v'_{\omega_1}) \not\propsubper
f_D(v''_{\omega_1})$.  Let us assume that this is not true, i.e.\ that
\pref{whmxx:4} holds.
\begin{equation}
\label{whmxx:4}
f_D(v'_{\omega_1}) \propsubper f_D(v''_{\omega_1})
\end{equation}
Clause $3'$ and \pref{whmxx:1} imply that for some $g'$:
\begin{gather}
\label{whmxx:120}
g' \in G \\
\label{whmxx:6}
\denot{M(st),g'}{\tau_1} = f_D(v''_1), \; \dots, \; 
\denot{M(st),g'}{\tau_k} = f_D(v''_k) \\
\label{whmxx:102}
\denot{M(st),st,f_D(v''_t),\pts,g'}{\phi'} = T
\end{gather}
I set $et'$ as in \pref{whmxx:2}. Then \pref{whmxx:102} implies
\pref{whmxx:103}.
\begin{gather}
\label{whmxx:2}
et' = f_D(v''_t) \\
\label{whmxx:103}
\denot{M(st),st,et',\pts,g'}{\phi'} = T
\end{gather}
\pref{whmxx:1} and clause $2'$ imply that $v_t'' \in D_P$, which in
turn implies that $f_D(v''_t) \in \periods$. Then, \pref{whmxx:2}
implies that:
\begin{equation}
\label{whmxx:119}
et' \in \periods
\end{equation}
For $i \in \{1,2,3,\dots,n\}$, \pref{whmxx:6} and \pref{wht2.68.2}
imply \pref{whmxx:106}. \pref{whmxx:106}, \pref{wht2.68}, and the fact
that $\beta_i \in \vars$ imply \pref{whmxx:107}.
\begin{gather}
\label{whmxx:106}
\denot{M(st),g'}{\tau_{\omega_i}} = f_D(v''_{\omega_i}) \\
\label{whmxx:107}
\denot{M(st),g'}{\beta_i} = g'(\beta_i) = f_D(v''_{\omega_i})
\end{gather}
\pref{wht2.54} (which holds because of \pref{wht2.64}) and the fact
that $i \in \{1,2,3,\dots,n\}$ imply \pref{whmxx:113}. \pref{wht2.57}
(proven above) and the fact that $i \in \{1,2,3,\dots,n\}$ imply
\pref{whmxx:114}. \pref{whmxx:113} and \pref{whmxx:114} imply
\pref{whmxx:115}.
\begin{gather}
\label{whmxx:113}
g(\beta_i) = f_D(v_i) \\
\label{whmxx:114}
v_i = v'_{\omega_i} \\
\label{whmxx:115}
g(\beta_i) = f_D(v'_{\omega_i})
\end{gather}
\pref{whmxx:115} and \pref{whmxx:107} were proven for $i \in
\{1,2,3,\dots,n\}$. For $i = 1$, from \pref{whmxx:115} we get
\pref{whmxx:5}, and from \pref{whmxx:107} we get \pref{whmxx:107b}.
\pref{whmxx:4}, \pref{whmxx:5}, and \pref{whmxx:107b} imply
\pref{whmxx:108}.
\begin{gather}
\label{whmxx:5}
g(\beta_1) = f_D(v'_{\omega_1}) \\
\label{whmxx:107b}
g'(\beta_1) = f_D(v''_{\omega_1}) \\
\label{whmxx:108}
g(\beta_1) \propsubper g'(\beta_1) 
\end{gather}
For $l \in \{2,3,\dots,n\}$, from \pref{whmxx:115} we get
\pref{whmxx:115b}, and from \pref{whmxx:107} we get \pref{whmxx:107c}.
\pref{whmxx:7} and the fact that $l \in \{2,3,\dots,n\}$ imply
\pref{whmxx:116}. \pref{whmxx:115b} and \pref{whmxx:116} imply
\pref{whmxx:117}.  
\begin{gather}
\label{whmxx:115b}
g(\beta_l) = f_D(v'_{\omega_l}) \\
\label{whmxx:107c}
g'(\beta_l) = f_D(v''_{\omega_l})
\label{whmxx:116}
v'_{\omega_l} = v''_{\omega_l} \\
\label{whmxx:117}
g(\beta_l) = f_D(v''_{\omega_l})
\end{gather}
\pref{whmxx:117} and \pref{whmxx:107c} imply that $g(\beta_l) =
g'(\beta_l)$. I have proven that for every $l \in \{2,3,\dots,n\}$,
$g(\beta_l) = g'(\beta_l)$. Therefore, \pref{whmxx:118} holds.
\begin{equation}
\label{whmxx:118}
g(\beta_2) = g'(\beta_2), \; g(\beta_3) = g'(\beta_3), \; \dots, \; 
g(\beta_n) = g'(\beta_n)
\end{equation}
\pref{whmxx:119}, \pref{whmxx:120}, \pref{whmxx:103},
\pref{whmxx:108}, and \pref{whmxx:118} are against
\pref{whm:600}. Therefore, the hypothesis that \pref{whmxx:4} is true
cannot hold, i.e.\ $f_D(v'_{\omega_1}) \not\propsubper f_D(v''_{\omega_1})$.
\pref{whm:601} has been proven. The proof that the RHS of
\pref{wht2.1} is a subset of the LHS has been completed. 



\end{document}